# Mining Insights from Weakly-Structured Event Data

# Mining Insights from Weakly-Structured Event Data

PROEFSCHRIFT

ter verkrijging van de graad van doctor aan de Technische Universiteit Eindhoven, op gezag van de rector magnificus prof.dr.ir. F.P.T. Baaijens, voor een commissie aangewezen door het College voor Promoties, in het openbaar te verdedigen op woensdag 19 juni 2019 om 11:00 uur

door

Niek Tax

geboren te Deventer

Dit proefschrift is goedgekeurd door de promotoren en de samenstelling van de promotiecommissie is als volgt:

| | |
|---|---|
| voorzitter: | prof.dr.ir. J.J. Lukkien |
| 1e promotor: | prof.dr.ir. Wil M.P. van der Aalst |
| 1e copromotor: | dr. Natalia Sidorova |
| 2e copromotor: | dr. Reinder Haakma |
| | |
| leden: | prof.dr. Milan Petkovic<br>(Eindhoven University of Technology & Philips Research) |
| | prof.dr. Barbara Weber<br>(Technical University of Denmark) |
| | prof.dr. Gianluigi Greco<br>(University of Calabria) |

*Het onderzoek of ontwerp dat in dit proefschrift wordt beschreven is uitgevoerd in overeenstemming met de TU/e Gedragscode Wetenschapsbeoefening.*

Dedicated to my parents.

# Abstract

Process mining techniques aim to extract actionable insights from event data. Traditionally applications of process mining have focused mainly on the business process management (BPM) domain, where the event data that is generated is often structured as an effect of normative process specifications that are in place during execution of a business processes. This thesis focuses on process mining on event data where such a normative specification is absent and, as a result, the event data is less structured. The thesis puts special emphasis on one application domain that fits this description: the analysis of smart home data where sequences of daily activities are recorded.

In this thesis we propose a set of techniques to analyze such data, which can be grouped into two categories of techniques. The first category of methods focuses on *preprocessing* event logs in order to enable process discovery techniques to extract insights into unstructured event data. In this category we have developed the following techniques:

- An unsupervised approach to refine event labels based on the time at which the event took place, allowing for example to distinguish recorded *eating* events into *breakfast*, *lunch*, and *dinner*.
- An approach to detect and filter from event logs so-called *chaotic activities*, which are activities that cause process discovery methods to overgeneralize.
- A supervised approach to abstract low-level events into more high-level events, where we show that there exist situations where process discovery approaches overgeneralize on the low-level event data but are able to find precise models on the high-level event data.

The second category focuses on mining *local process models*, i.e., collections of process model patterns that each describe some frequent pattern, in contrast to the single global process model that is obtained with existing process discovery techniques. When no precise process model exists that explains all behavior in the event data there can still exist several frequent patterns that each capture a local fragment of behavior. Several techniques are introduced in the area of local process model mining, including a basic method, fast but approximate heuristic methods, and constraint-based techniques.

All techniques have been implemented and evaluated on real-life data.

# Contents

# List of Figures

# List of Tables

# List of Algorithms

# 1 Introduction

A large share of the world's data naturally occurs in sequences. Examples of such sequence data includes texts in natural language (i.e., sequences of letters or of words), DNA sequences, web browsing behavior, and execution traces of business processes or of software systems. Several different research fields have independently focused on developing algorithms, tools, and techniques to model and to analyze such sequence data. Three research fields that have in particular contributed to the analysis of sequence data are the fields of *sequence mining*, *grammar inference*, and *process mining*, which we will briefly introduce in the upcoming paragraphs. In addition to these three research fields, many application-domain-specific techniques for sequence analysis have been developed in application-focused research fields. This thesis can be positioned in-between the research fields of process mining and sequence mining and additionally has links to the application domain of analysis of human behavior in smart environments.

*Sequence mining* [RS02], also called temporal data mining, is a research field that is concerned with the analysis of sequence data and the extraction of insights from it. Throughout the years, the sequence mining community has formulated many mining tasks that are related to sequence data and has proposed many algorithms that address those mining tasks. Mining tasks for sequence data includes *sequential pattern mining* [AS95; Fou+17], i.e., mining frequently occurring subsequences from a collection of sequences, *temporal rule mining* [Das+98; Fou+12], i.e., mining rules of the nature "if some event(s) occurred, some other event(s) are likely to occur with a given probability". We will soon see that this thesis is closely linked to both of these mining tasks. The sequence mining research field focuses on developing general methodologies that often are agnostic to their application domain. Sequence mining techniques have been successfully applied in a wide range of application domains, including crime analysis [Che+04], analysis of website usage profiles for web personalization [Mob+02; Mob07], analysis of information diffusion through blogs on the web [Gru+04], and analysis of public transport travel trajectories [Che+12b].

Several research fields have developed application-specific techniques for sequence analysis. An example of such a field is the research field of bioinformatics, in which techniques have been developed that are heavily specialized in analyzing genome, RNA, or DNA sequences [Bli+13; Rei+07; WGS09]. Another example of a research field that has developed application-specific techniques for sequence analysis are *ambient intelligence* [CAJ09] and the closely-related *ubiquitous computing* [Che+11] fields, which focus on analyzing sequences of on data that originates from human activity. The ambient intelligence community mainly focuses on smart homes or other types of smart environments (e.g. smart office spaces) while the ubiquitous computing community mainly focuses on data from mobile phones or on-body sensors. Sequence mining techniques from these research fields focus on extracting frequent patterns of human behavior (e.g., [Azt10; Eld+18; KHC10; LWV07]).

*Grammar inference* is a research field that makes the assumption that the sequences are generated from some formal grammar and aims to find such a formal grammar given a collection of sequences. Such a formal grammar that is generated by grammar inference techniques takes the form of an automaton (e.g., in [CO94; Dup96] in the case the to-be-discovered language is assumed to be from the class of regular languages, or alternatively, takes the form of a context-free grammar (e.g., in [EHJ07; NW97]) in the EBNF form [Int96] that is commonly used for the specification of programming languages (see for example the Java[1] and Python[2] EBNF specifications).

*Process mining* [Aal16] is a research field that originates from the business process management community and is concerned with extracting knowledge and insights from sequences of business process executions (called an *event log*). The aim of process mining is often to improve the performance of a business process with respect to some key performance indicator, e.g. reducing the cycle time of the process, reducing the workload of resources that operate the process, increasing the financial profit obtained by executing the process, or to improve compliance of the process with rules and regulations. *Process discovery* plays a central role in process mining and it addresses the task of automatic extraction of a process model that accurately describes the business process from an event log. There are several notations to represent a business processes, including BPMN [Obj11], YAWL [AH05], Petri nets [Mur89; Rei12]. Note that process discovery and grammar inference on a high level share the same main goal: extracting a formal language that describes some sequence data. An important difference between grammar inference and process discovery is that the process models that are used to capture the language, in contrast to the formalisms used by grammar inference, treat concurrency as a first-class citizen, i.e., they make explicit which branches of the process are executed concurrently.

While the process mining field originated from the business process management (BPM) research field, the applications of process mining have in recent years broadened to other fields, thereby bringing the process mining field a step closer to the sequence mining field in being application-agnostic. In Section 1.1 we will introduce process mining in a BPM context. In Section 1.2 we will briefly discuss the recent move of process mining to other application domains outside of the BPM field. One application domain that we will assign a special status is the field of ambient intelligence. We will thereby uncover certain data properties that differentiate event data from the BPM field from event data from non-BPM fields such as ambient intelligence. In Section 1.3 we will formulate research goals regarding the development of novel process mining techniques that are useful for analyzing event data from non-BPM application domains. These research goals follow from the differences in data properties between event data from BPM and non-BPM contexts that we identified in Section 1.2. Additionally, Section 1.3 lists the main contributions of this thesis and provides a high-level overview of the thesis structure. Section 1.4 provides a detailed overview of the thesis contents and structure.

[1]https://docs.oracle.com/javase/specs/jls/se7/html/jls-18.html

[2]https://docs.python.org/3/reference/grammar.html

## 1.1 Process Mining in Business Process Management

*Business processes* are vital for organizations to produce the services and products that they offer to their customers. A precise definition of a business process is given by Weske:

> "A business process consists of a set of activities that are performed in coordination in an organization and technical environment. These activities jointly realize a business goal. Each business process is enacted by a single organization, but it may interact with business processes performed by other organizations." [Wes12].

An important aspect of this definition is the *technical environment* in which the business activities are performed. The enactment of business processes in modern organizations is often supported by IT systems. Examples of such systems include Customer Relationship Management (CRM) systems to manage the interactions with current and potential customers, Enterprise Resource Planning (ERP) to track business resources and the status of business commitments, and Business Process Management (BPM) systems to support the enactment of business workflows. During the execution of the business process, such systems, as a side-effect of supporting process enactment, keep track of what was done, by whom, for whom, where, when, etc.

Each execution of a business activity is logged, resulting in a digital record of the execution that we refer to as an *event*. A sequence of events that belong to the same instance of the business process is called a *trace*, and the collection of all recorded traces we call an *event log*. Table 1.1 shows an example of an event log by providing an excerpt of an event log generated by some IT system.

While event logs were originally the side-product of IT systems that support business process enactment, in recent years they have sparked a new area of research focusing on event-based business process analytics, called *process mining*. Process mining focuses on mining interpretable and actionable insights into the business process with an aim to improve that business process, i.e., to make it more time-efficient, cost-efficient, or otherwise more efficient. Process discovery is an important task within the field of process mining and focuses on extracting a visual model of the business process from the event data.

**Table 1.1:** An excerpt from an event log generated by IT systems that support execution of a business process.

| Case | Activity | Worker | Timestamp | Other data |
|---|---|---|---|---|
| ... | ... | ... | ... | ... |
| 10102 | Register loan application | Paul | 12-03-2018 09:03 | ... |
| 10102 | Check completeness of application | Peter | 12-03-2018 10:15 | ... |
| 10102 | Perform checks | Emily | 15-03-2018 15:55 | ... |
| 10102 | Make decision | Emily | 01-04-2018 16:26 | ... |
| 10102 | Notify rejection | Paul | 02-04-2018 09:00 | ... |
| 10103 | Register loan application | Paul | 12-03-2018 16:45 | ... |
| 10103 | Check completeness of application | Emily | 13-03-2018 10:33 | ... |
| 10103 | Request additional information | Peter | 14-03-2018 12:31 | ... |
| 10103 | Check completeness of application | Peter | 15-03-2018 16:11 | ... |
| 10103 | Perform checks | Emily | 19-03-2018 14:03 | ... |
| 10103 | Make decision | Emily | 12-04-2018 11:10 | ... |
| 10103 | Notify acceptance | Paul | 12-04-2018 12:15 | ... |
| 10103 | Provide loan | Alice | 13-04-2018 15:30 | ... |
| ... | ... | ... | ... | ... |

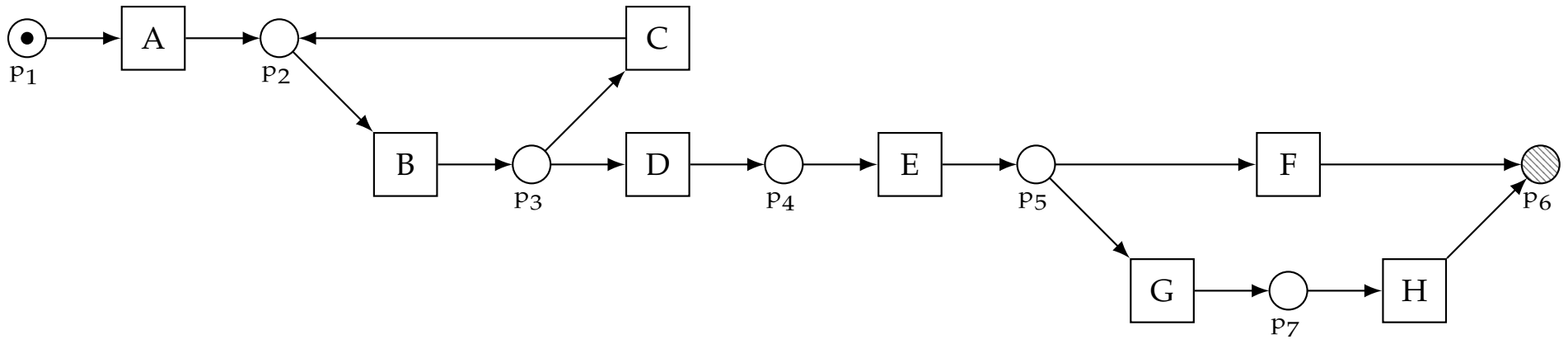

**Figure 1.1:** A process model of the process from which Table 1.1 is generated. A=*Register loan application*, B=*Check completeness of application*, C=*Request additional information*, D=*Perform checks*, E=*Make decision*, F=*Notify rejection*, G=*Notify acceptance*, H=*Provide loan*.

In this thesis, we will mostly represent process models in the form of Petri nets [Mur89; Rei12], since they have both a mathematical definition as well as an intuitive visual representation and their properties have been analyzed extensively over the years. Figure 1.1 shows an example of a Petri net, representing the behavior that was shown in the event log of Table 1.1. We will formally introduce Petri nets in detail in Chapter 2.

Figure 1.2 visualizes the context of process discovery in the context of business process management. Often a model of the business process, or possibly a set of business rules, are created at design time that specifies the behavior that executions of this business process are supposed to adhere to. This is often called a *normative process*, i.e., it specifies the intended process behavior. IT systems such as BPM or ERP systems are typically configured according to this normative process. By doing so, the IT system ends up *allowing* for the behavior of the business process and *disallowing* for execution paths that are in violation of the process model or the business rules. Therefore, instances of the business process that are executed on the IT system are constrained by the configuration of the IT system. As an example of how an IT system constrains the executions of the business process in the context of the loan application process of Figure 1.1, consider that the supporting IT system might disable (i.e., "gray out") the button to make the decision (E) to reject or accept the request until all checks have been performed (D).

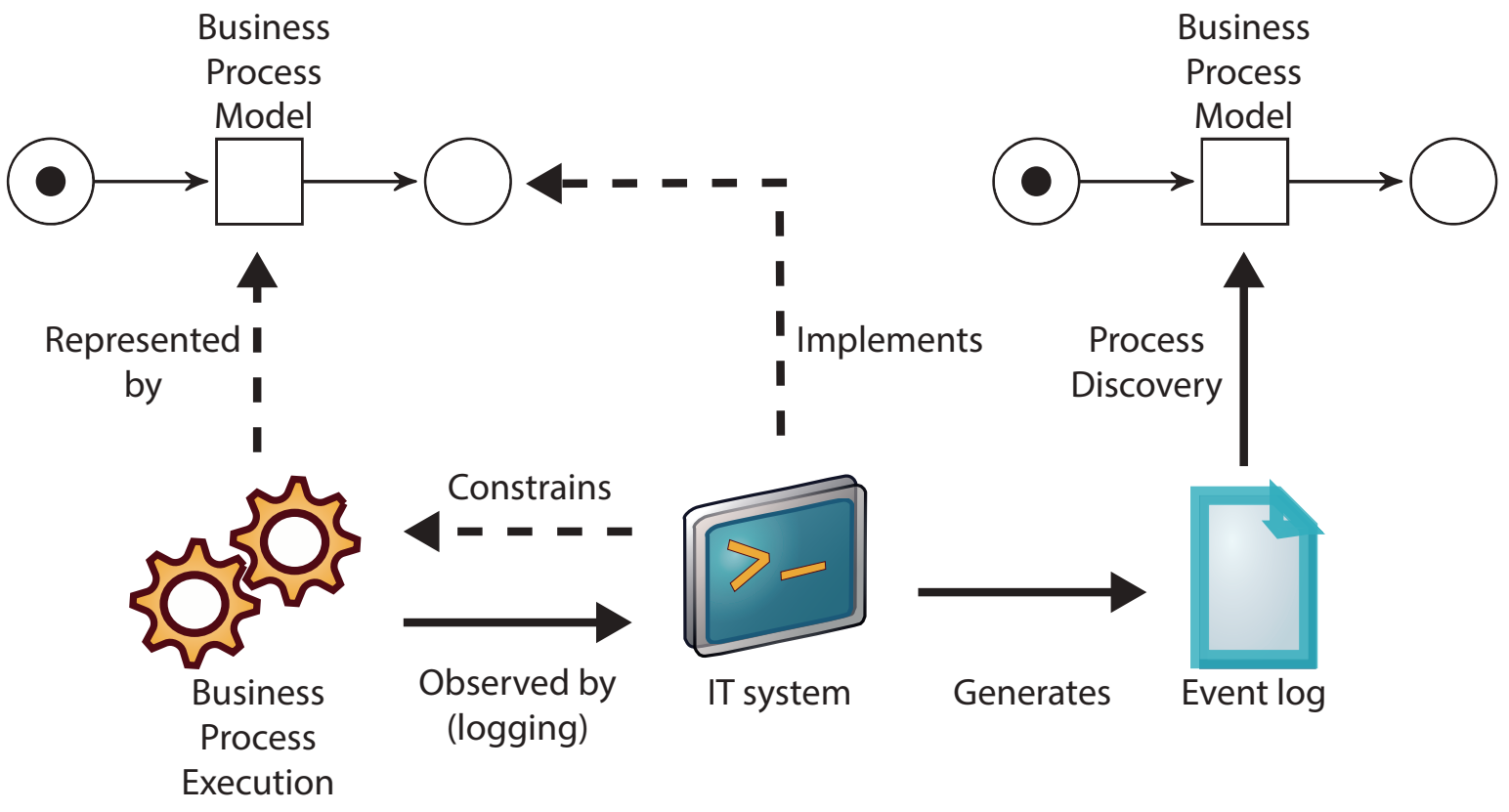

**Figure 1.2:** Process discovery in the context of business process management (BPM).

In addition to constraining the execution of instances of the business process, the IT system also records a trace for each instance of the business process that indicates how that instance has been executed. Event logs are generated by this logging of business process executions. Note however that the constraints that imposed on the business process by the IT system ensure are certain degree of structure in the resulting event logs, e.g., in the context of the loan application process of Figure 1.1, disabling the decision button until checks have been performed ensures that none of the recorded instances contain an E before a D.

Applying process discovery techniques of such event logs can aid organizations to get visual insights into their business processes. Process models that are discovered through process discovery can generate surprising insights into the *real execution behavior* of the business process, which might differ from the execution behavior that is expected by process stakeholders. Process models that are discovered through process discovery techniques can furthermore be used as a starting point for more in-depth types of follow up analysis, such as checking compliance of process executions with rules and regulations [RFA12], identifying bottlenecks in the business process [AAD12], and finding decision points in the process and the identifying the business rules that influence these decision outcomes [Man+16a].

A plethora of process discovery algorithms has been developed throughout the years [Aal+04; Aug+17; Aug+18; BDA12b; Goe+09; LFA13a; LFA13b], producing process models in a range of different process modeling notation. Process discovery techniques aim to find a process model from an event log such that it accurately represents the behavior that was observed in the event log. Whether or not this is the case is generally measured in four quality dimensions:

- The process model should allow for the behavior that was seen in the event log (called *fitness*).
- The process model should not allow for too much behavior that was not seen in the event log (called *precision*).
- The process model should generalize beyond the behavior seen in the event log, by allowing for behavior that is possible in the actual system (called *generalization*).
- The process model should not be unnecessarily complex (called *simplicity*).

## 1.2 Process Mining in non-BPM domains

In recent years, applications of process mining techniques have broadened to a wider range of application domains outside of BPM. Recent application domains in which process discovery techniques have been successfully applied include, but are not limited to, analyzing software based on recorded software executions [LA15; Lee18], analyzing human behavior based on recordings from smart devices, smart home environments, or ambient systems [Dim+16; LMM15; Med+04; Szt+15; Szt+16; Tax+16a; Tax+16c; Tax+18a; Tax+18e; Tax+18f; TSA18], and analyzing the learning behavior of students participating in massive open online courses (MOOCs) [AGG15; Muk+15].

Event logs that originate from the software domain generally contain events that represent certain method-calls in the software execution. In the smart home domain, events represent human activity in the house that is registered by certain sensors. Table 1.2 shows an example of such an event log that originates from a smart home environment. Event logs in an educational setting contain a sequence of events for each student that consists of his interactions with the online study material. The analysis of event logs from these new application domains with process mining techniques creates new links between the process mining field and application-oriented research communities. Examples of such application-oriented research fields are the fields of *Educational Data Mining* (EDM) [BY09] and *Learning*

**Table 1.2:** An excerpt of a log of event generated by a home environment that logs the activity of its resident.

| Sensor | Timestamp | Address | Heart rate | Other data |
|---|---|---|---|---|
| … | … | … | … | … |
| Motion in bedroom | 12-03-2018 06:48 | Surrey Lane 25 | 80 | … |
| Motion in kitchen | 12-03-2018 06:55 | Surrey Lane 25 | 85 | … |
| Open/close fridge | 12-03-2018 06:56 | Surrey Lane 25 | 83 | … |
| Start microwave | 12-03-2018 06:58 | Surrey Lane 25 | 86 | … |
| Start electric kettle | 12-03-2018 06:59 | Surrey Lane 25 | 86 | … |
| Motion in living room | 12-03-2018 07:04 | Surrey Lane 25 | 75 | … |
| Motion in kitchen | 12-03-2018 07:35 | Surrey Lane 25 | 84 | … |
| Motion in bathroom | 12-03-2018 07:42 | Surrey Lane 25 | 86 | … |
| Pressure on toilet seat | 12-03-2018 07:42 | Surrey Lane 25 | 88 | … |
| Start shower | 12-03-2018 07:45 | Surrey Lane 25 | 85 | … |
| Motion in bedroom | 12-03-2018 07:59 | Surrey Lane 25 | 83 | … |
| Motion in hallway | 12-03-2018 08:11 | Surrey Lane 25 | 84 | … |
| Open/close front door | 12-03-2018 08:15 | Surrey Lane 25 | 83 | … |
| … | … | … | … | … |

*Analytics* (LA) [BI14], two distinct research communities that both focus on the analysis of data related to learning and/or student behavior. The *Ambient Intelligence* [CAJ09] and *Ubiquitous Computing* [Che+11] research fields focus analysis on data that originates from human activity, where the ambient intelligence community focuses on smart homes or other smart environments while the ubiquitous computing community focuses on data from mobile phones or on-body sensors. A more specific research area is the *Activity Recognition* [Che+12a] research area, which focuses on detecting *human activities* from sensor readings, and on detecting patterns between those human activities.

In the business process management domain, there is typically either a normative process model in place that the execution of the process is supposed to adhere to, or, when this is not the case, there typically is at least a common understanding amongst process workers about the business goal of the process and the process steps that are needed to achieve those business goals. Such a normative process and a common understanding of the goals of the process are not always available in some of the novel application domains of process mining. Figure 1.3 visualizes the context of process discovery in non-BPM domains. Observe that in Figure 1.2, the IT system constrained the behavior in the recorded traces by constraining the behavior that the IT system allowed for. In contrast, the non-BPM context the IT system solely plays the role of *observer* and it does not influence or constrain the behavior. Using a smart home context as an example, the presence of sensors in the home does not influence the living behavior of the inhabitants of the home. This means that there are no orderings of activities that are explicitly disallowed for by the system: as far as the system is concerned, all possible sequences of activities have a non-zero probability of occurring. However, not all orderings of activities are equally likely, and therefore the event log that is collected by the logging system will contain certain orderings of activities more frequently than others. The goal of process discovery of such logs is to unveil and visualize the mainstream behavior.

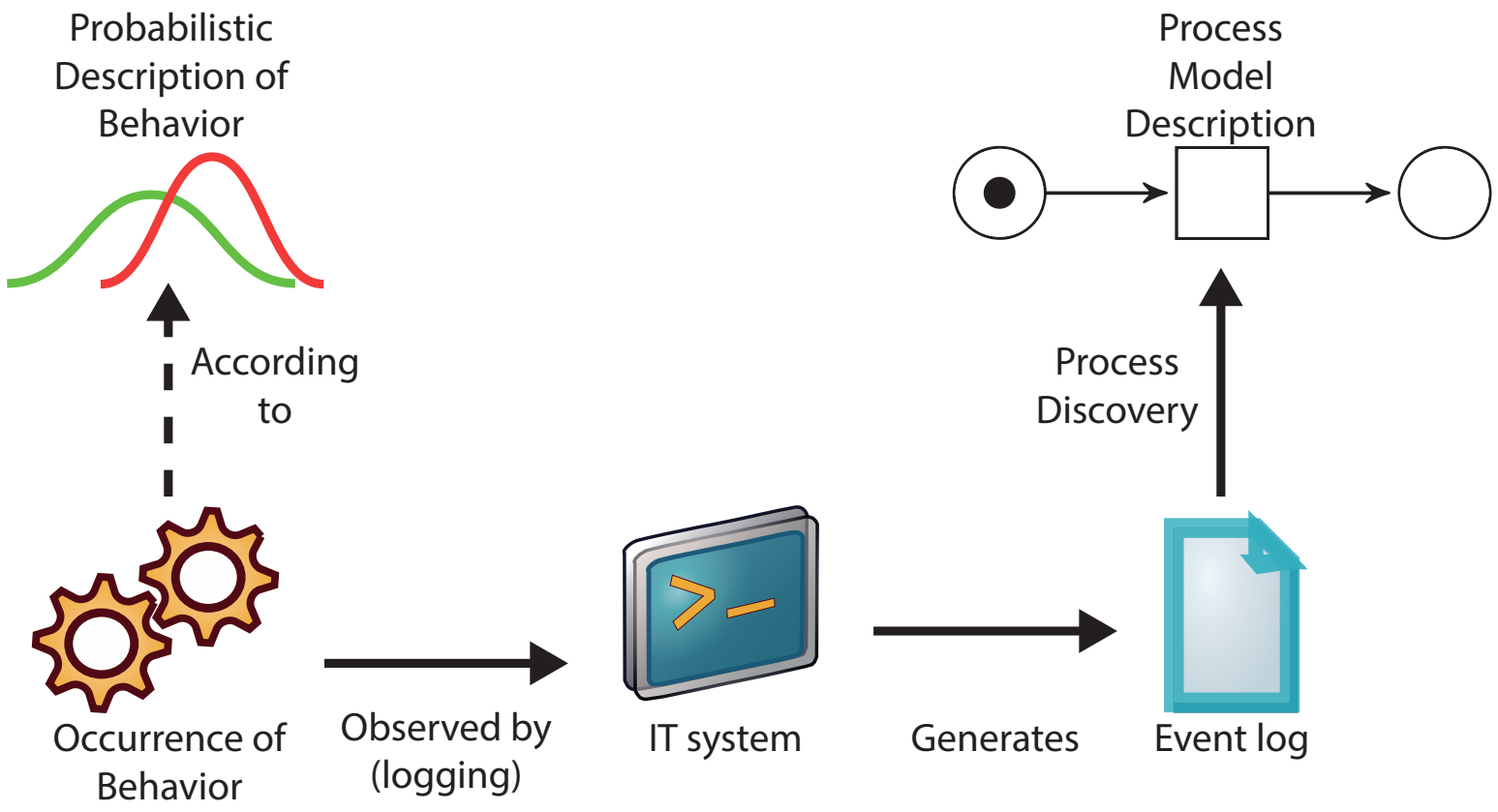

**Figure 1.3:** Process discovery in the context of non-BPM domain.

As a result of the absence of a normative process in the non-BPM domain of smart homes, the event logs tend to have a higher variability in the behavior of the log. Another example of a non-BPM domain where there is no normative process is in event logs of student behavior in MOOCs on platforms such as Coursera[3] and edX[4], as students can freely choose the order in which they watch the available video lectures and in which they work on the exercises. Furthermore, not all students of the MOOC share the same goal: while some students are motivated to fully pass the course and obtain a certificate of completion, others are solely interested in one of the topics covered in the course and will only study the material on this topic. Additionally, some students might not know a priori whether they are interested in joining the course, and are just having a look at the course material to decide to which degree they will study the course material. In the remainder of this thesis, we will refer to event logs that originate from application domains where there is no normative process in place and where as a result the ordering of activities in the log is highly variable as *weakly-structured event data*.

**Property 1.** Weakly-structured event logs contain highly variable behavior.

The higher variability in the behavior of the logs from weakly-structured application domains and the lack of a normative process from which the events were

[3]https://www.coursera.org
[4]https://www.edx.org/

generated make it hard to obtain insights into the behavior of the log using process discovery techniques. Depending of the process discovery technique that is applied, the resulting process models obtained from weakly-structured event logs tend to be generalizing too much, therefore allowing for a lot of behavior that was not observed in the data (i.e., low precision), or alternatively, they tend to be too complex for human interpretability (i.e., low simplicity)

Leotta et al. [LMM15] identified several challenges for applying process mining techniques in smart spaces. One of the main challenges that they identified was the existing gap between events on the sensor-level (e.g., some motion sensor registers motion, or some pressure sensor senses pressure) on the one hand and actual real-world events on the other hand (e.g., eating, sleeping). Note that there are two aspects to this challenge:

1. Sometimes we do not know the label of an event, i.e., an event from the log that was generated by the humidity sensor in the bathroom might correspond to a human activity *showering*, but this label is not directly available in the log.
2. Sometimes we do not know the events themselves, i.e., an event from the log that was generated by a pressure sensor on the toilet seat might not map one-to-one to am event of the human activity *going to the toilet*, as some other events that were generated by the motion sensor in the bathroom that preceded or succeeded this toilet seat pressure sensor event also belong to the same event in human activity terms.

We jointly summarize these two points in the following property of weakly-structured event data:

**Property 2.** Usable events or usable event labels are not always readily available in weakly-structured event logs.

Many process mining techniques focus on modeling the control-flow of the activities (i.e., the way in which activities are ordered) in the process. In human activity, there are many contextual factors that play a role and that do not directly have to do with the ordering of activities, such as the time of the day.

**Property 3.** Patterns in weakly-structured event data are not always limited to the control-flow.

## 1.3 Research Goals and Contributions

In this thesis aim for methods and techniques to get interpretable insights in the behavior that is contained in event logs generated from weakly-structured domains. We now map the three properties of weakly-structured event logs that we identified in Section 1.2 to three research goals for this thesis.

### 1.3.1 Research Goals

A special focus in this thesis will be put on weakly-structured event data that originates from smart home environments and ubiquitous systems that record events from human behavior. The first mention of the application of process mining in this application domain dates back to 2004 by de Medeiros et al. [Med+04]. In this work, de Medeiros et al. [Med+04] envisioned applications of process mining on data from ubiquitous environments in healthcare, by supporting an audit of medical protocols by tracking doctors and nurses though the hospital. In the same work, de Medeiros et al. [Med+04] make the claim that in order to successfully apply process discovery in ubiquitous environments, the process discovery algorithm should be able to deal with short loops in the data. Based on this observation, de Medeiros et al. [Med+04] propose an extension of an existing process discovery algorithm (i.e., the $\alpha^+$ Miner) that extends the original algorithm (i.e., the $\alpha$ Miner [Aal+04]) with capabilities to deal with short loops. However, the $\alpha^+$ Miner is not able to deal with high variability that is found in weakly-structured event data (i.e., property 1), therefore the algorithm is not very useful on real-life data from smart homes.

**Research Goal 1. (RG1)** Towards mining insights into the behavior of event logs that are highly variable.

More recently, several works [Dim+16; LMM15; SLM18] proposed the use of fuzzy models [GA07] instead of Petri nets to represent the behavior in event logs from smart home or ubiquitous systems. Fuzzy models do not specify the full behavior of the system. Instead, they specify the degree of association between activities. Fuzzy model do not have formal semantics, i.e., it is not defined which sequences of activities are and which are not possible according to the model. Additionally, fuzzy models do not contain information on what is executed concurrently. In this thesis, we set the goal of mining insights from highly variable event logs in such a way that we are also able to mine parallel execution explicitly.

Property 2 of weakly-structured event concerns the lack of usable events or usable event labels. Several techniques [BA09; BM13; Faz+15; FGP15; FSR14; Man+16b; Man+18] aim to address the preprocessing of events into more high-level events, thereby closely relating to the lack of usable events in the event log. However, none of these techniques leverage time information of the events as part of the abstraction approach.

Very little work has been done in the area of finding suitable event labels. However, there has been work in process discovery approaches that as a subprocedure of the process discovery algorithm automatically learns to distinguish between different versions of some event label based on the context of preceding and succeeding events (e.g., [BD17; BDA12a; Her00; Med+04; Van14]). However, it would be useful to develop stand-alone techniques to preprocess event labels into usable event labels that are not integrated into a process discovery approach but instead result in another event log that can then be analyzed with any type of analysis technique for event data. Furthermore, no techniques currently leverage time information or other data attributes of events in refining event labels.

**Research Goal 2. (RG2)** Towards mining insights from event logs where either the events or the event labels are not readily available.

The use of timestamp attributes in process mining is mostly limited to the use case of performance analysis [AAD12], where the event data is projected onto a process model and timestamps are used to detect which parts of the process model take time. In the smart home application domain, dependencies between activities van be time-dependent, e.g., for someone who has a habit of showering in the morning before breakfast there might be a relation "showering is followed by eating" that holds true in the morning but might not hold true for showering and eating events at other times of the day.

**Research Goal 3. (RG3)** Towards mining insights that unveil dependencies between time and control-flow.

### 1.3.2 Contributions

We categorize our contributions in two bodies of work. The first body of work focuses on *pre-processing event data* into more structured event data to aid mining insights from it. The second body of work focuses on *mining local process models* that describe frequent fragments of behavior from event logs expressed in control-flow constructs. This thesis is structured into two parts, with the first part focusing on pre-processing techniques and the second part focusing on local process models. We now proceed by giving an overview of the pre-processing part of this thesis, listing the contributions of that part, and relating these contributions to the three research goals.

#### *Event Log Pre-processing*

Figure 1.4 provides an overview of the pre-processing part of this thesis. We will now discuss the contributions of each element in this overview figure in relation to the research goals.

*RG1: Towards mining insights into the behavior of event logs that are highly variable.*
In order to mine insights from variable event logs, it can help to first transform the event log such that frequent structures in the event log become more apparent before applying any mining technique such as some process discovery algorithm.

In some instances, the high variability of the event log can be attributed to the labels of the events. For an event with the label *eating* it might be close to random which other types of events follow or precede it, thereby making it very hard for a process discovery or other type of mining algorithm to identify any relation between eating and other activities. However, if we assume that these eating events additionally have a data attribute that specifies the type of food that is being eaten, it might in fact be the case that after eating spaghetti the next activity typically is *doing the dishes*, while for example, the next activity after eating a banana is still close to random. If we would transform the event label of all the events that are labeled *eating* into more fine-grained event types *eating banana*, *eating spaghetti*, and so forth, this would enable process discovery techniques to afterward discover the relation between *eating spaghetti* and *doing the dishes*. We call such a transformation of an event label based on data attribute values of the event a *label refinement*.

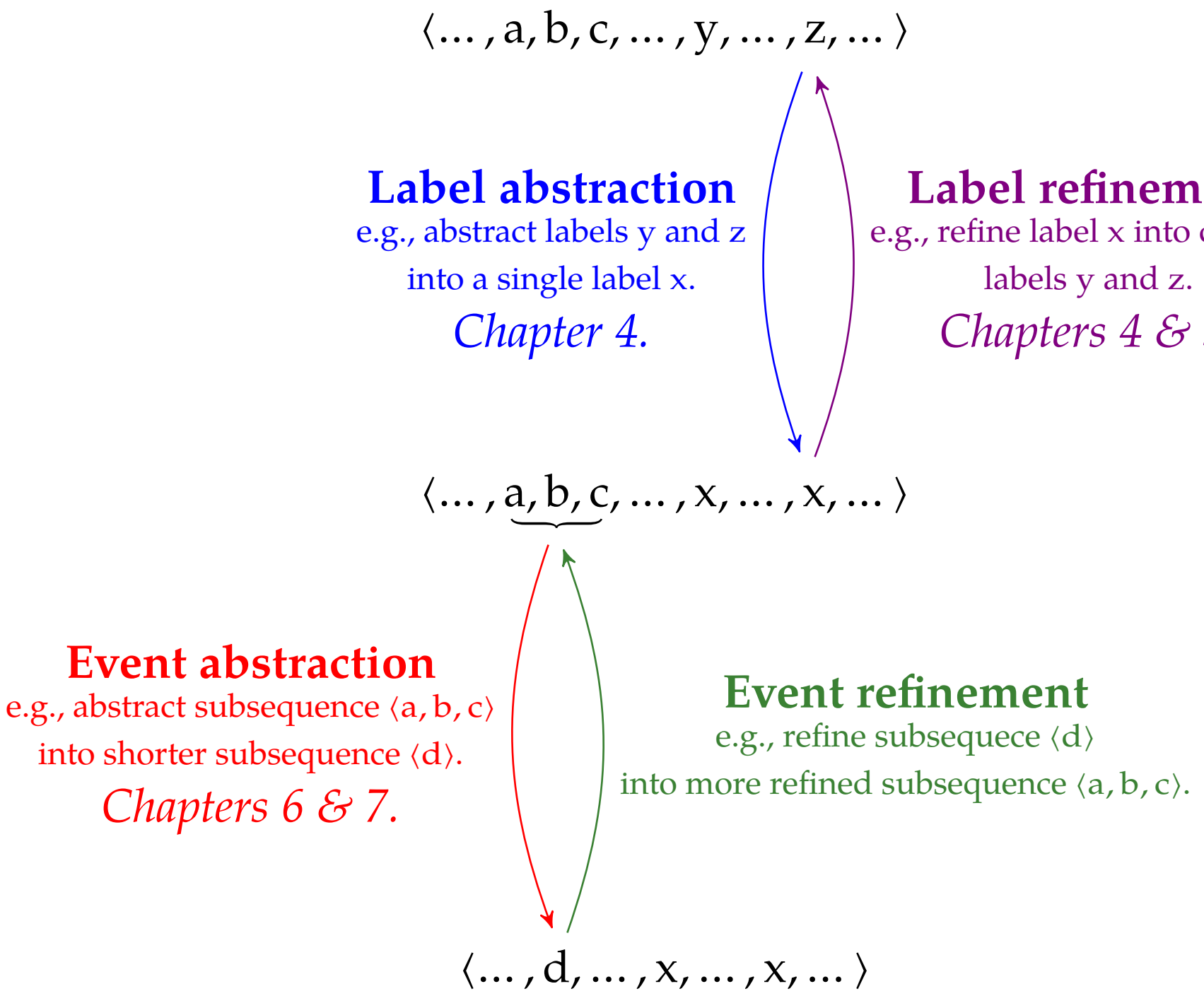

**Figure 1.4:** A taxonomy of event log pre-processing methods.

Note that in some instances, the inverse of a label refinement, which we call a *label abstraction* can also be of use. To motivate this, consider that the three events labels *eating spaghetti*, *eating lasagna*, and *eating boeuf bourguignon* are unlikely to differ much in their relations to other activities in the event log, and to avoid an explosion in the number of activities in the log it might be preferable to abstract those labels into a single label *eating dinner*.

We designed a statistical approach to evaluate whether a given proposal for a re-finement of a given event label into some more fine-grained event label contributes positively to the degree of structure in the event log (i.e., whether it reduces the degree of variability). This statistical test is described in Chapter 4. In Chapter 5 we propose an automated way to generate proposals to refine event labels in a way that is domain-specific to timestamp attributes. By combining this technique with the statistical test we are able to automatically refine event labels based on timestamp attributes. This enables us to make refinements of the sorts of *eating* into *breakfast*, *lunch*, *dinner*, *other eating* (or, snack) and of *sleeping* into *night sleep* and *daytime nap*.

Orthogonal to the search for the right event labels is the search for the right events. In this thesis we show that the degree of variability of an event log can depend on the level of granularity of events, i.e., abstracting events to higher-level events can unveil structure in the event log that is not observable from the original low-level events. In Chapter 7 and Chapter 12 we propose techniques to address the challenge of abstracting event to higher-level events that respectively are supervised (i.e., requiring both low-level event data as well as some event data that is on the desired level of abstraction) and unsupervised (i.e., requiring only the low-level events). Additionally, we argue that filtering of event logs is a special type of event abstraction and we propose a technique to filter out "chaotic" activities (i.e., activities of which the occurrence is highly random) from event logs in Chapter 6.

*RG2: Towards mining insights from event logs where either the events or the event labels are not readily available.*
We consider event logs where the event labels are not readily available to be event logs where the events consist of multiple event attributes and where the labeling that is needed to unveil structures in the event logs is a complex function that depends on multiple of those event attributes. We propose to address the challenge of mining insights from event logs where there are no event logs readily available by choosing one of the event attributes as the starting event label and then iteratively applying the label refinement techniques of Chapter 4 and Chapter 5.

The event abstraction techniques of Chapter 7 and Chapter 12 address this goal by enabling the analyst to abstract low-level events to events that are on a higher abstraction level.

*RG3: Towards mining insights that unveil dependencies between time and control-flow.*
The label refinement technique of Chapter 5 brings elements from the timestamp data into the event label and thereby enables the discovery of relations between time and control-flow.

### *Local Process Models*

Figure 1.5 provides an overview of the local process models part of this thesis. We will now discuss the contributions of each element in this overview figure in relation to the research goals.

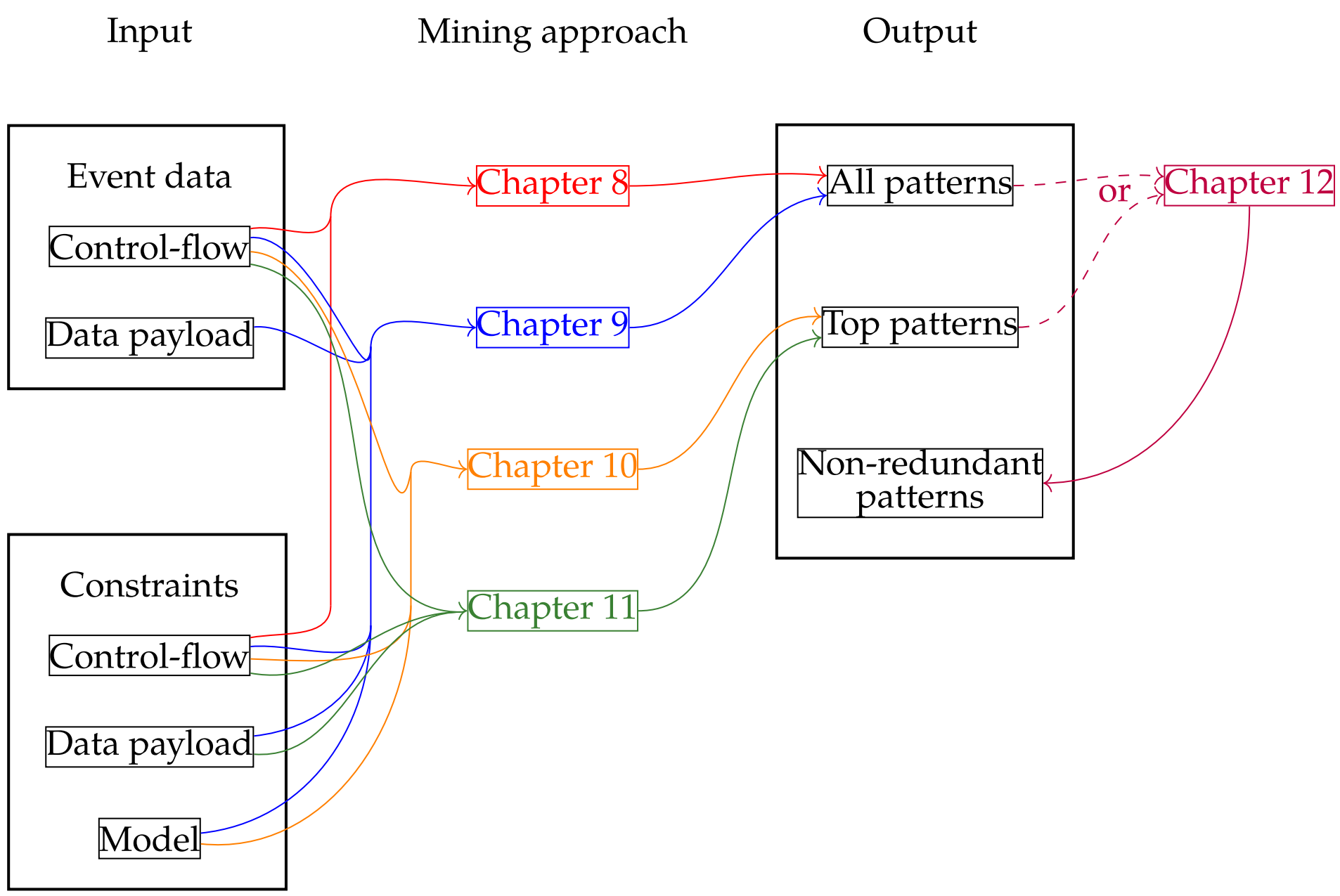

**Figure 1.5:** A taxonomy of local process model techniques.

*RG1: Towards mining insights into the behavior of event logs that are highly variable.* Process discovery techniques mine a global model, i.e., they return a single model that aims to represent the behavior that was seen in the event log. We will show in this thesis that for highly variable event log there might not exist a single process model that by itself accurately describes the behavior in the log without describing too much behavior that has not been observed. In contrast to process discovery techniques, pattern mining techniques, such as sequential pattern mining, mine a collection of patterns instead of a single model. This allows each pattern to describe frequent behavior from a highly variable event log without as a side-effect describing behavior that has never been observed in the log. In this thesis, we contribute a novel type of pattern mining, called *local process model* (LPM) mining, that focuses on mining small fragments of frequently occurring process behavior from event logs. These LPM patterns are able to express ordering constructs like sequential ordering, (exclusive) choice, concurrency, and loops, which are commonly used in the business process modeling community. At the same time, mining LPM pattern is are particularly well-suited to mining insights from highly variable event logs due to that fact that each pattern individually is only required to capture a single piece of frequent behavior. Therefore, Part II in its entirety closely links to RG1.

*RG2: Towards mining insights from event logs where either the events or the event labels are not readily available.*
The topic of LPM mining is less strongly related to RG2 than to RG1. However, in Chapter 12 we will show a particular application of LPM patterns that does closely link to RG2: we will show that LPMs can be used to abstract an event log to an event log with more abstract higher-level events in an unsupervised way.

*RG3: Towards mining insights that unveil dependencies between time and control-flow.*
In Chapter 11 we propose a type of LPMs that can be used to discover relations between control-flow perspective and the time aspect. These constraint-based LPMs work by formulating a maximum time gap that is allowed between two events in order to still be considered part of the same pattern instance. Thereby, it gives insight in how frequently certain control-flow patterns occur in the event log given a maximum time constraint.

## 1.4 Thesis Overview

This thesis is structured into two parts and 14 chapters. The first two chapters are positioned before the start of part I:

**Chapter 1.** This is the chapter that you are currently reading, introduces and describes the topic, the goals, and the contributions of this thesis.

**Chapter 2.** Introduces formal concepts and notation that are used throughout the remainder of this thesis.

### 1.4.1 Part I: Pre-processing of Event Data

This part discusses pre-processing techniques for event data that aim at enabling the discovery of insightful ordering relations between the activities in the event data. This part contains the following chapters:

**Chapter 3.** In this chapter we discuss the success criteria by providing a formal analysis of measures that capture when a process model accurately describes the behavior of an event log. This chapter is partly based on [Tax+18c].

**Chapter 4.** We introduce the concept of refinements for event labels and provide a statistical test to evaluate whether some given refinement from the original event labels in the log to some more refined event labels helps to discover an accurate process model. This chapter is partly based on [Tax+16d].

**Chapter 5.** We propose a method to mine proposals for label refinements for events from smart home environments based on the timestamp at which the events took place. This chapter is partly based on [Tax+16a; Tax+18a].

**Chapter 6.** We introduce an approach for a novel type of filtering of event logs: filtering chaotic activities, i.e., activities that are unrelated to the occurrence of other activities and therefore can happen at any arbitrary point in time. This chapter is partly based on [TSA18].

**Chapter 7.** We discuss an approach to abstract events by grouping together multiple low-level events that together form one single higher level event. Note that this part contains three orthogonal types of pre-processing that can be applied in combination: 1) label refinements, 2) event log filtering, and 3) event abstraction. This chapter is partly based on [Tax+16c; Tax+18f].

### 1.4.2 Part II: Local Process Models

This part discusses techniques to mine frequent patterns of behavior from an event log where the patterns are represented as process models. We refer to these type of patterns as *Local Process Models (LPMs)*. This part contains the following chapters:

**Chapter 8.** We introduce the foundations of LPM mining from event logs. Additionally, we show how LPMs can be used to used to abstract events in the event log to higher level events, thereby creating a link with Chapter 7. This chapter is partly based on [Tax+16e] and [MT17].
The basic LPM mining technique that we introduce in this chapter mines LPMs based solely on the control-flow aspect of the input event log and then generates the set of all LPM patterns that satisfy a certain set of conditions that depend on the control-flow in the log, such as *support* and *confidence*.

**Chapter 9.** We describe a conceptual framework to mine LPMs not based on their frequency, but based on utility functions that describe the *usefulness* of a pattern of process behavior based on the log. This chapter is partly based on [Tax+18b].
The extension to LPM mining that is described in this chapter makes in the data attributes of the events in the event log in-scope for LPM mining.

**Chapter 10.** We introduce heuristic, approximate but fast, techniques for LPM mining that are able to deal with event logs with large numbers of activities. This chapter is partly based on [GTZ18; Tax+16b; TGZ17], of which [TGZ17] has been awarded the best paper award at the 2017 International Symposium on Data-Driven Process Discovery and Analysis (SIMPDA).
The LPM mining techniques that are proposed in this chapter do not find all LPMs that satisfy the *support* constraint, instead, the provide *best-effort* to mine a set of LPMs that are good according to some quality criteria.

**Chapter 11.** We consider the constraint types of a maximum time that two events of an LPM instance are separated (called *time gap constraints*) and a maximum number of events that are not explained by an LPM (called *event gap constraints*). Additionally, this chapter proposed algorithms for efficient mining of LPMs with time gap constraints and event gap constraints. This chapter is partly based on [Tax+18e], which has been awarded the best paper award at the 2018 International Conference on Intelligent Environments (IE).

**Chapter 12.** We present methods to mine smaller collections of LPMs. The aim is to still capture the main frequent behaviors in the event log while without overloading the analyst with an overload of frequent patterns. This chapter is based on a paper that is currently under review.

**Chapter 13.** We describe the implementations of all techniques related to LPMs and is intended to additionally serve as a manual for the LPM mining tool. This chapter is partly based on [Tax+18d].

### 1.4.3 Other Publications

In addition to the publications included in this dissertation, I have also published several other papers over the course of this PhD project:

- Two papers on heuristic approaches to mining LPMs with utility functions [DTN17; DTN18] (as second author).
- A new process discovery algorithm, called the Indulpet Miner, that brings ideas from the area of ensemble learning to the field of process discovery by combining the results of several other process discovery algorithms. This algorithm can additionally be seen as an application of LPMs, which are used as a subprocedure of the Indulpet Miner. [LTH18] (as second author).
- A large-scale comparison of 87 machine learning algorithms for document ranking in web search [HTB17; TBH15].
- An evaluation of deep learning recurrent neural network architectures for the task of predicting the continuation and the remaining cycle time of business process instances [Tax+17].
- An alarm-based approach based on machine learning techniques to predict when running instances of business processes require intervention from process workers [Tei+18] (as second author).
- An evaluation of recurrent neural network architectures and several other sequence prediction algorithms for predicting upcoming human activity in smart home environments [Tax18] (as sole author).

- A method to use (possibly discovered) process models as a probabilistic sequence classifier and a comparison of the several process discovery algorithms and the process models that they generate with existing probabilistic sequence classification algorithms from the machine learning field [TZT18]. An extension is currently under review.

# 2 Preliminaries

In this chapter, we introduce the basic concepts, definitions, and notation that we use throughout the thesis. We start off by recalling basic concepts of mathematics, such as sets, relations, and functions. We then continue by defining concepts related to event logs, such as events, event labels, and traces. We discuss concepts related to process models, including Petri nets, and introduce automated discovery of process models from event logs. We discuss notions related to the quality of process models in relation to event logs. Finally, we introduce concepts from the area of pattern mining, including frequent itemset mining, association rule mining, sequential pattern mining, and episode mining.

## 2.1 Sets, Multisets, Relations, Functions, and Sequences

**Definition 2.1 (Set).** A **set** is a possibly infinite collection of distinct elements. $X = \{a, b, c\}$ denotes a finite set that consists of elements a, b, and c. Typically, we denote sets using capital letters. Two sets A and B are equal if and only if they have precisely the same elements. $\emptyset$ denotes the empty set, i.e., the set that does not contain any elements. $|X|$ denotes the *cardinality* of set X, i.e., the number of elements of the set, e.g., $|\{a, b, c\}| = 3$. $\mathscr{P}(X)$ denotes the *power set* of X, i.e., the set of all possible subsets of X. For example, $\mathscr{P}(\{a, b, c\}) = \{\emptyset, \{a\}, \{b\}, \{c\}, \{a, b\}, \{a, c\}, \{b, c\}, \{a, b, c\}\}$.

We define several operations on sets:

- $A \times B$ denotes the *Cartesian product* of two sets A and B consists of all ordered pairs of elements of A and B respectively, i.e., $A \times B = \{(a, b) \mid a \in A \wedge b \in B\}$. For example, $\{a, b, c\} \times \{d, e\} = \{(a, d), (a, e), (b, d), (b, e), (c, d), (c, e)\}$.

- $X \setminus Y$ denotes the *set difference* between set X and set Y, which consists of the set of elements that are in set X but not in set Y, e.g., $\{a, b, c\} \setminus \{a, c\} = \{b\}$.

- $X \cup Y$ denotes the *set union* between set X and set Y, which consists of the set of elements that are either in set X or in set Y, e.g. $\{a, b\} \cup \{a, c\} = \{a, b, c\}$.

- $X \cap Y$ denotes the *set intersection* between set X and set Y which is defined as the set of elements that are both in set X and in set Y, e.g., $\{a, b\} \cap \{a, c\} = \{b\}$. $\diamond$

**Definition 2.2 (Relation).** A **relation** $R$ between a collection of sets is a subset of their Cartesian product. A *unary relation* $R$ on a set $X$ is a collection of 1-tuples of $X$, i.e., it is simply a subset $R \subseteq X$. A *binary relation* $R$ on set $X$ and set $Y$ is a collection of pairs of from $X$ and $Y$, i.e., it is a subset $R \subseteq X \times Y$. For a binary relation $R \subseteq X \times Y$ and given $x \in X$ and $y \in Y$ we write $xRy$ if $(x, y) \in R$. A *tertiary relation* $R$ on sets $X$, $Y$, and $Z$ is a collection of triplets from $X$, $Y$, and $Z$, i.e., it is a subset $R \subseteq X \times Y \times Z$.$\diamond$

An example of a type binary relation that is frequently used is the concept of a **partial order relation**.

**Definition 2.3 (Partial order relation).** A **partial order relation** over a set $X$ is a binary relation $\leq \subseteq X \times X$ such that $\forall_{x \in X} x \leq x$ (*reflexivity*), $\forall_{x,y \in X} x \leq y \wedge y \leq x \implies x = y$ (*antisymmetry*), and $\forall_{x,y,z \in X} x \leq y \wedge y \leq z \implies x \leq z$ (*transitivity*). $\diamond$

An example of a partial order relation is the set inclusion relation $\subseteq$ on the powerset $\mathcal{P}(X)$ of some set $X$. It is trivial to see that the set inclusion relation indeed is reflexive, antisymmetric, and transitive. For example, for set $X = \{1, 2\}$ with powerset $\mathcal{P}(X) = \{\emptyset, \{1\}, \{2\}, \{1, 2\}\}$, the relation contains the following elements $\subseteq = \{(\emptyset, \emptyset), (\emptyset, \{1\}), (\{1\}, \{1\}), (\emptyset, \{2\}), (\{2\}, \{2\}), (\emptyset, \{1, 2\}), (\{1\}, \{1, 2\}), (\{2\}, \{1, 2\}), (\{1, 2\}, \{1, 2\})\}$. Note that this order relation is called *partial* since it does not require all elements of the set to be related. Taking again relation $\subseteq$ as example, we have $\{1\} \not\subseteq \{2\}$ as well as $\{2\} \not\subseteq \{1\}$, i.e., no ordering over elements $\{1\}$ and $\{2\}$ is defined.

**Definition 2.4 (Total order relation).** A **total order relation** over a set $X$ is a partial order relation relation $< \subseteq X \times X$ such that $\forall_{x,y \in X} x < y \vee y < x$ (*connex property*).$\diamond$

We have seen that $\{1\} \not\subseteq \{2\}$ and $\{2\} \not\subseteq \{1\}$, showing that $\subseteq$ is not a total order relation. Examples of total order relations are the less-than relation ($<$) and the less-than-or-equal relation ($\leq$) on the set of real numbers $\mathbb{R}$.

**Definition 2.5 (Functions).** A *total function* is a relation that maps every element from $X$ to an element of $Y$, denoted as $f : X \to Y$. $X$ is called the *domain* of function $f$ and is denoted by $dom(f)$ and $Y$ is called the *codomain* of $f$ and is denoted $cod(f)$. The *range*, denoted $rng(f)$, of function $f$, is set $rng(f) = \{f(x) | x \in X\} \subseteq Y$.

A *partial function*, denoted by $f : X \nrightarrow Y$, is a relation that maps a subset $dom(f) = X' \subseteq X$ to elements of $cod(f) = Y$, thereby not forcing every element of $X$ to map to an element of $Y$. In case $X' = X$, a partial function is called a *total function*. We denote function composition of function $f : X \to Y$ and $g : Y \to Z$ with $g \circ f$, with $g \circ f : X \to Z$ being defined as $(g \circ f)(x) = \{g(f(x)) \mid x \in dom(f)\}$. We denote a function composition of function $f : X \to X$ with itself as $f^2 = f \circ f$, $f^3 = f \circ f \circ f$, and in the general case for any $i \in \mathbb{N}^+$, $f^i$ denotes $i$ applications of $f$.

A *multivariate function* is a function that depends on several arguments. For sets $X_1, \dots, X_n$ the domains of the n arguments of a multivariate function, it maps every element from $X_1, \dots, X_n$ to an element of Y. A multivariate function is denoted as $f : X_1 \times \dots \times X_n \to Y$. The domain $dom(f)$ of a multivariate function consists of the Cartesian product of $X_1, \dots, X_n$. The *arity* is the number of arguments of a function and it is denoted $ar(f)$, i.e., for $f : X_1 \times \dots \times X_n \to Y$, $ar(f) = n$. For example, for $f : X \to Y$, $ar(f) = 1$. ◇

**Definition 2.6 (Multiset).** A **multiset** (or bag) over a set X is a function $M : X \to \mathbb{N}$ where $\mathbb{N}$ is the set of natural numbers. For a multiset $M : X \to \mathbb{N}$ we write $[a_1^{M(a_1)}, a_2^{M(a_2)}, \dots, a_n^{M(a_n)}]$, where $\{a_1, a_2, \dots, a_n\} = \{a \in X \mid M(a) > 0\}$. For example, multiset $[a^2, b^3, c]$ consists of 2 a elements, 3 b elements and one c element. The set of all multisets over X is denoted $\mathcal{B}(X)$. A multiset $M \in \mathcal{B}(X)$ is a *multisubset* of another multiset $M' \in \mathcal{B}(X')$, denoted $M \subseteq M'$, if and only if $X \subseteq X'$ and $\forall x \in X : M(x) \leq M'(x)$. The *cardinality* of sets is lifted to multisets in the following way: given multiset $M \in X$, $|M| = \sum_{x \in X} M(x)$, e.g., $|[a^2, b^3, c]| = 6$.

Let $M_1 \in \mathcal{B}(X)$ and $M_2 \in \mathcal{B}(X)$ be two multisets. We define several operations on multisets:

- The *multiset sum*, denoted $M_1 \uplus M_2$, yields the multiset where $\forall x \in X : (M_1 \uplus M_2)(x) = M_1(x) + M_2(x)$. For example, $[a^2, b^3, c^2] \uplus [a^3, c, d] = [a^5, b^3, c^3, d]$.
- The *multiset union*, denoted $M_1 \cup M_2$, yields the smallest multiset such that both $M_1$ and $M_2$ are multisubsets of it, i.e., $\forall x \in X : (M_1 \cup M_2)(x) = \max\{M_1(x), M_2(x)\}$. For example, $[a^2, b^3, c^2] \cup [a^3, c, d] = [a^3, b^3, c^2, d]$.
- The *multiset intersection*, denoted $M_1 \cap M_2$ and it yields the largest multiset that is both a submultiset of $M_1$ and a submultiset of $M_2$, i.e., $\forall x \in X : (M_1 \cap M_2)(x) = \min\{M_1(x), M_2(x)\}$. For example, $[a^2, b^3, c^2] \cap [a^3, c, d] = [a^2, c]$.
- The *multiset subtraction*, denoted $M_1 - M_2$, is the multiset where $\forall x \in X : (M_1 - M_2)(x) = \max\{M_1(x) - M_2(x), 0\}$. For example $[a^2, b^3, c^2] - [a^3, c, d] = [b^3, c]$. ◇

**Definition 2.7 (Sequence).** A **sequence** specifies a particular ordering in which certain elements follow each other. A (finite) sequence of elements from some set X is a function $\sigma : \{1, \dots, n\} \to X$, where $\sigma(i) = a_i$ for any $1 \leq i \leq n$. We denote a sequence $\sigma : \{1, \dots, n\} \to X$ as $\langle a_1, a_2, \dots, a_n \rangle$. The set X that is used as the codomain of a sequence is often referred to as its *alphabet*. The length of a sequence $\sigma : \{1, \dots, n\} \to X$ is denoted $|\sigma|$, and it takes the value $|\sigma| = n$. An example of a sequence is $\sigma = \langle a, b, c, c, b, a \rangle$, which is a sequence over alphabet $\{a, b, c\}$ and $|\sigma| = 6$. The *empty sequence*, i.e. the sequence without any elements, or the sequence of zero length, is denoted $\langle\rangle$. $\sigma_1 \cdot \sigma_2$ denotes $X^*$ denotes the set of all sequences over alphabet X and $X^+ = X^* \setminus \{\langle\rangle\}$ denotes the set of all nonempty sequences over set X. $\sigma \upharpoonright_X$ *projects* a sequence $\sigma$ on the elements of set X, e.g. $\langle a, b, c, d, a\rangle \upharpoonright_{\{a,c\}} = \langle a, c, a\rangle$.

We define two operations on sequences:

- $\sigma_1 \cdot \sigma_2$ denotes the *concatenation* of sequences $\sigma_1$ and $\sigma_2$. For example, $\langle a, b, c\rangle \cdot \langle c, d, e\rangle = \langle a, b, c, c, d, e\rangle$.
- $\sigma_1 \diamond \sigma_2$ denotes the *shuffle* of two sequences $\sigma_1$ and $\sigma_2$, i.e., the set of all sequences that are obtained by interleaving the elements of $\sigma_1$ and $\sigma_2$. formally we have $\langle\rangle \diamond \sigma = \sigma \diamond \langle\rangle = \sigma$ and $\langle a\rangle \cdot \sigma_1 \diamond \langle b\rangle \cdot \sigma_2 = \{\langle a\rangle \cdot x \mid x \in \sigma_1 \diamond \langle b \diamond \sigma_2\} \cup \{\langle b\rangle \cdot \sigma_2 \mid \sigma_2 \in \langle a\rangle \cdot \sigma_1 \diamond \sigma_2\}$. For example, $\langle a, b, c\rangle \diamond \langle a, d\rangle = \{\langle a, b, c, a, d\rangle, \langle a, b, a, c, d\rangle, \langle a, a, b, c, d\rangle, \langle a, b, a, d, c\rangle, \langle a, a, b, c, d\rangle, \langle a, a, b, d, c\rangle\}$. $\diamond$

Functions can be lifted to sequences. A partial function $f : X \nrightarrow Y$ with domain $dom(f)$ is lifted to sequences over X using the following recursive definition: (1) $f(\langle\rangle) = \langle\rangle$; (2) for any $\sigma \in X^*$ and $x \in X$:

$$f(\sigma \cdot \langle x\rangle) = \begin{cases} f(\sigma) & \text{if } x \notin dom(f), \\ f(\sigma) \cdot \langle f(x)\rangle & \text{if } x \in dom(f). \end{cases}$$

For example, for sequence $\sigma = \langle 1, 2, 3, a, b, 4\rangle$ and function $f(n) = n + 1$ with $dom(f) = \mathbb{N}$, $f(\sigma) = \langle 2, 3, 4, 5\rangle$.

Functions can also be lifted to sets. A partial function $f : X \nrightarrow Y$ with domain $dom(f)$ is lifted to a set $X' \subseteq X$ as follows: $f(X') = \{f(x) \mid x \in (X' \cap dom(f))\}$. For example, for set $X = \{1, 2, 3, a, b, 4\}$ and function $f(n) = n + 1$ with $dom(f) = \mathbb{N}$, $f(X) = \{2, 3, 4, 5\}$. Combining the definition of composition of a function with itself with lifting a function to sets allows for repeated execution of a function $f : X \nrightarrow \mathcal{P}(X)$. For example, for $X = \{1, 2, 3, a, b, 4\}$ and $f(n) = \{n - 1, n + 1\}$ with $dom(f) = \mathbb{Z}$, $f(X) = \{0, 1, 2, 3, 4, 5\}$, $f^2(X) = \{-1, 0, 1, 2, 3, 4, 5, 6\}$, and $f^3(X) = \{-2, -1, 0, 1, 2, 3, 4, 5, 6, 7\}$.

For a *set* S of *sequences over alphabet* X, we define its prefix-closure $preclo(S) = \bigcup_{\sigma \in S}\{\sigma' \in X^* \mid \exists_{\sigma'' \in X^*} \sigma = \sigma' \cdot \sigma''\}$ as the set of all prefixes of all elements of sequences S. For example, $preclo(\{\langle a, b, c\rangle, \langle a, b, d, e\rangle\}) = \{\langle\rangle, \langle a\rangle, \langle a, b\rangle, \langle a, b, c\rangle, \langle a, b, d\rangle, \langle a, b, d, e\rangle\}$.

## 2.2 Graphs

**Definition 2.8 (Directed graph).** A **directed graph** is a tuple $DG = (N, E)$ where N is a set of *nodes* and $E \subseteq N \times N$ a set of *edges*. For each edge $(n, n') \in E$ we say that it is an *outgoing edge* of n and an *incoming edge* of $n'$. A *path* in a directed graph $DG = (N, E)$ is a sequence of nodes $\langle n_1, n_2, \dots, n_k\rangle$ such that each $n_i \in N$ and for each $1 \leq i < k$, $(n_i, n_{i+1}) \in E$. A *directed acyclic graph* (DAG) is a directed graph *DG* such that for all $n \in N$ there is no path $\langle n, \dots, n \rangle$. A *weighted directed graph* is a tuple $WDG = (N, E, W)$ where N and E form a directed graph and $W : E \rightarrow \mathbb{R}$ assigns an arc weight to each edge. A directed graph is called *connected* if for every pair of two nodes $n_1, n_2 \in N$ there exists a path $\langle n_1, \dots, n_2 \rangle$ or a path $\langle n_2, \dots, n_1 \rangle$. A directed graph is a *strongly connected graph* if for every pair of two nodes $n_1, n_2 \in N$ there exists a path $\langle n_1, \dots, n_2 \rangle$. ◇

**Definition 2.9 (Undirected graph).** An **undirected graph** is a graph where, in contrast to a directed graph, edges have no orientation, i.e. it is a tuple $UG = (N, E)$ with N is a set of *nodes* and $E \subseteq \{\{n_1, n_2\} \mid n_1, n_2 \in N\}$ a set of *edges*. A *path* in an undirected graph $UG = (N, E)$ is a node sequence $\langle n_1, n_2, \dots, n_k \rangle$ such that each $n_i \in N$ and for each $1 \leq i < k$, $\{n_i, n_{i+1}\} \in E$. An *undirected acyclic graph*, also called a *forest*, is an undirected graph *UG* such that for all $n \in N$ there is no path $\langle n, \dots, n \rangle$. A *weighted undirected graph* is a tuple $WUG = (N, E, W)$ where N and E form an undirected graph and $W : E \rightarrow \mathbb{R}$ assigns an arc weight to each edge. An undirected graph is called *connected* if for every pair of two nodes $n_1, n_2 \in N$ there exists a path $\langle n_1, \dots, n_2 \rangle$. ◇

**Definition 2.10 (Tree).** A **tree** is a connected undirected acyclic graph. A *rooted tree* is a tree in which one of its nodes $r \in N$ is given special status, and is referred to as the *root* of the tree. The nodes $\{n \in N \mid \{r, n\} \in E\}$ that are connected to the root are called its *children* or *child nodes*. In general, each of the nodes $\{n_1, n_2, \dots, n_k\}$ that are connected to $n \in N$ and are one edge further away from the root are called is called a *child* of n. Nodes $\{n_1, n_2, \dots, n_k\}$ that are connected to the same node n and that share the same distance from the root node are called *siblings*. Nodes that do not have and children are called *leaves* or *leaf nodes*. ◇

## 2.3 Event Logs and their Components

In this thesis, we use some definitions regarding event logs in comparison to the majority of the process mining literature (e.g., the definitions used in [Aal16]). The motivation to deviate from standard definitions is the fact that part of this thesis concerns the question of how to appropriately label events in an event log (see research goal 2). Therefore, we will separately define the concept of an *unlabeled event log* and the concept of a *labeled event log* and we will introduce the concept of a *labeling function* to link the two.

**Table 2.1:** An example log from a smart home environment over.

| Event id | Timestamp | Sensor | Address | Heart rate |
|---|---|---|---|---|
| 1 | 12-03-2018 06:48 | Motion in bedroom | Surrey Lane 25 | 80 |
| 2 | 12-03-2018 06:55 | Motion in kitchen | Surrey Lane 25 | 85 |
| 3 | 12-03-2018 06:56 | Open/close fridge | Surrey Lane 25 | 83 |
| 4 | 12-03-2018 06:58 | Start microwave | Surrey Lane 25 | 86 |
| 5 | 12-03-2018 06:59 | Start electric kettle | Surrey Lane 25 | 86 |
| 6 | 12-03-2018 07:04 | Motion in living room | Surrey Lane 25 | 75 |
| 7 | 12-03-2018 07:35 | Motion in kitchen | Surrey Lane 25 | 84 |
| 8 | 12-03-2018 07:42 | Motion in bathroom | Surrey Lane 25 | 86 |
| 9 | 12-03-2018 07:42 | Pressure on toilet seat | Surrey Lane 25 | 88 |
| 10 | 12-03-2018 07:45 | Start shower | Surrey Lane 25 | 85 |
| 11 | 12-03-2018 07:59 | Motion in bedroom | Surrey Lane 25 | 83 |
| 12 | 12-03-2018 08:11 | Motion in hallway | Surrey Lane 25 | 84 |
| 13 | 12-03-2018 08:15 | Open/close front door | Surrey Lane 25 | 83 |
| … | … | … | … | … |

An **event** is the most elementary component of an event log. Let $\mathbb{I}$ be a set of event identifiers, $\mathbb{T}$ be a set of timestamps and $\mathcal{A}_1 \times \dots \times \mathcal{A}_n$ be an *attribute domain* consisting of n attributes. Examples of such attributes $\mathcal{A}_i$ can be the name of the activity, the resource or employee that executed the activity, costs that were involved in executing the activity (financial or otherwise), etc. In the context of events that are generated in a smart home environment, events attributes $\mathcal{A}_1 \dots \mathcal{A}_n$ usually include the type of sensor that generated the event, the location of the sensor that generated the event, or some value of measurement that was read by the sensor.

**Definition 2.11 (Event).** An event is a tuple $e = (i, t, a_1, \dots, a_n)$, with *identifier* $i \in \mathbb{I}$, *timestamp* $t \in \mathbb{T}$, and *data payload* (also called *attributes*) $(a_1, \dots, a_n) \in \mathcal{A}_1 \times \dots \times \mathcal{A}_n$. For an event e, $\pi_{id}(e)$ and $\pi_{time}(e)$ respectively denote the identifier and the timestamp of event e. $\mathcal{E} = \mathbb{I} \times \mathbb{T} \times \mathcal{A}_1 \times \dots \times \mathcal{A}_n$ is a universe of events over a given attribute domain $\mathcal{A}_1, \dots, \mathcal{A}_n$. ◇

Often, the insights that we want to obtain by analyzing events correspond to relations between events. Therefore, it is generally not interesting to consider an event in isolation, but instead, events are considered in the context of other events. We call a set of events $E \subseteq \mathcal{E}$ an **event set** when all events in E have the same data attributes (but not necessarily the same attribute values). An example of an event set is shown in Table 2.1, where the events have data attributes *sensor*, *address*, and *heart rate*.

Often it is useful to partition an event set into smaller sets such that the events that are grouped together belong together according to some criterion. We might, for example, be interested in discovering the typical behavior of households over the course of a day. In order to do so, we can for example group together events with the same *address* and the same day-part of the *timestamp*. In the context of a loan application process, one might be interested in the typical way in which loan applications are handled, therefore grouping together the events by the loan application in the context of which they were performed. For each of these event sets, we can construct a trace, and again timestamps define the ordering of events within the trace. For events of a trace having the same timestamps, an arbitrary ordering can be chosen within a trace. We call such an ordered group of events a trace and we call the set of attributes that share the same value amongst all events the **trace identifier**, with domain $T_{id}$. We refer to the grouping of an event set into subsets of related events as a *case notion* and define it as a function $case : E \rightarrow T_{id}$ that defines the partitioning of an arbitrary set of events $E \subseteq \mathcal{E}$ from a given event universe $\mathcal{E}$ into event sets $E_1, \dots, E_j, \dots$ where each $E_j$ is a maximal subset of E such that for any $e_1, e_2 \in E_j$, $case(e_1) = case(e_2)$; the value of *case* shared by all the elements of $E_j$ defines the value of the *trace attribute* $T_{id}$.

**Definition 2.12 (Trace).** A **trace** $\sigma$ is a finite sequence formed from the events of an event set $E \subseteq \mathcal{E}$ that respects the time ordering of events, i.e. for all $k, m \in \mathbb{N}$, $1 \leq k < m \leq |E|$, we have: $\pi_{time}(\sigma(k)) \leq \pi_{time}(\sigma(m))$. ◇

**Definition 2.13 (Trace universe).** We define the **universe of traces** over event universe $\mathcal{E} = \mathbb{I} \times \mathbb{T} \times \mathcal{A}_1 \times \dots \times \mathcal{A}_n$, denoted $\mathcal{C}(\mathcal{E})$, as the set of all possible traces over $\mathcal{E}$. In other words, $\mathcal{C}(\mathcal{E})$ is the set of all possible sequences of events over the attributes spaces of the data payload $\mathcal{A}_1 \times \dots \times \mathcal{A}_n$ of $\mathcal{E}$. ◇

For example, the events of Table 2.1 form an event set and can be interpreted as a single trace $\sigma = \langle (1, \text{12-03-2018 06:48}, \text{Motion in bedroom}, \text{Surrey Lane 25}, 80), \langle (2, \text{12-03-2018 06:55}, \text{Motion in kitchen}, \text{Surrey Lane 25}, 83), \dots \rangle$ when we apply a case notion that maps all events to the same trace identifier. The events in this trace come from $\mathcal{E}_1 = \mathbb{I} \times \mathbb{T} \times \mathit{Sensor} \times \mathit{Address} \times \mathit{Heartrate}$ and $\sigma \in \mathcal{C}(\mathcal{E}_1)$. We omit $\mathcal{E}$ in $\mathcal{C}(\mathcal{E})$ and simply write $\mathcal{C}$ when the event universe is clear from context.

Note that complex, multidimensional trace attributes are also possible, i.e. a combination of the name of the person performing the event activity and the date of the event, so that every trace contains activities of one person during one day. The event sets obtained by applying an event partitioning can be transformed into traces (respecting the time ordering of events).

**Definition 2.14 (Unlabeled event log).** An **unlabeled event log** U over an event universe $\mathcal{E}$ is a set of traces, i.e., $U \subseteq \mathcal{C}(\mathcal{E})$. An unlabeled event log U is obtained by applying a case notion function *case* to an event set $E \subseteq \mathcal{E}$. The event log consists of one trace per subset of events $E_i$ that was generated by applying *case* to E. For each trace that represents a subset of events $E_i$ generated by applying *case* to E the events are ordered by their timestamp. ◇

In many real-life scenarios it is sufficient to consider a more simple view of event logs in which we abstract away from the data attributes of events and simply consider traces to be sequences of event labels. In this simpler view, event logs simply become multisets of traces. We refer to this alternative, simpler, definition of an event log as a **labeled event log**, in which, in contrast to an unlabeled event log, events no longer contain an event id, a timestamp, or data payload.

**Definition 2.15.** A *labeling function* is a function $l : \mathcal{C} \rightarrow \Sigma^*$ that maps an unlabeled trace to a labeled trace over a finite set of event labels $\Sigma$ in such a way that for all unlabeled traces $\sigma, \sigma' \in \mathcal{C}$ it holds that if $\sigma$ is a prefix of $\sigma'$ then $l(\sigma)$ is a prefix of or equal to $(\sigma')$. ◇

If we consider all events of Table 2.1 that occurred within the same day to form a trace and take as example labeling function l that considers the value of the sensor attribute to be the event label then the result of applying l is trace $\langle$Motion in bedroom, Motion in kitchen, Open/close fridge, Start microwave, ... $\rangle$.

**Definition 2.16 (Labeled event log).** For a set of event labels (or activities) $\Sigma$, a labeled event log $L \subseteq \mathcal{B}(\Sigma^*)$ is a multiset of sequences over $\Sigma$. A labeled event log is created from an unlabeled event log U by applying a labeling function l to U. Labeled event log $L = l(U)$ is the multiset $L : cod(l) \rightarrow \mathbb{N}$ such that where $\forall_{\gamma \in cod(l)} L(\gamma) = |\{\sigma \in U \mid l(\sigma) = \gamma\}|$. ◇

Note that applying a labeling function l to an unlabeled log can turn multiple unlabeled traces into identical sequences of event labels. Therefore, an *unlabeled log is a set of traces* while a *labeled log is a multiset of traces*. Throughout the thesis we will abstain from referring to labeled and unlabeled event logs and simply use the term event log when it is clear from context which of the two we refer to.

For a labeled event log L over the alphabet of event labels $\Sigma$ we define several simple operations:

- $\tilde{L}=\{\sigma \in \Sigma^+ \mid L(\sigma)>0\}$ is the **trace set** of L, i.e., the set of all traces variants that occur at least once in a labeled log. For example, for log $L = [\langle a,b,c\rangle^2, \langle b,a,c\rangle^3]$, $\tilde{L}=\{\langle a,b,c\rangle\langle b,a,c\rangle\}$.
- *Activities*$(L)$ denotes the set of labels $\Sigma$ that occur in labeled log L. For example, *Activities*$([\langle a,b,c\rangle^2, \langle b,a,c\rangle^3]) = \{a,b,c\}$.
- $\#(a, L)$ denotes the number of occurrences of activity a in log L. e.g., $\#(a, [\langle a,b,c,a\rangle^2, \langle b,a,c\rangle]) = 5$.

## 2.4 Process Models & Process Discovery

The goal of process discovery is to discover a process model that accurately represents the behavior that was seen in an event log. Van der Aalst [Aal16] proposed a framework consisting of four quality dimensions to specify what 'accurately' means in the context of process discovery. The framework consists of the quality dimensions *fitness*, *precision*, *generalization*, and *simplicity*. First, the discovered process model should allow for all the behavior of the log (i.e., *fitness*). Furthermore, it should allow for not too much more behavior than what was seen in the log (i.e., *precision*), it should not be unnecessarily complex (i.e., *simplicity*), and it should generalize beyond the behavior that was seen in the log (i.e., *generalization*).

### 2.4.1 Petri Nets and Other Types of Process Models

One process model notation that is particularly used frequently in the field of process mining is the Petri net [RR98]. One of the reasons that Petri nets are popular in the process mining field is that they have an exact mathematical definition in addition to a visual representation. This contrasts many other process modeling notations, including Business Process Model and Notation (BPMN) and UML statechart diagrams, that lack a compact mathematical definition and for which there exist multiple proposals for their execution semantics (e.g., for BPMN: [DDO08; Nic+09; WG08], and for UML statechart: [EW00; LMM99; Var02]).

#### *Petri nets*

Petri nets are directed bipartite graphs that consist of transitions and places that are connected by arcs. Transitions represent activities, while places represent the enabling conditions of those transitions. Labels are assigned to transitions to indicate the type of activity that they model. Some transitions are left unlabeled and are only used for routing purposes, i.e., they do not correspond to an occurrence of an activity that is recorded in the execution log. Such transitions without label are called *silent transitions*, or $\tau$-transitions. A Petri net with labeled transitions is formally defined as follows.

**Definition 2.17 (Labeled Petri net).** A *labeled Petri net* $N = \langle P, T, F, \Sigma, \ell \rangle$ is a tuple where $P$ is a finite set of places, $T$ is a finite set of transitions such that $P \cap T = \emptyset$, $F \subseteq (P \times T) \cup (T \times P)$ is a set of directed arcs, called the flow relation, $\Sigma$ is an alphabet of labels that represent activities, with $\tau \notin \Sigma$ being a label representing invisible events, and $\ell : T \nrightarrow \Sigma$ is a partial function that assigns a label to each transition, or leaves it unlabeled (i.e., the $\tau$-transitions). $\diamond$

For a node $n \in (P \cup T)$ we use $\bullet n$ and $n \bullet$ to denote the set of input and output nodes of $n$, defined as $\bullet n = \{n' \mid (n', n) \in F\}$ and $n \bullet = \{n \mid (n, n') \in F\}$. An example of a Petri net can be seen in Figure 2.1, where circles depict places, rectangles depict transitions, and gray rectangles depict unlabeled transitions (i.e. $\tau$-transitions).

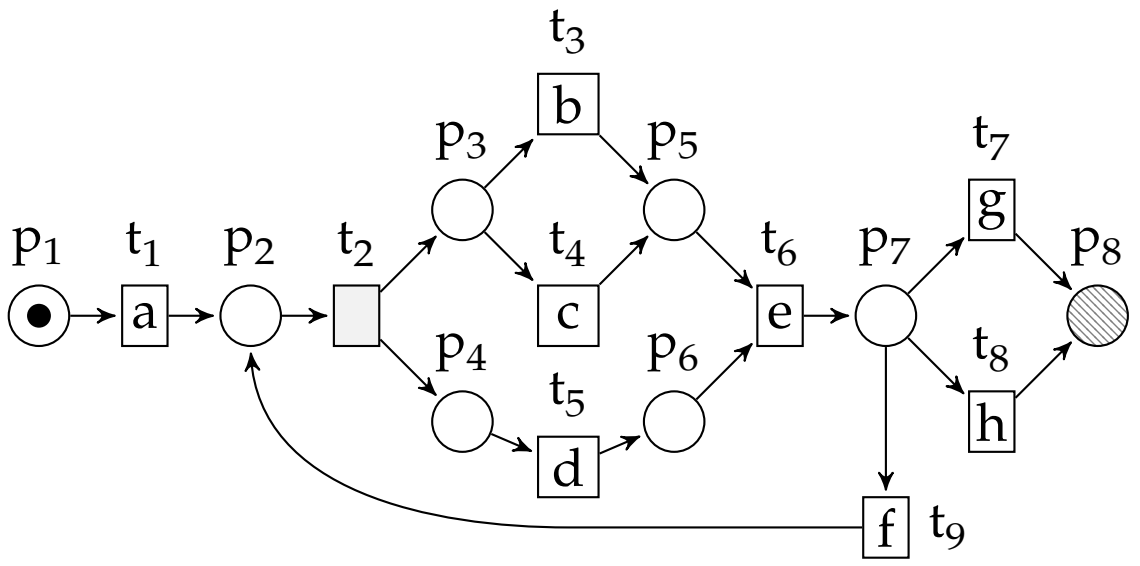

**Figure 2.1:** An example Petri net that describes the relations between activities a=register request, b=examine thoroughly, c=examine casually, d=check ticket, e=decide, f=re-initiate request, g=pay compensation, H=reject request.

A state of a Petri net is defined by its **marking** $M \in \mathcal{B}(P)$ that is a multiset of places. A marking is graphically denoted by putting $M(p)$ tokens on each place $p \in P$. A pair $(N, M)$ is called a marked Petri net. A state change in a marked Petri net occur when a transition fires. A transition t is enabled (i.e., it can fire) in a given marking M if each input place $p \in \bullet t$ contains at least one token, i.e. $\bullet t \subseteq M$. Once a transition t fires from a marking M from which t is enabled, one token is removed from each input place of t and one token is added to each output place of t, leading to a new marking $M'$ defined as $M' = M - \bullet t + t\bullet$.

The **firing** of a transition t leading from marking M to marking $M'$ is denoted as $M \xrightarrow{t} M'$, called a step. For example, in the Petri net of Figure 2.1, $[p_1] \xrightarrow{t_1} [p_2]$. We lift steps to sequences of firings of enabled transitions, writing $M_1 \xrightarrow{\gamma} M_2$, to indicate that $M_2$ can be reached from $M_1$ through a *firing sequence* of transitions $\gamma \in T^*$. For example, in the Petri net of Figure 2.1, $[p_1] \xrightarrow{\langle t_1, t_2, t_3 \rangle} [p_4, p_5]$. Since not firing any transition results in the marking remaining unchanged, we define $M \xrightarrow{\langle\rangle} M$ for any M.

A firing sequence in a Petri net corresponds to a sequence of observable activities $\ell(\gamma) \in \Sigma^*$. For example, $\ell(\langle t_1, t_2, t_3 \rangle) = \langle A, B \rangle$.[5] By defining an *initial marking* that indicates the state in which the process begins and a set of possible *final markings* in which process executions can end we can define the **language** that is accepted by a Petri net, i.e., the set of finite sequences of activities that it allows for. We call a labeled Petri net with an initial marking and a set of final marking an *accepting Petri net*, which we formally define as follows.

**Definition 2.18 (Accepting Petri net).** An *accepting Petri net* is a triplet $APN = \langle N, M_0, MF \rangle)$, where N is a labeled Petri net, $M_0 \in \mathcal{B}(P)$ is its initial marking, and

[5] Note that $\ell$ is a partial function (Definition 2.5) and $t_2 \notin dom(\ell)$.

$MF \subseteq \mathcal{B}(P)$ is its set of possible final markings. A sequence $\sigma \in \Sigma^*$ is a *trace* of an accepting Petri net *APN* if there exists a firing sequence $M_0 \xrightarrow{\gamma} M_f$ such that $M_f \in MF$, $\gamma \in T^*$ and $\ell(\gamma) = \sigma$. ◇

In the Petri nets that are shown in this paper (e.g., Figure 2.1), places that belong to the initial marking contain a token. Furthermore, for Petri nets where *MF* consists of a single possible final marking we will visually indicate the places belonging to that final marking as ◍. When *MF* consists of multiple possible final markings we will refrain from visually indicating the final marking and simply define set *MF* in addition to the Petri net.

**Definition 2.19 (Language of an accepting Petri net).** The *language* of an accepted Petri net *APN*, denoted $\mathfrak{L}(APN)$, is the set of all its traces, i.e., $\mathfrak{L}(APN) = \{\ell(\gamma) \mid \gamma \in T^* \wedge \exists_{M_f \in MF} M_0 \xrightarrow{\gamma} M_f\}$. ◇

Note that the language of an accepting Petri net *APN* can be of infinite size (i.e. $|\mathfrak{L}(APN)| = \infty$) when *APN* contains one or more cycles. An example of an accepting Petri net where this is the case is accepting Petri net $APN_1$ that given in Figure 2.1. The language of $APN_1$ is $\mathfrak{L}(APN_1) = \{\langle a,b,d,e,g\rangle, \langle a,c,d,e,g\rangle, \langle a,d,b,e,g\rangle, \langle a,d,c,e,g\rangle, \langle a,b,d,e,h\rangle, \langle a,c,d,e,h\rangle, \langle a,d,b,e,h\rangle, \langle a,d,c,e,h\rangle, \langle a,b,d,e,f,b,d,e,g\rangle, \langle a,c,d,e,f,b,d,e,g\rangle, \langle a,d,b,e,f,b,d,e,g\rangle, \langle a,d,c,e,f,b,d,e,g\rangle, \langle a,b,d,e,f,b,d,e,h\rangle, \langle a,c,d,e,f,b,d,e,h\rangle, \langle a,d,b,e,f,b,d,e,h\rangle, \langle a,d,c,e,f,b,d,e,h\rangle, \ldots\}$.

We denote the universe of process models as $\mathcal{M}$, and it consists of all process models in any process modeling notation that has formal operational semantics, i.e. its behavior is well-defined. While in this section we have defined the language of accepting Petri nets, in theory, $\mathfrak{L}(PM) \subseteq \Sigma^+$ can be defined for any process model $PM \in \mathcal{M}$ that has formal semantics. In this thesis we will mostly express process models in the Petri net notation. A subclass of Petri nets that is frequently used in process discovery is the workflow net.

A workflow net (or, WF-net) is an accepting Petri net with a unique source place and a unique sink place.

**Definition 2.20 (Workflow net).** Let $APN = \langle N, M_0, MF\rangle$ be an accepting Petri net consisting of labeled Petri net $N = \langle P, T, F, \Sigma, \ell\rangle$. *APN* is a workflow net if there are places $i, o \in P$ such that $\bullet i = \emptyset$, $o\bullet = \emptyset$, all nodes $P \cup T$ are on a path from i to o, $M_0 = [i]$, and $MF = \{[o]\}$. ◇

A desirable property of a workflow net is that is sound. **Soundness** in a workflow net implies that from each marking M that is reachable starting from the initial marking $[i]$ it is always possible to reach the final state $[o]$ (i.e., it is free from deadlocks). Furthermore, once $[o]$ is reached from $[i]$, all the other places need to be empty (i.e., the model guarantees proper completion). Finally, there are no dead transitions, i.e., for each transition $t \in T$ there exists a marking that is reachable from $[i]$ such that $t$ is enabled.

### *Process Tree*

A type of process model that is always sound by design is the **process tree** [BDA12a]. A process tree is a tree $PT = (N, E)$ with $N$ a set of nodes and $E$ a set of edges where leaf nodes represent activities. The non-leaf nodes represent *operators*, which specify the allowed behavior over the activity nodes. Allowed operator nodes are the *sequence* operator ($\rightarrow$) that indicates that the first child is executed before the second, the *exclusive choice* operator ($\times$) that indicates that exactly one of the children can be executed, the *concurrency* operator ($\wedge$) that indicates that every child will be executed but allows for any ordering, and the *loop* operator ($\circlearrowright$), which has one child node and allows for repeated execution of this node. Note that our definition of the loop operator ($\circlearrowright$) with one child node deviates from how this operator was defined in the original process tree definition [BDA12a], where it was defined as an operator with three child nodes. We will motivate this alternative definition of the loop operator in Chapter 8 and detail the differences between the two definitions. We now provide a recursive formal definition of process trees.

**Definition 2.21 (Process tree).** An activity node $a \in \Sigma$ is a process tree, with $\mathfrak{L}(a) = \{\langle a \rangle\}$. If $M_1$ and $M_2$ are process trees, then:

- $\rightarrow(M_1, M_2)$ is a process tree, with $\mathfrak{L}(\rightarrow(M_1, M_2)) = \{\sigma_1 \cdot \sigma_2 \mid \sigma_1 \in \mathfrak{L}(M_1), \sigma_2 \in \mathfrak{L}(M_2)\}$
- $\times(M_1, M_2)$ is a process tree, with $\mathfrak{L}(\times(M_1, M_2)) = \mathfrak{L}(M_1) \cup \mathfrak{L}(M_2)$
- $\wedge(M_1, M_2)$ is a process tree, with $\mathfrak{L}(\wedge(M_1, M_2)) = \{\sigma_1 \diamond \sigma_2 \mid \sigma_1 \in \mathfrak{L}(M_1), \sigma_2 \in \mathfrak{L}(M_2)\}$
- $\circlearrowright(M_1)$ is a process tree, with $\mathfrak{L}(\circlearrowright(M_1)) = \mathfrak{L}(M_1) \cup \{\sigma_1 \cdot \sigma_2 \mid \sigma_1 \in \mathfrak{L}(M_1), \sigma_2 \in \mathfrak{L}(M_1)\} \cup \{\sigma_1 \cdot \sigma_2 \cdot \sigma_3 \mid \sigma_1 \in \mathfrak{L}(M_1), \sigma_2 \in \mathfrak{L}(M_1), \sigma_3 \in \mathfrak{L}(M_1)\} \cup \ldots$ ◇

Figure 2.2a shows an example process tree $M_4$, with $\mathfrak{L}(M_4) = \{\langle a, b, c \rangle, \langle a, c, b \rangle, \langle d, b, c \rangle, \langle d, c, b \rangle\}$. Informally, it indicates that either activity a or d is executed first, followed by the execution of activities b and c in any order. $M_4$ can also be written using shorthand notation as $\rightarrow(\times(a, d), \wedge(b, c))$. A process tree can be trivially transformed into a Petri net that has the same language as the original process tree. For example, Figure 2.2b shows the resulting Petri net after transforming process tree $M_4$. Process models that are obtained by transforming a process tree into a Petri net are guaranteed to belong to the class of sound workflow nets.

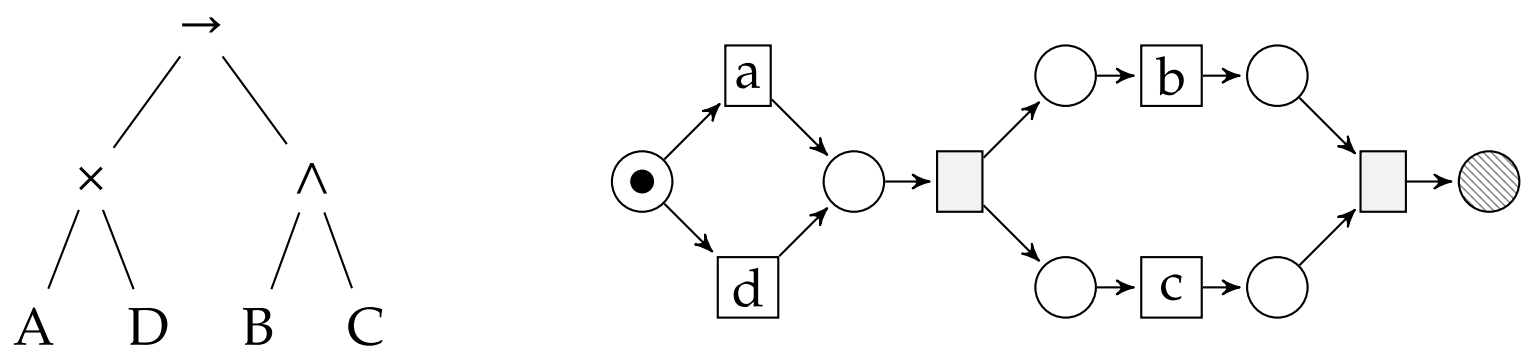

(a) (b)

**Figure 2.2:** *(a)* An example process tree over activities $\Sigma = \{a, b, c, d\}$, and *(b)* an equivalent accepting Petri net.

## 2.4.2 Discovering Process Models

A **process discovery** algorithm is a function $PD : \mathcal{B}(\Sigma^{+}) \rightarrow \mathcal{M}$ that provides a process model for a given labeled event log. The goal is to discover a process model that is a good description of the process from which the event log was obtained. An assumption that is frequently made in the area of process discovery is that there exists some ground truth process model that is often referred to as the *system*, denoted $PM^{sys}$, that is operational and by observing its behavior some event log L is obtained. Therefore, under ideal circumstances, $\tilde{L} \subseteq \mathfrak{L}(PM^{sys})$. However, in practice, the logging of the executions of $PM^{sys}$ is not always perfect and as a result the traces of L can contain some logging errors that are commonly referred to as *noise*.

### *Common Event Log Operations in Process Discovery*

We will now discuss several operations on labeled event logs that are frequently used within process discovery algorithms. In order to do so, we extend the function $\#(a, L)$ to the function $\#(\sigma, L)$ to count the number of occurrence of a subsequence $\sigma$, in L:

$$\#(\sigma, L) = \sum_{\sigma' \in L} |\{0 \leq i \leq |\sigma'| - |\sigma| \mid \forall_{1 \leq j \leq |\sigma|} \sigma'(i+j) = \sigma(j)\}| \qquad (2.1)$$

One of the most commonly used representation in process discovery is the *directly-follows relation*, which detects from an event log which activities b ∈ follow a certain activity $a \in \Sigma$. Formally, the directly-follows relation is defined as $df = \{(r, r') \in \Sigma \times \Sigma \mid \#(\langle r, r' \rangle, L) > 0\}$. One of the process discovery algorithms in which the directly-follows relation plays a central role is the Inductive Miner (IM) [LFA13b], which builds the directly-follows relation from an event log and specifies a procedure to transform the directly-follows relation into a process model. A more recent variant of the Inductive Miner, called the Inductive Miner infrequent (IMi) [LFA13a], takes into account noise in the event log by specifying a more noise-resilient variant to the directly-follows relation: $df_k = \{(r, r') \in \Sigma \times \Sigma \mid \#(\langle r, r' \rangle, L) > k\}$, for some frequency-threshold parameter $k \in \mathbb{N}^+$.

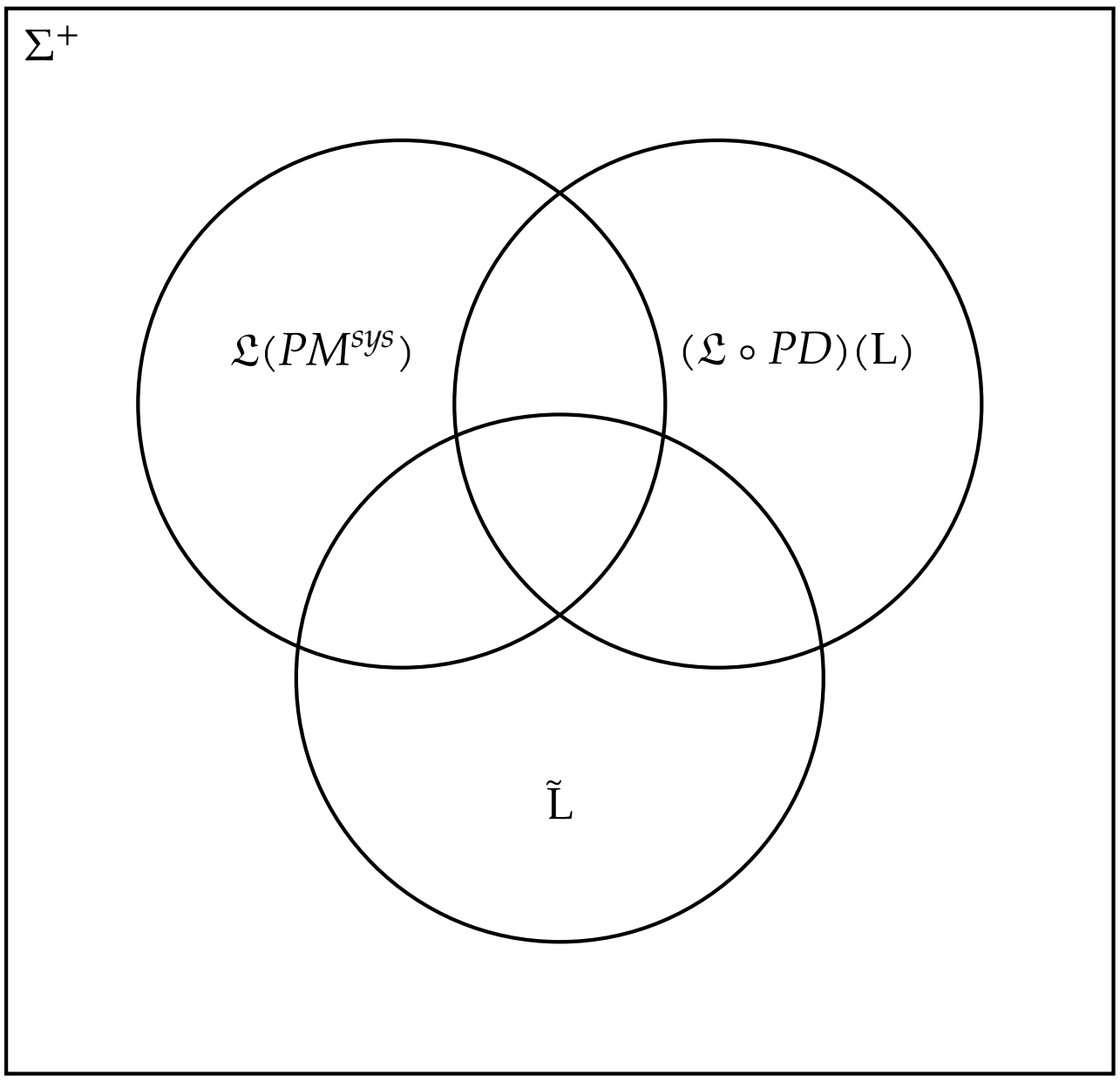

**Figure 2.3:** A Venn diagram containing the behavior of the system $\mathfrak{L}(PM^{sys})$, the behavior seen in the event log $\tilde{L}$, and the behavior of a discovered process model.

To be able to infer the start and and of the process it is common to pre-process the event log by adding explicit start and end events. $L^{\rfloor}$ contains the traces of event log L appended with an *artificial end event* that we represent with $\rfloor$. For each $\sigma = \langle e_1, e_2, \dots, e_n \rangle$ in log L, log $L^{\rfloor}$ contains a trace $\sigma^{\rfloor} = \langle e_1, e_2, \dots, e_n, \rfloor \rangle$. Likewise, $L^{\lfloor}$ contains the traces of event log L prepended with an *artificial start event* $\lfloor$, i.e., for each $\sigma = \langle e_1, e_2, \dots, e_n \rangle$ in log L, log $L^{\lfloor}$ contains a trace $\sigma^{\lfloor} = \langle \lfloor, e_1, \dots, e_n \rangle$. The artificial *start* and *end* events allow us to define the ratio of start events of an activity, e.g., $dfr(a, \rfloor, L^{\rfloor})$ and $dpr(a, \lfloor, L^{\lfloor})$ represent the ratio of events of activity a that respectively occur at the end of a trace and at the beginning of a trace.

### *The Aim of Process Discovery*

The aim of process discovery is to discover a process model that allows for all the behavior that was observed in the event log (called *fitness*) while it should not allow for too much behavior that was not seen in the event log (called *precision*). For a given process model $PD(L)$ that is discovered from a log L that was generated by executing a system $PM^{sys}$, Figure 2.3 shows a Venn diagram that contains the behavior that is described by the system, the log, and the discovered model. We say that the discovered model $PD(\mathrm{L})$ is *fitting* on a log L if all its traces are allowed for by the model, i.e. $\tilde{\mathrm{L}} \subseteq \mathfrak{L}(PD(\mathrm{L}))$. Note that in the ideal case the process discovery algorithm would rediscover the process model of the system, i.e. $PD(\mathrm{L}) = PM^{sys}$, in which case it is easy to see that the model is fitting when we do not consider noise.

**Fitness** measures the degree to which an event log is fitting on a process model, i.e. it is related to the size of $\tilde{\mathrm{L}} \cap \mathfrak{L}(PD(\mathrm{L}))$ with respect to the size of $\tilde{\mathrm{L}}$.

**Precision** measures how much of the behavior that is allowed for by the model was also observed in the log, i.e. it is related to the size of $\tilde{\mathrm{L}} \cap \mathfrak{L}(PD(\mathrm{L}))$ with respect to the size of $\mathfrak{L}(PD(\mathrm{L}))$.

Often, we are interested in models that provide a good trade-off between fitness and precision. This is analogous to the situation in supervised learning in data mining, where a desired predictive model strikes the right balance between having high *recall* (i.e., instances that belong to some "positive class" are predicted by the model as being of the positive class) and having high *precision* (i.e., instances that have been predicted to belong to the positive class are in fact belonging to the positive class). In the data mining field, the trade-off between recall and precision is often expressed in the form of F-score, i.e., the geometric mean between recall and precision. Analogous to the data mining field, De Weerdt [De +11] et al. proposed to use **F-score** to evaluate process models that are discovered from an event log by taking the geometric mean between fitness and precision.

In addition to fitness and precision, the following quality dimensions are often used in process mining:

**generalization** measure how much of the behavior of the system model is captured by the discovered process model, i.e. it is related to the size of $\mathfrak{L}(PD(\mathrm{L})) \cap \mathfrak{L}(PM^{sys})$ with respect to the size of $\mathfrak{L}(PM^{sys})$. However, in practice the system model $PM^{sys}$ is often not known (note that if it would have been known then it would not have been needed to discover a process model in the first place), therefore in practice it cannot be calculated. Several studies have tried to develop techniques that estimate generalization purely based on the log and the model, e.g. [AAD12; BDA14; RA08].

**simplicity** is related to the fact that the discovered process model should not be unnecessarily large in terms of model elements and should be easy to understand. Several measures have been proposed throughout the years to quantify simplicity, e.g. [BDA14; LA09].

Additional to these four quality dimensions of process discovery there are other desired properties of a discovered process model that are dependent on the process model notation that is used. An example of a desired property of process discovery results that is specific to workflow nets is soundness.

## 2.5 Conformance: Relating Event Logs and Process Models

*Conformance checking* is the task of relating an event log to a given process model with the aim to determine the degree to which the observed behavior conforms to the behavior that is specified by a process model. Alignments [AAD12] are a commonly used technique to map traces from event logs to firing sequences of a process model. Many of the existing measures for precision and fitness in process mining are based on alignments.

An **alignment** is a sequence of moves that relates a trace $\sigma$ from a labeled log to a Petri net, where each move in the sequence is one of the following types:

**Synchronous move** is a tuple $(e_i, t_i)$ where $e_i \in \sigma$ is an event from the trace and $t_i$ a transition in the Petri net. A synchronous move represents behavior that was both observed in the trace and allowed for by the process model.

**Log move** represents behavior that was observed in the trace but that was not allowed for by the process model. A log move is a tuple $(e_i, \gg)$ where $e_i \in \sigma$ an event from the trace and $\gg$ indicates that the behavior cannot be mimicked by the model.

**Model move** represents behavior that is performed by the model while no corresponding event was observed in the log. A model move is a tuple $(\gg, t_i)$ where $t_i$ is a transition of the process model and $\gg$ indicates that there is no corresponding event in the trace.

The alignment that we would ideally obtain minimize the number of log moves and model moves, as those moves represent deviations between the behavior that was observed in a trace and the behavior that is possible according to the model. The task of finding an alignment is obtained by formulating the problem as a shortest path optimization problem in which some *costs* are assigned to *log moves* and *model moves* while *[synchronous moves* are assigned zero cost. A search algorithm is then applied to find an alignment between log and model behavior that minimizes the costs. The result is generally referred to as an *optimal alignment* and it does not have to be unique. In practice, often the A* search algorithm [HNR68] is used to solve the optimization problem in a way that the globally optimal solution is guaranteed to be found.

Often, the costs for *model moves* are set differently for model moves on silent (unlabeled) transitions in comparison to model moves on labeled transitions. For silent transitions, it is impossible to map them to a synchronous move, since there is no matching event in the log. In some sense, a model move on a silent transition should not be seen as a real deviation between trace and model and should therefore not have any costs. In practice, model moves on silent transitions are assigned a very low but non-zero cost (called $\epsilon$-costs) in order to prevent us from finding optimal alignments that contain unnecessary loops of silent transitions. We will not go further into technical details regarding the construction of the search space of the shortest path problem and the computation of optimal alignments and instead refer the reader to [AAD12; ADA11].

| Log | a | a | ≫ | d | ≫ | e | f | ≫ | b | e | ≫ |
|---|---|---|---|---|---|---|---|---|---|---|---|
| Model | $t_1$<br>a | ≫ | $t_2$ | $t_5$<br>d | $t_3$<br>b | $t_6$<br>e | $t_9$<br>f | $t_2$ | $t_3$<br>b | $t_5$<br>e | $t_7$<br>g |

**Figure 2.4:** An example of an optimal alignment between labeled trace $\langle a, a, d, e, f, b, e \rangle$ and the accepting Petri net of Figure 2.1.

We denote an optimal alignment of trace $\sigma$ on model M with $\Omega_M(\sigma)$, i.e., the alignment with minimal cost, and $\overline{\Omega_M}(\sigma)$ denotes the worst possible alignment of $\sigma$ on M, i.e., the alignment with maximal cost. $\delta(\Omega_M(\sigma))$ and $\delta(\overline{\Omega_M}(\sigma))$ respectively denote the costs of the optimal and of the worst possible alignment of $\sigma$ on M.

Consider for example the accepting Petri net of Figure 2.1 and trace $\langle a, a, d, e, f, b, e \rangle$. Figure 2.4 shows an example of an optimal alignment between this trace and model. The trace starts with an event of activity a which the model can mimic from the initial marking $[p_1]$ by firing $t_1$, leading to marking $[p_2]$. The second event of the trace is another a event, which is impossible in the model, resulting in a log move on a (≫ in the model). The model then needs to fire $t_2$ from marking $p_2$ leading to marking $[p_3, p_4]$, but firing $t_2$ cannot be mimicked by the log since it is a silent ($\tau$) transition. Therefore, a model move $(\gg, t_2)$ is needed. Event d can be mimicked, but in order to mimic the e after that the model first needs either a b ($t_3$) or a c ($t_4$). The alignment that is shown performs a model move $(\gg, t_3)$, but note that alternatively a model move $(\gg, t_4)$ would have been equally optimal. This shows that there can exist more than one optimal alignment of a trace on a process model and often it is sufficient to obtain only one of the optimal alignments. e and f in the trace can be mimicked with synchronous moves by firing $t_6$ and $t_9$ and after that another model move is needed on silent ($\tau$) transition $t_2$. The b and e can be mimicked by firing $t_3$ and $t_5$ respectively, leading to marking $[p_7]$, from which a model move on either $t_7$ (g) or $t_8$ (h) is needed to end up in final marking $[p_8]$. The alignment of Figure 2.4 arbitrarily ended up with a model move on $t_7$ and not $t_8$.

## 2.6 Frequent Pattern Mining

**Frequent pattern mining** [Han+07] is the research field that focuses on extracting frequent substructures (i.e., patterns) from a data set. There are many types of frequent pattern mining that focus on different types of substructures and on different types of input data. One of the most basic forms of frequent pattern mining is *frequent itemset mining*.

### 2.6.1 Frequent itemset mining

**Frequent itemset mining** assumes as input a so called *transaction database*. A transaction database is a multiset of *transactions*, $D = [t_1, t_2, \ldots, t_n]$, in which each transaction (or *itemset*) $t_i = \{i_1, i_2 \ldots, i_m\}$ is a set of *items* that are included in that transaction, each set of items $i_j \in \Sigma$ comes from a fixed set of existing items $\Sigma$ (called *symbols*). The task of frequent itemset mining is to retrieve all sets of items $I \subseteq \Sigma$ that co-occur together in a given transaction database D more frequently than a given minimum number of times, called the *frequency threshold* (or called *support threshold*), *min_sup*. We start off by defining the support of an itemset $I \subseteq \Sigma$ in a transaction database.

**Definition 2.22 (Support).** Given a transaction database $D = [t_1, t_2, \ldots, t_n]$ and an itemset $I \subseteq \Sigma$, the support of I in D is $sup(D, I) = \sum_{t_i \in \{t_j | t_j \in D, I \subseteq t_j\}} D(t_i)$. ◇

Based on this definition of support, the task of frequent itemset mining is defined as follows.

**Definition 2.23 (Frequent itemset mining).** Given a transaction database $D = [t_1, t_2, \ldots, t_n]$ and a minimum support threshold $min_sup \in \mathbb{N}$, the task of frequent itemset mining is to retrieve the set $FIM(D, min_sup) = \{I \subseteq \Sigma \mid sup(D, I) \geq min_sup\}$. ◇

We will now show give an example of the frequent itemsets that satisfy $min_sup = 3$ in a transaction database of shopping transaction $D_1 = [\{\text{cigars}, \text{whiskey}\}^2, \{\text{beer}\}, \{\text{bacon}, \text{eggs}\}, \{\text{beer}, \text{cigars}, \text{whiskey}\}, \{\text{whiskey}\}]$. The *support* of itemset {cigars} is three, since there are three itemsets in $D_1$ that contain {cigars}: {cigars, whiskey} occurs twice as an itemset and {beer, cigars, whiskey} occurs once as an itemset. The support of itemset {wishkey} is four, since there are four itemsets in $D_1$ that contain {wishkey}: {cigars, whiskey} occurs twice as an itemset, {beer, cigars, whiskey} occurs once as an itemset, and {wishkey} occurs once as an itemset. Finally, the support of itemset {cigars, whiskey} is three, since there are three itemsets in $D_1$ that contain {cigars, whiskey}: {cigars, whiskey} occurs twice as an itemset and {beer, cigars, whiskey} occurs once as an itemset. None of the other subsets of Σ satisfy a minimum support of three in $D_1$, therefore, $FIM(D_1, 3) = \{\{\text{cigars}\}, \{\text{wishkey}\}, \{\text{cigars}, \text{whiskey}\}\}$.

Many algorithms have been developed that address the task of frequent itemset mining (see [Han+07] for an overview). Note that the output of frequent pattern mining is well-defined, therefore, the different algorithms for the task only differ from each other in aspects such as computational complexity and memory efficiency, but all provide the exact same result for identical transaction database and support threshold.

An property that plays an important role in frequent itemset mining algorithms to mine itemsets efficiently is the fact that for two itemsets $I_1$ and $I_2$ such that $I_1 \subset I_2 \subseteq \Sigma$ it is guaranteed to hold that $sup(D, I_1) \geq sup(D, I_2)$. This property is known as the *downward closure property*, or is sometimes alternatively named the *Apriori principle*, named after the first frequent pattern mining algorithm that leveraged this principle, the Apriori algorithm [AS94].

A frequent pattern mining task that is closely related to the task of frequent itemset mining but that is slightly more complex is the task of association rule mining.

### 2.6.2 Association rule mining

**Association rule mining**, like frequent itemset mining, takes a transaction database as input. The the task of association rule mining was originally defined by Agrawal et al. [Agr+93] as the task of mining a set of rules that each take the form:

$$I \implies a \tag{2.2}$$

In such a rule, $I \subseteq \Sigma$ is called the *antecedent* and $a \in \Sigma$ is called the *consequent*. Similar to frequent itemset mining, the rules that association rule mining aims to mine are such that itemset I satisfies some minimum support in transaction database D. Unlike frequent itemset mining, association rule mining additionally introduces a quality dimension that is called *confidence*, which is related to the likelihood that some transaction $t_i \in D$ additionally contains item a.

The *support* of an association rule $X \implies a$ in a transaction database D can be expressed in terms of itemset support:

$$sup(D, X \implies a) = sup(D, X \cup \{a\}) \tag{2.3}$$

Likewise, the *confidence* of an association rule $X \implies a$ in a transaction database D can be expressed in terms of itemset support in the following way:

$$conf(D, X \implies a) = \frac{sup(D, X \cup \{a\})}{sup(D, X)} \tag{2.4}$$

For example, in transaction database $D_1 = [\{\text{cigars}, \text{whiskey}\}^2, \{\text{beer}\}, \{\text{bacon}, \text{eggs}\}, \{\text{beer}, \text{cigars}, \text{whiskey}\}, \{\text{whiskey}\}]$, the association rule $\{\text{beer}, \text{whiskey}\} \implies \text{cigars}$ has a support of one (since it only occurs in transaction $\{\text{beer}, \text{cigars}, \text{whiskey}\}$) and has a confidence of one (since all $D_1$ does not contain any transaction that contains beer and whiskey but no cigars).

The task of association rule mining is to retrieve all rules from a given transaction database D that satisfy a given support threshold *min_sup* as well as a given confidence threshold *min_conf*.

**Definition 2.24 (Association rule mining).** Given a transaction database $D = [t_1, t_2, \dots, t_n]$ a minimum support threshold $min_sup \in \mathbb{N}$ and a minimum confidence threshold $min_conf \in [0, 1]$, the task of frequent itemset mining is to retrieve the set $ARM(D, min_sup, min_conf) = \{X \implies a \mid X \subseteq \Sigma, a \in \Sigma, sup(D, X \implies a) \geq min_sup \land conf(D, X \implies a) \geq min_conf\}$. ◇

Observe that the task of association rule mining is very closely link to the task of frequent itemset mining. In fact, lion's share of association rule mining algorithms take the approach of retrieving the set of frequent itemsets as a first mining step and then as a second mining step extract association rules from the obtained set of frequent itemsets.

In later work [AS94], the task of association rule mining has been generalized such that the consequent of an association rule is no longer limited to a single item, but can instead itself be an itemset, i.e., association rules now take the follow shape:

$$X \implies Y \tag{2.5}$$

with $X \subseteq \Sigma$ the *antecedent* itemset and $Y \subseteq \Sigma$ the *consequent* itemset. The term *association rule mining* in the pattern mining literature refers to both types of problem definitions.

### 2.6.3 Sequential pattern mining

**Sequential pattern mining** is a type of frequent pattern mining that takes similar type of input data as what is used in process discovery. In contrast to frequent itemset mining where the transactions are assumed to be unordered (and thus stored in a multiset of transactions), in sequential pattern mining some transactions are assumed to be ordered (and thus stored in a sequence of transactions).

The input data of sequential pattern mining is a so called a **sequence database** that takes the form $SD \in \mathcal{B}(\mathcal{P}(\Sigma)^*)$. An example of a sequence database is $SD_1 = [\langle\{cigars, whiskey\}, \{cheese, soap\}, \{grapefruit\}\rangle, \langle\{whiskey\}, \{grapefruit\}\rangle]$. Sequence database $SD_1$ represents a store that has two costumers: the first one bought cigars and whiskey in his first transaction at the store, bought soap and cheese in his second transaction, and finally bought a grapefruit in his third transaction. The second customer bought whiskey in his first transaction and grapefruit in his second.

A sequential pattern is a pattern of the form $\langle I_1, I_2, \dots, I_n\rangle$ with each $I_j \subset \Sigma$.

**Definition 2.25 (Sequence containment).** A sequential pattern $\langle I_1, I_2, \dots, I_n\rangle$ is *contained* in a sequence $\langle I'_1, I'_2, \dots, I'_m\rangle$ of transactions from a sequence database if there exists integers $i_1 < i_2 < \dots < i_n$ such that $I_1 \subseteq I'_{i_1}, I_2 \subseteq I'_{i_2}, \dots, I_n \subseteq I'_{i_n}$. We denote $\langle I_1, I_2, \dots, I_n\rangle \prec \langle I'_1, I'_2, \dots, I'_m\rangle$ if a sequential pattern $\langle I_1, I_2, \dots, I_n\rangle$ is contained in a sequence of itemsets $\langle I'_1, I'_2, \dots, I'_m\rangle$. ◇

For example, $\langle\{\text{cigars}\},\{\text{cheese},\text{soap}\},\{\text{grapefruit}\}\rangle \prec \langle\{\text{cigars},\text{whiskey}\},\{\text{milk}\},\{\text{cheese},\text{soap}\},\{\text{grapefruit}\}\rangle$ since $\{\text{cigars}\} \subseteq \{\text{cigars},\text{whiskey}\}$, $\{\text{cheese},\text{soap}\} \subseteq \{\text{cheese},\text{soap}\}$, and $\{\text{grapefruit}\} \subseteq \{\text{grapefruit}\}$. However, $\langle\{\text{cheese}\},\{\text{cigars}\}\rangle$ is not contained in $\langle\{\text{cheese},\text{cigars}\}\rangle$.

The support of a sequential pattern in a given sequence database *SD* is:

$$sup(SD,\langle I_1,\dots,I_n\rangle) = \sum_{\sigma \in \{\sigma' \in SD \mid \langle I_1,\dots,I_n\rangle \prec \sigma'\}} SD(\sigma) \tag{2.6}$$

Based on this definition of support we can define the task of sequential pattern mining as follows:

**Definition 2.26 (Sequential pattern mining).** Given a sequence database *SD* and a minimum support threshold $min_sup \in \mathbb{N}$, the task of sequential pattern mining is to retrieve the set $SPM(SD, min_sup) = \{\langle I_1, I_2, \dots, I_n\rangle \in \mathscr{P}(\Sigma)^* \mid sup(SD,\langle I_1, I_2, \dots, I_n\rangle) \geq min_sup\}$. ◇

Early influential work in the area of sequential pattern mining includes the work of Agrawal and Srikant [AS95], who have shown that the downward closure property for frequent itemset mining can be generalized to sequential patterns, leading to the following two guarantees:

1. $sup(SD,\langle I_1,\dots,I_i,\dots,I_n\rangle) \geq sup(SD,\langle I_1,\dots,I'_i,\dots,I_n\rangle)$ if $I_i \subseteq I'_i$, i.e., adding elements to one or more itemsets of a sequential pattern cannot increase its support.

2. $sup(SD,SP) \geq sup(SD,SP \cdot \langle I\rangle)$, i.e., a sequence cannot be more frequent than its subsequence.

In the same work, the AprioriSome and AprioriAll algorithms for sequential pattern mining have been introduced that leverage these downward closure properties for sequential pattern mining.

In some works from frequent pattern mining literature, the sequential pattern mining task that follows Definition 2.25 and Definition 2.26 is referred to as **gapped sequential pattern mining** instead of simply as sequential pattern mining. The word gapped is added to explicitly distinguish the task from earlier definitions of a related task (e.g., in [DM85]) where a sequential pattern is contained in a sequence of sequence database only if it occurs as a *consecutive* subsequence in a sequence, i.e., without the presence of gaps. Under this stricter definition of sequence containment (in comparison to to Definition 2.25), a sequential pattern $\langle I_1, I_2, \dots, I_n\rangle$ is *contained* in a sequence $\langle I'_1, I'_2, \dots, I'_m\rangle$ of transactions from a sequence database if there exists an integer $i_s$ such that $I_1 \subseteq I'_{i_s}, I_2 \subseteq I'_{i_s+1}, \dots, I_n \subseteq I'_{i_s+n}$. For example, $\langle\{\text{cigars}\},\{\text{cheese},\text{soap}\},\{\text{grapefruit}\}\rangle$ would under this alternative definition not be contained in $\langle\{\text{cigars},\text{whiskey}\},\{\text{milk}\},\{\text{cheese},\text{soap}\},\{\text{grapefruit}\}\rangle$ since there is no element in the sequential pattern that matches {milk}, while it is contained in it according to Definition 2.25.

### 2.6.4 Episode mining

**Episode mining** was initially introduced by Mannila et al. [MT95; MTV97] in 1995 and it is concerned with mining patterns from a single sequence consisting of pairs of event and timestamp. The timed sequence input data, called *event sequence* in episode mining terminology, takes the form $S = \langle (e_1, t_1), (e_2, t_2), \dots, (e_n, t_n) \rangle$ with each $e_i \in \Sigma$ and timestamps $t_i \in \mathbb{N}$ are represented by integers and are increasing, i.e., $\forall_{i \in \{1,\dots,n-1\}} t_i \leq t_{i+1}$.

The foundational work on episode mining [MT95] introduced a window-based approach to count the support of an episode and proposed an algorithm called WINEPI. A *window* on an event sequence $S = \langle (e_1, t_1), (e_2, t_2), \dots, (e_n, t_n) \rangle$ is an event sequence $W = \langle (e'_1, t'_1), (e'_2, t'_2), \dots, (e'_m, t'_m) \rangle$ such that $t_1 \leq t'_1$ and $t'_m \leq t_n$ and W consists of exactly those pairs $(e_i, t_i)$ from S where $t'_1 \leq t_i \leq t'_m$. The *width* of a window W is defined as $\mathit{width}(W) = t'_m - t'_1$.

The set of all windows W on S such that $\mathit{width}(W) = w$ for some $w \in \mathbb{N}$ is denoted $\mathit{aw}(S, w)$. Observe that $|\mathit{aw}(S, w)| = t_{|S|} - t_1 - w + 1$.

**Definition 2.27 (Episode).** An *episode* is a tuple $\phi = (V, \leq, g)$ consisting of a set of nodes V, $\leq$ a partial order relation on V, and $g : V \to \Sigma$ a labeling function that associates each node with an event type. An episode $\phi$ is called a *parallel episode* if $\forall_{x,y \in V} x \neq y \implies x \not\leq y$. An episode $\phi$ is called a *serial episode* if $\phi$ is a total order. Finally, an episode $\phi$ is called a *composite episode* if $\phi$ is neither a parallel not a serial episode. ◇

An episode $\phi = (V, \leq, g)$ is said to *occur* in an event sequence $S = \langle (e_1, t_1), (e_2, t_2), \dots, (e_n, t_n) \rangle$ if there exists an injective mapping $h : V \to \{1, \dots, n\}$ such that $\forall_{v \in V} g(v) = e_{h(v)}$ and $\forall_{v,w \in V} v \leq w \implies h(v) < h(w)$. We denote $\phi \models S$ if episode $\phi$ occurs in S.

The *support* of an episode $\phi$ in $S = \langle (e_1, t_1), (e_2, t_2), \dots, (e_n, t_n) \rangle$ is dependent on the choice of window size $w \in \mathbb{m}$ and is defined as follows:

$$\mathit{sup}(S, \phi, w) = \frac{|\{W \in \mathit{aw}(S, w) \mid \phi \models W\}|}{|\mathit{aw}(S, w)|} \tag{2.7}$$

The task of frequent episode mining is concerned with retrieving all frequent episodes from S that meet a given minimum support threshold *min_sup* given a window width w.

An alternative to the window-based support definition was later introduced by the same authors [MT96; MTV97] that is based on minimal occurrences, yielding the MINEPI algorithm. A minimal occurrence of an episode $\phi$ is a window W such that $\phi \models W$ but there does not exist a subwindow $W'$ of W such that $\phi \models W'$.

More recently, Leemans and van der Aalst [LA14] have extended the ideas behind episode mining from its initial definition on a single sequence to an event log consisting of multiple traces. In this variant to episode mining, windows are no longer extracted based on time information, but instead each trace from an event log is treated as a single window in which an episode $\phi$ either occurs or does not occur.

### 2.6.5 Closed patterns and maximal patterns

It is often a problem that frequent pattern mining techniques retrieve too many patterns to be interpreted by an analyst. This leads to an information overload to the analyst, who as a result does not anymore easily see the relevant patterns amongst the other patterns. There are multiple approaches to address this issue. Two of those approaches that are commonly used are the mining of *closed patterns* and the mining of *maximal patterns*. Both *closed* and *maximal* patterns are concepts that are independent of the type of patterns that are being mined, i.e., these ideas can in principle be applied in combination with any type of frequent pattern mining. See for example, [Pas+99] for mining closed frequent itemsets, [BCG01; GZ01; GZ05] for mining maximal frequent itemsets, [YHA03a] for mining closed sequential patterns, [Fou+14b; FWT13] for mining maximal sequential patterns, and [ZLC10] for mining closed episodes.

#### *Closed pattern mining*

A pattern P is considered to be a **closed pattern** if there does not exist a superpattern P′ of P such that the support of P′ is identical to the support of P. In the context of frequent itemset mining, a closed set of frequent itemsets *CFIM* is a subset of the set of frequent itemsets $FIM(D, min_sup)$ if $\forall_{I \in CFIM} \nexists_{I' \in CFIM} I \subset I' \land sup(D, I) = sup(D, I')$.

For example, if $I = \{a, b, c\}$ is a frequent itemset that meets the support threshold, then I is closed only if for all $x \in \Sigma$ it holds that $I' = \{a, b, c, x\}$ has *strictly lower support* in transaction database D than the support of I. The idea behind closed patterns is that if frequent itemset I′ already has a support of at least *min_sup*, then it is already implied by the downward closure property that $I \subset I'$ also has a support of at least *min_sup*. Therefore, itemset I is considered to provide the analyst with additional information over itemset I′ only if it has higher support.

In the case of sequential pattern mining, a sequential pattern $SP = \langle\{a\}, \{b\}, \{c\}\rangle$ with support k is a *closed sequential pattern* if and only if its support is at least *min_sup* and for all $x \in \Sigma$ the sequential pattern $SP' = \langle\{a\}, \{b\}, \{c\}, \{x\}\rangle$ has lower support frequently in the transaction database than $\langle a, b, c\rangle$.

It is easy to see that a set of closed patterns can be obtained by first simply mining the set of all patterns that satisfy *min_sup* and then filtering out those that are non-closed. However, more efficient algorithms direct algorithms exist that are able to mine the set of closed patterns without first generating the full set of frequent patterns [Pas+99; YHA03a; ZLC10]. Furthermore, it has been shown that the full set of frequent patterns that meet a support threshold *min_sup* can be reconstructed from the set of closed patterns. Therefore, the set of closed patterns is considered to a lossless compression of the set of frequent patterns.

### *Maximal pattern mining*

A pattern P is considered to be a **maximal pattern** if there does not exist a superpattern P′ of P such that the support of P′ meets the support threshold *min_sup*. In the context of frequent itemset mining, a frequent itemset I with support $sup(D, I) \geq min_sup$ is a *maximal itemset* if and only if for all $x \in \Sigma$, itemset $sup(D, I \cup \{x\}) < min_sup$. Maximal patterns are defined analogously for sequential patterns.

Trivially, maximal pattern mining reduces the set of patterns to a higher degree than closed pattern mining does. However, in contrast to the set of closed patterns, it is not always possible to reconstruct the full set of frequent patterns that meet a support threshold *min_sup* from the set of maximal patterns. Therefore, the set of maximal patterns is considered to a lossy compression of the set of frequent patterns.

Like with the set of closed patterns, it is easy to see that a set of maximal patterns can be obtained by first simply mining the set of all patterns that satisfy *min_sup* and then filtering out those that are non-maximal. Like for closed patterns, more efficient algorithms direct algorithms exist that are able to mine the set of maximal patterns without first generating the full set of frequent patterns [BCG01; Fou+14b; FWT13; GZ01; GZ05].

# Part I

# Pre-processing of Event Data

**Chapter 3** We analyze measures to express how well a process model represents the behavior seen in an event log. This chapter is based on the publication [Tax+18c].

**Chapter 4** We provide a method to assess whether some proposed relabeling of an event log into more fine-grained or more coarse-grained label would lead to a more structured event log. This chapter is based on the publication [Tax+16d].

**Chapter 5** We present a method to bring information from event data attributes into the event labels, thereby relabeling events into more fine-grained events, resulting in a more structured event log. This chapter is based on the publications [Tax+16a; Tax+18a].

**Chapter 6** We provide a technique to detect and filter out chaotic activities from event logs, thereby creating more structure in the resulting event log and enabling the discovery of better process models. This chapter is based on the publication [TSA18].

**Chapter 7** We provide a technique to use supervised learning to abstract low-level events into higher level events, thereby creating a more structured event log on a coarser level of granularity. This chapter is based on the publications [Tax+16c; Tax+18f].

# 3 On the Quality of Process Models with Respect to Event Logs

**Parts of this chapter have been published as:**

- Niek Tax, Xixi Lu, Natalia Sidorova, Dirk Fahland, and Wil M. P. van der Aalst: *The imprecisions of precision measures in process mining*. Information Processing Letters 135: pp. 1–8 (2018)

In this first part of the thesis that focuses on the preprocessing of event logs with the aim to enable or aid the mining of insights, we start with an analysis of the success criteria of mining insights from event data. Recall from Section 2.4 that the process model that is pursued by process discovery algorithms ideally allows for all the behavior that was observed in the event log (called *fitness*), while at the same time it should not be too general by allowing for much more behavior than what was seen in the event log (called *precision*). In this chapter, we assess whether existing measures for fitness and for precision correspond to the conceptual intuitions behind fitness and precision. Based on this assessment we will draw conclusions on how the quality of process models that are discovered from event logs should be assessed. Throughout this chapter, we will focus solely on labeled event logs (as defined in Chapter 2), i.e., we leave event logs where events consist of multiple event attributes out of scope.

Fitness, conceptually, should be high when all behavior of the event log fits the process model and low when no behavior of the event log fits the process model. An example of a simplistic fitness measure is the *ratio of fitting traces* (rft), i.e., the ratio of the traces of an event log that completely fit the process model. More formally, for a given process model $M \in \mathcal{M}$ and a log L, $rft(L, M) = \frac{\sum_{\sigma \in (\tilde{L} \cap \mathfrak{L}(M))} L(\sigma)}{|L|}$. The first mention of this measure in literature is by Greco et al. [Gre+04; Gre+06b], who refers to this measure as *completeness*. While rft-fitness measures fitness on a trace level, several more recent measures aim to take a step beyond the trace level by measuring it on the event level. The rationale behind measuring fitness on the event level is based on the reasoning that fitness, for traces that do not fit the model completely, should still quantify the degree to which the trace fits or does not fit the process model, and should yield a higher fitness value when the unfitting traces of an event log L are *close to* the behavior that is allowed by a process model. More formally, if for a trace $\sigma \in L$ that is unfitting on a process model $M \in \mathcal{M}$ (i.e., $\sigma \notin \mathfrak{L}(M)$) there exists a somehow similar trace of the process model $\sigma' \in \mathfrak{L}(M)$ such that $\sigma' \approx \sigma$, then the fitness of trace $\sigma$ should, according to the principles of event-level fitness, be higher compared to the situation when $\sigma$ is *nothing like* the behavior of the process model. However, the way in which similarity between traces (represented by the $\approx$ symbol) is measured differs between event-level fitness measures and there seems to be little discussion within the research community on the properties that are to be desired from such a similarity measure. Two fitness measures that measure fitness on the event level are *token-based fitness* [RA08] and *alignment-based fitness* [ADA11]. In Section 3.1 we formulate axioms for fitness measures, and in Section 3.2 we describe existing fitness measures in detail, and we validate whether they fulfill these axioms.

In contrast to the fitness dimension, for which only a handful of measures have been developed throughout the years, a wide range of measures have been proposed for precision [Bro+14; DCC16; Gre+06a; LFA16; Muñ+10; RA08]. To the best of our knowledge, there is currently no work on verifying whether precision measures actually quantify what they are aiming to quantify in a consistent manner. Conceptually, the precision of a process model in the context of an event log should be high when the model allows for only a few traces that were not seen in the log, and it should be low when it allows for many traces that were not seen in the log. In this chapter, we propose a set of axioms that formulate desired properties of precision measures in a more formal way and use this to systematically validate whether existing precision measures are in agreement with conceptual beliefs about precision.

After analyzing fitness measures in Section 3.1 and Section 3.2, we continue by analyzing precision measures. First, we introduce axioms for precision measures in Section 3.3 and in Section 3.4 we describe existing precision measures and analyze whether these are in concordance with the axioms. In Section 3.5 we describe two situations in which we believe that there is currently no agreement in the research field on what we expect from a precision measure. In Section 3.6 we conclude this chapter, thereby linking the insights on fitness and precision measures from this chapter to guidelines for the evaluation of mining results for the succeeding chapters of this thesis.

## 3.1 Axioms for Fitness Measures

We consider fitness to be a function $\mathit{fit}(L, M)$ that quantifies what parts of the event log L are possible in process model M. Below, we formalize desired properties of function *fit* that we believe should consistently hold for any model and any event log.

The axioms that we formulate partly overlap with the propositions for fitness measures that were recently listed by van der Aalst [Aal18]. However, some of the axioms that we state here are more specific while at the same time they are more strict. Additionally, we present a smaller set of requirements for fitness measures than van der Aalst [Aal18], as we show that several of van der Aalst's propositions are implied by others. Here, we distinguish between the core requirements for fitness measures, which we refer to as *axioms*, and the requirements that logically follow from the core requirements, which we refer to as *corollaries*.

The first axiom states that fitness is deterministic, i.e., for a given log and model it always returns the same result.

**Axiom A1.** A fitness measure is deterministic, i.e., it is a *function* $fit : \mathcal{B}(\Sigma^+) \times \mathcal{M} \rightarrow \mathbb{R}$. ◇

Existing fitness measures normalize $\mathbb{R}$ to a $[0,1]$-interval. This axiom overlaps with proposition *DetPro*$^+$ by van der Aalst [Aal18].

The second axiom formulates that for a given log L and two models $M_1$ and $M_2$, the fitness of $M_2$ on L should be at least as high as the fitness of $M_1$ on L when the part of $M_2$ that was seen in L is larger than the part of $M_1$ that was seen in L.

**Axiom A2.** For two models $M_1$ and $M_2$ and a log $\mathfrak{L}(M_1) \cap \tilde{L} \subseteq \mathfrak{L}(M_2) \cap \tilde{L} \implies fit(L, M_1) \leq fit(L, M_2)$. ◇

As a corollary, it follows that for a given log L and two models $M_1$ and $M_2$, the fitness of $M_2$ on L should be at least as high as the fitness of $M_1$ on L when the part of $M_2$ allows for more behavior than $M_1$.

**Corollary 3.1.** For two models $M_1$ and $M_2$ and a log L, $\mathfrak{L}(M_1) \subseteq \mathfrak{L}(M_2) \implies fit(L, M_1) \leq fit(L, M_2)$. ◇

It is easy to see that this follows as a corollary, as it is the special case of **A2** where $L \in \mathcal{B}(\Sigma^+)$ is chosen such that $\tilde{L} = \mathfrak{L}(M_1) \cup \mathfrak{L}(M_2)$. Corollary 3.1 overlaps with proposition *RecPro1*$^+$ of van der Aalst [Aal18]. However, note that **A2** puts a stricter requirement on fitness measures than *RecPro1*$^+$, as it additionally applies to a model $M_1$ that allows for traces that are not allowed for by $M_2$, as long as those traces are not part of log L.

Note that when $\mathfrak{L}(M_1) = \mathfrak{L}(M_2)$, it also holds that $\mathfrak{L}(M_1) \subseteq \mathfrak{L}(M_2)$, and therefore $fit(L, M_1) \geq fit(L, M_2)$. However, $\mathfrak{L}(M_1) = \mathfrak{L}(M_2)$ also implies $\mathfrak{L}(M_2) \subseteq \mathfrak{L}(M_1)$ and therefore $fit(L, M_2) \geq fit(L, M_1)$. Combining these two observations, we can formulate the following corollary that follows from **A2**, which overlaps with proposition *BehPro*$^+$ of van der Aalst [Aal18].

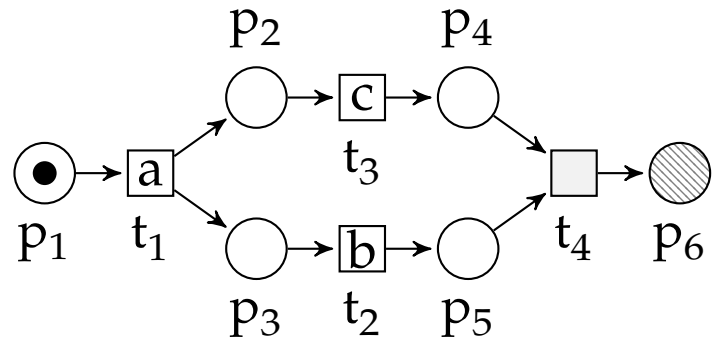

**Figure 3.1:** Model M with $\mathfrak{L}(M) = \{\langle a,b,c\rangle, \langle a,c,b\rangle\}$.

**Corollary 3.2.** For two models $M_1$ and $M_2$ and a log L, $\mathfrak{L}(M_1) = \mathfrak{L}(M_2) \implies \mathit{fit}(L,M_1) = \mathit{fit}(L,M_2)$, i.e., fitness depends only on the behavior of a model and not on its structure. ◇

It is easy to see that **A2** is a reasonable property to expect from a fitness measure. The behavior of L that process model $M_2$ allows for in addition to $M_1$, i.e., $(\mathfrak{L}(M_2) \cap \tilde{L}) \setminus \mathfrak{L}(M_1)$, might be non-empty, in which case the fitness of $M_2$ should be strictly higher than that of $M_1$. Alternatively, this set could be empty, in which case we would desire $\mathit{fit}(L,M_1) = \mathit{fit}(L,M_2)$, since the additional behavior of $M_2$ does not allow us to explain more log behavior. We now formulate a third axiom, a stricter version of **A2**.

**Axiom A3.** For two models $M_1$ and $M_2$ and a log L, $\mathfrak{L}(M_1) \cap \tilde{L} \subset \mathfrak{L}(M_2) \cap \tilde{L} \implies \mathit{fit}(L,M_1) < \mathit{fit}(L,M_2)$. ◇

As a corollary, it follows that if a model $M_2$ allows for more behavior than another model $M_1$, and there exists behavior in log L that is allowed for by $M_2$ but not by $M_1$, i.e., when $((\mathfrak{L}(M_2) \setminus \mathfrak{L}(M_1)) \cap \tilde{L}) \neq \emptyset$, then the fitness of $M_2$ on L should be higher than that of $M_1$ on L.

**Corollary 3.3.** For two models $M_1$ and $M_2$ and a log L, $\mathfrak{L}(M_1) \subseteq \mathfrak{L}(M_2) \wedge ((\mathfrak{L}(M_2) \setminus \mathfrak{L}(M_1)) \cap \tilde{L}) \neq \emptyset \implies \mathit{fit}(L,M_1) < \mathit{fit}(L,M_2)$. ◇

The fourth axiom states that adding fitting behavior to an event log increases fitness when the event log did not already completely consist of fitting traces.

**Axiom A4.** For event logs $L_1$ and $L_2$ and process model M, $\tilde{L}_1 \not\subseteq \mathfrak{L}(M) \wedge |L_2| > 0 \wedge \tilde{L}_2 \subseteq \mathfrak{L}(M) \implies \mathit{fit}(L_1 \uplus L_2, M) > \mathit{fit}(L_1, M)$ ◇

Note that a consequence of **A4** is that fitness is sensitive to the multiplicities of the traces in the log. To see that this is the case, consider the process model M of Figure 3.1 and two logs: $L_1 = [\langle a,b,c\rangle^{100}, \langle a,c,b\rangle^{100}, \langle a,d,e\rangle, \langle a,e,d\rangle]$ and $L_2 = [\langle a,b,c\rangle^2, \langle a,c,b\rangle^2, \langle a,d,e\rangle, \langle a,e,d\rangle]$. Since for $L_3 = [\langle a,b,c\rangle^{98}, \langle a,c,b\rangle^{98}]$ we have $L_1 = L_2 \uplus L_3$ with $\tilde{L}_3 \subseteq \mathfrak{L}(M)$ and $|L_3| > 0$, **A4** enforces that the fitness of M on $L_1$, where there are 2 non-fitting traces and 200 fitting traces, is higher than the fitness of M on $L_2$, where there are 2 non-fitting traces and 4 fitting traces, even though $\tilde{L}_1 = \tilde{L}_2$.

3

As the fifth axiom, we state that a given process model M yields an identical fitness value on all logs that completely fit M. This axiom overlaps with proposition *RecPro5*$^+$ of van der Aalst [Aal18].

**Axiom A5.** For two logs $L_1$ and $L_2$ and a process model M, $\tilde{L}_1 \subseteq \mathfrak{L}(M) \wedge \tilde{L}_2 \subseteq \mathfrak{L}(M) \implies \mathit{fit}(L_1, M) = \mathit{fit}(L_2, M)$ ◇

It follows from **A4** and **A5** that adding fitting behavior to an event log can never decrease fitness. This corollary overlaps with proposition *RecPro2*$^+$ of van der Aalst [Aal18].

**Corollary 3.4.** For event logs $L_1$, $L_2$ and process model M, $\tilde{L}_2 \subseteq \mathfrak{L}(M) \implies \mathit{fit}(L_1 \uplus L_2, M) \geq \mathit{fit}(L_1, M)$. ◇

It is easy to see that this corollary follows from **A4** and **A5**. Either $\tilde{L}_1 \not\subseteq \mathfrak{L}(M)$, or $\tilde{L}_1 \subseteq \mathfrak{L}(M)$. When $\tilde{L}_1 \not\subseteq \mathfrak{L}(M)$, then $\mathit{fit}(L_1 \uplus L_2, M) > \mathit{fit}(L_1, M)$ according to **A4**. Alternatively, when $\tilde{L}_1 \subseteq \mathfrak{L}(M)$, we know that $(L_1 \uplus L_2) \subseteq \mathfrak{L}(M)$ since $\tilde{L}_2 \subseteq \mathfrak{L}(M)$ is required by the corollary. Since $(L_1 \uplus L_2) \subseteq \mathfrak{L}(M)$ and $\tilde{L}_1 \subseteq \mathfrak{L}(M)$, **A5** requires $\mathit{fit}(L_1 \uplus L_2, M) = \mathit{fit}(L_1, M)$.

Where **A5** states that fitness is identical for all *completely fitting* logs, fitness should also be identical for all *completely non-fitting logs*. We formalize this in the sixth axiom.

**Axiom A6.** For two logs $L_1$ and $L_2$ and a process model M, $(\tilde{L}_1 \cap \mathfrak{L}(M)) = \emptyset \wedge (\tilde{L}_2 \cap \mathfrak{L}(M)) = \emptyset \implies \mathit{fit}(L_1, M) = \mathit{fit}(L_2, M)$ ◇

Adding non-fitting behavior to a log that is not already completely non-fitting should decrease fitness. This is captured in axiom **A7**.

**Axiom A7.** For event logs $L_1$ and $L_2$ and process model M, $(\tilde{L}_1 \cap \mathfrak{L}(M)) \neq \emptyset \wedge |L_2| > 0 \wedge ((\tilde{L}_2 \cap \mathfrak{L}(M)) = \emptyset) \implies \mathit{fit}(L_1 \uplus L_2, M) < \mathit{fit}(L_1, M)$. ◇

From **A6** and **A7** it follows that adding non-fitting traces to a log can never increase fitness. This corollary overlaps with proposition *RecPro3*$^+$ of van der Aalst [Aal18]. However, **A7** is more strict than *RecPro3*$^+$, as *RecPro3*$^+$ only requires $\mathit{fit}(L_1 \uplus L_2, M) \leq \mathit{fit}(L_1, M)$, and does not require it to be strictly lower.

**Corollary 3.5.** For event logs $L_1, L_2$ and process model M, $|L_2| > 0 \wedge (\tilde{L}_2 \cap \mathfrak{L}(M)) = \emptyset \implies \mathit{fit}(L_1 \uplus L_2, M) \leq \mathit{fit}(L_1, M)$. ◇

It is easy to see that this corollary follows from **A6** and **A7**. Trivially, either $(\tilde{L}_1 \cap \mathfrak{L}(M)) = \emptyset$, or $(\tilde{L}_1 \cap \mathfrak{L}(M)) \neq \emptyset$, has to hold. When $(\tilde{L}_1 \cap \mathfrak{L}(M)) \neq \emptyset$, then $\mathit{fit}(L_1 \uplus L_2, M) < \mathit{fit}(L_1, M)$ according to **A7**. Alternatively, when $(\tilde{L}_1 \cap \mathfrak{L}(M)) = \emptyset$, then $\mathit{fit}(L_1 \uplus L_2, M) = \mathit{fit}(L_1, M)$ according to **A6**.

It is important to note that while **A6**, **A7**, Corollary 3.5, and van der Aalst's proposition *RecPro3*$^+$ [Aal18] are all consistent with the intuitions behind trace-level fitness, they in fact directly contradict the conceptual ideas behind event-level fitness. For **A6** it is easy to see that this is the case: the main idea behind event-level fitness is that it quantifies the *degree* to which traces are fitting. Therefore, the fitness of a log $L = [\sigma]$ with $\sigma \notin \mathfrak{L}(M)$ depends on *how similar* $\sigma$ is to the behavior that is allowed by M. This directly contradicts **A6**, which states that all logs without a fitting trace should yield *the same* fitness value, thereby not leaving any space for the fitness measure to distinguish between *how similar* the non-fitting traces of a log are to the modeled behavior. To see that **A7**, Corollary 3.5, and van der Aalst's proposition *RecPro3*$^+$ [Aal18] contradict the conceptual notion of event-level fitness, consider that the non-fitting traces in $L_1$ could be very dissimilar from the model behavior while the later added traces of $L_2$ could be non-fitting but highly similar to behavior of the model. In that scenario, it would not be unthinkable for an event-level fitness measure to yield a higher fitness value for model M on combined log $L_1 \uplus L_2$ than on $L_1$, even though the added behavior $L_2$ is non-fitting.

In order to formulate event-level alternatives to **A6** and **A7** we would need to define a function $\mathit{sim} : \Sigma^+ \times \mathcal{M} \to [0, 1]$ that *quantifies* how close the behavior of a trace $\sigma \in \Sigma^+$ is to the behavior of a model $M \in \mathcal{M}$. The properties that are desired in such a function *sim* are so far largely left undiscussed in the process mining research field. However, it seems to be generally expected that *sim* should have the property that $\mathit{sim}(\sigma, M) = 1$ when $\sigma \in \mathfrak{L}(M)$ and that $\mathit{sim}(\sigma, M) < 1$ when $\sigma \notin \mathfrak{L}(M)$.

Using this generic notion of similarity *sim*, we can formulate the axiom that a model M has identical fitness on two logs $L_1$ and $L_2$ if both logs consist of traces that are equally similar to the behavior of M, in terms of *sim*. This axiom can be seen as an adaptation of **A6** that matches the conceptual notion of event-level fitness. $\mathit{sim}_L$ lifts function *sim* to the level of an event log and thereby yields a multiset of similarity scores.

**Axiom A8.** For a similarity function *sim*, two logs $L_1$ and $L_2$ and a process model M, $\mathit{sim}_L(\mathfrak{L}(M_1)) = \mathit{sim}_L(\mathfrak{L}(M_2)) \implies \mathit{fit}(L_1, M) = \mathit{fit}(L_2, M)$.

Furthermore, we can formulate the axiom stating that adding non-fitting traces to a log can never increase fitness if the added traces are less similar to the model behavior than the traces that are already present in the log. This axiom can be seen as an adaptation of **A7** that matches the conceptual notion of event-level fitness.

**Axiom A9.** For a similarity function *sim*, two logs $L_1$ and $L_2$ and a process model M, $\forall_{\sigma_1 \in \tilde{L}_1} \forall_{\sigma_2 \in \tilde{L}_2} sim(\sigma_1, M) > sim(\sigma_2, M) \implies fit(L_1 \uplus L_2, M) < fit(L_1, M)$. ◇

## 3.2 Fitness Measures

In this section we give an overview of the fitness measures and we validate the axioms for fitness measures introduced in Section 3.1 for each measure.

### 3.2.1 Token-based Fitness

Token-based replay [RA08] attempts to replay each trace of the log in the model in a log-focused manner. For each trace in the log, starting from the initial marking $m_0$, token-based replay iterates over the trace and for each event it attempts to fire the corresponding transition of which the label matches the events. When the trace contains an event that corresponds to a transition t that is not enabled from the marking m that the model resides in at that point in time, then the tokens that are needed to enable t are artificially created in order to enable it, and replay is continued. The artificial tokens that are needed to enable t when it is not enabled from m are equal to $\bullet t - m$. The number of artificially created tokens is counted while replaying the log and the number of artificial tokens needed to replay the complete log is referred to as the *number of missing tokens mt*. After reaching the end of a trace, the number of tokens is counted that reside in places that do not belong to the final marking $m_f$, and this is called the *number of remaining tokens rt*. Additionally, the *number of produced tokens* (*pt*) and the *number of consumed tokens* (*ct*) are counted, i.e., the number of tokens that are created and respectively removed by firing the transitions during log replay. *Token-based fitness* [RA08] defines fitness based on token-based replay, as a function over the number of *pt*, *ct*, *mt*, and *rt*:

$$tbf(L, M) = 1 - \frac{\frac{mt}{ct} + \frac{rt}{pt}}{2}$$

Note that missing tokens are by definition also included in the count of consumed tokens, as tokens are only added artificially if they are needed to fire a transition that matches the next event from the log. Therefore, $mt \leq ct$. Likewise, remaining tokens have to be produced at some point, therefore, $rt \leq pt$. Because $mt \leq ct$ and $rt \leq pt$, both ratios $\frac{mt}{ct}$ and $\frac{rt}{pt}$ yield a value in $[0, 1]$, and as an effect token-based fitness yields a value in $[0, 1]$. Since token-based replay is deterministic, token-based fitness is a deterministic measure. However, the token-based replay procedure described in [RA08] is defined only for traces $\sigma$ in which each event $e \in \sigma$ has a corresponding transition t in the model such that $\ell(t) = e$. The paper in which token-based replay was introduced states "replay is carried out in a non-blocking way and from a log-based perspective, i.e., for each log event in the trace the corresponding transition is fired, regardless of whether the current path of the model is followed or not."(from [RA08]) without specifying how to process events that do not have a corresponding transition in the model. Therefore, token-based fitness can be seen as a partial function $\mathcal{B}(\Sigma^+) \times \mathcal{M} \nrightarrow \mathbb{R}$ that is not properly defined for any arbitrary log, and **A1** does not hold.

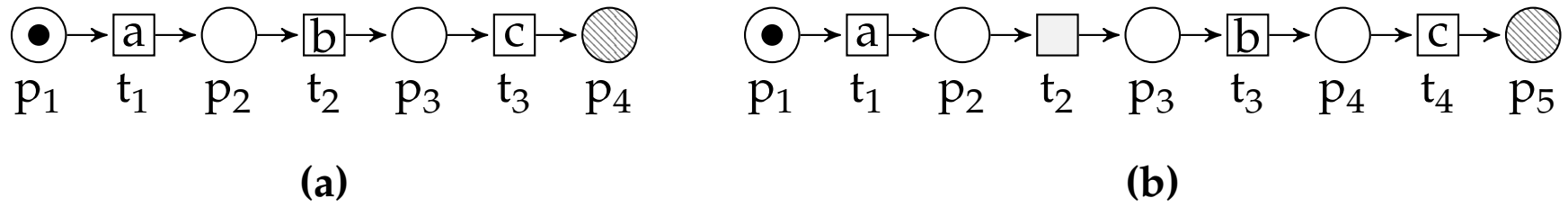

**Figure 3.2:** Two structurally different process models with language $\{\langle a, b, c\rangle\}$: *(a)* process model $M_1$ and *(b)* process model $M_2$, which contains an additional silent transition.

Figure 3.2 shows two process models $M_1$ and $M_2$ that both allow for a single trace: $\mathfrak{L}(M_1) = \mathfrak{L}(M_2) = \{\langle a, b, c\rangle\}$. Consider log $L = [\langle a, b\rangle]$, which does not fit on either process model. In order to replay the single trace of L on $M_1$, first, $t_1$ is fired to mimic the a in the trace and then $t_2$ is fired to mimic the b in the trace. After replaying a and b the model is not yet in the final marking, as it would need to fire $t_3$ to reach that. Therefore, the replay finishes with a remaining token in $p_3$. The firing of $t_1$ consumes one token and produces one token, and the firing of $t_2$ produces and consumes another token. The token in the initial marking is also counted as a produced token. Finally, the replay aims to consume a token from the final marking. However, there is no token in the final marking, therefore this token needs to be artificially produced (i.e., it is registered as a missing token). In total, $mt = 1$, $rt = 1$, $pt = 3$, $ct = 3$, and $tbf(L, M_1) = 1 - \frac{\frac{1}{3}+\frac{1}{3}}{2} = \frac{2}{3}$. When we replay the same trace on $M_2$, we have to fire silent transition $t_2$ in-between firing the transitions that correspond to events a and b from the log. Firing $t_2$ results in one consumed and one produced token. The count of remaining tokens does not change with respect to $M_1$, as the replay on $M_2$ also finishes with one remaining token, in $p_4$. The additional silent transition of $M_2$ results in $mt = 1$, $rt = 1$, $pt = 4$, $ct = 4$, and therefore the fitness changes to $tbf(L, M_2) = 1 - \frac{\frac{1}{4}+\frac{1}{4}}{2} = \frac{3}{4}$. The fact that $tbf(L, M_1) \neq tbf(L, M_2)$ while $\mathfrak{L}(M_1) = \mathfrak{L}(M_2)$ shows that token-based fitness is not only dependent on the *behavior* of a model, but also on its *structure*. Moreover, since $\mathfrak{L}(M_1) = \mathfrak{L}(M_2)$ implies $\mathfrak{L}(M_2) \cap \tilde{L} \subseteq \mathfrak{L}(M_1) \cap \tilde{L}$ but $tbf(L, M_2) > tbf(L, M_1)$, axiom **A2** does not hold. We have just shown that token-based fitness can be influenced by adding *one* silent transition to the model. In the limit, by creating a sequence consisting of *multiple* silent transitions, one can approach a perfect fitness score for an unfitting log. To see that this is the case, consider that each added silent transition to the chain of silent transitions results in one additional produced token and one additional consumed token. Therefore, the fitness of process model $M_n$ that consists of a chain of n silent transitions between the transitions that are labeled a and b (as shown in Figure 3.3a) on non-fitting log $L = [\langle a, b\rangle]$ results in a fitness value of $tbf(L, M_n) = 1 - \frac{\frac{1}{3+n} + \frac{1}{3+n}}{2}$, and in the limit, $\lim_{n\to\infty} tbf(L, M_n) = 1 - \frac{\frac{1}{3+n} + \frac{1}{3+n}}{2} = 1$.

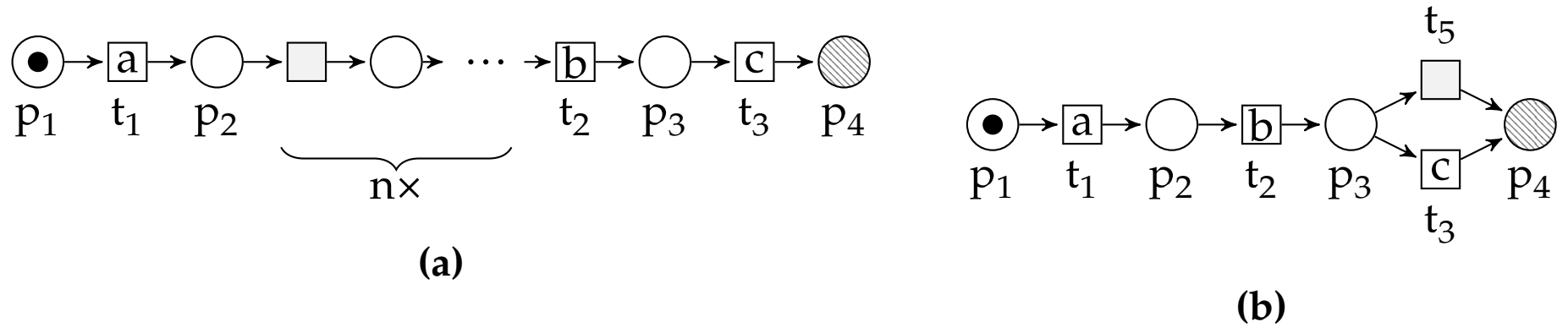

**Figure 3.3:** Two process models: *(a)* process model $M_1$ with language $\mathfrak{L}(M_1) = \{\langle a, b, c\rangle\}$, and *(b)* process model $M_2$ with language $\mathfrak{L}(M_2) = \{\langle a, b\rangle, \langle a, b, c\rangle\}$.

By making use of the behavior of token-based fitness in the limit, in the presence of chains of silent transitions, we will now show that **A3** does not hold for token-based fitness. Figure 3.3 shows process models $M_1$ and $M_2$ such that $\mathfrak{L}(M_1) \subseteq \mathfrak{L}(M_2)$. Now consider log $L = [\langle a\rangle, \langle a, b\rangle]$. Note that $(\mathfrak{L}(M_2) \setminus \mathfrak{L}(M_1)) \cap \tilde{L} = \{\langle a, b\rangle\}$, i.e., there is overlap between log L and the behavior that $M_2$ allows for additionally with respect to $M_1$. Replaying trace $\langle a\rangle$, which is non-fitting on both models, results on both $M_1$ and $M_2$ in one produced token from the initial marking and one from firing $t_1$, in one consumed tokens from firing $t_1$, and a remaining token in place $p_2$. In order to finally consume a token from the final marking, a missing token is produced in $p_4$. In total, replaying this trace on both models results in $mt = 1$, $rt = 1$, $pt = 2$, $ct = 2$. Replaying trace $\langle a, b\rangle$ on model $M_2$, on which it fits, results in one token produced in the initial marking, one token consumed and one produced by firing $t_1$, one token consumed and one produced by firing $t_2$, and one token consumed and one produced by firing $t_5$. The final marking is now reached without tokens in places other than $m_f$, so there are no remaining tokens and the token from the final marking can be consumed without generating a missing token. In total, replaying this trace on $M_2$ results in $mt = 0$, $rt = 0$, $pt = 4$, $ct = 4$. Adding the missing, remaining, produced, and consumed tokens for $M_2$ from both traces in L, $mt = 1$, $rt = 1$, $pt = 6$, $ct = 6$, and $tbf(L, M_2) = 1 - \frac{\frac{1}{6} + \frac{1}{6}}{2} = \frac{5}{6} \approx 0.917$. Replaying trace $\langle a, b\rangle$ on model $M_1$ results in one produced token from the initial marking, one consumed and one produced token from firing $t_1$, one consumed and one produced token for *each* of the n silent transitions, one consumed an one produced token from firing $t_2$, and finally a remaining token in $p_3$ and a missing token in the place of the final marking that is immediately consumed. Adding the missing, remaining, produced, and consumed tokens for $M_1$ from both traces in L, $mt = 1$, $rt = 2$, $pt = 5 + n$, $ct = 4 + n$, and $tbf(L, M_2) = 1 - \frac{\frac{1}{4+n} + \frac{2}{5+n}}{2}$. Observe that when $n \geq 5$, $1 - \frac{\frac{1}{4+n}+\frac{2}{5+n}}{2} \geq \frac{5}{6}$, and therefore $tbf(\mathrm{L}, \mathrm{M}_1) \geq tbf(\mathrm{L}, \mathrm{M}_2)$, while $\mathfrak{L}(\mathrm{M}_1) \subseteq \mathfrak{L}(\mathrm{M}_2)$ and $(\mathfrak{L}(\mathrm{M}_2) \setminus \mathfrak{L}(\mathrm{M}_1)) \cap \tilde{\mathrm{L}} \neq \emptyset$, showing that **A3** does not hold for token-based fitness.

Two non-trivial problems during token-based replay are, 1) whether a specific transition can be enabled via firing a sequence of invisible transitions, and 2) the decision on which transition to fire in the case of multiple identically labeled transitions. With a full state space exploration, these problems can be addressed in such a way that it is guaranteed that a fitting trace can be replayed without needing any missing or remaining tokens. However, Rozinat and van der Aalst [RA08] for reasons related to computational efficiency propose to implement the replay procedure in a heuristic way using a local search procedure and state themselves that "unfortunately, this means that it cannot be guaranteed that if the log fits the model it can be replayed correctly (and thus any mismatch really indicates a conformance problem)" (from [RA08]). Combining two logs $\mathrm{L}_1$ and $\mathrm{L}_2$ into a combined log $\mathrm{L}_1 \uplus \mathrm{L}_2$ such that a log of fitting traces $\mathrm{L}_2$ is added to a log $\mathrm{L}_1$ that is not perfectly fitting to some process model M results in fitness $tbf(\mathrm{L1} \uplus \mathrm{L}_2, \mathrm{M}) = 1 - \frac{\frac{mt_1+mt_2}{ct_1+ct_2}+\frac{rt_1+rt_2}{pt_1+pt_2}}{2}$ for $mt_1$, $mt_2$, $ct_1$, $ct_2$, $rt_1$, $rt_2$, $pt_1$, $rt_2$ being the missing, consumed, remaining, and produced tokens resulting from replaying the traces of $\mathrm{L}_1$ respectively $\mathrm{L}_2$. When a full state space exploration is used for token-based replay, then no missing or remaining tokens are needed for the replay of fitting traces, and the fitness of M on the combined log $\mathrm{L}_1 \uplus \mathrm{L}_2$ can be simplified to $tbf(\mathrm{L1} \uplus \mathrm{L}_2, \mathrm{M}) = 1 - \frac{\frac{mt_1}{ct_1+ct_2}+\frac{rt_1}{pt_1+pt_2}}{2}$ as $\tilde{\mathrm{L}}_2 \subseteq \mathrm{M}$. It is clear that $tbf(\mathrm{L1} \uplus \mathrm{L}_2, \mathrm{M}) > tbf(\mathrm{L1}, \mathrm{M})$ when additionally $|\mathrm{L}_2| > 0$ and $\tilde{\mathrm{L}}_1 \not\subseteq \mathfrak{L}(\mathrm{M})$, as $\tilde{\mathrm{L}}_2 \subseteq \mathrm{M}$ and $|\mathrm{L}_2| > 0$ together imply that $ct_2 > 0$ and $pt_2 > 0$ while $\tilde{\mathrm{L}}_1 \not\subseteq \mathfrak{L}(\mathrm{M})$ implies that $mt_1 > 0$ or $rt_1 > 0$. Therefore, **A4** holds for token-based fitness, however, it does not hold when using the heuristics for token-based replay that are used in the implementation. Another consequence of not needing any missing or remaining tokens to replay fitting traces on a process model is that for all completely fitting logs $\mathrm{L} \subseteq \mathfrak{L}(\mathrm{M})$, $tbf(\mathrm{L}, \mathrm{M}) = 1 - \frac{\frac{0}{ct}+\frac{0}{pt}}{2} = 1$. Therefore, **A5** holds also holds for token-based fitness when a full state space exploration is used. However, also **A5** does not hold when using the heuristics for token-based replay that are used in the implementation.

Consider logs $\mathrm{L}_1 = [\langle \mathrm{c}, \mathrm{b}, \mathrm{a} \rangle]$ and $\mathrm{L}_2 = [\langle \mathrm{a}, \mathrm{c} \rangle]$ and the process model M shown in Figure 3.1, such that $(\tilde{\mathrm{L}}_1 \cap \mathfrak{L}(\mathrm{M})) = \emptyset$ and $(\tilde{\mathrm{L}}_2 \cap \mathfrak{L}(\mathrm{M})) = \emptyset$. To replay $\mathrm{L}_1$ on M, there is one produced token in the initial marking. Then, to fire $\mathrm{t}_3$, one token is artificially created in $\mathrm{p}_2$ which is then immediately consumed by firing $\mathrm{t}_3$, producing a token in $\mathrm{p}_4$. This is followed by firing $\mathrm{t}_2$ for which one token is artificially created in $\mathrm{p}_3$, which is then immediately consumed in order to fire $\mathrm{t}_2$, and one token is produced in $\mathrm{p}_5$. Firing $\mathrm{t}_4$ consumes the tokens from $\mathrm{p}_4$ and $\mathrm{p}_5$ and produces one in $\mathrm{p}_6$. Firing $\mathrm{t}_1$ consumes the token from $\mathrm{p}_1$ and produces tokens in $\mathrm{p}_2$ and in $\mathrm{p}_3$, both of which are remaining tokens after completion of the replay. The token in place $\mathrm{p}_6$ of the final marking is finally consumed. In total, replaying this trace on M results in $mt = 2$, $rt = 2$, $pt = 5$, $ct = 5$, and $tbf(L_1, M) = 1 - \frac{\frac{2}{5}+\frac{2}{5}}{2} = \frac{3}{5} = 0.6$. To replay $L_2$ on M, there is one produced token in the initial marking, then firing $t_1$ consumes the token from $p_1$ and produces one in $p_2$ and one in $p_3$. Then, firing $t_3$ consumes the token from $p_2$ and produces one in $p_4$, after which replay ends and tokens remain in $p_3$ and $p_4$. Finally, a token is consumed from place $p_6$ of the final marking, which is missing and first has to be artificially produced. In total, replaying this trace on M results in $mt = 1$, $rt = 2$, $pt = 4$, $ct = 3$, and $tbf(L_2, M) = 1 - \frac{\frac{1}{3}+\frac{2}{4}}{2} = \frac{7}{12} \approx 0.583$. Since $tbf(L_1, M) \neq tbf(L_2, M)$ while $(\tilde{L}_1 \cap \mathfrak{L}(M)) = \emptyset$ and $(\tilde{L}_2 \cap \mathfrak{L}(M)) = \emptyset$, **A6** does not hold for token-based fitness.

Consider log $L_1 = [\langle c, b, a, a, a\rangle, \langle a, b, c\rangle]$, log $L_2 = [\langle a, c\rangle]$ and process model M shown in Figure 3.1. To replay trace $\langle c, b, a, a, a\rangle$ of $L_1$ on M, there is one produced token in the initial marking, then, to fire $t_3$, one token is artificially created in $p_2$ which is immediately consumed by firing $t_3$ and a token is produced in $p_4$. Next, $t_2$ is fired for which one token is artificially created in $p_3$ which is immediately consumed and one token is produced in $p_5$. Firing $t_4$ consumes the tokens from $p_4$ and $p_5$ and produces one in $p_6$. Firing $t_1$ consumes the token from $p_1$ and produces tokens in $p_2$ and in $p_3$. To mimic both the second and the third a, $t_1$ is fired again twice, for which two artificial tokens are created in $p_1$ which is immediately consumed by $t_1$ and two tokens are produced in $p_2$ and in $p_3$. Places $p_2$ and in $p_3$ now both contain three tokens that are remaining tokens after completion of the replay. Finally, the token in place $p_6$ is consumed as it is part of the final marking. To replay fitting trace $\langle a, b, c\rangle$ of $L_1$ on M, one token is produced in the initial marking $p_1$ which is consumed by firing $t_1$ producing a token in $p_2$ and in $p_3$. Firing $t_2$ consumes the token of $p_3$ and produces one in $p_5$. Firing $t_3$ consumes the token of $p_2$ and produces one in $p_4$. Firing $t_4$ consumes the tokens of $p_4$ and $p_5$ and produces a token in $p_6$ which is consumed since it is part of the final marking. No tokens remain after replay. In total, replaying both traces of $L_1$ on M results in $mt = 4$, $rt = 6$, $pt = 16$, $ct = 14$, and $tbf(L_1, M) = 1 - \frac{\frac{4}{14}+\frac{6}{16}}{2} = \frac{75}{112} \approx 0.670$. As explained in the counterexample for **A6**, to replay $\langle a, c\rangle$ of $L_2$ on M, $mt = 1$, $rt = 2$, $pt = 4$, $ct = 3$. To replay $L_1 \uplus L_2$ on M, $mt = 5$, $rt = 8$, $pt = 20$, $ct = 17$, and $tbf(L_1 \uplus L_2, M) = 1 - \frac{\frac{5}{17}+\frac{8}{20}}{2} = \frac{111}{170} \approx 0.630$. Since $(\tilde{L}_1 \cap \mathfrak{L}(M)) \neq \emptyset$, $|L_2| >$ and $(\tilde{L}_2 \cap \mathfrak{L}(M)) = \emptyset$, but $tbf(L_1 \uplus L_2, M) > tbf(L_1, M)$, **A7** does not hold for token-based fitness.

Token-based fitness is an event-level fitness measure, where the similarity between a trace and a process model is defined as $sim_{tbf}(\sigma, M) = 1 - \frac{\frac{mt}{ct}+\frac{rt}{pt}}{2}$. **A8** trivially holds: if two logs $L_1$ and $L_2$ have traces that are identical in their similarity to process model M, then M has identical fitness on those two logs. This is easy to see, since the definition of token-based fitness is simply $sim_{tbf}$ lifted to the level of event logs. **A9** also holds trivially: if all traces of a log $L_1$ have a higher similarity than all traces of $L_2$, then all the traces of $L_1$ have a lower ratio of missing tokens w.r.t. consumed token and a lower ratio of remaining tokens w.r.t. produced token than all the traces of $L_2$. Therefore, a process model M has a lower token-based fitness on combined event log $L_1 \uplus L_2$ than on $L_1$.

### 3.2.2 Alignment-based Fitness

One of the main limitations of token-based replay is that the replay approach does not create a corresponding path through the model for a given trace, i.e., it does not make a best effort to provide an explanation of a trace in terms of model behavior. Alignments [ADA11] were introduced to overcome this limitation of token-based replay. For a trace $\sigma$ from a log L that fits on an accepting Petri net *APN*, alignments [ADA11] give a sequence of transition firings $\gamma \in T^*$ such that $m_0 \xrightarrow{\gamma} m_f$ with $m_0$ the initial marking and $m_f$ a final marking of *APN* and $\ell(\gamma)=\sigma$.

Figure 3.4 shows multiple examples of alignments on the process model M of Figure 3.1. First, Figure 3.4a shows an alignment of trace $\sigma_1 = \langle a, b, c \rangle$ on M, which consists of synchronous moves on a, b, and c in the log which the model mimics by firing $t_1$, $t_2$, and $t_3$, finally followed by a model move on silent transition $t_4$ to reach the final marking. An alternative alignment of the same trace is shown in Figure 3.4b, which consists of three log moves a, b, and c, followed by four model moves on $t_1$, $t_2$, $t_3$, and $t_4$. Calculating *optimal alignments* consists of finding an alignment that is minimal in terms of its total *misalignment cost*. These costs for misalignment can be defined in a flexible way, but, in practice, often the *standard cost function* is used, which is defined in [ADA11] and proposes to assign a cost of 0 to synchronous moves, a cost of 1 to log moves, and for model moves a cost of 1 for labeled transitions and 0 for silent transitions. Using the standard cost function, Figure 3.4a is an optimal alignment of $\sigma_1$ on M, yielding a total cost of 0 since it consists solely of synchronous moves and of a model move on a silent transition. The alternative alignment of $\sigma_1$ on M that is shown in Figure 3.4b is suboptimal, as it yields a total cost of 7 (3 log moves and 4 model moves on non-silent transitions). Note that the alignment shown in Figure 3.4b is not only suboptimal, it is in fact the worst possible alignment. In general, the alignment with the highest cost for a given trace $\sigma$ and model M consists of log moves on all events in $\sigma$ and a sequence of model moves that corresponds to the shortest path through the model from initial marking $m_0$ to $m_f$. For a log L and model M, alignment-based fitness is defined as $abf(L, M) = 1 - \frac{\sum_{\sigma \in L} \delta(\Gamma_M(\sigma))}{\sum_{\sigma \in L} \delta(\overline{\Gamma_M}(\sigma))}$.

While $\Gamma_M(\sigma)$ can be non-deterministic in the sense that it returns an arbitrary optimal alignment from the set of optimal alignments, all optimal alignments have the same cost $\delta(\Gamma_M(\sigma))$. Furthermore, it is easy to see that $\delta(\overline{\Gamma_M}(\sigma))$ is also deterministic. Therefore, alignment-based fitness satisfies **A1**.

For models $M_1$ and $M_2$ such that $\mathfrak{L}(M_1) \subseteq \mathfrak{L}(M_2)$, trivially, for each $\sigma \in L$ such that $\sigma \in \mathfrak{L}(M_1)$ also $\sigma \in \mathfrak{L}(M_2)$. Therefore, all such traces can be aligned on both models without needing log moves and without model moves on non-silent transitions. For a trace $\sigma \in L$ such that $\sigma \notin \mathfrak{L}(M_1)$, the model trace $\sigma' \in \mathfrak{L}(M_1)$

| Log | a | b | c | ≫ |
|---|---|---|---|---|
| Model | $t_1$ | $t_2$ | $t_3$ | $t_4$ |
| | a | b | c | τ |

**(a)**

| Log | a | b | c | ≫ | ≫ | ≫ | ≫ |
|---|---|---|---|---|---|---|---|
| Model | ≫ | ≫ | ≫ | $t_1$ | $t_2$ | $t_3$ | $t_4$ |
| | | | | a | b | c | τ |

**(b)**

| Log | d | e | f | ≫ | ≫ | ≫ | ≫ |
|---|---|---|---|---|---|---|---|
| Model | ≫ | ≫ | ≫ | $t_1$ | $t_2$ | $t_3$ | $t_4$ |
| | | | | a | b | c | τ |

**(c)**

| Log | ≫ | ≫ | ≫ | ≫ | d | e | f |
|---|---|---|---|---|---|---|---|
| Model | $t_1$ | $t_2$ | $t_3$ | $t_4$ | ≫ | ≫ | ≫ |
| | a | b | c | τ | | | |

**(d)**

| Log | a | b | c | ≫ | d |
|---|---|---|---|---|---|
| Model | $t_1$ | $t_2$ | $t_3$ | $t_4$ | ≫ |
| | a | b | c | τ | |

**(e)**

**Figure 3.4:** Five alignments on the process model M of Figure 3.1:*(a)* an optimal alignment of trace $\sigma_1 = \langle a, b, c\rangle$ on M, *(b)* a suboptimal alignment of trace $\sigma_1$ on M, *(c) and (d)* two different optimal alignments of trace $\sigma_2 = \langle d, e, f\rangle$, and *(e)* an optimal alignment of trace $\sigma_2 = \langle a, b, c, d\rangle$.

that is most similar to $\sigma$ in terms of alignment cost is also part of $\mathfrak{L}(M_2)$, therefore, $\sigma$ can be aligned on $M_2$ with at most the same cost as on $M_1$. This shows that for alignment-based fitness it holds that $\mathfrak{L}(M_1) \subseteq \mathfrak{L}(M_2) \implies \mathit{fit}(L, M_1) \leq \mathit{fit}(L, M_2)$. Observe that this shows that **A2** holds for alignment-based fitness. Note that when $(\mathfrak{L}(M_2) \setminus \mathfrak{L}(M_1)) \cap \tilde{L} \neq \emptyset$, there is at least one trace $\sigma \in L$ such that $\sigma \notin \mathfrak{L}(M_1)$ while $\sigma \in \mathfrak{L}(M_2)$. Aligning this trace $\sigma$ on $M_1$ takes a nonzero cost, while $\sigma$ can be alignment with zero cost on $M_2$. This shows that **A3** also holds for alignment-based fitness.

For two logs $L_1$ and $L_2$ and a process model M, the fitness of the combined log $L_1 \uplus L_2$ is $\mathit{abf}(L_1 \uplus L_2, M) = 1 - \frac{\sum_{\sigma \in L_1} \delta(\Gamma_M(\sigma)) + \sum_{\sigma \in L_2} \delta(\Gamma_M(\sigma))}{\sum_{\sigma \in L_1} \delta(\overline{\Gamma_M}(\sigma)) + \sum_{\sigma \in L_2} \delta(\overline{\Gamma_M}(\sigma))}$. When $L_2 \subseteq \mathfrak{L}(M)$, we know that $\forall_{\sigma \in L_2} \delta(\Gamma_M(\sigma)) = 0$. Additionally, when $|L_2| > 0$, we know that $\sum_{\sigma \in L_2} \delta(\overline{\Gamma_M}(\sigma)) > 0$. When $L_1 \not\subseteq \mathfrak{L}(M)$, we know that $\sum_{\sigma \in L_1} \delta(\Gamma_M(\sigma)) > 0$. Therefore, $\mathit{abf}(L_1 \uplus L_2, M) > \mathit{abf}(L_1, M)$ when $L_2 \subseteq \mathfrak{L}(M)$, $|L_2| > 0$, and $L_1 \not\subseteq \mathfrak{L}(M)$, i.e., **A4** holds for alignment-based fitness.

For any trace $\sigma$ such that $\sigma \in \mathfrak{L}(M)$, $\delta(\Gamma_M(\sigma)) = 0$. Therefore, for any log L such that $\tilde{L} \subseteq \mathfrak{L}(M)$, $\mathit{abf}(L, M) = 1 - \frac{\delta(\Gamma_M(L))}{\delta(\overline{\Gamma_M}(L))} = 1 - \frac{0}{\delta(\overline{\Gamma_M}(L))} = 1$. This shows that **A5** holds for alignment-based fitness.

As a counterexample of **A6** for alignment-based fitness, consider process model M from Figure 3.1 and $L_1 = [\langle d, e, f\rangle]$ and $L_2 = [\langle a, b, c, d\rangle]$. Note that, in accordance with **A6**, $(\tilde{L}_1 \cap \mathfrak{L}(M)) = \emptyset$ and $(\tilde{L}_2 \cap \mathfrak{L}(M)) = \emptyset$. The optimal alignments of M on logs $L_1$ and $L_2$ are respectively shown in Figure 3.4c and Figure 3.4e. The alignment in Figure 3.4c consists of 3 log moves and it consists of 4 model moves out of which 3 are on non-silent transitions, therefore the alignment costs are 6. Trace $\langle d, e, f\rangle$ has length 3 and the shortest path through model M consists of 3 non-silent transitions, therefore $\delta(\overline{\Gamma_M}(\langle d, e, f\rangle)) = 6$, resulting in $abf(L_1, M) = 1 - \frac{\delta(\Gamma_M(\langle d,e,f\rangle))}{\delta(\overline{\Gamma_M}(\langle d,e,f\rangle))} = 1 - \frac{6}{6} = 0$. The alignment in Figure 3.4e consists of 1 log move and it consists of 1 model move, but it is on a silent transition, therefore $\delta(\Gamma_M(\langle a, b, c, d\rangle)) = 1$, i.e., the alignment costs are 1. Trace $\langle a, b, c, d\rangle$ has length 4 and the shortest path through model M consists of 3 non-silent transitions, therefore $\delta(\overline{\Gamma_M}(\langle a, b, c, d\rangle)) = 7$, resulting in $abf(L_2, M) = 1 - \frac{\delta(\Gamma_M(\langle a,b,c,d\rangle))}{\delta(\overline{\Gamma_M}(\langle a,b,c,d\rangle))} = 1 - \frac{1}{7} \approx 0.857$. Because $abf(L_2, M) > abf(L_1, M)$ while $(L_1 \cap \mathfrak{L}(M)) = \emptyset$ and $(L_2 \cap \mathfrak{L}(M)) = \emptyset$, **A6** does not hold.

A similar type of argument that we used to show that **A6** does not hold for alignment-based fitness can be used to show that **A7** does not hold. As a counterexample of **A7** for alignment-based fitness, consider the same process model M from Figure 3.1, but now and $L_1 = [\langle a, b, c\rangle, \langle d, e, f\rangle]$ and $L_2 = [\langle a, b, c, d\rangle]$. Log $L_1$ contains the same non-fitting trace as in the counter-example for **A6**, but now it has an additional fitting trace $\langle a, b, c\rangle$. $L_2$ is the exact same log as in the counter-example for **A6**. Note that, as required by **A7**, $(\tilde{L}_1 \cap \mathfrak{L}(M)) \neq \emptyset$ and $(\tilde{L}_2 \cap \mathfrak{L}(M)) = \emptyset$. The optimal alignments of the non-fitting traces in $L_1$ and $L_2$ (i.e., $\langle d, e, f\rangle$ respectively $\langle a, b, c, d\rangle$) were already discussed in the counter example of **A6**, and we had found $\delta(\Gamma_M(\langle d, e, f\rangle)) = 6$, $\delta(\overline{\Gamma_M}(\langle d, e, f\rangle)) = 6$, $\delta(\Gamma_M(\langle a, b, c, d\rangle)) = 1$, and $\delta(\overline{\Gamma_M}(\langle a, b, c, d\rangle)) = 7$. The optimal alignment for trace $\langle a, b, c\rangle$ is shown in Figure 3.4a and it consists of one model move on a silent transition such that $\delta(\Gamma_M(\langle d, e, f\rangle)) = 0$ and $\delta(\overline{\Gamma_M}(\langle d, e, f\rangle)) = 6$. Combining the two traces in the logs, the fitness values result in $abf(L_1, M) = 1 - \frac{\delta(\Gamma_M(\langle a,b,c\rangle))+\delta(\Gamma_M(\langle d,e,f\rangle))}{\delta(\overline{\Gamma_M}(\langle a,b,c\rangle))+\delta(\overline{\Gamma_M}(\langle d,e,f\rangle))} = 1 - \frac{6}{12} = 0.5$ and $abf(L_1 \uplus L_2, M) = 1 - \frac{\delta(\Gamma_M(\langle a,b,c\rangle))+\delta(\Gamma_M(\langle d,e,f\rangle))+\delta(\Gamma_M(\langle a,b,c,d\rangle))}{\delta(\overline{\Gamma_M}(\langle a,b,c\rangle))+\delta(\overline{\Gamma_M}(\langle d,e,f\rangle))+\delta(\overline{\Gamma_M}(\langle a,b,c,d\rangle))} = 1 - \frac{7}{19} \approx 0.632$. Because $abf(L_1 \uplus L_2, M) > abf(L_1, M)$ while $(\tilde{L}_1 \cap \mathfrak{L}(M)) \neq \emptyset$, $|L_2| > 0$, and $(L_2 \cap \mathfrak{L}(M)) = \emptyset$, **A7** does not hold. Note that this same counter-example also shows that corollary 3.5 and thereby also proposition *RecPro3*$^+$ of van der Aalst [Aal18] do not hold for alignment-based fitness.

Alignment-based fitness is an event-level fitness measure, where the similarity between a trace and a process model is defined as $sim_{abf}(\sigma, M) = \frac{\delta(\Gamma_M(\sigma))}{\delta(\overline{\Gamma_M}(\sigma))}$. **A8** trivially holds: if two logs $L_1$ and $L_2$ have traces that are identical in their similarity to process model M, then M has identical fitness on those two logs. This is easy to see, since the definition of alignment-based fitness is simply $sim_{abf}$ lifted to the level of event logs. **A9** also holds trivially: if all traces of a log $L_1$ have a higher similarity than all traces of $L_2$, then all the traces of $L_1$ have a lower misalignment cost compared to the traces of $L_2$. Therefore, a process model M has a lower alignment-based fitness on combined event log $L_1 \uplus L_2$ than on $L_1$.

### 3.2.3 Ratio of Fitting Traces

While the *ratio of fitting traces* measure is a simple and naive fitness measure, it is easy to see that it fulfills all the axioms. First, since it is deterministic whether a trace fits or does not fit a model, **A1** holds.

The ratio of fitting traces measure also fulfills **A2**. To see that this is indeed the case, suppose for a second that $\mathfrak{L}(M_1) \subseteq \mathfrak{L}(M_2)$ while $rft(L, M_1) > rft(L, M_2)$. This would imply that there exists a trace $\sigma \in L$ such that $\sigma \in \mathfrak{L}(M_1)$ but $\sigma \in \mathfrak{L}(M_2)$, otherwise the ratio of traces of L that fit on the model could not be higher for $M_1$ than for $M_2$. However, such a trace $\sigma$ cannot exist, since $\mathfrak{L}(M_1) \subseteq \mathfrak{L}(M_2)$. Therefore, **A2** holds.

To show that this fitness measure also fulfills **A3**, suppose for a moment that $rft(L, M_1) \geq rft(L, M_2)$ while $\mathfrak{L}(M_1) \subseteq \mathfrak{L}(M_2)$ and $((\mathfrak{L}(M_2) \setminus \mathfrak{L}(M_1)) \cap \tilde{L}) \neq \emptyset$. Note that we have already shown that $rft(L, M_1) > rft(L, M_2)$ contradicts with $\mathfrak{L}(M_1) \subseteq \mathfrak{L}(M_2)$. Therefore, to show that $rft(L, M_1) \geq rft(L, M_2)$ cannot hold when $\mathfrak{L}(M_1) \subseteq \mathfrak{L}(M_2)$, it only remains to be shown that $rft(L, M_1) = rft(L, M_2)$ contradicts with $\mathfrak{L}(M_1) \subseteq \mathfrak{L}(M_2)$ and $((\mathfrak{L}(M_2) \setminus \mathfrak{L}(M_1)) \cap \tilde{L}) \neq \emptyset$. If $rft(L, M_1) = rft(L, M_2)$, then either there is no trace $\sigma \in L$ such that $\sigma \in \mathfrak{L}(M_2)$ while $\sigma \notin \mathfrak{L}(M_1)$, or, when there is such a trace $\sigma$, there has to exist another trace $\sigma' \in L$ such that $\sigma' \in \mathfrak{L}(M_1)$ while $\sigma' \notin \mathfrak{L}(M_2)$. However, such a trace $\sigma'$ cannot exist because $\mathfrak{L}(M_1) \subseteq \mathfrak{L}(M_2)$. Therefore, there can be no trace $\sigma \in \mathfrak{L}(M_2)$ while $\sigma \notin \mathfrak{L}(M_1)$, but this contradicts with $((\mathfrak{L}(M_2) \setminus \mathfrak{L}(M_1)) \cap \tilde{L}) \neq \emptyset$. This shows that $rft(L, M_1) \geq rft(L, M_2)$ cannot hold, and thus $rft(L, M_1) < rft(L, M_2)$. Therefore, **A3** holds for the ratio of fitting traces fitness measure.

Axiom **A4** trivially holds. To see that this is the case, consider that when we add the traces of a fitting event $L_2$ to an unfitting event log $L_1$, the rft-fitness of the combined log $L_1 \uplus L_2$ is $rft(L_1 \uplus L_2, M) = \frac{\sum_{\sigma \in (\tilde{L}_1 \cap \mathfrak{L}(M))} L_1(\sigma) + \sum_{\sigma \in (\tilde{L}_2 \cap \mathfrak{L}(M))} L_2(\sigma)}{|L_1| + |L_2|}$. Since **A4** assumes a log $L_2$ such that $L_2 \subseteq \mathfrak{L}(M)$, $rft(L_1 \uplus L_2, M) = \frac{|L_2| + \sum_{\sigma \in (\tilde{L}_1 \cap \mathfrak{L}(M))} L_1(\sigma)}{|L_1| + |L_2|}$. The fitness of $L_1$ is $rft(L_1 \uplus L_2, M) = \frac{\sum_{\sigma \in (\tilde{L}_1 \cap \mathfrak{L}(M))} L_1(\sigma)}{|L_1|}$, which is $< 1$ when $L_1 \not\subseteq \mathfrak{L}(M)$. Since event log $L_2$ is fitting, $\lim_{|L_2| \to \infty} \frac{|L_2| + \sum_{\sigma \in (\tilde{L}_1 \cap \mathfrak{L}(M))} L_1(\sigma)}{|L_1| + |L_2|} = 1$. Therefore, **A4** holds.

In rtf-fitness, the fitness of any log $L \subseteq \mathfrak{L}(M)$ is $rft(L, M) = \frac{\sum_{\sigma \in (\tilde{L} \cap \mathfrak{L}(M))} L(\sigma)}{|L|} = \frac{|L|}{|L|} = 1$ and the fitness of any log $(L \cap \mathfrak{L}(M)) = \emptyset$ is $rft(L, M) = \frac{\sum_{\sigma \in (\tilde{L} \cap \mathfrak{L}(M))} L(\sigma)}{|L|} = \frac{0}{|L|} = 0$, therefore **A5** and **A6** hold.

The proof that **A7** holds for rft-fitness is similar to the proof for **A4**. When we add the traces of a *non-empty and completely non-fitting* event $L_2$ to an event log $L_1$ that is not completely unfitting, the rft-fitness of the combined log $L_1 \uplus L_2$ is $rft(L_1 \uplus L_2, M) = \frac{\sum_{\sigma \in (\tilde{L}_1 \cap \mathfrak{L}(M))} L_1(\sigma) + \sum_{\sigma \in (\tilde{L}_2 \cap \mathfrak{L}(M))} L_2(\sigma)}{|L_1| + |L_2|}$. Since **A7** assumes a log $L_2$ such that $(L_2 \cap \mathfrak{L}(M)) = \emptyset$, we know that $\sum_{\sigma \in (\tilde{L}_2 \cap \mathfrak{L}(M))} L_2(\sigma) = 0$ and therefore $rft(L_1 \uplus L_2, M) = \frac{\sum_{\sigma \in (\tilde{L}_1 \cap \mathfrak{L}(M))} L_1(\sigma)}{|L_1|+|L_2|}$. Since **A7** also assume $L_2$ such that $|L_2| > 0$, we know that $\frac{\sum_{\sigma \in (\tilde{L}_1 \cap \mathfrak{L}(M))} L_1(\sigma)}{|L_1|+|L_2|} > \frac{\sum_{\sigma \in (\tilde{L}_1 \cap \mathfrak{L}(M))} L_1(\sigma)}{|L_1|}$, and therefore that $rft(L_1 \uplus L_2, M) > rft(L_1, M)$, indicating that **A7** holds.

**Table 3.1:** An overview of the fitness axioms and whether they hold (✓) or do not hold (✗) for each fitness measure.

| Measure | A1 | A2 | A3 | A4 | A5 | A6[6] | A7[6] | A8 | A9 |
|---|---|---|---|---|---|---|---|---|---|
| Ratio of Fitting Traces | ✓ | ✓ | ✓ | ✓ | ✓ | ✓ | ✓ | ✓ | ✓ |
| Token-based Fitness | ✗[7] | ✗ | ✗ | ✓[8] | ✓[8] | ✗ | ✗ | ✓ | ✓ |
| Alignment-based Fitness | ✓ | ✓ | ✓ | ✓ | ✓ | ✗ | ✗ | ✓ | ✓ |

The rtf-fitness measure is a trace-level fitness measure. However, we can still define the similarity function between a trace $\sigma$ and a process model M as for event-level fitness measures. The similarity between a trace and a process model is defined as $sim_{rtf}(\sigma, M) = \begin{cases} 1, & \text{if } \sigma \in \mathfrak{L}(M) \\ 0, & \text{otherwise.} \end{cases}$

With this definition of $sim_{rtf}$, **A8** trivially holds. If two logs $L_1$ and $L_2$ have traces that are identical in their similarity to process model M, then M has identical fitness on those two logs. This is easy to see, since the definition of token-based fitness is simply $sim_{tbf}$ lifted to the level of event logs. **A9** also holds trivially: if all traces of a log $L_1$ have a higher similarity than all traces of $L_2$, then all the traces of $L_1$ have a lower ratio of missing tokens with respect to consumed tokens and a lower ratio of remaining tokens with respect to produced tokens than all the traces of $L_2$. Therefore, a process model M has a lower token-based fitness on the combined event log $L_1 \uplus L_2$ than on $L_1$.

### 3.2.4 Discussion and Conclusions on Fitness Measures

Table 3.1 lists the fitness measures and provides an overview of the axioms that we have shown to hold (✓) or have shown not to hold (✗) for each measure.

We have shown that token-based fitness does not fulfill many of the axioms and therefore it does not measure fitness in a consistent way. The frequently used alignment-based fitness measure fulfills all the axioms except for **A6** and **A7**.

The ratio of fitting traces measure fulfills all the axioms for fitness measures. However, in contrast to the other two measures, it not able to look beyond whether or not a trace $\sigma$ of the log fits the model M, i.e., whether $\sigma \in \mathfrak{L}(M)$. In other words, it is not able to distinguish between behavior that *almost* adheres to the modeled behavior and behavior that is *nothing like* the modeled behavior. This puts the ratio of fitting traces in the category of *trace-level fitness* measures, as opposed to *event-level fitness* measures that are able to make such a differentiation. While "adding non-fitting behavior to the log can not increase fitness" [Aal18] has been stated in earlier work by Van der Aalst as a desired property for fitness measures, represented here by **A6** and **A7**, we have shown that property to be inconsistent with the principles of event-level fitness measures. As a consequence, **A6** and **A7** were found to not hold for both event-level fitness measures.

[6] This axiom captures conceptual aspects of trace-level fitness, but it contradicts conceptual aspects of event-level fitness.

[7] Token-based fitness is deterministic, but it is not well-defined for any arbitrary log.

[8] Does not hold when using the heuristics that are used in the implementation.

Axioms **A8** and **A9** respectively formulate weaker versions of **A6** and **A7** and thereby make it possible for them to be fulfilled by event-level fitness measures through the introduction of a notion of similarity between a trace and the behavior that is modeled by a process model, represented by a similarity function *sim*. We argue that additional axioms are needed that specify the desired properties of such a function *sim*. The fact that it is possible for the ratio of fitting traces fitness measure to formulate a similarity measure (i.e., $sim_{rtf}$) such that **A8** and **A9** hold, even though the ratio of fitting traces is not an event-level fitness measure, substantiates this claim. However, what properties are desired from such a similarity function so far seem to be left undiscussed in the research field.

In the remainder of the thesis, we will measure fitness using the alignment-based fitness measure, because it is able to measure fitness on the event level and it fulfills all the axioms except for **A6** and **A7**, which cannot be fulfilled by event-level fitness measures.

## 3.3 Axioms for Precision Measures

The desired properties for precision measures are not clearly defined in existing work. However, they are often discussed informally. Van der Aalst et al. [Aal+11], describe the precision dimension as "measure determining whether the model prohibits behavior very different from the behavior seen in the event log. A model with low precision is *underfitting*". Vanden Broucke et al. [Bro+14] describe it as "precision (or: appropriateness), i.e., the model's ability to disallow unwanted behavior;". Mũnoz-Gama and Carmona [Muñ+10] describe it as "Precision: refers to overly general models, preferring models with minimal behavior to represent as closely as possible to the log.". Buijs et al. [BR14] describe it as "... precision quantifies the fraction of the behavior allowed by the model which is not seen in the event log.".

We consider precision to be a function $prec(L, M)$ that quantifies which part of the language of model M is seen in event log L. Below we formalize desired properties of function *prec* through axioms to *consistently* hold for any kind of model and any kind of log. Note that in the examples that we will show in this chapter all models M will be Petri nets. However, the formulated axioms are more general and apply to any process model $M \in \mathcal{M}$. Figure 3.5 visualizes two axioms in Euler diagrams.

The first axiom states that precision is deterministic, i.e., given a log and model always the same result is returned. Existing precision measures normalize $\mathbb{R}$ to a $[0,1]$-interval.

**Axiom A10.** A precision measure is deterministic, i.e., it is a *function* $prec : \mathcal{B}(\Sigma^+) \times \mathcal{M} \rightarrow \mathbb{R}$. ◇

The second axiom formulates the conceptual description of precision more formally: if a process model $M_2$ allows for more behavior not seen in a log L than another model $M_1$ does, then $M_2$ should have a lower precision than $M_1$ regarding L.

**Axiom A11.** For models $M_1$ and $M_2$ and a log L, $\tilde{L} \subseteq \mathfrak{L}(M_1) \subseteq \mathfrak{L}(M_2) \implies prec(L, M_1) \geq prec(L, M_2)$ ◇

Note that **A11** does allow $\tilde{L} \subseteq \mathfrak{L}(M_1) \subset \mathfrak{L}(M_2)$ with $prec(L, M_1) = prec(L, M_2)$. Ideally, since $\mathfrak{L}(M_1)$ is smaller than $\mathfrak{L}(M_2)$ we would like to see a higher precision for $M_1$, but this requirement might be too strict. However, for a process model M with $\tilde{L} \subseteq \mathfrak{L}(M)$, we would like the precision of M on L to be higher than the precision of M on any flower model (i.e., a model that allows for all behavior over its activities) on log L.

**Axiom A12.** For models $M_1$ and $M_2$ and a log L, $\mathfrak{L}(M_1) \subset \mathcal{P}(\Sigma^*) \wedge \mathfrak{L}(M_2) = \mathcal{P}(\Sigma^*) \implies prec(L, M_1) > prec(L, M_2)$ ◇

The precision of a log on two language equivalent models should be equal, i.e., precision should not depend on the model structure.

**Axiom A13.** For models $M_1$ and $M_2$ and a log L, $\mathfrak{L}(M_1) = \mathfrak{L}(M_2) \implies prec(L, M_1) = prec(L, M_2)$ ◇

**A13** was stated before in an informal manner by Rozinat and van der Aalst [RA08], who stated that precision should be independent of structural properties of the model.

Adding fitting traces to a fitting log can only increase the precision of a given model with respect to the log.

**Axiom A14.** For model M and logs $L_1$ and $L_2$, $\tilde{L}_1 \subseteq \tilde{L}_2 \subseteq \mathfrak{L}(M) \implies prec(L_2, M) \geq prec(L_1, M)$ ◇

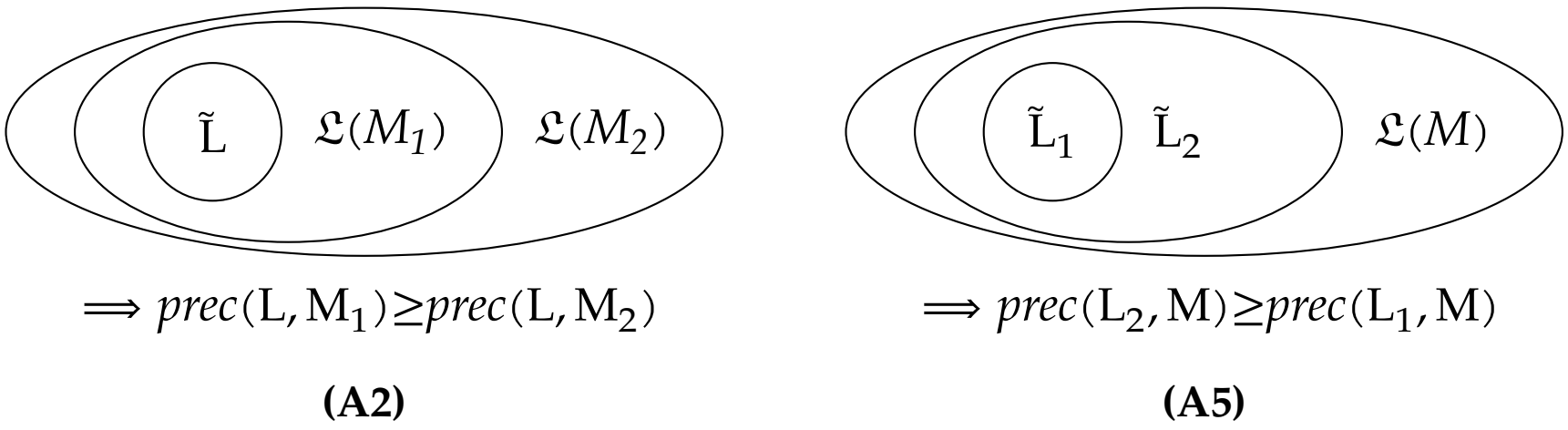

**Figure 3.5:** Two of the five axioms for precision measures visualized with Euler diagrams.

From **A14** it follows as a corollary that precision is maximal when the log contains all the traces allowed by the model, and minimal when it contains no traces allowed by the model.

In the coming sections, we will validate whether these axioms hold for several precision measures. Some articles that introduce precision measures explicitly mention that the measure is intended to be used only with a certain subclass of Petri nets. An example of such a subclass of Petri nets is the bounded Petri net, which have the restriction that all places must have a finite number of tokens in all reachable markings. When an article that introduces a precision measure states an explicit assumption on the subclass of Petri nets, then we only validate the axioms on this subclass of Petri nets. When no explicit assumption on a subclass of Petri nets is stated, we assume that the precision measure is intended for Petri nets in general.

## 3.4 Precision Measures

In this section, we give an overview of the precision measures that have been developed in the process mining field, and validate the axioms for precision measures introduced in Section 3.3 for each of those measures.

### 3.4.1 Soundness

Greco et al. [Gre+04; Gre+06a] were the first to propose a precision measure, defining it as the number of unique executions of the process that were seen in the event log divided by the number of unique paths through the process model. This measure is not usable in practice, because it is zero when the process model allows for an infinite number of paths through the model. Any process model having a loop has a precision of 0. More recent precision measures are capable of calculating the precision of a model for an event log even when the models allows for infinite behavior.

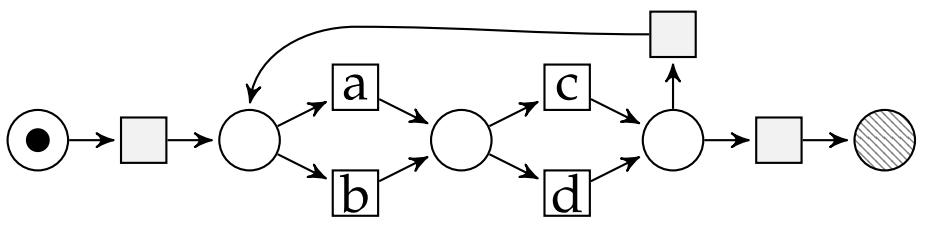

**Figure 3.6:** Model M.

### 3.4.2 Behavioral Appropriateness

Rozinat and Van der Aalst [RA08] proposed the *simple behavioral appropriateness* precision measure, which looks at the average number of enabled transitions during replay. The authors observed themselves that simple behavioral appropriateness is dependent on the structure of the model, and not solely dependent on the behavior that it allows. Therefore, **A13** does not hold for this measure. Furthermore, it is possible that multiple ways to replay a trace on a model exist when it contains silent transitions or duplicate labels, where the average number of enabled transitions can depend on this chosen replay path. This replay path through the model is chosen arbitrarily from the possible ways in which the trace can be replayed. This shows that **A10** does not hold for simple behavioral appropriateness, as it is not deterministic.

In the same paper, Rozinat and van der Aalst [RA08] propose *advanced behavioral appropriateness*, which is independent of the model structure. Advanced behavioral appropriateness calculates the sets $S_F \subseteq \Sigma \times \Sigma$ of pairs of activities that sometimes, but not always, follow each other. Likewise set $S_P \subseteq \Sigma \times \Sigma$ is calculated as the set of activities that sometimes, but not always, precede each other. $S_F^L$ and $S_P^L$ denote the sometimes-follows and sometimes-precedes relations on the log, and $S_F^M$ and $S_P^M$ denotes the sometimes-follows and sometimes-precedes relations according to the model. However, to calculate $S_F^M$ and $S_P^M$, exhaustive exploration of the state space of the model is required, prohibiting the application of this measure for large models or highly concurrent models, where the state-space explosion problem arises. Advanced behavioral appropriateness precision is defined as $a'_b = \left(\frac{|S_F^L \cap S_F^M|}{2 \cdot |S_F^M|} + \frac{|S_P^L \cap S_P^M|}{2 \cdot |S_P^M|}\right)$. Because $S_F^M$ and $S_P^M$ are obtained through exhaustive exploration of the state space of the model, it is easy to see that they depend only on the behavior of the model and not on its structure, therefore **A13** holds. A problem with advanced behavioral appropriateness occurs for deterministic models, where $|S_P^M| = |S_F^M| = 0$, leading to undefined precision. This shows that advanced behavioral appropriateness is a partial function, which conflicts with **A10**.

Rozinat and van der Aalst [RA08] state that simple behavioral appropriateness and advanced behavioral appropriateness assume the Petri net to be in the class of *sound workflow (WF) nets* [Aal97; Aal98].

Consider model M of Figure 3.6, which belongs to the class of sound WF-nets, and any log L such that $\tilde{L} \subseteq \mathfrak{L}(M)$. The loop in model M causes $S_F^M$ and $S_P^M$ to contain all pairs of activities of $\Sigma$. Therefore, $|S_F^M|$ and $|S_P^M|$ are identical to the some-times relations $|S_F^{M'}|$ and $|S_P^{M'}|$ of any model $M'$ with $\mathfrak{L}(M')=\wp(\Sigma^*)$, leading to $prec(L, M)=prec(L, M')$. As $\mathfrak{L}(M)\subset\wp(\Sigma^*)$, this is in conflict with **A12**.

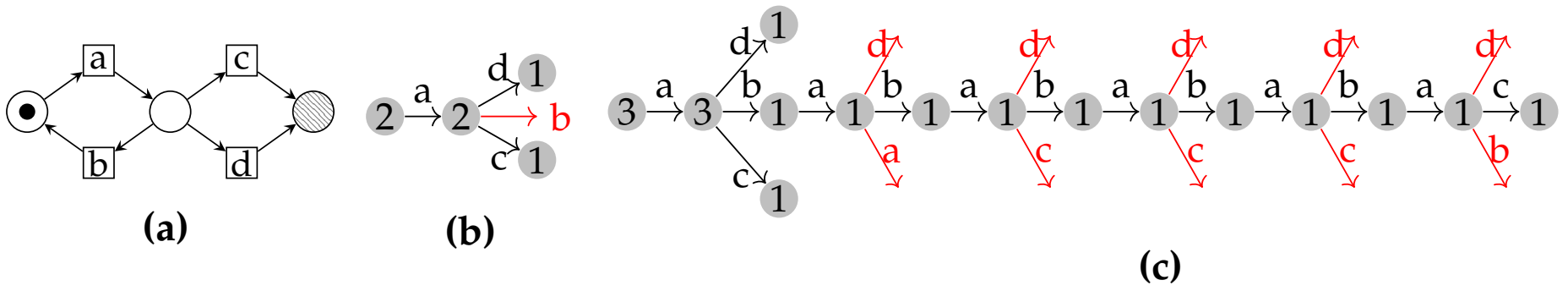

**Figure 3.7:** *(a)* Model M, and the alignment automata on M for *(b)* log $L_1 = [\langle a, c\rangle, \langle a, d\rangle]$, and for *(c)* log $L_2 = [\langle a, c\rangle, \langle a, d\rangle, \langle a, b, a, b, a, b, a, b, a, b, a, c\rangle]$. Red arcs correspond to escaping edges.

### 3.4.3 Escaping Edges Precision

Escaping Edges Precision (ETC) [Muñ+10] calculates precision by constructing a *prefix automaton*, which consists of one state per unique prefix of the event log. Figure 3.7b shows an example prefix automaton for an event log $L = [\langle a, c\rangle, \langle a, d\rangle]$. For each state in the prefix automaton, it is then determined which activities are allowed as next activities by the process model. Activities that are allowed as next activities for some prefix but that are never observed in the event log after this prefix are referred to as *escaping edges*.

In later work [AAD12; Adr+12], a version of ETC precision was developed that is based on the concept of alignments [ADA11], which were also used for alignment-based fitness (see Section 3.2.2). In the alignment-based ETC, the prefix automaton is calculated on the aligned event log instead of the original event log, making the precision measure applicable to non-fitting traces, i.e., traces that are not in the language of the model. Adriansyah et al. [Adr+12] describe two versions of the alignment-based escaping edges precision: *one-align ETC*, which calculates the precision based on one optimal alignment of log and model, and *all-align ETC*, which calculates the precision based on all optimal alignments between log and model. In practice, it is often computationally infeasible to calculate all optimal alignments. A later precision measure, *representative-align ETC* [Adr+15], calculates the escaping edges based on a sample of optimal alignments, and can, therefore, be seen as a trade-off between the computational efficiency of one-align ETC and the reliability of all-align ETC. The papers on ETC precision and its variants do not state an assumption on a subclass of Petri nets. ETC, one-align ETC, all-align, and representative-align ETC precision are all implemented in the package ETConformance[9] as part of the process mining framework ProM [Don+05].

The one optimal alignment that is used by one-align ETC is chosen arbitrarily from the set of optimal alignments of a log on a model. However, different opti-mal alignments result in different prefix automata, which can potentially lead to different precision values. This shows that **A10** does not hold for one-align ETC.

[9] https://svn.win.tue.nl/trac/prom/browser/Packages/ETConformance

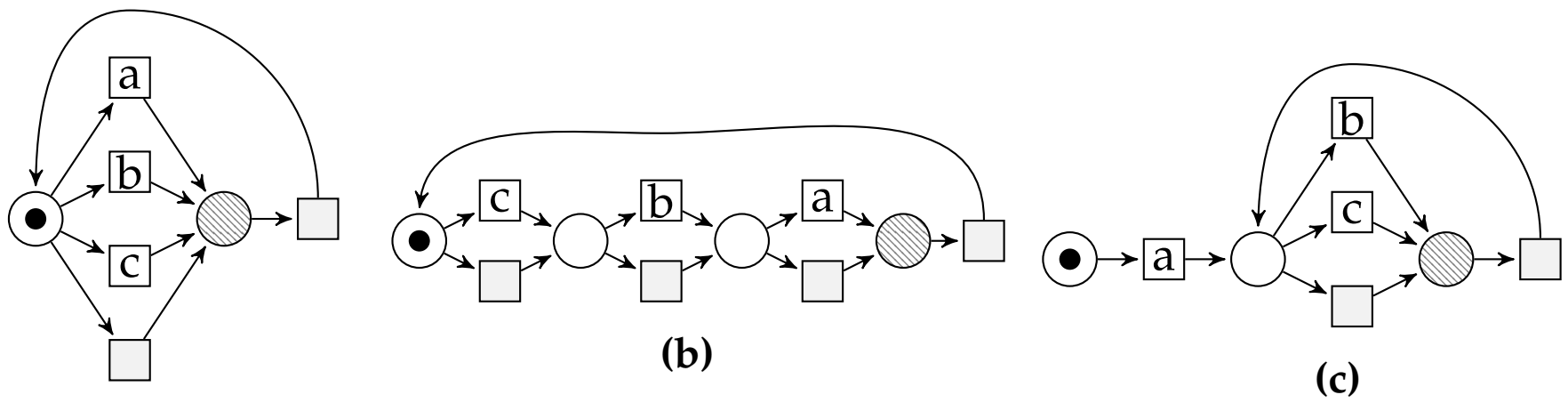

**Figure 3.8:** *(a)* The flower-model, which allows for all behavior over its set of activities, *(b)* an alternative representation of the flower-model, and *(c)* a constrained model which always starts with activity a.

Consider log $L_1=[\langle a,c\rangle,\langle a,d\rangle]$, log $L_2=[\langle a,c\rangle,\langle a,d\rangle,\langle a,c\rangle,\langle a,b,a,b,a,b,a,b,a,b,a,c\rangle]$ and model M be the Petri net of Figure 3.7a. Note that $\tilde{L}_1 \subset \tilde{L}_2$. The alignment automata that are generated for the calculation of $prec(L_1,M)$ and the calculation of $prec(L_2,M)$ are shown in Figure 3.7b and Figure 3.7c. The circles represent the states of the automaton, and the arrows the transitions. The numbers in the states represent the weights of the states for the precision calculation, i.e., the number of times that states are visited in the alignment of log L on model M [Adr+12]. In an alternative definition of one-align ETC [AAD12] the states are weighted by the number of times that events occurred while being in this state according to the alignment of L on M, instead of the number of times that this state was reached according to this alignment. Figure 3.7b shows that the initial state was visited twice, activity a occurred twice at the start in log $L_1$, resulting in a state from which activities b, d, and c were enabled. From this state, activities c and d were seen once, and activity b was never seen, thus it is an escaping edge. Escaping edges precision is then the (weighted) average ratio of non-escaping edges from all outgoing edges, where states are weighted by the number of times that they are visited. Counting the weighted number of non-escaping edges in the numerator and the weighted total number of edges in the denominator in our example, we find $prec(L_1,M)=\frac{2\times1+2\times2+1\times0+1\times0}{2\times1+2\times3+1\times0+1\times0}=\frac{6}{8}=0.75$. One-align ETC results in the following precision values for M on $L_1$ and $L_2$: $prec(L_1,M)=0.75$ and $prec(L_2,M)=0.7143$. This shows that **A14** does not hold for one-align ETC. By comparing the automata of Figures 3.7b and 3.7c it becomes clear that the single trace that is in $L_2$ but not in $L_1$ frequently brings the model to states with three escaping edges, thereby reducing precision. The prefix automata and the precision calculations for M on logs $L_1$ and $L_2$ were performed manually following the procedure from the paper and validation using the ETConformance plugin in ProM.

Now consider log $L=[\langle a,b,c\rangle]$, and the three Petri nets $M_1$, $M_2$, $M_3$ in Figures

3.8a, 3.8b, and 3.8c respectively. Note that $M_1$ and $M_2$ are language equivalent, as $\mathfrak{L}(M_1)=\mathfrak{L}(M_2)=\{a,b,c\}^*$. $M_3$ is more behaviorally constrained than $M_1$ and $M_2$, since all its traces start with activity a. The one-align precision of $M_1$, $M_2$, $M_3$ on L are: $prec(L,M_1)=0.3333$, $prec(L,M_2)=0.5238$, and $prec(L,M_3)=0.4444$. $\mathfrak{L}(M_3)\subset\mathfrak{L}(M_2)$, but $prec(L,M_2)>prec(L,M_3)$, implying that **A11** does not hold for one-align ETC. Additionally, since $\mathfrak{L}(M_2)=\{a,b,c\}^*$, this implies that **A12** does not hold for one-align ETC. Furthermore, $\mathfrak{L}(M_1)=\mathfrak{L}(M_2)$, but $prec(L,M_1)\neq prec(L,M_2)$, implying that **A13** does not hold for one-align ETC.

Analyzing the ETConformance plugin in ProM we found that the prefix automaton generated for one-align precision for calculation of $prec(L,M_1)$ results in 6 states, belonging to 3 firings of observable transitions and 2 firings of $\tau$-transitions. In 3 of the 6 states, which correspond to $M_1$ being in the initial marking, there are 4 possible next activities according to the model, of which only one is observed for that prefix. Furthermore, it shows that the alignment automaton generated for L and $M_1$ consists of 6 states, the automaton for L and $M_2$ consists of 12 states, and the automaton for L and $M_3$ consists of 5 states. This shows that the silent ($\tau$) transitions in $M_2$ generate additional states in the alignment automaton, leading to a higher precision value.

Computing the precision of $M_1$ and $M_2$ on L with all-align ETC and representative-align ETC did not finish after 8 hours of computation time. The long computation time of those measures on models where many optimal alignments exist is a known issue which hinders the application of those measures in practice.

### 3.4.4 Negative Event Precision

Goedertier et al. [Goe+09] proposed a method to induce *negative events*, i.e., sets of events that were prevented from taking place. Negative events are induced for each position in the event log, i.e., for each event e in each trace of the log, a set of events is induced that could not have taken place instead of event e. De Weerdt et al. [De +11] proposed a precision measure based on negative events, *behavioral precision* ($p_B$), which is closely linked to how precision is defined in the area of data mining. Negative event precision regards a process model as a binary classifier that determines whether a certain event can take place given a certain prefix, and then evaluates the precision of this classifier in data mining terms taking the induced negative events as ground truth. For a given trace prefix, *true positive* (TP) events are defined as events that are possible according to both the process model (i.e., a transition labeled with this event is enabled) and log (i.e., this event is not a negative event). *False positive* events (FP) are negative events that are induced for a given prefix while these negative events were in fact possible according to the model. Behavioral precision is defined as $p_B = \frac{TP}{TP+FP}$, which is in accordance with the definition of precision in the data mining field. In later work [Bro+12] induction of artificial negative events has been refined based on frequent temporal patterns which are mined from the event log. Finally, weighted artificial events,

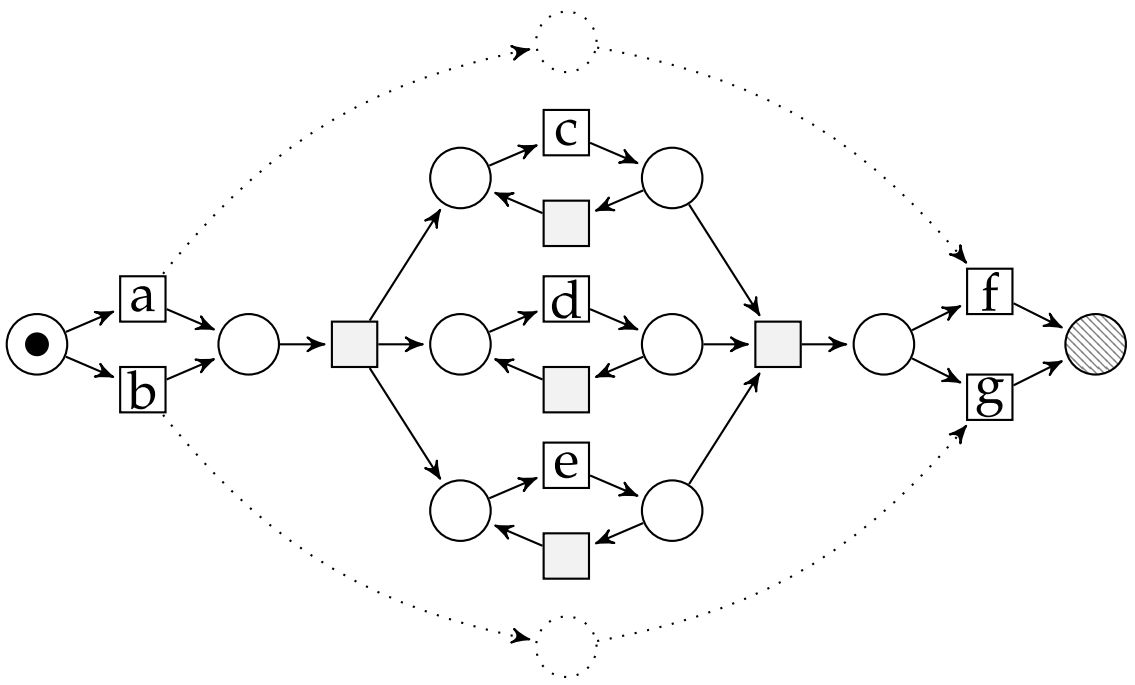

**Figure 3.9:** A process model $M_1$ and its behaviorally restricted variant $M_2$ which includes the dotted places and arcs.

where negative events are weighted according to their confidence, are proposed in [Bro+14].

*Weighted behavioral precision* induces negative events for an event e in the log by taking a window of events w that directly precede e, then calculating all subsequences of events in the log that exactly match w, and finally negative events are identified by calculating which events have never occurred in the log directly after any subsequence matching w. This procedure is repeated for different windows sizes, and the resulting negative events are weighted by the window size.

To induce the events that could not have happened after e.g. trace prefix $\sigma' = \langle a, c, c, d, e, c, d, e, e \rangle$, the method to induce weighted negative events described in [Bro+14] searches for subsequences of events in the log that are identical to the latest k events of $\sigma'$ in the event log. All the activities that have never succeeded such subsequences are considered to be negative events. Furthermore, the confidence of these negative events is based on the length k of those matching subsequences.

Negative event based precision measures, with the different methods for negative event induction, are implemented in the ProM package NEConformance[10]. We will now evaluate the precision measure that uses weighted negative events [Van14], which is the most recent approach to induce negative events and the recommended approach for measuring precision [Van14]. The paper on negative event precision does not mention any assumption on the subclass of Petri nets to which the measure is applicable. Therefore we assume that the measure is intended to be applicable to the whole class of Petri nets in general.

Consider models $M_1$ and $M_2$ of Figure 3.9 respectively excluding and including the arcs and places indicated in dotted lines. $\mathfrak{L}(M_2) \subset \mathfrak{L}(M_1)$, since $M_2$ contains a long-term dependency between activities a and f and between activities b and g, which $M_1$ does not have. Consider an event log L which consists of 10 traces from $M_2$, leading to L being fitting on both $M_1$ and $M_2$. We found the negative event precision of $M_1$ and $M_2$ on the same L to be non-deterministic, resulting in slightly different values every time that it is calculated. This shows that **A10** does not hold for negative event precision.

[10] http://processmining.be/neconformance/

Because negative event precision is non-deterministic, we calculated the precision of $M_1$ and $M_2$ on L both 20 times. The highest precision found in 20 repetitions for $M_1$ is 0.4876, while the lowest precision found for $M_2$ is 0.4545, showing that the non-determinism has the effect that **A11** does not hold for negative event precision. We found an average value of 0.4744 with a standard deviation of 0.0090 for the precision of $M_1$ on L and an average value of 0.4640 with a standard deviation of 0.0072 for the precision of $M_2$ on L. This shows that also in terms of average precision value **A11** does not hold.

To test whether the difference in mean precision between $M_1$ and $M_2$ is due to chance alone we formulate a null hypothesis:

$H_0$ : *The average negative event precision of* $M_2$ *on* L *is higher than or equal to the average negative event precision of* $M_1$ *on* L.

Testing this null hypothesis with a one-tailed Welch t-test [Wel47] we found a *p-value* of 0.0001801, indicating that we can reject the null hypothesis with significance level 0.01. This shows that, with statistical significance, the precision of $M_1$ on L is higher than the precision of $M_2$ on L, which is in disagreement with **A11**.

To see why **A11** does not hold for negative event precision, consider the negative event inducing procure being applied to trace prefix $\sigma' = \langle a, c, c, d, e, c, d, e, e \rangle$ from log L. Petri net $M_2$ generates many different traces because of the parallel length-one-loops on activities c, d and e, which allows for any sequence of any length over these activities. Therefore, the matching subsequences of $\sigma'$ in the log generated from $M_2$ are the subsequences that by chance ended in the same behavior over c, d, and e. Because the sequences of c, d and e events can be long and diverse, activity a and b are unlikely to be present in the matching subsequences, which makes it unlikely that the procedure can induce the negative event g for $\sigma'$. Because the negative events that reflect the constraint that $M_2$ introduces compared to $M_1$ cannot be induced from the log, negative event precision is not able to recognize that $M_2$ is more precise than $M_1$.

### 3.4.5 Projected Conformance Checking

Projected Conformance Checking (PCC) precision was developed by Leemans et al. [LFA16] as a computationally efficient precision measure that scales to event logs with billions of events. PCC precision projects both event log and model on all subsets of activities of size k, and generates minimal deterministic finite automata (DFA) for the behavior over these subsets of activities in the log (i.e., *log automaton*) and for the behavior over these events allowed by the model (i.e., *model automaton*). Based on the log automaton and model automaton it then builds a *conjunction automaton* which allows the behavior that was allowed both in log and model automaton. It then iterates over the states of the model automaton and calculates the share of transitions of each state that is also possible in the corresponding state in the conjunction automaton. It defines precision as the average of this share over the model automaton states.

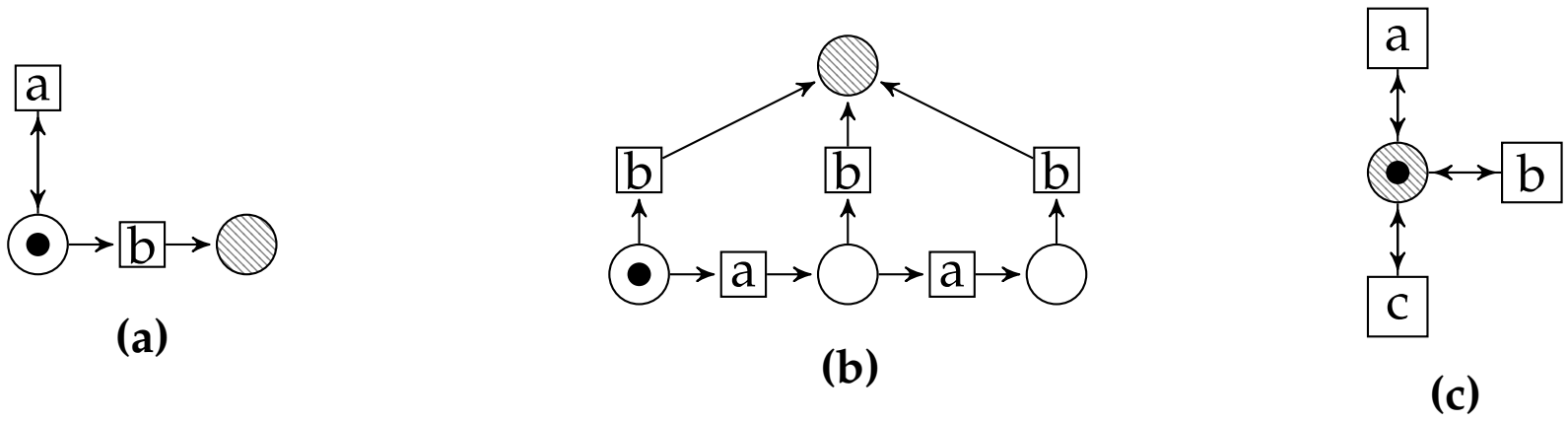

**Figure 3.10:** *(a)* A model with a length-one-loop, *(b)* the same model with the loop unrolled up to two executions. *(c)* A flower model over activities a, b and c.

PCC precision is implemented in ProM package ProjectedRecallAndPrecision[11]. PCC precision assumes the Petri net for which precision is calculated to be a bounded Petri net, i.e., it must have a finite number of tokens in every place for all reachable markings.

Consider log $L = [\langle a, b\rangle]$ and Petri nets $M_1$ and $M_2$ of Figures 3.10a and 3.10b respectively. $M_1$ starts with a length-one-loop on activity a, followed by activity b. $M_2$ unrolls the length-one-loop on activity a of $M_1$ to at most two executions, thereby limiting the behavior as it only allows at most two executions of activity a. It is easy to see that $M_1$ and $M_2$ are bounded Petri nets, as each place can have at most one token. For this log and these models, PCC precision results in $prec(L, M_1){=}0.6$, and $prec(L, M_2){=}0.5$. However, since $\mathfrak{L}(M_2){\subset}\mathfrak{L}(M_1)$, **A11** states that the precision of $M_2$ for fitting log L should be higher or equal to the precision of $M_1$. This shows that **A11** does not hold for PCC precision.

This drop in precision is an effect of the additional states that are created in the model DFA as an effect of unrolling the length-one-loop. The model DFA created from Petri net $M_2$ (Figure 3.10b) for example contains a state s that is reached after firing $\langle a, a\rangle$. This state however is never reached based on event log L, which only contains a trace $\langle a, b\rangle$, which has the effect that none of the enabled transitions from state s were observed in the log, bringing down the precision. In the DFA generated from Petri net $M_1$ (Figure 3.10a), this state s is merged with the state that one reaches after observing a single a event, as future behavior allowed by the model does not depend on the number of a-events seen.

Consider Petri net M of Figure 3.10c, and event logs $L_1{=}[\langle b, a, c\rangle, \langle a, a, c\rangle]$, and

[11]https://svn.win.tue.nl/trac/prom/browser/Packages/ProjectedRecallAndPrecision/

$L_2$=[⟨b, a, c⟩, ⟨a, a, c⟩, ⟨a, b, b, b, b, b, b, b, b, b, b, b, b, b, b, b⟩, ⟨b, a, a, a, a, a, a, a, a, a, a, a, a, a, a, a⟩]. The single place of M is bounded to one token, therefore M belongs to the class of bounded Petri nets. It is easy to see that $\tilde{L}_1 \subset \tilde{L}_2$, since the first two traces of log $L_2$ form log $L_1$. PCC precision results in $prec(L_1, M)=0.3125$ and $prec(L_2, M)=0.2727$, violating **A14**. The two traces of $L_2$ that are not in $L_1$ are very long traces to the traces that are in $L_1$, leading to additional states in the log automaton and the conjunction automaton. The additional states of the conjunction automaton have a low precision of $\frac{1}{4}$, since for each state the model allows for four options (firing activity a, b, c, or stopping), while only one is seen in the log. Therefore, if we would expand trace ⟨b, a, a, a, a, a, a, a, a, a, a, a, a, a, a⟩ with more events of activity a, then $prec(L_2, M)$ would approach $\frac{1}{4}$.

### 3.4.6 Anti-Alignment Precision

Van Dongen et al. [DCC16] proposed a precision measure based on *anti-alignments* [CC16], i.e., executions of the model that are as different as possible from the observed log. This notion of being maximally different is defined in terms of a distance measure, for which the authors propose to use Levenshtein distance. The precision measure consists of a weighted average between two components: *trace-based precision* and *log-based precision*. Trace-based precision of a trace $\sigma$ is based on the anti-alignment with a length of at most $|\sigma|$, while the log-based precision uses anti-alignments with a length of at most n, which is a parameter of the precision measure. The paper [DCC16] proposes to set $n = 2 \max_{\sigma \in L} |\sigma|$. This measure is implemented in the ProM package AntiAlignments[12].

Consider logs $L_k$ consisting of all traces $\langle a, (b,a)^*, b, c \rangle$ with up to k repetitions of $(b,a)^*$, i.e., $L_0 = [\langle a,b,c \rangle]$, $L_1 = [\langle a,b,c \rangle, \langle a,b,a,b,c \rangle]$, $L_2 = [\langle a,b,c \rangle, \langle a,b,a,b,c \rangle, \langle a,b,a,b,a,b,c \rangle]$, etc. Consider M shown in Figure 3.11. The anti-alignment of $L_0$ on M, using $n = 2 \max_{\sigma \in L} |\sigma| = 6$, is $\langle a,b,a,b,c \rangle$, as this is the longest trace that is allowed by M that is not in $L_0$. The Levenshtein distance between $\langle a,b,a,b,c \rangle$ and $\langle a,b,c \rangle$ is $\frac{2}{5}$, resulting in log-based precision $prec(L_0, M)=1-\frac{2}{5}=\frac{3}{5}$. For $L_1$, $n = 2 \max_{\sigma \in L} |\sigma| = 10$, allowing for the much longer anti-alignment $\langle a,b,a,b,a,b,a,b,c \rangle$, which has Levenshtein distance $\frac{4}{9}$ to closest log trace $\langle a,b,a,b,c \rangle$, therefore, $prec(L_1, M)=\frac{5}{9}$. Note that $prec(L_1, M) < prec(L_0, M)$, while $L_0 \subset L_1 \subset \mathfrak{L}(M)$, which shows that **A14** does not hold. For each log $L_k$, the longest trace has size 2k+3 and the size of the anti-alignment is $n=(2k+3) \cdot 2$, meaning that the smallest number of edit operations for Levenshtein distance to transform the anti-alignment into the longest trace consists of 2k+3 deletions. Therefore $\lim_{k \to \infty} prec(L_k, M)=\frac{1}{2}$, while in the limit, $L_k=\mathfrak{L}(M)$, for which we would expect a precision of 1.

[12] https://svn.win.tue.nl/trac/prom/browser/Packages/AntiAlignments/

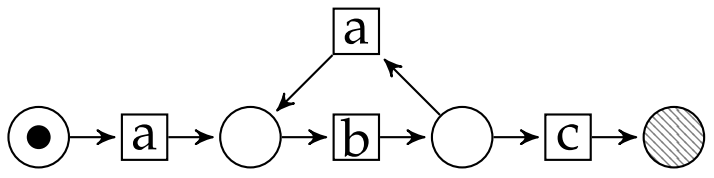

**Figure 3.11:** A model with $\mathfrak{L}(M) = [\langle a, b, c\rangle, \langle a, b, a, b, c\rangle, ...]$.

**Table 3.2:** An overview of the precision axioms and whether they hold (✓) or do not hold (✗) for each precision measure.

| Measure | A10 | A11 | A12 | A13 | A14 |
|---|---|---|---|---|---|
| Simple behavioral appropriateness | ✗ | | | ✗ | |
| Advanced behavioral appropriateness | ✗ | | ✗ | ✓ | |
| One-align ETC | ✗ | ✗ | ✗ | ✗ | ✗ |
| Negative Event Precision | ✗ | ✗ | | | |
| PCC precision | | ✗ | | | ✗ |
| Anti-alignment precision | | | | | ✗ |

### 3.4.7 Overview of the Properties of Precision Measures

We formulated five axioms that describe desirable properties for precision measures. Table 3.2 gives an overview of the axioms that we showed that do hold (✓) and that do not hold (✗) for each precision measure. We found that none of the existing precision measures fulfills all five axioms. Empty cells in the table are currently unknown, and no formal proof nor a counter example has been found that proves or disproves the axiom for the respective precision measure.

## 3.5 Contexts With Unclear Precision Requirements

The axioms introduced in Section 3.3 can be regarded as necessary conditions for precision measures, but they leave precision unspecified in some contexts. Figure 3.12a shows a situation in which $\tilde{L}\subseteq\mathfrak{L}(M_1)$, $\tilde{L}\subseteq\mathfrak{L}(M_2)$, but $\mathfrak{L}(M_1)\setminus\mathfrak{L}(M_2)\neq\emptyset$ and $\mathfrak{L}(M_2)\setminus\mathfrak{L}(M_1)\neq\emptyset$. In this setting, both $M_1$ and $M_2$ allow for (a possibly infinite amount of) different behavior that was not seen in L. Precision measures deal with this situation by quantifying the amount of behavior of $M_1$ and $M_2$. However, there are no obvious formal properties telling how the precision of $M_1$ and $M_2$ on L should relate.

Furthermore, all axioms define desired properties of precision measures when the event log L fits the behavior of the model M, i.e., $\tilde{L} \subseteq \mathfrak{L}(M)$. In practice, process discovery techniques will return process models with fitness below 1, i.e., there exists $\sigma\in L : \sigma\notin\mathfrak{L}(M)$. The discovery algorithm may deliberately abstract from infrequent behavior. Therefore, we do not formulate axioms for precision measures in the context of event logs that do not fit the process model, since we feel that there is not enough agreement in the process mining community on how a precision measure should behave in this context. Figure 3.12b shows an Euler diagram of a log L and two models $M_1$ and $M_2$ such that $\mathfrak{L}(M_1)\subset\mathfrak{L}(M_2)$ and $\tilde{L}\not\subseteq\mathfrak{L}(M_1)$, which is a non-fitting equivalent of **A11**. **A11** prescribes $prec(L,M_1)\geq prec(L,M_2)$, however, when the log does not fit the models, the behavior that fits $M_2$ but not $M_1$, $(\tilde{L}\setminus\mathfrak{L}(M_1))\cap\mathfrak{L}(M_2)$, makes it unclear how the precision of $M_1$ and $M_2$ should relate. Furthermore, even when $(\tilde{L}\setminus\mathfrak{L}(M_1))\cap\mathfrak{L}(M_2)=\emptyset$, it can be the case that the behavior in L that does not fit the models is behaviorally similar to the behavior of $M_2$.

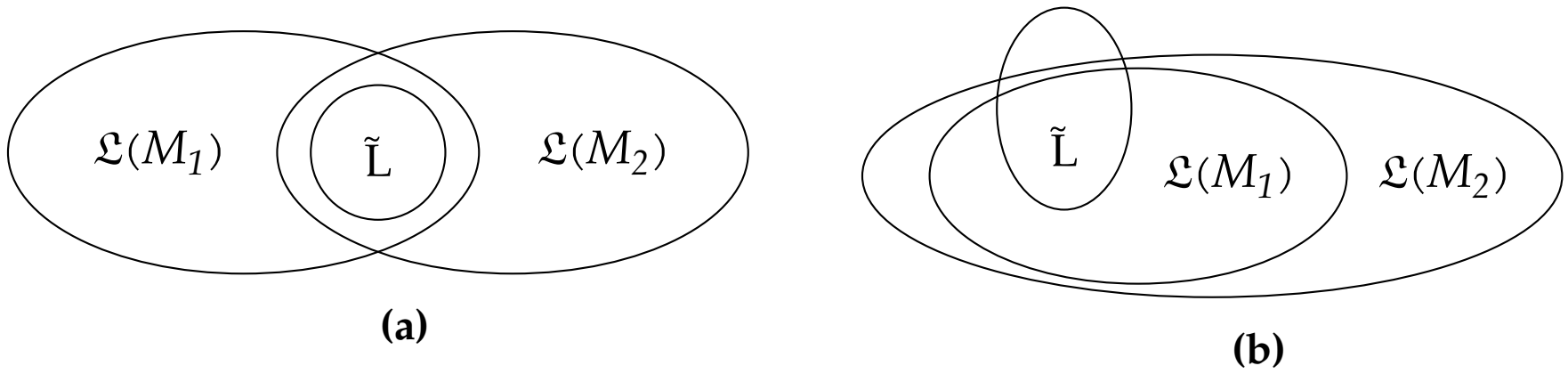

**Figure 3.12:** Two situations in which the desired properties of precision measures are unclear: *(a)* two models on which log L fits, with both models allowing for behavior that is not allowed by the other model, and *(b)* two models and a log on which the models do not fit.

In the remainder of the thesis, we will use both one-align ETC [Adr+12] and negative event precision, depending on the use case. One-align ETC precision is fast from a computational point of view and therefore suitable for large-scale experiments where the computation time of a precision measure matters. Even though one-align ETC fails to provide formal guarantees for any of the axioms, in practice, the value that it yields for most logs and models that are encountered in real-life situations can be regarded as at least indicative of the actual precision of the model, in the conceptual sense. Negative event precision is more computationally expansive on large models and logs than one-align ETC, and therefore less suitable for larger-scale experiments. While it violates **A10** and **A11**, the effect size with which it was found to violate those axioms is only minor, indicated by the fact that statistical testing was needed to demonstrate that it violates those axioms.

## 3.6 Conclusions

In this chapter, we have formulated axioms for fitness measures as well as for precision measures, in which we aim to formalize desired properties, properties that are widely believed to hold for existing measures, or properties that we believe are widely believed to be desirable by researchers from the process mining field. Surprisingly, we discovered that none of the existing precision measures fulfills all formulated axioms.

For fitness, we have shown that some desired properties that were previously stated by van der Aalst [Aal18] follow as a corollary from other properties. Furthermore, we have shown that one of the properties that were previously listed as a desired property for fitness, in fact, contradicts some of the conceptual notions of event-level fitness. We have formulated two alternative axioms for event-level fitness measures that rely on formulating a similarity measure between a trace and a process model. However, further discussion in the field is needed on the properties that are desired from such a similarity measure. Finally, we found that token-based fitness violates several axioms for fitness while alignment-based fitness fulfills all axioms except those that contradict with the notion of event-level fitness.

Based on the findings we will use alignment-based fitness to measure fitness throughout this thesis. To measure precision, we will use one-align ETC when fast computation of precision is required while we use negative event precision when the computation time of the measure is less of an issue.

# 4 Evaluating Label Refinements

Figure 4.1 visualizes the scope of this chapter by highlighting parts of the taxonomy of event log preprocessing methods. The first type of event log preprocessing that we will discuss in this part on event log preprocessing are *label refinements* and *label abstractions*. Label refinements address the task of splitting a label into more fine-grained subtypes, often with the aim of enabling the discovery of a high-quality process model. For example, in a labeled event log $L = [\langle a, b, c, a\rangle^{100}, \langle a, c, b, a\rangle^{100}]$ many process discovery algorithms will be unable to discover the behavior that the process starts with an a, is followed by b and c in parallel, and is ultimately followed by another a, because the majority of process discovery algorithms (we will discuss some exceptions in Section 4.5) is unable to discover a process model that contains multiple transitions with the same label. Simply relabeling L into $L' = [\langle a_1, b, c, a_2\rangle^{100}, \langle a_1, c, b, a_2\rangle^{100}]$ by changing every first a into $a_1$ and every second a into $a_2$ would enable many process discovery algorithms to discover a process model with the desired behavior of $a_1$, followed by b and c in parallel, and finally $a_2$. However, the approach of each n-th instance of activity a in a trace into $a_n$ does not seem to be a reasonable approach when the number of repetitions of an activity can be high, as this would explode the number of activities in the log and it fails to uncover patterns in the data in the case of loops[13]. Section 4.5.1 provides further discussion on the problems that occur with this simple type of relabeling.

In a parallel effort to the work on which this chapter is based [Tax+16a], Lu et al. [Lu+16a] developed a technique to refine the labels in a labeled event log purely based on the control-flow, i.e., on the ordering of event labels in the traces. Splitting labels purely on the control-flow context of the events leads to refined event labels that are not interpretable and take an alternative view. In contrast, this chapter discusses label refinements of a labeled event log L that are grounded in the data payload that can be found in the original unlabeled event log U from which L was originally generated. This leads to refined labels that are interpretable and can be explained in terms of data attributes of the event log. We introduce a technique based on statistical hypothesis testing to determine whether a given candidate for a label refinement is likely to be a useful one for the purpose of process discovery, and in Chapter 5 we propose an automated approach to refine the labels of an event log that bases refinements based on the timestamp of events in a way that makes use of specific properties of the time domain.

[13] Consider process tree $\rightarrow(\circlearrowright(a), \rightarrow(b, a))$ and traces $[\langle a, a, a, b, a\rangle, \langle a, a, b, a\rangle, \langle a, b, a\rangle]$ generated from it. The activities $a_2$ and $a_3$ in the relabeled traces $[\langle a_1, a_2, a_3, b, a_4\rangle, \langle a_1, a_2, b, a_3\rangle, \langle a_1, b, a_2\rangle]$ occur both before and after the b, thereby misleadingly steering a process discovery algorithm to believe that $a_2$ and $a_3$ are executed in parallel to b.

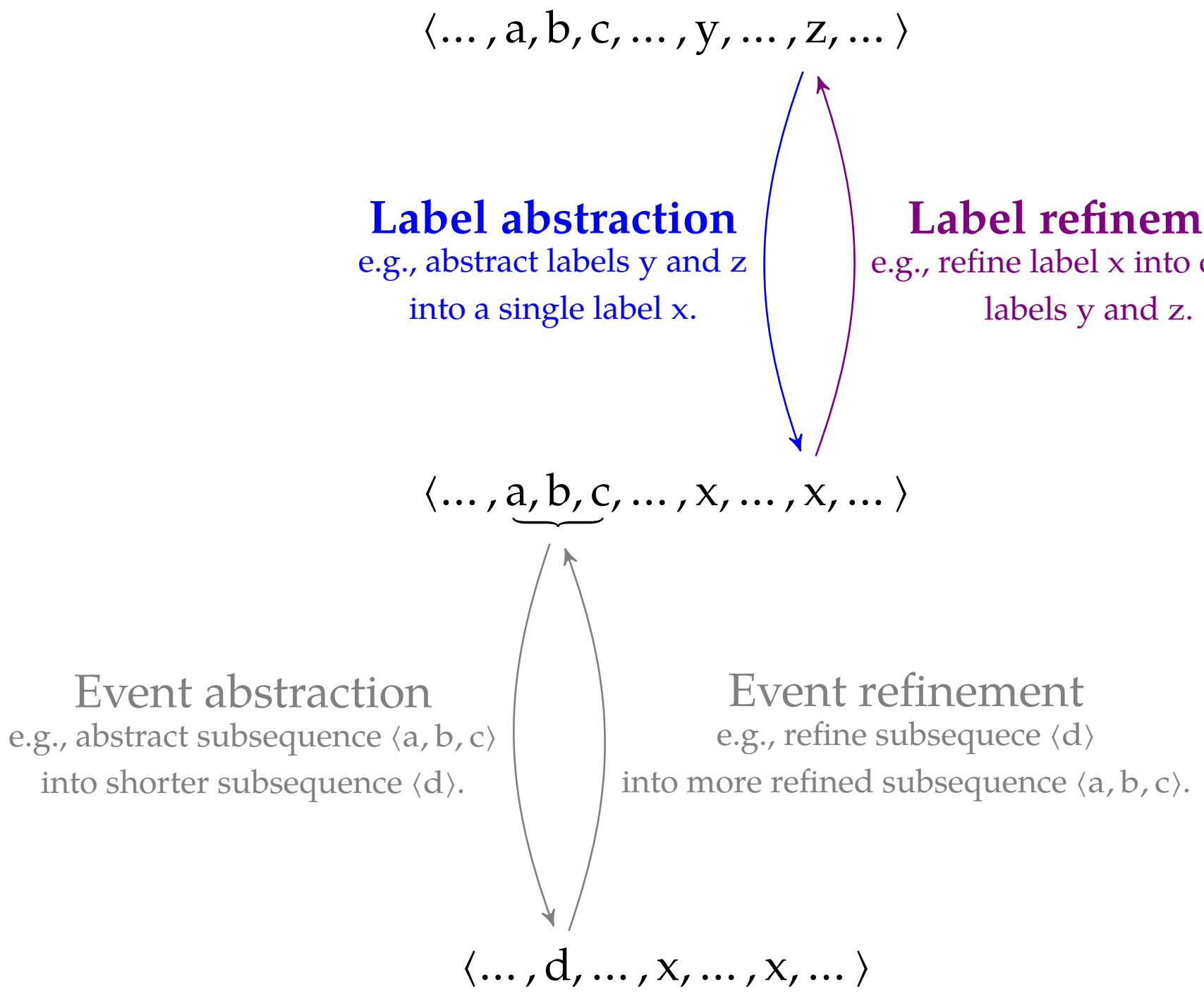

**Figure 4.1:** The positioning of this chapter in the taxonomy of event-log preprocessing methods. Colored preprocessing types are discussed in this chapter.

With the broadening of the process mining application domain in recent years from the original focus on business process management to a variety of domains where the event data does not originate from an underlying process model, the need for label refinements has increased. In many of those novel application domains, finding the right event labels of the events in the event log can be challenging. One of the application areas where finding the event labels is particularly challenging is the area of smart home environments, where the focus is on mining personal habits, routines, and frequent behavior, or on the detection of deviations from one's regular behavior. Finding the event labels of events consists of finding a *labeling function*[14] $l : \mathcal{C} \rightarrow \Sigma^*$ to transform smart home event data from its original *unlabeled event log*[15] form where the events are complex multi-dimensional objects to a *labeled event log*[16] where the events are simply an element from a finite set of event labels $\Sigma$. There are infinitely many labeling functions possible to create a labeled event log from an unlabeled event. The challenge is to find one that creates a labeled event log from which it is possible to mine a process model with process discovery techniques that unearth relations between activities.

As a concrete use case of the search for a suitable labeling function, imagine an elderly person of whom we want to discover a process model that describes his or her daily behavior. Events are generated by sensors, either periodically at regular intervals (e.g. by a temperature sensor or heart rate monitor), or triggered by some activity (e.g. motion in the proximity of a motion sensor, or sitting down on a chair with a pressure sensor). Table 4.1 shows an example of an unlabeled event log that is obtained by fusing the data from such sensors. The dots indicate that only a fraction of the logged events are shown. Assigning meaningful labels to these events is not straightforward. An event that was registered by the *bedroom motion* sensor can be caused by one out of multiple different human activities, for example, it could have been caused by a person *tossing & turning* in the bed at night, but it could also have been caused by *getting up*. For some application domains, it can be necessary to distinguish between the possible different underlying causes of the event. Differentiating between *tossing & turning* and *getting up* is, for example, necessary when we aim to generate a timely reminder to take medication that needs to be taken before breakfast. Based on contextual information (e.g., a specific increase in heart rate, a time stamp, etc.), the distinction between the two types of activities might be identified, and each event with label *bedroom motion* can be refined into either *tossing & turning* or *getting up*. The last column (i.e., *label*) in Table 4.1 shows some possible desired event labels. Figure 4.2 shows a process model that can be deduced from such a log using existing process discovery techniques.

One possible way would be to simply label each event with the sensor that generated it. This labeling function would, for example, result in events with labels: *bedroom motion*, *chair pressure*, *shower humidity*. However, these labels might be too coarse-grained to be used to mine insightful process models. Expert knowledge, data mining or machine learning techniques can be used to generate ideas for po-tential alternative labeling functions that refine this simple sensor-based labeling. The goal of a labeling function is to give "similar" events the same label. However, similarity is a relative notion, *so the initially chosen labeling function can be too abstract or too fine-grained to generate an informative process model*. Once a process discovery algorithm has been applied and a process model is obtained, one can assess whether the labeling function that was used actually allowed for the discovery of a process model that is both *fitting* and *precise*. However, instead of taking a trial-and-error approach of labeling an unlabeled event log in different ways and choosing the one from which the best process model can be discovered, in this chapter, we aim to find proper labeling functions *directly from the initial unlabeled event log*. The rationale behind a direct approach to finding a labeling function is that it prevents a computationally costly cycle of labeling, discover a process model, and evaluating that process model, that would otherwise be needed. The direct approach to discovering a labeling function that we explore in this chapter is rooted in statistics and uses statistical hypothesis testing to assess whether a candidate label refinement provides a benefit over another label refinement that more coarse-grained than this candidate.

[14] As defined in Definition 2.15
[15] As defined in Definition 2.14
[16] As defined in Definition 2.16

**Table 4.1:** Example event data originating from a smart home environment.

| Id | Timestamp | Address | Sensor | Heart rate | Label |
|---|---|---|---|---|---|
| 1 | 03/11/2015 02:45 | Mountain Rd. 7 | Bedroom motion | 74 | Tossing & turning |
| 2 | 03/11/2015 03:23 | Mountain Rd. 7 | Bedroom motion | 72 | Tossing & turning |
| 3 | 03/11/2015 04:59 | Mountain Rd. 7 | Bedroom motion | 71 | Tossing & turning |
| 4 | 03/11/2015 06:04 | Mountain Rd. 7 | Bedroom motion | 73 | Tossing & turning |
| 5 | 03/11/2015 08:45 | Mountain Rd. 7 | Bedroom motion | 85 | Getting up |
| 6 | 03/11/2015 09:10 | Mountain Rd. 7 | Living room motion | 79 | Living room motion |
| … | 03/11/2015 … | Mountain Rd. 7 | … | … | … |
| 7 | 03/12/2015 01:01 | Mountain Rd. 7 | Bedroom motion | 73 | Tossing & turning |
| 8 | 03/12/2015 03:13 | Mountain Rd. 7 | Bedroom motion | 75 | Tossing & turning |
| 9 | 03/12/2015 07:24 | Mountain Rd. 7 | Bedroom motion | 74 | Tossing & turning |
| 10 | 03/12/2015 08:34 | Mountain Rd. 7 | Bedroom motion | 79 | Getting up |
| 11 | 03/12/2015 09:12 | Mountain Rd. 7 | Living room motion | 76 | Living room motion |
| … | 03/12/2015 … | Mountain Rd. 7 | … | … | … |
| 12 | 03/13/2015 00:45 | Mountain Rd. 7 | Bedroom motion | 75 | Tossing & turning |
| 13 | 03/13/2015 02:29 | Mountain Rd. 7 | Bedroom motion | 75 | Tossing & turning |
| 14 | 03/13/2015 05:19 | Mountain Rd. 7 | Bedroom motion | 74 | Tossing & turning |
| 15 | 03/13/2015 05:34 | Mountain Rd. 7 | Bedroom motion | 79 | Tossing & turning |
| 16 | 03/13/2015 05:39 | Mountain Rd. 7 | Bedroom motion | 77 | Tossing & turning |
| 17 | 03/13/2015 08:37 | Mountain Rd. 7 | Bedroom motion | 79 | Getting up |
| 18 | 03/13/2015 08:52 | Mountain Rd. 7 | Living room motion | 78 | Living room motion |
| … | 03/13/2015 … | Mountain Rd. 7 | … | … | … |
| 19 | 03/14/2015 03:41 | Mountain Rd. 7 | Bedroom motion | 75 | Tossing & turning |
| 20 | 03/14/2015 05:00 | Mountain Rd. 7 | Bedroom motion | 74 | Tossing & turning |
| 21 | 03/14/2015 08:52 | Mountain Rd. 7 | Bedroom motion | 75 | Getting up |
| 22 | 03/14/2015 09:30 | Mountain Rd. 7 | Living room motion | 74 | Living room motion |
| … | 03/14/2015 … | Mountain Rd. 7 | … | … | … |
| 23 | 03/15/2015 02:11 | Mountain Rd. 7 | Bedroom motion | 77 | Tossing & turning |
| 24 | 03/15/2015 02:34 | Mountain Rd. 7 | Bedroom motion | 76 | Tossing & turning |
| 25 | 03/15/2015 08:35 | Mountain Rd. 7 | Bedroom motion | 79 | Getting up |
| 26 | 03/15/2015 08:57 | Mountain Rd. 7 | Living room motion | 77 | Living room motion |
| … | 03/15/2015 … | Mountain Rd. 7 | … | … | … |

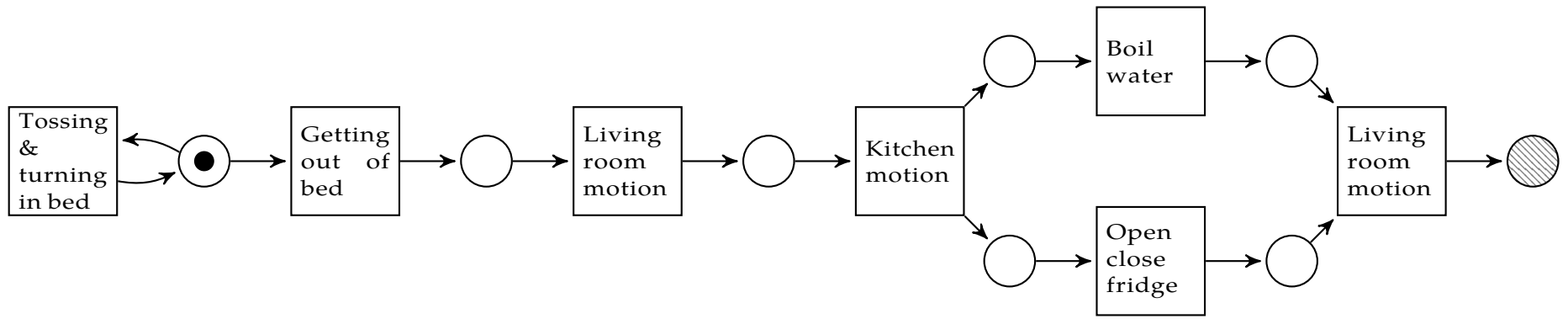

**Figure 4.2:** A Petri net derived from the event log in Table 4.1.

The remainder of this chapter is structured as follows. In Section 4.1 we formally introduce label refinements and related concepts. Section 4.2 provides a conceptual discussion on when a given label refinement is a useful one from the perspective of mining insights into the event data with process discovery. A statistical method to evaluate the usefulness of a label refinement is proposed in Section 4.3. Section 4.4 demonstrated the technique on real-life smart home data. Section 4.5 discusses the relation between label refinements and several existing related concepts in process mining. Section 4.6 provides concluding remarks.

## 4.1 Label Refinements

Table 4.1 shows an event set and, by partitioning the events into traces using a case notion, an unlabeled event log can be extracted from it. The partitioning that follows the horizontal lines in Table 4.1 provides an example of an unlabeled event that is created using a case notion function that groups events together in a trace when they occur on the same day. A naive way to create the labels for an unlabeled event log is to label each event by concatenating all of its attribute values from its data payload, i.e. $a_1, a_2, \dots, a_n \in \mathcal{A}_1 \times \mathcal{A}_2 \times \dots \times \mathcal{A}_n$. We refer to the event labeling function $l : \mathcal{C} \to \Sigma^*$ that labels events in this way as $l_{naive}$. For example, using $l_{naive}$ to label traces of the unlabeled event log shown in Table 4.1 resulting in the event e with $\pi(e)_{id} = 1$ being labeled as *"03/11/2015 02:45 Mountain Rd. 7 Bedroom motion 74"*. An alternative way to label event in an unlabeled event log is to label them based on the value of a single attribute from the data payload, which we refer to as $l_{attr}$, with *attr* the name of the attribute that is used to generate the labels. For example, event labeling function $l_{sensor}$, which labels events from a trace using the value of the sensor column, labels the event e with $\pi(e)_{id} = 1$ as simply *bedroom motion* when it is applied to the traces of Table 4.1.

Note that event labeling functions do not necessarily generate the label by simply only projecting the unlabeled log onto a subset of the attributes from the data payload. For example, one possible event labeling function would be to abstract the values of the heart rate attribute *low*, *normal*, and *high*, and label the event as such.

After this labeling step, some traces of the log can become identically labeled (the event id's would still be different). The information about the number of occurrences of a sequence of labels in an event log is highly relevant for process mining, since it allows differentiating between the mainstream behavior of a process (frequently occurring behavioral patterns) and exceptional behavior.

Let $\mathcal{C}$ be a universe of unlabeled traces over some set of data attributes and let $l_1 : \mathcal{C} \to \Sigma_1^*$ and $l_2 : \mathcal{C} \to \Sigma_2^*$ be two labeling functions that, when applied to an unlabeled event log $U \subseteq \mathcal{C}$, result in labeled event logs $L_1 = l_1(U)$ and $L_2 = l_2(U)$ such that $L_1 \subseteq \mathcal{B}(\Sigma_1^*)$ and $L_2 \subseteq \mathcal{B}(\Sigma_2^*)$ where $\Sigma_1 \neq \Sigma_2$, i.e., the two labeling function assign different labels to the events of U.

**Definition 4.1 (Labeled Refinement).** Labeling function $l_1$ is a *label refinement* of labeling function $l_2$, denoted by $l_1 \preceq l_2$, if and only if $\forall_{\sigma_1,\sigma_2 \in \mathcal{C}} : l_1(\sigma_1) = l_1(\sigma_2) \implies l_2(\sigma_1) = l_2(\sigma_2)$. ◇

An example of a label refinement can be found in Table 4.1, where a labeling function $l_1$ that labels according to the *label* column is a refinement of a label refinement $l_2$ that labels according to the *sensor* column. This is easy to see, as all events that are labeled *bedroom motion* by $l_2$ are labeled either *tossing & turning* or *getting up* by $l_1$, and therefore traces that are identical under $l_1$ are also identical under $l_2$. However, the inverse does not hold, as traces that are identical under $l_2$ are not necessarily identical under $l_1$. This is easy to see, as the *bedroom motion* events can be labeled into different labels by $l_1$.

When a labeling function $l_1$ is a *refinement* of a labeling function $l_2$, then we call $l_2$ a *label abstraction* of $l_1$. We call a refinement $l_1$ of $l_2$ a *strict label refinement*, denoted by $l_1 \prec l_2$, when $(\exists_{\sigma_1,\sigma_2 \in \mathcal{C}} : l_1(\sigma_1) \neq l_1(\sigma_2) \wedge l_2(\sigma_1) = l_2(\sigma_2)) \wedge (\forall \sigma \in \mathcal{C} : |l_1(\sigma)| = |l_2(\sigma)|)$.

For some universe of unlabeled traces $\mathcal{C}$ over some set of data attributes, for a some labeling function $l : \mathcal{C} \to \Sigma^*$, and for some labeled event log $L = l(U)$ for $U \subseteq \mathcal{P}(\mathcal{C})$, an *unlabeled traces function* $l^{-1} : \Sigma \to \mathcal{C}$ is defined as $l^{-1}(\sigma) = \{\sigma' \in \mathcal{C} \mid l(\sigma') = \sigma\}$. Intuitively, an unlabeled traces function returns the set $S \subseteq \mathcal{C}$ of all unlabeled traces from the universe of unlabeled traces C that would have been labeled into labeled trace $\sigma$ by labeling function l. *Language concretization* refers to the unlabeled traces function of a set of labeled traces $\mathfrak{L}$, i.e. $l^{-1}(\mathfrak{L}) = \bigcup_{\sigma \in \mathfrak{L}} l^{-1}(\sigma)$.

**(a)** Process model $M_1$

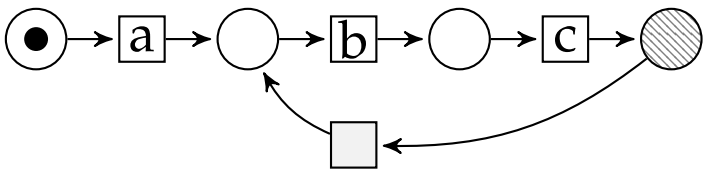

**(b)** Process model $M_2$

**Figure 4.3:** Two process models that are discovered from two different labeled event logs that are created by applying to different labeling functions to the same unlabeled event log, where the labeling function that generated the labeled log from which $M_2$ was discovered is a strict refinement of the labeling function that generated the labeled log from which $M_1$ was discovered.

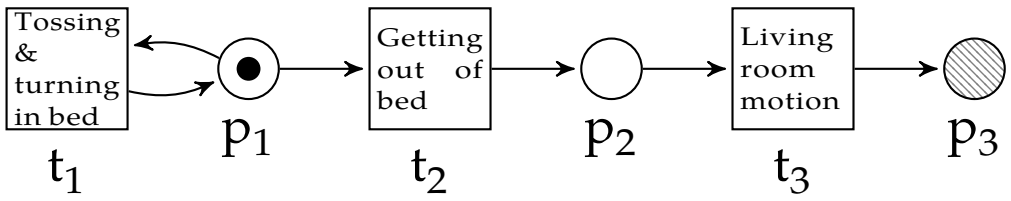

**(a)** Process model $M_1$

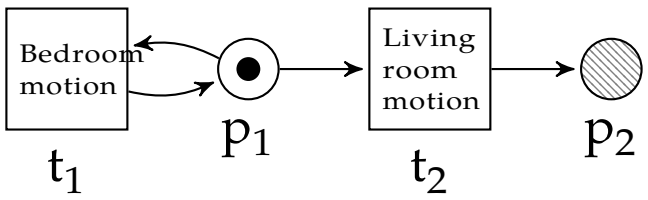

**(b)** Process model $M_2$

**Figure 4.4:** Another example of two process models that are discovered from two different labeled event logs that are created by applying to different labeling functions to the same unlabeled event log. In this case, the two labeled event logs are extracted from the unlabeled event log in Table 4.1.

## 4.2 On the Quality of Label Refinements for Process Mining

*Process discovery algorithms* discover a process model based on a labeled event log, where event labels are obtained by applying a labeling function to an unlabeled event log. The quality of the output of process discovery is generally expressed in terms of *fitness* and *precision*, as we have discussed in detail in Chapter 3. Therefore, intuitively, one event labeling function would be better than another one it improves the quality of the discovered model along these quality dimensions, for a given process discovery algorithm. However, the existing fitness and precision measures are defined in such a way that it can only be used to compare event logs and process models over *the same set of labels* while applying two different labeling functions to the same unlabeled event log can result in two labeled event logs that have *different*

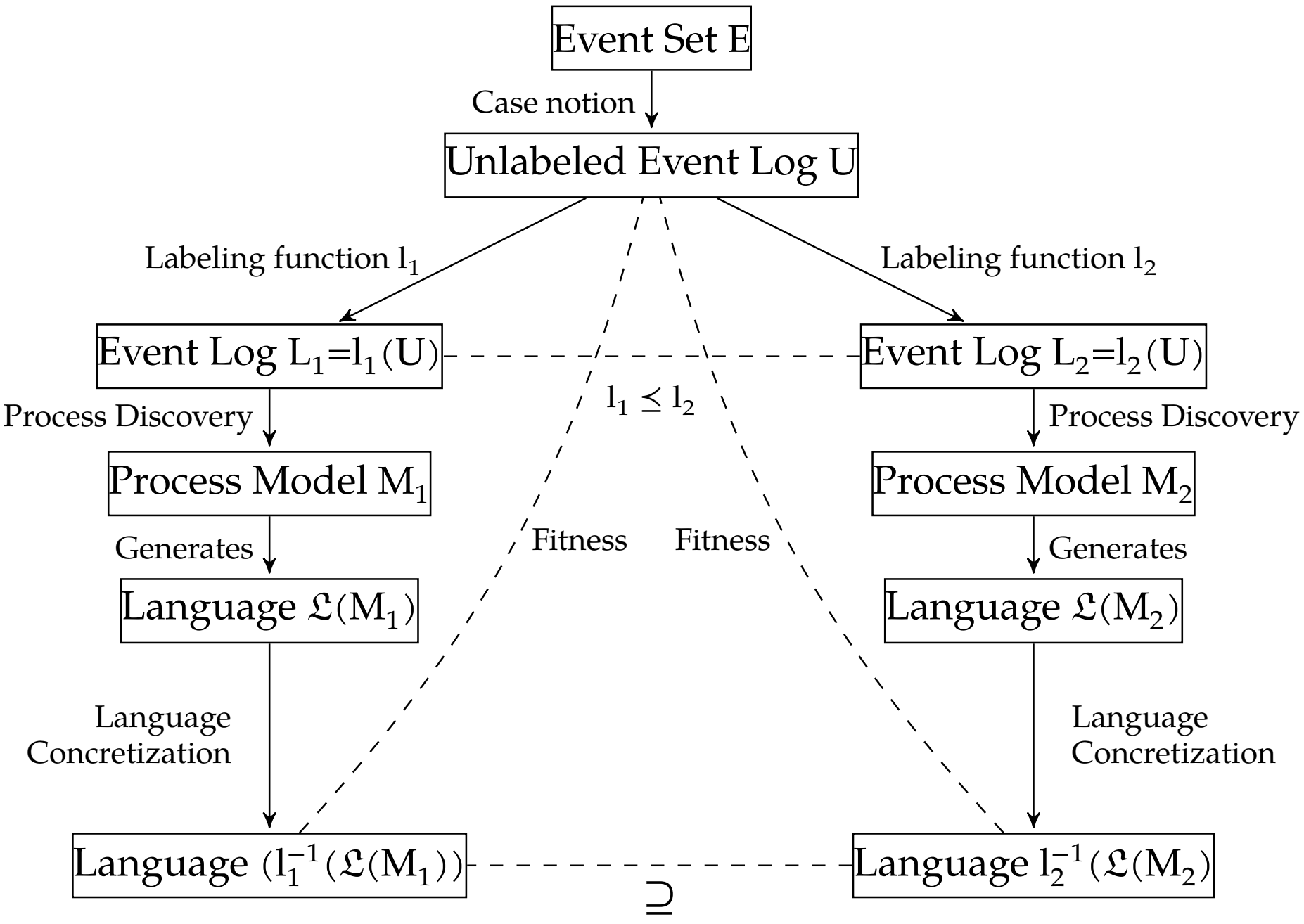

**Figure 4.5:** Comparing two labeling functions

*labels*. As a result, process discovery will on those two labeled event logs result in process models that describe different sets of activities from those two labeled event logs.

To see that the quality in terms of fitness and precision cannot be compared between different labeling functions, consider an unlabeled event log $U = \{\langle e_1, e_2, e_3, e_4, e_5\rangle, \langle e_6, e_7, e_8\rangle\}$ and two labeling functions, $l_1$ and $l_2$, such that $l_1(U) = [\langle a, a, a, a, a\rangle, \langle a, a, a\rangle]$ and $l_2(U) = [\langle a, b, c, b, c\rangle, \langle a, b, c\rangle]$. It is easy to see that labeling function $l_2$ is a strict label refinement of labeling function $l_1$. Figure 4.3 shows process models $M_1$ and $M_2$ that are respectively discovered from labeled event logs $l_1(U)$ and $l_2(U)$. Note that labeled event log $l_1(U)$ is actually fitting on $M_1$ (as $l_1(U) \subseteq \mathfrak{L}(M_1)$) and $M_1$ is quite precise with respect to $l_1(U)$. On the other hand, $l_2(U)$ is fitting on $M_2$ (as $l_2(U) \subseteq \mathfrak{L}(M_2)$) and $M_2$ is precise with respect to $l_2(U)$. From a point-of-view of fitness and precision measures, there seems to be not much difference between the quality of the two discovered process models $M_1$ and $M_2$. However, from the point of view of insights in the behavior that is going on in the event data, $M_2$ is much more informative than $M_1$.

To give a more concrete example, consider the unlabeled event log U of Table 4.1 and furthermore consider labeled function $l_1$ such that each event is labeled with the value of the *sensor* column and $l_2$ such that each event is labeled using a function that is such that its resulting labels are shown in the *label* column. Again, $l_2$ is a strict refinement of $l_1$. Process model $M_2$ in Figure 4.4a has perfect precision and fitness for the labeled event log that results from $l_2$. At the same time, process model $M_1$ in Figure 4.4b has perfect fitness and precision for the labeled event log that results from $l_2$. Process model $M_1$ is more specific than $M_2$ in the sense that it shows that *Getting out of bed* only occurs once a day while $M_2$ does not show this since the more coarse-grained labeled function $l_2$ does not differentiate between tossing and turning in bed and getting out of bed, labeling both types of events as *bedroom motion*. In the scenario of a reminder system for taking medication after getting out of bed that we introduced at the beginning of this chapter, $M_1$ is a useful process model for the purpose of sending a reminder message to take medicines after getting up, while Petri net $M_2$ is not. This suggests that Petri net $M_1$ is more useful than $M_2$, even though this is not reflected in fitness and precision.

4

As the examples in Figure 4.3 and Figure 4.4 have shown that label refinements cannot be evaluated in terms of fitness and precision on the resulting event log and process model, we have to make the comparison in the context of the unlabeled event log. Figure 4.5 visualizes this comparison. Suppose we have a set of events E, which is part of some universe of events $\mathcal{E}$. We choose a case identifier and build an unlabeled event log U from E. For two labeling functions $l_1$ and $l_2$ such that $l_1 \prec l_2$ we obtain two labeled event logs, $L_1 = l_1(U)$ and $L_2 = l_2(U)$. Applying process discovery to $L_1$ and $L_2$ results in two process models $M_1$ and $M_2$ respectively, which respectively accept languages $\mathfrak{L}(M_1)$ and $\mathfrak{L}(M_2)$. As discussed, these languages cannot be compared directly since they contain traces consisting of different event labels and precision measures asses the redundant behavior the discovered process models by using the event labels in the labeled event log that was used as input for the discovery algorithm.

By using the inverse functions $l_1^{-1}$, $l_2^{-1}$, every trace of $\mathfrak{L}(M_1)$ and every trace of $\mathfrak{L}(M_2)$ can be mapped to a set of traces that are built from the events from event universe $\mathcal{E}$. By taking the union of the sets obtained with $l_1^{-1}$ and $l_2^{-1}$ over the traces of the languages $\mathfrak{L}(M_1)$ and $\mathfrak{L}(M_2)$, i.e. $(l_1^{-1}(\mathfrak{L}(M_1))$ and $l_2^{-1}(\mathfrak{L}(M_2)))$, we obtain two comparable languages and it becomes theoretically possible to assess whether there is a gain in precision when we consider the combination of both the labeling function and the process discovery step. This contrasts existing precision measures which only measure the precision of the result of the process discovery step while assuming a given fixed labeling function.

The fitness and simplicity of the models depend mostly on the performance of the process discovery algorithm, and not on the choice of the labeling function. The precision that is defined in terms of events of the original universe $\mathcal{E}$ of events is however highly dependent on the appropriateness of the labeling function: choosing a more refined labeling function can increase the precision by eliminating the behavior that would be allowed in the model discovered with a more abstract labeling function.

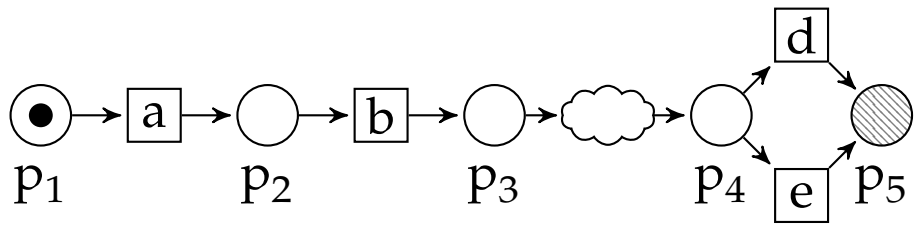

**(a)** Process Model $M_1$.

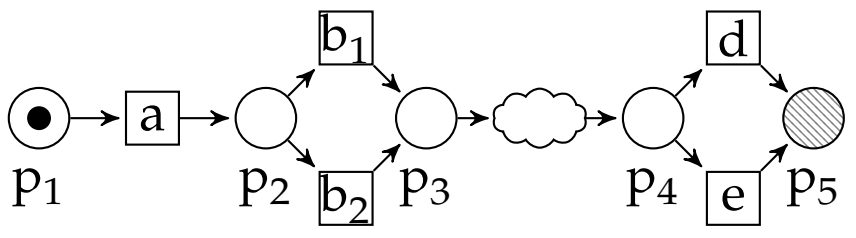

**(b)** Process Model $M_2$.

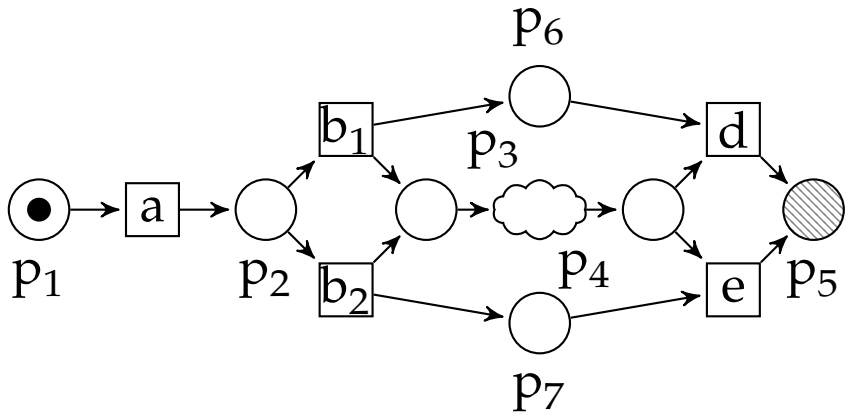

**(c)** Process Model $M_3$.

**Figure 4.6:** $M_2$ is a non-useful refinement and $M_3$ is a useful refinement of $M_1$.

### 4.2.1 Label Refinement Quality

The comparison of the languages generated by models is not feasible due to its complexity; for many classes of process models, including Petri nets, the problem of language inclusion is just not decidable [Hac64; Hir93]. Therefore, we need a different, practical approach to decide on the usefulness of switching from one labeling function $l_1$ to another labeling function $l_2$ that is a refinement of $l_1$. We start with discussing the usefulness of a label refinement conceptually by discussing a process model that is discovered from a log that is labeled with some labeling function and two alternative process models that are respectively discovered from the same log with alternative labeling functions that are refinements.

Consider event log L and labeling functions $l_1, l_2$, and $l_3$ such that $l_2 \prec l_1 \land l_3 \prec l_1$, and the resulting labeled event logs $L_1 = l_1(L), L_2 = l_2(L), L_3 = l_3(L)$. Let the $M_1, M_2, M_3$ in Figure 4.6 be the Petri nets obtained by applying process discovery to $L_1, L_2, L_3$ respectively. The cloud shape between places $p_3$ and $p_4$ indicates there is a subprocess in the middle of the Petri net that is abstracted away because it is not relevant to the usefulness of label refinements $l_2$ and $l_3$.

We can see that refinement $l_2$ does not lead to a more meaningful interpretation of the events that were labeled by $l_1$ as b (now labeled $b_1$ and $b_2$), since the behavior of the model is not related to the choice between $b_1$ and $b_2$. The transitions labeled with $b_1$ and $b_2$ have the same input and output places, and therefore, refinement $l_2$ does not provide new insights and unnecessarily harms the understandability of the Petri net by creating more transitions then needed. On the other hand, $l_3$ results in a behaviorally more specific model after applying process discovery. To see that this is the case, consider that $\mathfrak{L}(M_3)$ does not contain $\langle a, b_1, \ldots, e\rangle$ and $\langle a, b_2, \ldots, d\rangle$, while in contrast, $M_1$ does not distinguish between $b_1$ and $b_2$ and allows for $\langle a, b, \ldots, d\rangle$, which suggests that both types of traces are possible. Note that even though we have seen that the model that we discovered after applying $l_3$ (i.e., $M_3$) is behaviorally more specific than the model that we discovered after applying $l_1$ (i.e., $M_1$), this gain in "specificness" of the process model description will not be visible quantitatively when comparing the two models using one of the existing precision measures that we have discussed in detail in Chapter 3. The reason for this is that the languages of the two process models are defined in terms of different event labels. The precision of $M_1$ can only be measured with respect to the log generated by $l_1$, however, since this log does not distinguish between $b_1$ and $b_1$ we can say that its labels are imprecise in comparison to the log generated by $l_3$. When measuring the precision of $M_1$ on the log generated by $l_1$ this imprecision in the labels of $l_1$ is be accounted for.

## 4.3 An Evaluation Method for Label Refinements for Process Models

In the previous section, we have shown that we can compare the usefulness of refining a given labeling function $l_1$ to a given other labeling function $l_2$ such that $l_2 \preceq l_1$ by inspecting the Petri net obtained with process discovery. A naive way to evaluate a label refinement would be to apply process discovery to all possible label refinements. However, when the labeling function $l_1$ that was initially there creates a large number of unique labels, i.e., when $|cod(l_1)|$ is large, the many possible candidates for label refinements results in a combinatorial explosion of possibilities, therefore yielding this approach infeasible. We will now present an alternative approach that enables efficient estimation of the usefulness of a label refinement based on statistics and log relations without discovering the process model for each possible label refinement.

Algorithm 1 shows the steps of the label refinements evaluation method. The evaluation method consists of an entropy-based component that measures whether a label refinement makes the log statistics more unbalanced and a statistical test that tests whether there is a label statistic that tests whether the label refinement makes a statistically significant difference to at least one of the log statistics. In the following two sections we described the entropy-based measure and the statistical testing respectively.

**Table 4.2:** An overview of the use of log-based ordering relations (R) and statistics (S) in process discovery algorithms (ordered chronologically).

| Algorithm | Direct successor | Length-1-loop | Length-2-loop | Indirect successor |
|---|---|---|---|---|
| $\alpha$ miner [Aal+04] | R | | | |
| $\alpha^{+}$ [Med+04] | R | R | R | |
| $\alpha^{++}$ [WWS06] | R | R | | R |
| $\alpha^{\#}$ [Wen+10] | R | R | R | |
| $\alpha^{\$}$ [Guo+15] | R | R | R | R |
| Multiphase Miner [DA04] | R | R | R | |
| AGNES Miner [Goe+09] | S | S | | S |
| Heuristics miner [WR11] | S | S | S | S |
| Inductive Miner [LFA13b] | R | | R | |
| Inductive Miner infrequent [LFA13a] | S | | S | |
| Fodina [BD17] | S | S | S | S |
| Split Miner [Aug+17; Aug+18] | S | S | S | |

## 4.3.1 Log Relations and Statistics

Event ordering relations and statistics play an important role in many most process discovery algorithms as an intermediate step in going from event log to process model. *Ordering relations* express a relation between labels in a labeled event log. *Ordering statistics*, in addition to ordering relations, quantify the frequency of the occurrence of an ordering relation. Table 4.2 provides an overview of some well-known process discovery algorithms and the event log relations and statistics that they internally make use of. For each process discovery algorithm and each relation type the table contains an **R** when the relation is used in the discovery algorithm and it contains a **S** when it uses the corresponding statistic, i.e., when it uses the *counts* of how many events fulfill the relation. Let L be a labeled event log and let $a \in \Sigma$ and $b \in \Sigma$ be labels from the labeling function $cod(l)$ with which L was created. Formal definitions of these log-based ordering statistics are as follows:

- Relation $R^{L}_{df} \subseteq \Sigma \times \Sigma$, called *direct successor* or *directly-follows relation*, contains the pairs of labels such that $(a, b)$ in labeled log L such that $(a, b) \in R^{L}_{df} \iff \exists_{\sigma \in L, i \in \mathbb{N}} : \sigma(i) = a \wedge \sigma(i+1) = b$, i.e., the pairs of labels $a, b$ such that a is directly followed by b. $\overline{R}^{L}_{df} = (\Sigma \times \Sigma) \setminus R^{L}_{df}$ denotes the complement of $R^{L}_{df}$, i.e., the pairs $a, b$ such that a is never directly followed by b. Analogous to the directly-follows *log relation*, the directly-follows *log statistic* $\#^{L}_{df}(a, b)$ denotes the *number of occurrences* of a events in the traces of L that are directly followed by b and $\overline{\#}^{L}_{df}(a, b)$ denotes is the number of occurrences of a that are *not* directly followed by b. Observe that the *length-one loop* relation is a special case of the directly-follows relation where for the label pair $(a, b)$ we have $a = b$.

**Table 4.3:** A log statistic $s \in \{df, l2l, ef\}$ in contingency table form for logs $L_1 = l_1(U)$, $L_2 = l_2(U)$, and $l_2$ is a refinement of $l_1$ that splits a into $a_1$ and $a_2$ with respect to b.

| | $a_1$ | $a_2$ | $a$ |
|---|---|---|---|
| + | $\#_s^{L_2}(a_1, b)$ | $\#_s^{L_2}(a_2, b)$ | $\#_s^{L_1}(a, b)$ |
| − | $\overline{\#}_s^{L_2}(a_1, b)$ | $\overline{\#}_s^{L_2}(a_2, b)$ | $\overline{\#}_s^{L_1}(a, b)$ |

- Relation $R^L_{l2l} \subseteq \Sigma \times \Sigma$, called *length-two loop*, contains the pairs of labels such that $(a, b) \in R^L_{l2l} \iff \exists_{\sigma \in L, i \in \mathbb{N}} : \sigma(i) = a \land \sigma(i+1) = b \land \sigma(i+2) = a \land a \neq b$, i.e., the pairs of labels such that a is directly followed by b which in turn is again directly followed by a. In some of the process mining literature this relation is referred to as $R^+_{L,l2l}$. $\overline{R}^L_{l2} = (\Sigma \times \Sigma) \setminus R^L_{l2l}$ denotes the complement of $R^L_{l2l}$, i.e., the pairs a, b such that a is not in a length-two loop with b. Log statistics $\#^L_{l2l}(a, b)$ and $\overline{\#}^L_{l2l}(a, b)$ denote the number of occurrences of a that are, respectively, are not, followed by b;
- Relation $R^L_{ef} \subseteq \Sigma \times \Sigma$, called *indirect successor* or *eventually follows*, contains the pairs of labels such that $(a, b) \in R^L_{ef} \iff \exists \sigma \in L, i \in \mathbb{N}, j \in \mathbb{N} i < j \land \sigma(i) = a \land \sigma(j) = b$, i.e., the pairs of labels such that a is followed by b, but not necessarily directly. Relation $\overline{R}^L_{ef} = (\Sigma \times \Sigma) \setminus R^L_{ef}$ is the complement of $R^L_{ef}$ i.e., it contains the pairs of labels such that a is never followed by b. Log statistics $\#^L_{ef}(a, b)$ and $\overline{\#}^L_{ef}(a, b)$ denote the number of occurrences of a that are, respectively are not, eventually followed by b: for a trace $\sigma \in L, i, j \in \mathbb{N}$ with $i < j$, $\sigma(i) = a$ and $\sigma(j) = b$.

In the general sense, let $\#^L_s(a, b)$ and $\overline{\#}^L_s(a, b)$ denote the count of the number of a-events in labeled log L that *do*, respectively *do not*, satisfy relation $s \in \{df, l2l, ef\}$ with respect to b.

Let U be an unlabeled event log. Let $l_1$ and $l_2$ be two labeling functions with $l_2 \prec^= l_1$ for which we aim to determine whether $l_2$ is a *useful* refinement of $l_1$. Let $L_1 = l_1(L)$ and $L_2 = l_2(L)$. Let $l_1$ and $l_2$ have the property $\{a_1, a_2 \in cod(l_2)) | \exists_{\sigma_1, \sigma_2 \in L} : l_1(\sigma_1) = \lambda \cdot a \land l_1(\sigma_2) = \lambda' \cdot a \land l_2(\sigma_1) = \zeta \cdot a_1 \land l_2(\sigma_2) = \zeta' \cdot a_2\} \neq \emptyset$, that is, $l_2$ refines activity a into distinct activities $a_1$ and $a_2$. The difference in control flow between $a_1$ and $a_2$ can be expressed as the dissimilarity in log-based ordering statistics between event label $a_1$ and $b \in cod(l_2) \setminus \{a_1, a_2\}$ on the one hand, and $a_2$ and b on the other hand.

Each log-based ordering statistics of $a_1$ and $a_2$ with regard to any other activity b can be formulated in the form of a contingency table, as shown in Table 4.3.

## 4.3.2 Information Gain

The binary entropy function, $H_b(p) = -p \log_2 p - (1-p) \log_2 (1-p)$, is a measure of uncertainty, where the edge case $0 \log_2 0$ is taken to be 0. When it is applied to a log statistic, the binary entropy function represents a degree of nondeterminism. Non-deterministic, unbalanced, log statistics are helpful to process discovery algorithms that operate of log statistics, as it provides low uncertainty to the mining algorithm. Low entropy in the log statistics indicates high predictability of the process, making it easier for process discovery algorithms to return a sensible process model.

Consider the contingency tables in Table 4.4, based on log statistics obtained from Table 4.1 between the events labeled *tossing & turning* and *getting up* and the events labeled *living room motion*. On the right-hand side of the table, separated by the bar, are the log statistics of the before-split label in the before-split log. All five events with label *getting up* directly precede an event with label *living room motion*, while all sixteen events with label *tossing & turning* are *not* directly preceded by *living room motion*. Furthermore, all events with refined labels do *not* directly or eventually follow an event with label *living room motion*, and all events with refined labels do eventually precede an event with label *living room motion*.

Log statistics with a high degree of non-determinism, like the directly precedes statistic of the bedroom motion events before the split, might confuse a mining algorithm as there is no clear structure here: the *bedroom motion* event might directly precede *living room motion*, but most of the time it does not. After the split we see a completely deterministic directly precedes statistic, where *tossing & turning* never and *getting up* always directly precedes *living room motion*. This increased determinism is reflected by the entropy of the directly precedes statistic before and after the split. Before the split we have the following entropy (in bits) in the directly-precedes statistic:

$$-\tfrac{5}{5+16} \log_2 \tfrac{5}{5+16} - \tfrac{16}{5+16} \log_2 \tfrac{16}{5+16} = 0.7919$$

After applying the split, the entropy of the directly-precedes statistic reduces to the following for the *tossing & turning* events:

$$-\tfrac{0}{0+16} \log_2 \tfrac{0}{0+16} - \tfrac{16}{0+16} \log_2 \tfrac{16}{0+16} = 0$$

Furthermore, after the split, the entropy reduces to the following for the *getting up* events:

$$-\tfrac{0}{0+5} \log_2 \tfrac{0}{0+5} - \tfrac{5}{0+5} \log_2 \tfrac{5}{0+5} = 0$$

**Table 4.4:** Contingency tables for comparing the behavior of the two refined labels with respect to *Living room motion*.

| | Tossing & turning | Getting up | Bedroom motion |
|---|---|---|---|
| $\overset{+}{\rightarrow}$ | 0 | 0 | 0 |
| $\overset{-}{\rightarrow}$ | 16 | 5 | 21 |

**(a)** Directly follows

| | Tossing & turning | Getting up | Bedroom motion |
|---|---|---|---|
| $\overset{+}{\rightarrow}$ | 0 | 5 | 5 |
| $\overset{-}{\rightarrow}$ | 16 | 0 | 16 |

**(b)** Directly precedes

| | Tossing & turning | Getting up | Bedroom motion |
|---|---|---|---|
| $\overset{+}{\rightarrow}$ | 0 | 0 | 0 |
| $\overset{-}{\rightarrow}$ | 16 | 5 | 21 |

**(c)** Eventually follows

| | Tossing & turning | Getting up | Bedroom motion |
|---|---|---|---|
| $\overset{+}{\rightarrow}$ | 16 | 5 | 21 |
| $\overset{-}{\rightarrow}$ | 0 | 0 | 0 |

**(d)** Eventually precedes

The conditional entropy of the log statistic after the split is the weighted average of the entropy of the labels created in the split (i.e., of the entropy of *tossing & turning* and of *getting up*), which is $-\frac{16}{21} \times 0 \times \frac{5}{21} \times 0 = 0$.

The information gain of this label split with regard to the directly precedes *living room motion* statistic is equal to the total entropy of the log statistic prior to the split, minus the conditional entropy after the split, this $0.7919 - 0 = 0.7919$. Relative information gain [KL51] is a metric that provides insight into the ratio of bits of entropy reduced by a label refinement and can be calculated by dividing the information gain by the before-split entropy. The relative information gain of the directly precedes *living room motion* statistic is $\frac{0.7919}{0.7919} = 1$. Figure 4.4 shows the effect of this label refinement on the resulting Petri net obtained by process discovery.

So far we have calculated the Relative information gain for a single log statistic. A single label refinement, however, can impact multiple log statistics at the same time. We need a measure that integrates the information gain values of all log statistics to express the quality of a label refinement with respect to the determinism of the log statistics. Therefore, we sum over the entropy of all log statistics before the label split to obtain the total before-split entropy. We sum over the conditional entropies of all log statistics after the label split to obtain the total after-split entropy. Information Gain and Relative information gain are calculated as before. We let $\text{relative_information_gain}(L_1, L_2)$ be the function that returns the Relative information gain based on the pre-split log $L_1$ and post-split log $L_2$, where the set of refined label pairs in $L_2$ from which the log statistics are used corresponds to $\{a_1, a_2 \in cod(l_2) \mid \exists_{\sigma_1, \sigma_2 \in U} : l_2(\sigma_1) = \lambda \cdot a_1 \wedge l_2(\sigma_2) = \lambda' \cdot a_2 \wedge l_1(\sigma_1) = \lambda \cdot a \wedge l_1(\sigma_2) = \lambda' \cdot a\}$, where a is the corresponding label in $L_1$ that is refined into $a_1$ and $a_2$.

**Input:** Unlabeled Event log U, labeling functions $l_1$ and $l_2$ (such that $l_2 \prec^{=} l_1$),
Set of log-based ordering statistics S, Significance level $\alpha$
**Output:** The Relative information gain of $l_1$ w.r.t $l_2$
*Initialisation* :
*all_significant_different* = *true*; $L_1 = l_1(U); L_2 = l_2(U)$;
*split_set* = $\{a_1, a_2 \in cod(l_2) | \exists_{\sigma_1,\sigma_2 \in U} : l_1(\sigma_1) = \lambda \cdot a \wedge l_1(\sigma_2) = \lambda' \cdot a \wedge l_2(\sigma_1) = \zeta \cdot a_1 \wedge l_2(\sigma_2) = \zeta' \cdot a_2\}$;
*Main Procedure:*
**for** $a_1, a_2 \in$ *split_set* **do**
  *pair_significant_different* = *false*;
  **for** $\{b \in (cod(l_2) \setminus \{a_1, a_2\})\}$ **do**
    **for** $s \in S$ **do**
      p = *fisher_test*$(\#_s^{L_2}(a_1, b), \overline{\#}_s^{L_2}(a_1, b), \#_s^{L_2}(a_2, b), \overline{\#}_s^{L_2}(a_2, b))$;
      **if** $p < \alpha$ **then**
        *pair_significant_different* = *true*;
      **end if**
    **end for**
  **end for**
  **if** $\neq$ *pair_significant_different* **then**
    *all_significant_different* = *false*;
  **end if**
**end for**
**if** *all_significant_different* **then**
  **return** *relative_information_gain*$(L_1, L_2)$
**else**
  **return** 0.0
**end if**

**Algorithm 1:** Algorithm of the label refinement statistical evaluation method

### 4.3.3 Statistical Testing

The relative information gain measure can yield a high value for a refinement just by chance when the generated refined labels are infrequent. By applying statistical testing of the differences in log statistic in addition to calculating the relative information gain enables us to distinguish between information gain obtained by chance and actual information gain. Fisher's exact test [Fis34] is a statistical significance test for the analysis of contingency tables. When applied to the table above, it calculates a p-value for testing the null hypothesis that $a_1$ and $a_2$ events are equally likely to hold log relation s with regard to label b.

Fisher's exact test assumes individual observations to be independent and row and column totals to be fixed. Independence of individual observations might be affected by the grouping of events in traces. Here, we consider individual observations independence to be a working assumption. The test was designed for experiments where both the row and column totals where conditioned. In our setting, the column totals are conditioned by the labeling function, as the number of events of each label depends on the labeling. However, the row totals are not conditioned and are an observation. Strictly speaking, Fisher's exact test does not yield an exact result when one or both of the row or column totals are unconditioned, but will instead be slightly conservative [McD09], meaning that the probability of the p-value being less than or equal to the significance level when the null hypothesis is true is less than the significance level. Fisher's exact test is computationally expensive for large numbers of observations.

For large sample sizes, either the $\chi^2$ test of independence or the G-test of independence can be used, which are both found to be inaccurate for small sample sizes. A popular guideline is not to use the $\chi^2$ test of independence or the G-test for samples sizes less than one thousand [McD09]. The computational complexity of the evaluation procedure is $\mathcal{O}(|S| \times |cod(l)| \times |\text{split_set}|)$. Many process discovery algorithms are exponential in the number of labels [Aal12]. Based on this we can conclude that statistical evaluation of label refinements is computationally less expensive than checking label refinement usefulness through process discovery.

### 4.3.4 Correcting for Multiple Testing

The computational complexity indicates the number of hypothesis tests performed. When a large set of potential label refinements is evaluated, the evaluation method described is susceptible to the multiple comparison problem [Mil81]. The larger the set of hypotheses tested, the higher the probability of incorrectly rejecting the null hypothesis in at least one of the hypothesis tests. Applying a Bonferroni correction [Dun59; Dun61] to the hypothesis tests performed in the statistical evaluation method of label refinements keeps the familywise error rate constant, i.e., it adjusts the significance level of the individual statistical tests to stricter values such that the desired significance level for the set of significance tests as a whole is achieved.

**Figure 4.7:** The Results of the statistical tests for the evaluation of label refinement usefulness

| Log statistic | P-value |
|---|---|
| Directly follows | 1 |
| Directly precedes | $4.91 \times 10^{-5}$ |
| Eventually follows | 1 |
| Eventually precedes | 1 |

### 4.3.5 A Simplified Example

Consider the event log in Table 4.1 and imagine a scenario where a home care worker knows from experience that the elderly person always sets his alarm clock at 8:30 AM. Based on such expert knowledge we are able to define a label refinement such that all bedroom movements after 8:30 AM are considered as *getting up* events, while all other bedroom movements are considered to be *tossing & turning* events. The rightmost column shows the refined labels obtained through this expert knowledge labeling function. To evaluate the usefulness of this label refinement from a process model point of view, we apply the statistical evaluation method described in Section 4.3. As parameters we set the significance level threshold to the frequently used value of 0.01.

Table 4.7 shows the outcome of the statistical tests performed as part of the label refinement usefulness evaluation. Four hypothesis tests have been performed, after Bonferroni correction, each hypothesis test is tested at the significance level $\frac{0.01}{4} = 0.0025$. The direct following statistic of *tossing & turning* and *getting up* with *living room motion* is statistically significantly given this significance level. The label refinement constructed with expert knowledge is found to be a useful label refinement through statistical evaluation.

## 4.4 A Real Life Application

We will now apply our label refinement evaluation method to a real-life dataset from the smart home analysis domain. So far, we have discussed how to evaluate candidates for label refinements, but we have not yet discussed how to generate or how to obtain such candidates for label refinements for a given dataset. In Chapter 5, we will propose a method to automatically generate label refinement candidates for an event log. Therefore, we will also postpone a comparative evaluation with existing label refinement techniques to Chapter 5. In this section, we aim to validate whether the resulting refined events that are obtained by a applying a label refinement indeed are positioned in different parts of the model when the label refinement evaluation approach that we introduced in Section 4.3 assesses the label refinement as a useful one based on statistical significance.

We use the Van Kasteren smart home environment dataset [Kas+08] to illustrate the effects of label splits in the context of process mining of real-life processes. The van Kasteren dataset consists of 1285 events divided over fourteen different sensors. Events are segmented in days from midnight to midnight, to define cases in the event log. Figure 4.8 shows the *dependency graph* that we obtain by applying the Heuristics Miner [WR11] before we apply any label refinements to the dataset. A *dependency graph* depicts causal relations between activities that meet a certainty threshold. A *dependency graph* can be converted into a Petri net [WR11], however, for the sake of readability we included the *dependency graphs* instead of the Petri nets.

In order to perform this evaluation, we will use a simplistic set of label refinements candidates in the remainder of this section. The candidate set of label refinements consists of splitting each of the fourteen event types into two event types based on the their time in the day, where each event type is separated into a refined event type consisting of all its occurrences *before* the median time-in-the-day of the original event type and into another event type on or after the median time-in-the-day of the original event type.

Out of the fourteen candidate label refinements, two label refinements are selected by our approach that result in a in event log relations that are different with statistical significance.

The first label refinement found is the split of *Hall-bathroom door* into *Hall-bathroom door_1* and *Hall-bathroom door_2*, with a timestamp below, respectively above or equal to the median time in the day of *Hall-bathroom door* events. We calculated the median time-in-the-day of the *Hall-bathroom door* events to be at 12:37:31, thereby the label refinement is to label all *Hall-bathroom door* events that occurred before this time as *Hall-bathroom door_1* and the events that occurred on or after this time as *Hall-bathroom door_2*.

The second label refinement found is the split of *Cups cupboard* into *Cups cupboard_1* and *Cups cupboard_2* based on the median time of the events of the *Cups cupboard*. Somewhat surprisingly, we found the median time-in-the-day of the *Cups cupboard* events to be rather late in the day at 19:28:08.

We now continue by discussing these two label refinements in detail, thereby focusing on the results of the statistical evaluation framework and on the resulting changes in the process model that we obtain by applying a process discovery approach before and after the label refinement.

### 4.4.1 Discussion of the Hall-bathroom door Refinement

By applying the statistical testing framework that we described in Section 4.3 we found the two new labels to be behaviorally different with statistical significance. We found the two new event labels to be of significant difference in the following three event log relations:

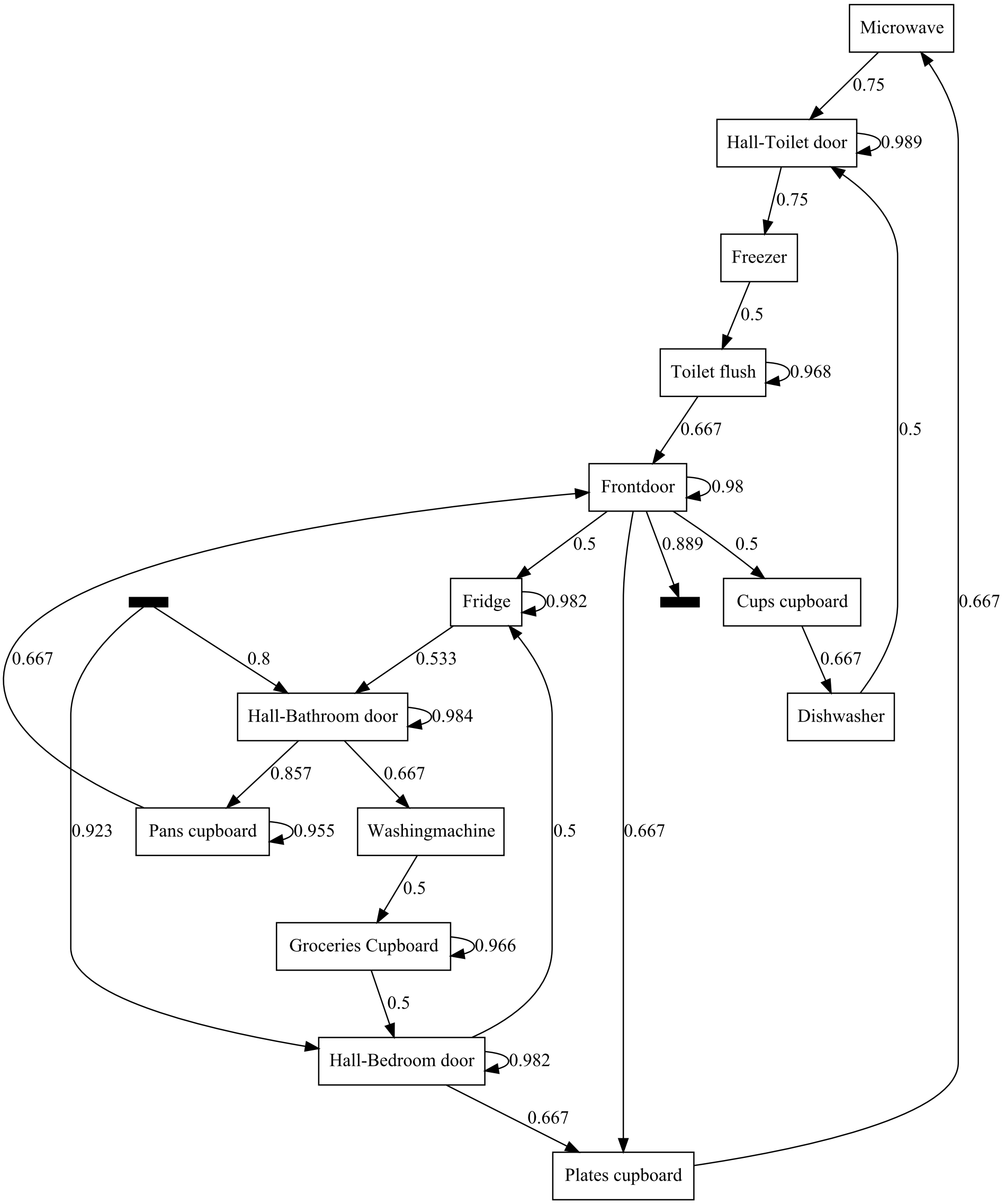

**Figure 4.8:** The dependency graph mined with the Heuristic Miner from the van Kasteren dataset before refining any event labels.

- The *eventually-follows relation* w.r.t the *Front door*, with a p-value of $3.06 \times 10^{-26}$.
- The *eventually-follows relation* w.r.t the *Plates cupboard*, with a p-value of $3.66 \times 10^{-23}$.
- The *eventually-follows relation* w.r.t. the *Microwave*, with a p-value of $1.85 \times 10^{24}$.

Overall, the relative information gain on the whole event log caused by this label refinement is 3.47%.

We now inspect the dependency graph that we mined using the Heuristics Miner [WR11] on the event log that we obtained by applying the median-split on the *Hall-bathroom door* events and validate whether the resulting heuristic net is in agreement with our statistical finding that *Hall-bathroom door_1* and *Hall-bathroom door_2* are behaviorally distinct.

Figure 4.9 shows this dependency graph. We would like to stress the following observations concerning the three statistically significant differences that we have found between *Hall-bathroom door_1* and *Hall-bathroom door_2*.

1. *Hall-bathroom door_1* is connected to the *Front door* while *Hall-bathroom door_2* is not. This is in line with our statistical finding that *Hall-bathroom door_1* and *Hall-bathroom door_2* differ in their eventually-follows relation with respect to the *Frontdoor*.
2. The statistically significant difference that we found between *Hall-bathroom door_1* and *Hall-bathroom door_2* with respect to the *Plates cupboard* is also directly represented by a difference between *Hall-bathroom door_1* and *Hall-bathroom door_2* in the dependency graph, as *Hall-bathroom door_1* is connected to *Plates cupboard* while *Hall-bathroom door_1* is not.
3. The third statistically significant difference that we found between *Hall-bathroom door_1* and *Hall-bathroom door_2* concerned the eventually-follows relation with respect to the *Microwave*. This third significant difference is not directly represented by an analogous difference in the dependency graph. However, observe that *Hall-bathroom door_2* is connected to the *Plates cupboard* which in turn is connected to the *Microwave*. It seems likely that the difference between *Hall-bathroom door_1* and *Hall-bathroom door_2* with respect to the *Microwave* is an indirect effect that results from their difference with respect to the *Plates cupboard*.

### 4.4.2 Discussion of the Cups cupboard Refinement

The relative information gain on the whole event log caused by this label refinement is 0.53%. The resulting labels of this refinement are significantly different with respect to the following log relations:

- The *eventually-precedes relation* w.r.t the *Freezer*, with a p-value of $2.53 \times 10^{-34}$.

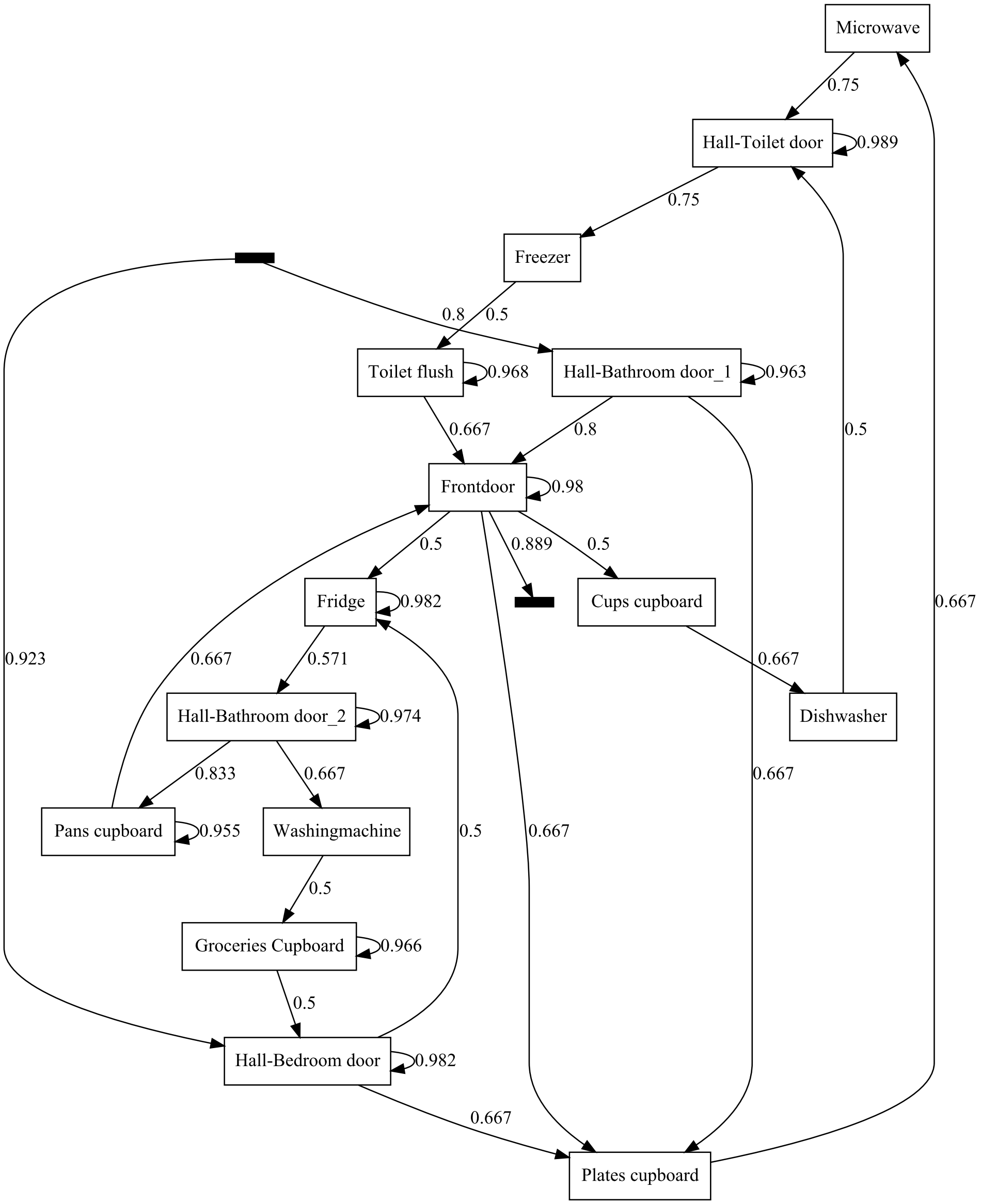

**Figure 4.9:** The dependency graph mined with the Heuristic Miner from the van Kasteren dataset after applying the label refinement on *hall-bathroom door* events.

- The *eventually-follows relation* w.r.t the *Dishwasher*, with a p-value of $2.2 \times 10^{-22}$.

Figure 4.10 shows the dependency graph that we obtained with with the Heuristics Miner on the van Kasteren log with the refined *Cups cupboard* label.

Observe that both of the statistically significant changes are represented in the dependency graph:

- *Cups cupboard_2* is preceded by the *Freezer* in the graph, while *Cups cupboard_1* is not.
- *Cups cupboard_2* is followed by *Dishwasher* in the graph, while *Cups cupboard_1* is not.

By combining these observations with our findings for the *Hall-Bathroom door*, where two of the three significantly different log relations were directly represented by a corresponding change in the dependency graph, we conclude that the set of log relations that are found to be different with statistical significance is a strong indicator of the changes in the dependency graph that are to be expected as a result of applying a label refinement.

Furthermore, it seems that even though we have used a very naive approach to generate the candidate label refinements, we have already found that the evening version of the *Cups cupboard* (i.e., *Cups cupboard_2*, consisting of the events that occurred on or later than 19:28:08) is preceded by the *Freezer* and followed by the *Dishwasher*, which seems to be a reasonable *Dinner* pattern. We conjecture that by using more involved data-driven approaches to generate candidate label refinements we can unearth more complex human routines. We will explore this direction in Chapter 5.

### 4.4.3 Effects on Precision

In Section 4.2 we had had shown the effect that improvements in the degree of behavioral specificness of a process model obtained with process discovery that result from the application of a label refinement can not always be quantified using precision measures. The cause for this effect is that the process models that are respectively obtained before and after the label refinement cannot be compared with respect to the same event log since the activity names need to match between log and model. Therefore, we can only measure the precision of the process model that is discovered after refinement with respect to the refined event log and can only measure the precision of the process model that is discovered before refinement with respect to the original event log.

Even though precision measures might not capture all of the gains in behavioral specificness of the process model, Table 4.5 presents the precision of the Petri nets that we obtained by transforming the dependency graphs of Figure 4.8, Figure 4.9, and Figure 4.10 to Petri net form. Observe that we still find a measurable increase in precision as an effect of applying the label refinements. Furthermore, the label refinement with higher information gain (the refinement of *Hall-Bathroom door*, with information gain of 3.47%) results in a larger precision increase than the *Cups cupboard* label refinement for which we found an information gain of 0.53%. This finding is in agreement with our intuition that more deterministic log statistics help the miner in mining structured, non-flower-like, models.

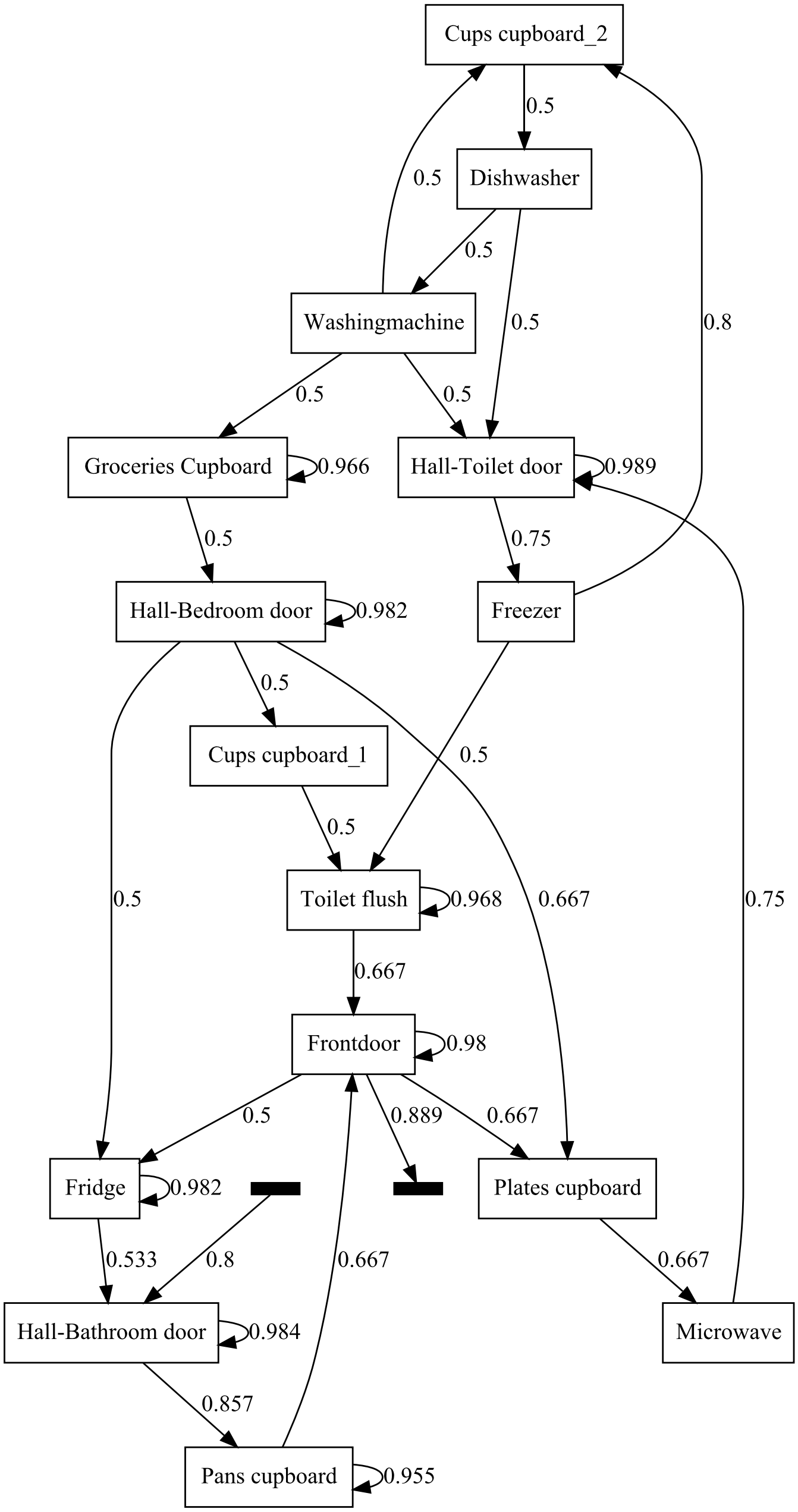

**Figure 4.10:** The dependency graph mined with the Heuristic Miner from the van Kasteren dataset after applying the label refinement on *cups cupboard* events.

**Table 4.5**

| Process model | ETC precision [Muñ+10] |
|---|---|
| Before refinement[17] | 0.56 |
| With *Hall-Bathroom door* refinement[18] | 0.69 |
| With *Cups cupboard* refinement[19] | 0.61 |

## 4.5 Relating Label Refinements to Related Techniques

We discuss four categories of techniques that are related to the evaluation framework for candidate label refinements that we introduced in this chapter. In Section 4.5.1 we discuss process discovery techniques that are able to discover process models with duplicate labels. In Section 4.5.2 we discuss existing work from the process mining domain in the area of label refinements. In Section 4.5.3 we discuss data-aware process mining. Finally, in Section 4.5.4 we discuss related techniques from outside the process mining research area.

### 4.5.1 Mining Process Models with Duplicate Labels

The task of finding refinements of labeling functions is closely related to the task of mining process models with duplicate activities, in which the resulting process model can contain multiple transitions/nodes with the same label. From the point of view of the behavior allowed by a process model, it makes no difference whether a process model is discovered on an event log with refined labels, or whether a process model is discovered with duplicate activities such that each transition/node of the duplicate activity precisely covers one version of the refined label. However, a refined label may also provide additional insights as the new labels are explainable in terms of time.

The first process discovery algorithm capable of discovering duplicate tasks was proposed by Herbst and Karagiannis back in 2004 [HK04], after which many others have been proposed. Two of those more recent approaches extend the foundational $\alpha$-algorithm with capabilities to mine process models with duplicate labels. These two extensions are called the $\alpha^*$-algorithm [LKP07] and the $\alpha^\#$-algorithm [GCY08].

[17]The Petri net corresponding to Figure 4.8
[18]The Petri net corresponding to Figure 4.9
[19]The Petri net corresponding to Figure 4.10

One class of process discovery techniques that is able to mine process models with duplicate tasks are based on Genetic Algorithms (GA). Genetic algorithms for process mining create a population of process models and iteratively evaluate the population of process models on the log, select the best process models from the population, and generate a new population based on combining (called *cross-over*) and modifying (called *mutation*) the selected process models from the old population. The genetic miner algorithm [AMW05; Med06; MWA06; MWA07] specifies the search space as well as the mutation and cross-over operations directly on Petri nets. Bratosin et al. [BSA10] proposed an approach to efficiently distribute the computation of genetic process mining across a cluster of processing units. More recently, Buijs et al. [BDA12b] proposed the Evolutionary Tree Miner (ETM), a genetic approach to process mining that defines the search space and the mutation and cross-over operations on process trees instead of on Petri nets. This use of process trees makes the genetic search more efficient since it reduces the redundancy in the search space. Two additional benefits are that a process tree is sound-by-design and the fact that the tree structure can be leverage to improve the A* heuristic that is used in alignments, thereby enabling faster evaluation of the process trees in the population [TC16].

The EnhancedWFMiner [Fol+09] algorithm for process discovery starts with a pre-processing step in which each activity $a \in \Sigma$ in the log is transformed into an activity $a_n$ with $a \in \Sigma$ and $n \in \mathbb{N}_{>0}$ based on the occurrence in the log, e.g., trace $\langle a, b, c, a, d, c \rangle$ is transformed into $\langle a_1, b_1, c_1, a_2, d_1, c_2 \rangle$. This pre-processing procedure enables the approach to mine process models with duplicate labels. However, this approach does not lead to interpretable results when the event log contains many repetitions of the same activity in the same trace. To illustrate this point, applying this procedure to the van Kasteren smart home dataset results in 190 activities (from 14 activities originally), many of which occur only once in the log. Figure 4.11 shows the resulting dependency graph that is obtained with the heuristic miner after applying this pre-processing procedure. The resulting model is so large and chaotic that is not interpretable.

The Fodina [BD17] algorithm extends the Heuristic Miner among others with a simple label splitting procedure. Labels are split based on their directly-precedes and directly-follows events, where the combination of the directly preceding and the directly following activity of an event is called its *context*. Unlike the technique that is proposed in this chapter, each unique context turns into a unique new label. For example, in trace $\langle a, b, c, a, b, c, b, d \rangle$ the first two b events would become identically labeled since both are preceded by a and followed by c, but the third b event would be labeled differently since it has a different context. In highly variable event logs where there is great variability in the following and preceding events this approach would lead to many unique activities.

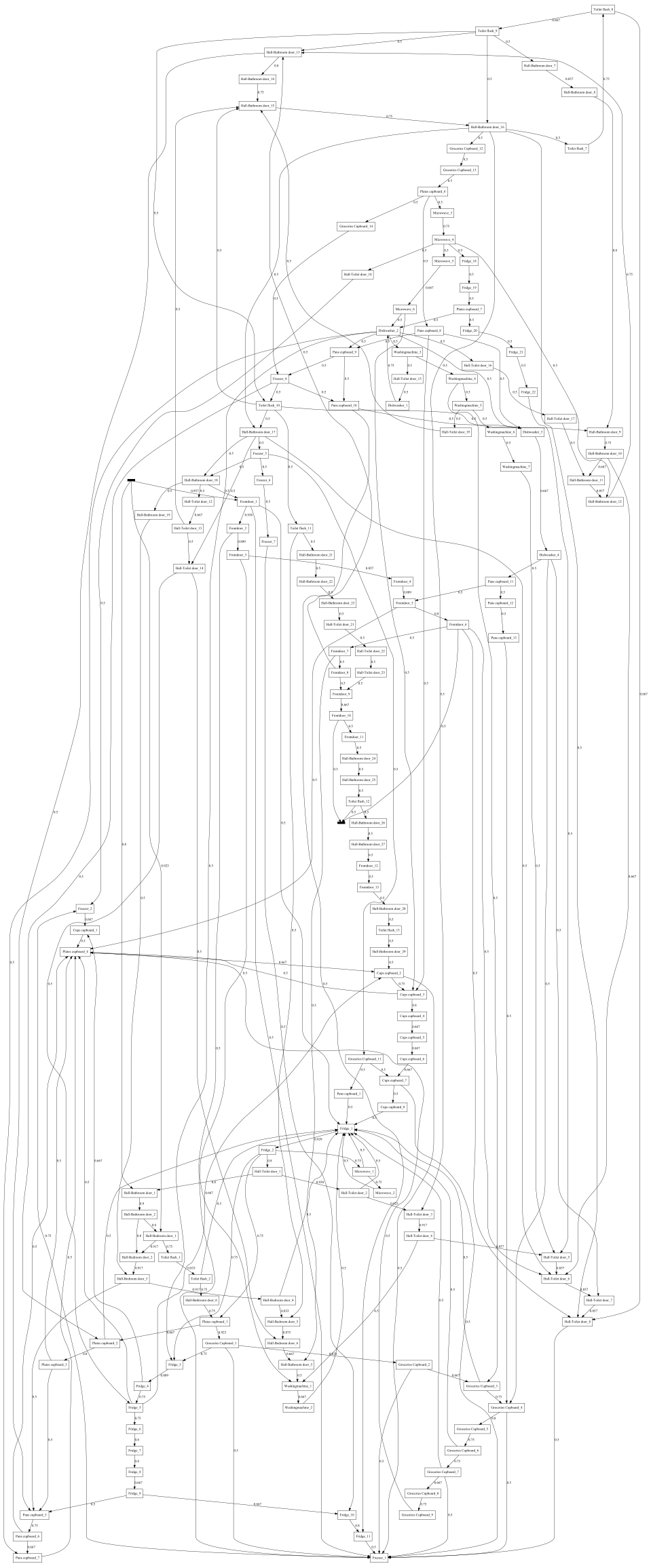

**Figure 4.11:** The dependency graph obtained with the Heuristic Miner from the van Kasteren dataset after relabeling each activity based on its number of occurrence in the trace.

An alternative approach has been proposed by Vázques-Barreiros [VML15] et al. who describe a local search approach based on process model repair that starts by discovering a process model without duplicate labels and then leverages process model repair techniques to bring duplicate labels into the process model.

Van der Aalst et al. [Aal+10] propose discover a Petri net by first extracting a labeled transition system from the event log and then in a second step using *state-based region theory* [Cor+98; DR96; ER90] to transform this transition system into a Petri net (a procedure call *synthesis*). Synthesis is guaranteed to result in a Petri net of which the reachability graph is bisimilar to the initial transition system, as long as this transition system satisfies a certain condition that is called *excitation-closedness*. When a transition system is not excitation-closed, it needs to be transformed such that it satisfies that property by splitting labels. This label splitting procedure is based on so-called *excitation regions* and *generalized excitation regions* [Cor+95; Cor+98]. However, in some circumstances it can be the case that many splittings must be performed to achieve excitation closure, which results in large numbers of transitions in the resulting Petri nets and high computation time (see [Car+08] for a discussion on these issues).

A general difference between all of the existing works on mining models with duplicate activities is that none of these techniques aim to find semantic differences between the different versions of the activities in the form of attribute differences. These approaches consider relabel the events based on the context in terms of control-flow, i.e., the preceding and following activities. In contrast, our perspective on label refinements is that a refinement of a label should have a semantic

### 4.5.2 Label Refinement Techniques in Process Mining

The work that is closest to our work is the work by Lu et al. [Lu+16a], who describe an approach to preprocess a labeled event log by refining the labels with the goal of discovering a process model with duplicate activities. The method proposed by Lu et al. [Lu+16a], however, does not base the relabelings on data attributes of those events and only uses the control flow context, leaving uncertainty whether two events relabeled differently are actually semantically different.

### 4.5.3 Data-Aware Process Mining

A method to discover guards in a given Petri net based net based on alignments is proposed by de Leoni and van der Aalst [LA13]. Guards put data-flow conditions on arcs, which creates a context dependence similar to a label refinement. Guards put data-flow conditions on arcs, putting enabling preconditions for transition firing. Consider a guard that puts a precondition on a transition that is labeled with a where the guard precondition is formulated in terms of single data attribute x from the log. This guard can be considered to be similar to a label refinement that refines activity a based on the value of x into two labels $a_1$ and $a_2$ such that all $a_1$ fulfill the guard condition and all $a_2$ do not. Note that in order to use guard discovery, one already needs to have a Petri net. Therefore, guard discovery can be classified as a post-processing technique to bring data into the Petri net *after* process discovery, where label refinements instead bring data into the event log *before* process discovery. This is a difference of significance, as for many unstructured event logs such as the ones from smart home environments it is not possible to discuss a high-quality process model from the initial data. When it is not possible to discover a process model from the original data such that the resulting process model has clear decision points, then guard discovery does not help to get insight into the dependencies of the process on data attributes.

### 4.5.4 Related Techniques Outside the Process Mining Domain

Similar to the approach that we take to label refinements, evaluating splits based on information gain is a well-known approach in the area of decision tree learning [Qui14]. The scenario of decision tree learning differs from label refinements in the assumption that ground truth labels are available, while finding a suitable label refinement is an unsupervised problem. Furthermore, label refinements draw similarities to automatic learning of ontologies [Mae12] in the sense that both are concerned with inferring multiple levels of semantic interpretations from data. Ontology quality evaluation techniques [BGM05] can be used to evaluate (automatically inferred) ontologies. However, techniques in the area of ontology quality evaluation are not process-centric, i.e., they do not take into account ordering relations between elements of the ontology in execution sequences.

## 4.6 Conclusions

In this chapter, we have provided a theoretical and conceptual notion of when label refinements and abstractions are useful from a process discovery point of view. Based on this conceptual notion of usefulness, we have introduced a framework based on statistics and information theory to evaluate the usefulness of a label refinement or abstraction. In addition, we have demonstrated the applicability of this statistical framework on real-life smart home data, where we have shown a label refinement that was constructed from domain knowledge that was useful according to the framework and also generated new insights into the behavior in the event log. In the next chapter, we will continue with the topic of label refinements and we will address the challenge of finding appropriate label refinements for an event log fully automatically.

# 5 Mining Time-based Label Refinements

Figure 5.1 visualizes the scope of this chapter by highlighting parts of the taxonomy of event log preprocessing methods. In the previous chapter, we have shown that, after starting from a labeling function that labels each event with the sensor in a smart home that registered the event, we can improve the insights that we get from process discovery by first refining the labeling function to take into account the time in the day at which the sensor registered the event. The refinements based on the time in the day that we demonstrated in the previous chapter were based on domain knowledge, and not identified automatically based on the data. In this chapter, we aim to make a step beyond refining labels based on domain knowledge by generating the refinement automatically based on event data such that the resulting labels are semantically interpretable and can be explained to the user. We explore methods to bring parts of the timestamp information to the event label in an intelligent and fully automated way, with the end goal of discovering behaviorally more precise and therefore more insightful process models.

We start this chapter by introducing a framework for the generation of event labels refinements based on the time attribute in Section 5.1. In Section 5.2, we apply this framework to a real-life smart home dataset and show the effect of the refined event labels on process discovery. In Section 5.3, we address the case of applying multiple label refinements together. We continue by describing related work in Section 5.4 and conclude in Section 5.5.

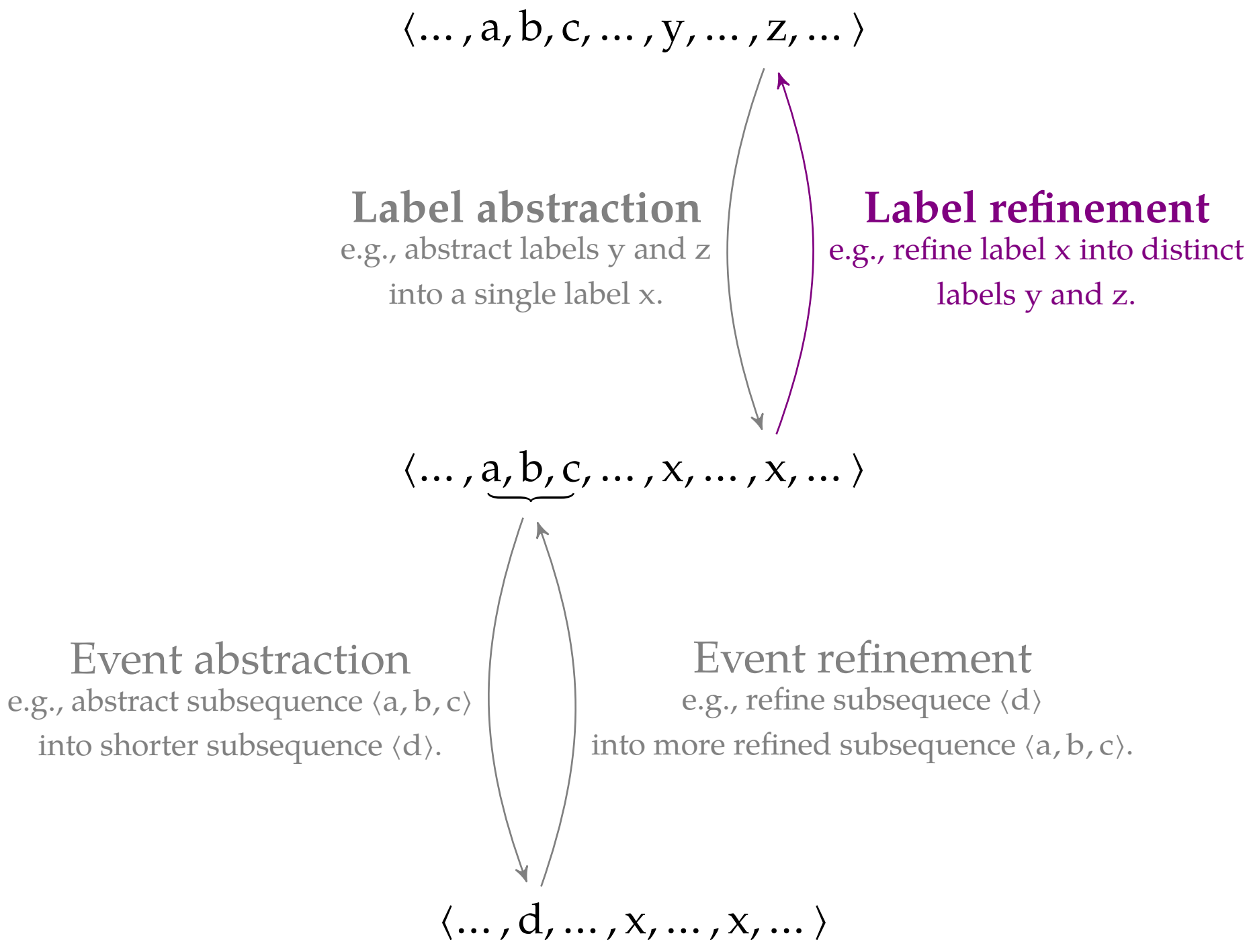

**Figure 5.1:** The positioning of Chapter 5 in the taxonomy of event-log preprocessing methods of Figure 1.4. Colored preprocessing types are discussed in this chapter.

## 5.1 A Framework for Time-based Label Refinements

In this section, we describe a framework to generate event labels that contain partial information about the event timestamp, in order to make the event labels more specific while preserving interpretability. The labeling function $l_{refined}$ is a refinement of $l_{sensor}$, i.e., $l_{refined} \preceq l_{sensor}$. Note that by bringing time-in-the-day information to the event label we aim at uncovering the daily routines of the person under study. We take a clustering-based approach by identifying dense areas in time-space for each label. The time part of the timestamps consists of values between *00:00:00* and *23:59:59*, i.e., it is equivalent to the timestamp attribute from Table 4.1 with the day-part of the timestamp removed. This timestamp can be transformed into a real number time representation in the interval $[0, 24)$. We cluster the set of timestamps, resulting in a label $a \in cod(l_{sensor})$ being refined into a set of new labels $\{a_{cluster-1}, a_{cluster-2}, \dots, a_{cluster-n}\} \subseteq cod(l_{refined})$, with n the number of clusters. The aim is to partition the time interval $[0, 24)$ into densely populated sub-intervals $[t_1^{start}, t_1^{end}), [t_2^{start}, t_2^{end}), \dots, [t_n^{start}, t_n^{end})$ such that the sub-intervals are non-overlapping and jointly span $[0, 24)$. The refined event labels $\{a_{cluster-1}, a_{cluster-2}, \cdots, a_{cluster-n}\}$ are then generated by relabeling each event of activity a with a timestamp in sub-interval $[t_i^{start}, t_i^{end})$ to cluster $a_{cluster-i}$ with $0 \leq i \leq n$.

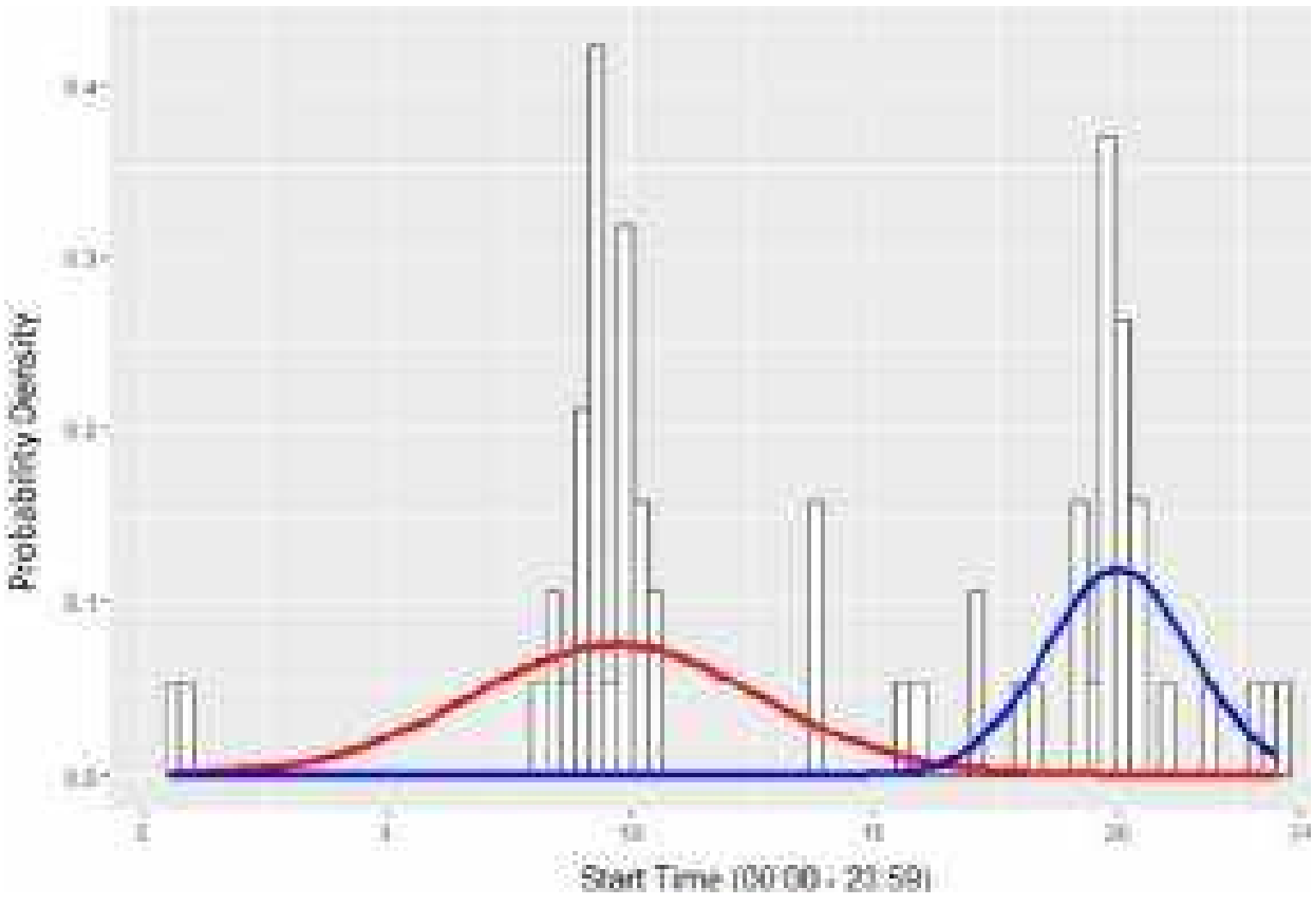

**Figure 5.2:** The histogram representation and a Gaussian Mixture Model fitted to timestamps values of the plates cupboard sensor in the Van Kasteren [Kas+08] dataset.

We apply soft clustering (also referred to as fuzzy clustering), because this approach has the benefit that it assigns to each event a likelihood of belonging to each cluster. A well-known approach to soft clustering is based on the combination of the Expectation-Maximization (EM) [DLR77] algorithm with mixture models, which are probability distributions consisting of multiple components of the same probability distribution. Each component in the mixture represents one cluster, and the probability of a data point belonging to that cluster is the probability that this cluster generated that data point. The EM algorithm is used to obtain a maximum likelihood estimate of the mixture model parameters, i.e., the parameters of the probability distributions in the mixture.

A well-known type of mixture model is the Gaussian Mixture Model (GMM), where the components in the mixture distributions are normal distributions. The data space of time is, however, non-Euclidean: it has a circular nature, e.g., 23.99 is closer to 0 than to 23. This circular nature of the data space introduces problems for GMMs. Figure 5.2 illustrates the problem of GMMs in combination with circular data by plotting the timestamps of the bedroom sensor events of the Van Kasteren [Kas+08] real-life smart home event log. The GMM fitted to the timestamps of the sensor events consists of two components, one with the mean at 9.05 (in red) and one with a mean at 20 (in blue). The histogram representation of the same data shows that some events occurred just after midnight, which on the clock is closer to 20 than to 9.05. The GMM, however, is unaware of the circularity of the clock, which results in a mixture model that seems inappropriate when visually comparing it with the histogram. The standard deviation of the mixture component with a mean at 9.05 is much higher than one would expect based on the histogram as a result of the mixture model trying to explain the data points that occurred just after midnight. The field of circular statistics (also referred to as directional statistics), concerns the analysis of such circular data spaces (cf. [MJ09]). In this chapter, we use a mixture of *von Mises* distributions to capture the daily patterns.

Here, we introduce a framework for generating refinements of event labels based on time attributes using techniques from the field of circular statistics. This framework consists of three stages to apply to the set of timestamps of a sensor:

**Data-model pre-fitting stage** A known problem with many clustering techniques is that they return clusters even when the data should not be clustered. In this stage, we assess if the events of a certain sensor should be clustered at all, and if so, how many clusters it contains. For sensor types that are assessed to be non-clusterable (i.e., the data consists of one cluster), the procedure ends and the succeeding two stages are not executed.

**Data-model fitting stage** In this stage, we cluster the events of a sensor type by timestamp using a mixture consisting of components that take into account the circularity of the data. The clustering result obtained in the fitting stage is now a candidate label refinement. The label can be refined based on the clustering result by adding the assigned cluster to the label of the event, e.g., *open/close fridge* can be relabeled into three distinct labels *open/close fridge 1*, *open/close fridge 2*, and *open/close fridge 3* in case the timestamps of the fridge where clustered into three clusters.

**Data-model post-fitting stage** In this stage, the quality of the candidate label refinements is assessed from both a cluster quality perspective and a process model (event ordering statistics) perspective. The label is only refined when the candidate label refinement is 1) based on a clustering that has a sufficiently good fit with the data, and 2) helps to discover a more insightful process model. If the candidate label refinement does not pass one of the two tests, the label refinement candidate will not be applied (i.e., the label will remain to only consist of the sensor name).

We now proceed with introducing the three stages in detail.

### 5.1.1 Data-model pre-fitting stage

This stage consists of three procedures: a test for uniformity, a test for unimodality, and a method to select the number of clusters in the data. If the timestamps of a sensor type are considered to be uniformly distributed or if they follow a unimodal distribution, then the data is considered to non-clusterable, and the sensor type will not be refined. If the timestamps are neither uniformly distributed nor unimodal, then the procedure for the selection of the number of clusters will decide on the number of clusters used for clustering.

### *Uniformity Check*

Rao's spacing test [Rao76] tests the uniformity of the timestamps of the events from a sensor around the circular clock. This test is based on the idea that uniform circular data is distributed evenly around the circle, and n observations are separated from each other $\frac{2\pi}{n}$ radii. The null hypothesis is that the data is uniform around the circle.

Given a sequence of n successive observations $\langle f_1, \dots, f_n \rangle$ that are ordered either clockwise or counterclockwise, the test statistics U for Rao's Spacing Test is defined as $U = \frac{1}{2}\sum_{i=1}^{n} \mid T_i - \lambda \mid$, where $\lambda = \frac{2\pi}{n}$, $T_i = f_{i+1} - f_i$ for $1 \leq i \leq n-1$ and $T_n = (2\pi - f_n) + f_1$.

### *Unimodality Check*

Hartigan's dip test [HH85] tests the null hypothesis that the data follows a unimodal distribution on a circle. When the null hypothesis can be rejected, we know that the distribution of the data is at least bimodal. Hartigan's dip test measures the maximum difference between the empirical distribution function and the unimodal distribution function that minimizes that maximum difference.

### *Selecting the Number of Mixture Components*

The Bayesian Information Criterion (BIC) [Sch78] introduces a penalty for the number of model parameters to the evaluation of a mixture model. Adding a component to a mixture model increases the number of parameters of the mixture with the number of parameters of the distribution of the added component. The likelihood of the data given the model can only increase by adding extra components, adding the BIC penalty results in a trade-off between the number of components and the likelihood of the data given the mixture model (). BIC is formally defined as $BIC = -2ln(\hat{L}) + k\,ln(n)$, where $\hat{L}$ is a maximized value for the data likelihood, n is the sample size, and k is the number of parameters to be estimated. A lower BIC value indicates a better model. We start with one component and iteratively increase the number of components from k to $k+1$ as long as the decrease in BIC is larger than 10, which is shown to be an appropriate threshold in [KR95].

### 5.1.2 Data-model fitting stage

A generic approach to estimate a probability distribution from data that lies on a circle or any other type of manifold (e.g., the torus and sphere) was proposed by Cohen and Welling [CW15]. However, their approach estimates the probability distribution on a manifold in a non-parametric manner, and it does not use multiple probability distribution components, making it unsuitable as a basis for clustering.

We cluster events generated by one sensor using a mixture model consisting of components of the von Mises distribution, which is the circular equivalent of the normal distribution. This technique is based on the approach of Banerjee et al. [Ban+05] that introduces a clustering method based on a mixture of von Mises-Fisher distribution components. The von Mises-Fisher distribution is an extension of the von Mises distribution to from the one-dimensional circle to n-dimensional spheres. The probability density function of a given datapoint x according to a von Mises distribution with mean direction $\mu$ and concentration parameter $\kappa$ is defined as follows.

$$pdf_{vonMises}(x \mid \mu, \kappa) = \frac{1}{2\pi I_0(\kappa)} e^{\kappa \cos(x-\mu)} \tag{5.1}$$

Both the mean $\mu$ and data point x are expressed in radians on the circle, such that $0 \leq x \leq 2\pi$ and $0 \leq \mu \leq 2\pi$. $I_0$ represents the modified Bessel function of order 0, defined as $I_0(k) = \frac{1}{2\pi}\int_0^{2\pi} e^{\kappa \cos(\theta)} d\theta$. The probability density function of the von Mises distribution is closely related to the probability density function of the normal distribution.

$$pdf_{normal}(x \mid \mu, \sigma^2) = \frac{1}{\sqrt{(2\pi\sigma^2)}} e^{-\frac{(x-\mu)^2}{2\sigma^2}} \tag{5.2}$$

Both probability distributions have a fraction a the first term which solely serves as a normalizing constant that makes sure that the distribution satisfies the requirement of a probability density function that it integrates to 1, i.e., $\int_{-\infty}^{\infty} pdf(x \mid \theta)\, dx = 1$ for any combination $\theta$ of distribution parameters. The $\mu$ parameter takes the role of both the mean and the mode in both probability distributions and both distributions define the probability density of a data point x in terms of its distance from $\mu$. Furthermore, by comparing the two probability density functions it becomes clear that the $\kappa$ parameter of the von Mises distribution is inversely related to the variance $\sigma^2$ of the normal distribution. As a result low values of $\kappa$ indicate a more spread out distribution around the circle and as $\kappa$ increases, the distribution becomes concentrated around the mean $\mu$. In the extreme, as $\kappa$ approaches 0, the distribution becomes uniform around the circle. In fact, these many similarities between the von Mises and the normal distribution are no coincidence, as the von Mises distribution has been shown to closely resemble a normal distribution that is wrapped around the unit circle [MJ09].

The package movMF [HG14] that is included in the statistical programming language R offers functionality to fit a range of circular distributions to a given dataset, including the von Mises distribution. We build upon this package to implement the procedure of fitting a mixture model of von Mises components using the expectation-maximization (EM) procedure where we select the number of components found with the BIC procedure of the pre-fitting stage. A candidate label refinement is created based on the clustering result, where the original label based on the sensor type is refined into a new number of distinct labels, each representing one von Mises component, where each event is relabeled according to the von Mises component that has the assigns the highest likelihood to the timestamp of that event.

### 5.1.3 Data-model post-fitting stage

This stage consists of two procedures: a statistical test to assess how well the clustering result fits the data, and a test to assess whether the ordering relations in the log become stronger by applying the relabeling function (i.e., whether it becomes more likely to discover a precise process model with process discovery techniques).

#### *Goodness-of-fit test*

After fitting a mixture of von Mises distributions to the sensor events, we perform a goodness-of-fit test to check whether the data could have been generated from this distribution. We apply the Watson $U^2$ statistic [Wat62]. The Watson $U^2$ test is a non-parametric test that calculates test statistic $U^2 = n \int_0^{2\pi} \left[F_n(x) - F(x) - \int_0^{2\pi} \{F_n(y) - F(y)\} dF(y)\right]^2 dF(x)$, which tests the null hypothesis that a random sample of size n with empirical distribution $F_n$ has been drawn from a population with the continuous distribution function F. While the Watson $U^2$ test statistic can in principle be used to test whether a data sample comes originates from any type of probability distribution, it's application is particularly popular in combination with circular probability distributions.

#### *Control-flow test*

The clustering obtained can be used as a label refinement where we refine the original event label into a new label for each cluster. We assess the quality of this label refinement from a process perspective using evaluation method for label refinements that we have introduced in Chapter 4.

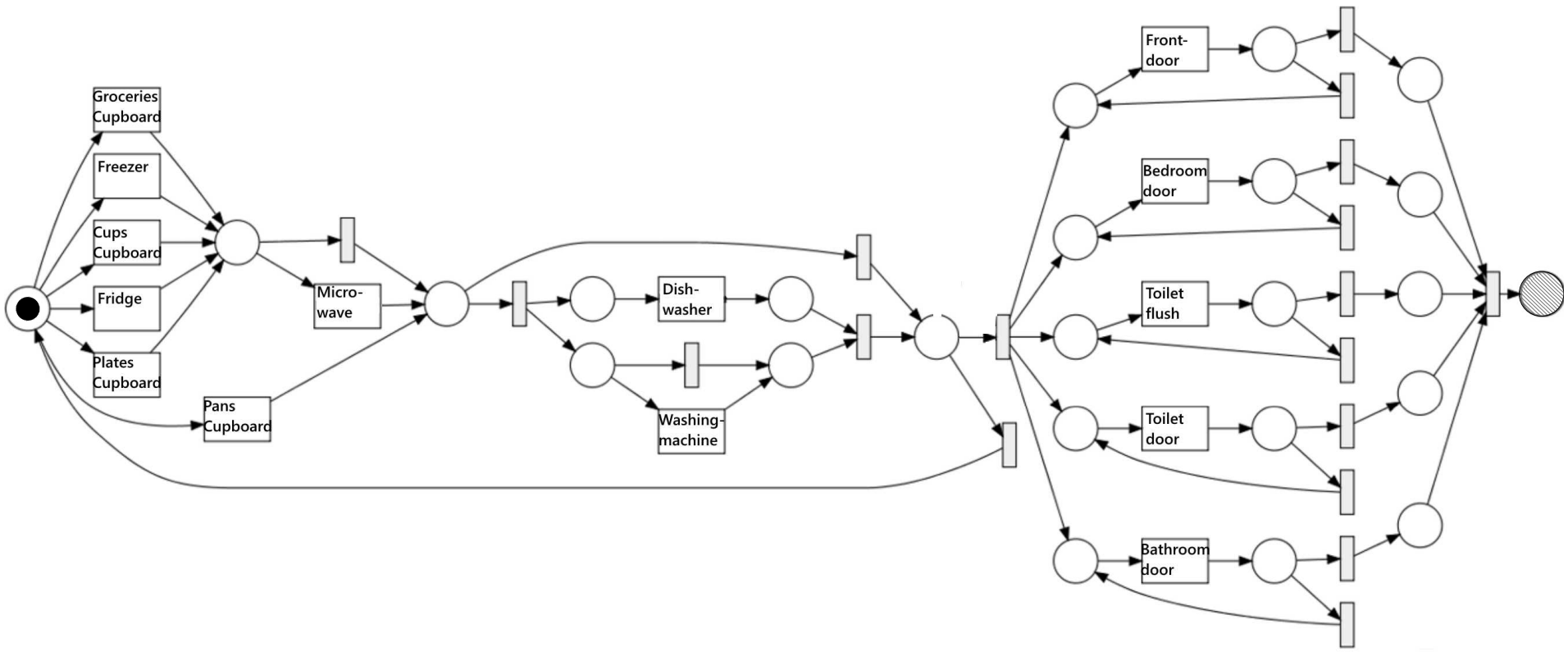

**(a)** Original event data

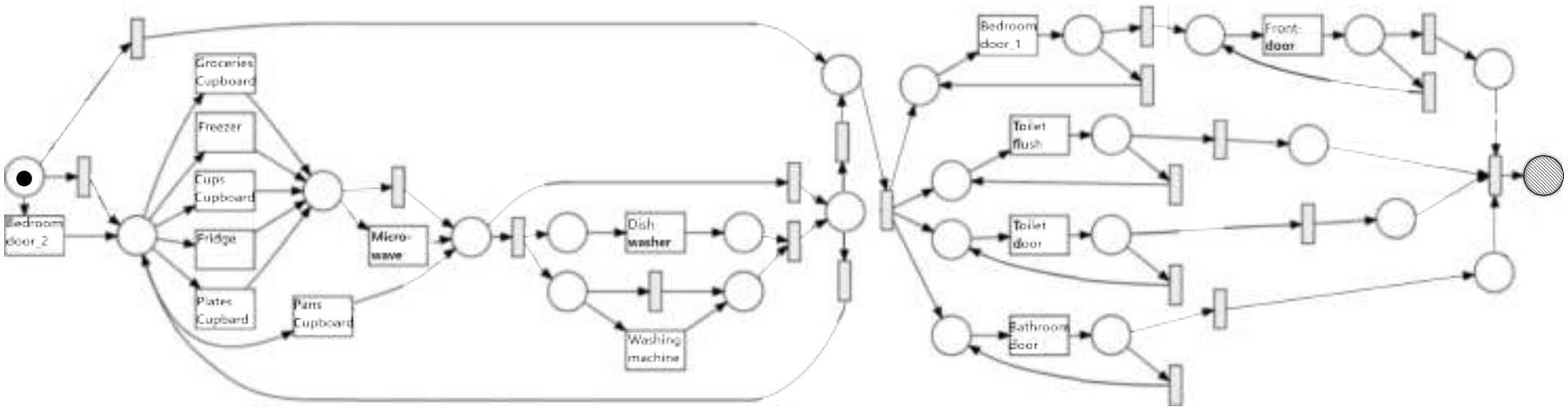

**(b)** Relabeled event data

**Figure 5.3:** The process models discovered on the Van Kasteren data with sensor-level labels (a) and refined labels (b) with the Inductive Miner infrequent (20% filtering).

## 5.2 Case Study

We apply our time-based label refinements approach to the real-life smart home dataset described in Van Kasteren et al. [Kas+08]. The Van Kasteren dataset consists of 1285 events divided over fourteen different sensors. We segment in days from midnight to midnight to define cases. Figure 5.3a shows the process model discovered on this event log with the Inductive Miner infrequent [LFA13a] process discovery algorithm with 20% filtering, which is a state-of-the-art process discovery algorithm that discovers a process model that builds a directly-follows graph from an event log, prunes away the 20% least frequent arcs from the graph, and extracts a Petri net from the remaining graph. Observe that the process model in the figure clearly overgeneralizes, i.e., it allows for too much behavior. At the beginning, a (possibly repeated) choice is made between five transitions. At the end of the process, the model allows any sequence over the alphabet of five activities, where each activity occurs at least once.

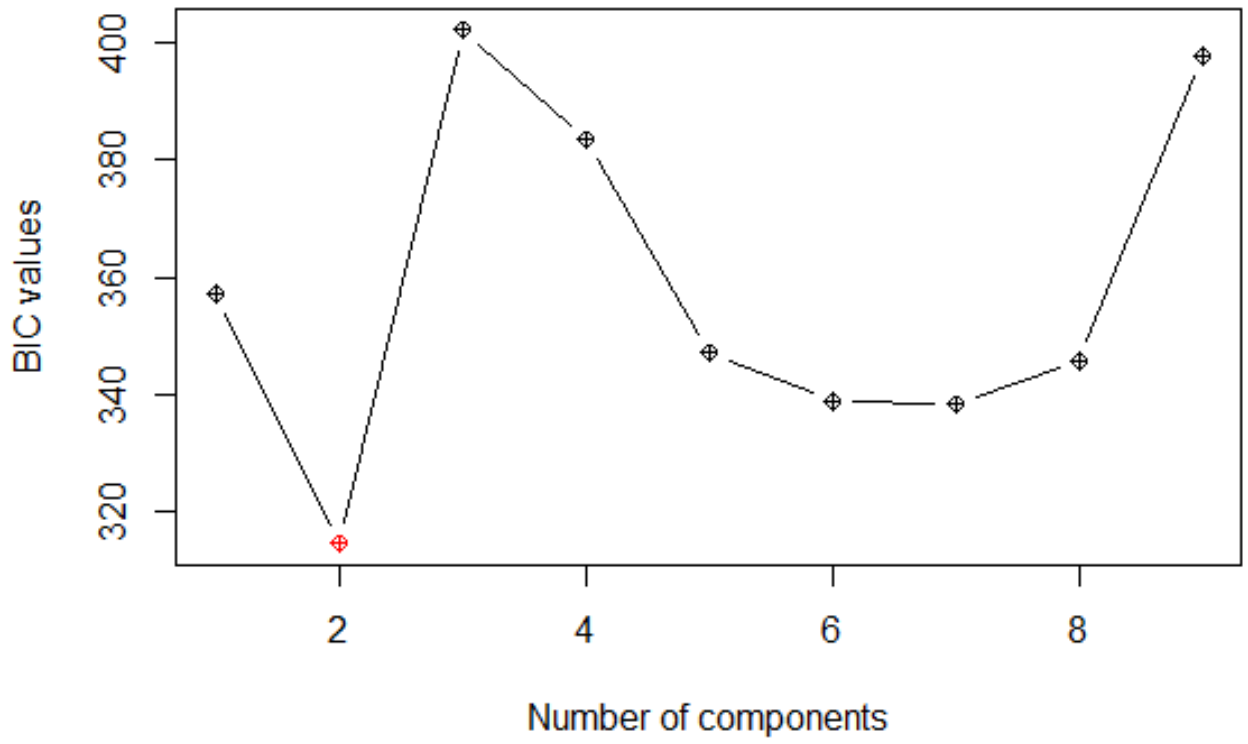

**Figure 5.4:** The BIC values for different numbers of components in the mixture model.

**Table 5.1:** Estimated parameters for a mixture of von Mises components for bedroom door sensor events.

| Cluster | $\alpha$ | $\mu$ (radii) | $\kappa$ |
|---|---|---|---|
| Cluster 1 | 0.76 | 2.05 | 3.85 |
| Cluster 2 | 0.24 | 5.94 | 1.56 |

We illustrate the framework by applying it to the *bedroom door* sensor. Rao's spacing test results in a test statistic of 241.0 with 152.5 being the critical value for significance level 0.01, indicating that we can reject the null hypothesis of a uniformly distributed set of *bedroom door* timestamps. Hartigan's dip test results in a p-value of $3.95 \times 10^{-4}$, indicating that we can reject the null hypothesis that there is only one cluster in the *bedroom door* data. Figure 5.4 shows the BIC values for different numbers of components in the model. The figure indicates that there are two clusters in the data, as this corresponds to the lowest BIC value. Table 5.1 shows the mean and $\kappa$ parameters of the two clusters found by optimizing the von Mises mixture model with the EM algorithm. A value of $0 \equiv 2\pi$ radii equals midnight. After applying the von Mises mixture model to the *bedroom door* events and assigning each event to the maximum likelihood cluster we obtain a time interval of $[3.08, 10.44)$ for cluster 1 and a time interval of $[17.06, 0.88)$ for cluster 2[20].

[20]Note that the circularity of the von Mises distribution enables the interval to cross the midnight point 0. In such scenarios $[17.06, 0.88)$ is short-hand notation for $[17.06 - 24) \cup [0, 0.88)$.

The Watson $U^2$ test results in a test statistic of 0.368 and 0.392 for cluster 1 and 2 respectively with a critical value of 0.141 for a 0.01 significance level, indicating that the data is likely to be generated by the two von Mises distributions found. The label refinement evaluation method of Chapter 4 finds statistically significant differences between the events from the two *bedroom door* clusters with regard to their control-flow relations with other activities in the log for 10 other activities using the significance level of 0.01, indicating that the two clusters are different from a control-flow perspective.

Figure 5.3b shows the process model discovered with the Inductive Miner infrequent with 20% filtering after applying this label refinement to the Van Kasteren event log. The process model still overgeneralizes the overall process, but the label refinement does help to restrict the behavior, as it shows that the evening *bedroom door* events are succeeded by one or more events of type *groceries cupboard*, *freezer*, *cups cupboard*, *fridge*, *plates cupboard*, or *pans cupboard*, while the morning *bedroom door* events are followed by one or more *frontdoor* events. It seems that this person generally goes to the bedroom in-between coming home from work and starting to cook. The loop of the *frontdoor* events could be caused by the person leaving the house in the morning for work, resulting in no logged events until the person comes home again by opening the *frontdoor*. Note that in Figure 5.3a *bedroom door* and *frontdoor* events can occur an arbitrary number of times in any order. Figure 5.3a furthermore does not allow for the *bedroom door* to occur before the whole block of kitchen-located events at the beginning of the net. The *bedroom door* label refinement described above improves the precision of the process model found with the Inductive Miner infrequent (20% filtering) [LFA13a] from 0.3577 when applied on the original event log to 0.4447 when applied on the refined event log and improves the F-score from 0.5245 to 0.6156. However, recall our discussion in Section 4.2 on the difficulties in comparing fitness and precision between two log-model combinations with that are not label-equivalent. The fitness and precision values of the model without the refinements are measured with respect to the original log, while the fitness and precision values of the model with the refinements are measured with respect to the refined log. As a result, not all of the imprecisions of the model without refinements might get accounted for in the precision value, since some of the imprecisions are in the labels of the log and not in the model itself.

The label refinement framework allows for refinement of multiple activities in the same log. For example, label refinements can be applied iteratively. Figure 5.5 shows the effect of a second label refinement step, where *Plates cupboard* using the same methodology is refined into two labels, representing time intervals $[7.98, 14.02)$ and $[16.05, 0.92)$ respectively[21].

[21] Note that the union of the intervals does not result to $[0, 24)$. In practice this is not problematic, since no data points were observed in the missing intervals ($[14.02, 16.05)$ and $[0.92, 7.98)$) and thus all *Plates cupboard*-events can be relabeled. If we would want to relabel a completely new observation that did not occur within the time intervals of the relabeling, then we assign the labels by comparing probability densities for this new data point according to each von Mises component.

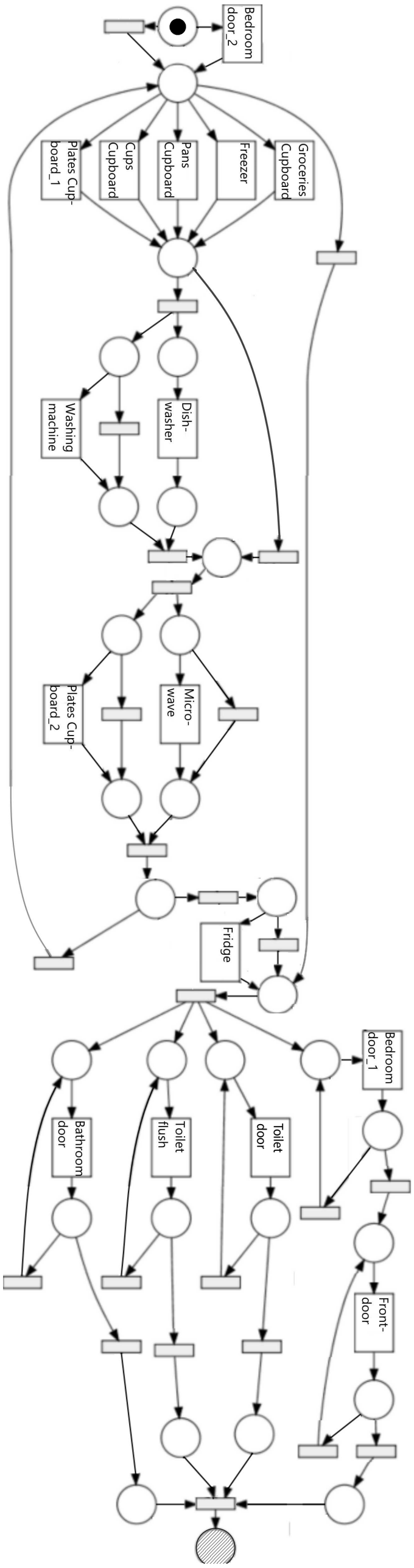

**Figure 5.5:** The result of the Inductive Miner infrequent (with 20% filtering) after a second label refinement.

The second refinement shows the additional insight that the evening version of the *Plates cupboard* occurs directly before or after the microwave. Generating multiple label refinements, however, comes with the problem that the control-flow test of Chapter 4 is sensitive to the order in which label refinements are applied. Because label refinements change the event log, it is possible that after applying some label refinement A, some other label refinement B starts passing the control-flow test that did not pass this test before, or fails the test while it passed before. Additionally, applying one label refinement can change the information gain of applying another label refinement afterwards. For example, when $\#^{L}_{df}(b, c) = \overline{\#}^{L}_{df}$, i.e., b is followed by c 50% of the time, the entropy of this log statistic is 1, equal to a coin toss.

Some label refinement A which refines b into $b_1, b_2$ where $b_1$ is always followed by c and $b_2$ is never followed by c is a good label refinement from an information gain point of view, as it decreases the entropy of the log statistic to zero. Some other label refinement B, which refines c into $c_1, c_2$ such that all b's are directly followed by $c_1$'s and never by $c_2$'s also leads to information gain. However, applying refinement B after having already applied refinement A, does not lead to any further information gain, since refinement A has already made it deterministic whether or not b is followed by any c. Ineffective label refinements might even harm process discovery, as each refinement decreases the frequencies with which activities are observed, thereby decreasing the amount of evidence for certain control-flow relations.

## 5.3 On the Ordering of Label Refinements

To explore the effect of the order in which label refinements are applied, we investigate four strategies to select a set of label refinements to apply to the event log and we evaluate these four strategies on the evaluation logs. Each of the strategies assumes the desired number k of label refinements to be given.

**All-at-once** This strategy can be seen as a naive strategy that ignores the influence of interplay between label refinements on the outcome of the control flow test. It simply applies the three stages of the framework to generate label refinements to each of the sensors in the event logs, and out of all the sensors that pass the statistical significance tests it selects the top k sorted on their information gain, i.e. all the information values are calculated using the original event log on which none of the other possible selected label refinements have been applied.

**Exhaustive Search** The exhaustive search strategy accounts for the effect of the order in which label refinements are applied on the information gain of a label refinement, thereby finding the optimal set of k label refinements. To achieve this goal, starting from the set of label refinements that have passed the statistical significance tests, the strategy investigates all possible orderings in which these label refinements can be applied to the log and calculates for each ordering of label refinements the total information gain that is achieved with it. To calculate this total information gain, the information gain of each label refinement in this ordering is calculated on the log to which all label refinements earlier in the ordering are already applied. While the label refinement sequence that is found with this strategy is optimal in terms of total information gain, this strategy can quickly become computationally intractable for event logs that contain many sensors.

**Greedy Search** This strategy performs a greedy search. It first applies the best label refinement in terms of information gain on the original log and refines the event log using this label refinement. Then, it iterates to find the next label refinement by calculating the information gain using the refined event log from the previous step. The procedure is continued until k refinements have been found. Note that this strategy leads to a considerably smaller search space compared to the exhaustive search strategy, as at each iteration only the best label refinement is applied and the next iteration continuous only from this best label refinement.

**Beam Search** The beam search strategy at is similar to the greedy search strategy, with the difference that at each iteration it keeps a predetermined number b (called the beam size) of best partial solutions are kept as candidates, while the greedy search only keeps the single best candidate. Only the best b combinations in terms of information gain that were found consisting of n label refinements are explored to search for a new set of $n + 1$ label refinements. This is an intermediate strategy in-between greedy and exhaustive search. Beam search is equivalent to greedy search when $b = 1$ and it is equivalent to the exhaustive search strategy when $b = \infty$.

### 5.3.1 Experimental Setup

Label refinements can lead to a decrease in the fitness of the resulting process model, while at the same time increasing its precision. The reason for this is that a very imprecise process model, i.e. one that allows for all behavior, is by definition perfectly fitting. As a result, changing to a process model that makes a more specific, i.e. more precise, description of the behavior can only bring fitness down. To see why refinements can increase precision while fitness can decrease, consider a refinement of a sensor type a into a refined $a_1$ and $a_2$ in such a way that 90% of the $a_1$ events are directly followed by b while $a_2$ is never followed by b. Before this refinement, the process discovery algorithm could not find any relation between a and b, as a could be followed by b, but that was not necessarily the case. After the refinement, it can find a sequential relation between $a_1$ and b, which is more precise than the model before the refinement because it is more specific about the behavior that takes place. However, the model discovered after refinement has lower fitness, since those 10% of the $a_1$ events were not followed by b were are not anymore adhering to the behavior that is described by the model.

We apply these four strategies on three event logs from smart home environments and measure the *fitness*, *precision*, and *F-score* of the models that are discovered with the Inductive Miner infrequent [LFA13a] with 20% filtering after each label refinement. The first event log is the Van Kasteren [Kas+08] event log which we introduced in Section 5.2. The other two event logs are two different households of a smart home experiment conducted by MIT [TIL04]. The log *Household A* of the MIT experiment contains 2701 events spread over 16 days, with 26 different sensors. The *Household B* log contains 1962 events spread over 17 days and 20 different sensors. Note that, like for the Van Kasteren dataset, we consider the events from the MIT A and B datasets on the sensor level instead of on the human activity level. Therefore, the event labels in the log each correspond to a sensor.

The choice between the four strategies can be seen as a trade-off between the degree in which they can take the effect of the ordering of label refinements into account and the computational effort that they require. Note that when the impact of the label refinements that are already applied to the log on the information gain of succeeding label refinements is large, then we expect the exhaustive search strategy to outperform the other strategies. In contrast, in the case that label refinements are mostly mutually independent and there is no strong effect of which label refinements have already been applied to the log on the usefulness in terms of information gain when applying a certain other label refinement, then the all-at-once strategy would suffice to find a set of label refinements and the considerable computational effort needed to perform an exhaustive search would be unnecessary, thereby enabling us to find sets of label refinements for logs with high numbers of sensors. The experiments in which we apply the four search strategies to generate label refinements are intended to investigate the degree to which the order of label refinements matter, i.e., the degree to which one label refinement impacts the usefulness of other label refinements.

### 5.3.2 Results

Figure 5.6 shows the fitness, precision, and F-score values of the process models that were discovered after selecting label refinements with each of the four strategies on the three event logs. The precision can be improved considerably through label refinements on all three event logs.

While the fitness decreased on the MIT A log as an effect of the refinements, fitness is not so much affected by refining the labels on the MIT B and the Van Kasteren event logs. The reason for this is that is in the case that the relation between sensors that is unveiled by a given label refinement holds in almost always, then there is almost no loss in fitness, i.e., if 100% of the $a_1$ in the previous label refinement example would have been directly followed by a b instead of only 90%, then the more precision sequential relation between $a_1$ and b can be modeled

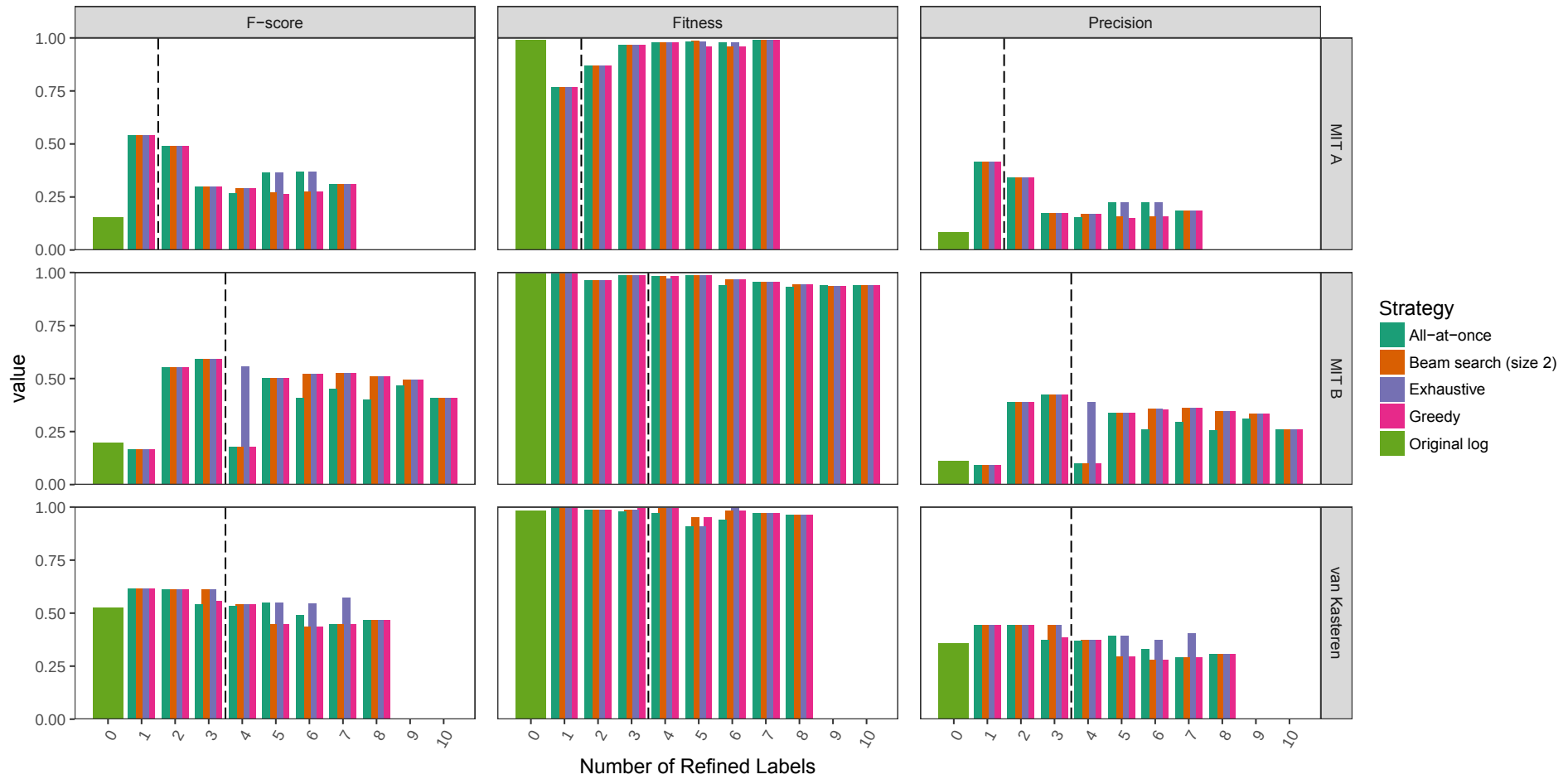

**Figure 5.6:** The fitness, precision, and F-score of the Inductive Miner infrequent (20% filtering) models obtained from the original and refined versions of three event logs.

without loss of fitness.

Note that all strategies find the exact same label refinement as first label refinement, since with only one label refinement there is no notion of the interplay between label refinements. When refining a second event label the four strategies all select the same label refinement on all three logs, hinting that in practice the effect of interplay between label refinements is not very strong. Since all four strategies selected the same second label refinement, the F-score, fitness, and precision are also identical for the four strategies. Figure 5.6 shows that for the MIT household A dataset there are 7 sensor types that can be refined, i.e., they passed the statistical tests of the pre-fitting stage and their obtained clustering passed the goodness-of-fit test. For the MIT household B dataset there are 10 sensor types that can be refined and there are 8 sensor types that can be refined for the Van Kasteren dataset. However, since the F-score for all strategies drops again after a few label refinements, not all of those label refinements lead to better process models. The four strategies perform very similar in terms of F-score.

Exhaustive search outperforms the other strategies for a few refinements on some logs, however, such improvements come with considerable computation times. On the MIT household B log, which has 10 possible label refinements, it takes about 25 minutes on an Intel i7 processor to evaluate all possible combinations of refinements. For logs with even more possible refinements, the exhaustive strategy can quickly become computationally infeasible. The computational complexity of the exhaustive strategy stems from the fact that the ordering in which label refinements are applied can influence their information gain, and therefore, the set of candidate label refinement sequences to consider is equal to the number of permutations on the number of activities (i.e. $\mathcal{O}(n!)$, with n the number of activities) when we restrict to at most one label refinement per original activity (i.e., when a label that was newly created by an earlier label refinement step is not considered in later refinement iterations), and even more when we do not put this restriction and allow label refinements on activities that themselves originate from refinements of other activities.

The all-at-once strategy is computationally very fast and only takes milliseconds to compute on all event logs. Note that with this strategy the information gain only needs to be calculated once for every event label in the log, therefore, the computational complexity is $\mathcal{O}(n)$, with n the number of activities. Yet, this strategy results in fitness, precision, and F-score values that are almost identical to what is obtained with the exhaustive strategy for MIT household A and Van Kasteren. When making six or more refinements on the MIT household B log, the quality of the obtained process model with the all-at-once strategy lags behind the other strategies, indicating that the label refinements that were applied earlier cause the later label refinements to be less effective. However, the optimum in F-score for this log lies at three refinements, therefore, the sixth refinement, where a performance difference between non-exhaustive strategies emerges should not be performed with any of the strategies in the first place.

Since the F-score decreases again when applying too many label refinements it is important to have a stopping criterion that prevents refining the event log too much. The dashed line in Figure 5.6 shows the results when we only refine a label when the information gain of the refinement is larger than zero. On the MIT households A and B logs this stopping criterion causes all strategies to stop at the best combination of label refinements in F-score, consists respectively of one and three refinements. This indicates that the control flow test of Chapter 4 provides a useful stopping criterion for label refinements.

As the greedy search strategy iteratively selects the best label refinement, its computational complexity is $\mathcal{O}(n^2)$, with n the number of activities, when we restrict to only one label refinement per event label. This is easy to see: picking the best a label refinement from all possible activities in the log that can be refined is $\mathcal{O}(n)$, and this procedure has to be repeated at most n times until there are no more activities left that can be refined. The computational complexity of the beam search strategy is depended on the beam size b, as $b = 1$ is equivalent to a greedy search and $b = \infty$ is equivalent to an exhaustive search. Note that on all logs the beam search approach with a beam size of only two resulted in the optimal process model in terms of F-score as found with the exhaustive search, while it reduces the computation time from around half an hour to milliseconds on all logs. Furthermore, the all-at-once and the greedy search strategies resulted in the optimal process model in terms of F-score as found with the exhaustive search on two of the three logs, while offering a better computational complexity than the beam search strategy, therefore, these strategies might be more suitable for large-scale event logs.

All strategies except the exhaustive search strategy suggest as the fourth refinement for MIT B a refinement that decreases the F-score sharply, to increase it again with a fifth refinement. This is caused by an unhelpful refinement being found as the fourth refinement by those strategies, which causes the frequencies to drop below the filtering threshold of the Inductive Miner, leading to a model that is less precise. At the fifth refinement, the follows statistics of other activities drop as well, causing the follows statistics that dropped in the fourth refinement to be relatively higher and above the threshold again. On the Van Kasteren log the optimum in F-score is to make only one refinement, although the F-score after applying the second and third refinement as found by the exhaustive and beam search is almost identical. The all-at-once strategy stops after applying only two refinements while the other strategies apply a third refinement. The best refinement combination found with the all-at-once strategy using the stopping criterion is identical to the refinement combination found with the other strategies, suggesting that in practice the differences between the four approaches are small. For real-life smart home environment event logs the effect that one label refinement influences the control flow test outcome of others is limited.

Figure 5.7 and Figure 5.8 shows the process model that are discovered with the Inductive Miner infrequent with 20% filtering respectively from the original MIT household A event log and the event log obtained after applying the optimal combination of label refinements found in the results of Figure Figure 5.6. Because of the silent transitions, the process model discovered from the original event log allows for almost all orderings over the sensor types. Even though the transition labels in the process model discovered from the refined event log are not readable because of the size, it is clear from the structure of the process model that it is much more behaviorally specific, containing a mix of sequential orderings, parallel blocks, and choices over the sensor types. Especially interesting is the part indicated by the blue dashed ellipse, which contains a parallel block consisting of a *cabinet*, the *oven* and *burner*, and the *dishwasher*, showing a clearly recognizable cooking routine. Furthermore, the part indicated by the red dotted ellipse indicates a sequentially ordered part, consisting of some *door* sensor registering the opening of a door, followed by starting the *washing machine* and then the *laundry dryer*.

The time-based label refinement generation framework as well as the four strategies to generate multiple label refinements on the same event log are implemented and publicly available in the process mining toolkit ProM [Don+05] as part of the *LabelRefinements*[22] package.

[22] https://svn.win.tue.nl/repos/prom/Packages/LabelRefinements/

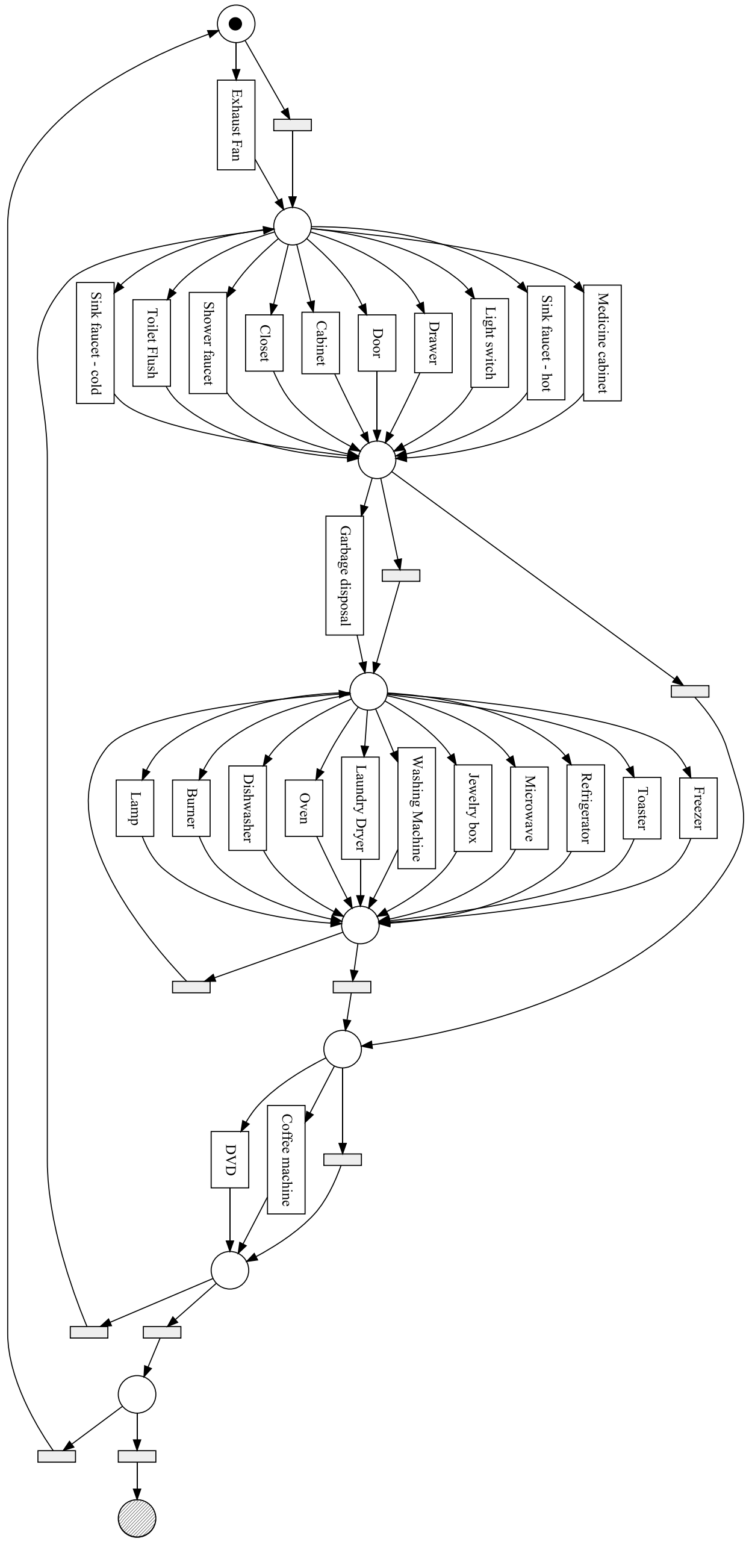

**Figure 5.7:** The Inductive Miner infrequent (20% filtering) process model discovered from the original MIT A event log.

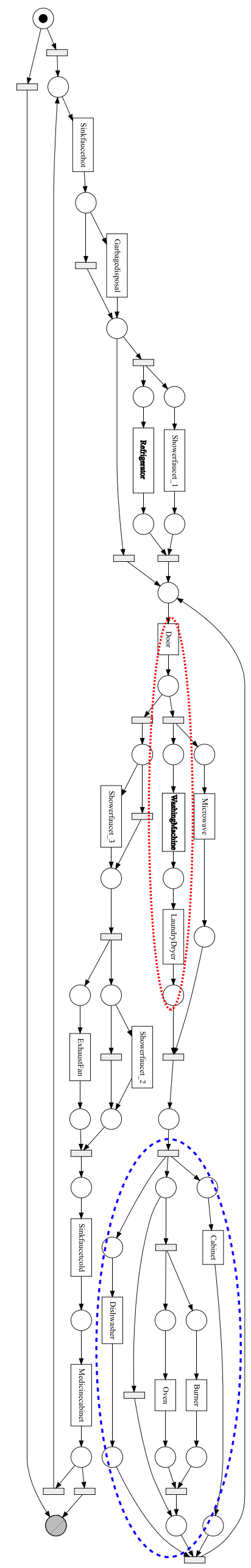

**Figure 5.8:** The Inductive Miner infrequent (20% filtering) process model discovered from the refined MIT A event log.

### 5.3.3 Comparison with Lu's Relabeling Approach

Now that we combined a label refinement generation strategy, a search strategy for generating multiple label refinements, and the label refinement evaluation approach from the previous chapter into a fully automated label refinement approach we will compare our approach with the label refinement approach by Lu et al. [Lu+16a] that we described in Section 4.5.2. The approach of Lu et al. [Lu+16a] starts by matching events from different traces, in order to identify similarities between events across traces. Two events $e_1, e_2$ that respectibely originate from traces $\sigma_1$ and $\sigma_2$ ($\sigma_1 \neq \sigma_2$) can be "mapped" if they share the same label. Furthermore, if events $e_1$ and $e_2$ are mapped, then there can be no other event from $\sigma_1$ that is mapped to $e_2$ and no other event from $\sigma_2$ that is mapped to $e_1$. An event mapping is calculated by minimizing an optimization problem that consists of the following components: *non-matching cost* specifies the cost of not matching an event to any other of another trace, *matching cost* specifies the costs of matching events based on the dissimilarity of the neighboring events of the two events that are mapped, and *structure cost* specifies the costs based on the dissimilarity of the position in their traces of the two events that are mapped. Finally, the three types of costs are weighted by the parameters *non-matching weight*, *matching weight*, and *structure weight*. The matching optimization problem is an NP-complete problem, which is addressed using a greedy heuristic search approach.

We have applied the label refinement approach from Lu et al. [Lu+16a] to the MIT household A dataset. With the recommended default parameters, the technique is not able to match any events from the event log. As a result, the process model that is discovered on the resulting event log is identical to the one shown in Figure 5.7. We hypothesize that the default parameters of the technique are not suitable for smart home event logs due to their high variability (resulting in high *matching costs*) and large trace length (possibly resulting in high *structure costs*).

Figure 5.9 shows the process model that we discovered with the Inductive Miner infrequent (with 20% filtering) after refining the event labels with the label refinement technique of Lu et al. [Lu+16a] when we set the weight of the non-matching cost to the maximal value that is allowed in the user interface. Choosing this maximal value for the non-matching cost leads to a maximal number of events that are mapped, therefore, leading to most fine-grained event log possible with this approach, i.e., the event log that maximally refines the event labels. Observe that this approach refines the *Jewelry box* and the *Washing Machine* into many refined variants. Almost all of those refined events are positioned in parallel to the rest of the process by the Inductive Miner with 20% filtering.

### 5.3.4 Limitations of the Approach

The combined application of the label refinement evaluation approach that we introduced in Chapter 4 and the approach to generate candidate label refinements

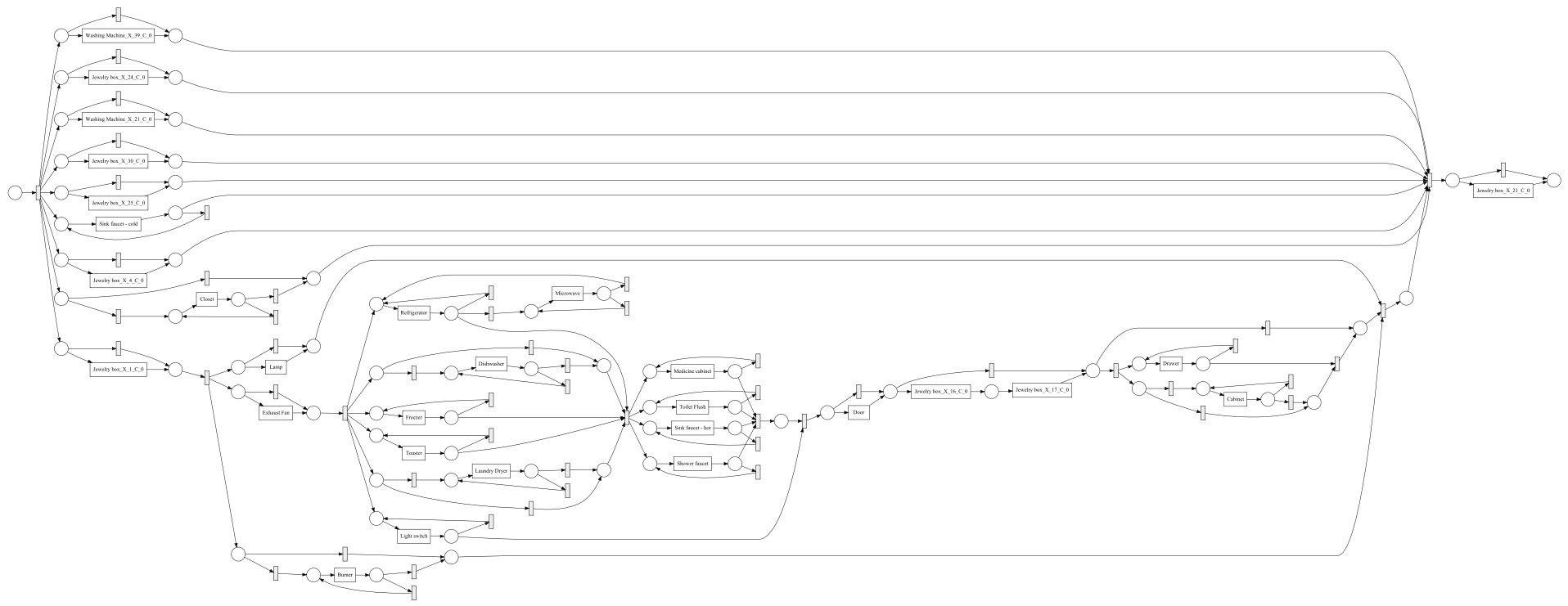

**Figure 5.9:** The Inductive Miner infrequent (20% filtering) process model discovered after relabeling the MIT A event log using the Lu et al. [Lu+16a] relabeling technique with maximal no-matching costs.

based on the data that we introduced in this chapter has several limitations.

1. While we have shown in the experiments in this chapter that the label refinements are able to increase the precision of the resulting process model, the precision of the resulting models after refining the event labels is still very low compared to what is typically observed in the BPM domain. We can therefore say that label refinements by themselves do not solve the problem of mining insights from highly variable event data, but it bring us a step towards achieving this goal by at least unveiling some relations between activities that remain hidden when we apply process discovery without refining event labels.

2. The label refinement approach contains many parameters: the label refinement evaluation approach of Chapter 4 can in principle be applied using any set of log statistics and with any desired significance level. Furthermore, the label refinement candidate generation approach of this chapter relies on many statistical tests each with its own significance level parameter and can be used with several different strategies to ultimately select the label refinements. In addition, the label refinements are agnostic of the process discovery approach, and most process discovery approaches that could be used to mine a process model from the refined event log themselves come with a set of algorithm parameters. While we have experimentally investigated the effects of the different selecting strategies in Section 5.3, the effect of the other parameters in the approach on the quality of the resulting process model remain mostly unexplored.

## 5.4 The Relation of Time-based Label Refinements to Temporal Relation Mining

Galushka et al. [GPR06] provide an overview of temporal data mining techniques and discuss their applicability to data from smart home environments. Many of the techniques described in the overview focus on real-valued time series data, instead of discrete sequences which we assume as input in this work. For discrete sequence data, Galushka et al. [GPR06] propose the use of sequential rule mining techniques, which can discover rules of the form *"if event **a** occurs then event **b** occurs with time **T**"*.

Huynh et al. [HFS08] proposed to use topic modeling to mine activity patterns from sequences of human events. Topic modeling originates from the field of natural language processing and addresses the challenge to find topics in textual documents and assign a distribution over these topics to each document. However, the discovered topics do not represent the human activities in terms of control-flow ordering constructs like sequential ordering, concurrent execution, choices, and loops.

Ogale et al. [OKA07] proposed an approach to describe the temporal relation between human behavior activities from video data using context-free grammars, using the human poses extracted from the video as the alphabet. Like Petri nets, context-free grammars define a formal language over its alphabet. However, Petri nets have a graphical representation, which is lacking for grammars. Furthermore, as shown by Peterson [Pet81], Petri net languages are a subclass of context-sensitive languages, and some Petri net languages are not context-free. This indicates that some relations over activities that can be expressed in Petri nets cannot be expressed in a context-free grammar.

One particularly related technique, called *TEmporal RElation Discovery of Daily Activities (TEREDA)*, was proposed by Nazerfard et al. [NRC11]. TEREDA leverages temporal association rule mining techniques to mine ordering relations between activities as well as patterns in their timestamp and duration. The ordering relations between activities that are discovered by TEREDA are restricted to the form *"activity* **a** *follows activity* **b**", where our proposed approach of modeling the relations with Petri nets allow for modeling of more complex relations between larger number of activities, such as: *"the occurrences of activity* **b** *that are preceded by activity* **a** *are followed by both activity* **d** *and* **e**, *but in arbitrary order"*. The patterns in the timestamps are obtained by fitting a Gaussian Mixture Model (GMM) with the Expectation-Maximization (EM) algorithm, thereby ignoring problems caused by the circularity of the 24-hour clock.

Jukkala and Cook [JC07] propose a method to mine temporal relations between activities from smart home environments logs where the temporal relation patterns are expressed in *Allen's interval algebra* [All83]. Allen's interval algebra allows the expression of thirteen distinct types of temporal relations between two activities based on both the *start* and *end* timestamps of these activities. The approach of Jukkala and Cook [JC07] is limited to describing the relations between pairs of activities, and more complex relations between three or higher numbers of activities cannot be discovered. The aim of mining the patterns in Allen's interval algebra representation is to increase the accuracy of activity recognition systems, while our goal is knowledge discovery.

Several papers from the process mining area have focused on mining temporal relations between activities from smart home event logs. Leotta et al. [LMM15] postulate three main research challenges for the applicability of process mining technique for smart home data. One of those three challenges is to improve process mining techniques to address the less structured nature of human behavior as compared to business processes. Our technique addresses this challenge, as the time-based label refinements help in uncovering relations between activities with process mining techniques that could not be found without applying time-based label refinements.

DiMaggio et al. [Dim+16] and Sztyler et al. [Szt+15; Szt+16] propose to mine Fuzzy Models [GA07] to describe the temporal relations between human activities. The Fuzzy Miner [GA07], a process discovery algorithm that mines a Fuzzy Model from an event log, is a process discovery algorithm that is designed specifically for weakly structured processes. However, Fuzzy Models, in contrast to Petri nets, do not define a formal language over the activities, and are therefore not precise on what activity orderings are allowed and which are not. While mining a Fuzzy Model description of human activities is less challenging compared to mining a process model with formal semantics, it is also limited in the insights that can be obtained from it.

## 5.5 Conclusions

We have proposed a framework based on techniques from the field of circular statistics to refine event labels automatically based on their timestamp attribute. We have shown on a real-life event log that this framework can be used to discover label refinements that allow for the discovery of more insightful and behaviorally more specific process models. Additionally, we explored four strategies to search combinations of label refinements. We found that the difference between an all-at-once strategy, which ignores that one label refinement can have an effect on the usefulness of other label refinements, and other more computationally expensive strategies is often limited. As a result of this finding, the fast but approximate label refinement strategies can in practice be applied instead of performing a full exhaustive search, thereby enabling label refinement search on smart home logs with high numbers of sensors.

# 6 Filtering out Chaotic Activities from Event Logs

**Parts of this chapter have been published as:**

- Niek Tax, Natalia Sidorova, and Wil M. P. van der Aalst: *Discovering more precise process models from event logs by filtering out chaotic activities*. Journal of Intelligent Information Systems 52(1): pp. 1–33 (2018)

6

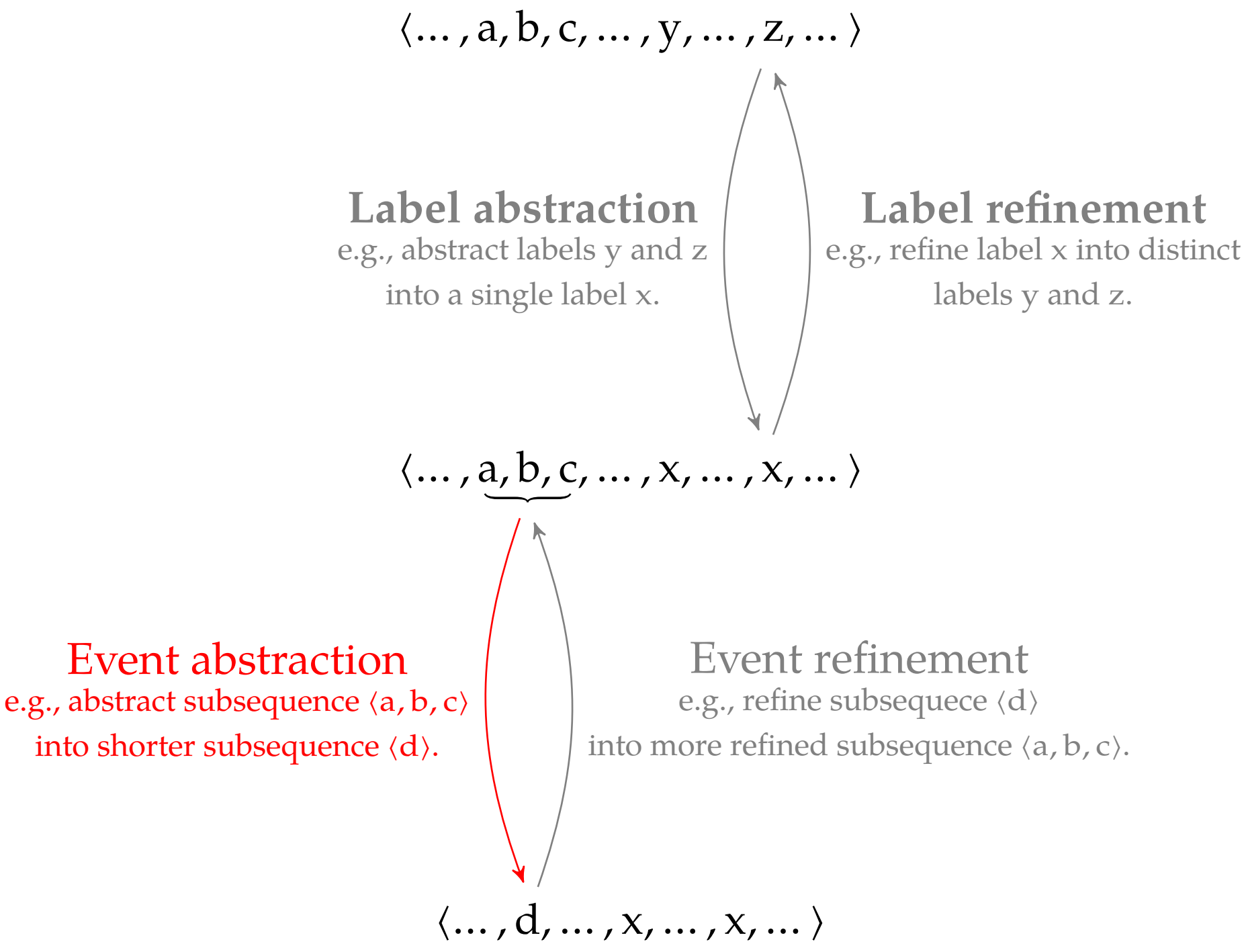

**Figure 6.1:** The positioning of this chapter in the taxonomy of event-log preprocessing methods. Colored preprocessing types are discussed in this chapter.

Figure 6.1 visualizes the scope of this chapter by highlighting parts of the taxonomy of event log preprocessing methods. After focusing on label refinements and abstractions in the previous chapters we will now focus this chapter on a different, orthogonal, type of event log preprocessing, which we refer to as the *filtering of chaotic activities*. Filtering event logs addresses the task of removing outlier behavior from an event log, by removing traces, events, or activities. Note that *event abstraction* is defined in the taxonomy as replacing subsequences of behavior with shorter (i.e., more abstract) subsequences. Therefore, filtering can be seen as a special case of event abstraction, where certain subsequences of behavior are replaced with the empty subsequence $\langle\rangle$, i.e., they are filtered out. This chapter focuses on filtering, while the more general case of event abstraction is addressed in Chapter 7.

In this chapter, the focus is on filtering chaotic activities. *Chaotic activities* refer to activities that can occur spontaneously at any point in the process execution, i.e., their occurrence has no relation with the occurrence of other activities happen before or after. In the context of human behavior, *going to the toilet* is an example of such an activity. People generally go to the toilet whenever they happen to feel the urge to go to the toilet, and it is unlikely that this urge is part of any habit or human routine that describes the relation between toilet and any other human activities. We will now proceed by introducing chaotic activities in more detail and by demonstrating the effect of the presence of chaotic activities in the event log on the quality of process discovery results.

Consider 6.2b, which shows an example process model from [Aal16] that describes a compensation request process. The process model consists of eight process steps (called activities): *(A) register request*, *(B) examine thoroughly*, *(C) examine casually*, *(D) check ticket*, *(E) decide*, *(F) re-initiate request*, *(G) pay compensation*, and *(H) reject request*. Figure 6.2a shows a small example event log consisting of six execution trails of the process model. The Inductive Miner [LFA13b] process discovery algorithm provides the guarantee that it can re-discover the process model from an event log given that all pairs of activities that can directly follow each other

| Event sequences |
|---|
| ⟨A,B,D,E,H⟩ |
| ⟨A,D,C,E,G⟩ |
| ⟨A,C,D,E,F,B,D,E,G⟩ |
| ⟨A,D,B,E,H⟩ |
| ⟨A,C,D,E,F,D,C,E,F,C,D,E,H⟩ |
| ⟨A,C,D,E,G⟩ |

**(a)**

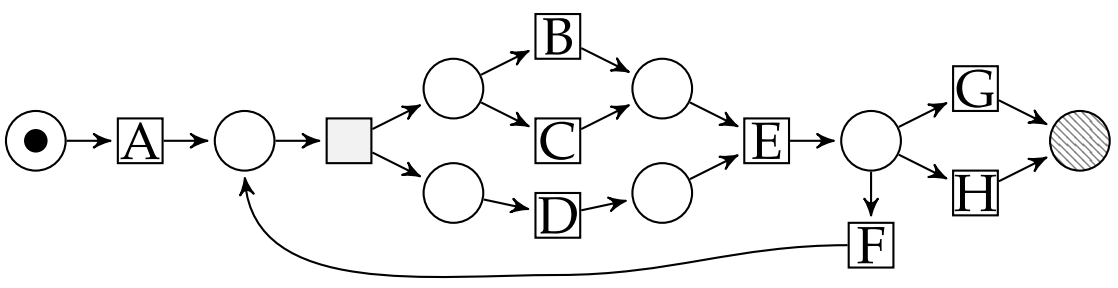

**(b)**

**Figure 6.2:** *(a)* Event log with A=register request, B=examine thoroughly, C=examine casually, D=check ticket, E=decide, F=re-initiate request, G=pay compensation, H=reject request, and *(b)* the Petri net mined from this log with the Inductive Miner [LFA13b].

| Event sequences |
|---|
| ⟨**X**,A,B,D,E,H⟩ |
| ⟨A,D,C,E,G⟩ |
| ⟨A,**X**,C,D,E,F,B,D,**X**,E,G⟩ |
| ⟨A,D,B,E,H⟩ |
| ⟨A,C,D,E,F,D,C,E,F,C,D,E,**X**,H⟩ |
| ⟨A,C,**X**,D,E,G⟩ |

**(a)**

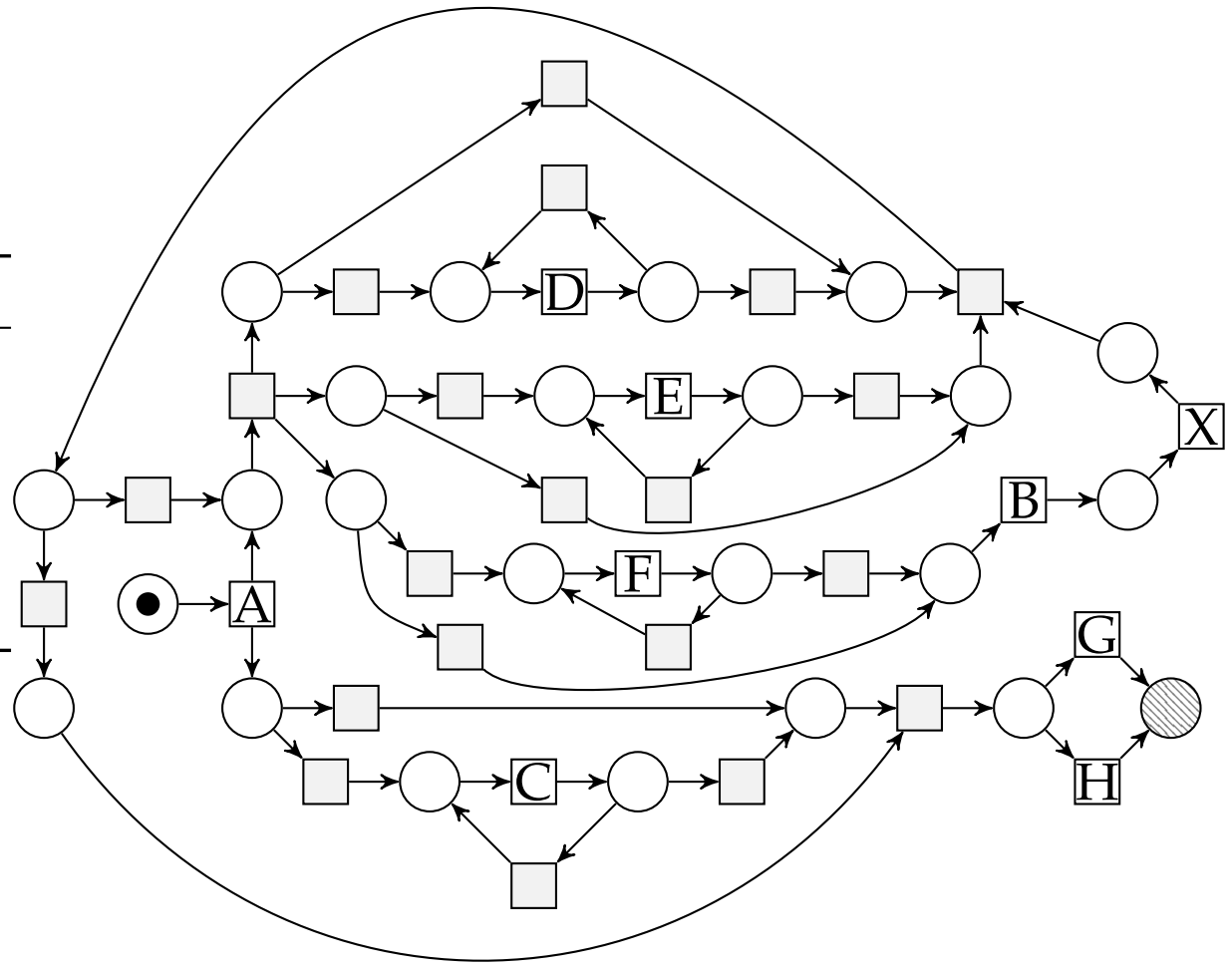

**(b)**

**Figure 6.3:** *(a)* The event log from Figure 6.2a with an added chaotic activity X, and *(b)* the Petri net mined from this log with the Inductive Miner [LFA13a].

in the process are present in the event log, i.e., the log is *directly-follows complete*. Note that the log in Figure 6.2a is directly-follows complete, which is a necessary condition (but not sufficient) for the Inductive Miner algorithm to rediscover the process model from which the event log was generated. In this case, we find that applying the Inductive Miner to this log correctly rediscovers the process model in Figure 6.2b from which the log was generated.

However, the presence of activities that can occur spontaneously at any point in the process execution, which we will call *chaotic activities*, substantially impacts the quality of the resulting process models obtained with process discovery techniques. Figure 6.3a contains the event log obtained from the one in Figure 6.2a by adding activity *(X) the customer calls* at random points, since customers can call the call center multiple times at any point in time during the execution of the process. Figure 6.3b shows the resulting process model discovered by the Inductive Miner [LFA13b] from the event log of Figure 6.3a. The process model discovered from the “clean” example log without activity X (Figure 6.2b) was very simple, interpretable, and accurate with respect to the behavior allowed in the process. In contrast, the process model discovered from the log containing X (Figure 6.3b) is very complex, hard to interpret, and it overgeneralizes by allowing for too much behavior that is not possible in the process. We consider X to be a so-called *chaotic activity* because it does not have a clear position in the process model and it complicates the discov-ery of the rest of the process. The reason for the decline in the quality of process models discovered from logs with chaotic activities is that the directly follows relations, which many process discovery algorithms operate on, are affected by chaotic activities. Examples of such process discovery algorithms include the Inductive Miner [LFA13b], the Heuristics Miner [WR11], and Fodina [BD17]. In a sequence of activities $\langle \dots, A, C, \dots \rangle$, where A was directly followed by C, the addition of a chaotic activity X can turn the sequence into $\langle \dots, A, X, C, \dots \rangle$, thereby obfuscating the causal relation between activities A and C.

In this chapter, we show that existing approaches do not solve the problem of chaotic activities and we present a technique to handle the issue. This chapter is structured as follows. In Section 6.1 we propose an approach to filter out chaotic activities. In Section 6.2 we evaluate our technique using synthetic data where we artificially insert chaotic activities and check whether the filtering techniques can filter out the inserted chaotic activities. Additionally, Section 6.2 proposes a methodology to evaluate activity filtering techniques in a real-life setting where there is no ground truth knowledge on which activities are truly chaotic, and motivates this methodology by showing that its results are consistent with the synthetic evaluation on the synthetic datasets. In Section 6.3, the results on a collection of seventeen real-life event logs are discussed. In Section 6.4, we discuss how the activity filtering techniques can be used in a toggle-based approach for human-in-the-loop process discovery. In Section 6.5, we discuss related techniques in the domains of process discovery and the filtering of event logs. Section 6.6 concludes this chapter.

## 6.1 Information-Theoretic Approaches to Activity Filtering

We consider a *chaotic activity* to be an activity that can occur at any point in the process and that thereby complicates the discovery of the rest of the process by obfuscating the directly-follows relations of the event log. In order to detect such activities, we specify the *directly-follows ratio*, denoted $\mathit{dfr}(a, b, L)$, which represents the ratio of the events of activity a that are directly followed by an event of activity b in event log L, i.e., $\mathit{dfr}(a, b, L) = \frac{\#(\langle a,b \rangle, L)}{\#(a,L)}$.

Likewise, the *directly-precedes ratio*, denoted $\mathit{dpr}(a, b, L)$, represents the ratio of the events of activity a that are directly preceded by an event of activity b in event log L, i.e., $\mathit{dpr}(a, b, L) = \frac{\#(\langle b,a \rangle, L)}{\#(a,L)}$.

Assuming an arbitrary but consistent order over the set of process activities $\mathit{Activities}(L)$, $\mathit{dfr}(a, L)$ represents the vector of values $\mathit{dfr}(a, b, L^{\rfloor})$ for all $b \in \mathit{Activities}(L) \cup \{\rfloor\}$ and $\mathit{dpr}(a, L)$ represents the vector of values $\mathit{dpr}(a, b, L^{\lfloor})$ for all $b \in \mathit{Activities}(L) \cup \{\lfloor\}$.

In this section, we propose a technique to detect chaotic activities in event logs and to filter them out from those event logs.

### 6.1.1 Direct Entropy-based Activity Filtering

We define the entropy of an activity in an event log L based on its directly-follows ratio vector and the directly-precedes ratio vector by using the usual definition of function for the categorical probability distribution: $H(X) = -\sum_{x \in X} x \log_2(x)$. From a probabilistic point of view, the directly-follows-ratio $\mathit{dfr}(a, L)$ and the directly-precedes-ratio $\mathit{dpr}(a, L)$ can be considered to be the empirical estimates of the categorical distributions over respectively the activities directly prior to a and directly after a, where the empirical estimates are based on $\#(a, L)$ trials. We define the entropy of activity $a \in \mathit{Activities}(L)$ in log L as: $H(a, L) = H(\mathit{dfr}(a, L)) + H(\mathit{dpr}(a, L))$. In case there are zero probability values in the directly follows or directly precedes vectors, i.e., $0 \in \mathit{dfr}(a, L) \vee 0 \in \mathit{dpr}(a, L)$, then the value of the corresponding summand $0 \log_2(0)$ is taken as 0, which is consistent with the limit $\lim_{p \to 0^+} p \log_2(p) = 0$.

For example, let event log $L = [\langle a, b, c, x\rangle^{10}, \langle a, b, x, c\rangle^{10}, \langle a, x, b, c\rangle^{10}]$, then $\mathit{dfr}(a, L) = \langle 0, \frac{20}{30}, 0, \frac{10}{30}, 0\rangle$, using the arbitrary but consistent ordering $\langle a, b, c, x, \rfloor\rangle$[23], indicating that 20 out of 30 events of activity a are followed by b and 10 out of 30 by x. Likewise $\mathit{dpr}(a, L) = \langle 0, 0, 0, 0, 1\rangle$, using the arbitrary but consistent ordering $\langle a, b, c, x, \lfloor\rangle$, indicating that all events of activity a are preceded by $\lfloor$. This leads to $H(\mathit{dfr}(a, L)) = 0.918$, $H(\mathit{dpr}(a, L)) = 0$, and $H(a, L) = 0.918$. Furthermore, $H(b, L) = 1.837$, $H(c, L) = 1.837$, and $H(x, L) = 3.170$, showing that activity x has the highest entropy of the probability distributions for preceding and succeeding activities. We conjecture that activities that are chaotic and behave randomly to a high degree have high values of $H(a, L)$.

**Input:** event log L
**Output:** list of event logs Q

*Initialisation* :

$L' = L$
$Q = \langle L' \rangle$

*Main Procedure:*

**while** $|\mathit{Activities}(L')| > 2$ **do**
  $\mathit{acts} = \mathit{Activities}(L')$
  $a' = \arg\max_{a \in \mathit{acts}} H(a, L')$
  $L' = L' \upharpoonright_{\mathit{acts} \setminus \{a'\}}$
  $Q = Q \cdot \langle L' \rangle$
**end while**
**return** Q

**Algorithm 2:** An activity filtering approach based on entropy.

Algorithm 2 describes a greedy approach to iteratively filter the most randomly behaving (chaotic) activity from the event log. The algorithm takes an event log L as input and produces a list of event logs, such that the first element of the list contains a version of L with one activity filtered out, and each following element of the list has one additional activity filtered out compared to the previous element.

[23] Recall that $\lfloor$ and $\rfloor$ respectively denote the *artificial start* and *artificial end* symbols that indicate the start and end of a trace. See Section 2.4.2 for a more elaborate discussion on these concepts.

In the example event log L, Algorithm 2 starts by filtering out activity x, followed by activity b or c. The algorithm stops when there are two activities left in the event log. The reason not to filter any more activities past this point is closely related to the aim of process discovery: uncovering relations between activities. From an event log with less than two activities, no relations between activities can be discovered.

## 6.1.2 The Entropy of Infrequent Activities and Laplace Smoothing

We defined entropy of the activities in an event log L is based on the directly-follows ratios *dfr* and the directly-precedes ratios *dpr* of the activities in L. The empirical estimates of the categorical distributions $\mathit{dfr}(a, L)$ and $\mathit{dpr}(a, L)$ become unreliable for small values of $\#(a, L)$. In the extreme case, when $\#(a, L){=}1$, $\mathit{dfr}(a, L)$ assigns an estimate of 1 to the activity that the single activity a in L happens to be preceded by and contains a probability of 0 for the other activities. Likewise, when $\#(a, L){=}1$, $\mathit{dpr}(a, L)$ assigns value 1 to one activity and value 0 to all others. Therefore, $\#(a, L){=}1$ leads to $H(\mathit{dfr}(a, L)){=}0$ and $H(\mathit{dfr}(a, L)){=}0$. This shows an undesirable consequence of Algorithm 2, infrequent activities are unlikely to be filtered out. In the extreme case, the activities that occur only once, which are the last in line activities to be filtered out. This effect is undesired, as very infrequent activities should not be the primary focus of the process model discovered from an event log.

We aim to mitigate this effect by applying Laplace smoothing [ZL01; ZL04] to the empirical estimate of the categorical distributions over the preceding and succeeding activities. Therefore, we define a *smoothed version* of the directly-follows and directly-precedes ratios, $\mathit{dfr}^s(a, b, L){=}\frac{\alpha\ +\ \#(\langle a,b\rangle,L)}{\alpha(|\mathit{Activities}(L)|+1)+\#(a,L)}$, with smoothing parameter $\alpha{\in}\mathbb{R}_{\geq 0}$. The value of $\mathit{dfr}^s(a, b, L)$ will always be between the empirical estimate $\mathit{dfr}(a, b, L)$ and the uniform probability $\frac{1}{|\mathit{Activities}(L)|+1}$, depending on the value $\alpha$. Similar to *dfr* and *dpr*, $\mathit{dfr}^s(a, L)$ represents the vector of values $\mathit{dfr}^s(a, b, L^{\rfloor})$ for all $b{\in}\mathit{Activities}(L) \cup \{\rfloor\}$ and $\mathit{dpr}^s(a, L)$ represents the vector of values $\mathit{dpr}^s(a, b, L^{\lfloor})$ for all $b \in \mathit{Activities}(L) \cup \{\lfloor\}$. From a Bayesian point of view, Laplace smoothing corresponds to the expected value of the posterior distribution that consists of the categorical distribution given by $\mathit{dfr}(a, L)$ and a Dirichlet distributed prior that assigns an equal probability to each of the possible next activities (including $\rfloor$). Parameter $\alpha$ indicates the weight that is assigned to the prior belief w.r.t. the evidence that is found in the data. An alternative definition of the entropy of log L, based on the smoothed distributions over the preceding and succeeding activities, is as follows: $H^s(a, L) = H(\mathit{dfr}^s(a, L)) + H(\mathit{dpr}^s(a, L))$. The smoothed direct entropy-based activity filter is similar to Algorithm 2, where function H in line 5 of the algorithm is replaced by $H^s$. Function $H(a, L)$ starts from the assumption that an activity is non-chaotic unless we see sufficient evidence in the data for it's chaoticness, function $H^s(a, L)$ in contrast starts from the assumption that is is chaotic, unless we see evidence sufficient evidence in the data for it's non-chaoticness.

Categorical distribution $dfr(a, L)$ consists of $|Activities(L)| + 1$, therefore, the maximum entropy of an activity decreases as more activities get filtered out of the event log. The keep the values of $H^s(a, L)$ comparable between iterations of the filtering algorithm, we propose to gradually increase the weight of the prior by setting weight parameter $\alpha$ to $\frac{1}{|Activities(L)|}$.

### 6.1.3 Indirect Entropy-based Activity Filtering

An alternative approach to the method proposed in Algorithm 2 is to filter out activities such that the *other* activities in the log become less chaotic. We define the total entropy of an event log L as the sum of the entropies of the activities in the log, i.e., $H(L) = \sum_{a \in Activities(L)} H(a, L)$.

Algorithm 3 describes a greedy approach that iteratively filters out the activity that results in the lowest total log entropy. We call this approach the *indirect entropy-based activity filter*, as opposed to the *direct entropy-based activity filter* (Algorithm 2), which selects the to-be-filtered activity directly based on the activity entropy, instead of based on the total log entropy after removal.

**Input:** event log L
**Output:** list of event logs Q
*Initialisation* :
$L' = L$
$Q = \langle L' \rangle$
*Main Procedure:*
**while** $|Activities(L')| > 2$ **do**
  $acts = Activities(L')$
  $a' = \arg\min_{a \in acts} H(L' \upharpoonright_{acts \setminus \{a\}})$
  $L' = L' \upharpoonright_{acts \setminus \{a'\}}$
  $Q = Q \cdot \langle L' \rangle$
**end while**
**return** Q

**Algorithm 3:** An indirect activity filtering approach based on entropy.

### 6.1.4 An Indirect Entropy-based Activity Filter with Laplace Smoothing

Just like the direct entropy-based activity filter, the indirect entropy-based activity filter is sensitive to infrequent activities. To deal with this problem, the ideas of the indirect entropy-based activity filtering method and Laplace smoothing can be combined, using the following definition for smoothed log entropy:

$H^s(L) = \sum_{a \in Activities(L)} H^s(a, L)$.

The algorithm for indirect entropy-based activity filtering with Laplace smoothing is identical to Algorithm 3, in which function H in line 5 is replaced by function $H^s$.

## 6.2 Evaluation using Synthetic Data

In this section, we evaluate the activity filtering techniques using synthetic data. Figure 6.4 gives an overview of the evaluation methodology. First, as step **(1)**, we generate a synthetic event log from a process model such that we know that all activities of this model are non-chaotic. We take the well-known process models introduced by Maruster et al. [Măr+06], which respectively consist of 12 and 22 activities and are commonly referred to as the Maruster A12, A22 models. The Maruster A12 and A22 models are shown respectively in Figures 6.5a and 6.6a. We generated 25 traces by simulation from Maruster A12 to form log $L_{A12}$ and generated 400 traces from Maruster A22 to form log $L_{A22}$. Then, in step **(2)**, we artificially insert activities that we position at random positions in the log. Since we chose the positions in the log of those activities randomly, we assume those activities to be *chaotic*. We vary the number (k) of randomly-positioned activities that we insert, to assess how well the chaotic activity filtering techniques are able to deal with different numbers of randomly-positioned activities in the event log. Furthermore, we vary the frequency of the randomly-positioned activities that we insert, where we distinguish between three types of randomly-positioned activities:

**Frequent randomly-positioned activities** the number of events inserted for all k randomly-positioned activities is $\max_{a \in Activities(L)} \#(a, L)$.

**Infrequent randomly-positioned activities** the number of events inserted for all k randomly-positioned activities is $\min_{a \in Activities(L)} \#(a, L)$.

**Uniform randomly-positioned activities** for each of the k inserted randomly-positioned activities the frequency is chosen at random from a uniform probability distribution with minimum value $\min_{a \in Activities(L)} \#(a, L)$ and maximum value $\max_{a \in Activities(L)} \#(a, L)$.

Note that in all three of the types of randomly-positioned activities, it is possible by chance for all inserted events to be in the same trace. However, when a high number number of events in inserted, such as in the frequent or the uniform randomly-positioned activities, this becomes statistically unlikely.

In step **(3)** we filter out activities from the event log one-by-one using the activity filtering approaches, until k activities have been removed. Ideally, this results in removing exactly those k activities that we had previously inserted in step 2 as randomly-positioned activities. We then count how many of the activities that were originally in the process model we also removed during this procedure (step

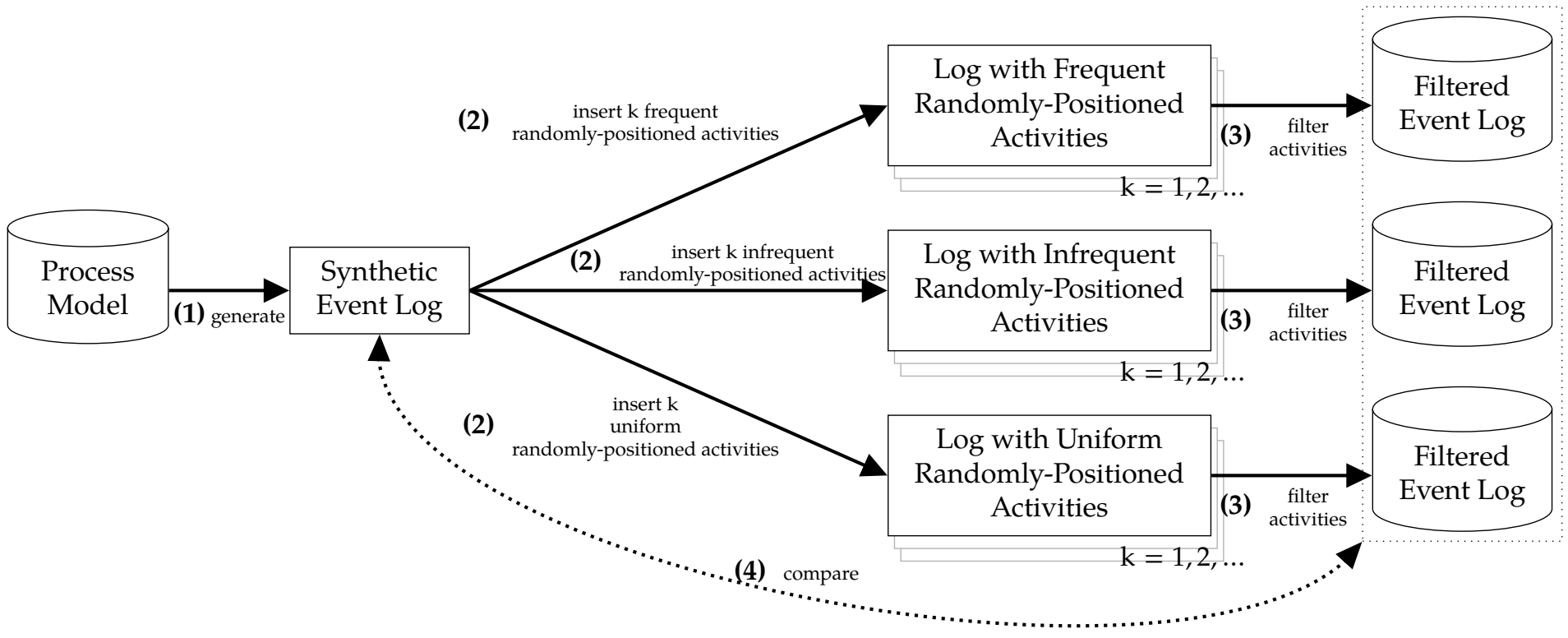

**Figure 6.4:** An overview of the proposed evaluation methodology on synthetic data.

**(4)**). Using this approach, we compare the direct entropy-based activity filtering approach (with and without Laplace smoothing) with the indirect entropy-based activity filtering approach (with and without Laplace smoothing). Furthermore, we compare those activity filtering techniques with activity filtering techniques that are based on the frequency of activities, such as filtering out the activities starting from the least frequent activity (least-frequent-first), or starting from the most frequent activity (most-frequent-first). Frequency-based activity filtering techniques are the current default approach for filtering activities from event logs.

The original process models A12 and A22 can be rediscovered from generated event logs $L_{A12}$ and $L_{A22}$ with the Inductive Miner [LFA13b] when there are no added randomly-positioned activities. Figure 6.5b shows the process model discovered by the Inductive Miner [LFA13b] after inserting one uniform randomly-positioned activity, activity X, into $L_{A12}$. The insertion of activity X causes the Inductive Miner to create a model that overgeneralizes the behavior of the event log, as indicated by many silent transitions in the process model that allow activities to be skipped. Adding a second uniform randomly-positioned activity Y to $L_{A12}$ results in the Inductive Miner discovering a process model (shown in Figure 6.5c) that overgeneralizes even further, allowing for almost all sequences over the set of activities. Figure 6.6b shows the process model discovered by the Inductive Miner after inserting two uniform randomly-paced activities (X and Y) into $L_{A22}$. The addition of X and Y has the effect that activity C is no longer positioned at the correct place in the process model, but it is instead put in parallel to the whole process, making the process model overly general, as it wrongly allows for activity C to occur before A and B, or after D, E, F, and G. Figures 6.5b, 6.5c and 6.6b further motivate the need for filtering out chaotic activities.

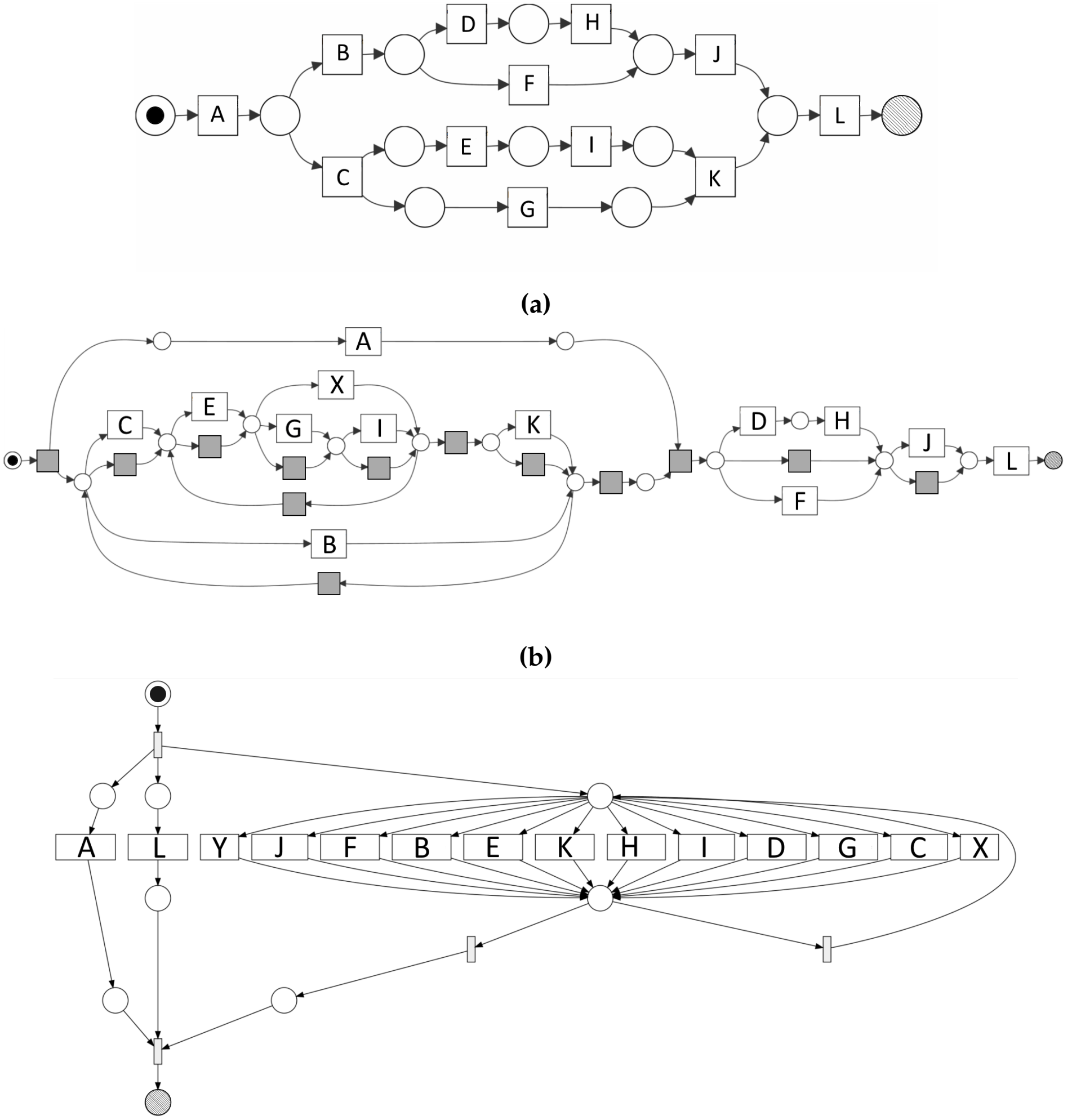

**(c)**

**Figure 6.5:** *(a)* The synthetic process model Maruster A12, from which we generate an event log L, consisting of 25 traces, from which the process model can be rediscovered with the Inductive Miner [LFA13b], *(b)* the process model discovered by the Inductive Mining after inserting one uniform randomly-positioned activity X to $L_{A12}$, and *(c)* the process model discovered by the Inductive Miner after inserting a second randomly-positioned activity Y to $L_{A12}$.

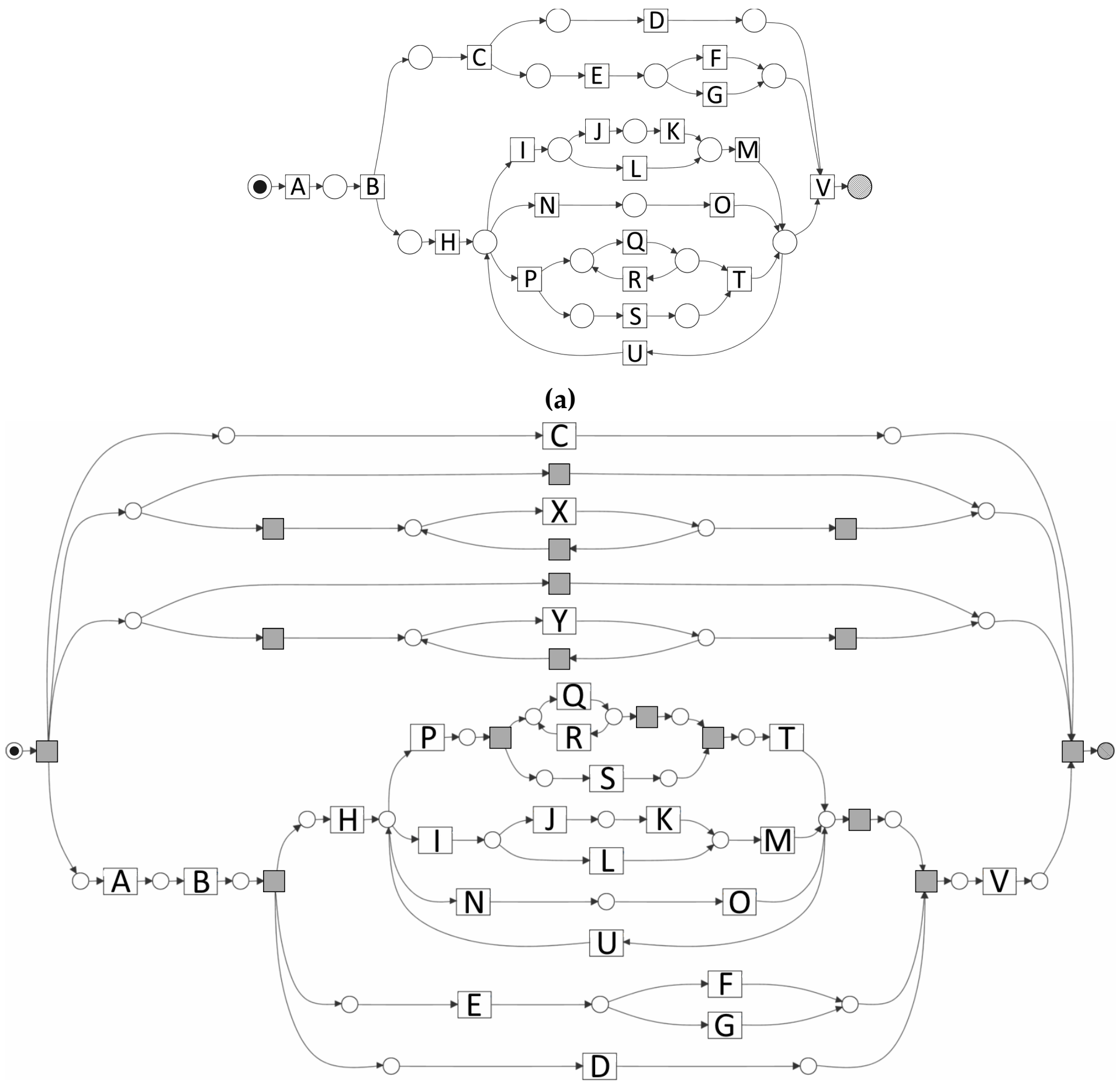

**(b)**

**Figure 6.6:** *(a)* The synthetic process model Maruster A22, from which we generate an event log $L_{A22}$, consisting of 400 traces, from which the process model is re-discoverable with the Inductive Miner [LFA13b], and *(b)* the process model discovered by the Inductive Miner after inserting a uniform randomly-positioned activity X to $L_{A22}$.

**Table 6.1:** The number of incorrectly filtered activities per filtering approach on $L_{A12}$ and $L_{A22}$ with k added Uniform (U) / Frequent (F) /Infrequent (I) chaotic activities.

| Approach | **Maruster A12** (Number of inserted randomly-positioned activities →) 1 | | | 2 | | | 4 | | | 8 | | | 16 | | | 32 | | | 64 | | | 128 | | |
|---|---|---|---|---|---|---|---|---|---|---|---|---|---|---|---|---|---|---|---|---|---|---|---|---|
| | U | F | I | U | F | I | U | F | I | U | F | I | U | F | I | U | F | I | U | F | I | U | F | I |
| Direct | 0 | 0 | 0 | 0 | 0 | 0 | 0 | 0 | 0 | 0 | 0 | 0 | 0 | 0 | 0 | 0 | 0 | 12 | 4 | 0 | 12 | 10 | 1 | 12 |
| Direct ($\alpha=\frac{1}{\vert A\vert}$) | 0 | 0 | 0 | 0 | 0 | 0 | 0 | 0 | 0 | 0 | 0 | 0 | 0 | 0 | 0 | 0 | 0 | 0 | 4 | 0 | 6 | 6 | 2 | 12 |
| Indirect | 0 | 0 | 0 | 0 | 0 | 0 | 0 | 0 | 0 | 0 | 0 | 1 | 1 | 0 | 1 | 1 | 0 | 1 | 2 | 0 | 1 | 3 | 1 | 6 |
| Indirect ($\alpha=\frac{1}{\vert A\vert}$) | 0 | 0 | 0 | 0 | 0 | 0 | 0 | 0 | 0 | 0 | 0 | 1 | 1 | 0 | 1 | 1 | 0 | 1 | 2 | 0 | 1 | 2 | 1 | 10 |
| Least-frequent-first | 9 | 12 | 0 | 11 | 12 | 0 | 6 | 12 | 0 | 11 | 12 | 0 | 11 | 12 | 0 | 12 | 12 | 0 | 12 | 12 | 0 | 12 | 12 | 0 |
| Most-frequent-first | 11 | 0 | 12 | 3 | 0 | 12 | 7 | 0 | 12 | 10 | 0 | 12 | 12 | 0 | 12 | 12 | 0 | 12 | 12 | 0 | 12 | 12 | 0 | 12 |
| **Approach** | **Maruster A22** (Number of inserted randomly-positioned activities →) **1** | | | **2** | | | **4** | | | **8** | | | **16** | | | **32** | | | **64** | | | **128** | | |
| | U | F | I | U | F | I | U | F | I | U | F | I | U | F | I | U | F | I | U | F | I | U | F | I |
| Direct | 0 | 0 | 0 | 0 | 0 | 0 | 0 | 0 | 0 | 0 | 0 | 1 | 0 | 0 | 0 | 0 | 0 | 0 | 0 | 0 | 0 | 0 | 0 | 5 |
| Direct ($\alpha=\frac{1}{\vert A\vert}$) | 0 | 0 | 0 | 0 | 0 | 0 | 0 | 0 | 0 | 0 | 0 | 1 | 0 | 0 | 0 | 0 | 0 | 0 | 0 | 0 | 0 | 0 | 0 | 5 |
| Indirect | 0 | 0 | 0 | 0 | 0 | 0 | 0 | 0 | 0 | 0 | 0 | 1 | 0 | 0 | 1 | 1 | 0 | 1 | 1 | 0 | 1 | 1 | 0 | 1 |
| Indirect ($\alpha=\frac{1}{\vert A\vert}$) | 0 | 0 | 0 | 0 | 0 | 0 | 0 | 0 | 0 | 0 | 0 | 1 | 0 | 0 | 1 | 1 | 0 | 1 | 0 | 0 | 1 | 1 | 0 | 1 |
| Least-frequent-first | 16 | 22 | 0 | 17 | 22 | 0 | 6 | 22 | 0 | 21 | 22 | 0 | 19 | 22 | 0 | 22 | 22 | 0 | 22 | 22 | 0 | 22 | 22 | 0 |
| Most-frequent-first | 7 | 0 | 22 | 8 | 0 | 22 | 19 | 0 | 22 | 17 | 0 | 22 | 19 | 0 | 22 | 22 | 0 | 22 | 22 | 0 | 22 | 22 | 0 | 22 |

Frequent randomly-positioned activities will impact the quality of process models discovered with process discovery to a higher degree than infrequent randomly-positioned activities. Each randomly-positioned activity that is inserted at a random position in the event log is placed in-between two existing events in that log (or at the start or end of the trace). By inserting randomly-positioned activity X in-between two events of activities A and C respectively, the directly-follows relation between activities A and C gets weakened. Therefore, the impact of randomly-positioned activity X is proportional to its frequency $\#(X, L)$.

## 6.2.1 Results

Table 6.1 reports the number of activities that were originally part of the synthetic process models A12 and A22 that were wrongly filtered out from $L_{A12}$ and $L_{A22}$ as an effect of removing all inserted randomly-positioned activities from these logs. If this number is 12 for Maruster A12 or 22 for Maruster A22, then all activities of the original process model had to be filtered out before the activity filtering technique was able to remove all inserted chaotic activities. The results show that the direct filtering approach can perfectly distinguish between actual activities from the process and artificial chaotic activities for up to 32 uniform randomly-positioned activities inserted activities to $L_{A12}$, up to 64 frequent randomly-positioned activities, and up to 16 infrequent randomly-positioned activities. Infrequent randomly-positioned activities are the hardest type of randomly-positioned activities to correctly filter out, as their infrequency can have the effect that the probability distributions over their surrounding activities can *by chance* have low entropy. Using Laplace smoothing with $\alpha = \frac{1}{|\text{Activities}(L)|}$ mitigates this effect, but does not completely solve it: the number of incorrectly removed activities drops from 12 to 0 as an effect of Laplace

smoothing for 32 added randomly-positioned activities, and from 12 to 6 for 64 added randomly-positioned activities. The indirect activity filter starts making errors already at lower numbers of added randomly-positioned activities than the direct activity filter; however, it is more stable to errors for higher numbers of added randomly-positioned activities, i.e., fewer activities get incorrectly removed for 64 and 128 added randomly-positioned activities. In contrast to direct activity filtering, Laplace smoothing does not seem to reduce the number of wrongly removed activities for indirect activity filtering. In fact, surprisingly, the number of incorrectly removed activities even increased from 6 to 10 as an effect of using Laplace smoothing for 128 infrequent randomly-positioned activities added to $L_{A12}$. The direct and indirect filtering approaches, both with and without Laplace smoothing, outperform the currently widely used approach of filtering out infrequent activities from the event log (least-frequent-first filtering). Furthermore, a second frequency-based activity filtering technique is included in the evaluation in which the most-frequent activities are removed from the event log (most-frequent first filtering). Both Frequency-based filtering approaches are not able to filter out the randomly-positioned activities inserted to $L_{A12}$ and $L_{A22}$, even for small numbers of added randomly-positioned activities.

### 6.2.2 An Evaluation Methodology for Event Data without Ground Truth Information

In a real-life data evaluation that we perform in the following section, there is no ground truth knowledge on which activities of the process are chaotic. This motivates a more indirect evaluation in which we evaluate the quality of the process model discovered from the event log after filtering out activities with the proposed activity filtering techniques. In this section, we propose a methodology for evaluation of activity filtering techniques by assessing the quality of discovered process models, we apply this evaluation methodology to the Maruster A12 and Maruster A22 event logs, and we discuss the agreement between the findings of Table 6.1 and the quality of the discovered process models.

When we filter out one or more activities from an event log, then the precision of the process discovery result from the filtered log is likely to increase, *regardless of which activities are removed from the log*. This effect is caused by two factors. First, many precision measures quantify the size of $\mathfrak{L}(M) \setminus \tilde{L}$ with respect to $\mathfrak{L}(M)$ in terms of the number of activities that are enabled at certain points in the process, w.r.t. the number of activities seen that were actually observed at these points in the process. With the log and model containing fewer activities after filtering, the number of enabled activities is likely to decrease as well. Secondly, activity filtering leads to log $L'$ that contains less behavior than original log $L$ (i.e., $\tilde{L'}$ is smaller than $\tilde{L}$), this makes it easier for process discovery methods to discover a process model with less behavior. These two factors make precision values between event logs with different numbers of activities filtered out incomparable. The degree to which the behavior of filtered log L′ decreases w.r.t. an unfiltered log L depends on the activities that are filtered out: when very chaotic activities are filtered from L the behavior decreases much more than when very structured activities are filtered from L. One effect of this is that too much behavior in a process model affects the precision of that model more for the log from which the non-chaotic activities are filtered out than for the log from which the chaotic activities are filtered out.

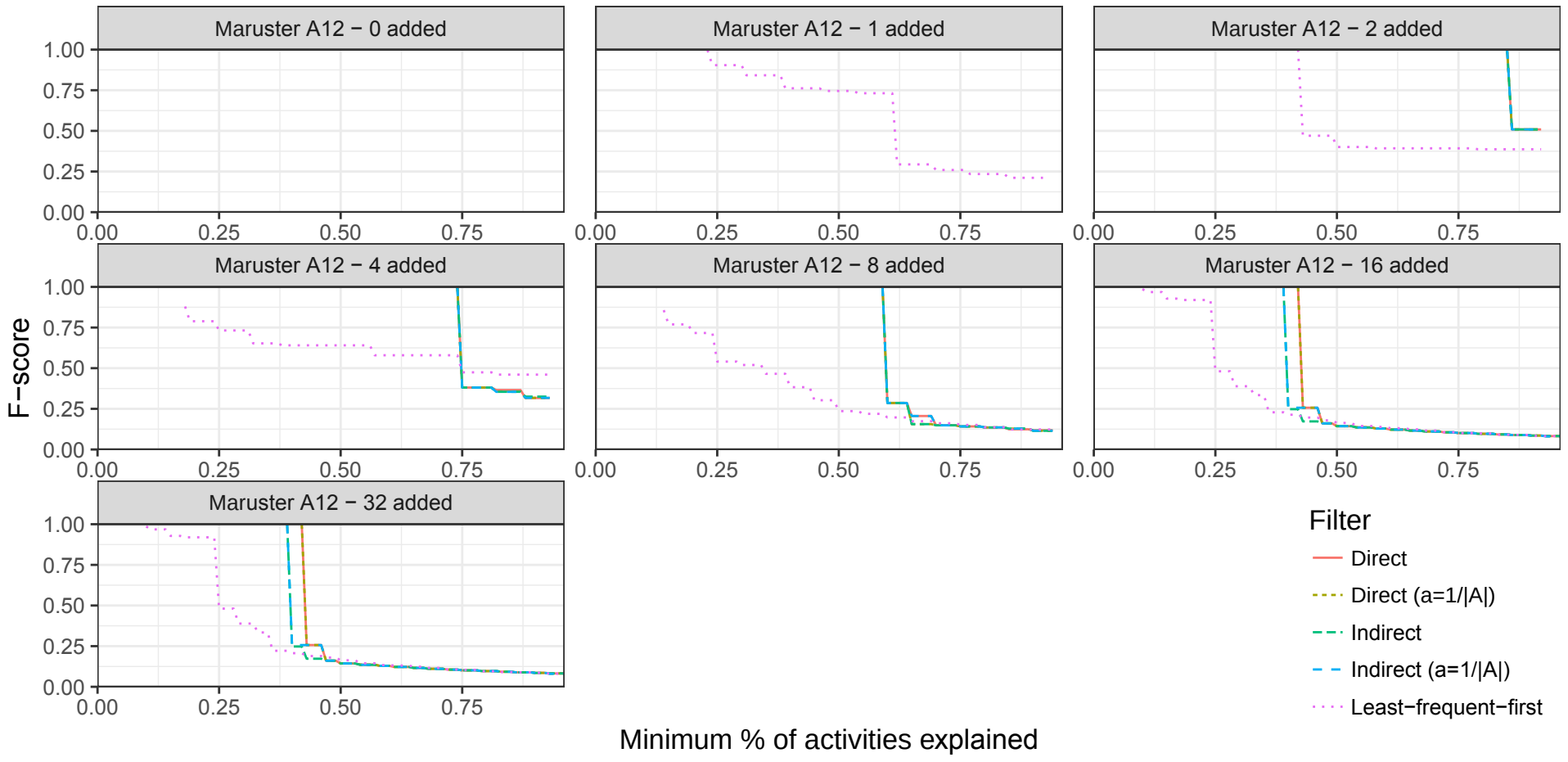

**Figure 6.7:** The F-score for the log generated from the Maruster A12 model with inserted artificial chaotic activities.

To measure the behavior that is allowed by the process model independent of which activities are filtered from the event log we determine the average number of enabled activities when replaying the traces of the log on the model. For an alignment $\Gamma_M(\sigma)$ of trace $\sigma$ on process model M, let $\Gamma^t_M(\sigma)$ be the sequence consisting of only the synchronous moves on visible transitions and let $\breve{\Gamma}^t_M(\sigma)$ denote the sequence of markings prior to each firing of a visible transition in $\Gamma^t_M(\sigma)$. Given a marking $m \in \mathcal{B}(P)$ we define the nondeterminism of that marking to be the number of reachable visible transitions that can be fired as first next visible transition from m, i.e., $\mathit{nondeterminism}(m) = |\{a \in \Sigma | m \xrightarrow{\gamma} m_i \wedge t \in \gamma \wedge l(t) = a \wedge \forall_{\gamma_i \in \gamma} \gamma_i \in \mathit{dom}(l) \implies \gamma_i = t\}|$. We define the nondeterminism of a model $M \in \mathcal{M}$ given a trace $\sigma \in \Sigma^+$ as the average nondeterminism of the markings $\breve{\Gamma}^t_M(\sigma)$ and define the nondeterminism for a model M and a log L as the average nondeterminism over the traces of L.

Figure 6.7 shows the F-scores measured for different percentages of activities filtered out from the Maruster $L_{A12}$ log with different numbers of uniform chaotic activities added. Note that the line stops when further removal of activities does not lead to further improvement in F-score. Note that on the original event log with

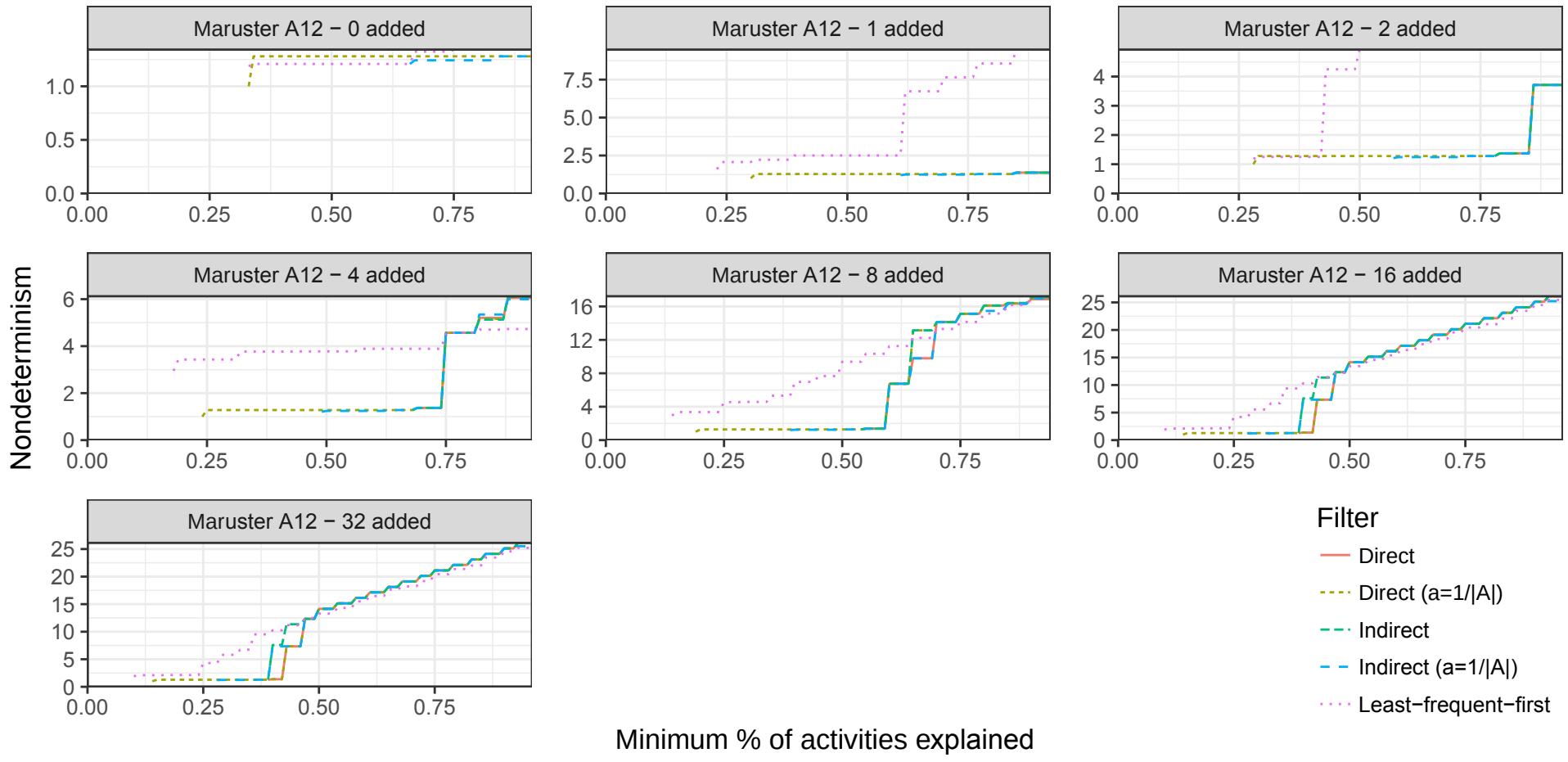

**Figure 6.8:** The nondeterminism on the log generated from the Maruster A12 model with inserted artificial chaotic activities.

0 chaotic activities added the F-score on the original log is already 1.0, resulting in no lines being drawn. With one chaotic activity added, the least-frequent-first filter needs to remove 75% of the activities before it ends up with F-score 1, which can be explained by the fact that 9 out of 12 non-chaotic needed to be removed in order with the least-frequent-first filter to remove all uniform chaotic activities, as shown in Table 6.1. All entropy-based activity filtering techniques remove the chaotic activity in the first filtering step, immediately leading to an F-score of 1.0. Up until 8 added chaotic activities there is no difference between the entropy-based activity filtering techniques in terms of F-score of the resulting process models, which is consistent with the fact that all these filtering techniques were found to filter without errors for these number of inserted chaotic activities in Table 6.1. For 16 and 32 activities, the direct filtering methods outperform the indirect filtering methods, consistent with the fact that the indirect approach made one filtering error according to the ground truth for these numbers of added chaotic activities. Note that the least-frequent-first filter is outperformed by the entropy-based filtering methods in terms of F-score of the discovered models, as would be expected given the filtering results according to the ground truth.

Figure 6.8 shows the results in terms of nondeterminism measured for different percentages of activities filtered out from the Maruster $L_{A12}$ log with various numbers of uniform chaotic activities added. The results show very clearly that when filtering out a number of activities that is identical to the number of added chaotic activities (this corresponds to 92% for one added activity, 86% for two added activities, 75% for 4 added activities, 60% for 8 added activities, 43% for 16 added activities, and 27% for 32 added activities), the nondeterminism reaches a value of 1.5, which is the nondeterminism value of the model discovered from the original log without added chaotic activities. The least-frequent-first filter, however, leads to process models where many activities are enabled on average, therefore overgeneralizing the process behavior, as an effect of filtering out nonchaotic activities instead of the added chaotic activities.

## 6.3 Evaluation using Real Life Data

For the experiments on real-life event logs we do not artificially insert chaotic activities to event logs, but instead filter directly on the activities that are present in these logs. Whether these logs contain chaotic activities that impact process discovery results is not known upfront. Therefore, we apply different activity filtering techniques to these logs and use them to filter out a varying number of activities, after which we assess the quality of the process model that is discovered from these filtered logs. Table 6.2 gives an overview of the real-life event logs that we use in the experiment. In total, we include five event logs from the business domain. Furthermore, we include twelve event logs that contain events of human behavior, recorded in smart home environments or through wearable devices. Mining process model descriptions of daily life is a novel application of process mining that has recently gained popularity [Dim+16; LMM15; Szt+15; Tax+16a; TGZ17]. Furthermore, human behavior event data are often challenging for process discovery because of the presence of highly chaotic activities, like *going to the toilet*. We perform the experiments with activity filtering techniques on real-life data with RapidProM [ABZ17], which is an extension that adds process mining capabilities to the RapidMiner platform for repeatable scientific workflows.

For each event log, we apply seven different activity filtering techniques for comparison: 1) direct entropy filter without Laplace smoothing, 2) direct entropy filter with Laplace smoothing ($\alpha = \frac{1}{|Activities(L)|}$), 3) indirect entropy filter without Laplace smoothing, 4) indirect entropy filter with Laplace smoothing ($\alpha = \frac{1}{|Activities(L)|}$), 5) least-frequent-first filtering, 6) most-frequent-first filtering, 7) filtering the activities from the log in a random order. Recall that the activity filtering procedure stops at the point where all but two activities are filtered from the event log because process models that contain just one activity do not communicate any information regarding the relations between activities. For each event log and for each activity filtering approach we discover a process model after each filtering step (i.e., after each removal of an activity). The process discovery step is performed with two process discovery approaches: the Inductive Miner [LFA13b], and the Inductive Miner infrequent (20%) [LFA13a].

**Table 6.2:** An overview of the event logs used in the experiments

| Name | Category | # traces | # events | # activities |
|---|---|---|---|---|
| BPI'12 [Don12] | Business | 13087 | 164506 | 23 |
| BPI'12 resource 10939 [Tax+16e] | Business | 49 | 1682 | 14 |
| Environmental permit [Bui14] | Business | 1434 | 8577 | 27 |
| SEPSIS [MT17] | Business | 1050 | 15214 | 16 |
| Traffic Fine [LM15] | Business | 150370 | 561470 | 11 |
| Bruno [Bru+13] | Human behavior | 57 | 553 | 14 |
| CHAD 1600010 [McC+00] | Human behavior | 26 | 238 | 10 |
| MIT A [TIL04] | Human behavior | 16 | 2772 | 27 |
| MIT B [TIL04] | Human behavior | 17 | 1962 | 20 |
| Ordonez A [OTS13] | Human behavior | 15 | 409 | 12 |
| van Kasteren [Kas+08] | Human behavior | 23 | 220 | 7 |
| Cook hh102 labour [Coo+13] | Human behavior | 18 | 576 | 18 |
| Cook hh102 weekend [Coo+13] | Human behavior | 18 | 210 | 18 |
| Cook hh104 labour [Coo+13] | Human behavior | 43 | 2100 | 19 |
| Cook hh104 weekend [Coo+13] | Human behavior | 18 | 864 | 19 |
| Cook hh110 labour [Coo+13] | Human behavior | 21 | 695 | 17 |
| Cook hh110 weekend [Coo+13] | Human behavior | 6 | 184 | 14 |

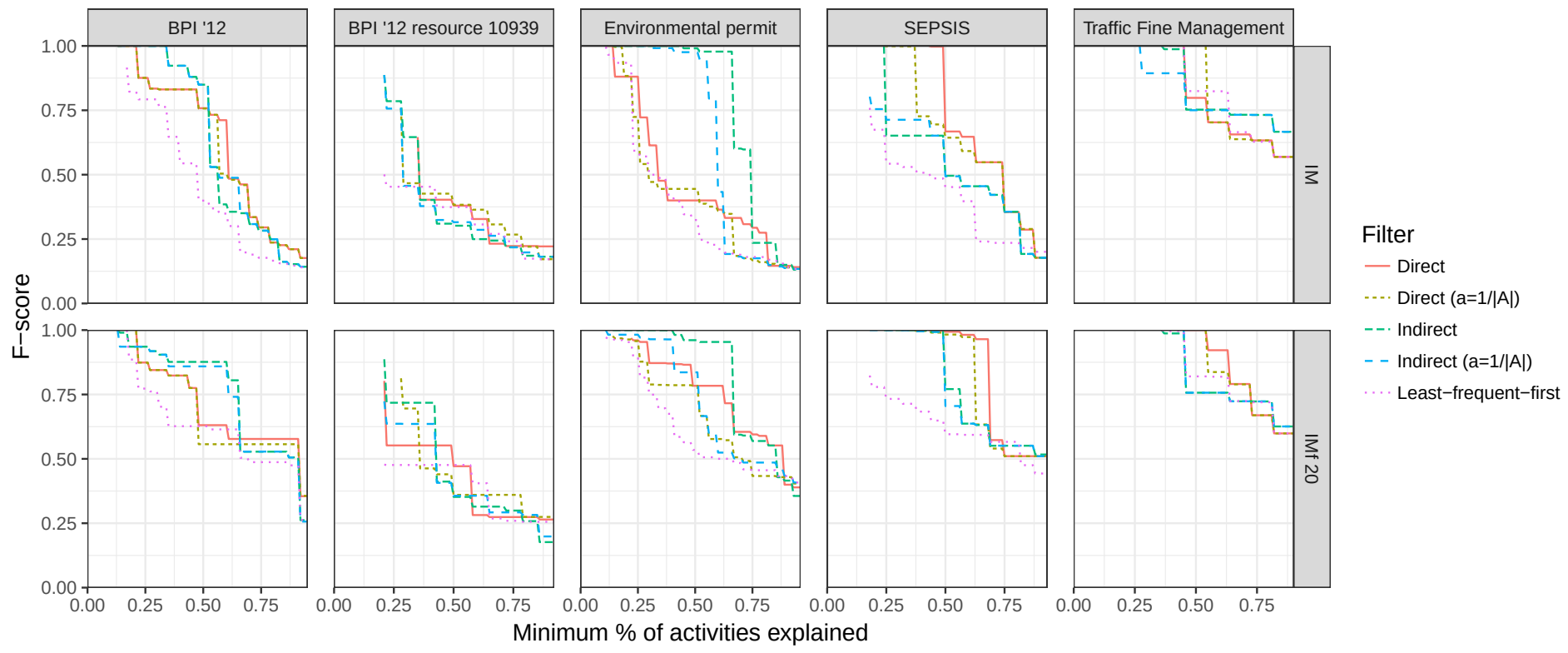

**Figure 6.9:** F-score on business logs dependent on the minimum share of activities remaining.

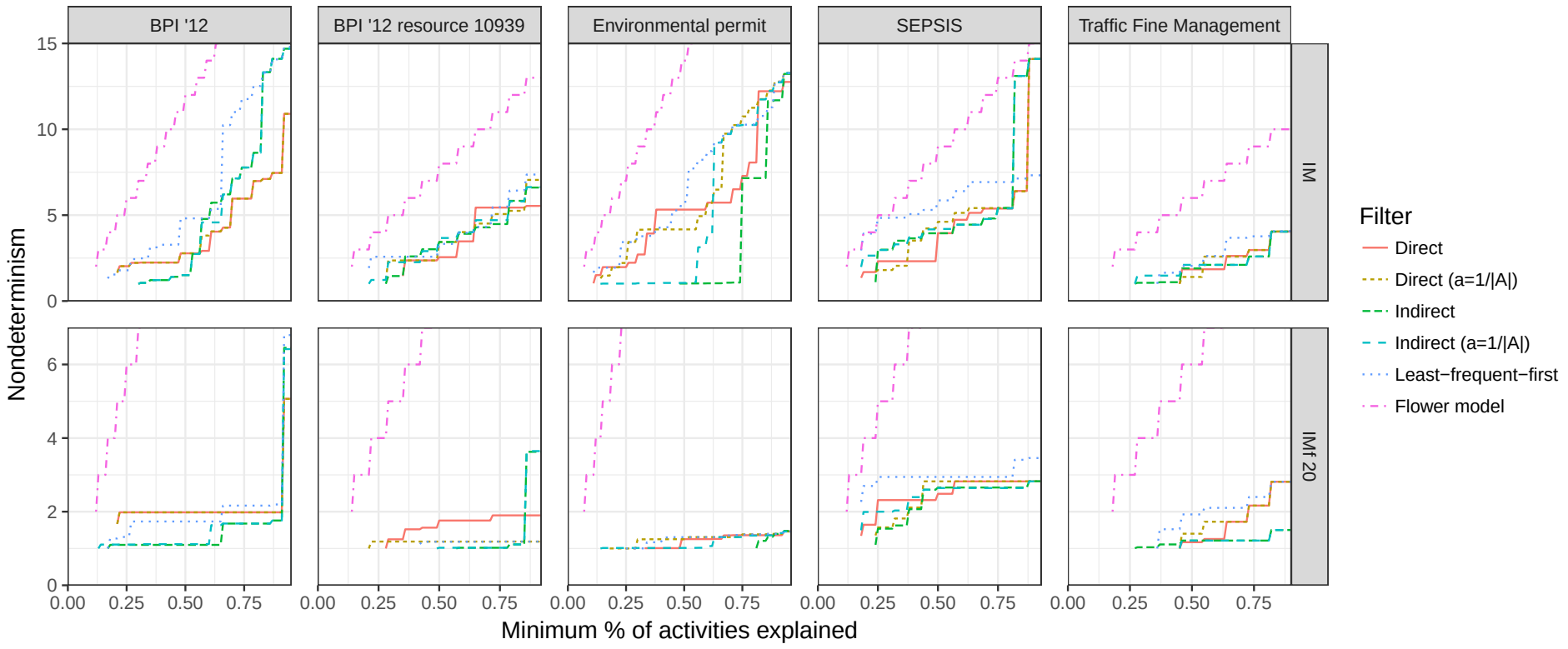

**Figure 6.10:** Nondeterminism on business behavior logs dependent on the minimum share of the activities remaining.

### *Results on Business Process Event Logs*

Figure 6.9 shows the F-score of the process models discovered with the Inductive Miner [LFA13b] and the Inductive Miner with infrequent behavior filtering [LFA13a] (20% filtering) on the five business event logs for different percentages of activities filtered out and different activity filtering techniques. The figure shows an increasing trend in F-score for all event logs when more activities are filtered from the event log. Furthermore, the line for the least-frequent-first filtering approach is below the lines of the entropy-based filtering techniques for most of the percentages of activities removed on most event logs, which shows that entropy-based filtering enables the discovery of models with higher F-score compared to simply filtering out infrequent activities. There are a few exceptions where filtering out infrequent activities outperforms the entropy-based techniques, e.g., the Inductive Miner on the BPI ’12 resource 10939 event log (around 40% of activities explained) and the traffic fines event log (around 55% of activities explained). It differs between event logs which of the entropy-based techniques performs best: for the environmental permit log the indirect filter without Laplace smoothing almost dominates the other techniques while for the SEPSIS log the direct filter without Laplace smoothing outperforms the other techniques. Generally, it seems that the use of Laplace smoothing harms F-score, as most parts of the lines of indirect filtering with Laplace smoothing are below the lines of the indirect approach without Laplace smoothing, and similar for the direct approach with and without Laplace smoothing. However, the detrimental effect of Laplace smoothing does not seem to be large, and in some cases, the usage of Laplace smoothing in filtering increases the F-score of the discovered models.

Figure 6.10 shows the nondeterminism of the process models as a function of the minimum percentage of activities. The green dashed line indicates the nondeterminism of the flower model, i.e., the process model that allows for all behavior over the activities. The lines stop when further removal of activities does not lead to further improvement of nondeterminism. It is clear that the filtering mechanism of the Inductive Miner helps to discover process models that are more behaviorally constrained, as the nondeterminism values are lower for the Inductive Miner infrequent 20% compared to the Inductive Miner without filtering. However, the results show even when already using the 20% frequency filter of the Inductive Miner infrequent, the chaotic activity filter can lead to an additional reduction of nondeterminism. Furthermore, the results on the environmental permit log and the SEPSIS log show that filtering several chaotic activities from the event log also enables the discovery of a model with low nondeterminism using the Inductive Miner without filtering. Which of the activity filtering approaches works best seems to be dependent on the event log: the indirect entropy-based filter leads to the models with the lowest nondeterminism on the traffic fine event log, the environmental permit event log, while the direct entropy-based filter works better for some percentages of remaining activities for the SEPSIS log and the BPI ’12 resource 10939 log.

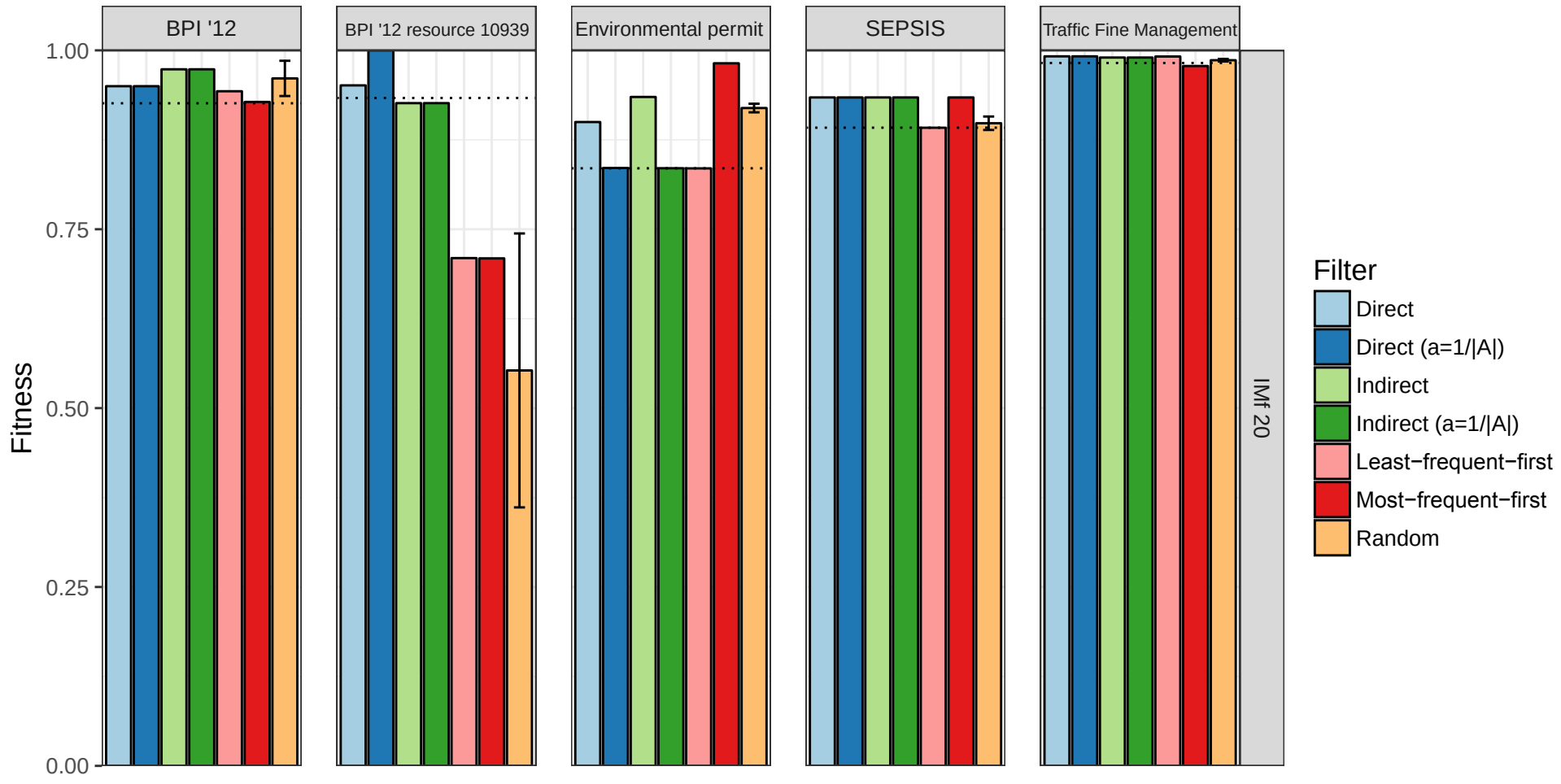

**Figure 6.11:** Fitness on business logs with least 75% of the activities remaining.

Figures 6.11 and 6.12 show the fitness and precision values for the business process event logs at the filtering step that leads to the highest F-score while describing at least 75% of the activities of the original log. In addition to the filtering techniques shown in Figure 6.9 it also shows the frequency-based activity filter where the *most* frequent activities are filtered out first, and a random baseline is shown which iteratively picks a random activity from the event log to filter out. The error bar for the random activity filter indicates one *standard error of the mean (SEM)* based on eight repetitions of applying the filter. The black dotted horizontal lines indicate the fitness and precision values of the process models discovered from the original event log without filtering any activities. Note that the fitness values are only shown for the Inductive Miner infrequent 20% [LFA13a] because the Inductive Miner without infrequent behavior filter [LFA13b] provides the formal ntee that the fitness of the discovered model is 1. Figure 6.11 shows that generally, the differences in fitness between the models discovered from the filtered logs are very minor, and very close to the fitness of the unfiltered log (i.e., the dotted line). Figure 6.12, however, shows that the entropy-based filtering approaches outperform filtering out activities based on frequency and filtering out random activities from the event log. The F-scores of the discovered process models is determined mostly by the precision of the models because the activity filtering impacts precision more than it impacts fitness. One exception is the BPI'12 resource 10939 log [Tax+16e], where the fitness decreases to below 0.75 as a result of applying one of the two frequency-based filters, while the precision increase as an effect of applying the filter is only minor.

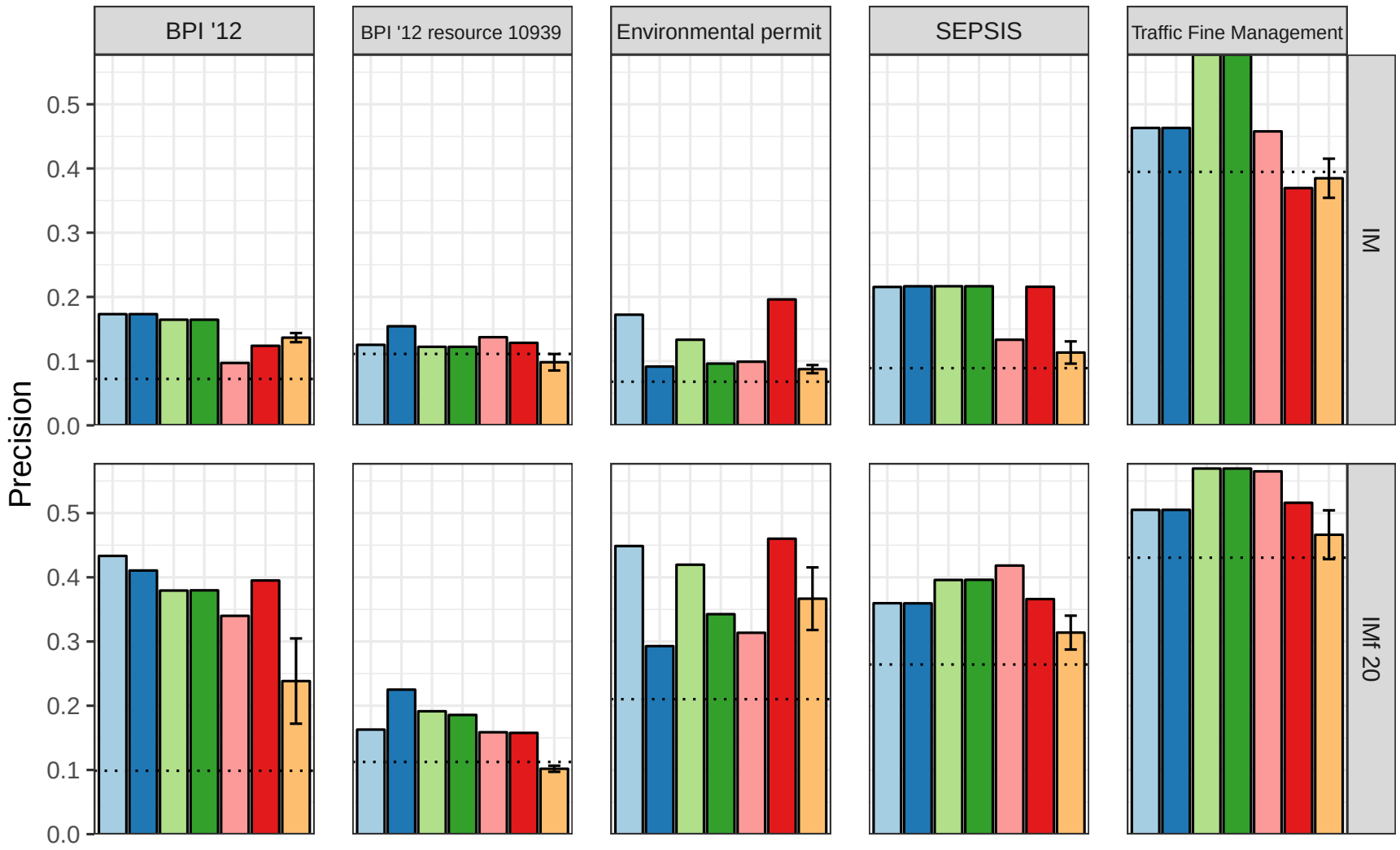

**Figure 6.12:** Precision on business logs with least 75% of the activities remaining.

### *Results on Human Behavior Event Logs*

Figure 6.13 shows the maximum F-score for different human behavior event logs as a function of the minimum percentage of activities that are remaining in the log. Again, the general pattern is that the F-score of the discovered process model decreases when the minimum percentage of events explained increases, as the process discovery task gets easier for smaller numbers of activities. The figure shows that filtering infrequent activities from the event log is dominated in terms of F-score by the entropy-based filtering techniques. Like on the business process event logs, there are mixed results on which of the four configurations of the entropy-based filtering technique leads to the highest F-score: on the CHAD event log the indirect activity filter outperforms the direct activity filter when using the Inductive Miner infrequent 20%; however, the direct activity filter leads to higher F-score for the Inductive Miner when filtering more than 50% of the activities.

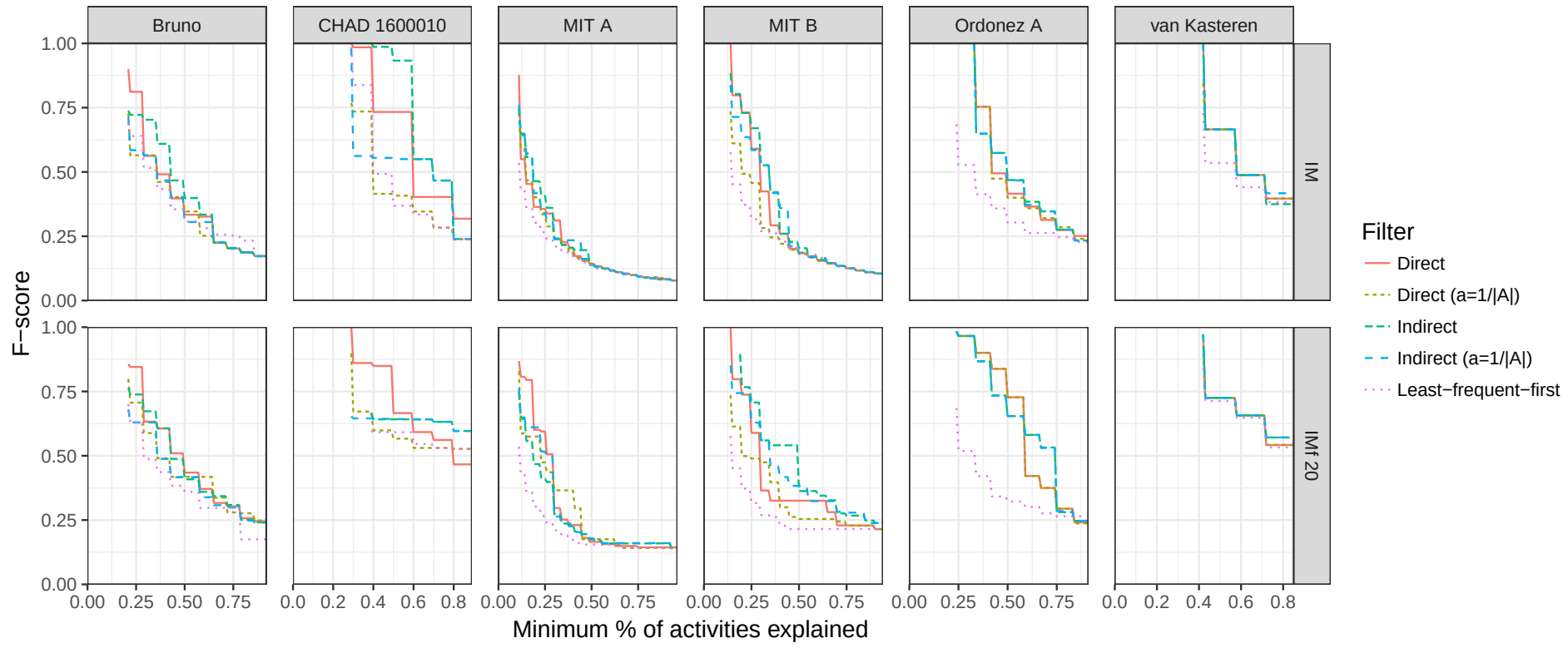

**Figure 6.13:** F-score on human behavior logs dependent on the minimum share of activities.

Figure 6.14 shows the nondeterminism results for the human behavior event logs. It is noticeable that the nondeterminism values of the process models that are discovered when filtering very few activities are much closer to the flower model compared to what we have seen before for the business process event logs. This is caused by human behavior event logs having much more variability in behavior compared to execution data from business processes, resulting in a much harder process discovery task. After filtering several chaotic activities, the nondeterminism drops significantly to ranges comparable to nondeterminism values seen for logs from the business process domain. This shows that the problem of chaotic activities is much more prominent in human behavior event logs than in business process event logs. The entropy-based activity filtering approaches lead to more deterministic process models compared to filtering out infrequent activities. Two clear examples of this are the MIT B log and the Ordonez A log, on which filtering out infrequent activities after several filtering steps results in a flower model (i.e., nondeterminism is identical to that of the flower model), while entropy-based activity filters enable the discovery of a model with nondeterminism close to one (i.e., very close to a sequential model) while at the same time keeping 75% of the activities in the event log.

Figure 6.15 shows the precision values for the human behavior logs for the filtering step that leads to the highest F-score while describing at least 50% of the activities of the original log. Similarly to what we have seen in the nondeterminism graph, removing random activities from the log and removing infrequent activities from the log results in smaller precision increases compared to the entropy-based activity filters. Furthermore, it is noticeable that removing frequent activities from the log works quite well to improve the precision of models discovered from the human behavior application domain. The reason for this is that some of the chaotic activities that are present in many of those event logs, including *going to the toilet* and *getting a drink*, also happen to be frequent. On the van Kasteren event log the indirect activity filter with Laplace smoothing leads to the largest increase in precision when mining a model with at least 50% of the activities (from 0.324 to 0.732 with the Inductive Miner infrequent 20%).

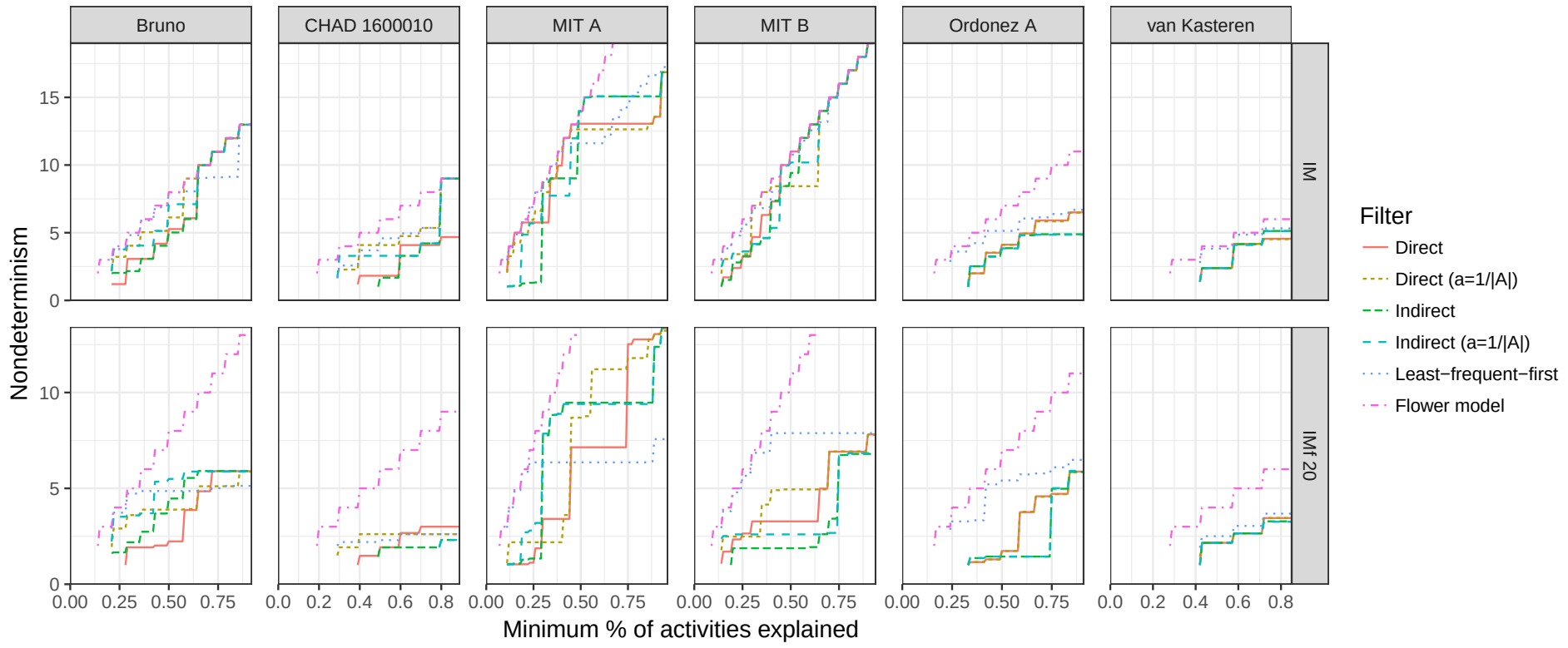

**Figure 6.14:** Nondeterminism on human behavior logs dependent on the minimum share of the activities remaining.

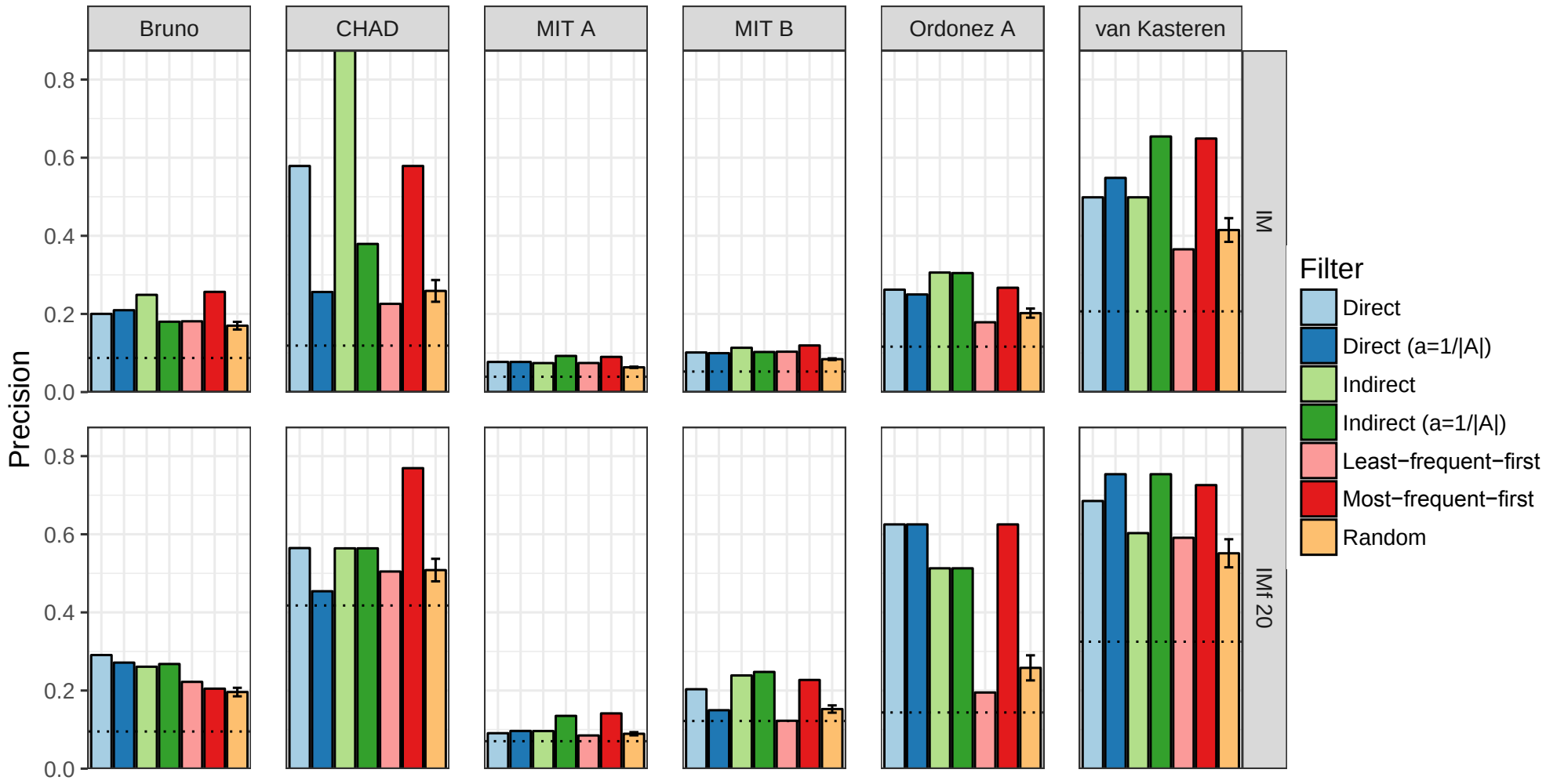

**Figure 6.15:** Precision on human behavior logs with at least 50% of the activities.

**Table 6.3:** Left: the order in which activities are filtered using the direct activity filter with Laplace smoothing ($\alpha = \frac{1}{|Activities(L)|}$) on the van Kasteren log. Right: the order in which the activities are filtered using the least-frequent-first filter.

| Order | Filtered activity (indirect entropy-based filter with Laplace smoothing) | Filtered activity (least-frequent-first filter) |
|---|---|---|
| 1 | Use toilet | Prepare dinner |
| 2 | Get drink | Get drink |
| 3 | Leave house | Prepare breakfast |
| 4 | Take shower | Take shower |
| 5 | Go to bed | Go to bed |
| 6 | Prepare breakfast | Leave house |
| 7 | Prepare dinner | Use toilet |

Table 6.3 shows in which order activities are filtered from the van Kasteren event log by 1) the indirect entropy-based activity filter with Laplace smoothing and 2) the least-frequent-first filter. It shows that the entropy-based filter filters *use toilet* as the first activity, which from domain knowledge we know to be a chaotic activity, as people generally just go to the toilet whenever they need to, regardless of which other activities they have just performed. For the infrequent activity filter *use toilet* would be the last choice of the activities to filter out, because it is the most frequent activity in the van Kasteren event log.

Figures 6.16a and 6.16b show the corresponding process models discovered with the Inductive Miner infrequent 20% from the logs filtered with the indirect activity filter with Laplace smoothing and the infrequent activity filter respectively. The process model discovered after filtering three activities with the Indirect entropy-based activity filter with Laplace smoothing is very specific on the behavior that it described: after *going to bed*, either the logging ends, or *prepare breakfast* occurs next, followed by *taking a shower*. After taking a shower, there is a possibility to either *go to bed* again or to *prepare dinner* before *going to bed*. The process model

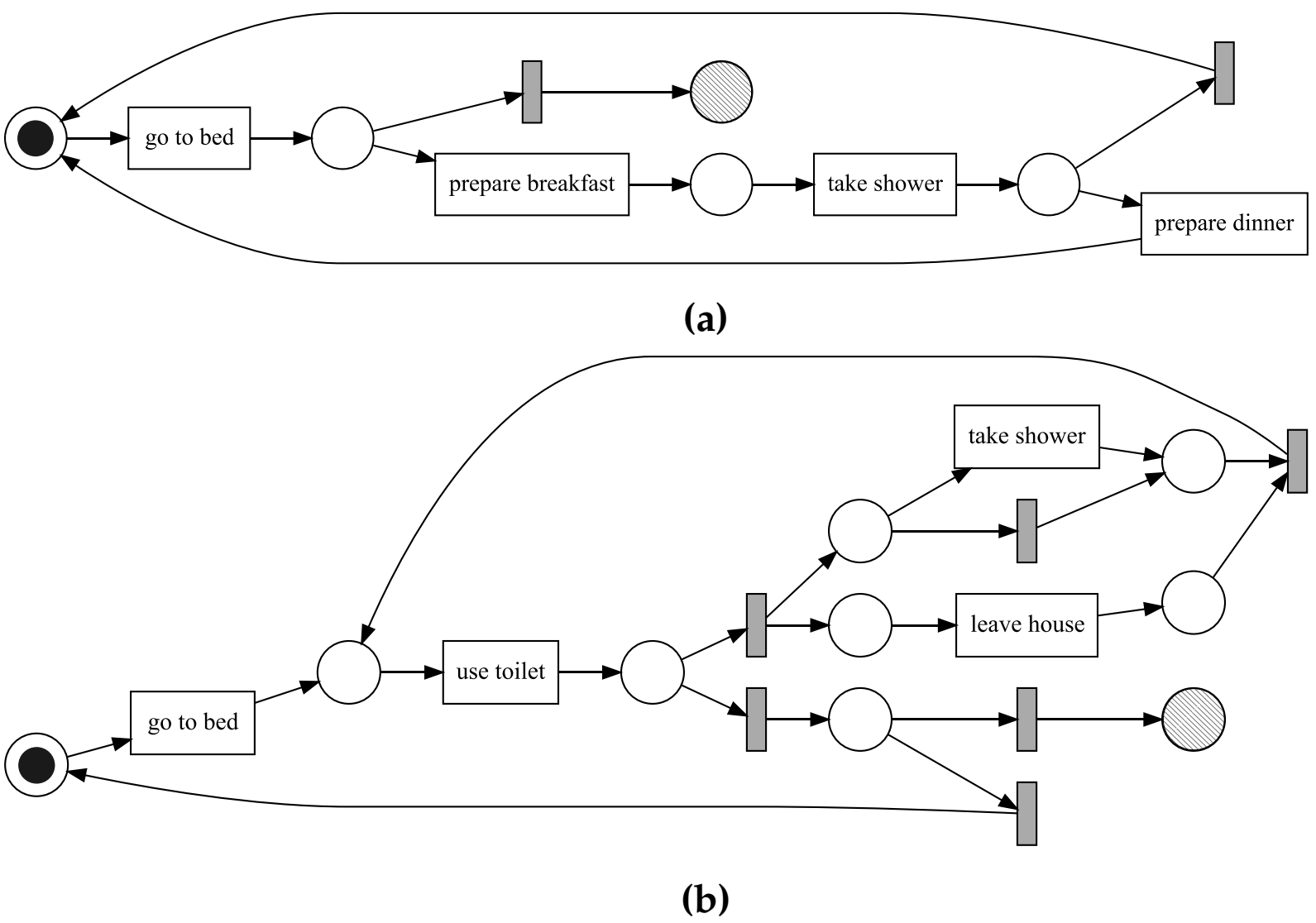

**Figure 6.16:** *(a)* The model discovered with Inductive Miner infrequent 20% on the Van Kasteren log after filtering all but four activities with the indirect approach with Laplace smoothing, and *(b)* the model discovered from the same log with the same miner when filtering all but four activities when filtering out the least frequent activities.

discovered after filtering three activities with the infrequent activity filter allows for many more traces: it starts with *go to bed* followed by *use toilet*, after which any of the activities *go to bed*, *take shower*, and *leave house* can occur as next event or the logging can end. Furthermore, the activities *leave house* and *take shower* can occur in any order, and *take shower* can also be skipped.

Figure 6.17 shows the results on F-score for the human behavior event logs by Cook et al. [Coo+13]. The results on the Cook event logs are in-line with the results on the human behavior event logs, however, on these event logs, it is even more clear that filtering out infrequent activities leads to suboptimal process models in terms of F-score. Which of the filtering approaches results in the optimal process model in terms of F-score is very dependent on the event log and the minimum number of activities to be remained after filtering: each of the four configurations of the entropy-based filtering approach is optimal for at least one combination of log and minimum percentage of activities explained.

Figure 6.18 shows the results in terms of nondeterminism for the same event logs. Filtering infrequent activities at high percentages of activities explained has much lower nondeterminism compared to the flower model, while further left on the graph, after filtering out more activities, the nondeterminism of filtering out infrequent activities gets closer to the flower model. This shows that filtering out infrequent activities can even be harmful to the quality of the obtained process discovery result. The nondeterminism values obtained with the four configurations of the entropy-based filtering approach are generally close to each other, where the optimal configuration is dependent on the log and the number of filtered activities.

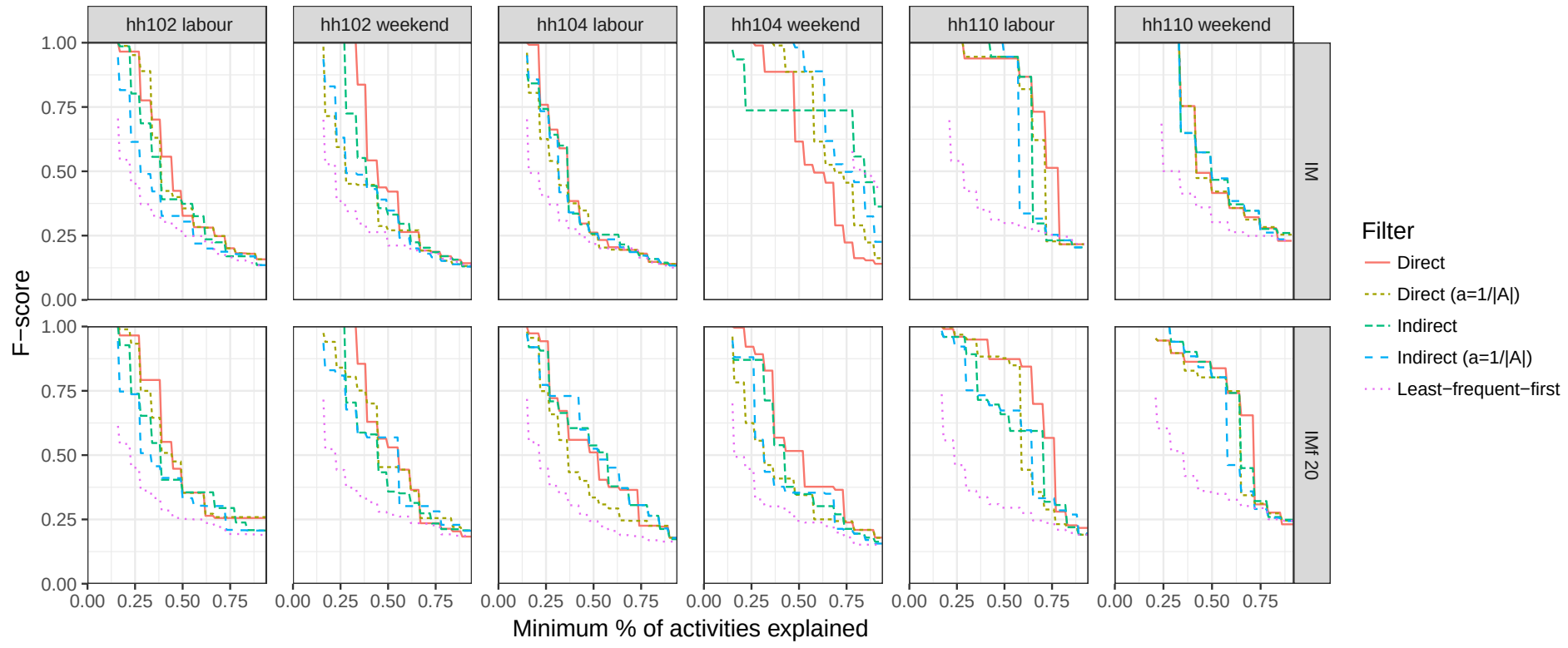

**Figure 6.17:** F-score on cook's human behavior logs dependent on the minimum share of the activities remaining.

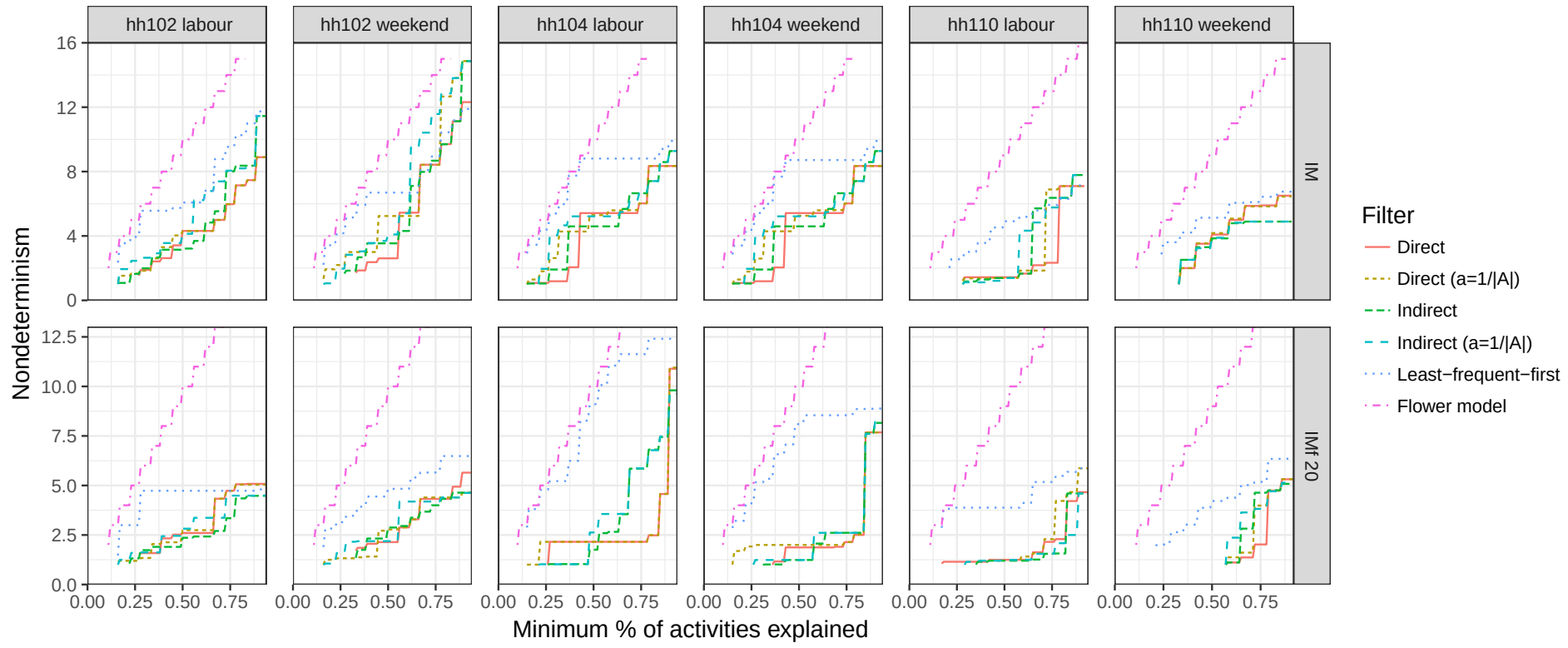

**Figure 6.18:** Nondeterminism on cook's human behavior logs dependent on the minimum share of the activities remaining.

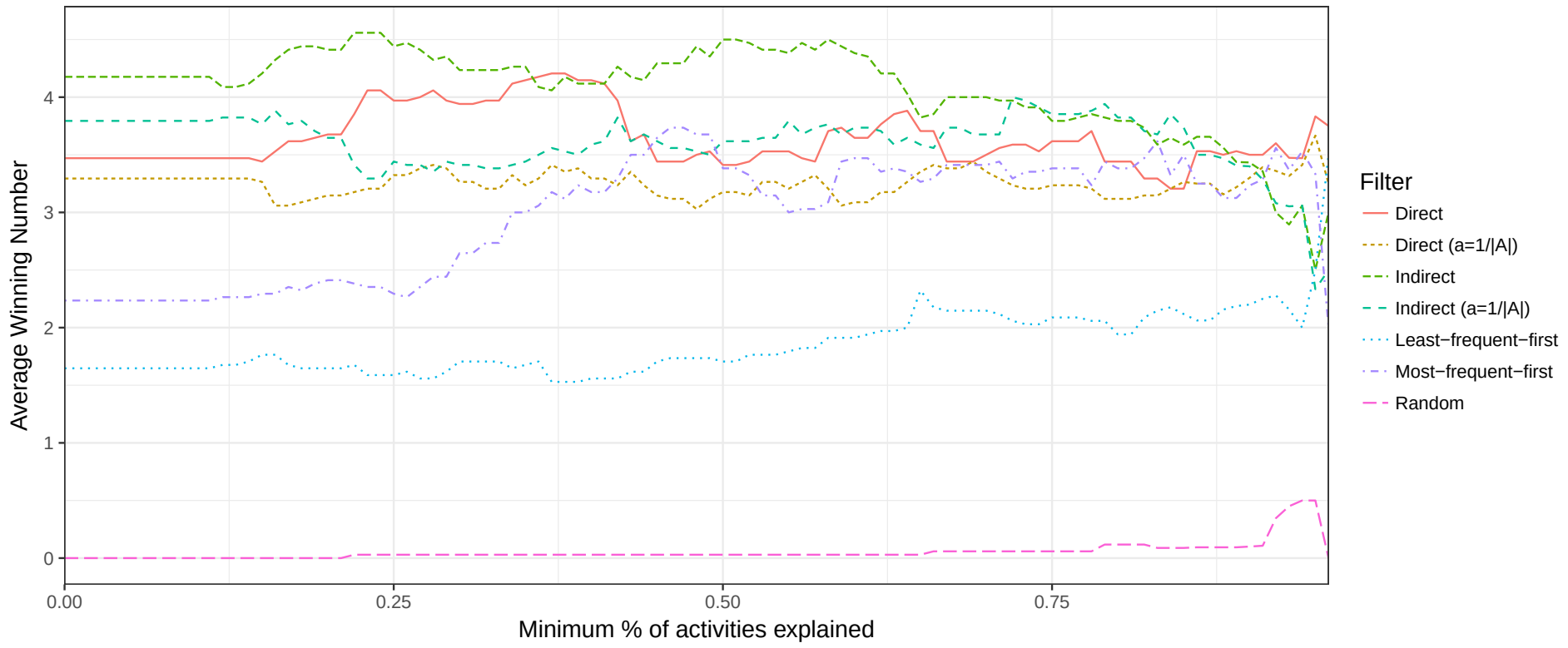

**Figure 6.19:** The average winning number for the seven activity filtering techniques dependent on the minimum ratio of activities explained, averaged over the 17 event logs used in the experiment.

**Table 6.4:** Kendall $\tau_b$ rank correlation between five activity filtering methods, mean and standard deviation over the 17 event logs.

| | Direct | Direct ($\alpha{=}\frac{1}{\lvert A\rvert}$) | Indirect | Indirect ($\alpha{=}\frac{1}{\lvert A\rvert}$) | Least-frequent-first |
|---|---|---|---|---|---|
| Direct | 1.0 | 0.2956 | 0.0829 | 0.1408 | 0.0504 |
| Direct ($\alpha{=}\frac{1}{\lvert A\rvert}$) | 0.2956 | 1.0 | 0.0698 | 0.0536 | 0.1454 |
| Indirect | 0.0829 | 0.0698 | 1.0 | 0.6852 | -0.0275 |
| Indirect ($\alpha{=}\frac{1}{\lvert A\rvert}$) | 0.1408 | 0.0536 | 0.6852 | 1.0 | -0.0392 |
| Least-frequent-first | 0.0504 | 0.1454 | -0.0275 | -0.0392 | 1.0 |

### *Aggregated Analysis Over All Event Logs*

We have observed in Figures 6.10, 6.14, and 6.18 that the entropy-based activity filtering techniques perform differently on different datasets and for different numbers of activities filtered. To evaluate the overall performance of activity filtering techniques, we use the number of other filtering techniques that it can beat over all the seventeen event logs of Table 6.2. This metric, known as *winning number*, is commonly used for evaluation in the Information Retrieval (IR) field [Qin+10; TBH15]. Formally, winning number is defined as

$$W_i^x = \sum_{j=1}^{17} \sum_{k=1}^{7} \mathbb{1}_{\{N_i^x(j)<N_k^x(j)\}}$$

**Table 6.5:** Number of event logs for which we can reject the null hypothesis that the orderings of activities returned by activity filters are uncorrelated, according to the tau test.

| | Direct | Direct ($\alpha=\frac{1}{\|A\|}$) | Indirect | Indirect ($\alpha=\frac{1}{\|A\|}$) | Least-frequent-first |
|---|---|---|---|---|---|
| Direct | 17 | 5 | 1 | 2 | 0 |
| Direct ($\alpha=\frac{1}{\|A\|}$) | 5 | 17 | 1 | 1 | 3 |
| Indirect | 1 | 1 | 17 | 17 | 3 |
| Indirect ($\alpha=\frac{1}{\|A\|}$) | 2 | 1 | 17 | 17 | 3 |
| Least-frequent-first | 0 | 3 | 3 | 3 | 17 |

where j is the index of an event log, i and k are indices of activity filtering techniques, $N_i^x(j)$ is the performance of the i-th algorithm on the j-th event log in terms of nondeterminism where each least x% of activities are explained and $\mathbb{1}_{\{N_i^x(j)<N_k^x(j)\}}$ is the indicator function

$$\mathbb{1}_{\{N_i^x(j)<N_k^x(j)\}} = \begin{cases} 1, & \text{if } N_i^x(j) < N_k^x(j), \\ 0, & \text{otherwise.} \end{cases}$$

We define $\overline{W}_i^x = \frac{W_i^x}{17}$ as the average number of other activity filtering techniques that are outperformed by filtering technique i at the point where at least x% of activities are explained.

Figure 6.19 shows the average winning number $\overline{W}_i^x$ for different values of x and for the seven different activity filtering techniques. We observe that for higher ratios of activities explained the differences between filtering techniques are smaller than for lower numbers of activities explained. Intuitively this can be explained by the fact that for lower ratios of activities explained more activities have been filtered out from the log. Therefore the effect of the filtering techniques is more clearly visible. The figure shows that, up until +-74% of activities explained, the indirect entropy-based activity filtering technique leads to the most deterministic process models averaged over all event logs included in the experiment, where it outperforms between 4 and 4.5 other filtering techniques. Between +-75% and +- 87.5% the indirect entropy-based activity filtering technique with Laplace smoothing results in the highest average winning number, although the difference with the indirect entropy-based filtering technique seems negligible. Filtering out random activities from the event log outperforms none of the 6 other activities filtering techniques for the most of the graph, indicating that frequency-based filtering clearly outperforms filtering random activities.

To investigate to what degree the order in which activities are removed from the logs differs between the activity filtering techniques we calculate Kendall's tau ($\tau_b$) rank correlation for each log between the activity filtering techniques in a pairwise way. Table 6.4 shows the rank correlation values found between the activity filters, averaged over the 17 event logs. The indirect activity filter with Laplace smoothing and the indirect activity filter without Laplace smoothing generate orderings over the activities of a log that are strongly correlated. Between the direct activity filter without Laplace smoothing and the direct activity filter without Laplace smoothing there is only a weak correlation. All the other activity filtering techniques are uncorrelated or very weakly correlated. Using the Kendall $\tau_b$ statistic, we apply a tau test for each pair of activity filtering techniques on each event log to test the null hypothesis that the two orderings in which activities are filtered by the two activity filtering techniques are uncorrelated, using a significance level $\alpha = 0.05$.

For each pair of activity filtering techniques Table 6.5 shows the number of event logs for which the null hypothesis was rejected, i.e., the number of event logs for which the order in which activities are filtered is statistically correlated. The indirect activity filters with and without Laplace smoothing create correlated orderings of activities for all seventeen event logs. For all other pairs of activity filtering techniques the orderings in which activities are filtered are only correlated with for low numbers of event logs.

## 6.4 Entropy-based Toggles for Process Discovery

In the previous section we have shown that all four configurations of the entropy-based activity filtering technique lead to more deterministic process models compared to simply filtering out infrequent activities. However, the differences in determinism of the process models that are discovered after applying any of the four configurations are small and dependent on the event log to which they are applied. Furthermore, all four configurations of the activity filtering technique simply impose an ordering over the activities, but do not specify at which step the filtering should be stopped. Additionally, the proposed filtering technique ignores the semantics of activities: activities that are chaotic may still be relevant for the process. Leaving them out of the process model to discover will harm the usefulness of the discovered process model.

To address the three issues we propose to use the filtering technique as a sorting technique over the activities in combination with toggles that interactively allow the process analyst to “disable” (filter out) or “enable” activities, and then rediscover and visualize the process model according to the new settings. This approach is similar to the Inductive Visual Miner [LA14], an interactive implementation of the Inductive Miner [LFA13a] algorithm which allows the process analyst to filter the event log interactively using a slider-based approach. The Inductive visual miner contains two sliders: with one slider activities can be filtered using the least-frequent-first filter, where the user can control how many activities are filtered out by moving the slider up and down. We propose to replace this slider with a sorted list of activities and toggles, as this allows the process analyst to override the ordering of the activities that is determined by the activity filtering technique with domain knowledge. Figure 6.20 shows a mockup of the proposed way to use the activity filter. Activities are by default sorted using the chaotic activity filter, showing the entropy to indicate the assessed degree of chaoticness of each activity. Based on this information, the process analyst can choose to rely on the filtering technique and filter out the top of the list or to override this list with domain knowledge. Furthermore, other activity filtering techniques, such as the least-frequent-first filter, can be included as an additional column on which the activities of the process can be sorted. This allows the process analyst to control how many activities, and which activities, are filtered out of the process model, and thereby also empowers the user to prevent the removal of semantically important activities that should not be removed. Furthermore, this approach allows the process analyst to explore himself which of the filtering techniques leads to the most useful process model from the event log that he is analyzing.

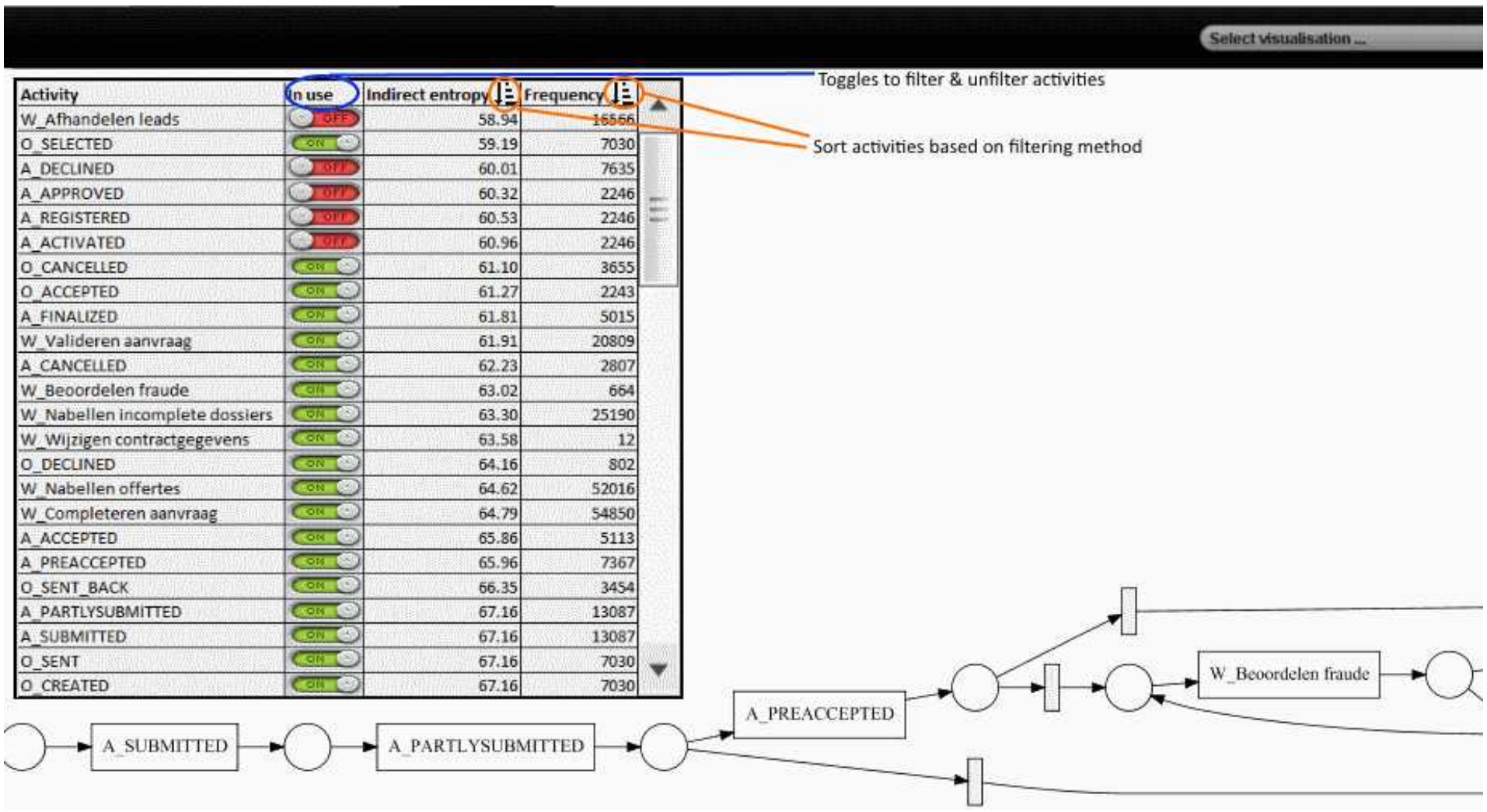

**Figure 6.20:** A mockup of the proposed way to use the activity filters in an interactive setting.

## 6.5 Related Work

Real life events logs often contain all sorts of data quality issues [Sur+17], include incorrectly logged events, events that are logged in the wrong order, and events that took place without being logged. Instances of such data quality issues are often referred to as *noise*. Many event log filtering techniques have been proposed to address the problem of noise. Existing filtering techniques in the process mining field can be classified into four categories: 1) event filtering techniques, 2) process discovery techniques that have an integrated filtering mechanism build in, 3) trace filtering techniques, and 4) activity filtering techniques. We use these categories to discuss and structure related work.

### 6.5.1 Event filtering

Conforti et al. [CRH17] recently proposed a technique to filter out outlier events from an event log. The technique starts by building a prefix automaton of the event log, which is minimal in terms of the number of arcs in the automaton, using an Integer Linear Programming (ILP) solver. Infrequent arcs are removed from the minimal prefix automaton, and finally, the events belonging to removed arcs are filtered out from the event log.

Lu et al. [Lu+16b] advocate the use of *event mappings* [LFA15b] to distinguish between events that are part of the mainstream behavior of a process and outlier events. Event mappings compute similar behavior and dissimilar behavior between each two executions of the process as a mapping: the similar behavior is formed by all pairs of events that are mapped to each other, whereas events that are not mapped are dissimilar behavior.

Fani Sani et al. [FZA18] proposes the use of sequential pattern mining techniques to distinguish between events that are part of the mainstream behavior and outlier events.

All three of the event filtering techniques listed above aim filter out outlier events from the event log, while keeping the mainstream behavior. Event filtering techniques model the frequently occurring contexts of activities and filter out the contexts of activities that occur infrequently in the log. For example, consider an activity B such that 98% of its occurrences are in context $\langle \dots, \mathrm{A}, \mathrm{B}, \mathrm{C}, \dots \rangle$, with the remaining 2% of the events of activity B are in context $\langle \dots, \mathrm{D}, \mathrm{B}, \mathrm{E}, \dots \rangle$, then the B events that occur between D and E will be filtered out by event filtering techniques. Note that our filtering technique is orthogonal to event filtering: it would consider activity B to be nonchaotic and would not filter out anything. However, when a log L contains a chaotic activity X, then event filtering techniques are not able to remove all events of this chaotic activity. One of the contexts of X will by chance be more frequent than other contexts, i.e., for some activity A, it will hold that $\forall \mathrm{B} \in \mathit{Activities}(L) : \#(\langle \mathrm{A}, \mathrm{X} \rangle, \mathrm{L}) > \#(\langle \mathrm{B}, \mathrm{X} \rangle, \mathrm{L})$, even though $\langle \mathrm{A}, \mathrm{X} \rangle$ might only be slightly more frequent. This will result in X events after a B being removed, while the X events after an A remain in the log. Applying a process discovery technique to this filtered log will then result in a process model where activity X is misleadingly positioned after activity A, while in fact X can happen anywhere in the process. The entropy-based activity filtering technique will instead detect that activity X is chaotic, and completely remove it from the event log, preventing the misleading effect of event filtering.

### 6.5.2 Process Discovery Techniques with Integrated Filtering

Several process discovery algorithms offer integrated filtering mechanisms as part of the approach. The Inductive Miner (IM) [LFA13b] is a process discovery algorithm which first discovers a *directly-follows graph* from the event logs, where activities are connected that directly follow each other in the log, from which in a second step a process model is discovered. The directly-follows relations are affected by the presence of a chaotic activity X: sequence $\langle \dots, A, X, C, \dots \rangle$ leads to false directly-follows relations between A and X and between X and C, while the directly-follows relation between A and C is obfuscated by X. The Inductive Miner infrequent (IMf) [LFA13a] is an extension of the IM where infrequent directly-follows relations are filtered out from the set of directly-follows relations that are used to generate to process models. The filtering mechanism of IMf can help to filter out the directly-follows relations between A and X and between X and C, but it does not help to recover the obfuscated directly-follows relation between A and C. Instead, the entropy-based activity filtering technique filters out the chaotic activity X, leading to sequence $\langle \dots, A, X, C, \dots \rangle$ being transformed into $\langle \dots, A, C, \dots \rangle$, thereby recovering the directly follows relation between A and C.

The Heuristics Miner [WR11] and the Fodina algorithm [BD17], in addition to the directly-follows relation, defines an *eventually-follows relation* between activities and allows the process analyst to filter out infrequent directly-follows and eventually follows relations. Two activities A and B are in an eventually-follows relation when A is eventually followed by B, before the next appearance of A or B. The eventually-follows relation, unlike the directly-follows relation, is not impacted by the presence of chaotic activities. The Heuristic Miner [WR11] and Fodina [BD17] both include filtering methods for the directly-follows and eventually-follows relations that are similar in nature to the filtering mechanism that is used in the Inductive Miner infrequent [LFA13a]. However, the use of sequential orderings and parallel constructs in the mining approaches of the Heuristic Miner [WR11] and Fodina [BD17] is based on the directly-follows relations only, with the eventually follows relations being used for the mining of long-term dependencies. Furthermore, in contrast to the Inductive Miner, the process models discovered with the Heuristic Miner [WR11] or Fodina [BD17] can be unsound, i.e., the can contain deadlocks.

The ILP-miner [Wer+09a; Wer+09b] is a process discovery algorithm where a set of behavioral constraints over activities is discovered for each prefix (called the *prefix-closure*) of the event log, based on which a process model is discovered that satisfies these constraints using *Integer Linear Programming* (ILP). Van Zelst et al. [ZDA15] proposed a filtering technique for the ILP-miner where the prefix closure of the event log is filtered prior to solving the ILP problem by removing infrequently observed prefixes. It is easy to see that a chaotic activity X affect the prefix-closure that is discovered from the event log: given log consisting of two traces $\langle A, X, C \rangle$ and $\langle X, A, C \rangle$, activity X causes the prefixes closures of the two traces to have no overlap in states, while without activity X the two traces are identical.

This makes the filtering method of the prefix-closure proposed by Van Zelst et al. [ZDA15] less effective, as frequent prefixes randomly get distributed over several infrequent prefixes when chaotic activities are present. Instead, the chaotic activity filtering technique presented here would remove chaotic activity X, leading to traces $\langle A, X, C \rangle$ and $\langle X, A, C \rangle$ becoming identical after filtering, therefore leading to a simpler process model while still describing the behavior of the event log accurately.

The Fuzzy Miner [GA07] is a process discovery algorithm that aims at mining models from flexible processes, and it discovers a process model without formal semantics. The Fuzzy Miner discovers this graph by extracting the eventually follows relation from the event log, which is not affected by chaotic activities. Similar to the Heuristics Miner [WR11] and Fodina [BD17] the Fuzzy Miner allows to filter out infrequent eventually-follows relations between activities. In practice, the lack of formal semantics of the Fuzzy Miner models hinders the usability of the models, as the models are not precise on what behavior is allowed in the process under analysis.

### 6.5.3 Trace filtering

Ghionna et al. [Ghi+08] proposed a technique to identify outlier traces from the event log that consists of two steps: 1) mining frequent patterns from the event log, and 2) applying MCL clustering [Don08] on the traces, where the similarity measure for traces is defined on the number of patterns that jointly characterize the execution of the traces. Traces that are not assigned to a cluster by the MCL clustering algorithm are considered to be outlier traces and are filtered from the event log. It is easy to see that trace filtering techniques address a fundamentally different problem than chaotic activity filtering: in the event log shown in Figure 6.3b there are only two traces that do not contain an instance of chaotic activity X, therefore, even if a trace filtering technique would be able to perfectly filter out traces that contain a chaotic event, the number of remaining traces will become too small to mine a fitting and precise process model when the chaotic activity is frequent.

### 6.5.4 Activity filtering

The modus operandi for filtering activities is to simply filter out infrequent activities from the event log. The plugin *'Filter Log using Simple Heuristics'* in the ProM process mining toolkit [Don+05] offers tool support for this type of filtering. The Inductive Visual Miner [LA14] is an interactive process discovery tool that implements the Inductive Miner [LFA13b] process discovery algorithm in an interactive way: the process analyst can filter the event log using sliders and is then shown the process model that is discovered from this filtered log. One of the available sliders in the Inductive Visual Miner offers the same frequency-based activity filtering functionality. The working assumption behind filtering out infrequent activities is that when there are just a few occurrences of an activity, there is probably not enough evidence to establish their relation to other activities to model their behavior. However, as we have shown, for frequent but chaotic activities, while they are frequent enough to establish their relation to other activities, complicate the process discovery task by lowering directly-follows counts between other activities in the event log. The entropy-based activity filtering technique is able to filter out chaotic activities, thereby reconstructing the directly-follows relations between the non-chaotic activities of the event log, at the expense of losing the chaotic activities.

## 6.6 Conclusions

In this chapter, we have shown the possible detrimental effect of the presence of chaotic activities in event logs on the quality of process models produced by process discovery techniques. We have shown through synthetic experiments that frequency-based techniques for filtering activities from event logs, which is currently the *modus operandi* for activity filtering in the process mining field, do not necessarily handle chaotic activities well. As shown, chaotic activities can be frequent or infrequent. We have proposed four novel techniques for filtering chaotic from event logs, which find their roots in information theory and Bayesian statistics. Through experiments on seventeen real-life datasets, we have shown that all four proposed activity filtering techniques outperform frequency-based filtering on real data. The indirect entropy-based activity filter has been found to be the best performing activity filter overall averaged over all datasets used in the experiments; however, the performance of the four proposed activity filtering techniques is highly dependent on the characteristics of the event log.

Because the performance of the filtering techniques was found to be log-dependent, we propose the use the activity filtering techniques in a slider-based approach where the user can filter activities interactively and directly see the process model discovered from the filtered event log. Ultimately, only the user can decide which activities to include.

# 7 Abstracting Events to Higher-Level Events using Supervised Learning

Figure 7.1 visualizes the scope of this chapter by highlighting parts of the taxonomy of event log preprocessing methods of Figure 1.4. In Chapter 6 we have addressed a special case of event abstraction by proposing a method for activity filtering. In this chapter, we will discuss a more general method for event abstraction. In Chapter 4 and Chapter 5 we have shown how more precise process models of human behavior can be discovered by taking the name of the sensor as a starting point for the event label and then refine the labels using the time of the day at which the event occurred. However, the labels in such process models still represent sensors, and they have no direct interpretation on the level of human activity. The process models would be more insightful if the transition labels in the process model would be formulated in terms of *human activities* instead of in terms of sensors. One approach to obtaining a process model with transition labels in terms of human activities is by first preprocessing the event log such that event labels are obtained that represent human activities and then applying process discovery to this event log. Note that multiple sensor events might belong to the same event on the human activity level, e.g., a *cooking* activity might result in a sensor event from the *open/close sensor on the fridge*, followed by an event from the *open/close sensor on the cutlery drawer*, then an event from the *power sensor on the water cooker*, and finally an event from the *open/close sensor on the oven*. Therefore, multiple events on the sensor level could be aggregated into a single event on the activity level. This need to abstract sensor-level events to human activity level events was also expressed by Leotta et al. [LMM15] back in 2015, where the need to bridge the gap between sensor events and human activities was stated as one of the main research challenges for the application of process mining to smart spaces.

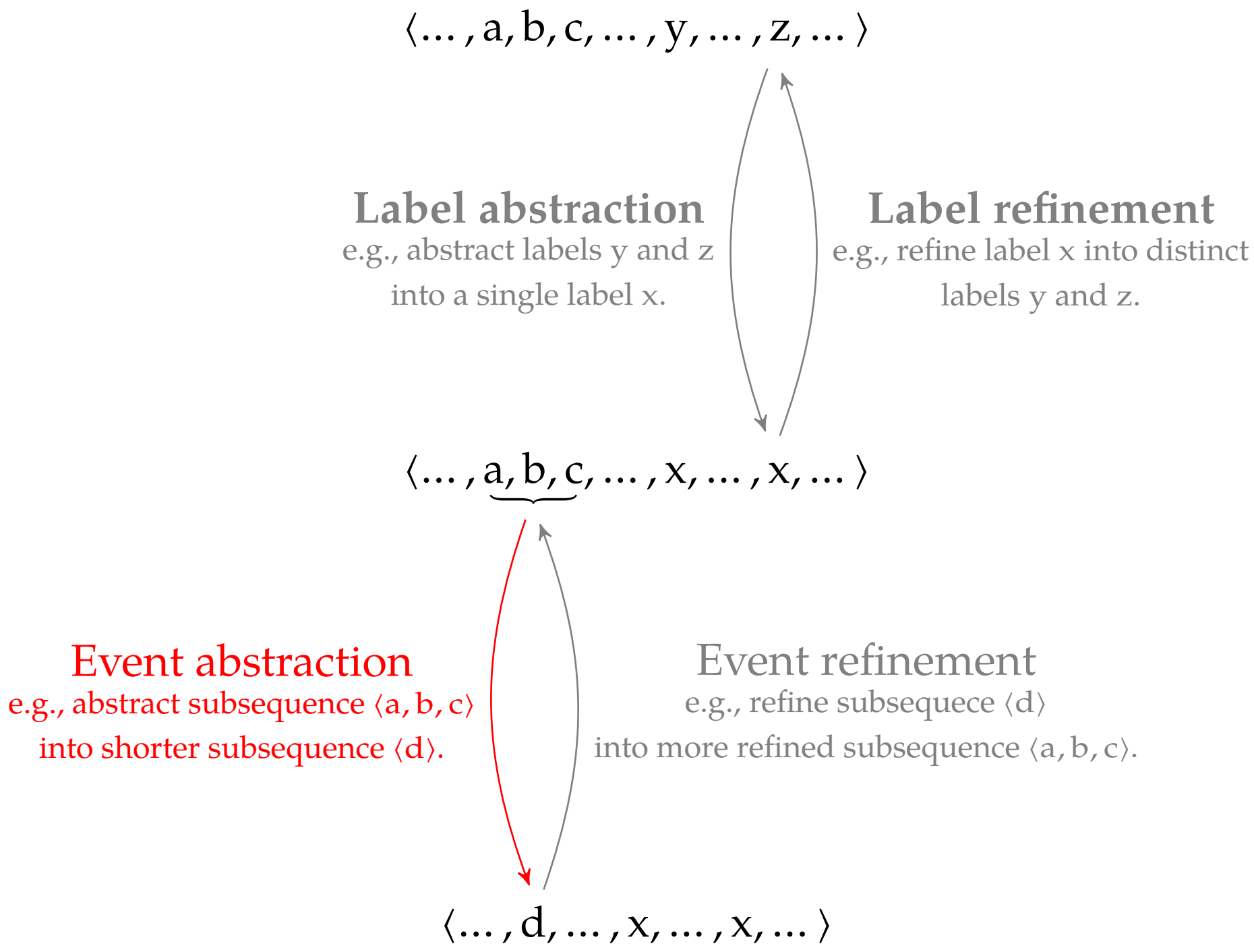

**Figure 7.1:** The positioning of Chapter 7 in the taxonomy of event-log preprocessing methods of Figure 1.4. Colored preprocessing types are discussed in this chapter.

When abstracting sensor events to events on a human behavior level, it is a nontrivial task to determine the desired set of labels on the human activity level. This desired set of labels on the human activity level also depends on the types of insights that one aims to obtain from analyzing the resulting event log on the human activity level. Two possible candidates for desired sets of activity labels are the *activities of daily living (ADL)* [Kat+63; Kat83] and the *instrumental activities of daily living* (IADL) [LB69]. Table 7.1 provides an overview of the types of activities that are included in ADL and IADL. Interestingly, both the ADL and the IADL activities originate from the research field of geriatrics and were initially formulated as part of a questionnaire to asses the degree to which elderly persons or people suffering from chronic illnesses are able to live independently. More recently, ADL and IADL activities have gained interest from the activity recognition research field, where the aim is to automatically detect the occurrence of ADL or IADL activities or to detect patterns and relations between ADL and IADL activities (see e.g., [BI04; Kas+08; KWM11; TIL04] In this chapter, we mostly focus on ADL activities.

**Table 7.1:** An overview of the ADL [Kat+63; Kat83] and the IADL [LB69] activities.

| ADL | IADL |
|---|---|
| Eating | Using telephone |
| Waking up | Shopping |
| Personal hygiene | Food preparation |
| Toilet use | Housekeeping |
| Taking a bath or a shower | Laundry |
| Walking on a corridor | Traveling |
| Ascend and descend stairs | Medication use |
| Dressing | Handling finances |
| Bowel continence | |
| Urine continence | |

In this chapter, we present a supervised approach to event abstraction which relies on the availability of training examples from which the mapping from sensor-level to activity-level events can be learned. Such training data can be collected in smart home experiments by collecting diary-like annotations in which the inhabitants of the smart home manually record their ADL-level events and their timestamps. This enables discovery of process models that describe the ADL-level activities directly, leading to more comprehensible and more precise descriptions of human behavior. Often it is infeasible or simply too expensive to obtain such diaries for periods of time longer than a couple of weeks. However, to mine a process model of human behavior more than a couple of weeks of data is needed. Therefore, there is a need to infer ADL-level interpretations from sensor-level events automatically for there data where there are no diary annotations available.

Methods from the area of supervised learning can be used to learn how to transform data at the sensor level to events at the ADL level through examples, without providing hand-made descriptions how human activities relate to sensor events. Comparable methods have been employed in the area of *activity recognition*, in which continuous-valued time series that are generated by sensors are mapped to time series of ADL-level events. Sensor-level events, such as *opening the freezer*, trigger change points in these time series, and corresponding ADL events (e.g. *preparing dinner*) are generally collected by manual logging of activity diaries. However, unlike the techniques from the activity recognition field, we operate on discrete events at the sensor-level instead of continuous time series. By operating on discrete events instead of on continuous-valued time series, the approach becomes a general event abstraction approach for process mining that is also applicable outside of the smart home domain.

**Definition 7.1 (Event Abstraction).** For an unlabeled event log $U \subseteq \mathcal{C}$, a labeling function $l_1 : \mathcal{C} \rightarrow \Sigma_1^*$ is an *event abstraction* of another labeling function $l_2 : \mathcal{C} \rightarrow \Sigma_2^*$ if $(\forall_{\sigma_1,\sigma_2 \in \mathcal{C}} l_2(\sigma_1) = l_2(\sigma_2) \implies l_1(\sigma_1) = l_1(\sigma_2)) \wedge (\forall_{\sigma \in \mathcal{C}} |l_2(\sigma)| \geq |l_1(\sigma)|)$. ◇

If $l_1$ is an event abstraction of $l_2$ then we call $l_2$ an *event refinement* of $l_1$. Note that the relations *event abstraction*, *event refinement*, *label abstraction*, and *label refinement* between two labeling function $l_1 : \mathcal{C} \rightarrow \Sigma_1^*$ and $l_2 : \mathcal{C} \rightarrow \Sigma_2^*$ are mutually exclusive, i.e., if $l_1$ is an event abstraction of $l_2$, then $l_1$ cannot at the same time be an event refinement, label refinement, or label abstraction of $l_2$. To see that $l_1$ cannot at the same time be a label abstraction and an event abstraction of $l_2$, consider that for it to be a label abstraction, $\forall_{\sigma \in \mathcal{C}} |l_1(\sigma)| = |l_2(\sigma)|$ while to be an event abstraction $\forall_{\sigma \in \mathcal{C}} |l_1(\sigma)| \leq |l_2(\sigma)|$. The event abstractions that we aim to discover in this chapter are event abstractions with respect to $l_{sensor}$, i.e., the labeling function that labels each event with the sensor that generated it.

In this chapter, we propose a framework for abstraction of events based on supervised learning. Our framework facilitates data preprocessing. We show that the event logs produced by our framework allow for the mining of more precise process models from smart home event logs. Additionally, the process models obtained represent ADL activities directly, thereby enabling direct analysis of human behavior itself, instead of indirect analysis through sensor-level models. In Section 7.4 we describe related work, focusing both on the process mining and the activity recognition areas. In Section 7.1 we explain conceptually why abstraction from sensor-level to ADL events can help to the process discovery step to find more precise process models. In Section 7.2 we describe a framework for retrieving features for abstraction of an event log that focuses on the semantically well-defined data attributes $\mathcal{A}_1, \ldots, \mathcal{A}_n$ for unlabeled event logs that are specified in the IEEE XES standard for event logs [IX16]. In Section 7.3 we apply our framework to three real-life smart home event logs and show that the discovered models are more precise compared to the models discovered on the unpreprocessed data at the sensor level. Additionally, we show in this section that the abstraction approach is more generally applicable by demonstrating it to an artificial dataset from a digital photocopier system where the discovered process model is incomprehensibly large when the event log is not abstracted while the discovered process model becomes within the limits of human cognitive capabilities when the event log is first abstracted. Section 7.5 concludes the chapter.

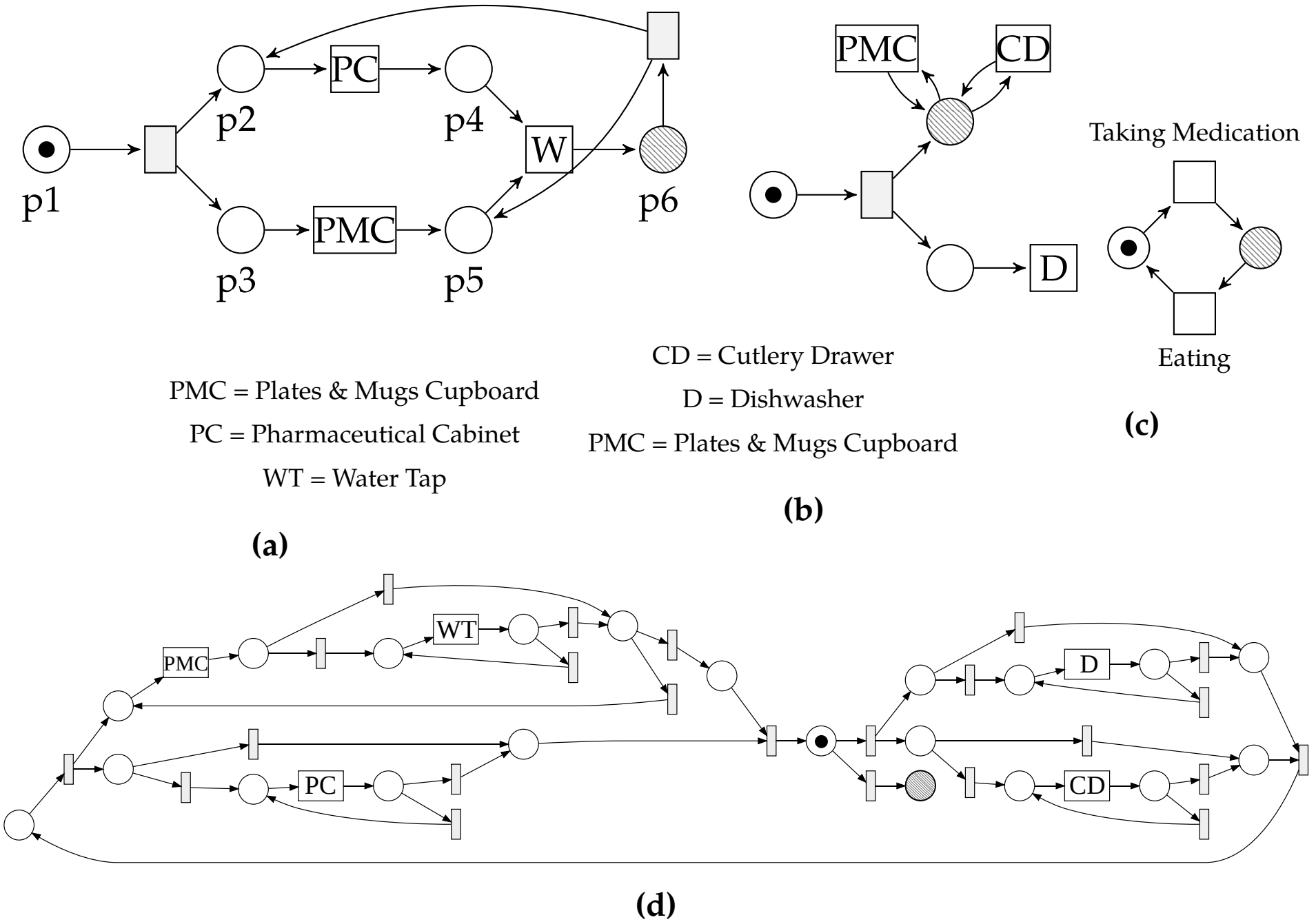

**Figure 7.2:** A human activity level process model (c) where the two transitions themselves are defined as process models (shown in a and b), and the Inductive Miner (IM) result on the sensor-level traces generated from this model (d).

# 7.1 The Discrepancy between Low-level and High-level Structure

Figure 7.2 demonstrates through a simplistic example that a process can seem unstructured at the sensor level of events, while being structured at a human behavior (ADL) level. The process model in Figure 7.2c depicts the actual process at the ADL level. The *taking medication* ADL activity is itself represented by a process, which is shown in Figures 7.2a. *Eating* is also defined as a process, which is shown in Figure 7.2b. The Inductive Miner (IM) [LFA13a] applied on traces of events on the sensor-level, simulated from the process of Figure 7.2c, results in the process model shown in Figure 7.2d. This process model allows for almost all possible traces over alphabet $\{WT, PMC, PC, D, CD\}$, as there is only one behavioral restriction specified by the process model: if a *W* occurs, then it has to be preceded by a *PMC* event. The silent transitions allow skipping the execution of the other transitions in the Petri net. The model from Figure 7.2d has a very low precision with respect to the simulated log, as it overgeneralizes, i.e., it allows for too much behavior that is not seen in the event log. Behaviorally this model is almost equivalent to the *flower model*, i.e., the model that does not specify any behavioral restrictions over the activities. The alternating pattern consisting of *taking medication* and *eating* in the ADL level process of Figure 7.2c cannot be found in Figure 7.2d. The high variance in the *start events* and *end events* of the sensor-level subprocesses of *taking medication* and *eating* as well as by the overlap in types of activities between the subprocesses are the cause of this effect. Both subprocesses contain *PMC*, and the miner cannot see that there are actually two different contexts for the *PMC* activity to split the label in the model. Abstracting the sensor-level events to their respective human activity level events before applying process discovery to the resulting human activity log unveils the alternating structure between *eating* and *taking medication* as shown in Figure 7.2c. The resulting model after abstraction is both *fitting*, as it allows for all the behavior that is should allow, and *precise*, as it does not allow for any more behavior than it should allow for.

## 7.2 Abstracting Events Through Sequence Labeling

This section presents the framework for supervised abstraction of events based on Conditional Random Fields (CRFs). We start this chapter by introducing basic concepts related to CRFs. We continue by introducing feature functions that can be extracted from unlabeled event logs and their data payload over their attribute space $\mathcal{A}_1, \dots, \mathcal{A}_n$. Every unlabeled event log has its own data payload with its attribute space. To extract feature functions from event attributes, it is of importance to be able to use domain knowledge about the attribute concerning the semantics of the attribute. In the general case, such semantic information of what an event attribute actually means might not be available. However, the IEEE XES standard [IX16] specifies a set of event attributes that are commonly found in the attribute space of unlabeled event logs, and it provides precise definitions of their semantics. In this section, we will show how feature functions can be defined for the event attributes that are defined in the IEEE XES standard.

### 7.2.1 Conditional Random Fields

We consider the detection of human activity (ADL) events from sensor-level events as a sequence labeling problem where each event on the sensor-level is classified into one of the human activity level events. Linear-chain Conditional Random Fields (CRFs) [LMP01] are a special type of Probabilistic Graphical Model (PGM) that has proven to be suitable for a variety of sequence labeling tasks in natural language processing and biological sequences. Conceptually, CRFs can be regarded as a sequential version of multiclass logistic regression, i.e., the predictions in the prediction sequence are dependent on each other. CRFs model the probability distribution over the possible labellings of an input sequence, conditional to that input sequence. Formally, linear-chain CRFs are formulated as follows:

$$P(y|x) = \frac{e^{\sum_{t=1}^{n} \sum_{k \in K} w(k) f_k(x(t), y(t-1), y(t))}}{Z(x)} \tag{7.1}$$

in which $Z(x)$ is a normalization term which makes sure that the resulting function is a valid probability distribution (i.e., it sums to one). $x$ with $|x| = n$ is an input trace from an unlabeled event log that consisting of the low-level events (i.e., on the sensor level), $y$ with $|y| = n$ is the associated sequence of higher, ADL-level, labels for the $n$ events in the $x$. Section 7.2.2 will address how the final resulting higher level trace will ultimately end up with $< n$ events. $K$ is the set of features that are defined for events from the unlabeled event log, where for each $k \in K$, $f_k$ denotes the k-th feature function and $w(k)$ represents the weight of the feature function $k$. Each feature function is a function over the current event (i.e., the t-th event of $x$), the previous predicted high-level label $y(t-1)$ and the current high-level label $y(t)$ that we are currently predicting. In the training phase, based on a set of example traces for which both low-level events $x$ and the high-level events $y$ are known, the values $w(k)$ for all $k \in K$ are optimized. This optimization procedure aims to select the weights $w(k)$ such that the cross-entropy error of the predicted labels for the input sequences on the training set is minimized. At prediction time, the predicted sequence of high-level events $\hat{y}$ is determined by calculating:

$$\hat{y} = \arg\max_{y} P(y|x) \tag{7.2}$$

with $y \in \{\sigma \in \Sigma^* || \sigma| = |x|\}$. CRFs are conceptually close to Hidden Markov Models (HMMs) [RJ86]. However, where HMMs assume the feature functions to be uncorrelated, CRFs do not make this assumption.

Now we specify how each of the XES extensions can be used to extract helpful feature functions for event abstractions. In the training phase, we search for values of weight vector $w$ that minimize the cross-entropy between the ground truth label and the predicted label on the training data. A commonly used algorithm to optimize the weights of a CRF is the Broyden–Fletcher–Goldfarb–Shanno (BFGS) algorithm, or its later version with bounded-memory guarantees called Limited-memory BFGS (L-BFGS) [Noc80].

The set $K$ of feature functions for the attributes defined in IEEE XES logs that we will define in Section 7.2.2 could generate a feature space where $|K|$ is unknown and dependent on the specific event log, depending on which of the attributes from the IEEE XES standard are present in the data payload of the log. Potentially, in case $|K|$ is high, and the number of training samples is low, this could result in problems related to overfitting. To address the issue of overfitting, we apply a common technique from machine learning called $\ell_1$ regularization, which slightly alters the goal function of the weight vector $w$ in the training phase. $\ell_1$ regularization is widely applied in the machine learning and the statistics research fields, with

7

another popular application being LASSO regression [Tib96], which is simple linear regression with additional $\ell_1$ regularization. The resulting goal function of the CRF when applying $\ell_1$ regularization consists of the cross-entropy between the ground truth label and the predicted label and it adds to that a terms $\lambda \sum_{k \in K} w(k)$, and minimizes the resulting goal function. The effect of adding the terms $\sum_{k \in K} w(k)$ to the goal function is that preference is given to a weight vector w that provides a simple explanation for the sequences in the training data, i.e., an explanation where many of the features are unused, or have weight 0. The $\lambda$ term is a parameter that specifies the strength of the regularization effect. Unfortunately, it has been shown that L-BFGS is not able to successfully optimize the weight vector w when there is an $\ell_1$ regularization term in the goal function of the optimization procedure. This is caused by the fact that L-BFGS procedure is based on differentiation and the added $\sum_{k \in K} w(k)$ term the error non-differentiable for certain parts of the parameter space. Therefore, we apply the OWL-QN [AG07] optimization procedure, which has been developed specifically for $\ell_1$-regularized CRFs.

### 7.2.2 The IEEE XES Standard and Extracting Features from it

The IEEE standard XES, which is an abbreviation for *eXtensible Event Stream*, is the IEEE standard for process mining event logs. Figure 7.3 provides an overview of the IEEE XES file structure. The IEEE XES standard defines an *event log* as a set of *traces*. A trace is defined as sequences of *event*s. Logs, traces and events can contain *attribute*s, containing a *key* and *value*. *Global* attributes are a set of attribute keys for traces or events, which indicate respectively that all traces respectively all events in the log contain an attribute with that key. *Classifier*s define viewing perspectives on logs, and can be used to transform traces from complex sequences of events to simple sequences of labels, taking the labels from one or more global event attributes that are defined by the classifier. A special collection of global attributes for logs, traces, or events are provided by *extension*s, which have clear, globally defined, semantics for some attribute keys. The IEEE XES standard [IX16] lists the following set of extensions, and their semantic interpretation:

**Concept** stores a generally understood name for any hierarchy element. For logs, the name attribute may store the name of the process having been executed. For traces, the name attribute usually stores the case ID. For events, the name attribute represents the name of the event, e.g. the name of the executed activity represented by the event (*key:* **concept:name**).

**Lifecycle** specifies the lifecycle phase (*key:* **lifecycle:transition**) that the event represents in a transactional model of its producing activity. The lifecycle extension also specifies a standard transactional model for activities.

**Organizational** the name, or identifier, of the resource having triggered the event (*key:* **org:resource**), the role of the resource having triggered the event, within the organizational structure (*key:* **org:role**), and the group within the organizational structure, of which the resource having triggered the event is a member. (*key:* **org:group**).

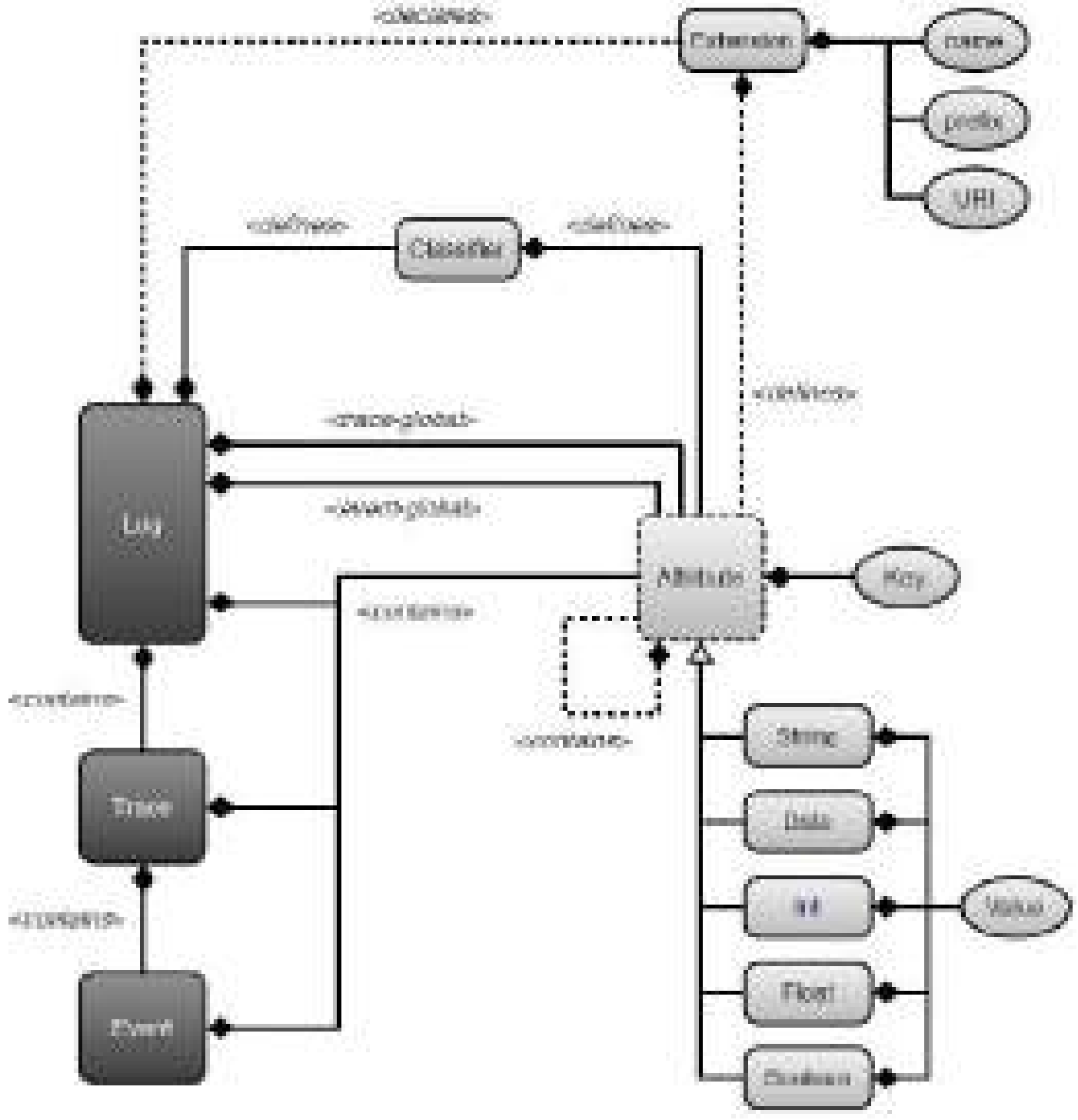

**Figure 7.3:** The metamodel of IEEE XES event logs (from [IX16]).

**Time** The date and time, specifying the point in time when the event has taken place (*key:* **time:timestamp**).

In addition to the existing *concept:name* attribute key, we propose attribute key *label*, representing the activity at the human activity (ADL). We use the existing key *concept:name* sto represent sensor-level activity. Each event then contains a low-level *concept:name* attribute and a high-level *label* attribute, linking a sensor event (e.g., *plates & mugs cupboard*) to an ADL event (e.g., *taking medication*), and potentially linking another sensor event of the same type to a different ADL event (e.g. *eating*). Some traces can consist of events with empty *label* attributes, in case ADL level annotations are not available for the trace. We aim to provide accurate estimations of activities on the ADL level for traces with missing *label* attributes, thereby replacing the missing *label* value with an estimated value, by using a model learned from traces where *label* attributes are available. This allows traces with missing *label* attributes to be used for process mining applications.

**Table 7.2:** A case consisting of events on the sensor level and their respective predicted label (i.e., human activity annotations)

| trace | time:timestamp | concept:name | label |
|---|---|---|---|
| 23 | 04/12/2016 08:46:24 | Pharmaceutical cabinet | Taking medication |
| 23 | 04/12/2016 08:47:12 | Plates & mugs cupboard | Taking medication |
| 23 | 04/12/2016 08:47:46 | Water tap | Taking medication |
| 23 | 04/12/2016 08:48:60 | Plates & mugs cupboard | Eating |
| 23 | 04/12/2016 08:48:90 | Dishwasher | Eating |
| 23 | 04/12/2016 17:11:59 | Plates & mugs cupboard | Taking medication |
| 23 | 04/12/2016 17:11:70 | Pharmaceutical cabinet | Taking medication |
| 23 | 04/12/2016 17:12:19 | Water tap | Taking medication |

**Table 7.3:** The resulting ADL-level log, obtained after merging in Table 7.2 consecutive events in a trace with identical values for the label attribute. Lifecycle:transition s='start' and c='complete'.

| trace | time:timestamp | concept:name | lifecycle:transition |
|---|---|---|---|
| 23 | 04/12/2016 08:46:24 | Taking medication | s |
| 23 | 04/12/2016 08:47:46 | Taking medication | c |
| 23 | 04/12/2016 08:48:60 | Eating | s |
| 23 | 04/12/2016 08:48:90 | Eating | c |
| 23 | 04/12/2016 17:11:59 | Taking medication | s |
| 23 | 04/12/2016 17:12:19 | Taking medication | c |

A schema of the abstraction approach is provided in Figure 7.4. The method uses as input a collection of traces with *label* attribute (i.e., the ADL activity for the sensor events are known) and a collection of traces without *label* attribute (i.e., only the sensor activity, but no ADL activity, is known). A CRF is trained on the traces with *label* attributes, resulting in a function from sensor-level events to human activity level events. The trained CRF model can be used to estimate the *label* attributes for traces where these are missing. Generally, multiple consecutive sensor-level events will have identical *label* attribute values. We assume that multiple human activity level events cannot occur concurrently. This assumption allows merging of a sequence of events with identical *label* values into a two ADL events, where the first event has value *start* and the second has value *complete* for attribute *lifecycle:transition*. Tables 7.2 and 7.3 illustrate this collapsing procedure with an example.

The CRF-based abstraction approach is implemented as part of the process mining framework ProM [Don+05] and available as package *AbstractEventsSupervised*.

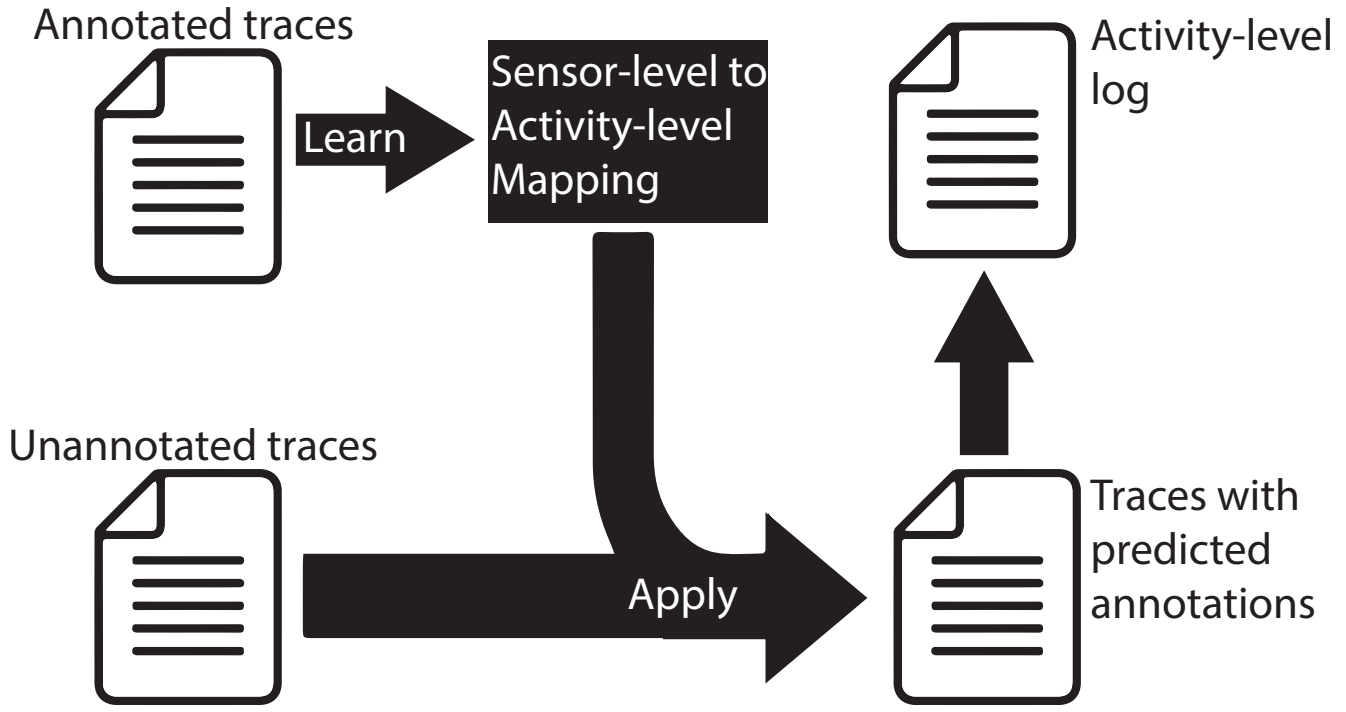

**Figure 7.4:** An overview of the Supervised Event Abstraction technique.

### 7.2.3 Extracting a Feature Vector from an IEEE XES event log

We now discuss per XES extension how feature functions can be obtained.

**Concept** Based on the k-gram $\langle a_1, a_2, \dots, a_k \rangle$ consisting of the sensor-level activities (*concept:name*) of the k last-seen events, we estimate a categorical probability distribution over the classes of human activity level activities from the training log, such that the probability of class l is equal to the number of times that the n-gram was observed while the k-th event was annotated with class l, divided by the total number of times that the n-gram was observed. A feature function based on the concept extension has two parameters, k and l, and is valued with the estimated categorical probability density of the current sensor-level event having human activity level label l given the n-gram with the last k sensor-level event labels. It can be useful to combine multiple features that are based on the concept extension, where the features have different values for k and l.

**Organizational** Analogous to the concept extension, a categorical probability distribution over the ADL activities can be estimated on the training data for *resource*, *role*, or *group* attribute k-grams. Similarly, a three parameter organizational extension based feature function, with k-gram size parameter k, parameter $o \in \{\text{group}, \text{role}, \text{resource}\}$, and label parameter l, is valued with the estimated probability density according of the categorical probability distribution for ADL activity l given the last k values of o.

**Time** Certain types of ADL activities could have particularly high probability density in certain parts of the day (e.g. *sleeping* at night), week (e.g. *working* during business days), month, etc. A naive attempt to model the probability density of ADL activities over the time-of-the-day, time-of-the-week, or time-of-the-month could be based on a Gaussian Mixture Model (GMM), where each component of the mixture is defined by a normal distribution. The circular, non-Euclidean, nature of the data space of time-of-the-day, time-of-the-week, or time-of-the-month, how-ever, introduces problems for the GMM, as, using time-of-the-day as an example, 00:00 is actually very close to 23:59. We had already illustrated this problem back in Figure 5.2 back in Chapter 5. The Gaussian component with a mean around 10 o'clock has a standard deviation that is much higher than what one would expect when looking at the histogram, as the GMM tries to explain the data points just after midnight with this component. These data points just after midnight would, however, have been much better explained with the Gaussian component with the mean around 20 o'clock, which is much closer in time. Alternatively, we use a mixture model with components of the von Mises distribution, which is a close approximation of a normal distribution wrapped around the circle. To determine the correct number of components of such a von Mises Mixture Model (VMMM) we use Bayesian Information Criterion (BIC) [Sch78], which choses the number of components that explains the data with the highest likelihood while at the same time adding a penalty for the number of model parameters. A VMMM is learned from training data and models the probabilities of each type of human activity based on the amount of time that has passed since the day start, week start, month start. A feature function for the time extension with parameters $t \in \{day, week, month, \dots\}$ and label $l$, is valued with the probability density of label $l$ according to the learned VMMM model given the view on the timestamp of the event according to $t$. An alternative approach to estimate the probability density on data that lies on a manifold, such as a circle, is described by Cohen and Welling [CW15].

**Lifecycle & Time** The IEEE XES standard [IX16] describes lifecycle phases of activities. Lifecycle values that are commonly found in real life logs are *start* and *complete* which respectively represent when this activity started and ended However, a larger set of lifecycle values is defined in the IEEE XES standard, including *schedule*, *suspend*, and *resume*. The time differences between different stages of an activity lifecycle can be calculated for event logs that contain the lifecycle extension as well as and the time extension. For example, when observing the *complete* of an activity, the time between this *complete* and the corresponding *start* of this activity can contain useful information for predicting the correct human activity label. Identifying the corresponding *start* event is non-trivial when multiple instances of the same activity run concurrently. We make the assumption that only one sensor-level event of the same sensor type can be on-going at any point in time.

The IEEE XES standard defines an ordering over the lifecycle values. For each type of human activity, we fit a GMM to the set of time differences between each two consecutive lifecycle steps. A feature based on both the combination of the lifecycle and the time extension with activity label parameter $l$ and lifecycle $c$ is valued the probability density of activity $l$ as estimated by the GMM given the time between the current event and lifecycle value $c$. We decide on the number of components of the GMM using BIC [Sch78]. Note that while these features are time-based, regular GMMs can be used instead of VMMMs since time duration is a Euclidean, non-circular, space.

### 7.2.4 Evaluating the Estimated Human Activity Events for Discovering ADL Models

A well-known metric for the distance of two sequences is the Levenshtein distance. However, Levenshtein distance is not suitable to compare sequences of human actions, as human behavior sometimes includes branches in which it does not matter in which order two activities are performed. For example, most people *shower* and *have breakfast* after waking up, but people do not necessarily always perform the two in the same order. Indeed, when $\langle a, b \rangle$ is the sequence of predicted human activities, and $\langle b, a \rangle$ is the actual sequence of human activities, we consider this to be only a minor error because it is often not relevant in which order two parallel activities are executed. Levenshtein distance would assign a cost of 2 to this abstraction, as transforming the predicted sequence into the ground truth sequence would require one deletion and one insertion operation. For example, most people *shower* and *have breakfast* after waking up, but people do not necessarily always perform the two in the same order. An evaluation measure that better reflects the prediction quality of event abstraction is the Damerau-Levenstein distance [Dam64], which adds a swapping operation to the set of operations used by Levenshtein distance. Damerau-Levenshtein distance would assign a cost of 1 to transform $\langle a, b \rangle$ into $\langle b, a \rangle$. To obtain comparable numbers for different numbers of predicted events we normalize the Damerau-Levenshtein distance by the maximum of the length of the ground truth trace and the length of the predicted trace and subtract the normalized Damerau-Levenshtein distance from 1 to obtain Damerau-Levenshtein Similarity (DLS).

## 7.3 Case Studies

In this section we evaluate the supervised event abstraction framework on three case studies on real-life smart home datasets and on one artificial dataset from an digital photocopier system.

### 7.3.1 Experimental setup

We include three real-life smart home event logs in the evaluation: the Van Kasteren event log [Kas+08], and two event logs from a smart home experiment conducted by MIT [TIL04]. All three event logs used in for the evaluation consist of time series of multiple dimensions, in which every time series contains the binary value that characterizes the state of one sensor over time. These datasets include motion sensors, open/close sensors, and power sensors (discretized to binary states). We have preprocessed sensor data into events, such that each change point in the value of a sensor results in an event. Events that have occurred in the same day are grouped together to form a case. Annotations at the ADL level are included in all three logs. The following XES extensions can be used for these event logs:

**Concept** Specifies which sensor in the smart home environment caused this event.

**Time** Specifies the timestamp at which the changepoint in the sensor occurred.

**Lifecycle** Has the value *Start* when this event represents change in sensor value from 0 to 1 and *Complete* otherwise.

For evaluation of the abstraction quality, we split the data into separate parts for training and testing. The predicted ADL labels are evaluated by comparing them the actual ADL labels in a Leave One Trace Out Cross Validation (LOTOCV) setup, i.e., we iteratively leave out one trace from the dataset to evaluate on, while training on the other traces of the dataset. We measure the accuracy of the ADL level traces compared to the ground truth ADL traces in terms of Damerau-Levenshtein similarity [Dam64]. Additionally, we evaluate the quality of the process model that can be discovered from the ADL level traces. We use the Inductive Miner (IM) [LFA13b] to mine a Petri net from the estimated ADL level log and evaluate it on fitness, precision, and f-score.

## 7.3.2 Case Study 1: Van Kasteren Event Log

For the first case study, we use an event log from a smart home setting with in-house sensors [Kas+08]. This Van Kasteren log contains 1285 events divided over fourteen different sensors. The log contains 23 days of data. The average Damerau-Levenshtein similarity between 1) the ADL level traces that were predicted in the Leave One Trace Out Cross Validation (LOTOCV) experimental setup and 2) the actual ADL level traces is 0.7516, which shows that the estimated ADL level traces as produced by the approach are reasonably similar to the ground truth.

Figure 7.5 depicts the resulting Petri net obtained with the IM algorithm [LFA13a] from the sensor-level events. The process model starts with a choice between four activities: *hall-toilet door*, *hall-bedroom door*, *hall-bathroom door*, and *frontdoor*. After this choice the model branches into three parallel blocks, where the upper block consists of a large choice between eight different activities. The other two parallel blocks respectively contain a loop of the *cups cupboard* and the *fridge*. This model allows for many traces, and it has high similarity to the flower model. The data seems to be unstructured at the sensor level granularity events, or, at least the IM algorithm seems to be unable to find such structure.

Figure 7.6 depicts the resulting process model obtained by applying the IM algorithm on the aggregated set of predicted test sequences. The process model shows a clear daily routine, which starts with having *breakfast*, followed by *leaves the house*, presumably to go to work. When back from work, the person *prepares dinner* and *goes to bed*. Activities *use toilet* and *take shower* are put in parallel to this sequence of activities, indicating that they occur at different places in the sequences of activities.

Table 7.4 shows the effect of the abstraction on the fitness, precision, and F-score of the models discovered by the IM algorithm of the three case studies shown in this section and the following sections. It shows that the precision of the model discovered on the abstracted log is much higher than the precision of the model discovered on the sensor data, indicating that the abstraction helps to discover a model that is more behaviorally constrained and more specific. At the same time, the drop in fitness as a result of abstracting events is limited, indicating that the obtained models on the higher level still allow for most of the behavior seen in the log. The F-score values show that the trade-off between fitness and precision improves as a result of the abstraction of events for all three smart home event logs.

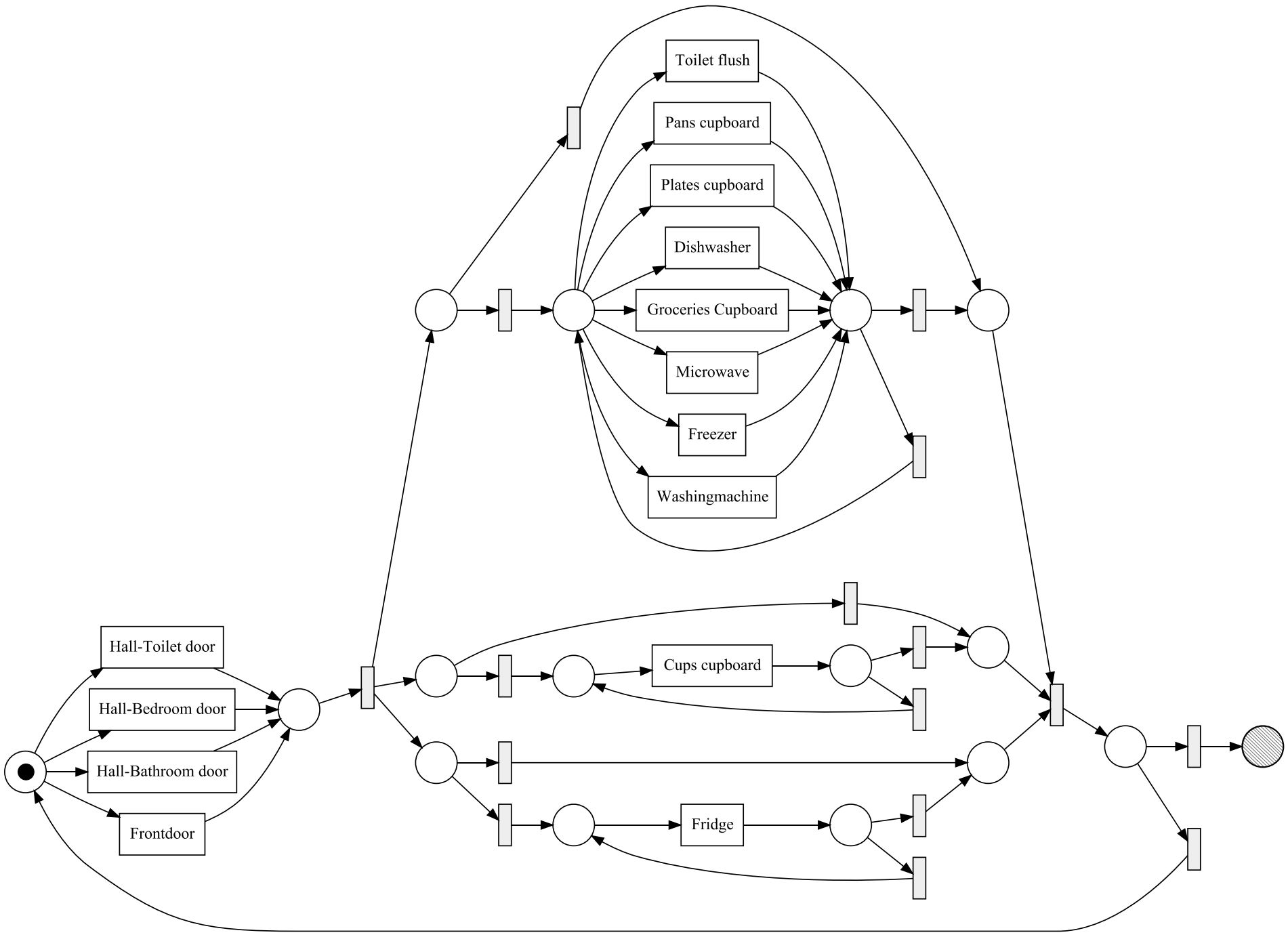

**Figure 7.5:** Inductive Miner output on the sensor-level events of the Van Kasteren log.

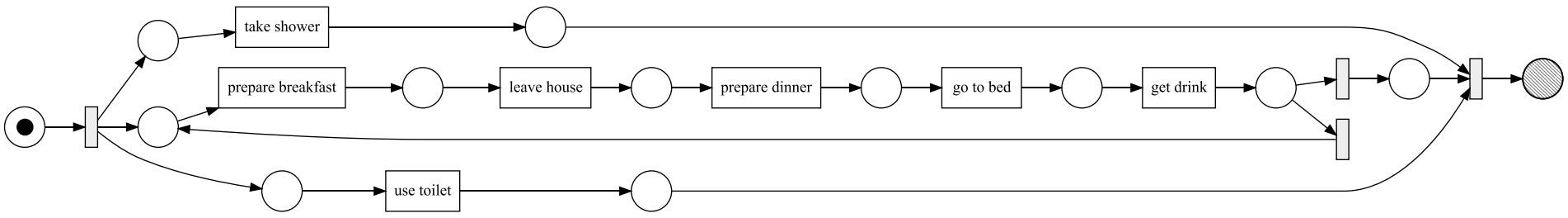

**Figure 7.6:** Inductive Miner output on the human activity level Van Kasteren log.

**Table 7.4:** Effect of abstraction on fitness, precision, and F-score of the process model discovered by the Inductive Miner.

| Event log | Abstraction | Fitness | Precision | F-score |
|---|---|---|---|---|
| Van Kasteren | No (Figure 7.5) | 0.9111 | 0.3308 | 0.4854 |
| Van Kasteren | Yes (Figure 7.6) | 0.7918 | 0.7804 | 0.7861 |
| MIT household A | No (Figure 7.7) | 0.9916 | 0.2289 | 0.3719 |
| MIT household A | Yes (Figure 7.8) | 0.9880 | 0.3711 | 0.5395 |
| MIT household B | No (Figure 7.9) | 1.0 | 0.2389 | 0.3857 |
| MIT household B | Yes (Figure 7.10) | 0.9305 | 0.4319 | 0.5900 |

### 7.3.3 Case Study 2: MIT Household A Event Log

For the second case study, we use the data of *household A* of a smart home experiment conducted by MIT [TIL04]. Household A contains data from 16 days of living, 2701 sensor-level events registered by 26 different sensors. The human level activities are provided in the form of a taxonomy of activities on three levels, called *heading*, *category* and *subcategory*. On the *heading* level the human activities are very general in nature, such as the activity *personal needs*. The eight different activities on the *heading* level branch into 19 different activities on the *category* level, where *personal needs* branches into e.g. *eating*, *sleeping*, and *personal hygiene*. The 19 *categories* are divided over 34 *subcategories*, which contain very specific human activities. At the *subcategory* level the *category meal cleanup* is for example divided into *washing dishes* and *putting away dishes*. At the *subcategory* level there are more types of human activities than there are sensors-level activities, which makes the abstraction task very hard. Therefore, we set the target label to the *category* level.

The model that is discovered with the IM algorithm on the sensor events in the MIT household A log is shown in Figure 7.7. The model obtained allows for too much behavior, as it contains two large choice blocks. We found a Damerau-Levenshtein similarity of 0.6348 in the LOTOCV experiment. Note that the abstraction accuracy on this log is lower than the abstraction accuracy on the Van Kasteren event log. However, the MIT household A log contains more different types of human activity, resulting in a more difficult prediction task with a higher number of possible target classes. Figure 7.10 shows the process model discovered with the IM algorithm from the ADL level traces that we predicted from the sensor-level events. Even though the model is too large to print in a readable way, from its shape it is clear that the abstracted model is much more behaviorally constrained than the sensor-level model. The precision and fitness values in Table 7.4 show that indeed the process model after abstraction has become behaviorally more specific while the portion of behavior of the data that fits the process model remains more or less the same.

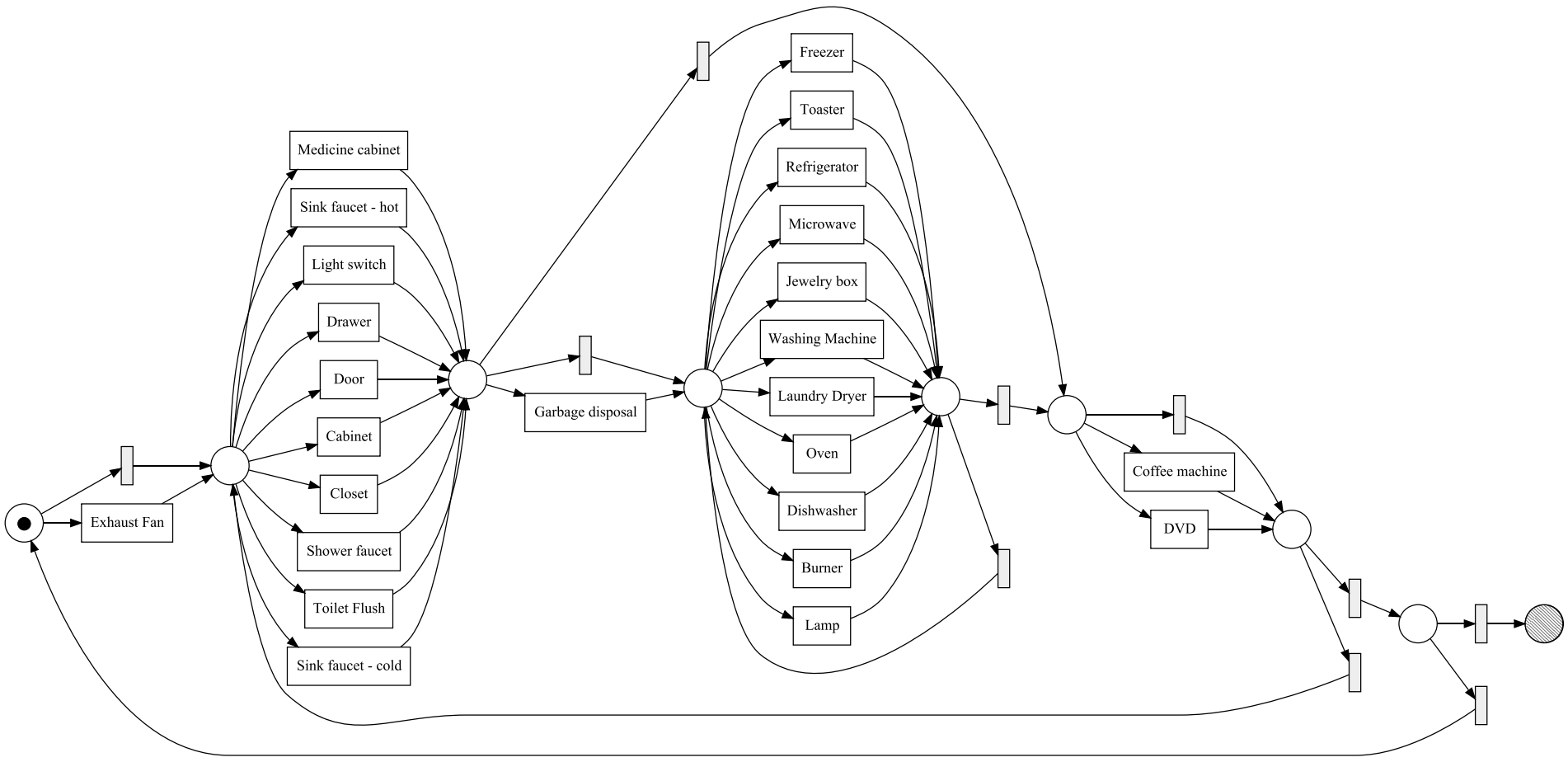

**Figure 7.7:** The resulting process model of the Inductive Miner applied on the sensor-level MIT household A event log

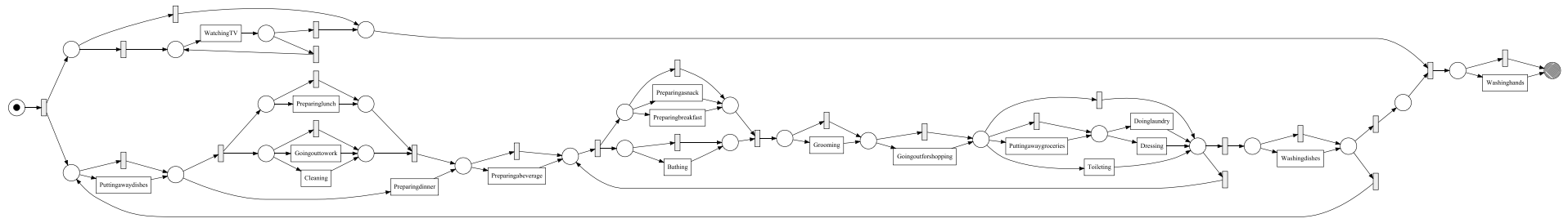

**Figure 7.8:** The resulting process model of the Inductive Miner applied on the discovered human activity level events on the MIT household A log.

## 7.3.4 Case Study 3: MIT Household B Event Log

For the third case study, we use the data of *household B* of the MIT smart home experiment [TIL04]. Household B contains data from 17 days of living, 1962 sensor-level events registered by 20 different sensors. Identically to MIT household A the human-level activities are provided as a three-level taxonomy. Again, we use the *subcategory* level of this taxonomy as target activity label.

The model discovered with the IM algorithm [LFA13a] from the sensor events is shown in Figure 7.9. The model obtained allows for too much behavior, as it contains two large choice blocks. We found a Damerau-Levenshtein similarity of 0.5865 in the LOTOCV experiment, which is lower than the similarity found on the MIT A dataset while the target classes of the abstraction are the same for the two datasets. This can be explained by the fact that there is less training data for this event log, as household B contains 1932 sensor-level events where household A contains 2701 sensor-level events. Figure 7.10 shows the process model discovered from abstracted log. Again this model is not readable due to its size, but its shape shows it to be behaviorally quite specific. The precision and fitness values in Table 7.4 also that process model after abstraction has indeed become behaviorally more specific while the portion of behavior of the data that fits the process model decreased only slightly.

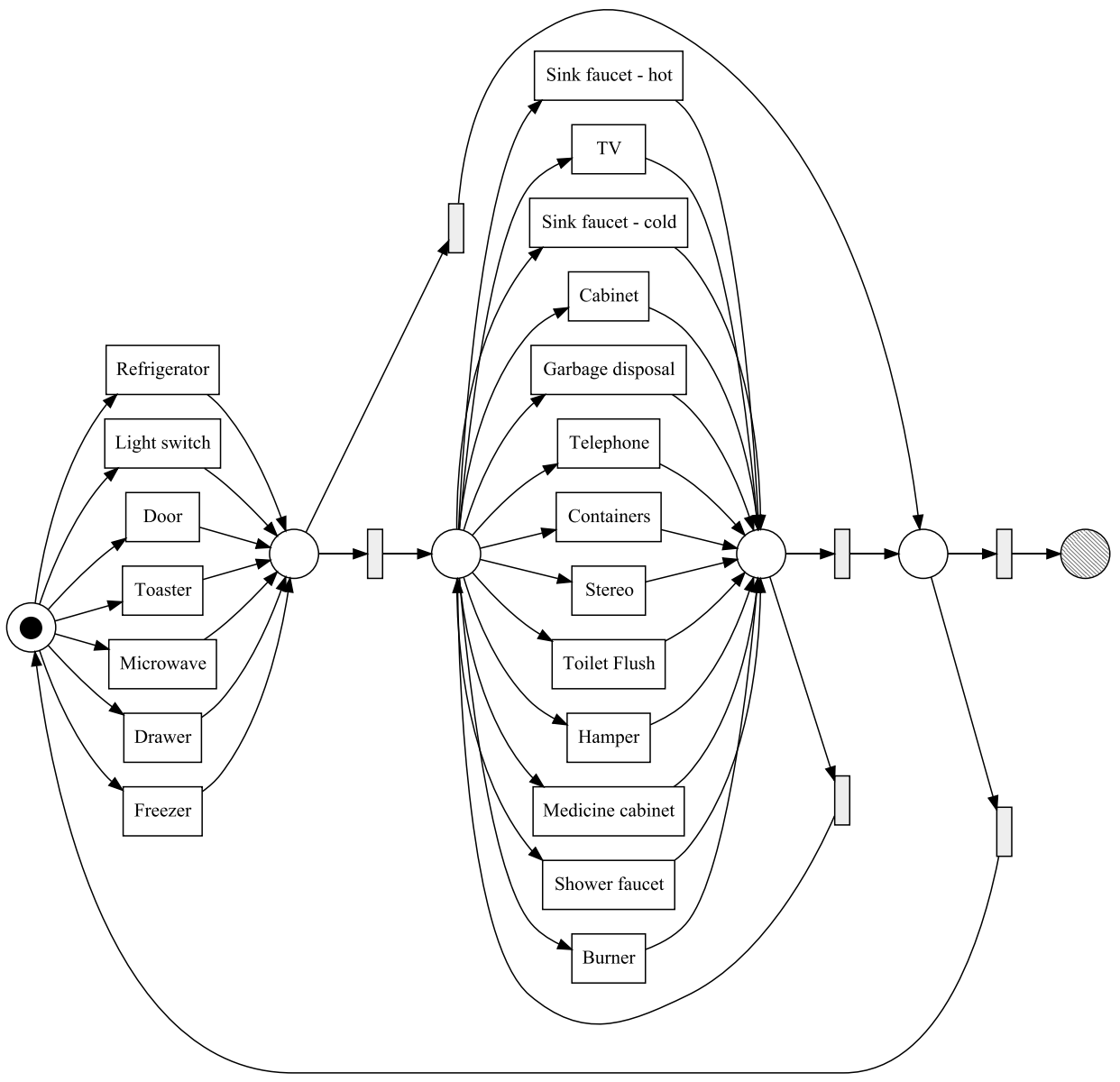

**Figure 7.9:** The resulting process model of the Inductive Miner applied to the sensor-level MIT household B event log.

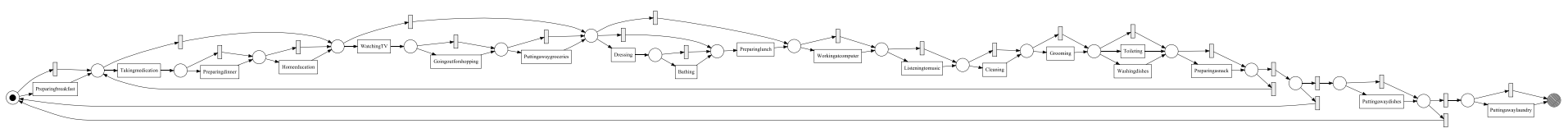

**Figure 7.10:** The resulting process model of the Inductive Miner applied to the discovered human activity level events on the MIT household B log

## 7.3.5 Case Study 4: Artificial Digital Photocopier

Bose et al. [Bos12; BVA12] created a synthetic event log based on a digital photocopier to evaluate his unsupervised methods of event abstraction. The purpose of this case study is to show that event abstraction using conditional random fields is not limited to application in smart home environments and can be applied in process mining more generally.

We annotated each low-level event with the correct high-level event manually using domain knowledge from the actual process model as described by Bose et al. [Bos12; BVA12]. This event log is generated by a hierarchical process, where high-level events *capture image*, *rasterize image*, *image processing* and *print image* are defined in terms of a process model. The *print image* subprocess amongst others contains the events *writing*, *developing* and *fusing*, which are themselves defined as a subprocess. In this case study we set the task to transform the log such that subprocesses *capture image*, *rasterize image* and *image processing*, *writing*, *fusing* and *developing*. Subprocesses *writing* and *developing* both contain the low-level event types *drum spin start* and *drum spin stop*. In this case study we focus in particular on the *drum spin start* and *drum spin stop* events, as they make the abstraction task non-trivial in the sense that no one-to-one mapping from low-level to high-level events exists.

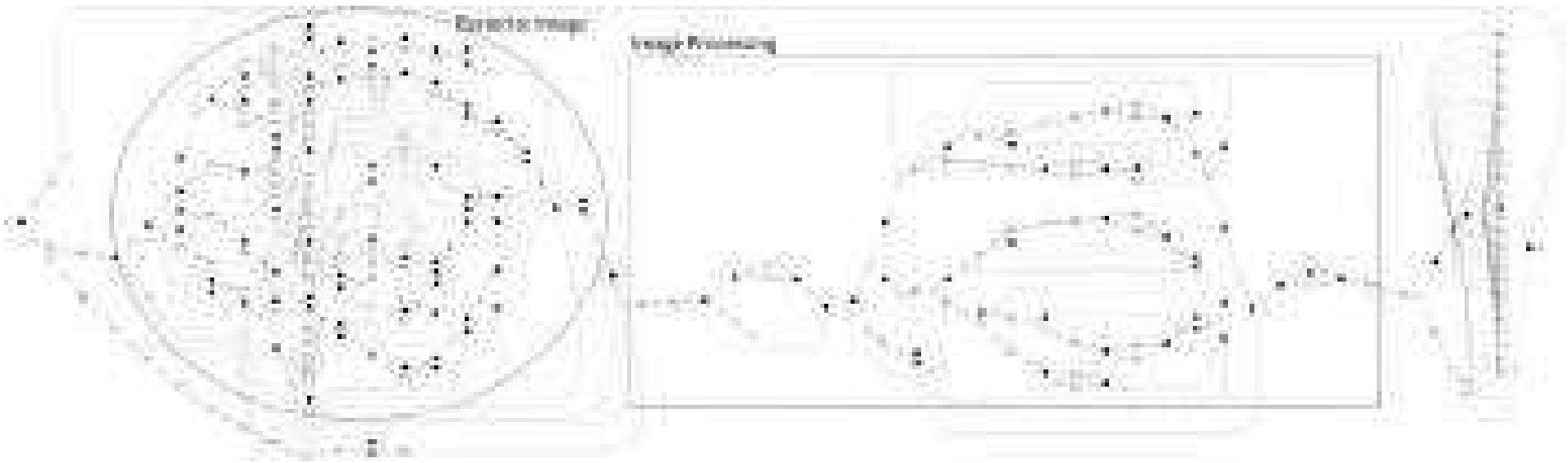

**Figure 7.11:** The resulting process model of the Inductive Miner applied to the low-level Artificial Digital Photo Copier event log.

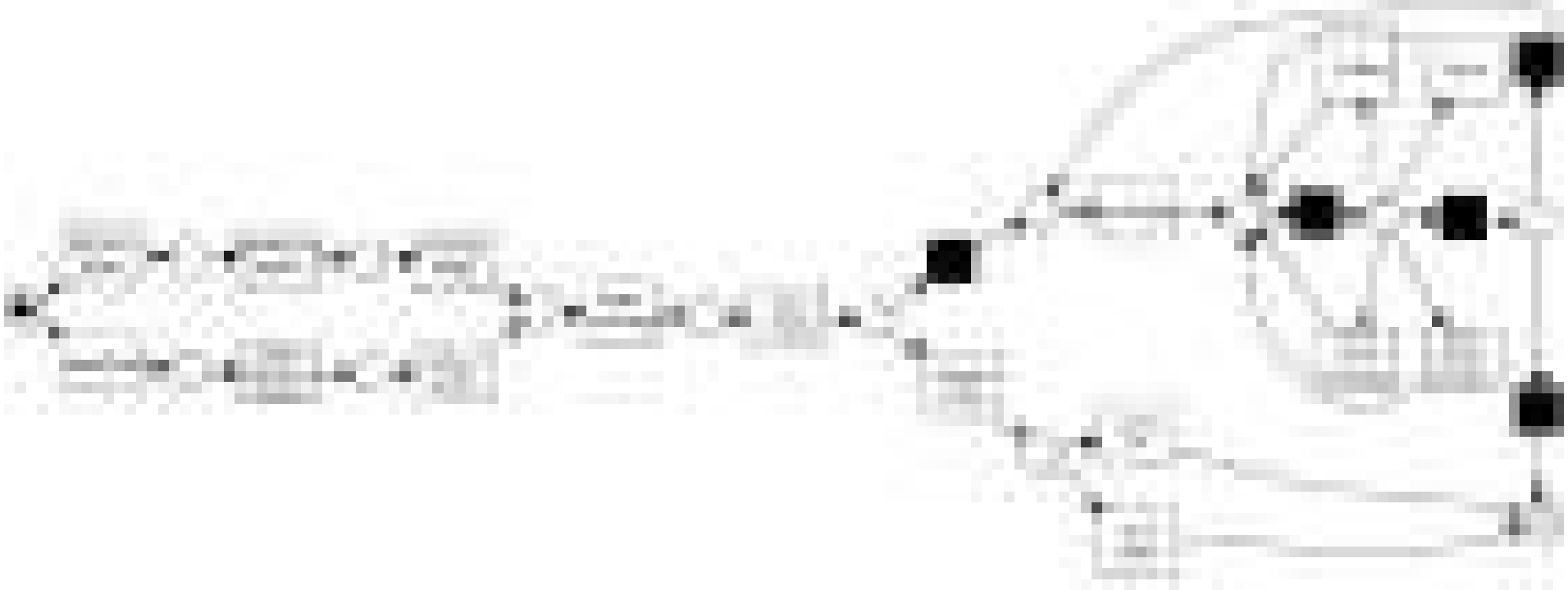

**Figure 7.12:** The resulting process model of the Inductive Miner applied to the discovered high-level events of the Artificial Digital Photo Copier event log.

The artificial digital photocopier dataset has the concept, time and lifecycle XES extensions. On this event log annotations are available for all traces. On this dataset we use a 10-Fold Cross-Validation setting on the traces to evaluate how well the supervised event abstraction method abstracts low-level events to high-level events on unannotated traces, as this dataset is larger than the smart home dataset that we used in the other case studies and as a result the LOTOCV setup is too computationally expensive.

**Table 7.5:** The confusion matrix for classification of low-level events *drum spin start* and *drum spin stop* into high-level events *writing* and *developing*.

| | Developing | Writing |
|---|---|---|
| Developing | 6653 | 0 |
| Writing | 0 | 917 |

The confusion matrix in Table 7.5 shows the aggregated results of the mapping of low-level events *drum spin start* and *drum spin stop* to high-level events *developing* and *writing*. The results show that the supervised event abstraction method is capable of detecting the many-to-many mappings between the low-level and high-level labels, as it maps these low-level events to the correct high-level event without making errors. The Damerau-Levenshtein similarity between the aggregated set of test fold high-level traces and the ground truth high-level traces is close to perfect: 0.9724.

Figure 7.11 shows the process model obtained with the Inductive Miner on the low-level events in the artificial digital photocopier dataset. The two sections in the process model that are surrounded by dashed lines are examples of high-level events within the low-level process model. Even though the low-level process contains structure, the size of the process model makes it hard to comprehend. Figure 7.12 shows the process model obtained with the same process discovery algorithm on the aggregated high-level test traces of the 10-fold cross validation setting. This model is in line with the official artificial digital photocopier model specification, with the *print Image* subprocess unfolded, as provided in [Bos12; BVA12]. In contrast to the event abstraction method described by Bose et al. [Bos12] which found the high-level events that match specification, supervised event abstraction is also able to find suitable event labels for the generated high-level events. This allows us to discover human-readable process models on the abstracted events without performing manual labeling, which can be a tedious task and requires domain knowledge.

## 7.4 Related Work

The majority of the work on abstractions from sensor-level to human activity level events can be found activity recognition research area, which addresses the challenge of recognizing different types of *activities of daily living* (ADL) [Kat83] from e.g. in-house sensors [Kas+08; TIL04], or on-body sensors [BI04; KWM11].

Techniques for activity recognition typically operate on discretized time windows from series of continuous-valued sensor readings and aim to classify each window onto the correct ADL type, such as *showering* or *eating*. Activity recognition methods can be classified into two categories: probabilistic approaches [BI04; Kas+08; KWM11; TIL04] and ontological reasoning approaches [CN09; RB11]. The advantage of probabilistic approaches over ontological reasoning approaches is their capability to deal with sensor data that is noisy, uncertain and/or incomplete [CN09].

The first application of supervised learning to the problem of ADL activities detection based on the data from in-house sensors [TIL04] was based on a naive Bayes classifier. Recently, a variety of activity recognition approaches were proposed that are based on probabilistic graphical models (PGMs) [Kas+08; KK07], including [Kas+08] (based on Conditional Random Fields (CRFs) [LMP01] and Hidden Markov Models (HMMs) [RJ86]), and [KK07] (based on Bayesian Networks [FGG97]). [KHC10] applied both HMMs and CRFs to the activity recognition problem, and observed that HMMs have difficulties to capture long-range dependencies between observations as well as transitive dependencies. The consequence of this is that HMMs have trouble to detect activities that are performed concurrently, or interleaved. Contrary to HMMs, CRFs have no difficulties to capture transitive and long-term dependencies.

Our work differentiates itself from existing activity recognition work in the form of the input data on which they operate and in the goal that it aims to achieve. Activity recognition methods take multi-dimensional time series data as input; this input consists thus of consecutive sensor values. Sliding window segmentation methods are then used to identify ADL activities to each window. Choosing an appropriate size for such a window, however, requires domain expertise and is particular to the dataset. This dependence on the domain specifics hinders the generality of window-based approaches. Our goal is a generic method that does not require this domain knowledge, and that works in general for any event log. Any chosen time window size t would be too small for some event logs (i.e., when the typical time between two consecutive high-level events in the event log is much larger than t, resulting in high computation time), while being too large for other event logs (i.e., when the typical time between two consecutive high-level events in the log is smaller than t, resulting in short high-level events not being detected). Therefore, time range based approaches are not suitable for supervised abstraction of events in the general case. Instead of operating on time ranges, we make a prediction of the ongoing ADL activity for each event on the low, sensor, level. Each change in the value of a binary-valued in-house sensors can be regarded as an occurrence of a low-level event. A second difference with existing activity recognition techniques is that our framework aims to find an abstraction of the data that enables discovery of more precise process models, where classical activity recognition methods do not have a link with the application of process mining.

Other related work can be found in the area of process mining, where several techniques address the problem of dealing with low-level events (e.g. sensor-level) by abstracting them to events at a higher level [BA09; DA10; GRA10; Man+16b; MT17]. Most existing event abstraction methods rely on clustering methods where each cluster of low-level events is interpreted as one single event on the higher level. Abstraction methods that are based on unsupervised learning have two fundamental limitations. First, no labels are generated for the created high-level events, and it is often not trivial to label them manually, as this requires domain knowledge that might not be available. Secondly, unsupervised learning does not provide any guidance concerning the degree to which the log should be abstracted. Normally, such unsupervised abstraction approaches are parameterized to control the size of the obtained clusters, thereby impacting the degree of abstraction. However, finding the right settings of such parameters, so that abstraction methods produce meaningful results, generally comes down to trial and error.

Two abstraction techniques from the process mining field that rely on the domain knowledge are proposed in [BMW14] and [Man+16b]. The technique from [BMW14] requires knowledge about a single model of the overall process, and uses this model to map events in the log to the correct granularity level that is present in the model. However, for our application domain in mining on ADL activities, no overall process model on the desired granularity is available. The technique from [Man+16b] relies on domain knowledge to perform the abstract step, requiring the user to specify a low-level process model for each high-level activity. However, in the context of human behavior, it is unreasonable to expect the user to provide the process model in sensor terms for each human activity.

Instead of event abstraction on the level of the event log, unsupervised abstraction methods that work on the level of a model (e.g. [VVK09]) can also be applied to make large complex models more comprehensible. Note that such methods also do not give guidance on how to label resulting transitions in the process model. Furthermore, such approaches do not help when the event log is unstructured in a way that it is infeasible to discover a structured process model from the original event log.

## 7.5 Conclusions

We presented a novel framework to abstract events using supervised learning which has been implemented in the ProM process mining toolkit. An important part of the framework is a generic way to extract useful features for abstraction from the extensions defined in the XES IEEE standard for event logs. We propose the Damerau-Levenshtein Similarity for evaluation of the abstraction results, and motivate why it fits the application of process mining. Finally, we showed on three real life smart home datasets that application of the supervised event abstraction framework enables us to mine more precise process model description of human life compared to what could be mined from the original data on the sensor-level. Ad-ditionally, these process models contain interpretable labels on the human behavior activity level.

# Part II

# Discovering Local Process Models

**Chapter 8** We introduce the foundations of local process model mining. This chapter is based on [MT17; Tax+16e].

**Chapter 9** We present extensions of local process model mining based on utility functions and constraints. This chapter is based on [Tax+18b].

**Chapter 10** We present a set of heuristic techniques that aim to speedup local process model mining. This chapter is based on [GTZ18; Tax+16e; TGZ17].

**Chapter 11** We present techniques to efficiently mine local process models for two specific types of constraints: event gap and time gap constraints. This chapter is based on [Tax+18e].

**Chapter 12** We present a set of techniques to filter a set of local process models into a smaller set of patterns in order to reduce the pattern overload for the analyst.

**Chapter 13** We introduce the tools and implementations for local process mining. This chapter is based on [Tax+18d].

# 8 Foundations of Local Process Models

**Parts of this chapter have been published as:**

- Niek Tax, Natalia Sidorova, Reinder Haakma, and Wil M. P. van der Aalst: *Mining Local Process Models*. Journal of Innovation in Digital Ecosystems 3(2): pp. 183-196 (2016)
- Felix Mannhardt and Niek Tax: *Unsupervised Event Abstraction using Pattern Abstraction and Local Process Models*. Joint Proceedings of the Radar tracks at the 18th International Working Conference on Business Process Modeling, Development and Support (BPMDS), and the 22nd International Working Conference on Evaluation and Modeling Methods for Systems Analysis and Development (EMMSAD), and the 8th International Workshop on Enterprise Modeling and Information Systems Architectures (EMISA). CEUR-ws.org, 1859: 55-63 (2017)

In Part I, we have introduced a collection of preprocessing techniques that aid the mining of a process model from an event log that is unstructured, i.e., where the relations between activities are weak. In Part II of the thesis, we will propose a different, orthogonal, approach to mine insights from such unstructured event data that is based on mining local patterns of behavior. This contrasts the global models that are obtained with process discovery approaches.

Two existing classes of approaches to mine local patterns of behavior instead of global end-to-end process models are sequential pattern mining and episode mining. Sequential pattern mining and episode mining are limited to the discovery of *sequential orderings* (in the case of sequential pattern mining) or *partially ordered sets* of events (in the case of episode mining), while process discovery methods aim to discover a larger set of event relations that includes sequential orderings, (exclusive) choice relations, concurrency, and loops. At the same time, process discovery is limited to the discovery of a *single model* to represent the behavior of the process instances that were observed in the event, i.e., it does not address the mining of local patterns that occur *within* those process instances. In this chapter, we will introduce the basic method to mine *local* process models that are positioned in-between simple patterns (e.g. frequent sequential patterns and frequent episodes) and start-to-end models. *Local process models* (LPMs) focus on a subset of the process activities and they describe some behavioral pattern that occurs frequently *within* event sequences without necessarily describing those sequences completely.

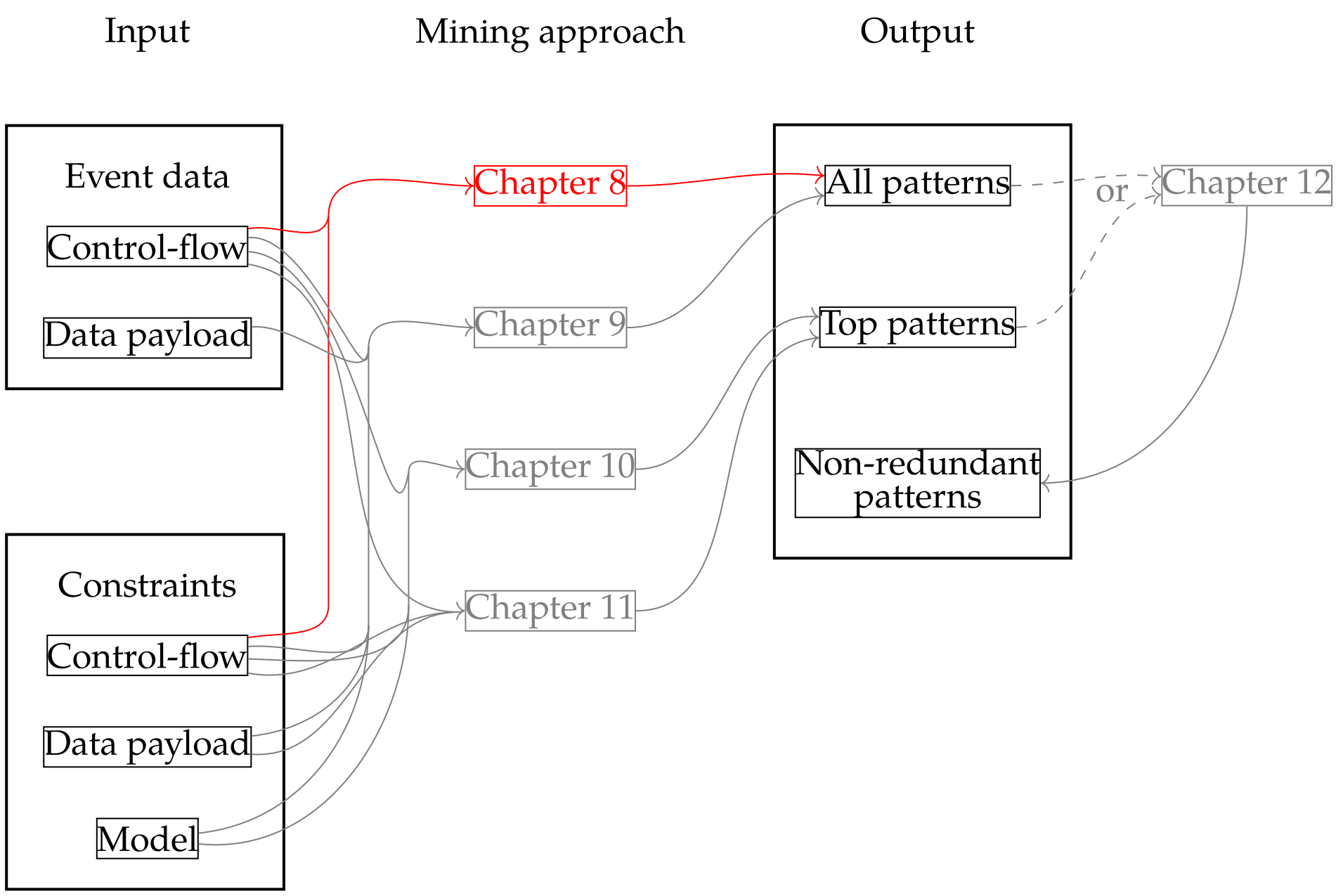

**Figure 8.1:** A taxonomy of local process model techniques.

In this chapter, we introduce the foundations of LPM mining by introducing the purpose of LPMs, quality criteria for LPMs, and an algorithm to mine LPMs from an event log. In later chapters of Part II we extend LPM mining in several directions. Chapter 9 introduces a framework for question-driven LPMs, where the definitions of pattern quality go beyond the quality criteria that we introduce in this chapter and can additionally depend on the data attributes of events and cases. Chapter 10 addresses the computational difficulties of LPM mining by proposing fast but approximate heuristic LPM mining methods that do not necessarily find all LPMs that satisfy the set of constraints, but instead mine a set of LPMs that are good according to some quality criteria.

We will now proceed by sketching a scenario to motivate the need for LPM mining where process discovery and sequential pattern mining both do not suffice. Imagine a sales department where multiple sales officers perform four types of activities: (a) register a call for tender, (b) investigate a call for tenders from the business perspective, (c) investigate a call for tenders from the legal perspective, and (d) decide on participation in the call for tender. The event sequences (Figure 8.2a) contain the activities performed by one sales officer throughout the day. The sales

| Event sequences |
|---|
| ⟨a,**a,c,b**,a,**a,c,b**,b,c⟩ |
| ⟨c,**a,c,b**,a,a,**a,b,c**,b⟩ |
| ⟨a,**a,b**,d,**c**,d,**a,b,c**,b⟩ |
| ⟨c,**a,c,b**,b,b,**a**,d,**b,c**⟩ |
| ⟨b,**a,b,c**,c⟩ |
| ⟨d,**a,c,b**,c,a,**a,c**,a,**b**⟩ |
| ⟨d,**a,b,c**,d,c,a,c,**a,b,c**⟩ |

**(a)**

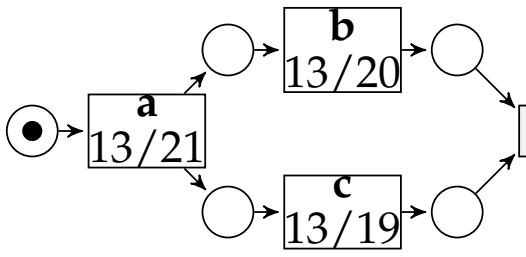

**(b)**

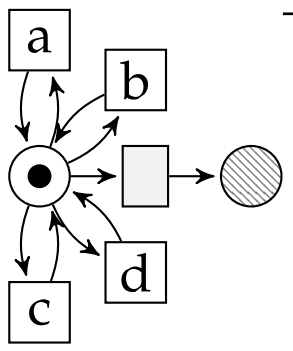

**(c)**

| Sequential patterns |
|---|
| ⟨a, b, a⟩ |
| ⟨a, b, c⟩ |
| ⟨a, c, a⟩ |
| ⟨a, c, b⟩ |
| ⟨b, a, b⟩ |
| ⟨b, a, c⟩ |
| ⟨a, a, c⟩ |

**(d)**

**Figure 8.2:** *(a)* A log L of sales officer event sequences with highlighted instances of the frequent pattern. *(b)* The local process model showing frequent behavior in L. *(c)* The Petri net discovered on L with the Inductive Miner algorithm [LFA13b]. *(d)* The sequential patterns discovered on L with the PrefixSpan algorithm [Pei+01].

officer works on different calls for tenders and not necessarily performs all activities for a particular call himself. Applying discovery algorithms, like the Inductive Miner [LFA13b], to an event log from such an unstructured domain yields process models that allow for any sequence of events (Figure 8.2c), i.e., a model that *is* or *is close to* the so-called flower model. When we apply any sequential pattern mining algorithm using a threshold of six occurrences, we obtain the seven sequential patterns of length three that are depicted in Figure 8.2d. Those sequential patterns in Figure 8.2d were obtained using the implementation of the PrefixSpan algorithm [Pei+01] that is included in the SPMF [Fou+14a] pattern mining toolkit. However, the data contains a frequent non-sequential pattern where a sales officer first performs *a*, followed by *b* and a *c* in arbitrary order ( Figure 8.2b). This pattern cannot be found with existing process discovery and sequential pattern mining techniques. The two numbers shown in the transitions (i.e., rectangles) represent (1) the number of events of this type in the event log that fit this LPM and (2) the total number of events of this type in the event log. For example, 13 out of 19 events of type *c* in the event log fit the transition that is labeled *c* in the model, which are indicated in **bold** in the log in Figure 8.2a. Underlined sequences indicate non-consecutive instances, i.e. instances with non-fitting events in-between the events forming the instance of the LPM. For example, the third sequence of Figure 8.2a, ⟨..., **a**, **b**, d, **c**, ...⟩ indicates that the a event and the succeeding b event fit the LPM of Figure 8.2b, however, the d event does not fit the behavior of this model (the process model has no transition that is labeled d).

The method that we will describe in this chapter to extract frequently occurring LPMs that allow for choice, concurrency, loops, and sequence relations leverages process trees [BDA12b] to define the search space. The approach is based on recursive exploration of candidate process trees. In this thesis, we will generally visually depict LPMs in their Petri net representation. Note that the transformation from the process tree representation of an LPM in the backend to the Petri net representation that is used for visualization is trivial. However, this is merely for visualization purposes, and we should keep in mind that the LPMs that we are able to mine are restricted to the class of behavior that can be expressed by process trees, thereby some models from the class of Petri nets remain unreachable. As a second note, observe that we could trivially represent LPMs in other process model notations, such as BPMN [Obj11], EPC [KSN92], UML activity diagram [Int12], or UML statechart diagram [Int12], as process trees can be converted into such modeling notations. We define five quality dimensions that express the degree of representativeness of a local process model with regard to an event log: *support*, *confidence*, *language fit*, *coverage*, and *determinism*. We describe pruning approaches for some of those quality measures and show how they can be used to make the search over process trees more efficient.

The chapter is organized as follows. In Section 8.1 we provide a definition of LPM mining and describe quality measures for LPMs in the context of an event log. Section 8.2 introduces an LPM mining approach. Section 8.3 introduces a selection and ranking mechanism for LPMs and discusses pruning methods. Section 8.4 describes how instances of an IPM van be detected in an event log using the alignment algorithm [ADA11]. Section 8.5 describes related work in the fields of process discovery and pattern mining. Section 8.6 demonstrates the relevance of the proposed technique using two real-life data sets and compares the results with the results obtained with several related techniques. In Section 8.7 we will show a specific application of LPMs, thereby showing that LPMs can be useful other than for the purpose of finding frequent sub-processes in the event log as a form of exploratory analysis of the event data. This application addresses the task of abstracting events into higher level events, thereby closely linking to the work that we presented in Chapter 7. Section 8.8 discusses several limitations of the LPM mining approach that is described in this chapter. Section 8.9 concludes this chapter.

## 8.1 Defining Local Process Models

A local process model (LPM) aims to describe frequent behavior in an event log in local patterns, i.e. smaller patterns. An LPM does not aim to describe all behavior in a trace completely, but instead aims to describe traces partially, focusing on frequently occurring patterns. An LPM *LPM* represents behavior over some alphabet of activities $\Sigma$ and it accepts a language $\mathfrak{L}(LPM) \subseteq \mathcal{P}(\Sigma^*)$. We will now start by defining several concepts related to instances of a process model in an event log and then continue by defining local process models by using those concepts. All definitions that we will provide in this section are independent of the process model notation or formalism that is used to express a local process model, i.e., they can be used for any type of process model for which the language is defined. However, all examples that we provide in this section will be in the form of process trees.

**Definition 8.1 (Subsequence and Landmark).** A sequence $\sigma' \in \Sigma^*$ is a *subsequence* of another sequence $\sigma \in \Sigma^*$ (with $|\sigma'| \leq |\sigma|$), denoted $\sigma' \sqsubseteq \sigma$, if there exists a sequence of positions $\lambda = \langle i_1, i_2, \dots, i_{|\sigma'|} \rangle$ with each $i_j \in \mathbb{N}$ and $1 \leq i_1 < i_2 < \dots < i_{|\sigma'|} \leq |\sigma|$ such that $\forall_{i_j \in \lambda} \sigma(i_j) = \sigma'(j)$. Such a sequence $\lambda$ we call a *landmark* of $\sigma'$ in $\sigma$. Function $\xi$ projects a sequence on a landmark, i.e., $\xi(\sigma, \lambda) = \langle \sigma(\lambda(1)), \sigma(\lambda(2)), \dots, \sigma(\lambda(|\lambda|)) \rangle = \sigma'$.

Observe that multiple landmarks for a single subsequence can exist within a single sequence. For example, for sequences $\sigma = \langle a, b, c, a, b, c \rangle$ and $\sigma' = \langle a, c \rangle$, we have $\sigma' \sqsubseteq \sigma$ with the following landmarks: $\langle 1, 3 \rangle$, $\langle 1, 6 \rangle$, $\langle 4, 6 \rangle$.

**Definition 8.2 (Instances of a Process Model in a Trace).** A pair $(\sigma, \lambda)$ of trace $\sigma$ and landmark $\lambda$ is said to be an instance of process model $M \in \mathcal{M}$ if $\xi(\sigma, \lambda) \in \mathfrak{L}(M)$.

For example, for process model $M = \to(a, \wedge(b, c))$ and traces $\sigma_1 = \langle a, b, c \rangle$ and $\sigma_2 = \langle a, b, c, a, c, b \rangle]$, we have the following instances of M: $\{(\sigma_1, \langle 1, 2, 3 \rangle), (\sigma_2, \langle 1, 2, 3 \rangle), (\sigma_2, \langle 1, 2, 5 \rangle), (\sigma_2, \langle 1, 3, 6 \rangle), (\sigma_2, \langle 4, 5, 6 \rangle), (\sigma_2, \langle 1, 5, 6 \rangle)\}$.

Often, we are interested in the instances of a process model $M \in \mathcal{M}$ in an event log $L \in \mathcal{B}(\Sigma^*)$ such that each event of the log is only explained by a single instance of M. By preventing the same event to be part of multiple instances of the same pattern we prevent overcounting the number of instances of long patterns.

**Definition 8.3 (Strongly Overlapping Instances).** Given two instances $(\sigma, \langle i_1, \dots, i_n \rangle)$ and $(\sigma', \langle i'_1, \dots, i'_{n'} \rangle)$ of a process model $M \in \mathcal{M}$ respectively in traces $\sigma \in \Sigma^*$ and $\sigma' \in \Sigma^*$, $(\sigma, \langle i_1, \dots, i_n \rangle)$ and $(\sigma', \langle i'_1, \dots, i'_{n'} \rangle)$ are *strongly overlapping* if $\sigma = \sigma'$ and $\{i_1, \dots, i_n\} \cap \{i'_1, \dots, i'_m\} \neq \emptyset$.

For process model $M = \to(a, \wedge(b, c))$ and traces $\sigma_1 = \langle a, b, c \rangle$ and $\sigma_2 = \langle a, b, c, a, c, b \rangle\}$, instances $(\sigma_2, \langle 1, 2, 3 \rangle)$ and $(\sigma_2, \langle 1, 2, 5 \rangle)$ are strongly overlapping instances since they share events $\sigma_2(1)$ and $\sigma_2(2)$. In contrast, $(\sigma_2, \langle 1, 2, 3 \rangle)$ and $(\sigma_2, \langle 4, 5, 6 \rangle)$ are two instances that are not strongly overlapping, since they do not share any events.

Definition 8.3 follows the standard definition of *overlapping* pattern instances that is common in the pattern mining research field. Under this definition, for $M = \to(a, \to(b, c))$ and trace $\sigma = \langle a, b, a, b, c, c \rangle$, landmarks $\langle 1, 2, 5 \rangle$ and $\langle 3, 4, 6 \rangle$ are considered to be non-overlapping since they do not share any event. However, while these occurrences are non-overlapping in the traditional sense, they are interleaved in the sense that the first landmark starts at index 1 and continues until index 5, while the second landmark starts at position 3, i.e., halfway the execution of the first occurrence. In contrast to sequential patterns, our choice to use process models to represent frequent patterns enables us to express concurrent executions explicitly *within* the structure of the pattern. For example, if the behavior of pattern M frequently occurs in parallel to itself, then we would instead want to mine pattern $M' = \wedge(\rightarrow(a, \rightarrow(b, c)), \rightarrow(a, \rightarrow(b, c)))$ instead of M. In order to favor to represent concurrency as much as possible *within* the pattern, we define a weaker notion of overlapping instances than the one that is traditionally used in the pattern mining field. This weaker notion of overlapping instances considers two instances of a pattern to be overlapping if they are executed concurrently.

**Definition 8.4 (Weakly Overlapping Instances).** Given two instances $(\sigma, \langle i_1, \dots, i_n \rangle)$ and $(\sigma, \langle i'_1, \dots, i'_{n'} \rangle)$ of a process model $M \in \mathcal{M}$ respectively in traces $\sigma \in \Sigma^*$ and $\sigma' \in \Sigma^*$, $(\sigma, \langle i_1, \dots, i_n \rangle)$ and $(\sigma, \langle i'_1, \dots, i'_{n'} \rangle)$ are *weakly overlapping* if $\sigma = \sigma'$ and $[i_1, i_n] \cap [i'_1, i'_{n'}] \neq \emptyset$.

It follows from Definition 8.4 that in order for two instances $(\sigma, \langle i_1, \dots, i_n \rangle)$ and $(\sigma', \langle i'_1, \dots, i'_m \rangle)$ to be *non-overlapping*, $\sigma \neq \sigma' \vee i'_m < i_1 \vee i_n < i'_1$. It trivially holds that two instances that are not weakly overlapping are also not strongly overlapping. Therefore, weak overlap is a weaker notion of overlapping than strong overlap. Often, we are concerned with the instances of a process model in a collection of traces.

**Definition 8.5 (Set of Non-Overlapping Instances in a Trace Set).** Given a set of traces $S = \{\sigma_1, \sigma_2, \dots\}$ with each $\sigma_k \in \Sigma^*$ and a process model $M \in \mathcal{M}$, a set of instances I of M in S is a set of *non-overlapping* instances if and only if $\nexists i, i' \in I$ with $i \neq i'$ such that i and i′ are weakly overlapping.

We are interested in the largest set of non-overlapping instances of some process model $M \in \mathcal{M}$ in a given set of traces $L \in \mathcal{P}(\Sigma^*)$ in order to count the frequency of the behavior that is specified by M in L.

**Definition 8.6 (Set of Maximal Non-Overlapping Instances in a Trace Set).** Given a set of traces $S = \{\sigma_1, \sigma_2, \dots\}$ with each $\sigma_k \in \Sigma^*$ and a process model $M \in \mathcal{M}$, a set of non-overlapping instances I of M in S is a set of *maximal* non-overlapping instances if there does not exist another non-overlapping set of instances I′ such that $\sum_{(\sigma,\lambda) \in I} |\lambda| < \sum_{(\sigma',\lambda') \in I'} |\lambda'|$.

For example, for a set of traces $\{\sigma_1 = \langle a, b \rangle, \sigma_2 = \langle a, a, b, a, b, b \rangle\}$ and for process model $M = \rightarrow(a, b)$, a maximal non-overlapping instance set is $I = \{(\sigma_1, \langle 1, 2 \rangle), (\sigma_2, \langle 2, 3 \rangle), (\sigma_2, \langle 4, 5 \rangle)\}$. Note that index 1 and index 6 of trace $\sigma_2$ also contain a and b events. However, using these events as an instance of M disallows us from counting the instances with landmarks $\langle 2, 3 \rangle$ and $\langle 4, 5 \rangle$ at the same time as $\langle 1, 6 \rangle$, as that would lead would make the set of instances weakly overlapping. Trading in the events at indices $\langle 2, 3 \rangle$ and $\langle 4, 5 \rangle$ for the events at indices 1 and 6 leads to a non-overlapping set of instances $I' = \{(\sigma_1, \langle 1, 2 \rangle), (\sigma_2, \langle 1, 6 \rangle)\}$ with $\sum_{(\sigma',\lambda') \in I'} |\lambda'| = 4$. However, $\sum_{(\sigma,\lambda) \in I} |\lambda| = 6$ and, thus I′ is not maximal. Furthermore, observe

that a maximal set of non-overlapping instances is not necessarily unique. The non-overlapping set of instances $I'' = \{(\sigma_1, \langle 1,2\rangle), (\sigma_2, \langle 2,3\rangle), (\sigma_2, \langle 4,5\rangle)\}$ illustrates this, since $\sum_{(\sigma,\lambda)\in I''} |\lambda| = \sum_{(\sigma,\lambda)\in I} |\lambda| = 6$, therefore, $I''$ and $I$ are both maximal.

In many cases we are not concerned with which set of maximal non-overlapping instances we retrieve for a given process model $M \in \mathcal{M}$ and trace $\sigma' \in \Sigma^*$. However, we do care which maximal non-overlapping instances we retrieve when we are dealing with process models that contain loops. Consider for example process model $M = \rightarrow(a, \rightarrow(b, \circlearrowleft(\rightarrow(c, b))))$ and trace $\sigma = \langle a, b, c, a, b\rangle$. Both $I = \{(\sigma, \langle 1,2\rangle), (\sigma, \langle 4,5\rangle)\}$ and $I' = \{(\sigma, \langle 1,2,3,5\rangle)\}$ are maximal non-overlapping sets of instances (consisting of four events in total). However, the loop in the process would remain unused if we interpret $\sigma$ as $I = \{(\sigma, \langle 1,2\rangle), (\sigma, \langle 4,5\rangle)\}$. In fact, the much simpler model $M' = \rightarrow(a, b)$ suffices in order to satisfy instances I. Therefore, in order to obtain added value of the loop operator, we prefer the maximal non-overlapping set of instances $I'$ over $I$. We do so by defining the notion of a *minimal set of maximal non-overlapping instances*.

**Definition 8.7 (Minimal Set of Maximal Non-Overlapping Instances in a Trace Set).** Given a set of traces $S = \{\sigma_1, \sigma_2, \dots\}$ with each $\sigma_k \in \Sigma^*$ and a process model $M \in \mathcal{M}$, the set of maximal non-overlapping instances I of M in S is a *minimal* set of maximal non-overlapping instances if there does not exist another set of maximal non-overlapping instances $I'$ such that $|I| > |I'|$.

Observe that for process model $M = \rightarrow(a, \rightarrow(b, \circlearrowleft(\rightarrow(c, b))))$, trace $\sigma = \langle a, b, c, a, b\rangle$, and maximal non-overlapping sets of instances $I = \{(\sigma, \langle 1,2\rangle), (\sigma, \langle 4,5\rangle)\}$ and $I' = \{(\sigma, \langle 1,2,3,5\rangle)\}$, only $I' = \{(\sigma, \langle 1,2,3,5\rangle)\}$ is a minimal set of maximal non-overlapping instances (since $|I'| = 1$ and $|I| = 2$.

In practice, we generally want to detect a minimal set of maximal non-instances of a process model $M \in \mathcal{M}$ in an event log $L \in \mathcal{B}(\Sigma^*)$, while Definitions 8.3, 8.4, 8.6 and 8.7 are all defined for given traces $\sigma, \sigma' \in \Sigma^*$. In order the obtain the instances in an event log $L \in \mathcal{B}(\Sigma^*)$, we start by detecting the set of instances I in each trace variant of $\tilde{L}$. We now lift the notion of instances I of a process model M on a set of traces $L' \in \mathcal{P}(\Sigma^*)$ to a multiset of instances $I'$ of M on an event log $L' \in \mathcal{B}(\Sigma^*)$.

**Definition 8.8 (Minimal Multiset of Maximal Non-Overlapping Instances in an Event Log).** Given an event log L and a process model $M \in \mathcal{M}$, the minimal multiset of maximal non-overlapping instances of M in L is the multiset $I(\sigma, \langle i_1, \dots, i_n\rangle) = L(\sigma)$ with $dom(I)$ the set of maximal non-overlapping instances in the trace set $\tilde{L}$.

For example, for $L = [\sigma_1 = \langle a, b\rangle^{10}, \sigma_2 = \langle a, a, b, a, b, b\rangle^2]$ and $M = \rightarrow(a, b)$, a minimal set of maximal non-overlapping instance for the set of traces $\tilde{L}$ is $I = \{(\sigma_1, \langle 1,2\rangle), (\sigma_2, \langle 2,3\rangle), (\sigma_2, \langle 4,5\rangle)\}$ and therefore the minimal multiset of maximal non-overlapping instances of the event log are $[(\sigma_1, \langle 1,2\rangle)^{10}, (\sigma_2, \langle 2,3\rangle)^2, (\sigma_2, \langle 4,5\rangle)^2]$.

| Event sequences |
|---|
| ⟨a,**a,c,b**,a,**a,c,b**,b,c⟩ |
| ⟨c,**a,c,b**,a,a,**a,b,c**,b⟩ |
| ⟨a,**a,b**,d,**c**,d,**a,b,c**,b⟩ |
| ⟨c,**a,c,b**,b,b,**a**,d,**b,c**⟩ |
| ⟨b,**a,b,c**,c⟩ |
| ⟨d,**a,c,b**,c,a,**a,c**,a,**b**⟩ |
| ⟨d,**a,b,c**,d,c,a,c,**a,b,c**⟩ |

**(a)**

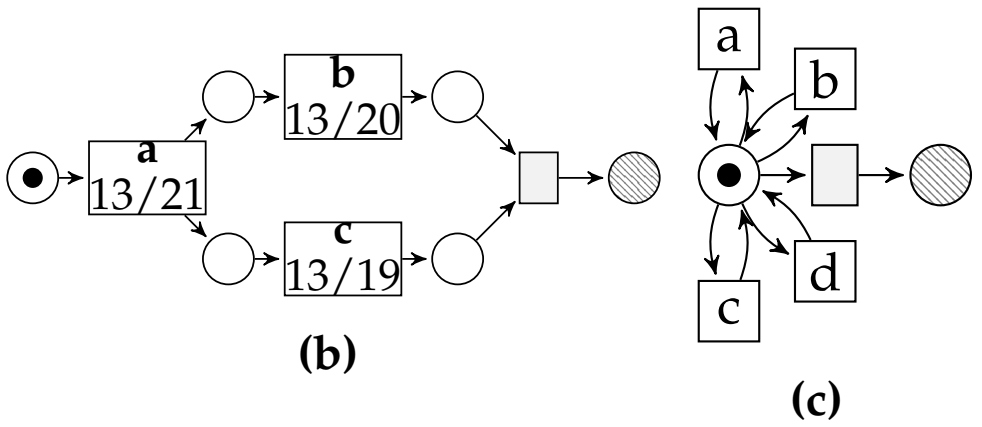

**Figure 8.3:** An event log and two process models, repeated from Figure 8.2.

**Definition 8.9 (Minimal Multiset of Maximal Non-Overlapping Instances Oracle).** Given an event log $L \in \mathcal{B}(\Sigma^*)$ and a process model $M \in \mathcal{M}$, we define $\Gamma_M$ to be an oracle function that generates a minimal multiset of maximal non-overlapping instances of M in L.

Section 8.4 addresses the question how $\Gamma_M$ can be computed in practice. Based on the minimal multiset of maximal non-overlapping instances I that we obtain by applying $\Gamma_M$ to L we can define several quality measures that expresses how well a process model M describes the behavior in an event log L. For example, the larger the minimal multiset of maximal non-overlapping instances of $M \in \mathcal{M}$ in $L \in \mathcal{B}(\Sigma^*)$, the more frequently we can say M to be occurring in L. This brings us to the first quality measure for LPMs, called *support*, which is is of special importance to LPM mining and even to pattern mining in general.

**Definition 8.10 (Support).** For a process model $M \in \mathcal{M}$ and a labeled event log $L \in \mathcal{B}(\Sigma^*)$, $\mathit{support}(M, L) = |\Gamma_M(L)|$.

This definition of *support* is analogous to the notion of *repetitive support* as used in the field of sequential pattern mining, which is concerned with counting not only pattern instances (occurrences) that are repeating in *different* sequences but also those repeating *within* each sequence. Furthermore, like repetitive support in frequent pattern mining, our support definition is *non-overlapping* as it is based on $\Gamma_M$, i.e., for a trace $L = [\langle a, b, a, b, c, c\rangle]$ a pattern M with $\mathfrak{L}(M) = \{\langle a, b, c\rangle\}$ the support of $\mathit{support}(M, L) = 1$. This non-overlapping requirement for pattern instances prevents over-counting the support of long patterns and it makes the set of instances that are counted as part of the support informative to users.

The support of the process model of Figure 8.3b on Figure 8.3a is thirteen (with the thirteen counted occurrences highlighted in Figure 8.3a). The support of the process model Figure 8.3c on Figure 8.3a is seven: the model of Figure 8.3c allows for all behavior and as a result, each trace $\sigma$ of the log is one single occurrence of the pattern with landmark $\langle 1, 2, \ldots, |\sigma| \rangle$.

A process model $M \in \mathcal{M}$ is a local process model for a given log L, a given set of LPM quality measures and a set of thresholds for these quality measures when M satisfies the threshold for each quality measure.

**Definition 8.11 (Local Process Model).** A process model $M \in \mathcal{M}$ is a *local process model* for a given labeled event log $L \in \mathcal{B}(\Sigma^*)$ given a set of LPM quality measures $\{q_1, \ldots, q_n\}$ with each quality measure $q_i \in \mathcal{M} \times \mathcal{B}(\Sigma^*) \rightarrow \mathbb{R}$ and set of thresholds $T = \{t_1, \ldots, t_n\}$ with each $t_i \in \mathbb{R}$ if and only if $\forall_{q_i \in Q} q_i(M, L) \geq t_i$.

We have already introduced the *support* quality measure and will now continue by introducing several additional quality measures for LPMs using the log and the two process models of Figure 8.2 as a running example. In order to define additional quality measures, we lift function $\xi$ to multisets of process model instances in such a way that it returns the multiset of corresponding trace fragments, i.e., $\xi([(\sigma_1, \lambda_1)^{w_1}, (\sigma_2, \lambda_2)^{w_2}, \ldots]) = [\xi(\sigma_1, \lambda_1)^{w_1}, \xi(\sigma_2, \lambda_2)^{w_2}, \ldots]$.For example, for $\sigma_1 = \langle a, b, d, c \rangle$ and $\sigma_2 = \langle a, c, b, a, a, c, b \rangle$ and process model $M = \rightarrow(a, \wedge(b, c))$ resulting in $\Gamma_M([\sigma_1, \sigma_2]) = [(\sigma_1, \langle 1, 2, 4 \rangle), (\sigma_2, \langle 1, 2, 3 \rangle), (\sigma_2, \langle 4, 6, 7 \rangle)]$, we have $\xi(\Gamma_M(L)) = [\langle a, b, c \rangle, \langle a, c, b \rangle^2]$ where $\langle a, b, c \rangle$ originates from $(\sigma_1, \langle 1, 2, 4 \rangle)$ and the two instances $\langle a, c, b \rangle$ originate from $(\sigma_2, \langle 1, 2, 3 \rangle)$ and $(\sigma_2, \langle 4, 6, 7 \rangle)$.

## 8.1.1 Confidence

A useful LPM is not only frequent in the absolute sense, but also explains a high ratio of the events in the log with the activities that are represented by the LPM, i.e., an LPM that describes activities a, b and c should ideally explain a large share of the a, b, and c events of the log. We call this quality dimension *confidence*.

An event e from a log L fits an model M if there is an instance in $(\sigma, \lambda) \in \Gamma_M(L)$ such that $\exists_{i \in \lambda} \sigma(i) = e$. The confidence of a process model M on event log $L \in \mathcal{B}(\Sigma^*)$ with respect to an activity $a \in \Sigma$ is the ratio of events of type a in L that fit M:

$$\mathit{confidence}(a, M, L) = \frac{\#(a, \xi(\Gamma_M(L)))}{\#(a, L)}$$

In order to obtain a single confidence value for a process model on a log, we aggregate the confidence values over all activities in the pattern. We are interested specifically in patterns where none of the activities represented in the pattern has a low confidence, because we argue that when one activity in the pattern has low confidence, a more suitable pattern for this event log could be obtained by removing the low-confidence activity from the pattern. Therefore, we aggregate the confidence values of the activities using the harmonic mean, which compared to the geometric and arithmetic means is more sensitive to a single low value. We define the confidence of an process model M on an event log L to be the harmonic mean of the individual confidence scores of the event types of M.

**Definition 8.12 (Confidence).** For a process model $M \in \mathcal{M}$ that describes the set of activities $\Sigma_M$ and a labeled event log $L \in \mathcal{B}(\Sigma^*)$, $\mathit{confidence}(M, L) = \frac{|\Sigma_M|}{\sum_{a \in \Sigma_M} \frac{1}{\mathit{confidence}(a, M, L)}}$.

The confidence of the process model of Figure 8.2b on Figure 8.2a is thirteen out of twenty-one occurrences of activity a, thirteen out of twenty occurrences of activity b, and thirteen out of nineteen occurrences of activity c. The harmonic mean evaluates to $\frac{3}{\frac{21}{13}+\frac{20}{13}+\frac{19}{13}} = \frac{3}{\frac{1067}{260}} = \frac{780}{1067} \approx 0.731$. The confidence of the process model Figure 8.2c on Figure 8.2a is 1, since all events of activities a, b, and c fit the model.

### 8.1.2 Language Fit

Language fit expresses the ratio of the behavior that is allowed by the LPM that is observed in the event log. This quality criterion provides a counterbalance to support and fitness: while support and confidence favor patterns that are more general in the behavior that they allow for, language fit punishes patterns that are overly general. We can say that LPMs that allow for much more behavior than what is observed are likely to overgeneralize and therefore do not describe the behavior in the event log well. The language fit of an LPM M given log L is:

$$\mathit{language_fit}(M, L) = \frac{|\{\sigma \in \mathfrak{L}(M) \mid \exists_{\sigma' \in \xi(\Gamma_M(L))} \sigma = \sigma'\}|}{|\mathfrak{L}(M)|}$$

Since $|\mathfrak{L}(M)| = \infty$ in case M contains a loop, $\mathit{language_fit}(M, L) = 0$ for any M containing a loop. Restricting the language and the LPM instances to sequences of a bounded length allows us to approximate language fit for models with infinite size language. We define *n-restricted language* for a process model $M \in \mathcal{M}$, consisting of all language traces of at most length $n \in \mathbb{N}$, as $\mathfrak{L}_n(M) = \{\sigma \in \mathfrak{L}(M) \mid |\sigma| \leq n\}$. The n-restricted language is guaranteed to be of finite size, independent of the operators in the process tree. The 5-restricted language of the process tree $M = \rightarrow(a, \times(\wedge(c, d), \circlearrowleft(\rightarrow(b, a))))$ is $\mathfrak{L}_5(M) = \{\langle a \rangle, \langle a, b, a \rangle, \langle a, b, a, b, a \rangle, \langle c, d \rangle, \langle d, c \rangle\}$.

**Definition 8.13 (Language Fit).** For a process model $M \in \mathcal{M}$ and a labeled event log $L \in \mathcal{B}(\Sigma^*)$, $\mathit{language_fit}_n(M, L) = \frac{|\{\sigma \in \mathfrak{L}_n(M) \mid \exists_{\sigma' \in \xi(\Gamma_M(L))} \sigma = \sigma'\}|}{|\mathfrak{L}_n(M)|}$

The multiset of occurrences of the process model of Figure 8.2b that were observed in Figure 8.2a are $[\langle a, c, b \rangle^6, \langle a, b, c \rangle^7]$ (instances are highlighted in Figure 8.2a). Therefore, the set of the types of occurrences of the LPM is $\{\langle a, c, b \rangle, \langle a, b, c \rangle\}$. The language of the model $\mathfrak{L}_n(M)$ for all $n \geq 3$ is also $\{\langle a, c, b \rangle, \langle a, b, c \rangle\}$. Therefore, the language fit is $\frac{|\{\langle a,c,b \rangle, \langle a,b,c \rangle\}|}{|\{\langle a,c,b \rangle, \langle a,b,c \rangle\}|} = \frac{2}{2} = 1$.

To calculate the language fit for the process model Figure 8.2c for $\mathfrak{L}_n(M)$ with $n = 11$ (i.e., the length of the longest trace), observe that $\mathfrak{L}_{11}(M) = 4^{11} = 4194304$.

Since each trace fits the model, each trace corresponds to one observation of the model and since each trace is unique the set of the types of occurrences of the model contains 11 elements. The language fit therefore is $\frac{11}{4194304}$. This shows that the language fit measure heavily favors process models that do not generalize too much beyond the observed behavior.

### 8.1.3 Determinism

Language-fit by itself is not sufficient to counterbalance support and confidence and prevent overgeneralizing patterns. Note that when the event log becomes highly random and as a result almost all orderings over some set of activities is observed at least once in the event log, then language fit becomes high even for flower models. It could be the case however that some of these orderings are very frequent and others occur only once. Determinism is a measure to punish LPMs when they allow for many behaviors.

LPMs that are highly deterministic have more predictive value in with regard to future behavior. When the language of LPM M contains traces of type $\sigma \cdot \langle a \rangle \cdot \gamma_1$ and $\sigma \cdot \langle b \rangle \cdot \gamma_2$, the continuation of the trace after observing prefix $\sigma$ can be either a or b, leaving some uncertainty. LPMs with a high degree of certainty are preferable over LPMs with a low degree of certainty. A measure for the determinism quality dimension is dependent on the process model structure and not only on its language. Note that this makes the determinism measure dependent on the notation in which the process model is represented. In the remainder, we introduce determinism in a generic way, by defining it in terms of the *states* of the process model. What are the states of a process model and how these states correspond to prefixes of sequences is dependent on the process model notation. How to calculate determinism concretely for the case of Petri nets will be discussed in Section 8.4.3, after introducing how to compute $\Gamma_M(L)$, as the determinism calculation for Petri nets is dependent on and closely linked to the calculation of $\Gamma_M(L)$.

Let $\mathcal{R}(M)$ be a set of reachable states of an LPM M. $\mathcal{W}_L : \mathcal{R}(M) \rightarrow \mathbb{N}$ represents a function assigning the number of times a state is reached while replaying the instances $\xi(\Gamma_M(L))$ on M. $\mathcal{D} : \mathcal{R}(M) \rightarrow \mathbb{N}$ represents a function assigning the number of transitions enabled in a certain state in *LPM*. Determinism is defined as:

**Definition 8.14 (Determinism).** For a process model $M \in \mathcal{M}$ and a labeled event log $L \in \mathcal{B}(\Sigma^*)$, *determinism*$(M, L) = \frac{\sum_{m \in \mathcal{R}(M)} \mathcal{W}_L(m)}{\sum_{m \in \mathcal{R}(M)} \mathcal{W}_L(m) \cdot \mathcal{D}(m)}$.

The determinism of the process model of Figure 8.2c on Figure 8.2a is easy to calculate even without having introduced the concrete calculation for Petri nets. There is only one marking from which transitions are fired, and from that marking there are five transitions enabled, yielding a determinism score of $\frac{1}{5} = 0.2$.

The determinism of the process model of Figure 8.2b on Figure 8.2a is calculated as follows: from the initial marking, there is only one transition enabled. After firing a, a marking is reached where two transitions are fired: b and c. If b is fired, after that only c is enabled and if alternatively c is fired then after that only b is enabled. Therefore, the third transition firing of the process model is always the only enabled transition. The determinism is $\frac{1+1+1}{1+2+1} = \frac{3}{4} = 0.75$.

The quality dimension of determinism closely relates to the quality dimension of precision that is used in process discovery (See Chapter 3 for an in-depth discussion of precision and precision measures). Precision measures in process discovery aim to quantify the share of the behavior that is allowed by the process model that was observed in the event log. In contrast to precision, determinism does not take into account which of the possible paths in the model have actually been seen observed in the event data.

### 8.1.4 Coverage

Larger LPMs can represent a larger number of events from the log in a single pattern instance. Support calculates the number of pattern instances of an LPM, but it does not represent the share of behavior of the event log that is represented by the pattern. For a given M and event log L, coverage is defined as the ratio of the events in the log that are part of a pattern instance.

**Definition 8.15 (Coverage).** For a process model $M \in \mathcal{M}$ and a labeled event log $L \in \mathcal{B}(\Sigma^*)$, $coverage(M, L) = \frac{\sum_{\sigma \in \xi(\Gamma_M(L))} |\sigma|}{\sum_{\sigma \in L} |\sigma|}$.

The coverage of the process model of the process model of Figure 8.2c on the event log of Figure 8.2a is simply 1, since the flower model explains all events. The coverage of the process model of the process model of Figure 8.2b on the event log of Figure 8.2a is calculated as follows: 39 events fit the pattern (13 of activity a, b, and of c) out of 56 events in total in the log, yielding a coverage of $\frac{39}{56} \approx 0.696$.

## 8.2 Local Process Model Mining Approach

LPM mining aims to discover a *collection* of LPM patterns instead of a *single* model. Each LPM individually captures only a subset of the activities in the event log, but the set of all mined LPMs together might span all activities in the log. The *support* measure that we introduced in Definition 8.10 plays a central role in LPM mining: we aim to obtain the set of all LPMs that meet a given threshold in support.

Our definitions for the fundamentals of LPM mining in Section 8.1 were independent of the process model notation or formalism, i.e., they can be used for any process model notation that defines the language of a process model. In this remainder of this chapter, we will restrict ourselves to process trees. Process trees are a pragmatic choice to define a search space over LPMs due to their tree structure. Therefore, in the remainder of the chapter, the universe of process models $\mathcal{M}$ will refer to the set of all possible process trees.

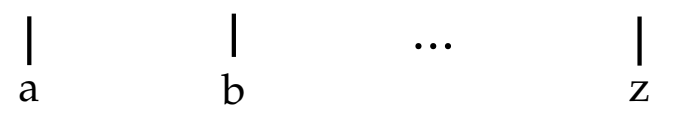

**Figure 8.4:** Set of elementary process trees over $\Sigma = \{a, b, \ldots, z\}$.

The design choice of using process trees for LPM mining has the implication that some types of behavior that are outside the expressive power (often called *representational bias*) of process trees can no longer be discovered. Examples of such types of behavior that are outside of the representational bias of process trees are *long-term dependencies* and the *milestone pattern*. However, we believe that supporting arbitrary combinations of sequential ordering ($\rightarrow$), (exclusive) choices ($\times$), concurrent execution ($\wedge$), and loops / repeated execution ($\circlearrowright$) is already a significant step beyond the expressive power of related types of pattern mining techniques that we will describe in Section 8.5. Furthermore, the process tree notation is easily extensible and in principle, new process tree operators could be defined for constructs such as long-term dependencies and milestone patterns.

We now introduce an approach to mine a collection of LPM models that consists of four main steps:

**1) Generation** We start by generating the *initial set of candidate LPMs* $CL_1 = \{a \mid a \in Activities(L)\}$, i.e., each candidate LPM in $CL_1$ is a process tree that consists of a single node: a leaf node representing an activity from the log. Figure 8.4 shows this set of elementary process trees for an event log over alphabet $\Sigma = \{a, b, \ldots, z\}$. Next, we create an empty set of selected LPMs $SL = \emptyset$ that we will use in later steps to store all *selected LPMs*, i.e., the LPMs that meet some set of quality criteria.

**2) Evaluation** We evaluate LPMs in the most recent candidate set $CL_i$ based on a set of quality criteria.

**3) Selection** We select a set $SCL_i \subseteq CL_i$ based on the evaluation results. We keep track of all selected LPMs in set $SL$, i.e., $SL = SL \cup SCL_i$. We stop the mining procedure if $SCL_i = \emptyset$ or if the number of iterations exceeds some parameter *max_iterations* that specifies the maximum number of mining iteration, i.e. $i \geq max_iterations$.

**4) Expansion** We expand $SCL_i$ into a set of larger, expanded, candidate process models, $CL_{i+1}$ using a set of expansion rules. After expanding $SCL_i$ into $CL_{i+1}$, we go back to step 2 using the newly created candidate set.

Conceptually, we define the expansion operation of an LPM as a function $exp : \mathcal{M} \rightarrow \mathcal{P}(\mathcal{M})$ that specifies for each process tree a set of process trees that it can be expanded into. Such a function can then be used in step 4 of the LPM mining procedure to generate the candidate LPMs, i.e., $CL_{i+1} = \{m \mid m' \in SCL_i \wedge m \in exp(m')\}$.

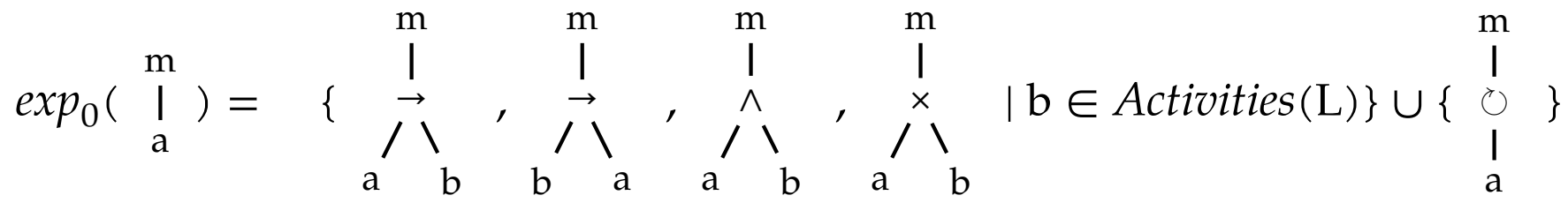

**Figure 8.5:** The specification of $exp_0$, which specifies the set of process tree expansion operations of leaf node a of process tree m.

Note that this high-level description of LPM mining does not detail how LPMs should be evaluated, selected, or expanded. We now continue by presenting a naive initial attempt to the expansion function in Section 8.2.1, which we will refine in Section 8.2.2 through Section 8.2.5. The evaluation and selection of LPMs for an event log will be discussed in Section 8.3.

### 8.2.1 An Initial Definition of the Expansion Step: $exp_0$

We will now define $exp_0$, an initial definition of the expansion step *exp* which is simple, but as we will show has problems with redundancy in the search space. Expansion function $exp_0$ replaces a leaf node $a \in Activities(L)$ (recall from the definition of process trees that leaf nodes represent activities) of the process tree by an operator node n. This operator node n is of one of the operator node types $n \in \{\rightarrow, \wedge, \times, \circlearrowright\}$ and the replaced activity node a is added as one of the child nodes of n. According to our process tree definition of Chapter 2, the loop operator $\circlearrowright$ contains one child node while the other operators $\rightarrow$, $\wedge$, and $\times$ need two child nodes. In case $n \in \{\rightarrow, \wedge, \times\}$, the second child node is created by selecting a new activity $b \in Activities(L)$ and adding an activity node for activity b. Figure 8.5 shows the set of expansion operations to leaf node a, consisting of five possible expansions. Two types of expansion operations are defined for the sequence ($\rightarrow$) operator. In contrast, only one expansion operation is defined for the $\wedge$ and $\times$ operators due to the commutativity of those operators. Furthermore, only one expansion operation is defined for the loop operator since $ar(\circlearrowright) = 1$.

For the loop ($\circlearrowright$) operator there is only a single expansion possible since this operator by definition only has one child node. Note that this definition of the loop operator deviates from the definition of the loop operator in the original process tree definition by Buijs et al. [BDA12a], where the loop operator was defined as having three child nodes, which respectively are called the *do*, *redo*, and *exit* nodes. We denote the loop operator that follows the original definition of Buijs et al. [BDA12a] as $\overline{\circlearrowright}(M_1, M_2, M_3)$ with $M_1, M_2, M_3 \in \mathcal{M}$ subtrees that represent the three child nodes, with the semantics according to the following definition.

**Definition 8.16 (Buijs et al. [BDA12a] Process Tree Loop Operator).** If $M_1$, $M_2$, and $M_3$ are process trees, then $\overline{\circlearrowright}(M_1, M_2, M_3)$ is a process tree, with $\mathfrak{L}(\overline{\circlearrowright}(M_1, M_2, M_3)) = \{\langle \sigma_1^1 \cdot \sigma_3 \rangle, \langle \sigma_1^1 \cdot \sigma_2^1 \cdot \sigma_1^2 \cdot \sigma_3 \rangle, \langle \sigma_1^1 \cdot \sigma_2^1 \cdot \sigma_1^2 \cdot \sigma_2^2 \cdot \sigma_1^3 \cdot \sigma_3 \rangle, \langle \sigma_1^1 \cdot \sigma_2^1 \cdot \sigma_1^2 \cdot \sigma_2^2 \cdot \sigma_1^3 \cdot \dots \cdot \sigma_3 \rangle, \dots\}$ with $\sigma_1^i \in \mathfrak{L}(M_1)$, $\sigma_2^i \in \mathfrak{L}(M_2)$, and $\sigma_3 \in \mathfrak{L}(M_3)$. ◇

Intuitively, $\overline{\circlearrowright}$ allows for the execution of the *do* child (i.e., $M_1$), after which it has the possibility to execute the *redo* child (i.e., $M_2$) but there is no obligation to do so. If the redo-child is executed, the do-child has to be executed again, after which there is another possibility to execute the redo-child. Finally, after a sequence of executions of do and redo children, the operator has to finish by executing the exit-child (i.e., $M_3$).

In contrast, our definition of the loop operator has only one child node and can be representing in terms of the tradition loop-operator as $\circlearrowright(M) = \overline{\circlearrowright}(M, \tau, \tau)$. To illustrate why define the loop operator for LPMs as $\circlearrowright$ instead of $\overline{\circlearrowright}$ consider that for for all process trees M it holds that the following four process trees that include M are language-equivalent $M_1 = \overline{\circlearrowright}(M, \tau, \tau)$, $M_2 = \overline{\circlearrowright}(\tau, M, M)$, $M_3 = \rightarrow(M, \overline{\circlearrowright}(\tau, M, \tau))$, and $M_4 = \rightarrow(\overline{\circlearrowright}(\tau, M, \tau), M)$. While with operator $\circlearrowright$ we can construct these four models such that $\mathfrak{L}(M_1) = \mathfrak{L}(M_2) = \mathfrak{L}(M_3) = \mathfrak{L}(M_4)$, in contrast, there is only one way to represent this language using the $\circlearrowright$ operator: $\circlearrowright(M)$. From the fact that there are multiple identical ways to represent the same behavior using the $\overline{\circlearrowright}$ operator, it follows that using this operator would lead to redundancy in the search space of LPM mining by creating language equivalent LPMs in with expansion steps. If these language equivalent LPMs fulfill the quality criteria and themselves get expanded, then it results in even larger numbers of language equivalent LPMs in the next expansion step.

The number of possible expansion operations for a process tree $m \in \mathcal{M}$ when using $exp_0$ grows with the size of the alphabet of the event log $|Activities(L)|$ and the number of activity nodes in m. This is easy to see, as each type of expansion operation can be applied to each activity node in m, leading to $5 \times |Activities(L)|$ expansion operations per activity node. At every point in the expansion process, the number of activity nodes in the tree is equal to the number of expansion operations performed plus one, as each expansion operation adds one activity node to the process tree.

The approach of iteratively expanding, selecting, and expanding process trees described above is not, in principle, limited to the set of operators $\{\rightarrow, \wedge, \times, \circlearrowright\}$ and can easily be extended to other operators, such as an inclusive choice operator or a long-term dependency operator. Adding extra operators, however, comes with the price of increased computational complexity as it increases the number of ways to expand a process tree.

### 8.2.2 Desired Properties of LPM expansion

In the previous section we have introduced expansion function $exp_0$. We will now show that there are several properties that are to be desired of an expansion function $exp$ that our initial attempt $exp_0$ does not satisfy. These properties are related to the absence of redundancy and resulting duplication of work in the expansion procedure and they are independent of the selection step, i.e., we aim for an expansion step that satisfies these properties even when all candidate LPMs get selected in the selection step ($SCL_i = CL_i$).

First of all, we aim to be able to generate every possible process tree language over $Activities(L)$ through a series of expansion operations by starting from $CL_1$. This property is captured by Requirement 1.

**Requirement 1.** $\forall_{m_1 \in \mathcal{M}} \mathfrak{L}(m_1) \subseteq \mathcal{P}(Activities(L)^*) \implies \exists_{i \in \mathbb{N}^+} \exists_{m_2 \in CL_1} \exists_{m_3 \in exp^i(m_2)} \mathfrak{L}(m_1) = \mathfrak{L}(m_3)$ ◇

Secondly, we aim for function $exp$ such that when we start from the initial LPMs in $CL_1 = \{a | a \in Activities(L)\}$ we do not create multiple identical LPMs by expanding them. This property is captured by Requirement 2.

**Requirement 2.** $exp(m_1) \cap exp(m_2) = \emptyset$ for $m_1, m_2 \in CL_1$ when $m_1 \neq m_2$. ◇

When Requirement 2 does not hold, this implies that there exists an $m_3 \in \mathcal{M}$ such that $m_3 \in exp(m_1) \wedge m_3 \in exp(m_2)$ for some $m_1, m_2 \in \mathcal{M}$. This results in redundant computational work for the LPM mining algorithm since $m_3$ is now generated more than once and possibly even evaluated and further expanded more than once.

Furthermore, ideally, we would like to maintain the property of not generating identical LPMs over multiple successive expansion steps. This is captured by Requirement 3.

**Requirement 3.** $exp^i(m_1) \cap exp^i(m_2) = \emptyset$ for $i \in \mathbb{N}^+$ and $m_1, m_2 \in CL_1$ when $m_1 \neq m_2$. ◇

Where $exp^i(m)$ combines the definitions of function composition and lifting functions to sets, as introduced in Chapter 2, and thereby denotes the set of models that can be generated through i consecutive expansions starting from m. For example, when $exp^2(m) = \bigcup\{exp(m') | m' \in exp(m)\}$, and thus Requirement 3 implies $exp(exp(m_1)) \cap exp(exp(m_2)) \neq \emptyset$ for $m_1, m_2 \in CL_1$ when $m_1 \neq m_2$. In fact, we would even desire the more general version of this property that is formulated by Requirement 4.

**Requirement 4.** $exp^i(m_1) \cap exp^j(m_2)$ for $i, j \in \mathbb{N}^+$ and $m_1, m_2 \in CL_1$ when $m_1 \neq m_2$. ◇

| | |
|---|---|
| $\mathfrak{L}(\rightarrow(a, \rightarrow(b,c))) = \mathfrak{L}(\rightarrow(\rightarrow(a,b),c))$ | (Associativity of the sequence operator) |
| $\mathfrak{L}(\wedge(a, \wedge(b,c))) = \mathfrak{L}(\wedge(\wedge(a,b),c))$ | (Associativity of the concurrency operator) |
| $\mathfrak{L}(\times(a, \times(b,c))) = \mathfrak{L}(\times(\times(a,b),c))$ | (Associativity of the choice operator) |
| $\mathfrak{L}(\wedge(a,b)) = \mathfrak{L}(\wedge(b,a))$ | (Commutativity of the concurrency operator) |
| $\mathfrak{L}(\times(a,b)) = \mathfrak{L}(\times(b,a))$ | (Commutativity of the choice operator) |

**Table 8.1:** Language equivalence properties for process trees.

When we define *exp* in such a way that each possible expansion of an LPM m introduces a constant number c of new operator nodes to m, then it is easy to see that Requirement 4 holds when Requirement 3 holds. Given $m_1, m_2 \in CL_1$ and $m_1 \neq m_2$, if there would exist two LPMs $m_3 \in exp^i(m_1)$ and $m_4 \in exp^j(m_2)$ such that they are identical (i.e, $m_3 = m_4$), then they need to have the same number of operator nodes. In order for $m_3$ and $m_4$ to have the same number of operator nodes k, $i = j = kc$ needs to hold.

Next to preventing the generation of *identical* LPMs we ideally also aim for a definition of *exp* such that we prevent the generation of *language-equivalent* LPMs. This is captured in Requirement 5.

**Requirement 5.** $\forall_{m_3 \in exp^i(m_1), m_4 \in exp^j(m_2)} \mathfrak{L}(m_3) \neq \mathfrak{L}(m_4)$ for $i, j \in \mathbb{N}^+$ and $m_1, m_2 \in CL_1$ when $m_1 \neq m_2$. ◇

Note that Requirement 5 puts a stronger requirement on *exp* than Requirement 4, as $(m_3 = m_4) \implies (\mathfrak{L}(m_3) = \mathfrak{L}(m_4))$, but $(\mathfrak{L}(m_3) = \mathfrak{L}(m_4)) \not\Longrightarrow (m_3 = m_4)$. To see that $(\mathfrak{L}(m_3) = \mathfrak{L}(m_4)) \not\Longrightarrow (m_3 = m_4)$, consider $m_3 = \rightarrow(a, \rightarrow(b,c))$ and $m_4 = \rightarrow(\rightarrow(a,b),c)$, for which $m_3 \neq m_4$ but $\mathfrak{L}(m_3) = \mathfrak{L}(m_4) = \{\langle a,b,c \rangle\}$. Table 8.1 provides a set of equality relations for process trees, i.e., process tree structures that are language-equivalent.

The equality relations of Table 8.1 are closed under context, i.e., if $x = y$ and $f \in \{\rightarrow, \wedge, \times, \circlearrowleft\}$, then $f(S_1, \dots, S_{i-1}, x, S_{i+1}, \dots, S_{ar(f)}) = f(S_1, \dots, S_{i-1}, y, S_{i+1}, \dots, S_{ar(f)})$. This means that for all process trees $m \in \mathcal{M}$ that contains one of the process trees on the left-hand side of Table 8.1 as subtree, replacement of this subtree with the right-hand side process tree results in a language equivalent process tree.

### 8.2.3 Refining the Expansion Function Into $exp_1$

We now proceed by demonstrating a type of redundancy that occurs in $exp_0$ with expansions that use commutative operators, and another type of redundancy that occurs with expansions with non-commutative operators. For both types of redundancy we will show how they can be countered by restricting the set of expanded process trees that $exp_0$ generates, leading to a new expansion function that we call $exp_1$. Furthermore, we will show that all process tree languages remain reachable under $exp_1$, i.e., that Requirement 1 holds.

### *Redundancy in Commutative Operator Expansions*

Consider process trees $m_1, m_2 \in CL_1$ with $m_1 = a$ and $m_2 = b$ and a log with $Activities(L) = \{a, b, c\}$. For $m_3 = \times(a, b)$ and $m_4 = \times(b, a)$, we have $m_3 \in exp_0(m_1)$ and $m_4 \in exp_0(m_2)$, yet, $\mathfrak{L}(m_3) = \mathfrak{L}(m_4)$. Note that the same redundancy arises for $m'_3 = \wedge(a, b)$ and $m'_4 = \wedge(b, a)$ (with $m'_3 \in exp_0(m_1)$ and $m'_4 \in exp_0(m_2)$). Note that $exp_0$ did already restrict expansions with the $\wedge$ and $\times$ operators to only *right-side expansions*, i.e., expansions where the new activity node $b \in Activities(L)$ was added as the second argument of the operator. We have now shown that this is not sufficient to prevent redundancy that arises with expansions that use commutative operators, since both variants of the language equivalences $\wedge(a, b) = \wedge(b, a)$ and $\times(a, b) = \times(b, a)$ can still be generated by $exp_0$ by expanding different activity nodes.

To prevent this type of redundancy with commutative operators in $exp_0$ we define an arbitrary but consistent ordering over the set of activities $\Sigma$, i.e., we define a function $index : \Sigma \to \{1, \dots, |\Sigma|\}$ such that $\forall_{a,b \in Activities(L)} a \neq b \implies index(a) \leq index(b)$. One possible example of such an arbitrary but consistent ordering function *index* would be one that follows the alphabetical order according to the activity names in $Activities(L)$. Using *index*, we now restrict the expansion of $m_1 = a$ to $\wedge(a, b) \in exp_0(m_1)$ and $\times(a, b) \in exp_0(m_1)$ by making the expansions conditional on $index(a) \leq index(b)$, i.e., the right-side expansion with commutative operators is only applied when the new activity occurs later in the alphabet than the replaced activity.

Note that this restriction does not pose any limitations on the class of process tree languages that can be mined with LPM mining. Let the expansion function that consists of $exp_0$ with the aforementioned constrained by $exp_{0'}$.

**Property 1.** Requirement 1 holds under expansion function $exp_{0'}$. ◇

$\oplus$ denotes a commutative operator. For any process tree $m_1$ that contains a subtree $s = \oplus(b, a)$ and $m_2$ that is identical to $m_1$ except for subtree s, which is $s = b$ for $m_2$, $m_1 \notin exp_{0'}(m_2)$ when $index(b) > index(a)$, but there always is $m_3$ such that $\mathfrak{L}(m_3) = \mathfrak{L}(m_1)$ that can be reached through $exp_{0'}$. Let $m_4$ be identical to $m_1$ except that $s = b$. Now, for $m_3$, $s = \oplus(a, b)$ and $m_3 \in exp(m_4)$. This shows that $\forall_{m \in \mathcal{M}} \{\mathfrak{L}(m') | m' \in exp_0(m)\} = \{\mathfrak{L}(m') | m' \in exp_{0'}(m)\}$. It now remains to be shown that this language equivalence holds for multiple consecutive extensions, i.e., that $\forall_{m \in \mathcal{M}} \forall_{i \in \mathbb{N}^+} \{\mathfrak{L}(m') | m' \in exp_0^i(m)\} = \{\mathfrak{L}(m') | m' \in exp_{0'}^i(m)\}$. In order for this to hold, the expansions under $exp_1$ of $m_3$ with subtree $s = \oplus(a, b)$ would need to allow for a language equivalent model for each extension that the prevented process model $m_1$ with subtree $s = \oplus(b, a)$ would have allowed for under $exp_0$. Note that both activity nodes a and b of $m_3$ can both be expanded and $\oplus$ is commutative, therefore, Requirement 1 holds when the restriction of $exp_{0'}$.

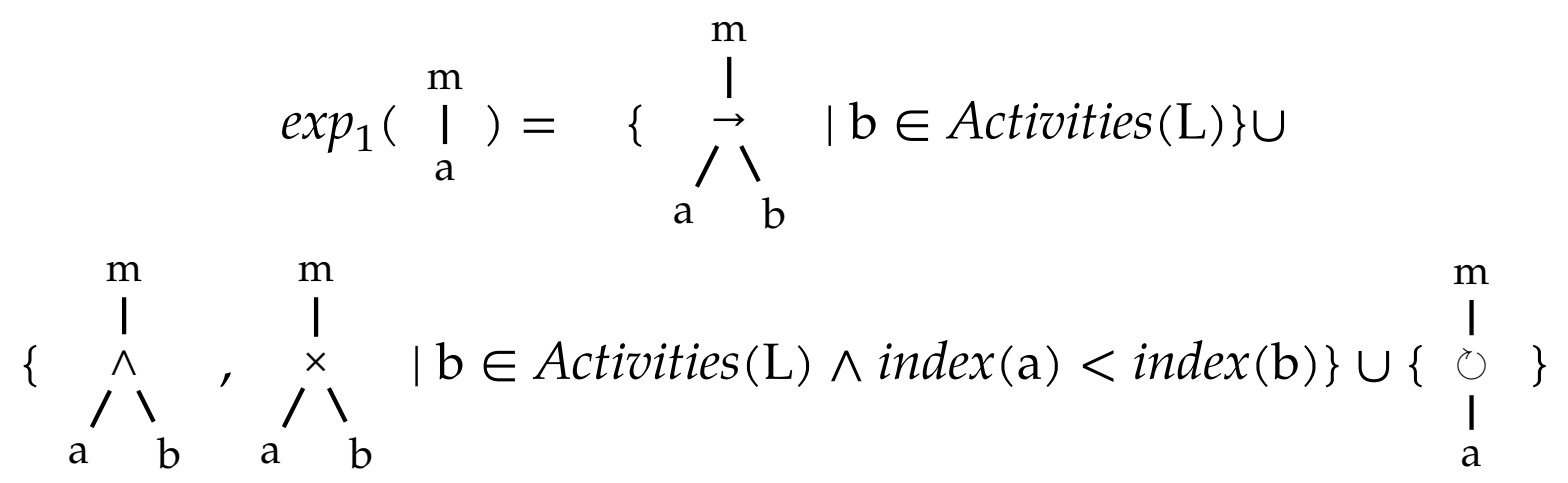

**Figure 8.6:** The specification of $exp_1$, which specifies the set of process tree expansion operations of leaf node a of process tree m.

### *Redundancy in Non-Commutative Expansions*

For process trees $m_1, m_2 \in CL_1$ with $m_1 = a$ and $m_2 = b$, $m_3 = \rightarrow (a, b)$ and $m_4 = \rightarrow (b, a)$ can be reached through expansion under $exp_0$ as $m_3 \in exp_0(m_1)$ and $m_4 \in exp_0(m_2)$. This shows that, while we initially thought that the non-commutativity of $\rightarrow$ required us to expand $m_1$ into both $m_3$ and $m_4$, this turns out to lead to redundancy as $m_3$ and $m_4$ can also be reached by expanding $m_2$. Therefore, it suffices to expand with non-commutative operators only using *right-side expansions*, i.e., where the added new activity node is added as the second child node of the new operator node.

Let the expansion function that consists of $exp_0$ with the aforementioned constraint be denoted by $exp_{0''}$. Using this expansion function, $m_3 \in exp_{0''}(m_1)$ and $m_4 \in exp_{0''}(m_2)$ while $m_3 \notin exp_{0''}(m_2)$ and $m_4 \notin exp_{0''}(m_1)$. This restriction on the search space does not pose a limitation on the class of process tree languages that can be mined with LPM mining.

**Property 2.** Requirement 1 holds under expansion function $exp_{0''}$. $\diamond$

To see that this is the case, consider that for a given $m \in CL_1$ and $i \in \mathbb{N}^+$ we have $exp^i_{0''}(m) \subseteq exp^i_0(m)$. Therefore, when we consider the whole set initial LPMs, $\bigcup_{\{exp^i_{0''}(m) | m \in CL_1\}} = \bigcup_{\{exp^i_0(m) | m \in CL_1\}}$. Since $exp_0$ and $exp_{0''}$ generate the same process trees they also generate the same set of languages.

Figure 8.6 defines $exp_1$, which applies both the search space reduction method of $exp_{0'}$ and of $exp_{0''}$. It is easy to see that $exp_1$ is computationally more efficient than $exp_0$: the number of expansion possibilities per leaf node decreased from $5 \times |Activities|$ to 4 times a factor that is upper bounded by but not always equal to $|Activities|$.

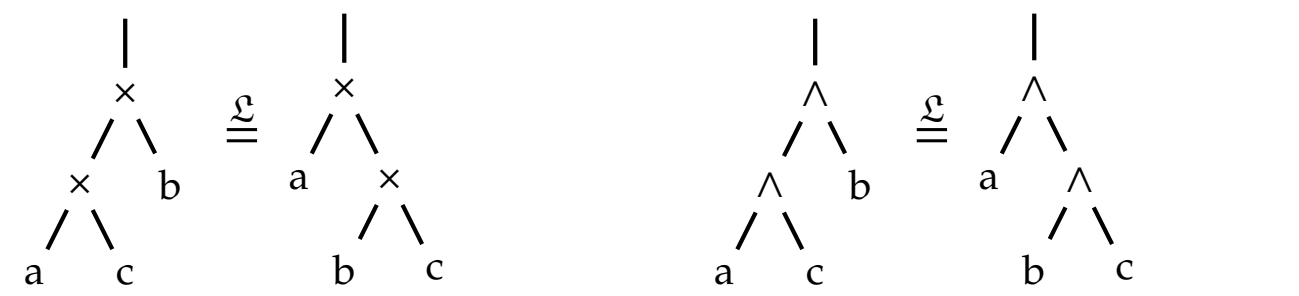

**(a)** $m_3$ and $m_4$, two language equivalent process trees.

**(b)** $m_5$ and $m_6$, two language equivalent process trees.

**Figure 8.7:** Two examples of language equivalent process trees in the search space under $exp_1$.

## 8.2.4 Further Refinements Into $exp_2$

Several further refinements of $exp_1$ are possible when we allow search space restrictions to be based not solely on the to-be-expanded activity node $a \in Activities(L)$ but additionally also on its parent node $pa(a)$.

### *Consecutive Commutative Expansions*

Figure 8.7 shows two examples of two language equivalent LPMs that are generated with $exp_1$. For process models $m_3$ and $m_4$ in Figure 8.7a with $\mathfrak{L}(m_3) = \mathfrak{L}(m_4)$, it holds that for $m_1 = \times(a, b)$, $m_3 \in exp_1(m_1)$ (by replacing activity node a by $\times(a, c)$) and $m_4 \in exp_1(m_1)$ (by replacing activity node b by $\times(b, c)$). Similarly, for process models $m_5$ and $m_6$ in Figure 8.7a with $\mathfrak{L}(m_5) = \mathfrak{L}(m_6)$, it holds that for $m_2 = \wedge(a, b)$, $m_5 \in exp_1(m_1)$ (by replacing activity node a by $\wedge(a, c)$) and $m_6 \in exp_1(m_1)$ (by replacing activity node b by $\wedge(b, c)$). This shows that it is necessary to constrain the set of allowed expansion steps in a way that is conditional on the structure of the LPM that is being expanded. The number of expansions can be reduced by performing an expansion with a commutative operator $\oplus_1 \in \{\times, \wedge\}$ to an existing subtree $m = \oplus_2(a, b)$ only to the second child of m but not to the first. With $exp_{1'}$ denoting the expansion function that uses $exp_1$ with this additional search space restriction, we have $\forall_{c \in Activities(L)} \oplus_2 (a, \oplus_1(b, c)) \in exp_{1'}(m) \wedge \oplus_2(\oplus_1(a, b), b, c) \notin exp_{1'}(m)$. Expansion function $exp_{1'}$ complies with Requirement 1.

**Property 3.** Requirement 1 holds under expansion function $exp_{1'}$. ◇

$\forall_{m_1 \in \mathcal{M}} \forall_{m_2 \in exp_1(m_1)} \exists_{m_3 \in exp_{1'}(m_1)} \mathfrak{L}(m_2) = \mathfrak{L}(m_3)$, as for every expansion that $exp_{1'}$ prevents there is a language-equivalent extension still possible under $exp_{1'}$ where instead of extending a (in the prevented extension), instead, the second child of $pa(a)$ is extended. Therefore, Requirement 1 holds for $exp_{1'}$.

### 8.2.5 Consecutive Loop Expansions

$\mathfrak{L}(\circlearrowright(m)) = \mathfrak{L}(\circlearrowright(\circlearrowright(m)))$ for any subtree $m \in \mathcal{M}$. Therefore, we restrict expansions of the loop operator to activity nodes a where $pa(a) \neq \circlearrowright$. With $exp_{1''}$ denoting the expansion function that uses $exp_1$ with this additional search space restriction, it is trivial to see that $exp_{1''}$ satisfies Requirement 1.

**Property 4.** Requirement 1 holds under expansion function $exp_{1''}$. $\diamond$

$\forall_{m \in \mathcal{M}} \mathfrak{L}(\circlearrowright(m)) = \mathfrak{L}(\circlearrowright(\circlearrowright(m)))$, therefore Requirement 1 holds.

$exp_2$ is the expansion function that restricts $exp_1$ with both the search space restrictions that we introduced in $exp_{1'}$ and in $exp_{1''}$. In the remainder of this thesis we assume that the expansion step of the LPM mining algorithm is performed using $exp_2$.

## 8.3 Local Process Model Selection & Ranking

The quality dimensions and measures that we defined in Section 8.1 are used to select LPMs from the LPMs that are generated through the recursive expansion of process trees. Additionally, the quality measures can be used to rank the selected LPMs.

### 8.3.1 State Space Pruning

Often one is not interested in LPMs with low support, confidence, or determinism. We can set a minimum threshold for the quality measures that we introduced in Section 8.1, thereby retrieving only the patterns that meet the specified interest of the process analyst. Additionally, setting a minimum threshold for these quality criteria allows us to prune away those parts of the search space where we know that expansions of a candidate LPM cannot meet the criteria, resulting in a speedup of the proposed recursive process tree exploration procedure.

#### *Support Pruning*

Any expansion $m' \in exp(m)$ of process tree $m \in \mathcal{M}$ where a leaf node $a \in m$ is replaced by subtree $\rightarrow(a, b)$, $\wedge(a, b)$, or $\circlearrowright(a)$ for any $b \in \Sigma$ is guaranteed to be at most as frequent, as m itself. The intuition behind this is these types of expansions put additional requirements on the behavior. As a result, for some instances of model m in a trace $\sigma$ given by $\xi(\Gamma_m^{(}\sigma))$ it might not be able to extend them into an instance of some expanded model $m'$. Therefore, when m does not meet support threshold $min_{sup}$, its expansions of any activity node a into $\rightarrow(a, b)$, $\wedge(a, b)$, and $\circlearrowright(a)$ can be pruned from the search space as $m'$ also cannot meet $min_{sup}$. Note that the expansion $m' \in exp(m)$ of process tree $m \in \mathcal{M}$ that results by replacing leaf node $a \in m$ by subtree $\times(a, b)$ in contrasts to the other types of expansions can lead to a model $m'$ with higher support than $m$.

While for $m_1 \in \mathcal{M}$, and $m_2 \in exp(m_1)$ the support of $m_2$ is smaller than or equal to the support of $m_1$ on some log $L$ when $m_2$ was created by expanding $m_1$ with $\rightarrow$, $\wedge$, or $\circlearrowleft$, there can still exist a model $m_3 \in exp(m_2)$ using $\times$ such that its support exceeds that of $m_1$. As example, consider $m_1 = \rightarrow(\circlearrowleft(a), b)$, $m_2 = \rightarrow(\circlearrowleft(\rightarrow(a, c)), b)$, $m_3 = \rightarrow(\circlearrowleft(\rightarrow(a, c)), \times(b, d))$ and log $L = [\langle a, c, a, c, d, a, c, a, c, b, a, b, a, c, d\rangle]$. On $m_1$ we have the LPM instances $\langle \underline{\mathbf{a}, c, \mathbf{a}, c, d, \mathbf{a}, c, \mathbf{a}, c, \mathbf{b}}, \underline{\mathbf{a}, \mathbf{b}}, a, c, d\rangle$ resulting in a support of 2. On $m_2$ we have the LPM instances $\langle \underline{\mathbf{a}, \mathbf{c}, \mathbf{a}, \mathbf{c}, d, \mathbf{a}, \mathbf{c}, \mathbf{a}, \mathbf{c}, \mathbf{b}}, a, b, a, c, d\rangle$ resulting in a support of 1, decreasing from 2 since the occurrence $\langle a, b\rangle$ does not anymore satisfy the LPM language. On $m_3$ we have the LPM instances $\langle \underline{\mathbf{a}, \mathbf{c}, \mathbf{a}, \mathbf{c}, \mathbf{d}}, \underline{\mathbf{a}, \mathbf{c}, \mathbf{a}, \mathbf{c}, \mathbf{b}}, a, b, \underline{\mathbf{a}, \mathbf{c}, \mathbf{d}}\rangle$ with support 3. LPM $m_3$ has a higher support than $m_2$ since the expansion of node $b$ into $\times(b, d)$ allowed instances to end with a $d$, resulting in the situation where the final three events of the trace satisfy the behavior of $m_3$ and become an instance, and the first pattern instance of $m_2$ is split into two instances because the $d$ halfway the instance of $m_2$ allowed the pattern to end in $m_3$. Note that, as long as no expansion with the $\times$-operator is used, monotonicity in support holds for any sequence of consecutive expansions with the operators other than the choice operator and support can only decrease. Therefore, if $m$ does not meet the support threshold, none of its expansions other than those with the $\times$-operator meet the support threshold.

#### *Determinism Pruning*

Process tree $m \in \mathcal{M}$ is guaranteed to be at least as deterministic as its expansion $m' \in exp(m)$ if $m'$ was expanded by replacing activity node $a \in m$ with subtree $\times(a, b)$ or $\wedge(a, b)$ for any $b \in \Sigma$. Therefore, when $m$ does not meet determinism threshold $min_{det}$, its expansions of any activity node $a$ into $\times(a, b)$, and $\wedge(a, b)$, we also know that $m'$ cannot meet the determinism threshold and it be pruned from the search space without generating and evaluating it. Combining this property with the support threshold, we know that any LPM $m$ that does not meet $min_{sup}$ and does not meet $min_{det}$ cannot be further expanded: expansions with $\times$ would bring the determinism further down while any other expansion would bring the support further down.

### 8.3.2 LPM Pattern Ranking

The LPM mining procedure can return a large number of LPM patterns, depending on the thresholds that were used for the LPM quality measures during mining. Therefore, we rank the LPMs according to the quality measures before presenting them to the process analyst. Generally, one is interested in multiple quality criteria at the same time. An LPM with high support but low determinism (e.g., a small flower pattern) does not generate much process insight, while at the same time an LPM with high determinism but with low support does not represent any frequent behavior of the log. It is also useful to rank patterns according to a weighted average over the quality criteria. The appropriate weighting of the quality dimensions depends on the business questions and the situation at hand, and should therefore be set by the analyst himself.

Before calculating the weighted average over the quality measures that we introduced in Section 8.1 it is important that all measures use a similar scale. Observe that confidence, language fit, determinism, and coverage are measures on a $[0, 1]$-interval. However, support is the odd one out and has codomain $\mathbb{N}$. Ranking LPM patterns requires a *normalized support* measure that transforms support into a $[0, 1]$-interval. In the pattern mining field, a *normalized support* measure can be trivially obtained by dividing support by the number of sequences/traces when at most one pattern instance can be counted per trace. However, the support measure that we use for LPM mining is a *repetitive support* measure, i.e., multiple instances of the pattern can be counted in the same sequence as long as those instances do not overlap. Therefore, we resort to a *squashing function*, e.g., a functions that "squashes" values from $\mathbb{N}$ into $[0, 1]$. Many squashing functions exist, with $f(x) = \frac{x}{x+c}$ for some fixed constant $c \in \mathbb{N}$ being a well-known example. Other examples of squashing functions include the logistic function and the hyperbolic tangent, which are both known from their use as activation functions in neural networks. Any squashing function can be used to obtain a *normalized support* measure from our original definition of support, after which the mined LPMs can be ranked according to a weighted average over quality measures that are all in $[0, 1]$-interval.

## 8.4 Alignment-Based Evaluation of Local Process Models

We now describe a way to calculate function $\Gamma_{LPM} : \mathcal{B}(\Sigma^*) \to \mathbb{N}$ (as described in Section 8.1) to identify the minimal multiset of maximal non-overlapping multiset of instances I for a local process model *LPM* and event log L. The calculation approach to $\Gamma_{LPM}$ that we will introduce is based on the concept of alignments that we introduced in Chapter 2. To be able to evaluate an LPM using alignments, we first transform the process tree to its Petri net representation. Recall that alignments aim to match a trace from an event log to a run through a Petri net. Observe that, conceptually, the task of detecting occurrences of an LPM in an event log is close to the concept of alignments. However, there are two fundamental differences between detecting LPM occurrences in a log and calculating alignments.

First, alignments have the possibility to execute *moves on model*, and as a result, the run through the model that results from the alignment can contain transition firings that have no corresponding event in the log. When we are detecting occurrences of an LPM, we want a detected occurrence to be a *complete execution* of the LPM and not an *approximate execution*. To avoid counting approximate executions of an LPM as LPM occurrences, we modify the standard alignments approach by removing model moves on non-silent transitions from the alignment search space. Note that removing model moves on silent transitions from the search space would have the effect that LPMs with a silent transition on all possible paths from initial to final marking cannot have occurrences (e.g., Figure 8.8a, where $t_4$ is in all such paths). The reason for this is that silent transitions by definition have no corresponding event in the event log and therefore can only be fired by a model move.

The second difference between the alignment calculation and counting occurrences of an LPM is that a single trace from an event log can contain more than one occurrence of an LPM. To be able to detect a second, third, or later occurrence of an LPM in a trace, we modify the Petri net representation of the LPM such that we connect each final marking to the initial marking through a silent transition, allowing the alignment to loop through the whole LPM multiple time and detect an occurrence of the LPM in each iteration of the loop. Figure 8.8a shows an example LPM and Figure 8.8b shows the corresponding Petri net after transformation. We transform the LPM in the following way.

**Definition 8.17 (Petri net transformation for LPM instance counting).** Given Petri net $LPM(N, M_0, MF)$ with $N = (P, T, F, \Sigma_M, \ell)$ the transformed Petri net $LPM_{BL}(N_{BL}, M_0, \{M_0\})$ with $N_{BL} = (P, T_{BL}, F_{BL}, \Sigma_M, \ell_{BL})$ for counting LPM instances is defined as follows:

- $T_{BL} = T \cup \{t_{bl_M} \mid M \in MF\}$,
- $F_{BL} = F \cup \{(p, t_{bl_M}) \mid M \in MF \wedge p \in M\} \cup \{t_{bl_M} \mid M \in MF \wedge p \in M_0\}$,
- $\ell_{BL} \in T_{BL} \to \Sigma_M \cup \{\tau\}$ with:
  $\ell_{BL} = \begin{cases} \ell(T), & \text{if } t \in T, \\ \tau, & \text{otherwise.} \end{cases}$ ◇

$LPM_{BL}$ contains a set of added silent transitions, $\{t_{bl_M} | M \in MF\}$, consisting of one silent transition for each final marking $M \in MF$. Backloop : $MF \to T_{bl}$ is a bijective mapping from a final marking $M \in MF$ to a silent transition $t_{bl_M}$. A silent transition $t_{bl_M}$ has all places in final marking M as input and place $M_0$ as output. The number of executions of backloop transitions $\{t_{bl_M} | M \in MF\}$ in the alignments of L on *LPM* indicates the number of executions of traces of *LPM* in L. Note that alignments require the model to be in a marking $M \in MF$ at the end of the alignment. This is impossible to obtain when pattern *LPM* is absent in log L. Therefore, we set the final marking to $\{M_0\}$, allowing the alignments to make a complete sequence of *moves on log*, resulting in zero occurrences of the LPM in a trace.

Note that the Petri net transformation of Definition 8.17 in the general case is not safe. To see this, consider for example the setting where *MF* contains a final marking that is a subset of another final marking that is also in *MF*. In such a setting it is possible that firing the backloop transition does not remove all of the tokens that

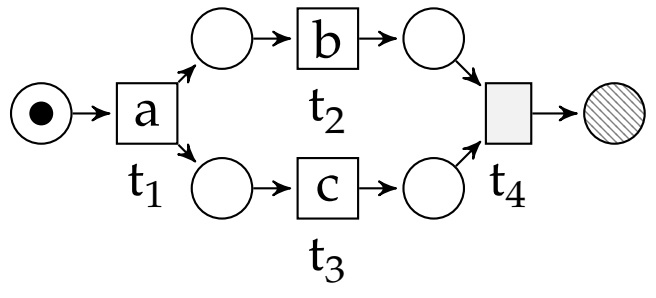

**(a)**

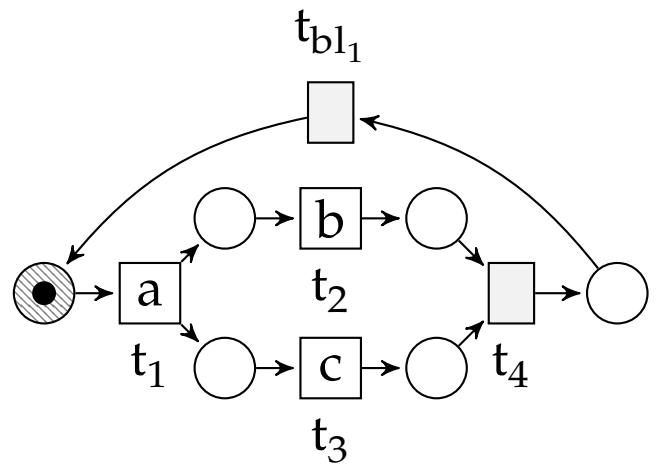

**(b)**

| Log | a | a | c | b | ≫ | ≫ | a | a | c | b | ≫ | ≫ | b | c |
|---|---|---|---|---|---|---|---|---|---|---|---|---|---|---|
| Model | a<br>$t_1$ | ≫ | c<br>$t_3$ | b<br>$t_2$ | τ<br>$t_4$ | τ<br>$t_{bl_1}$ | a<br>$t_1$ | ≫ | c<br>$t_3$ | b<br>$t_2$ | τ<br>$t_4$ | τ<br>$t_{bl_1}$ | ≫ | ≫ |

**(c)**

**Figure 8.8:** *(a)* An example local process model, $LPM = \rightarrow(a, \wedge(b, c))$, transformed to Petri net representation. *(b)* Process model $LPM_{BL}$, constructed from *LPM* by adding a silent connection from final to initial marking, and the final marking set to the initial marking. *(c)* Alignment of the trace $t_1$ on $LPM_{BL}$ when disallowing model moves on non-silent transitions.

are present in the Petri net[24]. However, in the specific case of Petri nets that are the result of a transformation of a process tree into a Petri net, all final markings have size one. Therefore, the Petri net transformation of Definition 8.17 works when we define LPM expansion operations in terms of process trees, as we did in Section 8.2.

## 8.4.1 An Example

Figure 8.8c illustrates through an example how alignments on the modification of the Petri net representation of an LPM can be used to detect the occurrences of that LPM. The example shows the alignment of the non-fitting trace $\sigma = \langle a, a, c, b, a, a, c, b, b, c \rangle$ on the model of Figure 8.8b that is obtained by modifying the Petri net representation of the LPM as shown in Figure 8.8a. The top row of the alignments represents the behavior of the log, while the middle row and the bottom row represent the behavior of the model. ≫ indicates *no move*, with a ≫ in the top row indicating a *move on model* and in the middle row indicating a *move on log*. The model is able to mimic the first event of the trace by executing $t_1$ with label a, but is not able to mimic the second a in the log, resulting in a *move on log*. The c and b in the log can be mimicked (by $t_3$ and $t_2$ respectively). Next event a in the log can only be mimicked by the model by first firing $t_{bl_1}$, resulting in a *move on model*, represented by the ≫ in the log. Afterwards, a can be mimicked and another *move*

[24]When firing the backloop transition of the smaller final marking

*on log* is needed for the second a. Events c and b can again be mimicked, after which a *move on log* is again needed as the log cannot mimic $t_{bl_1}$. The alignment could now have executed a *move on model* on a if we would not have prohibited *moves on models* on non-silent transition. This would have enabled synchronous moves on both b and c, falsely giving the impression that the LPM would have a third occurrence in the trace. Because we prohibited the *model move* on a, the only option is to decide a *move on log* on b and c, thereby not counting the incomplete occurrence of the pattern.

LPM *LPM* is evaluated on event log $L = [\langle a, a, c, b, a, a, c, b, b, c\rangle]$ by projecting L on the set of labels $\Sigma_M$ of *LPM*, i.e., $L' = L \upharpoonright_{\Sigma_M}$. The minimal multiset of maximal non-overlapping instances $\Gamma_{LPM}(L)$ can be derived from this alignment, where each synchronous move is part of an instance and each τ-move on transition $t_{bl_i} \in \{t_{bl_m} | M \in MF\}$ indicates the start of a new instance. The alignment in Figure 8.8c shows that $\Gamma_{LPM}(L) = [(\sigma, \langle 1, 3, 4\rangle), (\sigma, \langle 5, 7, 8\rangle)]$, which we can represent as $\langle \underline{\mathbf{a}, a, \mathbf{c}, \mathbf{b}}, \underline{\mathbf{a}, a, \mathbf{c}, \mathbf{b}}, b, c\rangle$ to highlight the position of the instances. Furthermore, $\xi(\Gamma_{LPM}(L)) = [\langle a, c, b\rangle^2]$.

## 8.4.2 Properties of alignment-based instance identification

**Theorem 8.1.** An optimal alignment of L on $LPM_{BL}$ yields a multiset of maximal non-overlapping instances of *LPM* on L. ◇

Since we have prohibited moves on model on labeled transitions, $\sum_{(\sigma,\lambda)\in\Gamma_{LPM}(L)} |\lambda|$ equals the total number of synchronous moves in the alignment. The optimal alignment is guaranteed to minimize the number of moves on model and number of moves on log (see [ADA11] for proof). Each event in the log is either represented by a synchronous move or a log move. Therefore, the optimal alignment maximizes the number of synchronous moves (by minimizing the number of log moves) and hence maximizes $\sum_{(\sigma,\lambda)\in\Gamma_{LPM}(L)} |\lambda|$. Therefore, the set of non-overlapping instances of *LPM* on L that is returned by alignments is a set of maximal non-overlapping instances of *LPM* on L.

**Theorem 8.2.** An optimal alignment of L on $LPM_{BL}$ yields a minimal multiset of maximal non-overlapping instances of *LPM* on L. ◇

Identifying an additional instance of an LPM beyond its first occurrence in a trace requires a move on model on a silent transition $t_{bl_M}$ for some $m \in MF$ of *LPM*. Since $t_{bl_M}$ is a silent transition, this model move has cost $\epsilon$. If it is possible to obtain the same number of synchronous moves without firing $t_{bl_M}$, this has costs that are lower by $\epsilon$. Therefore, alignments return the minimal multiset of maximal non-overlapping instances.

| Log | a | a | c | b | ≫ | ≫ | a | a | c | b | ≫ | ≫ | b | c |
|---|---|---|---|---|---|---|---|---|---|---|---|---|---|---|
| Model<br>Label | $t_1$<br>a | ≫ | $t_3$<br>c | $t_2$<br>b | $t_4$<br>$\tau$ | $t_{bl_1}$<br>$\tau$ | $t_1$<br>a | ≫ | $t_3$<br>c | $t_2$<br>b | $t_4$<br>$\tau$ | $t_{bl_1}$<br>$\tau$ | ≫ | ≫ |
| # enabled trans. | 1 | - | 2 | 1 | 1 | 1 | 1 | - | 2 | 1 | 1 | 1 | - | - |

**Table 8.2:** The number of transitions enabled at each point in the alignment

### 8.4.3 Calculating Determinism on Petri nets

To calculate the determinism based on the alignment of the log on the Petri net representation of an LPM, each transition firing in a Petri net that results from a synchronous move or from a model move on a silent transition corresponds to a change in the marking of the net. From each of those transitions firings, the marking of the net prior to the transition firing had a certain number of transitions enabled, shown in the bottom row of Table 8.2. For example, when $t_3$ fired in the alignment, the Petri net was in a marking where both $t_2$ and $t_3$ were enabled. The determinism of the net is the inverse of the average number of enabled transitions during replay. In the example in Table 8.2, the alignment consists of 10 transition firings, and therefore, $\mathit{determinism}(LPM, \mathrm{L}) = \frac{10}{1+2+1+1+1+1+2+1+1+1} = \frac{10}{12} \approx 0.83$.

## 8.5 Related Work

In this section, we discuss two areas of related work. First, we discuss existing work in process discovery and position Local Process Model (LPM) discovery in the process discovery landscape. Secondly, we discuss related work from the pattern mining field.

### 8.5.1 Process Discovery

Process discovery techniques can be classified in several dimensions. Some process discovery techniques discover process models with *formal semantics*, where the behavior that is allowed by the model is formally defined, while others discover models that lack such formal semantics. Orthogonally, process discovery algorithms can be classified in *end-to-end (global)* techniques that produce models that describe the logged process executions fully from start to end, and *pattern-based (local)* techniques that produce models that describe the behavior of the log only partially. A plethora of process discovery techniques have been developed over the years, an extensive overview of which is given in [Aal16]. Table 8.3 provides a selection of process discovery methods and their classification on the *formal semantics/no formal semantics* and the *global/local* dimensions. The table shows that LPM discovery is the only technique available that provides process models that are local and also have formal semantics.

**Table 8.3:** A classification of a selection of process discovery methods.

| Algorithm | Formal semantics | Global/Local | Algorithm | Formal semantics | Global/Local |
|---|---|---|---|---|---|
| Declare Miner [MMA11] | ✓ | Global | DPIL Miner [Sch+16a] | ✓ | Global |
| MINERful [DM15] | ✓ | Global | TBDeclare Miner [DMM16] | ✓ | Global |
| Language-based regions [Ber+07] | ✓ | Global | Inductive Miner [LFA13b] | ✓ | Global |
| Split Miner [Aug+17] | ✓ | Global | ILP miner [Wer+09a; Wer+09b; ZDA15] | ✓ | Global |
| **LPM Discovery** (this chapter) | ✓ | Local | Fuzzy Miner [GA07] | ✗ | Global |
| Episode Miner [LA14] | ✗ | Local | Instance graphs[DGP16] | ✗ | Local |
| Behavioral PM [DGP16] | ✗ | Local | SUBDUE patterns [Dia+13] | ✗ | Local |

The majority of the existing process discovery techniques focus on mining a global model with formal semantics. However, these techniques have the drawback that they are often not able to capture the frequent types of behavior in unstructured event data, as shown in Figure 8.2. The Fuzzy Miner [GA07] is a process discovery technique that aims to address this problem by dropping the requirement to obtain a model with formal semantics. The result of the Fuzzy Miner is a graph with activities as vertices and edges between activities that are somehow related. However, the language of this graph is not formally defined, and therefore, it is unclear what behavior actually does and what behavior does not occur in the event log.

### *Trace Clustering*

A specific area within the area of process mining that is related to LPM mining is the area of trace clustering [BA09; Fol+11; Gre+06a; HVA14; SGA08]. Trace clustering techniques are related to LPM discovery in the sense that both aim to enable extraction of process insight from event logs where the process is too unstructured for existing process discovery techniques to find a structured process model. Trace clustering, first proposed by Greco et al. [Gre+06a], aims to solve this by clustering similar traces together to prevent mixing different usage scenarios into one single unstructured process model. Process discovery is then applied to each cluster of traces separately, and thereby one process model is obtained per cluster of similar traces.

Trace clustering techniques work well when the original event log does not originate from one single process, but in fact originates from multiple processes. However, not in all types of complex and flexible event data there is a cluster tendency in the data. An example for such non-clusterable event data can be found in the log shown in Figure 8.2a, where no clustering over the traces would enable the discovery of the frequent pattern shown in Figure 8.2b. The traces in the log have large parts of randomness *within* the traces, while trace clustering helps for cases where there is a large degree of variety *between* traces.

*Declarative Process Discovery*

A set of process discovery techniques that are especially close to LPM discovery are *declarative process discovery* techniques, which discover so-called *declarative process models*. Declarative process models are process models that specify the behavior of the process through a set of rules that executions of the process should adhere to, using a declarative constraint template language such as Declare [APS09; PA06; PSA07], DPIL [ZSJ14], or SCIFF [Alb+08]. This contrasts imperative, or procedural, process models that explicitly model the behavior that is allowed. An example of such a template from Declare is *Response(*a, b*)*, indicating that if activity a occurs, then activity b follows eventually. Another example of a Declare template is *CoExistence(*a, b*)*, indicating that if a occurs, then b occurs within the same trace.

Table 8.4 shows the set of *constraint templates* that is supported by Declare, and describes their semantics expressed as regular expressions. The variable 'a' of a binary Declare constraint template is called the *activation* variable while variable 'b' is called the *target* variable. A constraint template can be instantiated into a concrete constraint by setting the activation and target variables to concrete activities from the event log. *Negation constraints* extend the set of constraint templates of Table 8.4 with the negated version of the relations. The Declare Miner extracts a set of rules in the form of Declare constraints from the event log based on a template matching.

Several extensions of the Declare miner exists, which we will discuss in the following paragraphs. These extensions generally extend the set of *constraint templates* in different ways, including some extensions that go beyond binary relations. All these approaches have in common that they extract the constraints (i.e., instantiations of the constraint templates) from the log that adhere to a given *support* threshold. This means that the set of control-flow structures that the analyst is interested in should be specified *apriori*, before starting the mining. Here lies a fundamental difference with LPM mining, where unary relations are expanded into binary relations, then into tertiary relations, and then iteratively into more complex n-ary relations, where the expansion of a continues as long as the minimum support threshold is met. As an effect, the search space for LPMs is pruned, i.e., a certain n-ary relation will only be in the search space if a weaker (n-1)-ary relation meets the support threshold.

Figure 8.9 illustrates a second difference between declarative approaches and LPMs. Consider an event log $L_1$, such that projected on the activities of the Petri net of Figure 8.9a, $L_1 = [\langle a,b,c\rangle^{40}, \langle a,b,d\rangle^{50}, \langle b,c\rangle^{10}, \langle b,d\rangle^{20}, \langle b\rangle^{10}, \langle c\rangle^{25}, \langle d\rangle^{31}]$, where $\langle a,b,c\rangle^{40}$ indicates this trace occurs 40 times. Furthermore, consider event log $L_2$ such that projected on the activities of the Petri net of Figure 8.9a, $L_2 = [\langle a,b\rangle^{60}, \langle a,b,c\rangle^{5}, \langle a,b,d\rangle^{5}, \langle b,c\rangle^{35}, \langle b,d\rangle^{95}, \langle c\rangle^{40}, \langle d\rangle^{100}]$. Using a minimum support of 0.7, the mining of target-branched Declare constraints from both $L_1$ and $L_2$ results in the Declare model shown in Figure 8.9d. The Declare model consists of target-branched Declare rules *Response*(a,b), as for both logs all a events are at some point followed by a b. Furthermore, the Declare model consists of *Response*(a,{b,c}), as in both logs at least 70% of the instances of b are followed by

**Table 8.4:** The semantics of Declare constraints as regular expressions.

<table>
<tr><th>Constraint</th><th>Semantics (as regular expression)</th></tr>
<tr><td>Existence(m,a)</td><td><code>[^a]*(a[^a]*)m,[^a]*</code></td></tr>
<tr><td>Absence(n,a)</td><td><code>[^a]*(a[^a]*)m,[^a]*</code></td></tr>
<tr><td>Init(a)</td><td><code>a.*</code></td></tr>
<tr><td>RespondedExistence(a,b)</td><td><code>[^a]*((a.*b)|(b.*a))*[^a]*</code></td></tr>
<tr><td>Response(a,b)</td><td><code>[^a]*(a.*b)*[^a]*</code></td></tr>
<tr><td>AlternateResponse(a,b)</td><td><code>[^a]*(a[^a]*b)*[^a]*</code></td></tr>
<tr><td>ChainResponse(a,b)</td><td><code>[^a]*(ab[^a^b]*)*[^a]*</code></td></tr>
<tr><td>Precedence(a,b)</td><td><code>[^b]*(a.*b)*[^b]*</code></td></tr>
<tr><td>AlternatePrecedence(a,b)</td><td><code>[^b]*(a[^b]*b)*[^b]*</code></td></tr>
<tr><td>ChainPrecedence(a,b)</td><td><code>[^b]*(ab[^a^b]*)*[^b]*</code></td></tr>
<tr><td>CoExistence(a,b)</td><td><code>[^a^b]*((a.*b)|(b.*a))*[^a^b]*</code></td></tr>
<tr><td>Succession(a,b)</td><td><code>[^a^b]*(a.*b)*[^a^b]*</code></td></tr>
<tr><td>AlternateSuccession(a,b)</td><td><code>[^a^b]*(a[^a^b]*b)*[^a^b]*</code></td></tr>
<tr><td>ChainSuccession(a,b)</td><td><code>[^a^b]*(ab[^a^b]*)*[^a^b]*</code></td></tr>
<tr><td>NotChainSuccession(a,b)</td><td><code>[^a]*(a[^a^b])*[^a]*</code></td></tr>
<tr><td>NotSuccession(a,b)</td><td><code>[^a]*(a[^b]*)*[^a^b]*</code></td></tr>
<tr><td>NotCoExistence(a,b)</td><td><code>[^a^b]*((a[^b]*)|(b[^a]*))?</code></td></tr>
</table>

either b or a c. Figures 8.9b and 8.9c respectively show the support counted from the structure of Figure 8.9a in $L_1$ and $L_2$. While many a events are followed by b events, those b events have little overlap with the b events that are followed by either c or d. Because LPMs count the support of the process model *as a whole*, the support of the LPM on $L_2$ is very low, and as a result it will not be found unless the support threshold is set very low. This shows that combinations of Declare constraints, including target-branched Declare constraints, are not equivalent in the insights that they communicate to a single LPM that consists of the same activities.

A third important difference between declarative process discovery and LPM mining can be found in the notion of support. LPM mining uses the notion of *repetitive support*, as introduced in the sequential pattern mining field by Ding et al. [Din+09]. Repetitive support means that not only pattern instances (i.e., occurrences) that are repeating in *different traces* are counted, but also those that are *repeating within the same trace*. An additional constraint of repetitive support is that pattern instances are *non-overlapping*, i.e., two instances of the same pattern cannot share the same event. For example, the support of the LPM of Figure 8.9b on trace $\langle a, b, c, a, a, b, c\rangle$ is 2, as the last instance of a is not followed by an instance of b *that is not already matched to an* a. Furthermore, its support on trace $\langle a, b, c\rangle$ is 1. In contrast, the support for Declare constraints is generally defined on the number of traces in which the constraint is satisfied, therefore, the support of both Declare constraints in Figure 8.9d is 1 on both traces $\langle a, b, c, a, a, b, c\rangle$ and $\langle a, b, c\rangle$, as both constraints hold for all instances of a in both traces. Di Ciccio et al. [DM15] introduced an *event-based* support measure for declarative process discovery that has similarities to the repetitive support used for LPMs. However, in contrast to repetitive support,

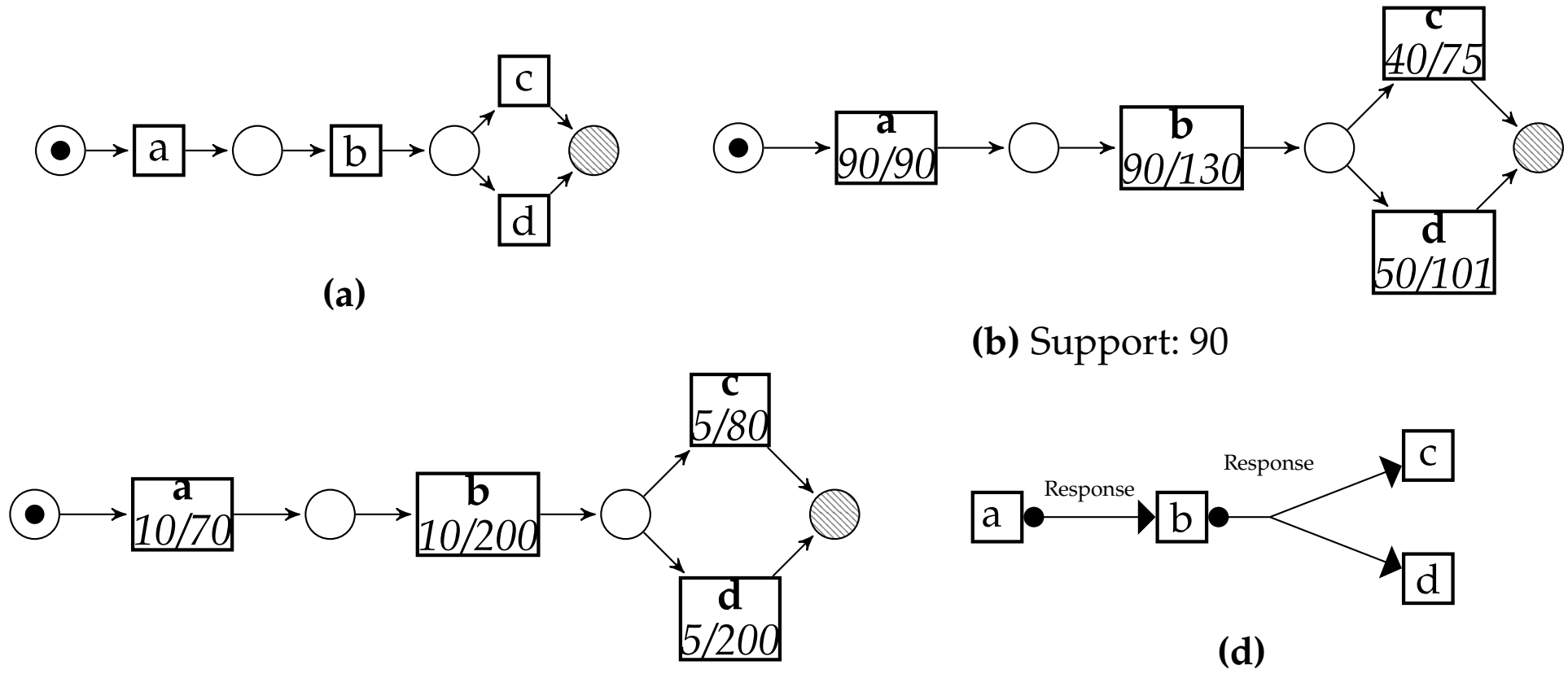

**(a)** **(b)** Support: 90 **(c)** Support: 10 **(d)**

**Figure 8.9:** *(a)* a Petri net describing a sequential relation between activities a, b, followed by a choice between c and d, *(b)* A Local Process Model mined from $L_1$, with the same structure as Figure 8.9a, *(c)* a Local Process Model with the same structure mined $L_2$, now showing a weaker relation between the four activities, and *(c)* the resulting target-branched Declare model mined from both $L_1$ and $L_2$ with minimum support 0.7, which consists of two declare constraints over the same set of activities: Response(a,b), and Response(b,{c,d}).

it does not require instances of the pattern to be *non-overlapping*, therefore all three of a would be counted as satisfying instances of constraint Response(a,b) in trace $\langle a, b, c, a, a, b, c \rangle$.

The target-branched Declare (TBDeclare) Miner [DMM16] is a recent declarative process discovery technique that mines TBDeclare constraints, i.e., Declare constraints where the *target* variable is extended to contain a set of activities instead of a single activity. An example TBDeclare constraint is *Response(a, {b,c})*, which expresses that if an activity *a* occurs, then eventually it is followed by either activity *b* or activity *c*. However, while TBDeclare constraints allow the target variable to contain multiple activities, the constraints themselves are still binary, and the relation between the activities in the target set is limited to an *exclusive choice*. Therefore, the fundamental difference between LPMs and declarative process model mining that is shown in Figure 8.9 is not bridged by TBDeclare. For example, a pattern such as: activity *a* is followed by both *b* and *c* in arbitrary order, cannot be discovered with TBDeclare. Note that the combination of Declare constraints Response(a,b) and Response(a,c) models somewhat similar behavior, but does not enforce that the instances of a that are followed by b are the same instances that are also followed by c.

DPIL [ZSJ14] is a declarative process modeling notation that like Declare specifies a set catalogue of constraint templates, which are called *macros*. For instance, the

*sequence(a,b)* macro states that the existence of the start of task b implies a previous occurrence of a completion of task a, similar to the precedence constraint template in Declare. Unlike Declare, DPIL offers extensive support for modeling constraint that involve the organizational aspect of the process, using macros to specify the resources or the roles of the resources that have to execute certain activities. DPIL rules can be mined from an event log using the DPIL Miner [Sch+16a]. However, the set of macros for control-flow constructs is limited, as DPIL for example has no support to express a parallel split [ZSJ14].

Alberto et al. [Alb+08] proposed a more general rule specification language that allows for the specification of rules that go beyond the traditional Declare templates by using a constraint language called SCIFF. SCIFF allows for the formulation of constraints that involve more than two activities, by having the precondition and/or the postcondition of the constraint define relations over multiple activities. Chesani et al. [Che+09] proposed a mining approach to mine SCIFF statements from event logs, however, this approach is limited to mining SCIFF statements that are identical to Declare constraints. While this mining approach is easily extensible to SCIFF arbitrary templates, this would be less computationally efficient than LPM mining. LPM mining iteratively growths patterns into larger patterns, allowing to prune the search space using the *downward closure property*, i.e., if a pattern does not meet the support threshold, then certain extensions of this pattern can theoretically also not meet the support threshold, therefore, these extensions do not have to be explored. An extension of [Che+09] to a more general set of constraints would not be able to leverage such pruning. The SQLMiner [Sch+16c] is an declarative process discovery approach that maps declare templates to SQL statements to leverage the query planner of a relational database system for efficient calculation of the support. In the paper the statement is made that any arbitrarily complex control-flow construct can be mapped to such an SQL statement, therefore, the method could be extended to constraints beyond the standard Declare templates. However, using the downward closure property for pruning like the LPM miner does is not trivial for such an extension.

#### *Hybrid Process Discovery*

Hybrid process discovery [MSR14; Sch+18] aims at discovering a process model that consists partially of procedural process model constructs and partially of declarative process model constructs. Existing hybrid process discovery approaches consist of a first step where activities are separated into a group of structured activities and a group of unstructured activities, based on the number of unique predecessors and successors of an activity. However, some activities, such as activity a in the event log of Figure 8.2a, are part of a frequent pattern, but also to occur as noise at random point in the traces. Such activities would be classified as noisy activities by existing hybrid process discovery approaches, resulting in this activity being modeled with binary declarative constructs.

### 8.5.2 Pattern Mining

*Sequential pattern mining*, which we briefly introduced in Chapter 2, is a mainstream research area in data mining. Some definitions of sequential pattern mining focused on extracting patterns that capture consecutive subsequences that recur in many sequences of a sequence database[Pei+01; Zak01]. Alternatively, algorithms have been proposed to mine *gapped* sequential patterns [AS95; Din+09; Ton+09; Wu+14], which allow gaps between two successive events of the pattern, and *repetitive* sequential patterns [Din+09; Ton+09], which capture not only subsequences that recur in multiple sequences, but also subsequences that recur frequently within the same sequence. Note that the combination sequential patterns that are both gapped as well as repetitive can be seen as mining LPMs in which all operator nodes in the process tree are →-nodes.

Some techniques in the pattern mining field focus on the mining of patterns that are more general than just sequential structures. For example, *episodes* [MTV97] extend sequential patterns with parallelism by allowing a pattern to incorporate *partial order relations*. [Pei+06] showed that episode mining techniques cannot mine any arbitrary partial order, and therefore proposes a technique to mine closed *partial order patterns*. In contrast to sequential pattern mining techniques, standard episode mining techniques focus on mining patterns from *a single sequence* and are not able to take an event log that consists of multiple sequences as input. Leemans et al. [LA14] later extended the concept of episode mining from the single trace input to event logs.

Sætrom and Hetland proposed several methods to [HS04; HS05; SH03] to mine patterns that are expressed in the regular-expression-like Interagon Query Language (IQL), thereby allowing the patterns to generalize to a higher degree than episodes and partial order patterns by additionally allowing patterns to incorporate choice and Kleene star (i.e., repetition) constructs. The mining techniques proposed by Sætrom and Hetland all make use a dedicated hardware device that was developed by the authors to make the mining computationally feasible. We will refrain from making comparisons between IQL pattern mining methods and LPM mining in this thesis since IQL pattern mining requires dedicated hardware.

Lu et al. propose a method called Post Sequential Patterns Mining (PSPM) [Lu+08] that takes as input a set of sequential patterns and post-processes them into a single graph consisting of sequential and exclusive choice constructs, which they call a Sequential Pattern Graph (SGP) [Lu+04]. In this work, *concurrency relations* between sequential patterns means that those patterns *occur together in the same sequences*. This notion of concurrency only works for sequence-based support, while for repetitive support patterns that can occur multiple times per sequence occurring in the same sequence is not enough to infer a concurrency relation. A later extension [LCK11] improves the procedure to extract concurrency relations between sequential patterns and proposes a visual notation, called a ConSP-Graph, to represent the concurrent relations between sequential patterns. An SGP can be discovered from an event log by first applying any existing sequential pattern mining algorithm followed by PSPM on the discovered set of sequential patterns.

Chen et al. [CLK10] extended the initial work on PSPM by Lu et al. [LCK11] by extracting *exclusive relations* between sequential patterns, i.e., patterns that do not occur in the same sequences, and propose a visual graph called an ESP-graph to visually represent such relations. However, while [CLK10] mention the extension of ConSP-Graphs to a richer set of workflow patterns as an important area of future work, no work has been done in the area of PSPM to mine graphs that can contain arbitrary combinations of concurrency, choices, sequential orderings, and loops in a single graph. Furthermore, the reliance of PSPM methods on the output of sequential pattern mining techniques can also be considered to be a drawback of the approach. When two activities a and b are in parallel then both the orderings $\langle a, b \rangle$ and $\langle b, a \rangle$ will be present in the complete log. However, if one of the two orderings is more frequent than the other due to chance, one of the two orderings might not reach the support set in the sequential pattern mining, making it impossible for PSPM methods to discover the concurrency of a and b. A more fundamental distinction between PSPM and Local Process Model (LPM) discovery is that PSPM merges all relations into one single pattern while LPM discovery aims at discovering a collection of patterns. Merging separate patterns into a single graph could result in one single overgeneralizing graph. For example in the case of log $L = [\langle b, a, c \rangle^{100}, \langle d, a, e \rangle^{100}]$, sequential pattern mining techniques will find two sequential patterns: $b, a, c$ and $d, a, e$. Merging them into a single graph where a is followed by either c or e and is preceded by either b or d results in the loss the long term dependency where b already determines the occurrence of a c after the a.

Jung et al. [JBL08] describe a method to mine frequent patterns from a collection of process models by transforming each business process to a vector format and then applying agglomerative clustering. Diamantini et al. [Dia+13; DPS12] take a similar approach, but apply graph clustering techniques instead of a traditional clustering approach. These techniques differ from LPM discovery as they take as input a set of process models instead of an event log. However, in many situations there are no existing process models available and, as shown in the introduction, it is not always possible to mine structured process models from an event log using process discovery techniques.

In later work, Diamantini et al. [Dia+16] describe a method to mine frequent patterns in process model notation through a two-step approach. First, each trace from the event log is transformed into a so-called instance graph [DA04], which is graph representation of a trace that shows which steps in the trace are performed sequentially and which steps are performed in parallel (i.e. overlapping in time). In the second step, they apply a graph clustering technique to obtain frequent subgraphs from this set of instance graphs. However, since instance graphs are limited to sequential and parallel constructs, other process constructs, such as choices and loops, cannot be discovered with the approach described in Diamantini et al. [Dia+16], while they can be discovered with LPM discovery.

## 8.6 Case Studies

We now illustrate the usefulness of Local Process Models (LPM) on real-life data using two case studies.

### 8.6.1 BPIC '12 Data Set

The Business Process Intelligence Challenge (BPIC) 2012 data set originates from a personal loan or overdraft application process in a global financial institution. Many existing process discovery algorithms are able to extract a fitting and precise process model from the BPIC'12 data, since the data set does not have the high level of variability on which existing process discovery algorithms fail. In order to obtain a data analysis task in which high-variability event data is involved, we take a different perspective on the BPIC'12 data: instead of analyzing the overall business process at the bank, we now analyze the day-to-day work routine of a specific employee at the bank. We therefore transform the event log by first merging all process instances into a single trace, then filtering on the events that were performed by a single randomly selected bank employee (with resource id 10939), ordering the events chronologically on their timestamp, and splitting the data into one trace for every working day. The obtained event log for resource 10939 consists of 49 cases (working days), 2763 events, and 14 types of business activities.

Figure 8.10 shows the Petri net discovered for resource 10939 with the Inductive Miner infrequent with a noise threshold of 20%. The discovered model only contains 13 non-silent transitions, as the activity *W_valideren aanvraag* is filtered out by the Inductive Miner because of its low frequency. The process model in Figure 8.10 is very close to a "flower model", which is the model that allows all behavior over its activities. The Inductive Miner without noise filtering returns exactly the flower model over the 14 activities in the log. The discovered process is unstructured because of a high degree of variance of the event log, which is caused by 1) the resource performing work on multiple applications interleaved, and 2) the resource only performing only a subset of the process steps for each application, and which process steps he performs might differ per application. For such a high-variance event log, it is likely that no start-to-end process model exists that accurately describes the behavior in the event log.

Figure 8.11 shows the top five LPMs discovered with the approach described in this chapter that we obtained by ranking the LPMs according to the default weighing scheme of the five quality criteria and filtering out the LPMs that describe the exact same set of activities as a higher-ranked LPM. These LPMs provide process insights that cannot be obtained from the start-to-end process model in Figure 8.10. Discovering the LPMs took 34 seconds on a machine with a 4-core 2.4

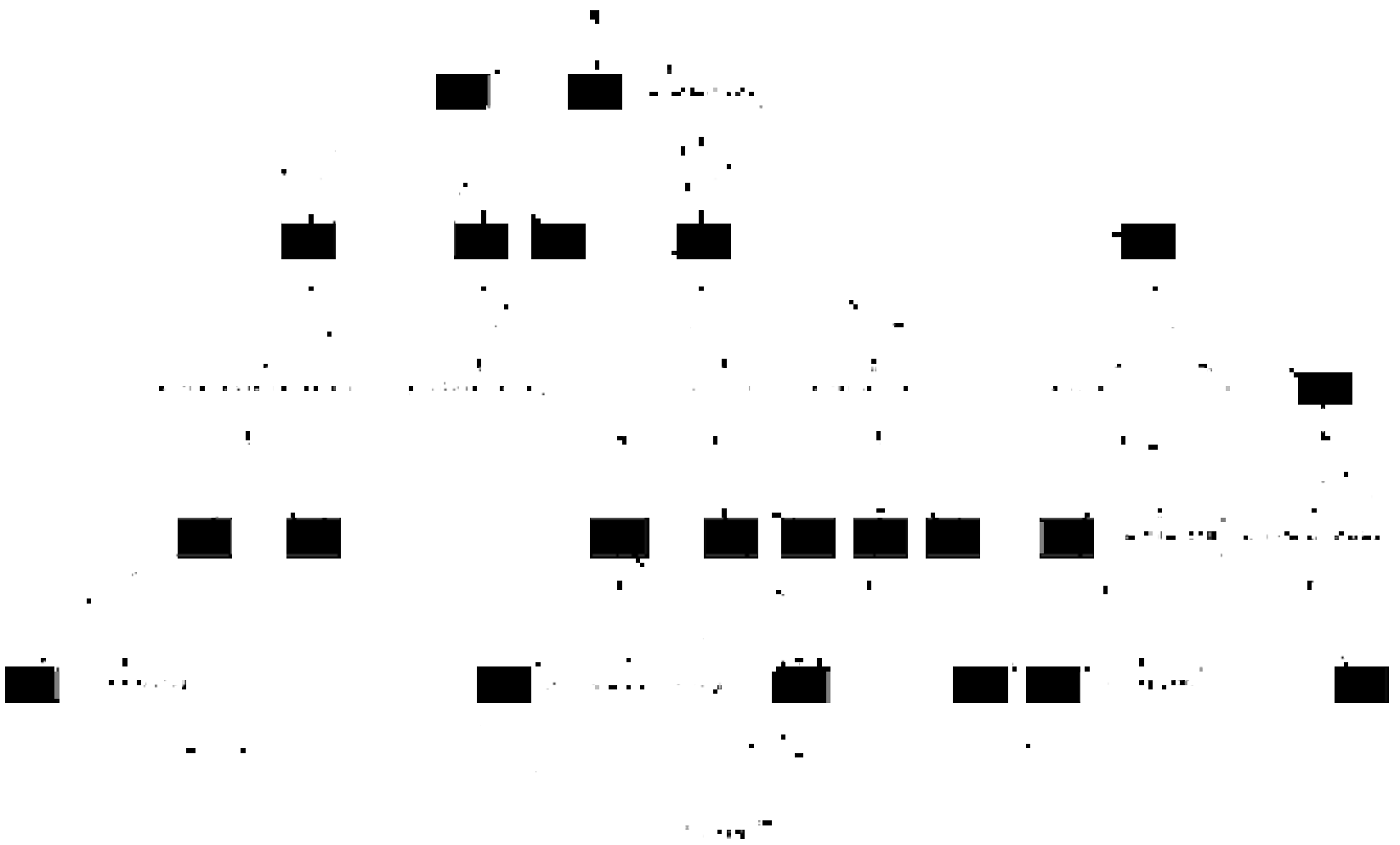

**Figure 8.10:** A process model representing the behavior of resource 10939 in the BPIC'12 log, obtained using the Inductive Miner infrequent (20%).

GHz processor using a support threshold of 0.7. Local process model *(a)* shows that all occurrences of events of type *O_SELECTED*, *O_CREATED*, and *O_SENT*, occur in this exact order. Figure 8.10 overgeneralizes by suggesting that for example *O_SELECTED* can be followed by three skip (black) transitions, after which another *O_SELECTED* or a *A_ACCEPTED* can be performed, which never happens in reality. *O_SELECTED* and *O_CREATED* in Figure 8.10 can be separated by *A_FINALIZED*, which makes the dependency between *O_SELECTED* and *O_CREATED* a long-term dependency, of which discovery is still one of the open problems in process mining [VW04] The LPM mining method does find this long term dependency, because each LPM candidate is evaluated on a version of the event log that is projected on the set of labels of candidate under evaluation.

LPM *(b)* is an extension of LPM *(a)* as the last three activities in the sequence are the same, therefore, each occurrence of LPM *(b)* in the log will also be an occurrence of *(a)*. LPM *(b)* starts with an additional activity *A_ACCEPTED* of which 103 out of 104 events follow this sequential pattern. The confidence of LPM *(b)* is lower than the confidence of *(a)*, because only 103 out of 124 events of the last three activities of the sequence in LPM *(b)* can be explained by the model while each event of these activities is explained by LPM *(a)*. From this, we can conclude that there are 21 occurrences of the sequence *O_SELECTED, O_CREATED, O_SENT* that are not preceded by *A_ACCEPTED*. Partly this can be explained by *A_ACCEPTED* only occurring 104 times, however, the model also shows that there is one *A_ACCEPTED* event that is not followed by *O_SELECTED, O_CREATED*, and *O_SENT*. It might

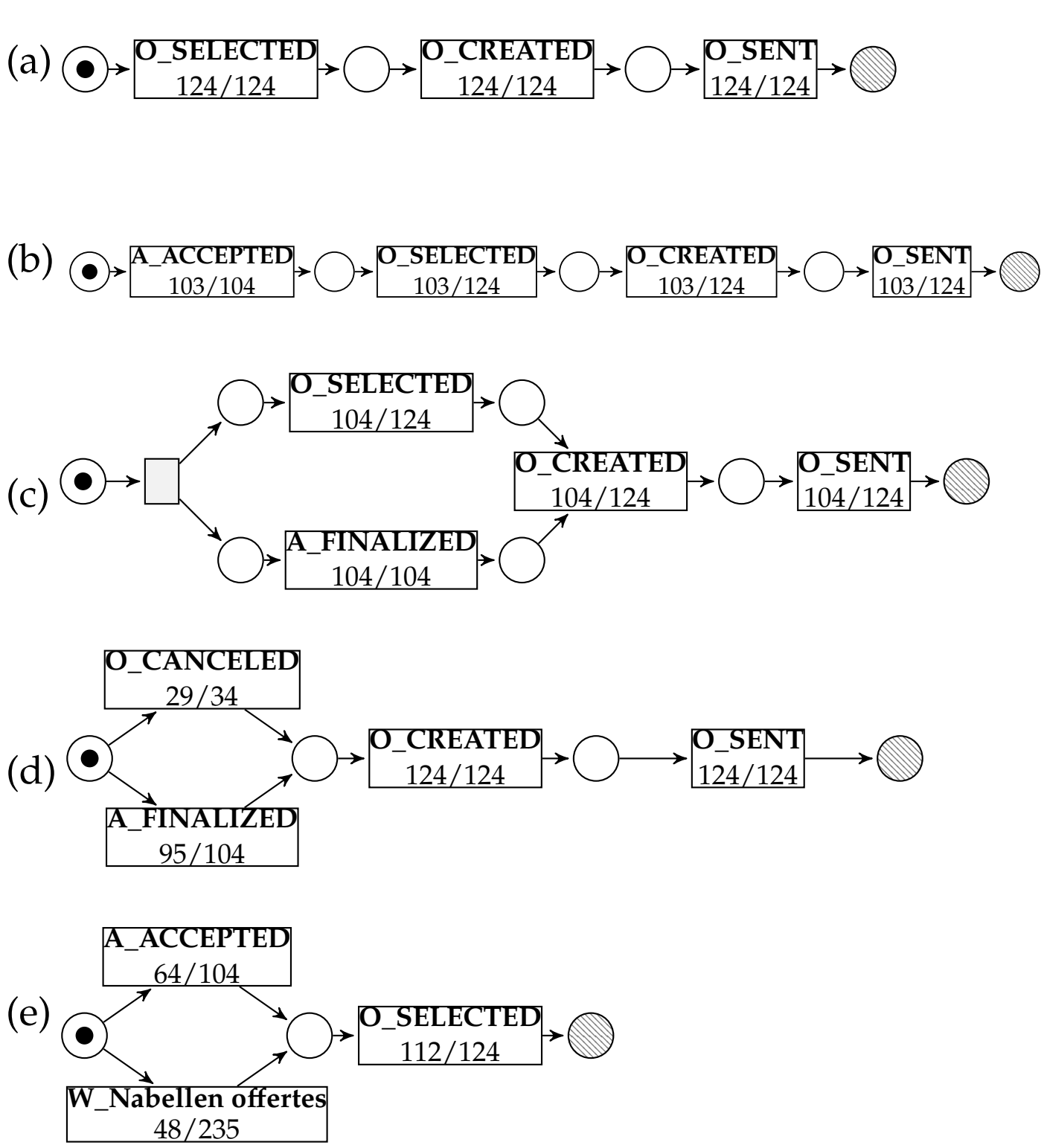

(a)

| | |
|---|---|
| Support | 0.9920 |
| Confidence | 1.0000 |
| Language $fit_5$ | 1.0000 |
| Determinism | 1.0000 |
| Coverage | 0.1346 |

(b)

| | |
|---|---|
| Support | 0.9904 |
| Confidence | 0.8655 |
| Language $fit_5$ | 1.0000 |
| Determinism | 1.0000 |
| Coverage | 0.1723 |

(c)

| | |
|---|---|
| Support | 0.9905 |
| Confidence | 0.8739 |
| Language $fit_5$ | 1.0000 |
| Determinism | 0.8811 |
| Coverage | 0.1723 |

(d)

| | |
|---|---|
| Support | 0.9920 |
| Confidence | 0.9374 |
| Language $fit_5$ | 1.0000 |
| Determinism | 0.7591 |
| Coverage | 0.1397 |

(e)

| | |
|---|---|
| Support | 0.9912 |
| Confidence | 0.3933 |
| Language $fit_5$ | 1.0000 |
| Determinism | 0.6666 |
| Coverage | 0.2753 |

**Figure 8.11:** Five local process models discovered on the BPI'12 log using the technique presented in this chapter. Clearly these models provide more insight than Figure 8.10.

be the case that this *A_ACCEPTED* event does not fit the regular workflow, or alternatively it might be the case that the other process steps of after *A_ACCEPTED* were executed by a different resource. Note that the determinism of LPMs *(a)* and *(b)* is 1.0, since both LPMs are sequential. The language fit of both LPMs is also 1.0, since both allow for only one execution path, which is observed in the log.

Local process model *(c)* shows that all instances of *A_FINALIZED* are in parallel with *O_SELECTED*, and ultimately followed by *O_CREATED* and *O_SENT*. This is more informative than Figure 8.10, which allows for much more behavior over activities *A_FINALIZED*, *O_SELECTED*, *O_CREATED*, and *O_SENT*.

Local process model *(d)* shows that each *O_CREATED* and *O_SENT* is preceded by either *O_CANCELED* (29 times) or *A_FINALIZED* (95 times). Also most of the *O_CANCELED* events (29 out of 34) and most of the *A_FINALIZED* events (95 out of 104) are followed by *O_CREATED* and *O_SENT*. Figure 8.10 does not provide the insight that *O_CANCELED* is followed by *O_CREATED* and *O_SENT*. Note that the determinism of LPM *(d)* is lower than the determinism of LPM *(c)*. This is in agreement with the intuition of determinism, as the concurrency at the start of LPM *(c)* can be regarded as a choice between two activities followed by a deterministic step of executing the other activity, while LPM *(d)* starts with a choice between two activities. After the concurrency in LPM *(c)* and the choice in LPM *(d)* respectively, the two models proceed identically. Local process model *(d)* has higher confidence than LPMs *(b)* and *(c)* as only five of the *O_CANCELED* and nine of the *A_FINALIZED* events cannot be explained by the model. LPM *(d)* has a higher confidence than LPM *(c)*, mostly because all occurrences of *O_CREATED* and *O_SENT* could be aligned in LPM *(d)* while only 104 out of 124 could be aligned in LPM *(c)*.

Notice that the number of events that were aligned on *A_FINALIZED* is lower in LPM *(d)* than in LPM *(c)*. This indicates that there are six occurrences where the alignments aligned on *O_CANCELED* while it was possible as well to align on *A_FINALIZED* (as both occurred). Therefore, an inclusive choice construct would have been a more correct representation than the exclusive choice that is currently included in the LPM. Note that our process tree based discovery approach allows for easy extension with additional operators, like e.g. an inclusive choice operator.

LPM *(e)* shows an example of a weaker pattern that performs lower on some quality measures but can still be discovered with the described approach. The coverage of LPM *(e)* is much higher than the other models as *W_Nabellen offertes* (Dutch for "Calling after call for tenders") is a frequently occurring event in the log. The confidence of LPM *(e)* is however much lower it explains only a fraction of the *W_Nabellen offertes* events.

### 8.6.2 Comparison with Related Techniques

In this section, we apply some of the related techniques described in Section 8.5 to the event log of BPI'12 resource 10939 and compare the insights that can be obtained with those methods with the insights that we obtained with LPM discovery.

We start with the Declare miner [MMA11], which mines a set of constraints from the data based on a set of constraint templates. Figure 8.12b shows the result of the Declare miner [MMA11] on the BPI'12 resource 10939 event log with a support threshold of 90%, requiring that the constraints hold in 90% of the cases. The model shows that a *choice* constraint holds between *O_SELECTED* and *W_Nabellen offertes*, indicating that on each working day either at least one event of type *O_SELECTED* or *W_Nabellen offertes* occurs. The same can be said about the pairs of event *W_Nabellen offertes* and *O_SENT*, *W_Nabellen offertes* and *O_CREATED*, and *W_Nabellen offertes* and *O_Completeren aanvraag*. Furthermore a *not chain succession* constraint is discovered between *W_Nabellen offertes* and *O_SENT*, indicating that *W_Nabellen offertes* and *O_SENT* never directly follow each other. *Not chain succession* constraints are also discovered between *W_Nabellen offertes* and *O_SELECTED*, and between *W_Nabellen offertes* and *O_CREATED*. Note that the none of the insights that we obtained from the LPMs in Figure 8.11 could be obtained from this Declare model.

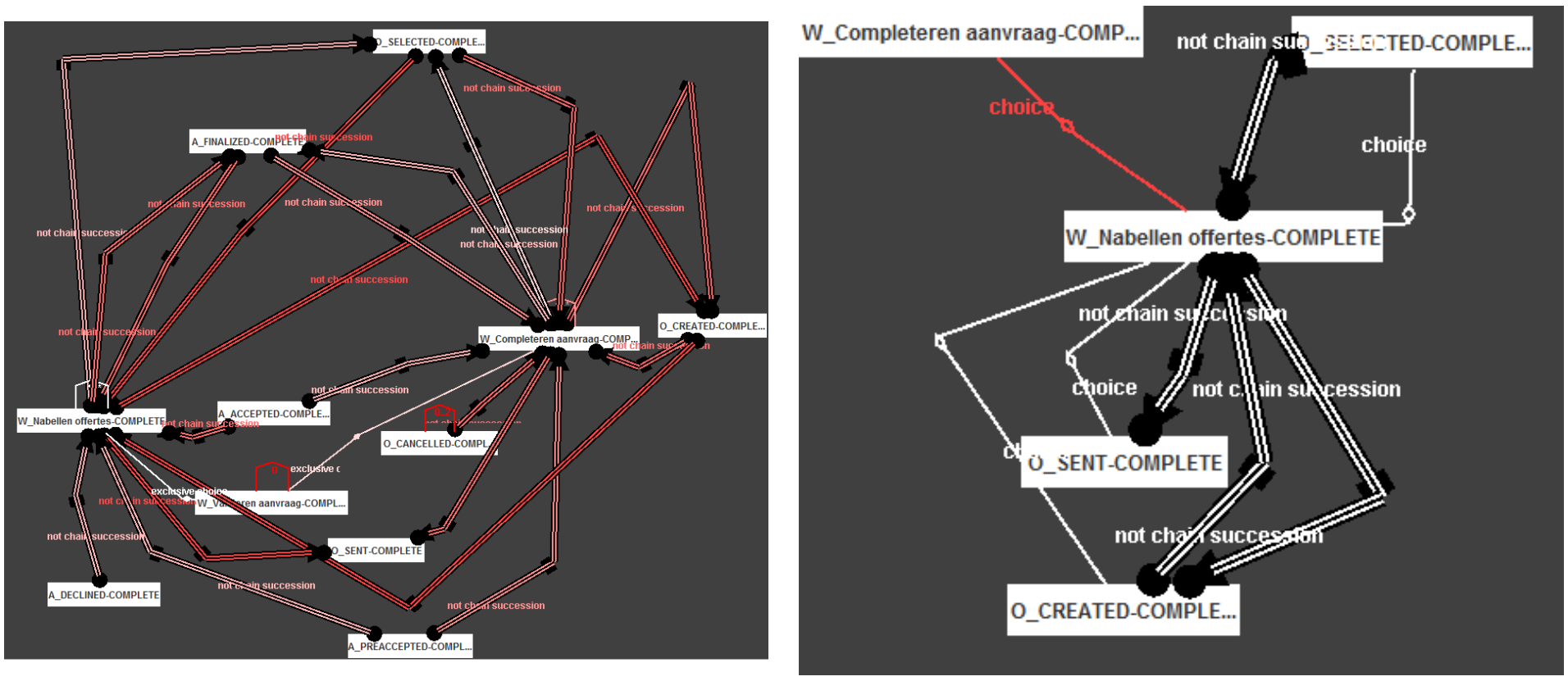

(a) (b)

**Figure 8.12:** *(a)* The result of the Declare miner [MMA11] with a support threshold of 80% on the BPI'12 resource 10939 event log. *(b)* The result of the same method on the same log with support threshold 90%.

By lowering the support threshold parameter of the Declare miner to 80%, we obtain a larger set of constraints. An *exclusive choice* constraint is found between *W_Valideren aanvraag* and *W_Nabellen offertes*, indicating that 80% of the cases contain one of the two activities but not both. The same type of constraint is found between *W_Valideren aanvraag* and *W_Completeren aanvraag*. The rest of the constraints found are *not chain succession* constraints.

To find constraints that can be deduced from the LPMs of Figure 8.11, such as the sequential ordering between *O_SELECTED* and *O_CREATED* from LPM *(a)*, the support threshold would need to be lowered even further, leading to an overload of constraints being found by the Declare miner. The Declare miner evaluates the constraints based on the ratio of cases in which the constraint holds. However, when activities are often repeated *within* cases, this is not a useful evaluation criterion. Employee 10939 performs most of the activities multiple times during a working day, therefore, to assess whether an activity a is generally followed by an activity b it is more useful to count the ratio of *occurrences* of activity a that are followed by b as in LPM discovery, instead of the number of cases that contain an a event that is followed by a b event.

Even more important is the fact that Declare miner is limited to a predefined set of rule templates, which often are limited to binary relations, while the LPM mining can mine arbitrary n-ary relations. That is likely to be the cause of Declare mining not finding any of the LPM relations found in Figure 8.11. At the same time this difference provides an explanation why Declare mining finds so many uninteresting *not chain succession* constraints between activities: when there are multiple a events in a trace, you are likely to find at least one a that is in a *not chain succession* relation with some activity b, leading to a high ratio of traces that fulfill such a *not chain succession* constraint.

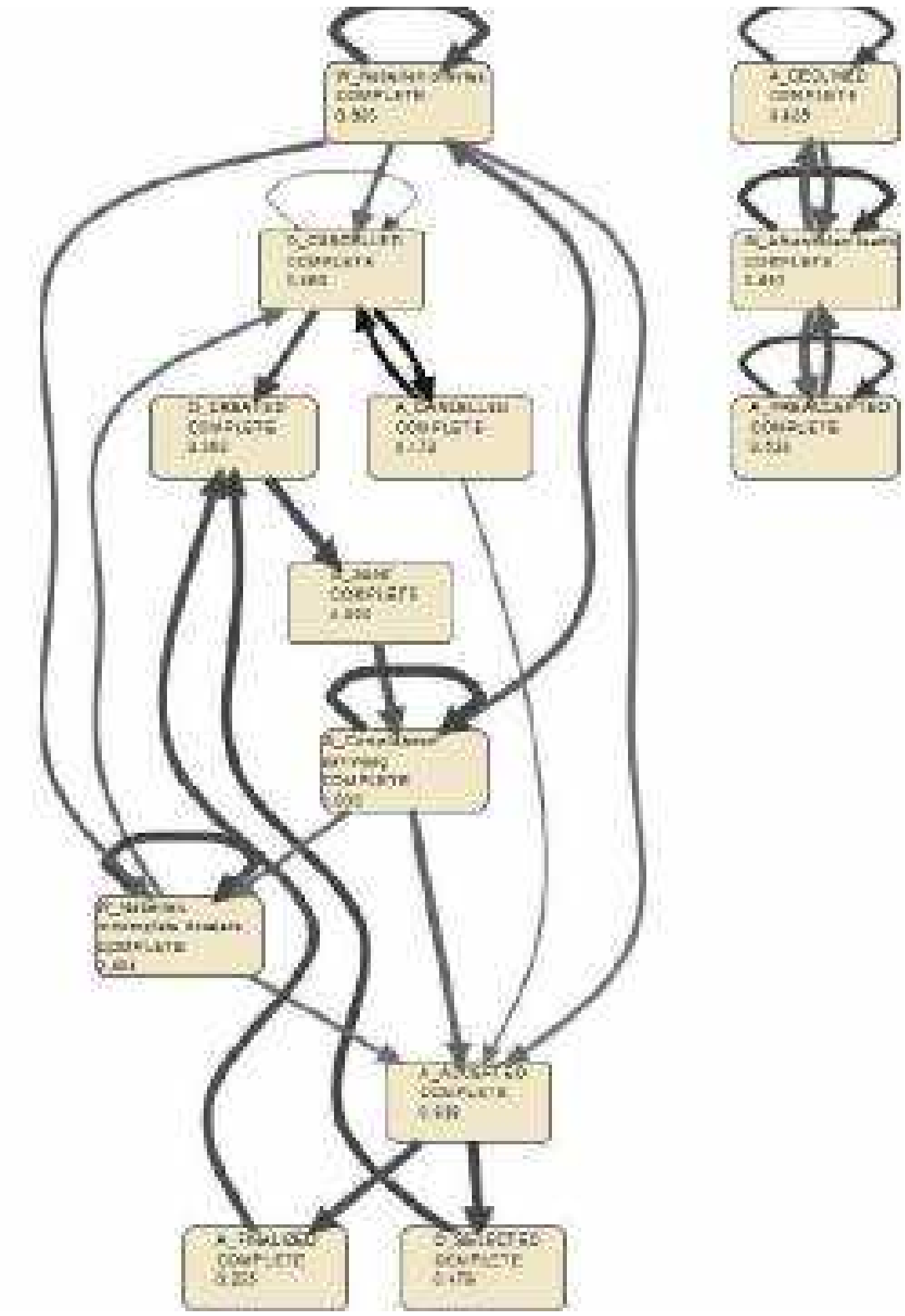

**Figure 8.13:** The result of the Fuzzy miner [GA07] with default parameters on the BPI'12 resource 10939 event log.

The Fuzzy Miner [GA07] is a process discovery technique developed to deal with complex and flexible process models. It connects nodes that represent activities with edges indicating follows relations, taking into account the relative significance of follows/precedes relations and allowing the user to filter out edges using a slider. However, the process models obtained using the Fuzzy Miner lack formal semantics, e.g. when a node has two or more outgoing edges, it is undefined whether this represents a choice, an exclusive choice, or parallel execution of the connected nodes.

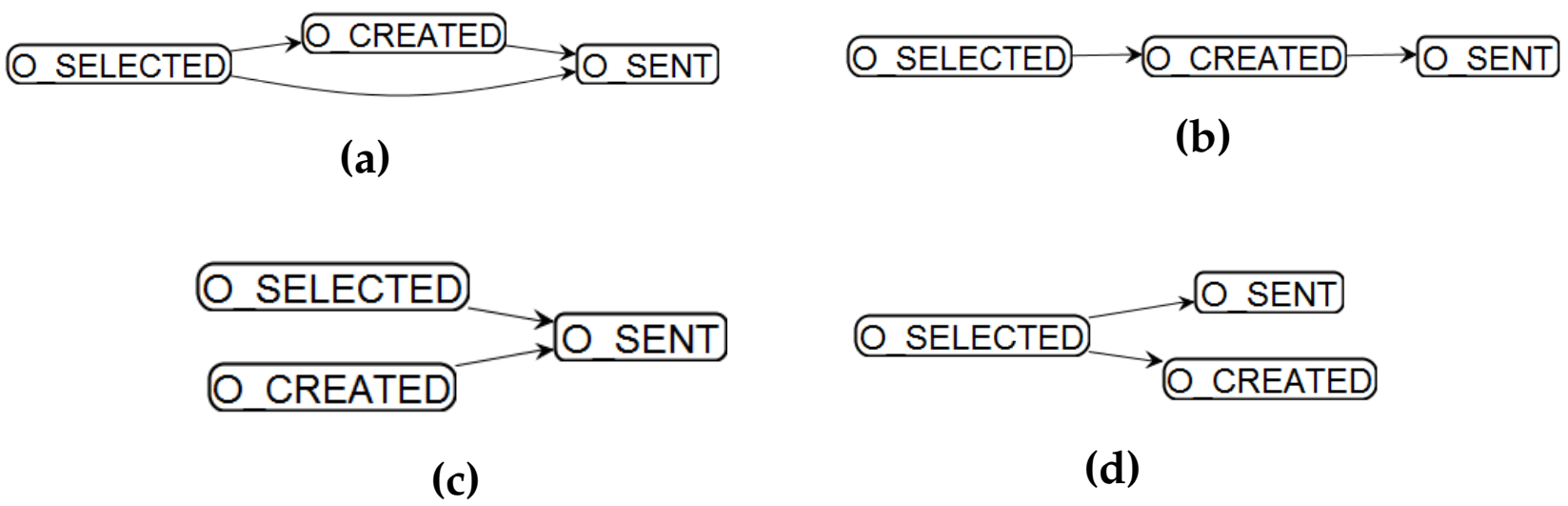

**Figure 8.14:** The first four episodes discovered by ProM's Episode Miner [LA14] on the BPI'12 resource 10939 event log.

Figure 8.13 shows the result of the Fuzzy miner on the BPI'12 resource 10939 event log with default parameters. The discovered Fuzzy model does contain a path from *O_SELECTED* through *O_CREATED* to *O_SENT*, which were shown to be in a sequential relation by LPM *(a)*. However, the Fuzzy model allows for many more paths, therefore the sequential relation between those three activities cannot be inferred from the Fuzzy model. LPM *(c)* showed a sequential path between *O_CREATED* and *O_SENT* that is preceded by an arbitrary ordering of activities *O_SELECTED* and *A_FINALIZED*. The Fuzzy model also shows arrows from both *O_SELECTED* and *A_FINALIZED* to *O_CREATED*, however, as a Fuzzy model does not make a distinction between parallelism, inclusive choice constructs and exclusive choice constructs, it does not answer the question whether *O_SELECTED* is preceded by *both* *O_SELECTED* and *A_FINALIZED*, or just by one of the two.

ProM's Episode Miner [LA14] is a method that can be considered to be in-between episode mining and process mining, as it discovers a collection of patterns from an event log where each pattern consists of partial order constructs. However, contrary to the technique that we describe in this chapter, ProM's Episode Miner does not support loop and exclusive choice constructs and is not easily extensible to include new types of constructs.

Figure 8.14 shows the first four episodes found with ProM's Episode Miner on the BPI'12 resource 10939 event log. The first two episodes show the same sequential ordering from *O_SELECTED*, *O_CREATED*, and *O_SENT* that is represented by LPM *(a)*. In fact, the output of the Episode miner in this case is redundant, since episodes *(a)* and *(b)* represent exactly the same behavior. LPM *(a)* however shows that all of the *O_SELECTED* events are followed by an *O_CREATED* event, therefore it is never skipped. Episode *(c)* indicates that *O_SELECTED* and *O_CREATED* can happen in any order, but both of them have to occur before *O_SENT* occurs and episode *(d)* indicates that *O_SELECTED* has to happen before *O_SENT* and *O_CREATED* can occur. Episodes *(a)*, *(c)* and *(d)* can be considered to be less specific versions of episode *(b)*. ProM's Episode Miner is not able to discover patterns with choice constructs like LPM *(d)*, or patterns with loops.

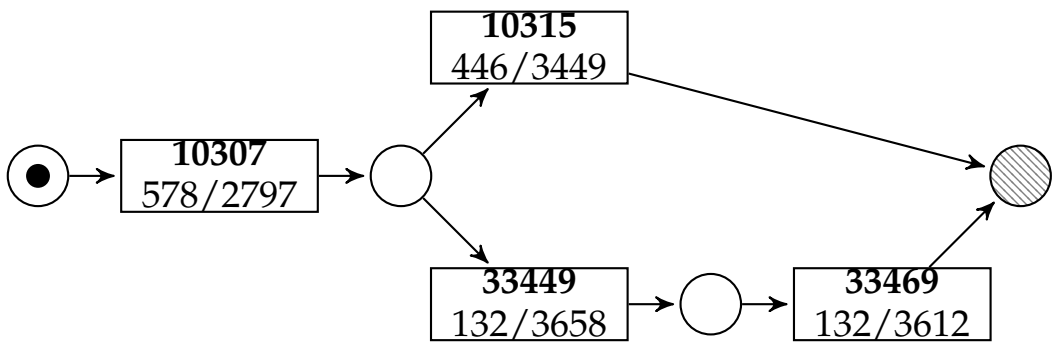

**Figure 8.15:** A non-sequential local process model discovered on the Gazelle data set.

### 8.6.3 Gazelle Data Set

The *Gazelle* data set is a real-life data set used in the KDD-CUP'2000 and contains customers' web click-stream data provided by the Blue Martini Software company. The Gazelle data set has been frequently used for evaluating sequential pattern mining algorithms. For each customer, there is a series of page views, in which each page view is treated as an event. The data set contains 59602 sequences (customers), 149638 events (page views), and 497 distinct event types (web pages). On average, each trace only contains 2.5 events. However, the lengths of the traces in the event log has a high variance, with 1678 traces containing ten or more events and the longest trace containing 267 events. More detailed information on the Gazelle data set can be found in [Koh+00]. We compare the LPM that we found on this data set with the sequential patterns obtained with the well-known sequential pattern mining algorithm PrefixSpan [Pei+01] as implemented in the SPMF [Fou+14a] sequential pattern mining library. We set the minimal support parameter of the sequential pattern mining algorithms to 10% of the number of input sequences. Since sequential patterns are just a specific form of LPM (using only the →-operator), sequential patterns that are obtained with a sequential pattern mining algorithm are also found when mining LPMs using the same support threshold.

Additionally, several non-sequential patterns were discovered that cannot be discovered with sequential pattern mining techniques, an example of which is shown in Figure 8.15. This shows that this well-known sequential pattern mining evaluation data set contains frequent and high-confidence patterns that cannot be found with sequential pattern mining approaches, but can be found with the LPM mining algorithm. This indicates the applicability of LPM mining to the field of pattern mining.

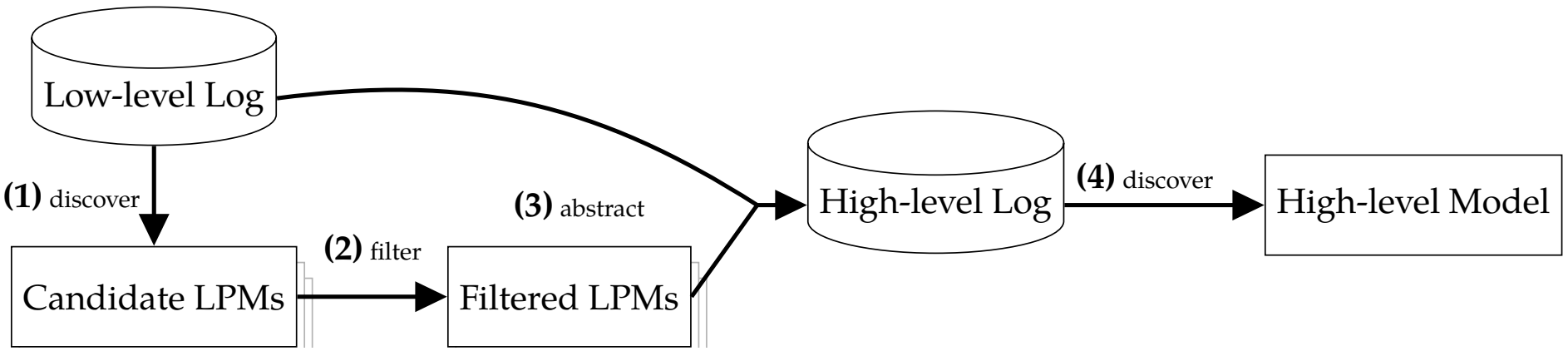

**Figure 8.16:** An overview of the proposed unsupervised abstraction technique.

## 8.7 An Example Application: Event Abstraction

In the previous section we have seen different use cases where LPM helped to gain an understanding of the behavior that is present in an event log. In this section, we will briefly discuss an application of LPMs other than for the sake of mining process model patterns in itself. The application of LPMs that we cover in this section is the abstraction of low-level events to higher level events. This closely links this chapter to Chapter 7 in which we introduced a supervised approach to abstracting low-level events to higher level events. In contrast, the approach that we will introduce in this section does not need labeled training data and can be used in an unsupervised way.

A recent approach to abstract recorded events to high-level activities [Man+16b; Man+18] uses *activity patterns* to capture the domain knowledge about the relation between the high-level activities and the low-level recorded events. Each activity pattern is a process model that describes the possible behavior in terms of low-level events that are conjectured to be observed during the execution of a certain high-level activity. However, such domain knowledge might not be available, and when the process contains many activities it becomes a tedious task to model each of them manually. We conjecture that LPMs can help to replace the domain knowledge that is needed in the approach of [Man+16b; Man+18] by using mined LPMs as activity patterns for the abstraction step.

In this section, we explore the application of automatically discovered LPMs to replace the domain knowledge in pattern-based abstraction to form a *novel, completely automated, abstraction technique*. Figure 8.16 gives an overview of the proposed method. Each LPM discovered by the LPM discovery method is assumed to represent a high-level activity. Then, we propose a technique to filter the set of LPMs, and use this filtered set of LPMs as activity patterns with the event abstraction method proposed in [Man+16b; Man+18] and discover a high-level process model using off-the-shelf discovery methods. Our proposed, integrated approach aims to *improve the precision* of process models found by process discovery techniques by abstraction of the event log while not sacrificing too much fitness.

In Section 8.7.1, we briefly introduce the pattern-based abstraction as proposed by [Man+16b; Man+18]. Section 8.7.2 explains how LPM discovery and pattern-based abstraction can be combined to form a fully automated abstraction technique. In Section 8.7.3, we present and discuss some preliminary results, and provide some concluding remarks on the use of LPMs for event abstraction in Section 8.7.4.

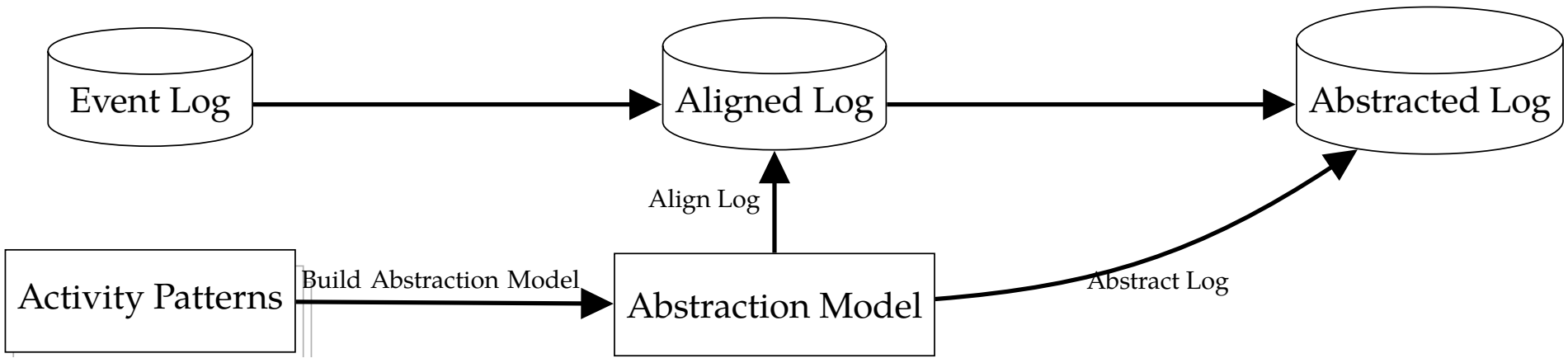

**Figure 8.17:** An overview of the pattern-based abstraction method described in [Man+16b; Man+18]. These steps jointly form step 3 of Figure 8.16."

### 8.7.1 Pattern-based Abstraction

An overview of the pattern-based abstraction method proposed in [Man+16b; Man+18] is depicted in Figure 8.17. The input is an event log and a set of activity patterns. Each activity pattern is a process model that represents the behavior expected for one execution of a high-level activity. Moreover, a mapping from activities to life-cycle transitions of the high-level activity is required, i.e., $\theta : T \nrightarrow LT$ with, e.g., $LT = \{start, complete\}$. Mapping $\theta$ allows to obtain information on when activities started and when they were completed, which some process discovery algorithms are able to leverage (for example, [LFA15a]).

Activity patterns are composed to an abstraction model and, then, an alignment technique [ADA11] is used to obtain a high-level event log. Through the use of the alignment technique, we can capture approximate executions of activity patterns. Then, we use the alignment information to create a corresponding high-level event log by only retaining those events that were aligned to activities $t \in T$ that are mapped to high-level activities, i.e., $t \in dom(\theta)$. Activity patterns may use any kind of process models with clear semantics, thus, the abstraction method also works with LPMs. For example, assume that LPM $LPM_1$ represents the behavior expected for some high-level activity. Mapping function $\theta$ could be defined such that the transition corresponding to A and transition $\tau_1$ are mapped to *start* and the transition corresponding to C is mapped to *complete*. It is possible to automatically obtain such mapping function $\theta$, e.g., by mapping the source activities to the start and the and the sink activities to the complete life-cycle transitions.

### 8.7.2 Unsupervised Abstraction Technique

We use discovered LPMs to replace the domain knowledge originally used in the pattern-based event abstraction method. Our proposed method discovers a high-level process model in the following four steps shown graphically in Figure 8.16.

1. We discover a *fixed number of candidate LPMs* based on their ranking on support (i.e., frequency). When setting the number of LPMs to use for abstraction to k, we select the top k LPMs of the discovered ranking of LPMs *LPMS*.

2. The LPMs in ranking *LPMS* can overlap in the activities that they describe. For the purpose of event abstraction this can be undesirable, because this results in multiple similar patterns for pattern-based abstraction, making it unclear which low-level event belongs to which high-level activity. Therefore, we filter *LPMS* using Algorithm 4. Algorithm 4 provides to diversifies the set of mined LPMs in a simple naive way by looking only at the set of activities that are used by the LPMs. This approach starts by selecting the top LPM from the ranking of LPMs obtained by the original LPM mining procedure, and then iterates over the ranking of LPMs, thereby selecting each LPM where the minimal Jaccard-distance of the alphabet of activities in the LPM with the alphabet of activities in one of the already selected LPMs exceeds a minimum *diversity threshold*. Note that we will in more detail explore algorithms for constructing a set of LPMs that takes into account the overlap of the LPMs in Chapter 12. In principle, any of the algorithms that we will propose there can be used to select the LPMs, instead of Algorithm 4.

3. We use those LPMs as *activity patterns* and obtain a high-level event log with the *abstraction method* described in [Man+16b]. In [Man+16b], several composition functions are presented that allow fine grained-control of the interaction between activities. In our case, we do not assume any domain knowledge on the process, thus, we limit the choice of composition parameter for our method to: *interleaving* and *parallel*. Interleaving means that any two high-level activities cannot occur at the same time, whereas parallel means that no such restriction is placed. Additionally, use the *loop* composition function to allow for repeated occurrence of the same high-level activity in a single trace.

4. Based on the high-level event log, we *discover a high-level process model* using existing methods such as the Inductive Miner [LFA13b].

For example, take LPM $LPM_1 = \rightarrow(\times(\circlearrowleft(B), A), C)$ and assume that we apply the proposed method to trace $\sigma = \langle A, \overline{\mathbf{B}, X, \mathbf{B}, \mathbf{C}}, C, \overline{\mathbf{A}, B, \mathbf{C}}, \overline{\mathbf{B}, \mathbf{B}, X, A, \mathbf{C}} \rangle$ leading to instances $\Gamma_{LPM_1}(\sigma) = \{(\sigma, \langle 2, 4, 5\rangle), (\sigma, \langle 7, 9\rangle), (\sigma, \langle 10, 11, 14\rangle)\}$ (highlighted with overline and in bold). We assume that LPM $LPM_1$ represents the behavior of some high-level activity H. Three instances of the LPM $LPM_1$ are present in trace $\sigma$, thus, the high-level activity H was executed three times. When applying pattern-based abstraction, we obtain the high-level trace $\langle H, H, H\rangle$[25]. Some low-level events in

[25] The abstraction technique also provides events for the start and complete life-cycle transitions. So far, we only use the complete transition.

**Input:** event log L, non-empty LPM list *LPMS*, diversity threshold $t \in [0, 1]$
**Output:** filtered LPM list *LPMS′*
*Initialisation* :
$i = 2, LPMS' = \langle LPMS(1) \rangle$
*Main Procedure* :
**while** $i \leq |LPMS|$ **do**
  $min_div = \min_{LPM \in LPMS'} \frac{|Activities(LPM) \cup Activities(LPMS(i))| - |Activities(LPM) \cap Activities(LPMS(i))|}{|Activities(LPM) \cup Activities(LPMS(i))|}$
  **if** $min_div \geq t$ **then**
    $LPMS' = LPMS' \cdot \langle LPMS(i) \rangle$
  **end if**
  $i = i + 1$
**end while**
**return** *LPMS′*

**Algorithm 4:** Heuristic LPM filtering.

$\sigma$ could not be matched to a high-level activity, e.g., the first event of the trace: A. Since we cannot assume that the discovered LPMs represent the entire behavior of the process, we add all those low-level events to the resulting trace. When we are planning to use a process discovery algorithm that can leverage lifecycle information of events (e.g., such as [LFA15a]), we generate the following trace: $\langle A, H_{start}, X, H_{complete}, C, H_{start}, B, H_{complete}, H_{start}, X, A, H_{complete} \rangle$. Alternatively, when we follow up this abstraction step by applying a process discovery algorithm that is not lifecycle-aware, we only induce the *complete*-steps of the high-level events, i.e., we generate $\langle A, X, H, C, B, H, X, A, H \rangle$.

Note that our technique does not discover the names of the high-level activities represented by the LPMs. However, LPMs can be labeled to its corresponding business activity based on domain knowledge.

### 8.7.3 Preliminary Results and Discussion

To be able to deal with the computational complexity of LPM mining for logs with many activities, we discover LPMs using activity clustering based on Markov clustering, as proposed in Chapter 10. As proposed in [Man+18], we expand high-level activities in the discovered process model with the corresponding LPMs to provide a fair comparison based on the low-level event log. The resulting *expanded process model* can be related to the low-level events with existing techniques that determine the models quality in terms of fitness and precision. We evaluate the quality of the expanded process models in terms of F-score [De +11], i.e., the harmonic mean between fitness and precision.

Figure 8.18 shows the F-score of the process models discovered with the IM infrequent [LFA13a] process discovery algorithm with 20% noise filtering. Horizontally it shows the results for different LPM *diversity thresholds* (0.2 to 0.9). Vertically it shows the results for five different event logs: the BPI challenge 2013 incidents (I) and closed problems (C) log [Ste13], the CoSeLoG receipt phase log [Bui14], the road fines log [LM15], and the Sepsis log [MB17]. The results are shown for 1 to 5 LPMs used in abstraction and parallel and interleaving composition. The dotted line indicates the F-score of the process model discovered from the original event log, i.e., without abstraction.

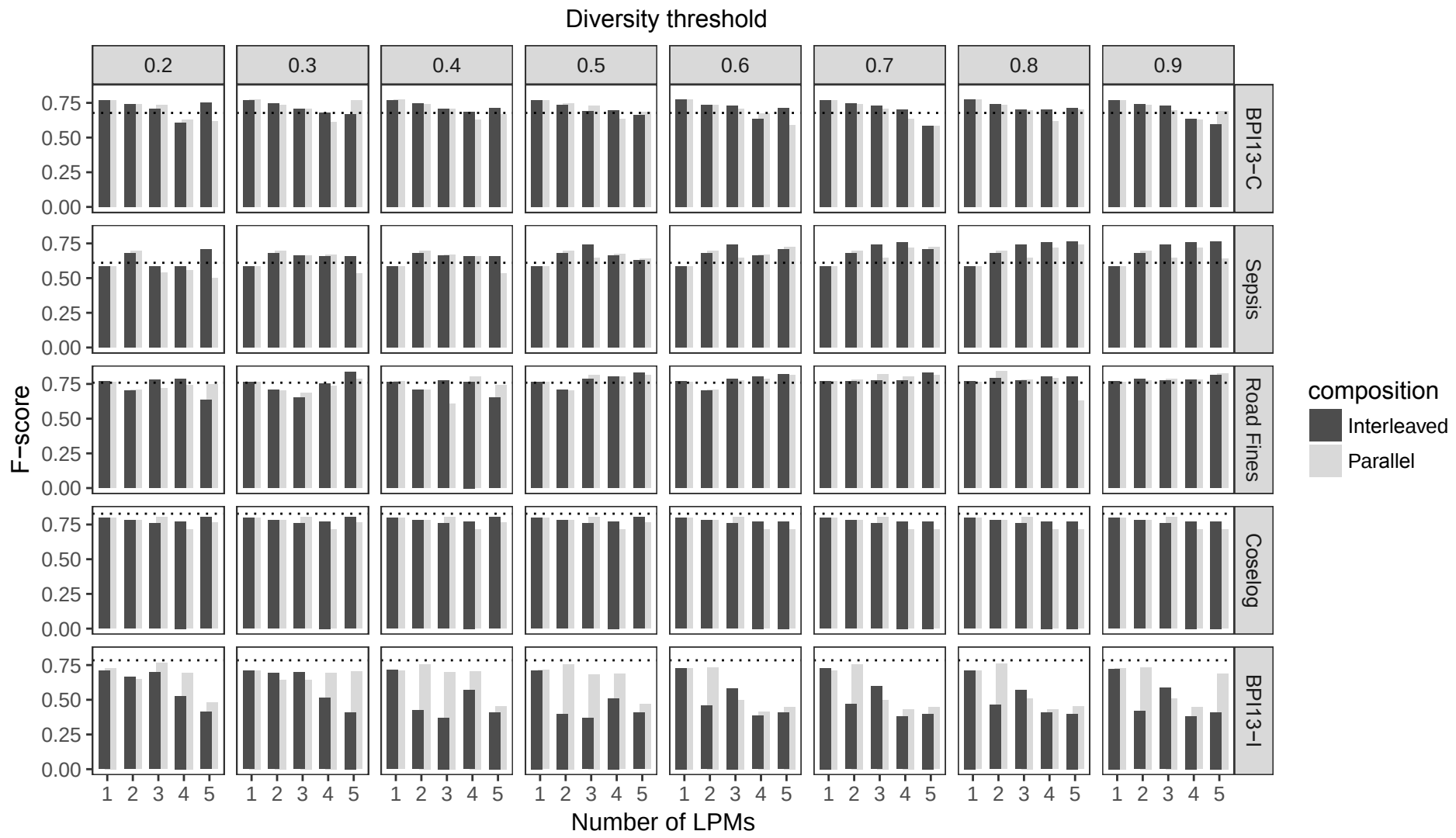

**Figure 8.18:** The F-scores of models discovered with the IM infrequent after applying the proposed method compared to the F-score obtained without the method (dotted line).

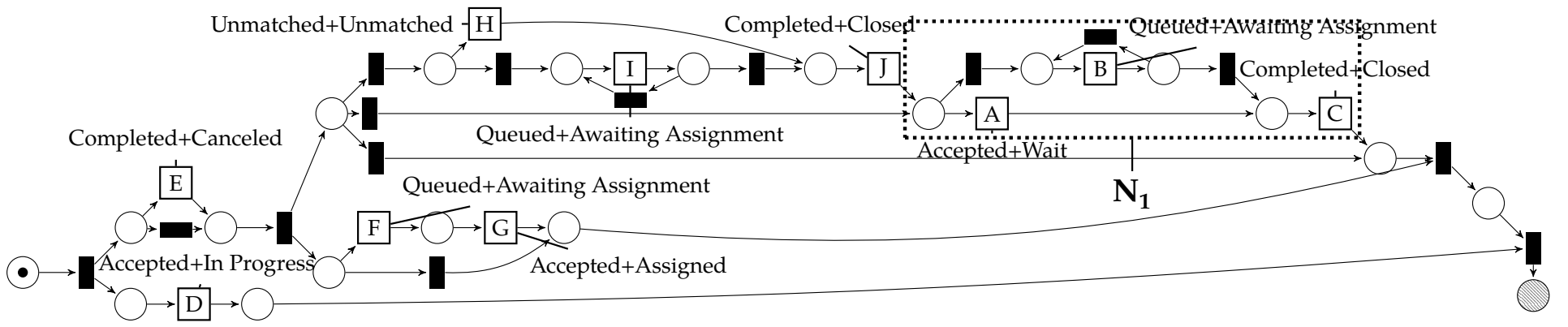

**Figure 8.19:** A Petri net discovered for the BPI13-C event log with parallel composition.

Figure 8.19 shows an expanded process model that is discovered for the BPI13-C log based on the proposed method. The LPM $N_1$ was used as an activity pattern and is part of the process model. Thus, the process model can hierarchically decomposed into sub-processes based on the activity patterns. Its precision score is improved from 0.53 to 0.86 at the expense of fitness, which drops from 0.84 to 0.65. The F-score improved from 0.65 to 0.74.

In our preliminary results, we found that for three out of five event logs the F-score of the process model can be improved by abstracting the event log prior to process discovery. Furthermore, it seems that abstracting with parallel composition does only improve the process model over abstraction with interleaving composition in one case, which is beneficial since the interleaving composition is computationally less expensive. The optimal number of LPMs used for abstraction differs between event logs, with the optimal number of LPMs being 5 for the sepsis log and 1 for the BPI13-C log. To make an abstraction technique based on LPM discovery and pattern-based abstraction fully automated, further experimentation would be needed. We need to analyze whether the optimal number of LPMs depends on properties of the event log and for which logs the good results can be expected.

### 8.7.4 Concluding Remarks on LPMs for Event Abstraction

In this section, we have shown how LPMs can be used together with the pattern-based abstraction technique of [Man+16b; Man+18] as a fully automated and unsupervised approach to abstract event logs into higher-level event logs. We have shown on five real-life event logs that the abstraction approach applied prior to process discovery can result in more precise process models compared to the process models that were discovered on the original low-level event logs.

We found that (1) the number of LPMs that should be used for abstraction, (2) the diversity threshold, and (3) the composition method which result in good process models being discovered are very dependent on the event log on which the technique is applied. In future work, we want to investigate this interplay between event log properties and the parameters of the abstraction approach that are needed to discover process models that strike a good balance between precision and fitness. A second direction for future work is to investigate the suitability of other algorithms that we will propose in Chapter 12 to select a non-redundant set of LPMs from a collection of LPMs for the purpose of selecting LPMs for event log abstraction.

## 8.8 Limitations of Local Process Models

We have proposed several improvements over the naive expansion function $exp_0$, resulting in expansion function $exp_2$. However, expansion function $exp_2$ still generates language equivalent process models (i.e, it does not satisfy Requirement 5). A consequence of this is the large search space of LPM expansions, which results in high computational complexity. An open challenge is to develop more efficient expansion functions that are able to satisfy Requirement 1 while preventing language equivalent process trees, or at least reducing the generation of language-equivalent process trees with respect to $exp_2$.

Furthermore, the current procedure to count the number of instances of an LPM in an event log is based on alignments, which is known to be of exponential time complexity in the length of the trace as well as in the number of activities in the log. This can be seen as a limitation of the approach. We consider it to be an open challenge and an interesting direction for future work to develop faster approaches to count/identify the instances of an LPM in an event log. Such approaches do not necessarily need to be dependent on alignments.

Finally, we have only explored LPMs that are expressed as process trees. Each type of process model notation comes with certain types of process behavior that can and some that cannot be described in that notation, which is commonly referred to as the *representational bias* of the process model notation. By choosing process trees as the underlying representation in the mining procedure of LPMs we rule out the possibility of some types of behavioral constructs that can just inherently not be expressed in the form of a process tree. Two examples of behavior that cannot be expressed by process trees are *long-term dependencies* and the *milestone pattern* (for an extensive overview of types of behavioral patterns we refer to [Aal+03]). An interesting direction for follow-up work would be to develop techniques for mining LPMs in a fundamentally different way, by using a different representation than process trees, thereby enabling LPMs that express other types of behavior that can currently not be discovered.

## 8.9 Conclusion

This chapter presents a method to discover local process models (LPMs) that can express the same rich set of relations between activities as business process models, but describe frequent fragments instead of complete start-to-end processes. We presented five quality criteria and corresponding measures that quantify the degree of representativeness of an LPM with respect to an event log. Furthermore, we defined a search space based process tree expansion rules that significantly reduces the degree of redundancy in the search space compared to a naive definition of the search space. Further improvement in the computational efficiency of LPM mining has been obtained by using properties of the quality measures that allow us to prune the search space. We have illustrated in two real-life case studies that the proposed method enables the user to obtain process insight in the form of valuable patterns when the degree of randomness/variance of the event data prevents traditional process discovery techniques to discover a structured start-to-end process model.

In the next chapter, we will present a generalization of LPMs where the usefulness of LPMs is allowed to be dependent on data attributes of the events that fit the LPM, thereby generalizing beyond frequency-based selection of LPM patterns.

The computational time that is required to mine LPMs rapidly grows with the number of activities in the event log. Therefore, in Chapter 10, we consider automatic discovery of projections on the event log. Such projections limit the search space to a promising subset of the activities, thereby enabling the mining of LPMs on logs with larger numbers of activities.

# 9 Extending Local Process Models with Utility Functions

**Parts of this chapter have been published as:**

- Niek Tax, Benjamin Dalmas, Natalia Sidorova, Wil M. P. van der Aalst, and Sylvie Norre: *Interest-Driven Discovery of Local Process Models*. Information Systems 77: pp. 105–117 (2018)

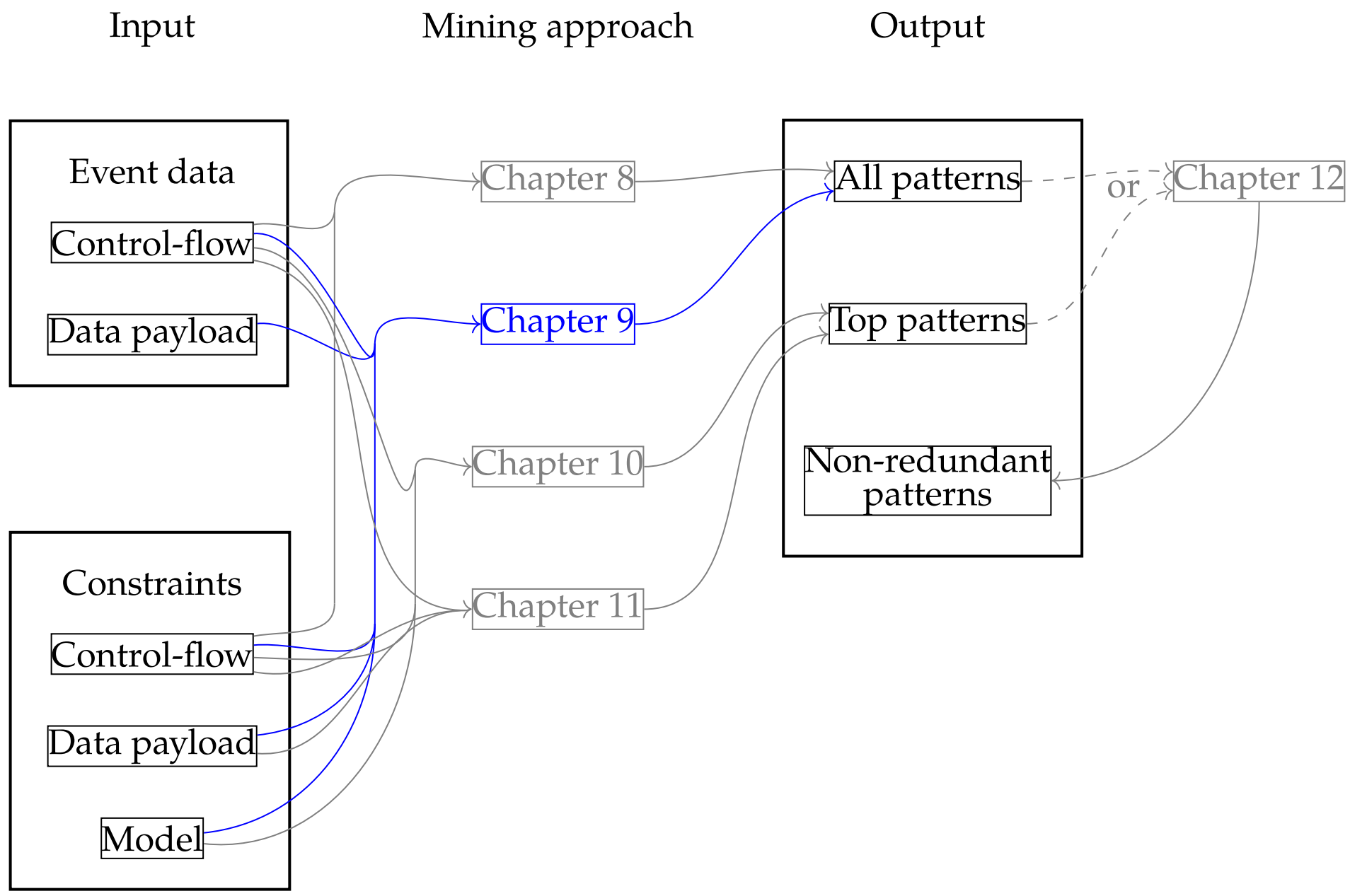

**Figure 9.1:** A taxonomy of local process model techniques.

The Local Process Model (LPM) mining technique that we proposed in Chapter 8 selects and ranks LPMs simply based on how frequent the behavior of the pattern

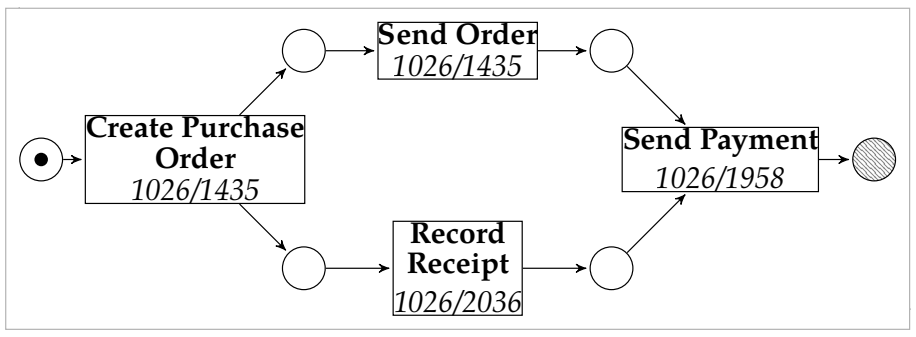

**(a)** (a)

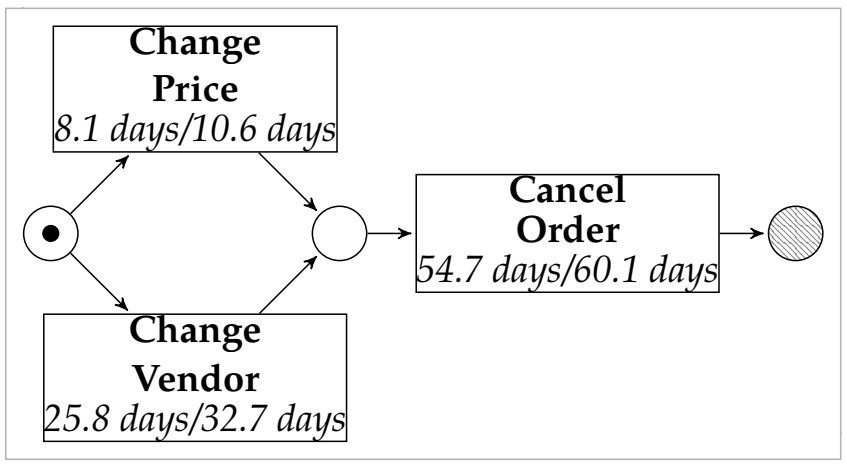

**(b)** (b)

**Figure 9.2:** *(a)* A frequent Local Process Model with low utility, and *(b)* a non-frequent Local Process Model with high utility.

is in the log. The support, confidence, language fit, and coverage quality measures that we proposed there are all derived from the number of observations of a pattern in the log. However, in the context of a business process, a process analyst might, in fact, be more interested in patterns of behavior that are somehow financially expensive, of which the execution is time-consuming, or that are in any other sense of special business interest.

A recent trend [Lin+15] in the frequent pattern mining field is to incorporate *utility* into the pattern selection framework, such that not just the most frequent patterns are discovered, but instead patterns are discovered that address typical business concerns, such as the patterns that represent high financial costs. Shen et al. [SZY03] were the first to introduce utility-based itemset mining. Since then, utility-based pattern mining has spread to different types of pattern mining, including sequential pattern mining [YZC12]. Utility-based pattern mining techniques assume that the value of each data point with respect to a certain business question is known and then discover the optimal patterns in terms of the value that they represent. In this chapter, we will consider the mining of *high-utility patterns* in the context of LPM patterns.

To motivate the need for utility-based patterns in LPM mining, imagine a Purchase-to-Pay process in which we take up the role of a process analyst who is interested in mining the fragments of the business process the employees of the company spend most of their time on. Figure 9.2 shows two LPMs that could be discovered from such an event log. Figure 9.2a is a process fragment that describes the creation of a purchase order, which is followed by both the sending of the order and the recording of the receipt in an arbitrary order, finally followed by a payment being sent. This process fragment is frequent, as indicated by the two numbers in each rectangle, indicating respectively the number of logged events that fit the behavior described by the LPM and the total number of events of these activities. Figure 9.2b describes a process fragment where the order is canceled after the price or the vendor of the order is changed. Even though the process fragment of Figure 9.2b is infrequent, it consumes considerable resources from the department, as the cancellation of an order is an undesired action and considerable time will be spend trying to prevent it. The support-based LPM mining approach of Chapter 8 would not be able to discover Figure 9.2b because of its low frequency, motivating the need for utility-based LPM discovery.

In this chapter, we propose a framework to discover LPMs based on their utility in the context of a particular business question. This chapter is organized as follows. Section 9.1 introduces utility functions and constraints in the context of LPMs. In Section 9.2 we demonstrate utility-based LPM discovery on three real-life event logs and show that we can obtain actionable insights. Section 9.3 describes related work. We conclude this chapter in Section 9.4.

## 9.1 Local Process Model Constraints and Utility Functions

The discovery of Local Process Models (LPMs) can be steered towards the business needs of the process analyst by using a combination of *constraints* and *utility functions*. Constraints can, for example, be used to find fragments of process behavior that lead to a loan application getting declined, or to find fragments that only describe loan applications above €15 K and never those below. Utility functions can be used to discover LPMs that give insight in which fragments of process behavior are associated with high financial costs or long time delays.

Constraints are requirements that the LPM has to satisfy, therefore, we define constraints as functions that result in 1 when the requirement holds and is 0 when it does not hold. In a general sense, they are defined as a function $c : X \rightarrow \{0, 1\}$ where X is the *scope* on which the function operates. We distinguish four different scopes on which X can be defined: *trace-level* (T), *event-level* (E), *activity-level* (A), and *model-level* (M). The class diagram in Figure 9.3 contains annotations which indicate which classes are included in each of the scopes.

Utility functions indicate to what degree an LPM is expected to be interesting and helpful to answer the business question of the process analyst at hand, and are defined as functions $f : X \rightarrow \mathbb{R}$. Like constraints, utility functions can be defined on the four scopes indicated in the class diagram of Figure 9.3. Where constraints can be used to formulate hard requirements, i.e., LPMs that do not fulfill them are deemed uninteresting, utility functions can be used to formulate soft preferences.

Multiple utility functions and constraints can be combined to form one composite function that describes the total utility of an LPM given log L. Given utility functions $f_1, f_2, \ldots, f_k$, an LPM is deemed more useful the higher it scores on utility functions $f_i$. Additionally, given constraints $c_1, c_2, \ldots, c_n$, and LPM needs to satisfy all constraints $c_i$ in order to be considered interesting in the first place. Where the original LPM discovery method of Chapter 8 selects and ranks LPMs based on support, utility allows for goal-oriented selection and ranking, i.e., LPMs are ranked based on their utility.

In Chapter 8, we assessed the quality of an LPM with respect to a labeled event log. Utility-based LPMs generalize the concept of LPMs to unlabeled event logs, thereby enabling the use of the data payload of events and traces in the utility function definition. However, event labels are still needed in order to represent the transitions of LPM patterns, i.e., the behavioral pattern of an LPM is still specified on a labeled log and it is merely the utility of that LPM that depends on the data payload. Therefore, utility functions and constraints of the trace, activity, and event-level scopes are all defined on a combination of an LPM, an unlabeled event log $U \in \mathcal{C}(\mathcal{E})$, as well as a labeling function $l$. They define the utility of an LPM as a function over the data payload of the events in U that fit the LPM.

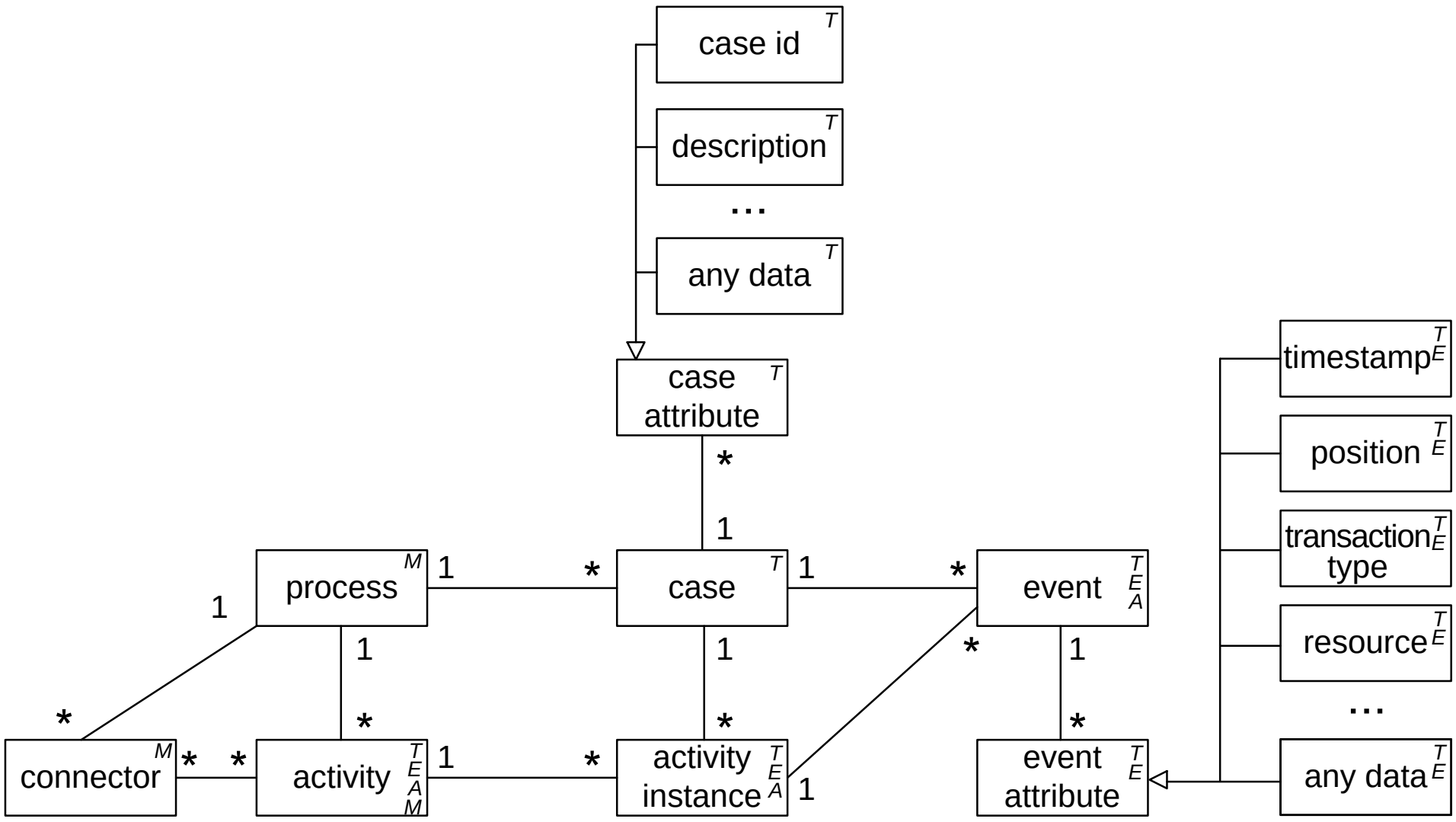

**Figure 9.3:** Basic logging concepts conceptualized in a class diagram. (Based on [Aal16]).

In order to obtain the fitting segments of an *unlabeled event log* $U \in \mathcal{C}(\mathcal{E})$ on LPM *LPM* with the data attributes that are given by event universe $\mathcal{E}$, we define function $\eta$ that determines the fitting segments of U for some labeling function $l$ by determining the fitting events of *labeled event log* $l(U)$ on *LPM* using $\Gamma_{LPM}(l(U))$ and then mapping the landmarks of the instances back to the corresponding traces in U.

**Definition 9.1 (Mapping landmarks to unlabeled traces).** Given a local process model *LPM*, unlabeled event log $U \in \mathcal{C}(\mathcal{E})$, an labeling function $l$, $\eta_{LPM}(U, l) = \{\xi(\sigma, \lambda) \mid \sigma \in U \wedge (l(\sigma), \lambda) \in \Gamma_{LPM}(l(U))\}$. ◇

For example, for an unlabeled event log $U = \{\sigma_1^u = \langle e_1^1, e_2^1, e_3^1, e_4^1, e_5^1 \rangle, \sigma_2^u = \langle e_1^2, e_2^2, e_3^2, e_4^2 \rangle, \sigma_3^u = \langle e_1^3, e_2^3, e_3^3, e_4^3 \rangle\}$ and some labeling function $l$ such that $l(U) =$ $[\sigma_1^l = \langle a, c, b, d, e\rangle, \sigma_2^l = \langle a, d, b, c\rangle^2]$ in a way that the unlabeled traces $\sigma_2$ and $\sigma_3$ become the identically labeled trace variant $\sigma_2^l$, and for some LPM $LPM$ with $\mathfrak{L}(LPM) = \{\langle a, b, c\rangle, \langle a, c, b\rangle\}$ such that $\Gamma_{LPM}(l(U)) = [(\sigma_1^l, \langle 1, 2, 3\rangle), (\sigma_2^l, \langle 1, 3, 4\rangle)^2]$, we have $\eta_{LPM}(U, l) = \{\langle e_1^1, e_2^1, e_3^1\rangle, \langle e_1^2, e_3^2, e_4^2\rangle, \langle e_1^3, e_3^3, e_4^3\rangle\}$.

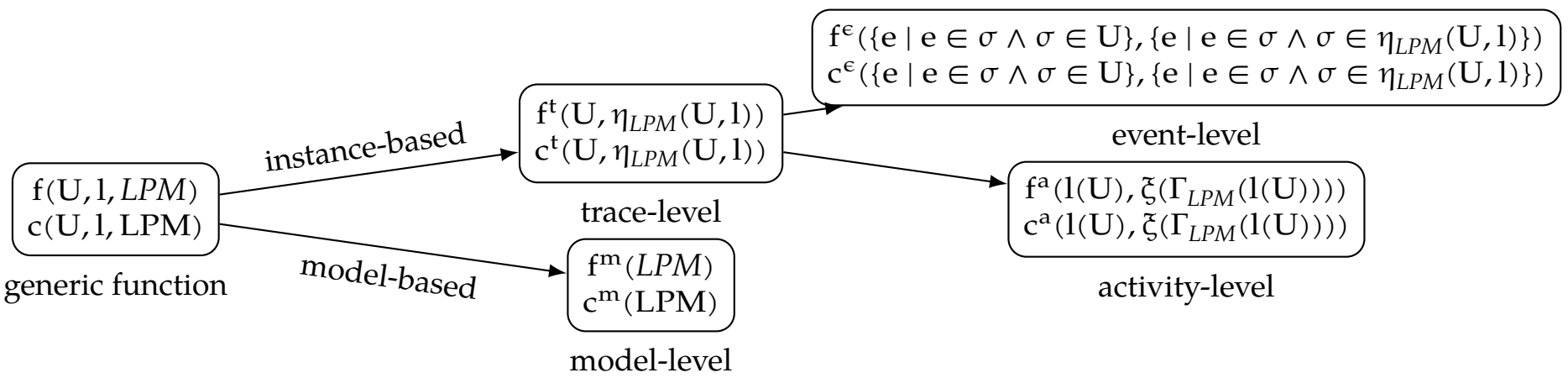

**Figure 9.4:** A taxonomy of utility and constraint function specifications on different scopes.

The trace, event, and activity-level scopes differ in the perspectives on log U and the instances $\eta_{LPM}(U, l)$ on which they operate. The model-level scope is defined solely on the LPM itself, and ignores the log argument U. It depends on the business question of the process analyst which concrete utility/constraint functions should be used. Figure 9.4 shows a taxonomy of the four scopes for constraint and utility functions, and describes the arguments of the functions at each of the four scopes. In the sections that follow we discuss definitions and properties of constraints and utility functions on each of these scopes and discuss several of their use cases.

As a running example throughout the descriptions of each of the scopes, we take the case that we had already introduced at the start of this chapter: a process analyst wants to analyze her Purchase-to-Pay process, where she is specifically interested in the stages in the process where high numbers of man-hours are spent by the office workers. She considers each stage in the business process to be a sequence of process steps that is executed within a certain amount of time. Her starting point is an event log consisting of logged executions from the process where the total number of man-hours that was spent is logged for each event in the process. In particular, the process analyst is interested in unnecessary re-work, i.e., activities or sequences of activities that are performed more than once for a single case. Therefore, she focuses her analysis on patterns that contain a loop. Some of the logged steps in the process are automated steps that are performed by the supporting information system, without any involvement of an office worker. Since no man-hours are involved in such automated steps, the process analyst does not want to include them in her analysis.

## 9.1.1 Trace-level Constraints and Utility Functions

Trace-level utility functions are the most general class of utility functions that are defined on the event log. Trace-level utility functions calculate the utility of an LPM

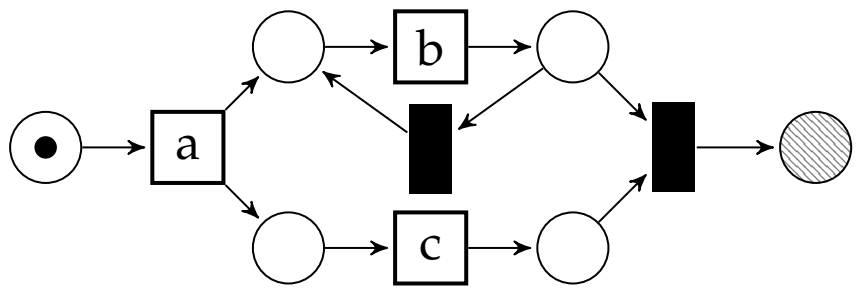

**(a)**

| event id | activity | time | cost |
|---|---|---|---|
| 1 | a | 26-3-2017 13:00 | €100 |
| 2 | b | 26-3-2017 13:25 | €500 |
| 3 | x | 26-3-2017 13:27 | €60 |
| 4 | b | 26-3-2017 13:30 | €400 |
| 5 | c | 26-3-2017 13:35 | €100 |
| 6 | c | 26-3-2017 13:42 | €500 |
| 7 | a | 26-3-2017 15:26 | €300 |
| 8 | b | 26-3-2017 15:27 | €50 |
| 9 | c | 26-3-2017 15:47 | €100 |
| 10 | b | 26-3-2017 16:10 | €250 |
| 11 | b | 26-3-2017 16:52 | €300 |
| 12 | x | 26-3-2017 16:59 | €10 |

**(b)**

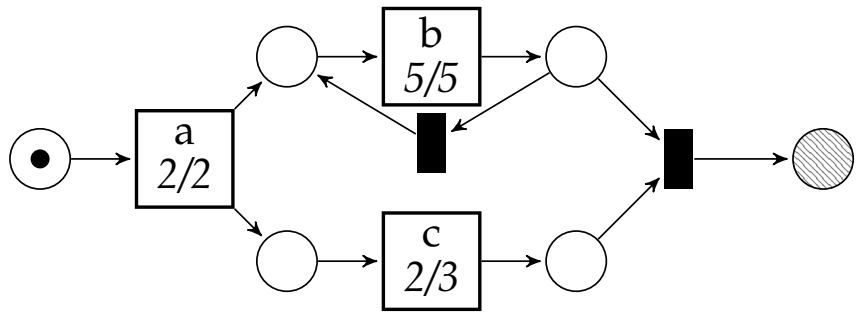

**(c)**

**Figure 9.5:** *(a)* Example accepting Petri net $APN_1$. *(b)* Trace $\sigma$ of an event log L. *(c)* LPM $LPM$ obtained by scoring $APN_1$ on $\sigma$.

by aggregating the utility over the trace-fragments that fit the LPM behavior and allow the utility of a fitting trace fragment to depend on the events in the trace fragment, their event properties, and properties of the case itself. Trace level-utility can, for example, be used to discover LPMs that describe the events that explain a high share of the total financial cost associated to a case, or a high share of the total running time of a case.

A trace-level utility function is a function $f^t(U, \eta_{LPM}(U,l))$ that indicates the utility of the instances of $LPM$ in $\eta_{LPM}(U,l)$. Consider the unlabeled event log U in Figure 9.5b that consists of just one trace $\sigma$. Furthermore, consider the process model M in Figure 9.5a. Calculating maximal non-overlapping instances of $LPM$ on the labeled log $l(U)$ that is obtained by applying labeling function $l = \pi_{\text{activity}}$ to U yields, $\xi(\Gamma_{LPM}(l(U))) = [\langle a,b,b,c\rangle, \langle a,b,c,b,b\rangle]$, and therefore, the instances of the unlabeled log U are $\eta_{LPM}(U,l) = \{(\sigma, \langle e_1, e_2, e_4, e_5\rangle, \langle e_7, e_8, e_9, e_{10}, e_{11}\rangle)\}$.

Assume that $\sigma$ has a case attribute *total_cost* that indicates the total cost of the trace. An example of a trace-level utility function is:

$$f^t_1(U, \eta_{LPM}(U,l)) = \sum_{\sigma \in \eta_{LPM}(U,l)} \sum_{e \in \sigma} \frac{\pi_{cost}(e)}{\phi_{\text{total_cost}}(\sigma)} \qquad (9.1)$$

$f^t_1$ discovers LPMs that explain a large share of the total trace costs. Another

example is the following function:

$$f_2^t(U, \eta_{LPM}(U, l)) = \sum_{\sigma \in \eta_{LPM}(U,l)} \pi_{time}(\sigma(|\sigma|)) - \pi_{time}(\sigma(1)) \quad (9.2)$$

$f_2^t$ results in LPMs where the behavior described typically occurs in long time intervals. Trace-level constraints put requirements on the instances of the unlabeled log, i.e., on $\eta_{LPM}(U, l)$. All trace-level utility functions can be transformed into trace-level constraints by adding thresholds for a minimal or maximal value of the function. For the Purchase-to-Pay running example, the requirement of the process analyst that the average duration of the instances of a resulting LPM is less than what he considers to be the maximal time for a stage (i.e., *max_time*) can be modeled using the following trace-level constraint:

$$c^t_{example}(U, \eta_{LPM}(U, l)) = \frac{f_2^t(U, \eta_{LPM}(U, l))}{|\eta_{LPM}(U, l)|} \leq max_time \quad (9.3)$$

## 9.1.2 Event-level Constraints and Utility Functions

Event-level utility functions and constraints can be used when the business question of the process stakeholder concerns certain event properties, but does not concern the trace-context of those events. Such utility functions can e.g. be used to discover LPMs that describe process behavior fragments with high financial costs. Constraints on this level can, for example, be used to limit LPMs that solely describe events that are executed by certain resources, or, in certain time periods. The trace-level scope is more expressive than the event-level scope, allowing the analyst to formulate more complex utility functions and constraints, but at the same time, it makes it harder to formulate such functions compared to the more restricted event-level scope.

An event-level utility function is a function $f^\epsilon(S_1, S_2)$ with $S_1 = \{e \mid e \in \sigma \wedge \sigma \in U\}$ the set of unlabeled events in U and $S_2 = \{e \mid e \in \sigma \wedge \sigma \in \eta_{LPM}(U, l)\}$ the set of unlabeled events that are part of an instance of an LPM, i.e., part of $\eta_{LPM}(U, l)$. For the LPM, trace, and the instances of the trace in the LPM shown in Figure 9.5, an example event-level utility function is:

$$f_1^\epsilon(S_1, S_2) = \sum_{e \in S_2} \pi_{cost}(e) \quad (9.4)$$

Utility function $f_1^\epsilon$ discovers LPMs with high costs. Note that $\{e \mid e \in \sigma \wedge \sigma \in U\}$ itself is also in the domain, allowing us to formulate a utility function that optimizes the share of utility explained per activity.

$$f_2^\epsilon(S_1, S_2) = \sum_{a \in \Sigma} \frac{\sum_{e \in \{e | e \in S_2 \wedge \pi_{activity}(e) = a\}} \pi_{cost}(e)}{\sum_{e \in \{e | e \in S_1 \wedge \pi_{activity}(e) = a\}} \pi_{cost}(e)} \quad (9.5)$$

Since set of events $S_1$ contains all events in U and $S_2$ contains only those events that are part of an instance of the LPM, the utility function $f_2^{\epsilon}$ yields the ratio of the financial costs of the activities that are explained by the LPM.

For the Purchase-to-Pay running example, the desired utility of an LPM as the total man-hours spent can be modeled with the following event-level utility function:

$$f^{\epsilon}_{example}(S_1, S_2) = \sum_{e \in S_2} \pi_{man_hours}(e) \tag{9.6}$$

An event-level constraint is a function $c^{\epsilon}(S_1, S_2)$ with $S_1 = \{e \mid e \in \sigma \wedge \sigma \in U\}$ the set of unlabeled events in U and $S_2 = \{e \mid e \in \sigma \wedge \sigma \in \eta_{LPM}(U, l)\}$ the set of unlabeled events that are part of an instance of an LPM, i.e., part of $\eta_{LPM}(U, l)$. An example is the following function:

$$c^{\epsilon}_1(S_1, S_2) = f^{\epsilon}_1(S_1, S_2) \geq 500 \tag{9.7}$$

Constraint $c^{\epsilon}_1$ results in LPMs with a total value of at least €500. An second example is:

$$c^{\epsilon}_2(S_1, S_2) = \forall_{e \in S_2} \pi_{cost}(e) \geq 100 \tag{9.8}$$

Constraint $c^{\epsilon}_2$ results in LPMs that never represent events with a value of less than €100. Note that $c^{\epsilon}_2$ does not hold for our example trace and LPM, where the event with id 8 fits the LPM but only has value €50. Therefore, this LPM will not be found by the LPM discovery technique when we use constraint $c^{\epsilon}_2$.

### 9.1.3 Activity-level Constraints and Utility Functions

Activity-level utility functions and constraints define the utility of an LPM based on the occurrences of each activity in the labeled log $l(U)$ and in $\Gamma_{L,LPM}$, i.e., they do not use the data payload of the events in the unlabeled log U. Activity-level utility functions can for example be used by the process analyst to specify that he is more interested in some activities of high impact (e.g., lawsuits, security breaches, etc.) than in others, resulting in LPMs that describe the frequent behavior before and after such events. With activity-level constraints, the process analyst can set a hard constraint on activity occurrences. The stakeholder or analyst can for example specify a function *relevance* : $\Sigma \rightarrow \mathbb{R}$ that indicates how interested he is in each activity of the log and then specify function $f^{a}(l(U), \xi(\Gamma_{LPM}(l(U))))$ that assigns the utility based on the *relevance* of the events that are part of the instances of *LPM* in $l(U)$. Note that when such as function *relevance* assigns equal importance to all activities, then the following activity-level utility function becomes equivalent to the support-based LPM discovery of Chapter 8:

$$f^{a}_1(l(U), \xi(\Gamma_{LPM}(l(U)))) = \sum_{\sigma \in \xi(\Gamma_{LPM}(l(U)))} \sum_{e \in \sigma} relevance(e) \tag{9.9}$$

Such utility functions can, for example, be used to get insight in the relations with other activities of some particular high-impact activities that the process analyst is interested in. Note that the occurrence of such activities in the log can be infrequent, in which case traditional LPM discovery without the use of utility functions and constraints is unlikely to return LPMs that concern those activities.

Adding a transition representing a zero-utility activity to an LPM can never increase its total utility, therefore, activity-level utility functions that assign zero value to some activities speed up discovery by allowing to limit the LPM expansion function *exp* to expand LPMs only using activities with non-zero utility, as activities with utility zero will never contribute to the overall utility of an LPM.

For the Purchase-to-Pay running example, an activity-level constraints can be used to model the desire of the analyst to focus his analysis on the non-automated activities of the process. Let *automated_activities* denote the set of automated activities in the Purchase-to-Pay process. The following activity-level constraint focuses the analysis on non-automated activities:

$$c^{a}_{example}(l(U), \xi(\Gamma_{LPM}(l(U)))) = \forall_{e \in \{e \in \Sigma | e \in \sigma, \sigma \in \xi(\Gamma_{LPM}(l(U)))\}} e \notin \mathit{automated_activities} \tag{9.10}$$

### 9.1.4 Model-level Utility Constraints and Utility Functions

Model-level utility functions and constraints can be used when a process analyst has preference or requirements for specific structural properties of the LPM. They have the form $f^{m}(LPM)$ and $c^{m}(LPM)$, i.e., they are independent of the log and dependent only on the LPM itself. When a process analyst, for example, wants to analyze the behavior that leads to the execution of a certain activity a which he is interested in, he can use a model-level constraint that enforces that all elements of $\mathfrak{L}(LPM)$ end with a.

Generally, we are interested in models that represent certain aspects of the process behavior based on the event log. Therefore, model-level utility functions and constraints are often not very useful on their own, but they become useful when combining them with utility functions and constraints on the event log, i.e., on the activity-level, event-level, or trace-level.

$\Gamma_{L,LPM}$ does not need to be calculated to determine whether a model-level constraint is satisfied, because model-level constraints are defined solely on the model. Therefore, model-level constraints can also be used to speed up LPM discovery by limiting the search space that is generated by expansion function *exp* to its subspace for which the model-level constraints hold. For the Purchase-to-Pay running example, the desired focus of the analysis on re-work activities can be modeled as a model-level constraint, by formulating a constraint $c^{m}_{example}$ that requires the resulting LPMs to contain a loop.

### 9.1.5 Composite Utility Functions

AS introduced informally in the beginning of this section, utility functions and constraints on the different levels can be combined into one single utility function. Formally, the total utility of an LPM is defined as the product of all utility functions and constraints. This results in the following equation, where for brevity we have omitted the function signatures for each of the four types of utility functions and constraints[26]:

$$u(U, l, LPM) = \prod_{i=1}^{n} c_i^t \times \prod_{i'=1}^{n'} c_{i'}^{\epsilon} \times \prod_{i''=1}^{n''} c_{i''}^{a} \times \prod_{i'''=1}^{n'''} c_{i'''}^{m} \times \sum_{j=1}^{k} f_j^t + \sum_{j'=1}^{k'} f_{j'}^{\epsilon} + \sum_{j''=1}^{k''} f_{j''}^{a} + \sum_{j'''=1}^{k'''} f_{j'''}^{m} \quad (9.11)$$

This has the effect that LPMs only have non-zero utility is they satisfy *all* constraints. If all constraints are satisfied, then the total utility u is defined as an unweighted sum over the individual utility functions $f_i$. Note that this still allows the process analyst to give priority to one utility function over another, as weights can be included as a part of the utility function itself by multiplying the utility function with a constant.

In the previous sections we have shown how we can model each of the elements of our Purchase-to-Pay analysis using the utility functions and constraints defined on the four scopes. To recap, we have modeled trace-level constraint $c^t_{example}$ that enforces the LPM to model behavior that is executed within a certain amount of time, $f^{\epsilon}_{example}$ assigns the utility to an LPM based on the man-hours that are spent by the office workers, $c^a_{example}$ restricts the analysis to non-automated activities, and $c^m_{example}$ focuses the analysis on re-work. Combined, the following utility function models the analysis question of the process analyst, where we have again omitted the function signatures for brevity:

$$u_{example}(L, LPM) = c^t_{example} \times c^a_{example} \times c^m_{example} \times f^{\epsilon}_{example} \quad (9.12)$$

The presented framework of utility functions and constraints generalizes the method described in Chapter 8 and all LPM quality measures that we had presented there are instantiations of utility functions. An example is the *support* measure, which is an activity-level utility function. The minimum threshold for support that we proposed in Chapter 8 is an example of an activity-level constraint. Another quality measure that we introduced is *determinism*, which is inversely proportional to the average number of enabled transitions in LPM during replay of the aligned event log. *Determinism* is an example of a composite utility function, consisting of a model-level and a trace-level component.

[26]See Figure 9.4 for an overview of the function signatures of the four types of constraints and utility functions.

The mining procedure of utility-based LPMs follows the same iterative expansion procedure for LPMs as introduced in Chapter 8. However, where the pruning of the state space was defined on support in Chapter 8, we now use an identical pruning procedure using a minimum utility value *minimum_utility* to define a minimum value for composite utility function $u(L, LPM)$. A more in-depth study into efficient heuristics to mine utility-based LPMs is outside the scope of the thesis, but can be found in [DTN17; DTN18]. In Chapter 11 we will discuss efficient mining strategies for LPM mining when using a specific type of constraints, namely, constraints on that are defined on the maximum time gap between two succeeding events in an instance of an LPM.

Note that the trace-level and event-level utility functions are not limited to continuous-valued properties. Many event logs contain ordinal event properties, such as *risk*, or, *impact*, which can take values such as *low*, *medium*, or *high*. Event-level utility functions can be applied to such event properties by specifying a mapping from the possible ordinal values to continuous values.

As a final note, we would like to remark that many information systems already register many trace-level and event-level data attributes in the event logs. Therefore, the applicability of the technique does not require any additional logging other than what is what is already being logged in practice. Traditional, non-utility-based LPM mining, however, does not make use of such data attributes.

9

## 9.2 Case Studies

In this section, we describe three case studies on real-life event logs. The first log originates from the traffic fine handling process by the Italian police. This event log is extensively used in literature for the evaluation of data-aware process mining approaches, enabling us to compare the insights obtained with our technique with the reported insights that were obtained with existing data-aware process mining techniques. The second log originates from an IT service desk of a large Dutch financial institution. The third log is an event log of sensor events in a smart home environment.

### 9.2.1 Traffic Fines

The road traffic fine management event log[27] is an event log where each case refers to a traffic fine. Each case starts with a *create fine* event, which has a property *amount* that specifies the amount of the fine. *Payment* events have a *paymentAmount* property, which indicates how much has been paid. Payment of the total fine amount can be spread out over multiple payments. Some fines are paid directly to the police officer when the fine is given, and some are sent by mail, in which case there is a *send*

[27] http://dx.doi.org/10.4121/uuid:270fd440-1057-4fb9-89a9-b699b47990f5

*fine* event. *Send fine* events have a property *expense*, which contains an additional administrative cost which adds to the total amount that has to be paid. When a fine is not paid in time, an *add penalty* event occurs which has an *amount* property that updates the fine amount set in the *create fine* event. If a fine is still not paid after the added penalty, it is *send for credit collection*. Furthermore, a fine can be appealed at the prefecture and, at a later stage, can be appealed in court. In total, the traffic fines event log contains 150,370 traces, 561,470 events, and 11 activities.

Assume we are a process analyst concerned with the business question: "which process fragments describe fines where the remaining amount to be paid is high?". Figure 9.6 shows the top three LPMs that we discovered from the traffic fine log using a trace-level utility function that defines utility as the remaining amount that is still to be paid at the time of the event. This utility function has a trace-level scope, as it is calculated from the latest seen *amount* in the case (either from a *create fine* or *add penalty* event) plus the *expense* fee if there is a *send fine* event minus the *paymentAmount* values of all *payment* events seen in the trace so far. In addition, we add a minimum *support* constraint to at least 10,000 pattern instances to make sure that the patterns that we find are not merely anomalies. In addition to showing the utility of the three LPMs, Figure 9.6 additionally shows their *support*, *confidence*, *language fit*, *determinism* and *coverage* values.

Mining the list of LPMs ranking on utility from the traffic fine event log that takes 39 minutes using an Intel i7 CPU @ 2.4GHz with 16 GB of memory. The first LPM in Figure 9.6 shows that in total €5.83 million has been created in fines, out of which €5.60 million was either send to credit collection or payments have been received for them. The remaining €200 thousand correspond to recent fines that have not yet been paid but are not yet due to be sent for credit collection. The total amount of payments received after a *create fine* event is €731 thousand, which is surprisingly low, comparable to the total of €5.83 million. Noteworthy is that fines representing a total value of €6.43 million are sent to credit collection, which is even more than the fines representing a value of €5.83 million that were created in the first place. That the total value of fines at credit collection is higher than the total value of fines that were handed out is because of added penalties and, to a lesser degree, added expenses. Finally, fines representing a value of €6.32 million that were send to credit collection out of the total €6.43 million value of fines that were send to credit collection fit the control flow pattern of the LPM, i.e., they occur after a *create fine* and are in an XOR-construct with *payment* events. Since we know that all traces start with a *create fine* event, the only possible explanation left is that fines representing a value of €6.32 million that were sent to credit collection have not received a single payment before being sent to credit collection, while for the remaining fines representing a value of €110 thousand that were sent to credit collection at least one partial payment of the fine was already received.

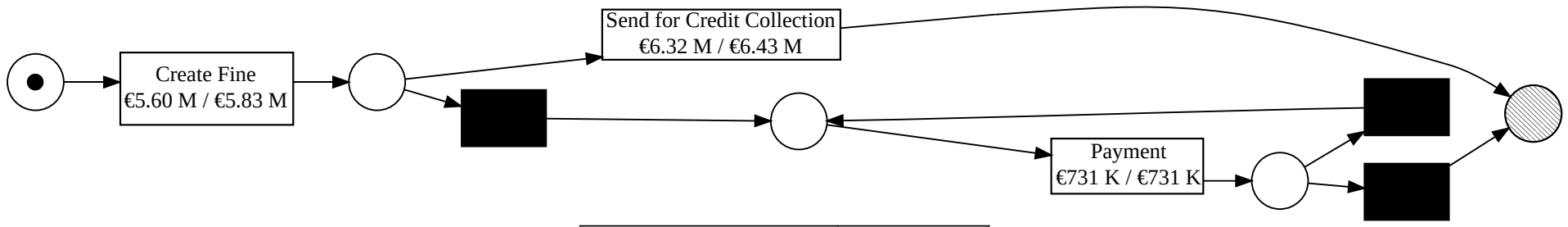

| | |
|---|---|
| Support | 126602 |
| Confidence | 0.9302 |
| Language fit$_5$ | 1.0000 |
| Determinism | 0.8211 |
| Coverage | 0.4735 |

**(1)** total utility= €12.6 million

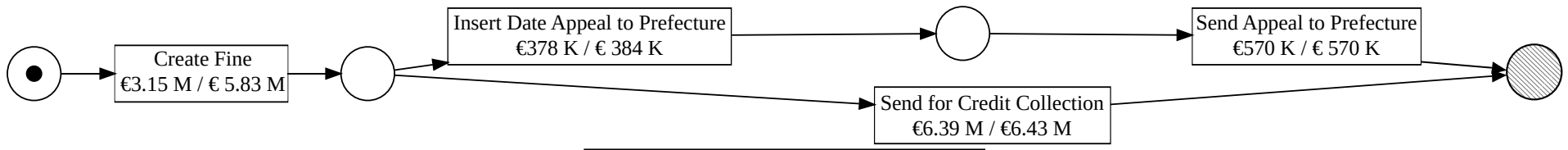

| | |
|---|---|
| Support | 62868 |
| Confidence | 0.7394 |
| Language fit$_5$ | 1.0000 |
| Determinism | 0.6738 |
| Coverage | 0.2463 |

**(2)** total utility= €10.5 million

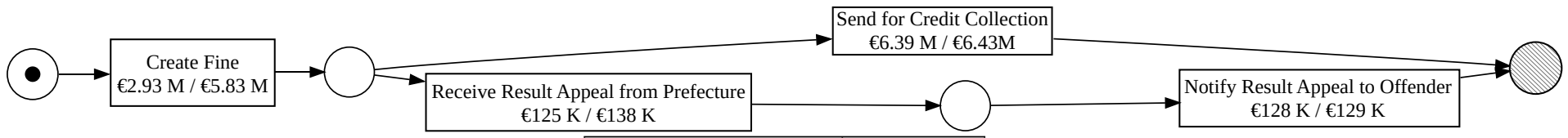

| | |
|---|---|
| Support | 59623 |
| Confidence | 0.7084 |
| Language fit$_5$ | 1.0000 |
| Determinism | 0.6683 |
| Coverage | 0.2109 |

**(3)** total utility= €9.6 million]

**Figure 9.6:** The top three LPMs in terms of total utility as discovered from the traffic fine log, using a *remaining amount to pay* utility function.

The second LPM in Figure 9.6 shows that fines representing a value of €3.15 out of the total value of fines of €5.83 million was either sent to credit collection or appealed at the prefecture. The appeal procedure starts with an *insert data appeal to prefecture* event which is later followed by a *send appeal to prefecture* event. The LPM shows that for the appealed fines, the total value is €378 thousand at the time of *insert date appeal to the prefecture*, but added penalties for late paying raise the total appealed amount to €570 thousand at the time that the appeal is actually send to prefecture. This means appealed fines on average increase by a factor of 1.5 in value during the appeal procedure as a result of penalties being added for the payment term being overdue. Finally, the numbers in the *send for credit collection* transition show that only a very small portion of the total value of fines that were sent to credit collection were followed by an appeal procedure.

The third LPM in Figure 9.6 shows a pattern similar to the second LPM, with the difference that it now describes two later steps in the *appeal to prefecture* procedure. At the *receive result appeal from prefecture* step there is a total value of € 138 thousand. This is considerably lower than the fines representing values of € 378 thousand and € 570 thousand respectively that we found for the earlier steps of the appeal procedure. This shows that there are appealed fines representing a value of € 570 thousand - € 138 thousand = € 432 thousand which did not receive the appeal results. These appeals were either withdrawn before the verdicts on these appeals were made, or these appeals are still waiting for a verdict. Furthermore, fines representing a value of € 125 thousand out of the total value of € 138 thousand of fines at *receive result appeal from prefecture* fit the control flow of the pattern, indicating that fines representing a value of €13 thousand which received a result for appeal were either not (yet) notified, or they have been *sent for credit collection* prior to the appeal.

Note that while the support of the second and third LPMs are considerably lower than the support of the first LPM, the second and third LPM still have comparable financial impact compared to the first LPM.

*Comparison with Existing Techniques*

This same event log was used for the evaluation of three other data-aware process discovery techniques: the MP-Declare Miner [Sch+16b], the Data-Aware Heuristics Miner [Man+17], and the Multi-Perspective Process Explorer [MLR15]. Three insights into the process were obtained with the MP Declare Miner as reported in [Sch+16b]: *(1) "In 74% of the cases where a penalty was given, the case was sent for credit collection."*, *(2) "When add penalty was performed, the penalty amount had a value between €470 and €795."*, and *(3) "The higher the penalty amount is, the lower the probability that the fine is paid is."*. Note that these three insights are different types of insights than the insights obtained with utility-based LPM mining: while the insights mined with the MP-Declare miner state correlations between data and control-flow, the LPMs also show insights in the *change of the value of data attributes* between different steps in the control-flow (e.g., the change of the total amount of the fines over different steps in the process). Note that correlations between data and control-flow could easily be discovered in LPMs as a post-processing procedure using standard decision mining techniques [Man+16a].

The following three insights were discovered with the Data-Aware Heuristics Miner as reported in [Man+17]: *(1) "The occurrence of the activity* Add Penalty *mainly depends on the value of the dismissal attribute. Cases with values G do not receive a penalty, whereas cases with value* NIL *receive a penalty depending on the fine amount."*, *(2) "Unpaid fines that have with a small amount of less than €35 receive a penalty."*, and *(3) "The process ends after* Notify Result Appeal to Offender *for cases with dismissal value of # or G, since those cases are dismissed by the prefecture."*. Finally, the single reported insight that extracted with the Multi-Perspective Process Explorer [MLR15] is the following: *"The activity* Send for Credit Collection *occurs for fines with an amount > €71."*. Note that, like the insights mined with the MP-Declare Miner, the insights mined with the Data-Aware Heuristics Miner and the Multi-Perspective Process Explorer both mine insights that concern the relation between data attributes and control-flow or the other way around, these three insights are different types of insights than the insights obtained with utility-based LPM mining: while the insights mined with the MP-Declare miner state correlations between data and control-flow (or the other way around), while utility-based LPMs, depending on the formulation of the utility function, can also be used to get insights into the change of the data value over the different process steps.

### 9.2.2 IT Service Desk

The IT service desk event log[28] is an event log that was made publicly available as part of the *Business Process Intelligence Challenge 2014*. The data set contains *incident* events which represent disruptions of IT-services within a large financial institution. Each incident event is associated with one or more *interactions*, which represent the calls and e-mails to the service desk agents that are related to this incident. When an incident occurs, it is assigned to an operator, who either solves the issue, or reassigns it to a colleague having more knowledge. For each incident event several properties are recorded, amongst others:

**Service Component WBS** This is a number that identifies the service component involved in the incident.

**Configuration Item** This contains the type (i.e., laptop, server, software application, etc.) of the service component that the incident concerns. Each *service component WBS* belongs to one *configuration item*.

**Impact** The impact of service disruption to the customers as assessed by the operator. This property takes integer values from 1 to 5.

**Closure code** A code which classifies the cause of the service disruption, e.g., user error, software error, hardware error.

**CausedBy** Incidents of a service component have another service component as root cause of the service disruption. This field contains the *service component WBS* number of the root cause service component.

**Number of interactions** The number of calls to the IT service desk that are related to this incident.

[28] http://dx.doi.org/10.4121/uuid:c3e5d162-0cfd-4bb0-bd82-af5268819c35

**Number of reassignments** The number of times that this incident was reassigned from one IT service desk operator to another.

To create an event log consisting over multiple traces we group together events by their *service component WBS* number, creating one case per service component consisting of incidents that these service components are involved in. We set the activity label of each incident event to a combination of the *closure code* and the *causedBy* attribute separated by the pipe character (|). The resulting event log contains 313 traces, 25,262 events and 944 activities.

Assume now that we take up the role of a process analyst concerned with the business question: "which process fragments are related to high numbers of e-mails and phone calls to the IT service desk?". To answer this question we formulate the following utility function:

$$f^{e}(events(L), events(\Gamma_{L,LPM})) = \sum_{e \in events(\Gamma_{L,LPM})} \pi_{number_of_interactions}(e) \quad (9.13)$$

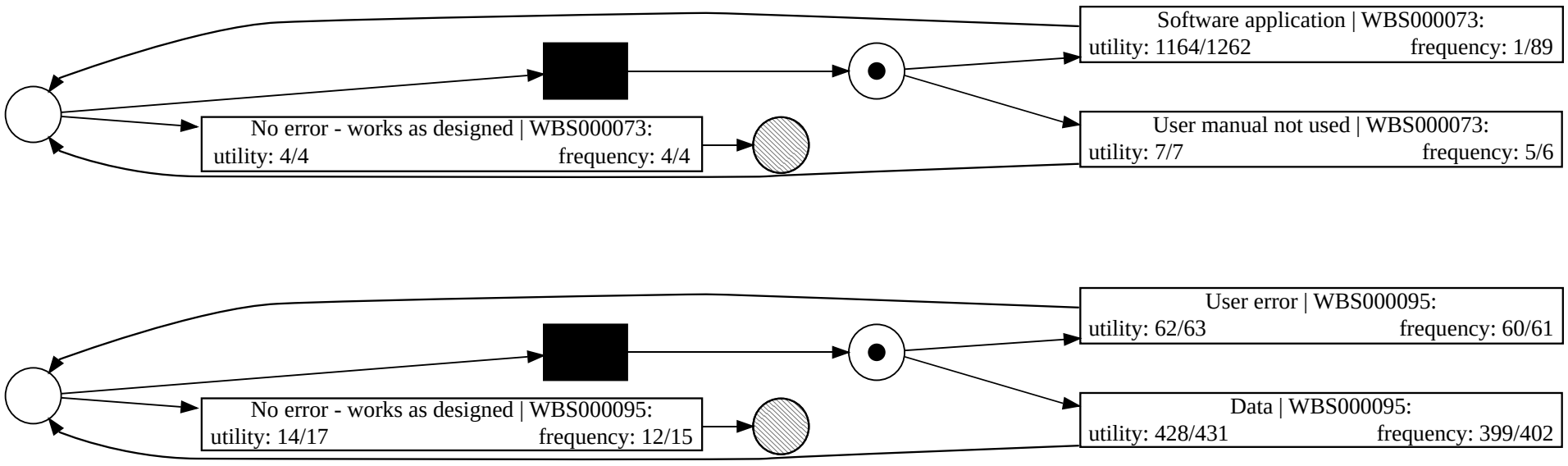

**Figure 9.7:** Two LPMs discovered from the BPI'14 event log when using the *number of interactions* event property as utility function.

Figure 9.7 shows the two LPMs with the highest utility that we discovered from the IT service desk event log using this utility function. Mining the full list of LPMs ranked on utility from this event log takes 3 minutes using an Intel i7 CPU @ 2.4GHz with 16 GB of memory.

The total utility of the top LPM of Figure 9.7 is 1175, indicating that the behavior of this LPM explains 1175 calls and e-mails to the IT service desk. The LPM shows that in total 1262 calls and e-mails to the IT service desk were associated with incidents with closure code *software application* that were caused by *WBS000073*, 1164 of which were associated with such an incident that fits the LPM. Only one single incident with this closure code and causedBy number out of the total 89 in the log fit the LPM behavior, however, this single incident had a large impact as it caused 1164 of the 1262 interactions. Finally, the LPM ends with an incident with closure code *No error - works as designed*. This indicates that this software application contains a feature that is working properly according to the IT service desk ("works as designed"), while bank employees perceive it as a problem, resulting in a lot of traffic to the IT service desk. Because only one single *software application | WBS000073* incident fits the behavior of this LPM, one could say that it describes an anomaly rather than a pattern. Note that the discovery of such anomaly-type LPMs is a result of the way we defined our utility function, which allows for highly skewed distribution of utility over events.

The second LPM shows a similar pattern, but it concerns incidents caused by server-based software application *WBS000095*. This LPM has a total utility of 504, meaning that it describes in total 504 calls and e-mails to the IT service desk. The model shows that 60 out of 61 *user errors* are eventually followed by an incident with closure code *No error - works as designed*. These 60 incidents together generate 62 interactions with the IT service desk. Note that 399 data incidents related to this software application, causing 428 interactions with the service desk, also resulted in an incident with this closure code.

In the two LPMs we see that two service components (*WBS000073* and *WBS000095*) are the main cause of call and e-mails to the IT service desk. Furthermore, for both service components, the LPMs end in *No error - works as designed* events, which could indicate that such problems could be prevented.

### 9.2.3 Smart Home

The process mining field has recently broadened its scope from analyzing business processes to other areas, including software analysis [LA15] and the analysis of human behavior [LMM15; Szt+15; Tax+16a; Tax+18f]. Applications of process mining on human behavior data is so far mostly based on discovery techniques that generate models without formal semantics [LMM15; Szt+15], while discovery techniques that generate models with formal semantics require extensive preprocessing of the event log to enable the mining of well-fitting models of human behavior [Tax+16a; Tax+18f]. In this case study, we focus on the analysis of human behavior with LPMs using constraints, enabling the mining of models with formal semantics without requiring extensive log preprocessing.

Tapia et al. [TIL04] collected event data from two households using a collection of *open/close sensors*, *pressure sensors*, and *power sensors*, with the goal of identifying *activities of daily living (ADL)* [Kat83]. The event logs from the two households are respectively referred to as *MIT A* and *MIT B*. Each case contains the one day of household sensor data, in which each event represents a trigger of a sensor where the activity label represents the sensor that generated the event. In this case study we focus on the *MIT A* event log, which consists of 16 cases (days), 2772 events (sensor triggers), and 72 activities (sensors).

Figure 9.8a shows the top three LPMs discovered from the MIT A event log when we use the traditional LPM mining approach that ranks LPMs based on *support*

**Table 9.1:** The *exhaust fan* and *kitchen drawer 3* events in the trace for the 4th of April 2003 of the MIT A event log.

| activity | time |
|---|---|
| Exhaust fan | 04-04-2003 07:23:14 |
| Kitchen drawer 3 | 04-04-2003 11:07:18 |
| Kitchen drawer 3 | 04-04-2003 13:48:39 |
| Kitchen drawer 3 | 04-04-2003 14:52:46 |
| Kitchen drawer 3 | 04-04-2003 17:49:08 |
| Kitchen drawer 3 | 04-04-2003 17:49:15 |
| Kitchen drawer 3 | 04-04-2003 17:52:13 |

and *confidence*, using a *support threshold* of 10, i.e., LPMs in the resulting list of LPMs a required to have at least 10 instances in the log. The first LPM shows that 61 out of 79 *toilet flushes* in the log are followed within the same day by getting hot water from the *sink faucet*, and 156 out of 169 of those *toilet flushes* fit this pattern. The second LPM shows that 28 out of 34 *exhaust fan* events are followed by a looping behavior of opening the *third drawer in the kitchen*. On average, each *exhaust fan* event is followed by 3.5 (=98/28) events of the *third kitchen drawer* in the same day. The third LPM shows that 28 out of the 34 *freezer* events are followed by a loop of events of the *first bathroom cabinet*. While the first LPM intuitively seems to make sense, as people tend to wash their hands after visiting the toilet, the sensors described by the other two LPMs seem to be unrelated. Table 9.1 shows the *exhaust fan* and *kitchen drawer 3* events of one of the days of the MIT A log. The sequence of events shown in the table together forms one instance of the second LPM in the ranking. However, the time gaps between the events seem to be so large that there does not seem to be any relation between the *exhaust fan* (at 07:23) and the *kitchen drawer* (first occurrence at 11:07). Therefore, the only information communicated by this LPM seems to be that events of the *exhaust fan* tend to occur mostly early in the morning and that *kitchen drawer 3* events are frequent.

The problem of unrelated activities ending up in an LPM can be mitigated by setting a *time gap constraint* on the maximum time between two consecutive events that fit the LPM in the log, and by setting a *total time constraint* on the maximum total duration of an instance of an LPM in the log from start to end. Figure 9.8b shows the resulting top three LPMs when we apply a time gap constraint of 2 minutes and a total time constraint of 5 minutes. Observe that in terms of the traditional LPM quality measures that we introduced in Chapter 8, the top three LPMs when we mine with the time gap constraints have lower *support*, *confidence*, and *coverage*, but tend to have higher *determinism*.

The first LPM shows that both *kitchen drawer 1* and 2 are almost always followed by *kitchen cabinet 1* within two minutes. When we look up the instances of this LPM in the event log, they all occur between 17:00 and 19:00, making it likely that this LPM is part of a *cooking* process. There are 16 instances of this LPM found in the event log, which is the same as the number of days in the event log. Therefore, this LPM can be used as a detector for the *cooking* activity. The second LPM shows that half of the *washing machine* events are followed within two minutes by the *laundry dryer*. This half of the washing machine events are likely to be the events where the washing machine opened after the machine finished it's washing program, while the other half of the washing machine events are the events where the washing machine was opened to put in the laundry. The third LPM shows that the *medicine cabinet* is followed by getting cold water from the *sink faucet* in slightly fewer than half of cases, where some of the *medicine cabinet* events are followed by a sequence of multiple events of getting cold water. While the second and third LPM describe trivial pieces of daily routine, it does show that the constraints help in finding sensible pieces of daily habits where the activities in the LPM are related to each other.

The computational complexity of LPM mining is combinatorially dependent on the number of activities in the event log, therefore, mining LPMs from the MIT A dataset using all 72 activities is computationally infeasible. In [Tax+16b] an approach is introduced to enable LPM mining from event logs with larger numbers of activities for which LPM mining otherwise would not be computationally feasible. This approach works by mining separately for LPMs in several heuristically chosen subsets of the activities of event log, and finally combining the results. Using this technique, the list of LPMs without using time gap constraints (i.e., Figure 9.8a) is mined in 5 minutes from the MIT A event log using an Intel i7 CPU @ 2.4GHz with 16 GB of memory. The list of LPMs obtained when using the time gap constraints (i.e., Figure 9.8b) is mined in less than 1 minute, showing that the use of constraints in LPM mining can not only make the results more useful, but can also speed up the mining procedure. The pruning mechanism based on the minimum support threshold as introduced in Chapter 8 removes parts of the search space where extending a certain LPM cannot result in an LPM that meets the threshold. The speedup obtained by using the time gap constraints results from the decrease in support of the LPMs where LPM instances do not meet the time gap constraint, therefore, leading to a higher number of LPMs that do not meet the support threshold and more patterns being pruned from the search space by the pruning mechanism.

We will further explore LPMs with time gap constraints in Chapter 11. In this chapter, we will introduce efficient algorithms for constraint-based LPM mining that are specific to time gap constraints. Additionally, this chapter further explores applications of time gap constraints in the area of smart home analysis.

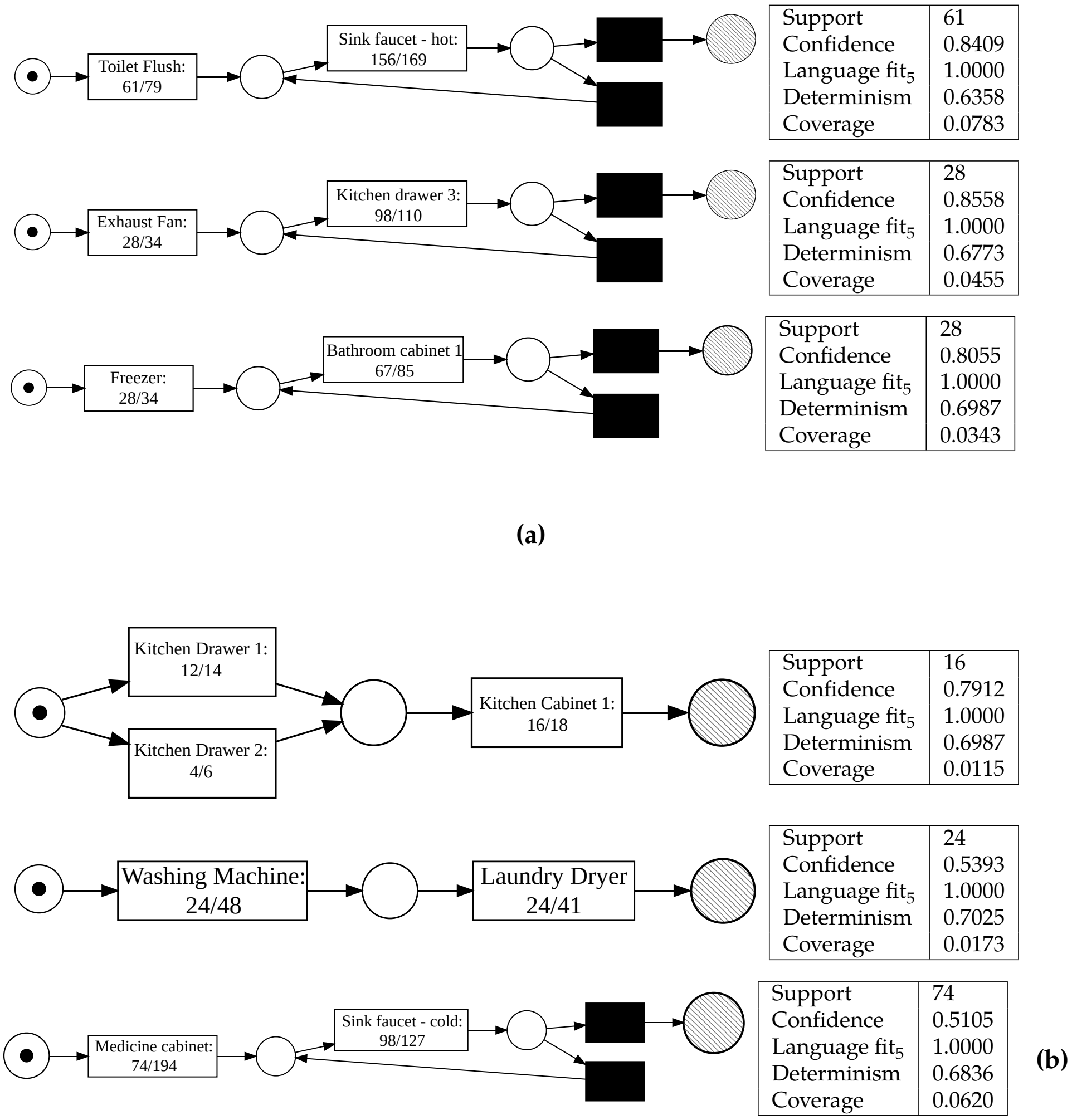

**Figure 9.8:** *(a)* The top three LPMs discovered without using time constraints, and *(b)* the top three LPMs discovered when we set a maximum time gap between two consecutive fitting events of 2 minutes and a maximum LPM instance time of 5 minutes.

## 9.3 Related Work

We now describe three classes of techniques that are closely related to LPM mining using utility functions and constraints.

### *Data-Aware Declare*

One area of work within declarative process discovery that is particularly close to this work is the discovery of data-aware declarative process models [Mag+13; Sch+16b]. MP-Declare [BMS16] extends the Declare templates to include conditions based on data attributes of events. Maggi et al. [Mag+13] proposed a technique to discover a MP-Declare model from an event log. MP-Declare consists of three types of data conditions: *activation conditions*, *target conditions*, and *correlation conditions*. Furthermore, there is the possibility to define *time conditions*. Activation conditions can be either *discriminative* or *descriptive*. Discriminative activation conditions specify a proposition on a data attribute that must hold for the events of the activation variable of the Declare template. Instead, descriptive activation conditions are initially mined without taking data attributes into account, and then the values of the some data attribute of all the activation events that fulfill the constraints are analyzed with some descriptive analysis technique. Correlation conditions relate the values of the attribute of the activation and the target event, and consists of a propositional formula containing both the activation and target event.

The main difference between utility-based LPM mining and multi-perspective Declare mining is that utility-based LPM mining focuses on question-driven analysis, allowing the process analyst to query the log for patterns of process behavior specifically based on the specified aspects of interest. In contrast, multi-perspective Declare mining provides tool that is more focused on exploratory analysis, that mines strong relations between control-flow and data attributes without taking a business need in the form of a utility function as input.

### *High-Utility Sequential Pattern Mining*

LPMs are able to express a richer set of relations (e.g., loops, XOR-constructs, concurrency) than sequential pattern mining techniques, which are limited to the sequential constructs. The mining of patterns driven by the interest of an analyst is known in the sequential pattern mining field as *high-utility sequential pattern mining*. Several high-utility sequential pattern mining algorithms have been proposed [Lan+14; Yin+13; YZC12]. In contrast to more traditional sequential pattern mining approaches, high-utility sequential pattern mining techniques assume a *cost* attribute to be present for each item, indicating the interestingness of that item.

#### *Constraint-Based Sequential Pattern Mining*

A second approach to interest-driven sequential pattern mining is *constraint-based sequential pattern mining*. In contrast to high-utility pattern mining, which gives preference to more useful patterns, constraint-based pattern mining completely removes non-useful ones. Constraint-based sequential pattern mining techniques additionally assume the *timestamp* of each item to be logged. Pei et al. [PHW07] provides a categorization of pattern constraints used in the pattern mining field, which consists of four types of constraints on the ordering of items in the pattern: *Item constraints* specify a subset of the activities (or, items) that must occur in the pattern, *length constraints* specify a requirement on the length of the sequential patterns, *super-pattern constraints* which specify a sub-pattern that the sequential pattern should contain, *regular expression constraint* which specify a regular expression that the sequential pattern should follow. Furthermore, Pei et al. [PHW07] specify two types of constraints on the timestamps of the items in the log that are instances of a pattern: *duration constraints* specify that the total time duration of instances of the sequential pattern must be longer or shorter than a given period, and *gap constraints* specify that the time difference between two adjacent activities that fit the pattern must be longer or shorter than a given period.

This richer notion of an event log allows us to define utility and constraints in a more flexible way (e.g., using event or case properties) compared to high-utility pattern mining techniques which define utility as mapping from an item type to utility and constraint-based pattern mining which limit the utility definition to ordering information and the time domain. The constraint types listed by Pei et al. [PHW07] can all be formulated in the utility-function and constraint framework that is described in this chapter, however, the framework is more general, and allows for the specification of other types of constraints.

Event gap constraints were first addressed in the sequential pattern mining field by Zaki [Zak00] by proposing the cSPADE algorithm, extending the SPADE sequential pattern mining algorithm [Zak98] to incorporate event gap constraints. More recent approaches to mine sequential patterns with event gap constraints include [AO03; Kem+16], and [Lel+03]. However, none of these techniques focus on counting *repetitive support* and patterns that go beyond sequential patterns.

#### *Process Cubes*

Process cubes [Aal13b] bring the notion of OLAP cubes to the process mining domain by allowing the user to slice and dice the event data based on event attributes or case attributes and presenting the user with a one discovered process model for each configuration of these attributes. The concept of process cubes was first described in [Aal13b] and was first illustrated on real-life data in [AGG15] in the form of a case-study on educational data from an online learning environment. Later work [BA15] improved the formalization of process cubes by solving a limitation of the original definition that was related to concurrency in the process and by allowing process cubes cells to depend on combinations of data attributes (i.e., multidimensional process cube cells). Furthermore, [BA15] offers an implementation of process cubes. Vogelgesang and Appelrath [VA16] used the multidimensional process cube approach of [BA15] as starting point and added several forms of visual highlighting to ease the process analyst to find differences between the process models in different cells of the process cube ([VA15] describes the tools & implementation of [VA16]).

Process cubes are related to utility-based LPMs in the sense that they are able to relate the control-flow dimension to the data dimension of the process. However, where process cubes allow the process analyst to see a model of the control-flow for different configurations of certain data attributes, it does not show which fragments or parts of the process are the fragments or parts that are the main contributors to certain key performance indicators (KPIs) of the business. In contrast, utility-based LPMs allow the process analyst to answer such KPI-related questions by finding the process fragment where the most money, time, or any other type of resource is spent.

### *Process Querying*

Gonzalez Lopez de Murillas et al. [GRA16] developed a meta-model for event logs and a query language that they call *Data-Aware Process Oriented Query Language (DAPOQ-Lang)* to query an event log directly for answers to business questions using an SQL-like syntax. A DAPOQ-Lang query always returns either a set of process instances that satisfy the conditions of the query, a set of data objects related to those process instances, (parts of) the database schema of the underlying database system that supports the process execution, or a set of temporal rules that relate several of these objects. An older process querying language (BPQL)is the *business process query language* [MS04], which in comparison to the more recent DAPOQ-Lang misses support to use both data attributes and temporal relations between activities as part of a query. Beheshti et al [Beh+11] proposed another querying language to analyze business process executions.

Querying languages for business process executions allow the process analyst to obtain concrete answers to her questions when she is upfront able to formulate concrete business questions. In the second case study of this chapter where we analyzed data from an IT service desk, it would have been possible to obtain the same insights as we had obtained using LPMs by formulating a query like that captures the question *"which software applications are related to more than [a given number of] interactions"*. However, when the insights are related mostly to the temporal relations between business activities such as those of Figure 9.6, then process querying techniques fail to provide the same insights as the queries are not answered in process-model-form and therefore do not show the control-flow relations between the activities.

## 9.4 Conclusions

This chapter presents a framework of utility functions and constraints for Local Process Models (LPMs) that allows for combinations of utility functions and constraints on different scopes: on the activity, event, trace, and model level. We formalize utility functions on each of the levels and provide examples of how they can be used. Finally, we show on real-life event logs that the utility functions and constraints can be used to discover insightful LPMs that cannot be obtained using existing support-based LPM discovery.

When a model e.g. allows for one execution of activity ‘a’ while multiple executions of ‘a’ are observed in the log, different alignments are possible depending on the choice of ‘a’ for the synchronous move and the ones for log moves. However, these events do not necessarily have equal utility. Thus, the utility of the LPM depends on the alignment returned by the alignment algorithm.

# 10 Heuristic Mining of Local Process Models

**Parts of this chapter have been published as:**

- Niek Tax, Natalia Sidorova, Wil M. P. van der Aalst, and Reinder Haakma: *Heuristic approaches for generating local process models through log projections*. Proceedings of the IEEE Symposium Series on Computational Intelligence (SSCI), IEEE: pp. 1–8 (2016)
- Niek Tax, Laura Genga, Nicola Zannone: *On the Use of Hierarchical Subtrace Mining for Efficient Local Process Model Mining*. Proceedings of the 7th International Symposium on Data-driven Process Discovery and Analysis (SIMPDA), CEUR-WS.org, 2016: pp. 8–22 (2016)
- Laura Genga, Niek Tax, Nicola Zannone: *Mining Local Process Models and their Correlations*. Data-Driven Process Discovery and Analysis, Lecture Notes in Business Information Processing (LNBIP), 340, Springer: pp. 65–88 (2018)

The computational time that is required to mine Local Process Models (LPMs) using the $exp_2$ expansion function and with the pruning mechanisms that we introduced in Chapter 8 grows rapidly as a function of the number of activities in the event log. This complicates the application of LPM mining on datasets with large numbers of activities. In this chapter, we will explore heuristic LPM mining techniques. Such heuristic LPM mining methods are able to mine LPMs fast, but, as a consequence, they might return approximate results (i.e., they might fail to discover some patterns). The goal of heuristic LPM mining is twofold:

1. the mining procedure should be as fast as possible
2. the mining procedure should approximate the results of the traditional LPM mining procedure as closely as possible.

In order to closely approximate the LPM mining results we consider it to be vital that at least the highest scoring LPMs in terms of the LPM quality criteria that we introduced in Section 8.1 are found.

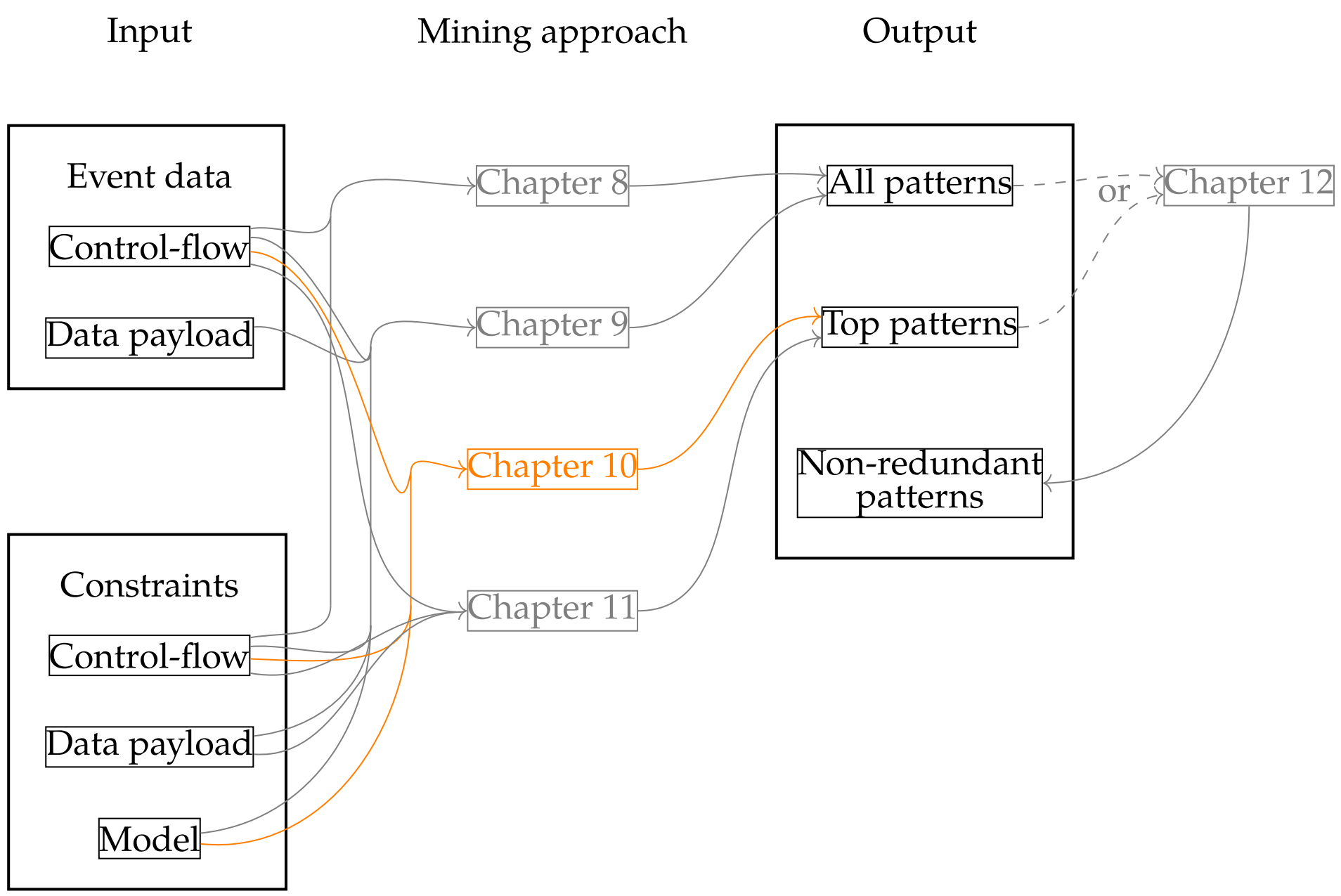

**Figure 10.1:** A taxonomy of local process model techniques.

One example of an event log that has so many activities that LPM mining using the traditional approach is computationally infeasible is the MIMIC-II dataset [Sae+02], a medical database containing 147461 logged medical procedure events from 1734 activities for 28280 patients collected between 2001 and 2008 from several intensive care units (ICUs) in the USA. Figure 10.2 shows some of the LPMs that we discovered from this event log using one of the heuristic mining techniques that we will introduce later in this chapter.

The first LPM indicates that the placement of continuous invasive mechanical ventilation on a patient is 3638 out of 8291 times followed by either arterial catheterization or by one or more instances of infusion of concentrated nutrition. The second LPM shows that the placement of invasive mechanical ventilation frequently co-occurs with the insertion of an endotracheal tube. The third LPM shows that a sequence of one or more hemodialysis events is always followed by the placement of a venous catheter for renal dialysis, and from the numbers, we can deduce that there are on average almost 4 hemodialysis events before one placement of a venous catheter. The fourth LPM in Figure 10.2 shows that all placements of a continuous invasive mechanical ventilation are eventually followed by the placement of a non-invasive mechanical ventilation. However, on average 8 continuous invasive mechanical ventilation have been placed (and probably removed again, but this was not logged) before the placement of a non-invasive mechanical ventilation.

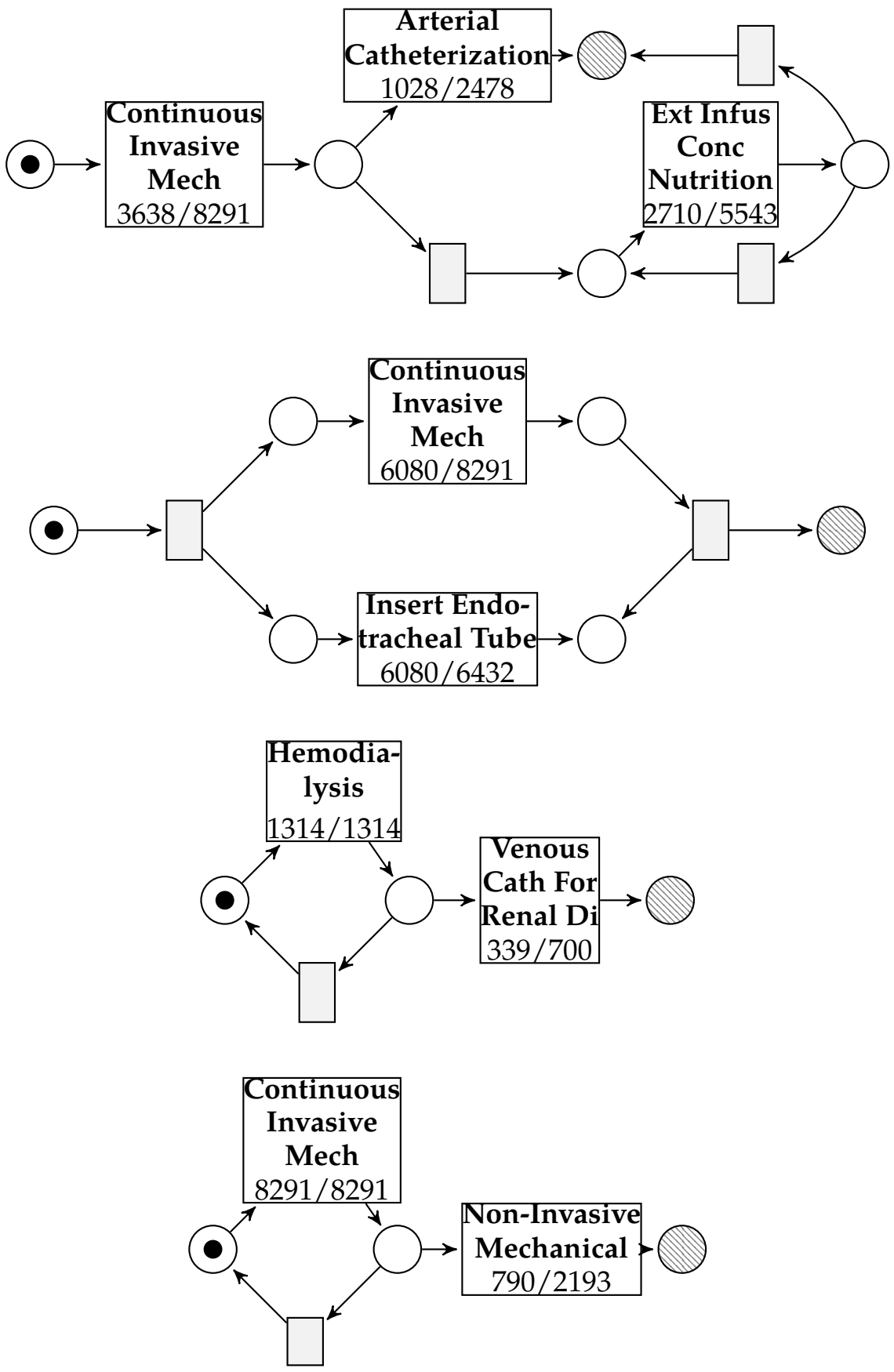

**Figure 10.2:** The top four local process models that were mined from the MIMIC II medical procedure data set using a heuristic projection set technique.

Due to computational issues it is not possible to discover the LPMs in Figure 10.2 using the brute-force approach based on expansion function $exp_2$ (as defined inChapter 8) given the large number of activities in the log. There are 1734 activities in the log, yielding an incredible number of possible candidate models. The computational complexity of LPM discovery grows combinatorially as a function of the number of activities in the event log, hindering the practical application to many real-life event logs. Observe however that for a given LPM *LPM* it would have been possible to discover it from a projection on the log that contains only a subset of all the activities that are present in the log, as long as this subset of activities contains at least the activities that are represented in the LPM. For example, the top LPM in Figure 10.2 could have been discovered using a projection on the log on just the activities *Continuous Invasive Mech*, *Arterial Catheterization*, and *Ext Infus Conc Nutrition*.

The heuristic mining techniques that we discuss in this chapter all follow the same paradigm: finding subsets of activities in the event log that are behaviorally related and then mine the LPMs only within these activity sets by projecting the log on these activities. The computational time of discovery decreases considerably when we apply LPM discovery only to an modified copy of the log that has been projected onto a subset of the activities. For example, the LPMs shown in Figure 10.2 were discovered in 12 seconds using the Markov clustering approach to finding projection sets if activities that we will introduce in Section 10.1, while the brute-force LPM mining approach did not finish mining in a full day.

In Section 10.1, we introduce the three heuristic approaches to discover projection sets of activities. In Section 10.2 we describe the experimental setup to evaluate these heuristics. Section 10.3 presents and discusses the results of the experiments. In Section 10.4 we will show that there exist pattern mining techniques that are orthogonal to LPMs in terms of the insights that they provide which that are themselves computationally expensive, but once they have been mined their results can be used to mine LPM patterns more efficiently. We discuss related work in Section 10.5 and conclude this chapter Section 10.6.

## 10.1 Finding Log Projection Sets for Heuristic LPM Mining

In this section, we describe three heuristics for the discovery of *projection sets* – subsets of activities to be used for projecting the log. Each heuristic takes an event log $L \in \mathcal{B}(\Sigma^*)$ as input and produces a set of projection sets $Q = \{q_1, \dots, q_n\}$ such that each projection set $q_i \subseteq \Sigma$. The intuition behind this is that Local Process Models (LPMs) only contain a subset of the activities of the log and each LPM can in principle be discovered on any projection of the log containing the activities used in this LPM. Since the computational complexity of LPM mining depends combinatorially on the number of activities in the log, mining LPMs on projections $q_i \in Q$ of the log instead of on the full log will significantly lower the running time of LPM mining since $|q_i| \leq |\Sigma|$. However, this results in a partial exploration of the LPM search space, since for each run the expansion function *exp* is constrained to expanding with activities from $q_i$, potentially leading to good LPMs not being found.

In principle, when we have a set of projection sets $Q = \{q_1, \dots, q_n\}$ such that activities that frequently follow each other are in the same projection set $q_i$, then the search space can be constrained almost without loss of quality since for each LPM with alphabet $\Sigma_M$ there exists a projection set $q_i \subseteq \Sigma_M$. Such projection sets $Q = \{q_1, \dots, q_n\}$ could potentially be overlapping. Such overlaps are in fact desired, since interesting patterns might for example exist within a set of activities $\{a, b, c, d\}$, as well as within a set of activities $\{a, b, c, e\}$, and discovering on both $L \upharpoonright_{\{a,b,c,d\}}$ and $L \upharpoonright_{\{a,b,c,e\}}$ and merging the results is faster than discovering on the full log with $\mathit{Activities}(L) = \{a, b, c, d, e\}$.

The desired set of projection sets $Q = \{q_1, \dots, q_n\}$ should fulfill the requirement that none of its elements is contained in another element, i.e., $\forall_{q_i,q_j \in Q=\{q_1,\dots,q_n\}} q_i \subseteq q_j \implies i = q$. This is important to avoid unnecessary computational work, since if there would be two projection sets $q_i$ and $q_j$ such that $q_i \subseteq q_j$ then any LPM that can be mined from $q_i$ can also be mined from $q_j$ and therefore mining from contained projection set $q_i$ is redundant.

### 10.1.1 Projection Sets based on Markov Clustering

Markov clustering [Don00; Don08] is a fast and scalable clustering algorithm for graphs that is based on simulation of flow in graphs. The main intuition behind Markov clustering is that, while performing a random walk on a graph, the likelihood of transitioning between two members of the same cluster is higher than the likelihood of transitioning between two nodes that are in different clusters. Markov clustering takes as input a Markov matrix, i.e. a matrix that describes the transition probabilities in a Markov chain. We generate a square matrix M of size $\Sigma \times \Sigma$ in which each element represents the *connectedness* of two activities.

Let $\mathit{index} : \Sigma \to \{1, \dots, |\Sigma|\}$ be a function that specifies an arbitrary but consistent ordering over the activities of the log, i.e., $\forall_{a,b \in \mathit{Activities}(L)} a \neq b \implies \mathit{index}(a) \leq \mathit{index}(b)$. The value in the $\mathit{index}(a)$-th row and $\mathit{index}(b)$-th column of M is defined using the directly-follows and directly-precedes ratios in the following way: $M_{\mathit{index}(a),\mathit{index}(b)} = \sqrt{\mathit{dpr}(a, b, L)^2 + \mathit{dfr}(b, a, L)^2}$. In machine learning terms, this comes down to using the $L_2$ norm of categorical probability distributions $\mathit{dpr}(a, b, L)$ and $\mathit{dfr}(b, a, L)$. A Markov matrix $M'$, i.e. a matrix where every value comes from interval $[0, 1]$ and each row sums to 1, is obtained by normalizing M row-wise.

The intuition behind using both *dpr* and *dfr* combined instead of either of the two is that it can be both of importance that activity a is often followed by b and that b is often preceded by a; if either of the two is true than there is apparently some relation between a and b. Applying Markov clustering to $M'$ results in a set of clusters of activities, where we use each activity cluster as a projection set.

Markov clustering simulates a random walk over a graph by alternating *expansion*, i.e. taking the power of this matrix, and *inflation*, i.e. taking the entrywise power of this matrix. The clusters generated by Markov clustering can overlap, which is a desired property in projection set discovery. The inflation parameter of the Markov clustering algorithm is known to be the main parameter in determining the granularity of the clustering obtained [Don00] with Markov clustering.

### 10.1.2 Projection Sets based on Log Entropy

Another approach to generate projections sets is to form groups of activities in such a way that for each activity in a group the categorical probability distribution over activities that are directly preceding it or are directly following it are very peaked, i.e. they are far away from the discrete uniform distribution. If after an occurrence of a given activity $a \in Activities(L)$ each other activity in the set of activities $Activities(L)$ is more or less equally likely to observe as following activity in the log, then this activity can be considered to be independent from the other activities in a probabilistic sense. Observe that such an activity matches the definition of a *chaotic activity* as introduced in Chapter 6. We can re-use this notion of chaotic activities by searching for subgroups of activities in the event log such that the activities within each group are non-chaotic to a high degree.

In Chapter 6 we had proposed to measure the "chaoticness" of an activity by measuring the information entropy of the probability distributions over the *following* and over the *preceding* activities. Furthermore, recall that we had defined the entropy of activity $a \in Activities(L)$ in log L as: $H(a, L) = H(dfr(a, L)) + H(dpr(a, L))$ where $H(X) = -\sum_{x \in X} x \log_2(x)$. Furthermore, $dfr(a, L)$ and $dpr(a, L)$ represented the probability distribution over the activities that follow respectively precede activity a (see Section 6.1). In case there are probability values of zero in the directly follows or directly precedes vectors, i.e., $0 \in dfr(a, L) \lor 0 \in dpr(a, L)$, then the value of the corresponding summand $0 \log_2(0)$ is taken as 0, which is consistent with the limit $\lim_{p \to 0^+} p \log_2(p) = 0$. Additionally, in Chapter 6 we had defined the entropy of an event log as the sum of the entropy values of the activities in the event log, i.e., $H(L) = \sum_{a \in Activities(L)} H(a, L)$. We aim to find subsets of activities S such that event log L projected on these activities meets some threshold in terms of the log entropy, i.e., we set a maximal entropy threshold parameter $t_H$ and search for subsets of activities S such that $H(L \upharpoonright_S) \leq t_H$.

In order to find sets of activities that meet these criteria we perform a bottom-up search, i.e., we start from a set of elementary log projection sets, $S_1 = \{\{a\} \mid a \in Activities(L)\}$. Then, iteratively we expand this set of log projections into larger sets of log projections, i.e., $S_{i+1} = \{s_i \cup s_1 \mid s_i \in S_i, s_1 \in S_1 \land H(L \upharpoonright_{s_i \cup s_1}) \leq t_H\}$. The procedure stops if $S_{i+1} = \emptyset$ or if $S_{i+1} = Activities(L)$. LPMs that can be discovered using some projection set s can also be discovered using a superset s′, therefore, projection sets that are subsets of other projection sets are undesirable as they lead to unnecessary double work. However, note that $S_i$ can contain projection sets that are fully contained in projection sets in $S_j$ when $j > i$. Therefore, we aggregate $S_1, S_2, \ldots$ into a final set of projection sets in the following way: $S_{final} = \{s_i \mid s_i \in S_i \land \nexists_{s_j \in S_j} s_i \subset s_j\}$.

### 10.1.3 Projection Sets based on Maximal Relative Information Gain

A more local perspective on entropy-based projection set discovery would be to compare a projection set to the projection set of the previous time step, instead of to a fixed entropy threshold like in the entropy-based method. In the Maximal Relative Information Gain (MRIG) approach, we add an activity to a projection set if adding it decreases the entropy of at least one of the categorical probability distributions *dpr* or *dfr*, even when the entropy of other categorical probability distributions might increase. The intuition behind this is that any decrease in entropy of a categorical probability distribution *dpr* or *dfr* indicates that some pattern is getting stronger by adding that activity.

We define the (MRIG) of projection set A over projection set B, with $A \supset B$, on event log L as the maximal relative information gain over all the log statistics on L, that is the ratio of bits for encoding the log statistic that is most decreased by growing the projection set:

$$MRIG(A, B, L) = \max_{a \in B} \max\{\frac{H(dfr(a, L\restriction_B)) - H(dfr(a, L\restriction_A))}{H(dfr(a, L\restriction_B))}, \frac{H(dpr(a, L\restriction_B)) - H(dpr(a, L\restriction_A))}{H(dpr(a, L\restriction_B))}\}$$

We choose a threshold parameter $t_M$ to indicate the minimum value of MRIG for considering a projection set as potentially interesting. Like in the log entropy based approach, we start from the set of elementary log projection sets, $S_1 = \{\{a\} \mid a \in \Sigma_L\}$ and $S'_1 = S_1$. We proceed iteratively defining $S_{i+1} = \{s_i \cup s_1 | s_1 \in S_i, s_1 \in S_1, MRIG(s_i \cup s_1, s_1, L) > t_M\}$. The procedure stops if $S'_{i+1} = \emptyset$ or $S_{i+1} = \Sigma_L$. Like in the entropy-based approach, projection sets that are subsets of other projection sets are ultimately removed: $S_{final} = \{A | \exists S'_i : A \in S'_i \land (\forall S'_j, \forall B \in S'_j : A \not\subset B)\}$.

10

## 10.2 Experimental setup

Figure 10.3 gives an overview of the evaluation methodology for the three heuristics. We assess their usefulness by comparing the quality of the LPMs obtained using the heuristic with the LPMs using obtained using the full LPM search. First, we apply LPM discovery to the original, unprojected, event log L, resulting in the "ideal" top k of LPMs that we aim to approximate using the heuristic techniques. Then we apply one of the projection set discovery methods on event log L, resulting in a set $Q = \{q_1, \ldots, q_n\}$ of projection sets. On each of the projected event logs in $L\restriction_{q_i}$ for $q_i \in Q$ we apply LPM Discovery, resulting in $|Q| \times k$ LPMs (i.e., one ranking of k LPMs for each $q_i \in Q$), which we aggregate into a single ranking of k LPMs by selecting the top k unique best LPMs in terms of weighted average over the quality criteria. We compare these LPMs to the "ideal" ranking of LPMs that we obtained without projections. Theoretically, they can coincide, or be equally good, if for every "ideal" LPM there is a projection set containing its activities. If it is not the case, some of the best scoring models will be missing and we compare how

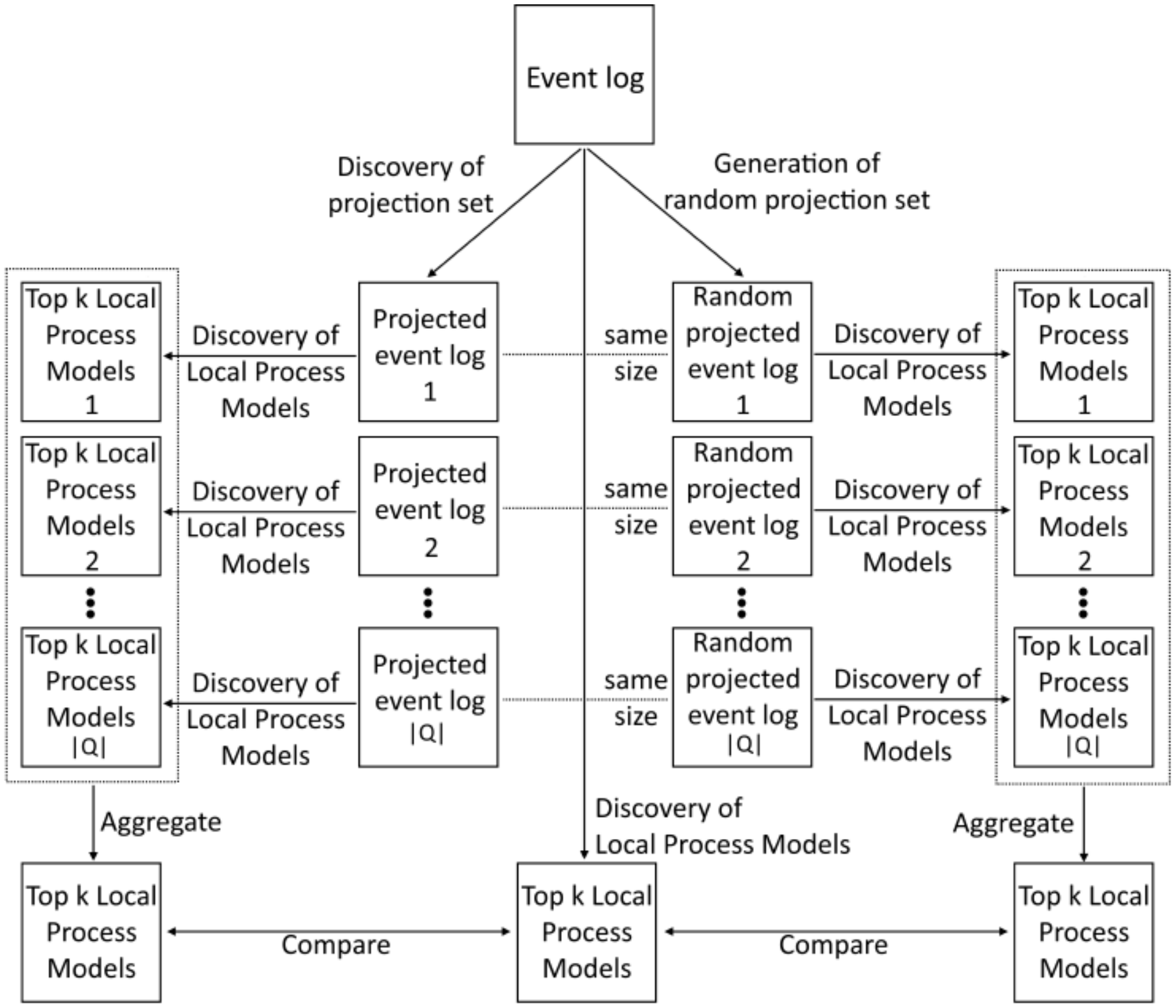

**Figure 10.3:** The evaluation methodology for projection set discovery approaches

close are the scores of the models present in the set to the scores of the "ideal" LPMs.

Even with the loss of quality in LPMs, projections could be a good option to opt for, since they allow to significantly improve the computation time and make LPM discovery possible for logs with many different activities. We evaluate how our projection discovery methods perform compared to random projections as follows: We create a set $Q'$ of random projection sets, consisting of $|Q|$ projection sets where for each projection set $q \in Q$ we create a random projection set $q' \subseteq \Sigma_L$ of the same size as q: $|q'| = |q|$. We apply LPM discovery on each event logs in $L\upharpoonright_{q'}$ for each $q' \in Q'$ aggregate the results by selecting the unqie top k best LPMs from the discovered ones. The projection discovery method works well if the LPMs from top k score closer to the "ideal" top k than the LPMs generated based on the random projection sets. To obtain statistically relevant results, we create the random projection sets ten times.

### 10.2.1 Measures for Comparison of Local Process Model Rankings

A simplistic way to evaluate the quality of a ranked result is by measuring *recall*, i.e., by simply measuring how many of the elements of the top of the ranking in the "ideal" (i.e., ground truth) ranking are also present in the top of the ranking that was obtained. More concretely we define *recall@k*, as the fraction of LPMs from the "ideal" top k that are also discovered with the use of projections.

A more nuanced way of comparing rankings is based on the idea of quantifying the *relevance* of each element in the ranking. By using such relevance information, it is in contrast to *recall@k* possible to take into account the fact that for certain elements of the ground truth ranking it might be worse to not have them in the obtained ranking than for others. In the case of LPM rankings, we can use the weighted average over the support, confidence, language fit, determinism, and coverage as its relevance. Another benefit of quantifying the relevance of each element in the ranking is that not being able to find LPMs from the "ideal" top k is less severe in the case that the alternative LPMs that were found instead are also of good quality.

Furthermore, missing the top LPMs of the ground truth ranking is more severe than missing some lower ranked LPMs from the ground truth ranking. Therefore, we additionally use Normalized Discounted Cumulative Gain (NDCG@k) [Bur+05; JK02], one of the most widely used metrics for evaluation of a ranking with respect to a given "ideal" (i.e., ground truth) ranking in the field of Information Retrieval [TBH15]. The NDCG measure compares the rankings in such a way that it gives more weight to the top of the rankings than to the lower parts of the rankings.

Discounted Cumulative Gain (DCG) aggregates the relevant scores of the individual LPMs in the ranking of LPMs in such a way that the graded relevance is reduced in the logarithmic proportion to the position of the result; due to that, more weight is put on the top of the ranking than on the lower parts of the ranking. DCG is formally defined as:

$$DCG@k(\mathrm{R}) = \sum_{\mathrm{i}=1}^{min(\mathrm{k},|\mathrm{R}|)} \frac{rel(\mathrm{R}(\mathrm{i}))}{\log_2(\mathrm{i}+1)} \tag{10.1}$$

where $rel(\mathrm{R}(\mathrm{i}))$ is the relevance of the LPM at position i in ranking R (i.e, its weighted average of the model's scores on the quality measures for the LPM). The @k part of DCG@k indicates that it is calculated only over the first k elements of the ranking (or up to the length of the ranking, whichever is smaller). Note that DCG@k returns a value of which the scale is dependent on k and on the scale of the relevance scores of the elements in the ranking. In order to obtain a measure that can be used to compare ranking approaches it is useful to normalize the DCG scores into $[0,1]$-range. This Normalized Discounted Cumulative Gain (NDCG) is obtained by dividing the DCG value by the theoretically highest possible DCG@k score, i.e., the DCG@k score of the ground truth ranking itself. This theoretically highest possible DCG@k score is called Ideal Discounted Cumulative Gain (IDCG@k). In the context of evaluating LPM rankings using projection set discovery approaches, the IDCG value is the DCG value that is obtained when mining the LPMs using the brute-force LPM mining approach of Chapter 8, i.e., without projecting the log on subsets of activities. This yields a proper ground truth, since all local process models that can be discovered from projected event logs can also be discovered from the original event log. Formally, Normalized Discounted Cumulative Gain (NDCG) is defined as:

$$NDCG@k(\mathrm{R}) = \frac{DCG@k(\mathrm{R})}{IDCG@k(\mathrm{R})} \tag{10.2}$$

### 10.2.2 Data Sets for Projection Discovery Experiments

We perform the evaluation on five different data sets using the methodology described above. All of the data sets used originate from the human behavior logging domain, with four data sets consisting of *Activities of Daily Life* (ADL) and one consisting of activities performed by an employee in a working environment. Event logs from the human behavior domain are generally too unstructured to allow for discovery of informative process models with process discovery techniques, while LPM discovery allows to discover some relations between activities at a local level that are not discoverable with usual process mining techniques.

Table 10.1 gives an overview of the main event log characteristics used in the evaluation. Even though the goal of the project set discovery approaches that we have introduced in this chapter are to enable the mining of LPMs on logs with large numbers of activities, the event logs that we will use in the experiments of this chapter have a limited number of activities (at most 14). The motivation for this is that this allows us to determine the ground truth ranking of LPMs, that is, the ranking of LPMs obtained by LPM discovery on the full log without using projections. However, our goal of enabling the mining of LPMs from logs with higher numbers of activities is still reached, and an example of this we have already seen in Figure 10.2, where we mined and displayed the top LPMs from the MIMIC-II dataset with 1734 activities. For the Van Kasteren data set [Kas+08] we show and discuss the ranking of LPMs for which we aim to speed-up discovery through projections. For each data set, we identified a support threshold used for pruning that allows us to run LPM discovery on the unprojected event log within a reasonable time (max. 10 minutes on a 4-core 2.4 GHz CPU). This value depends on the number of activities as well as the length of the traces within the event logs, i.e. more activities and/or longer traces result in a need for a higher support threshold to finish the experiment within the time limit.

The **Business Process Intelligence Challenge (BPIC)'12** data set originates from a personal loan application process in a global financial institution. We use a version of this event log that takes the perspective of a single employee (resource id 10939) of the bank where the cases are his working days and the events correspond to the activities that are executed during these working days by this employee. We refer back to Section 8.6.1 for a detailed description of how this event log is obtained from the original BPIC'12 data set. We set the support pruning parameter to 0.675 for this data set.

**Table 10.1:** An overview of the data sets used in the evaluation experiment

| Data set | # of activities | # of cases | # of events |
|---|---|---|---|
| BPI '12 resource 10939 | 14 | 49 | 1682 |
| Bruno | 14 | 57 | 553 |
| CHAD subject 1900010 | 10 | 26 | 238 |
| Ordonez A | 12 | 15 | 409 |
| Van Kasteren | 14 | 23 | 1285 |

The **Bruno** et al. [Bru+13] data set is a public collection of labeled wrist-worn accelerometer recordings. The data set is composed of fourteen types of ADL events performed by sixteen volunteers. We set the support pruning parameter to 0.6 for this data set.

The **CHAD** database [McC+00] consists of 22 exposure and time-use studies that have been consolidated in a consistent format. In total, the database contains 54000 individual days of human behavior, from which we extract an event log from a randomly chosen study subject such that each case represents a day. For this data set, we set the support pruning parameter to 0.4.

The **Ordonez** et al. [OTS13] data set consists of ADL events that are performed by two users in their own homes, recorded through smart home sensors. We use an event log obtained from sensor events of subject A, with each case representing a day. We set the support pruning parameter to 0.6.

The **Van Kasteren** et al. [Kas+08] data set is a smart home environment data set consisting of multidimensional time series data. Each dimension in the time series represents the binary state of an in-home sensor. These sensors include motion sensors, open/close sensors, and power sensors (discretized to on/off states). We transform the multidimensional sensor time series into events by considering the change points of sensor states as events. We create cases by grouping events by day, with a cut-off point at midnight. We set the support pruning parameter to 0.675 for this data set.

## 10.3 Results & Discussion

Figure 10.4 shows the results of the evaluation using the five evaluation data sets and Table 10.2 shows the speed-up of projection-based LPM discovery. The speed-up mainly depends on the size of the projection sets. Note that we do not compare the computation time of the discovered and the randomly generated projections, as both consist of equally sized projections. The time needed for discovering the projection set is included in the computation time shown in Table 10.2, however,

**Table 10.2:** Local Process Model discovery speed-up obtained with each projection set discovery method on the evaluation data sets

| Data set | Speed-up Markov | Speed-up entropy | Speed-up MRIG |
|---|---|---|---|
| BPI'12 resource 10939 | **42.9** | 5.6 | 6.9 |
| Bruno | 222.9 | 111.5 | **336.1** |
| CHAD subjection 1900010 | **1060.1** | 1041.2 | 1 |
| Ordonez A | 7.3 | **33.7** | 1.1 |
| Van Kasteren | **111.2** | 83.5 | 96.7 |

the projection set discovery time is negligible compared to the time needed for LPM discovery.

Each dark gray bar in Figure 10.4 represents the performance on the measure indicated by the column of the projection set discovery method indicated by the row on the dataset indicated below. Each gray bar indicates the average performance over ten random projection sets of the same size as the discovered projection set belonging to the dark gray bar to its left. The error bars indicate the Standard Error (SE), indicating the uncertainty of the mean performance of the random projection sets.

Our first observation is that all three projection set discovery methods are better than random projections on almost all data sets. The only exception is the Markov-based projection set discovery approach in case of the Bruno data set, which performs worse than random projections. In this case, remarkably, none of the LPMs that were discovered using the full brute-force-search were also found by using the projection sets of the Markov-based approach. The Markov clustering based approach to projection set discovery achieves the highest speed-up on three of the five data sets, and second highest speed-up on the other two data sets. The high speed-up indicates that the projections generated by Markov clustering are relatively small, which also explains the quality loss of the LPMs with regard to the ground truth which is typically larger than with the other projection set discovery methods. The size of the clusters created with Markov clustering can be influenced through its inflation parameter, which we set to 1.5. We found lower values to result in all activities being clustered into one single cluster, meaning that the only projection created is the original log itself.

The entropy-based approach shows a higher gain than the Markov based approach on the NDCG@{5,10,20} metrics for all data set, when compared to random projections of the same size. However, the obtained speed-up of LPM discovery with the entropy based approach is lower than speed-up with the Markov based approach on all but one data set.

The MRIG-based approach shows significant improvements on all metrics on the BPI'12 and Bruno data sets. On these two data sets, it also results in the second highest and highest speed-up respectively. However, on the CHAD data and the Ordonez data the discovered projection sets consist of a single projection that contains almost all log activities, resulting consequently in no speed-up with close to perfect recall scores.

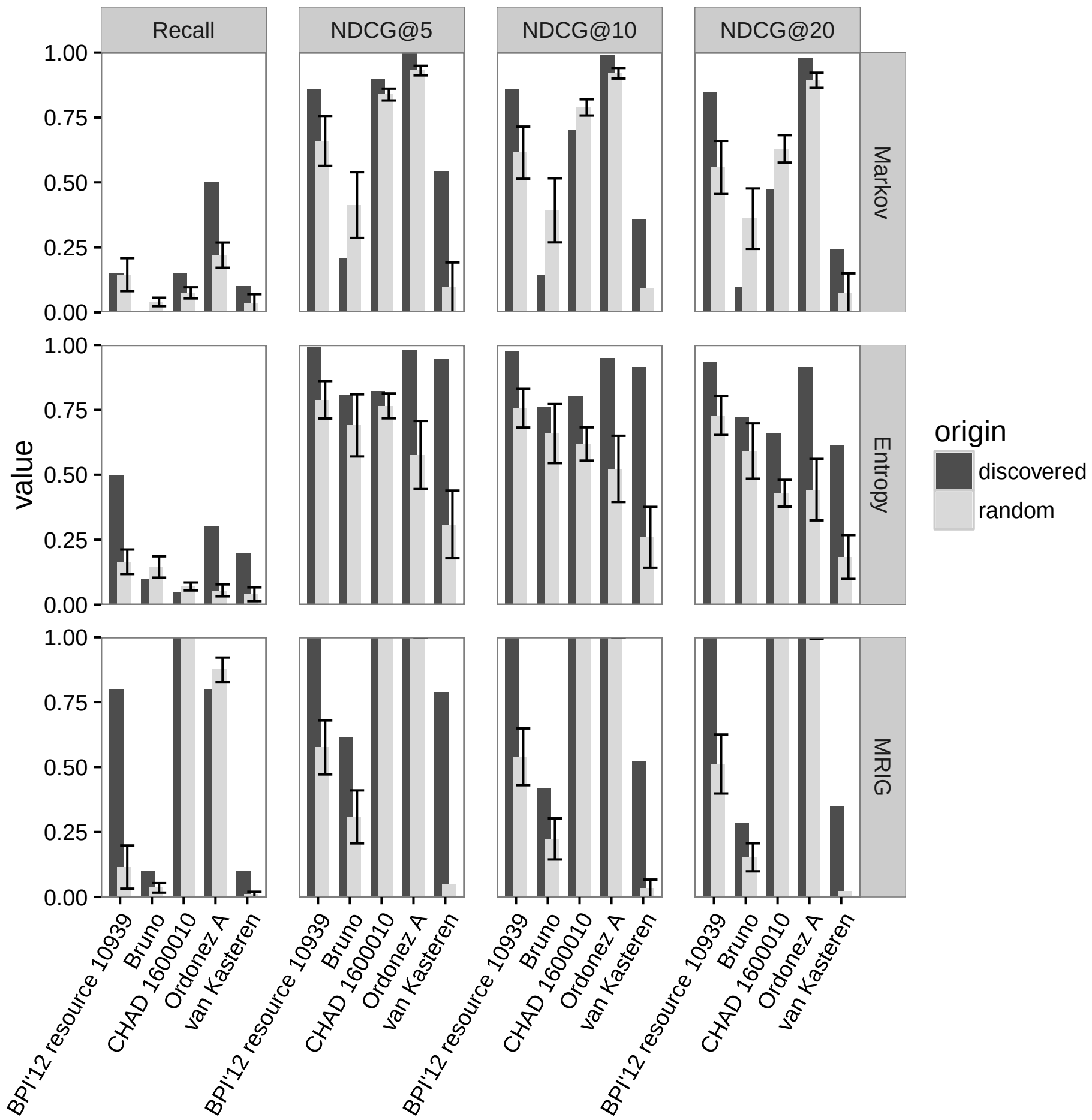

**Figure 10.4:** The performance of the three projection set discovery methods on the six data sets on the four metrics. Each column of the grid represents one measure and each row represents one projection set discovery method.

## 10.4 Using Subgraph Patterns for LPM Mining

As we have seen, LPM mining results in patterns that contain rich control-flow constructs such as sequential ordering, parallelism, choices, and loops. In contrast, subgraph mining techniques applied to traces that are represented as sequential graphs (i.e., subtrace mining [DGP16], discussed in Section 8.5) only allow for the mining of sequential behavior. However, using hierarchical subgraph mining technique such as the SUBDUE algorithm [JCH02], it is possible to mine hierarchical relations between patterns, which is infeasible with LPM patterns. This motivates that subtrace mining and LPM mining can be orthogonal techniques that can both be used as complementary techniques during exploratory analysis of an event log. However, both techniques can themselves be computationally expensive. Therefore, in this section we explore whether the results of subtraces that are mined with the SUBDUE algorithm can be used to speed-up LPM mining computation.

We conjecture that we can obtain a speed-up in LPM mining by using SUBDUE patterns in two different ways:

1. SUBDUE patterns can be used at projection set discovery approach by taking each set of activities that co-occur in a SUBDUE pattern to be a projection set.

2. The ordering over activities that is implied by a SUBDUE pattern can be used as a model-level constraint, thereby excluding LPM patterns that directly contradict the ordering of the SUBDUE pattern from search space $exp_2$.

Note that we separate this section from Section 10.1 for two reasons. First of all, the approach that we present in this section is not solely an approach to find projection sets of activities, but, in contrast to the approaches in Section 10.1, additionally constrains the search space for LPMs. Secondly, mining the SUBDUE patterns is itself a very computationally expensive operation due to the hierarchical nature of the SUBDUE patterns. Therefore, in contrast to the techniques that we described in Section 10.1, projection sets and constraints based on SUBDUE patterns *only speed up the mining of LPMs when the SUBDUE patterns are already available*. In other words, there is no longer any computational benefit to LPM mining if the computation time of mining the SUBDUE patterns is taken into account.

We introduce subtrace mining and present an approach to speed-up LPM mining using subtraces that are mined with the SUBDUE algorithm in Section 10.4.1. In Section 10.4.2, we will evaluate this technique using the same evaluation methodology that we described in Section 10.2.

### 10.4.1 Approach

We now investigate the application of the subtraces inferred by the technique proposed in [DGP16] as input for LPM mining, i.e., as projection sets for LPM mining. The approach presented in [DGP16] proposes to apply *Frequent Subgraph Mining* (FSM) algorithms to derive most relevant subprocesses. As a first step, the event log L is transformed into a directed graph $g = (V, E, \phi)$ where the set of nodes V consists of one node for every event in log L, where set of edges E contains an edge for each pair of events that directly follow each other, and labeling function $\phi$ associates each node with the activity of the corresponding event. As a result, each trace $\sigma \in L$ forms a connected component in graph g. Figure 10.5 gives an example of what the resulting graph looks like for log $L = [\langle a, b, c\rangle^2, \langle d, b, e\rangle]$.

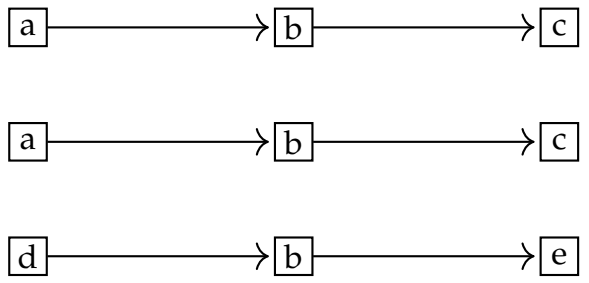

**Figure 10.5:** An example of the resulting graph for log $L = [\langle a, b, c\rangle^2, \langle d, b, e\rangle]$. The graph consists of three connected components, each corresponding to one trace from the event log.

Once this graph is obtained, an FSM algorithm is applied to derive frequent subgraphs from it. Since the graph is generated out of the traces from the log, the resulting frequent subgraphs can be interpreted to be frequent subtraces from the log. FSM algorithms usually rely on a metric to evaluate the *relevance* of the subgraphs. The work in [DGP16] uses the SUBDUE algorithm [JCH02], which is a FSM algorithm that employs *Description Length* (DL) to assess the relevance of subgraphs. Given a graph g and a subgraph s of g, SUBDUE uses the following relevance measure:

$$\nu(s, g) = \frac{DL(g)}{DL(s) + DL(g \mid s)} \tag{10.3}$$

Where $DL(g)$ is the number of bits that are needed to encode graph g, $DL(s)$ is the number of bits that are needed to encode subgraph s, and $DL(g \mid s)$ is the number of bits that are needed to encode a *compressed* version of graph g where every occurrence of subgraph s is replaced by a single node. By doing so, SUBDUE relates the relevance of a subgraph with its compression capability. The higher the value of $\nu(s, g)$, the lower the stronger the compression ability of subgraph s for graph g.

The SUBDUE algorithm works iteratively. At each step, it extracts the subgraph with to highest compression capability, i.e., $\bar{s} = \arg\max_s \nu(s, g)$. This subgraph is then used to compress graph g using the subgraph $\bar{s}$ with the strongest compression capability on g, and in next iteration the algorithm proceeds with this compressed graph. The subgraph $\bar{s}$ with the strongest compression capability is added to the set of mined SUBDUE patterns. The SUBDUE algorithm continues until either no more compression is possible or until a user-defined maximum number of iterations is reached. The outcome of SUBDUE is a hierarchical structure, where mined subgraphs are ordered according to their relevance, showing the existing inclusion relationships among them.

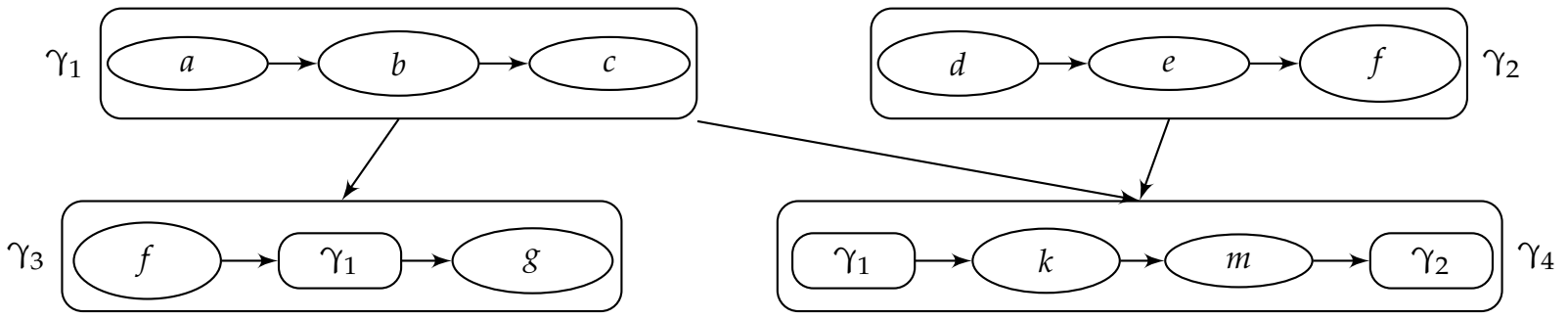

**Figure 10.6:** An example of the SUBDUE hierarchy

Top-level subgraphs are defined only through elements that belong to input graphs (i.e., nodes and edges), while lower-level subgraphs additionally contain upper-level subgraphs as nodes. Descending the hierarchy, we pass from subgraphs that are very common in the input graph set to subgraphs occurring in a few input graphs. Therefore, we can reasonably expect to be able to capture most of the process behaviors by only considering top-level subgraphs.

Figure 10.6 shows a simple example of the hierarchical structure mined by SUBDUE. We have two subgraphs at the top level, i.e., $\gamma_1$ and $\gamma_2$, together with two subgraphs at the lower level, i.e., $\gamma_3$ and $\gamma_4$. We can see that $\gamma_3$ is a child of $\gamma_1$, since it involves $\gamma_1$ in its definition. Similarly, $\gamma_4$ is a child of both $\gamma_1$ and $\gamma_2$.

### *Projection Sets based on SUBDUE patterns*

To extract projection sets for LPM mining based on SUBDUE patterns we create one projection set per top-level SUBDUE pattern that consists of all activities that occurred in that SUBDUE pattern. In the example SUBDUE patterns of Figure 10.6, we would extract two projection sets: projection set $\{a, b, c\}$ for top-level SUBDUE pattern $\gamma_1$ and projection set $\{d, e, f\}$ for top-level SUBDUE pattern $\gamma_2$. Note that just like with the projection set discovery approaches that we presented in Section 10.1 it is possible for the projection sets to partly overlap. This partial overlap occurs when two or more SUBDUE patterns partially overlap in their activities.

### *Ordering Constraints based on SUBDUE patterns*

*Ordering constraints* limit the set of possible expansions of an LPM to some pre-specified set of possible expansions that is smaller than the full set of expansions that expansion function $exp_2$ allows for. We refer to the LPM expansions that are prevented by ordering constraints as *illegal expansions*. We base these illegal expansions on the behavior from which we know that it is either impossible or, at least, very unlikely, based on the set of SUBDUE patterns. The following definition specifies which expansions we consider to be illegal expansions.

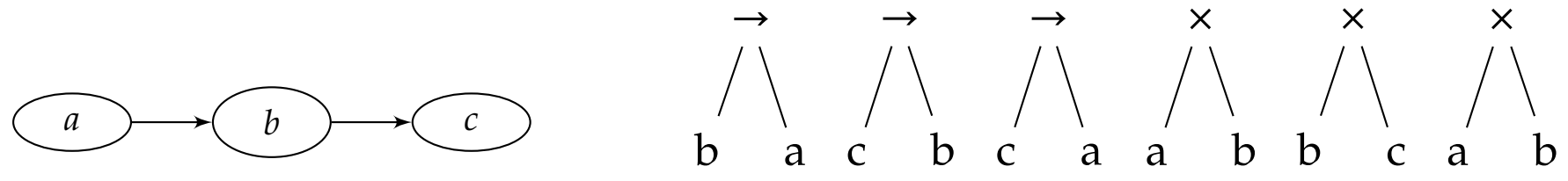

**(a)** Subtrace **(b)** Illegal expansions

**Figure 10.7:** An example of subtrace and the corresponding set of illegal expansions.

**Definition 10.1 (Illegal Expansion based on a SUBDUE pattern).** Given a subgraph $s = (V, E, \phi)$ that is mined by SUBDUE algorithm and an LPM $M \in \mathcal{M}$, a expansion $M' \in exp_2(M)$ is an illegal expansion if $M'$ replaces a leaf node of M by either $\to(b, a)$ or $\times(a, b)$ for some activities $a, b \in \Sigma$ for which there is a path from a to b in SUBDUE pattern s. $\diamond$

Figure 10.7 provides an example of a set of process trees that are inferred as illegal expansions based on an example subtrace. For instance, the subprocess in Figure 10.7a shows that most of the occurrences of activity *a* occurred before *b*, which, in turn, mostly occurred before *c*. Transitively, this means that *a* occurred before *c* as well. Therefore, from this subgraph, we obtain the six illegal expansions that are shown in Figure 10.7b.

Ordering constraints can help reduce the number of process trees to explore, as it narrows down the set of expansions $exp_2$ to those elements that are *not in the set of illegal expansions*. Combinations of multiple ordering constraints can reduce the search space even further. For example, let us assume that SUBDUE finds two subgraphs $\gamma_1, \gamma_2$ that involve activities a and b but reported in opposite execution order such that we have the following two sets of illegal expansions: $IE_{\gamma_1} = \{\to(b, a), \times(a, b)\}$ and $IE_{\gamma_2} = \{\to(a, b), \times(a, b)\}$. By merging these elements, we obtain $IE = IE_{\gamma_1} \cup IE_{\gamma_2} = \{\to(b, a), \to(a, b), \times(a, b)\})$.

The hierarchical structure returned by SUBDUE can be exploited to determine which subgraphs have to be considered during the LPM mining. As explained in Section 10.4.1, low-levels subgraphs in the SUBDUE hierarchy correspond to detailed and often infrequent process variants.

## 10.4.2 Experiments

As a baseline technique, we applied the Markov-based projection discovery technique presented in Section 10.1. We compared the search space reduction and the NDCG obtained when mining with the Markov-based projections with the search space reduction and NDCG obtained when mining with projections and constraints extracted from SUBDUE subgraphs. As a second baseline, we compare with a full search space exploration based on the LPM mining procedure of Chapter 8.

In this section, we explore the effect of exploiting subtrace mining to mine Local Process Models (LPM). To this end, we performed two sets of experiments. In the first, we investigated the effects of the only use of SUBDUE projections. Namely, we used the LPM mining procedure of Chapter 8 and adopting the activities involved in SUBDUE subgraphs as projections, thus neglecting their ordering. In the second set of experiments, we exploited both projections and ordering constraints, using the LPM mining procedure in Section 10.1.

*Dataset* We evaluated our technique using three real-life event logs. The first event log contains execution traces from a financial loan application process at a large Dutch financial institution, commonly referred to as the *BPI 2012* log [Don12]. This log consists of 13087 traces (loan applications) for which a total of 164506 events have been executed, divided over 23 activities. The second event log contains traces from the receipt phase of an environmental permit application process at a Dutch municipality, to which we will refer as the *receipt phase WABO* log [Bui14]. The receipt phase WABO log contains 1434 traces, 8577 events, and 27 activities. The third event log contains medical care pathways of sepsis patients from a medium size hospital, to which we will refer as the *SEPSIS* log [MB17]. The SEPSIS log contains 1050 traces, 15214 events, and 16 activities.

*Settings* We use the implementation of the Markov LPM mining algorithm available in the *LocalProcessModelDiscovery* package[29] of the ProM framework [Ver+10]. For the LPM mining approach that uses SUBDUE projections and constraints, we made an implementation available in ProM package *LocalProcessModelDiscovery-WithSubdueConstraints*[30]. Both for Markov-based LPM mining and SUBDUE-based LPM mining we use the standard configuration in ProM. For SUBDUE we used the implementation provided in `http://ailab.wsu.edu/subdue/`, varying the number of iterations. Note that, roughly speaking, a high number of SUBDUE iterations is expected to have benefits in terms of quality of LPM mining. Higher numbers of iterations lead to a higher number of process behaviors captured by SUBDUE subgraphs and, thus, taken into account during LPM mining. However, this has a negative impact on the speed-up of the LPM mining procedure. Moreover, it is worth noting that, by construction, SUBDUE extracts the largest among the more frequent subgraphs in the earlier iterations. Hence, in practice, we expect the subtraces obtained using SUBDUE even for a low number of iterations to be able to represent most of the process behaviors. According to this observation, we used 1, 5 and 10 iterations for our experiments. Nonetheless, to verify our assumption, we also performed for each dataset an experiment with a high number of iterations (i.e., 10000). Note that for these experiments we derived the LPMs only for the subgraphs of the top-level of SUBDUE hierarchy. All the experiments were performed

[29]`https://svn.win.tue.nl/repos/prom/Packages/LocalProcessModelDiscovery/`

[30]`https://svn.win.tue.nl/repos/prom/Packages/LocalProcessModelDiscoveryWithSubdueConstraints`

on an Intel i7 CPU.

*Results* Table 10.3 shows the results obtained on the three event logs. The results obtained when exploring the full search space, i.e., without applying any projections or constraints, are considered to be the ground truth LPM ranking and therefore have NDCG@k values of 1.0 by definition. For the BPI 2012 log and the receipt phase log the full search space consists of approximate 1.5 million LPMs, while for the smaller SEPSIS log the search space consists of 315k LPMs. The results obtained by the best heuristic configuration(s) are reported in bold.

When using projections mined with Markov projection mining, the search space of LPM mining is reduced by a factor between 43.50x (BPI 2012 log) and 120.21x (Receipt phase log), while the majority of the top 20 LPMs of the ground truth can still be found, resulting in high NDCG values. The table shows that using the constraints extracted from SUBDUE subgraphs after running the SUBDUE algorithm for 10k iterations together with the Markov projections further increases the speed-up of LPM mining on all three logs while resulting in identical LPM rankings.

The number of iterations used in the SUBDUE algorithms when using projections extracted from SUBDUE results is indicated between parentheses. The search space size of LPM mining when using SUBDUE projections depends on the number of iterations used in the SUBDUE mining phase: when a larger number of iterations is used, then a larger number of unique sets of activities will be found to be used as projection sets, leading to a larger LPM search space. At the same time, increasing the number of iterations used in the SUBDUE mining phase increases the quality of the mined LPMs in terms of NDCG. Notice that the quality of the LPM mining results is not very stable when we run the SUBDUE algorithm for only one iteration: on the receipt phase log and the BPI 2012 log it leads to results comparable with or better than the results obtained with Markov projections, but on the SEPSIS log it results in a very small search space and NDCG values. It is worth noting that running the SUBDUE algorithm for 10 iterations leads to a higher speed-up than Markov projections on all logs, even when no SUBDUE constraints are used, while at the same time the quality of the obtained rankings in terms of NDCG is higher. This seems to indicate that using the activity sets of SUBDUE subgraphs is a more effective approach to mine clusters of activities for projection based LPM mining than the Markov based approach. No increase in the quality of the mined LPMs has been found by increasing the number of SUBDUE iterations over 10 iterations; on all three event logs, the quality when using 10k iterations is identical to 10 iterations.

Using constraints extracted from SUBDUE subtrace mining in combination with SUBDUE-based projections has resulted in considerably higher speed-up on all three logs without any loss in quality of the mined LPMs. On all event logs, the highest quality LPMs are mined when using projections and the constraints ex-tracted from the subsequence mined with at least 10 iterations of the SUBDUE algorithm. Additionally, as expected, the speed-up obtained using these projections and constraints is considerably higher than the speed-up obtained when mining LPMs without constraints and with Markov projections, which was the current state-of-the-art approach in LPM mining. Compared to LPM mining with Markov projections, mining with SUBDUE (10) projections and constraints results in an increase in speed-up from 43.50x to 545.32x on the BPI 2012 log, from 120.21x to 668.71x on the receipt phase log, and from 50.04x to 6190.63x on the SEPSIS log. To put these results into perspective: this brings down the mining time on the BPI 2012 log from 24 minutes to less than two minutes.

**Table 10.3:** The experimental results of LPM mining with and without SUBDUE projections and constraints in terms of search space size and quality of the ranking.

| **Event Log** | **Projections** (iterations) | **Constraints** (iterations) | **Search Space Size** | **Speed-up** | **NDCG@5** | **NDCG@10** | **NDCG@20** |
|---|---|---|---|---|---|---|---|
| BPI 2012 | None | None | 1567250 | - | 1.0000 | 1.0000 | 1.0000 |
| BPI 2012 | Markov | None | 36032 | 43.50x | 0.9993 | 0.9987 | 0.9865 |
| BPI 2012 | Markov | SUBDUE (10k) | 21084 | 74.33x | 0.9993 | 0.9987 | 0.9865 |
| BPI 2012 | SUBDUE (1) | None | 10608 | 147.74x | 0.9993 | 0.9987 | 0.9830 |
| BPI 2012 | SUBDUE (5) | None | 10740 | 145.93x | **1.0000** | **0.9994** | 0.9870 |
| BPI 2012 | SUBDUE (10) | None | 10904 | 143.73x | **1.0000** | **0.9994** | **0.9903** |
| BPI 2012 | SUBDUE (10k) | None | 12666 | 123.74x | **1.0000** | **0.9994** | **0.9903** |
| BPI 2012 | SUBDUE (1) | SUBDUE (1) | 2718 | 576.62x | 0.9993 | 0.9987 | 0.9830 |
| BPI 2012 | SUBDUE (5) | SUBDUE (5) | **2620** | **598.19**x | **1.0000** | **0.9994** | 0.9870 |
| BPI 2012 | SUBDUE (10) | SUBDUE (10) | 2874 | 545.32x | **1.0000** | **0.9994** | **0.9903** |
| BPI 2012 | SUBDUE (10k) | SUBDUE (10k) | 4012 | 390.64x | **1.0000** | **0.9994** | **0.9903** |
| Receipt phase | None | None | 1451450 | - | 1.0000 | 1.0000 | 1.0000 |
| Receipt phase | Markov | None | 12074 | 120.21x | 0.9418 | 0.8986 | 0.8238 |
| Receipt phase | Markov | SUBDUE (10k) | 10610 | 136.80x | 0.9418 | 0.8986 | 0.8238 |
| Receipt phase | SUBDUE (1) | None | 8176 | 177.53x | **1.0000** | **0.9994** | 0.9903 |
| Receipt phase | SUBDUE (5) | None | 8256 | 175.81x | **1.0000** | **0.9994** | 0.9903 |
| Receipt phase | SUBDUE (10) | None | 8264 | 175.64x | **1.0000** | **0.9994** | **0.9958** |
| Receipt phase | SUBDUE (10k) | None | 8504 | 170.68x | **1.0000** | **0.9994** | **0.9958** |
| Receipt phase | SUBDUE (1) | SUBDUE (1) | 4012 | 390.64x | **1.0000** | **0.9994** | 0.9903 |
| Receipt phase | SUBDUE (5) | SUBDUE (5) | **1862** | **779.51x** | **1.0000** | **0.9994** | 0.9903 |
| Receipt phase | SUBDUE (10) | SUBDUE (10) | 2170 | 668.71x | **1.0000** | **0.9994** | **0.9958** |
| Receipt phase | SUBDUE (10k) | SUBDUE (10k) | 2178 | 666.41x | **1.0000** | **0.9994** | **0.9958** |
| SEPSIS | None | None | 315451 | - | 1.0000 | 1.0000 | 1.0000 |
| SEPSIS | Markov | None | 6304 | 50.04x | 0.9332 | 0.9148 | 0.8613 |
| SEPSIS | Markov | SUBDUE (10k) | 3768 | 83.72x | 0.9332 | 0.9148 | 0.8613 |
| SEPSIS | SUBDUE (1) | None | 12 | 26287.58x | 0.5763 | 0.3771 | 0.2489 |
| SEPSIS | SUBDUE (5) | None | 334 | 994.46x | 0.9916 | 0.9671 | 0.9472 |
| SEPSIS | SUBDUE (10) | None | 394 | 800.64x | **0.9923** | **0.9692** | **0.9534** |
| SEPSIS | SUBDUE (10k) | None | 1034 | 305.08x | **0.9923** | **0.9692** | **0.9534** |
| SEPSIS | SUBDUE (1) | SUBDUE (1) | **10** | **31545.10x** | 0.5763 | 0.3771 | 0.2489 |
| SEPSIS | SUBDUE (5) | SUBDUE (5) | 174 | 1812.94x | 0.9916 | 0.9671 | 0.9472 |
| SEPSIS | SUBDUE (10) | SUBDUE (10) | 144 | 2190.63x | **0.9923** | **0.9692** | **0.9534** |
| SEPSIS | SUBDUE (10k) | SUBDUE (10k) | 470 | 671.17x | **0.9923** | **0.9692** | **0.9534** |

Before concluding this section, we provide an example of LPM mined from a SUBDUE subgraph, to provide some insights on the benefits of introducing a higher level of structure on subtrace mining output. Figure 10.8a shows the subtrace

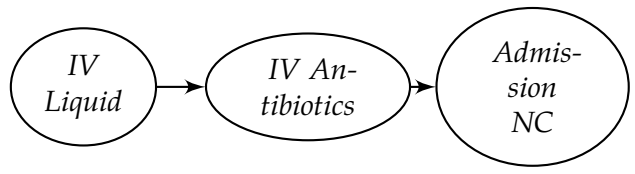

**(a)** Subgraph mined with SUBDUE

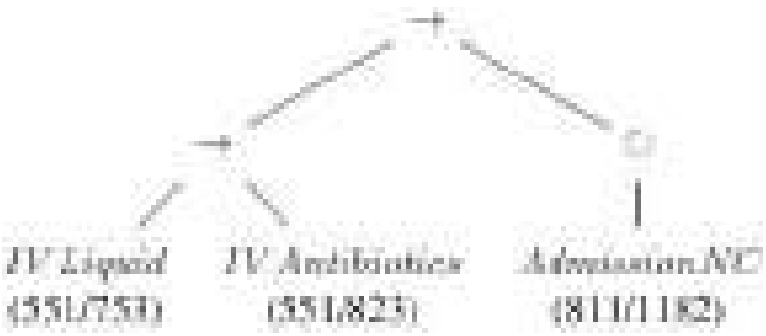

**(b)** Local Process Model

**Figure 10.8:** Subgraph mined with SUBDUE from the SEPSIS log and the Local Process Model showing the true relation between activities *IV Liquid*, *IV Antibiotics*, and *Admission NC*.

mined with SUBDUE from the SEPSIS log from which the projection on activity set $\{IVLiquid, IVAntibiotics, AdmissionNC\}$ was extracted and Figure 10.8b shows one of the LPM mined from that set of activities. While SUBDUE can only mine sequential relations, as in Figure 10.8a, the LPM shows the true relation between the three activities: 551 out of 753 occurrences of *IV Liquid* are followed by *IV Antibiotics* and then one or more occurrences of *Admission NC*. However, 811 out of 1182 instances of *Admission NC* are preceded by the sequence $\langle IVLiquid, IVAntibiotics \rangle$, showing that there are on average almost 1.5 admissions for each liquid and antibiotics.

## 10.5 Related Work

The task of discovering projections plays an important role within the area of decomposed process discovery and conformance checking. Decomposed process discovery aims at partitioning the activities in the event log such that after applying process discovery to each partition of the events, the start-to-end process model can be constructed by stitching together the process models discovered on the individual partitions. In [Aal13a] an approach was introduced to decompose process mining by using a *maximal decomposition* of a causal dependency graph, where the activities associated with each edge in the causal dependency graph end up in one cluster. Hompes et al. [HVA14] describe an approach to make more coarse-grained activity clusters by recombining the clusters of the maximal decomposition by balancing three quality criteria: cohesion, coupling, and balance. Van der Aalst and Verbeek [AV14] introduced a decomposed process mining approach based on *passages*. A passage is a pair of two non-empty sets of activities (X,Y) such that the set of direct successors of X is Y and the set of direct predecessors of Y is X. Munoz-Gama et al. [MCA14] proposed a decomposed conformance checking approach that discovers clusters of activities based on identifying Single-Entry Single-Exit (SESE) blocks in a Petri net model of the process. A SESE block in a Petri net is a set of edges that has exactly two boundary nodes: one entry and one exit. Clustering activities using this approach assumes the availability of a structured process model describing process instances from the beginning to the end, which is different from the application area of LPM discovery.

Carmona et al. [CCK09] describe an approach to generate overlapping sets of activities from a causal dependency graph and uses it to speed-up region theory based Petri net synthesis from a transitions system representation of the event log. The activities are grouped together in such a way that the synthesized Petri nets can be combined into a single Petri net generating all the traces of the log. More recently, Carmona [Car12] described an approach to discover a set of projections from an event log based on Principle Component Analysis (PCA).

All the related projection discovery methods mentioned above aim to discover models that can be combined into a single process model. In the context of LPM discovery, we are not interested in combining process models into one single process model; instead, each LPM is assumed to convey interesting information about a relationship between activities itself. Projection methods in the context of decomposed process mining all aim to minimize overlap between projections, leaving only the overlap needed for combining the individual process models into one. In our case, overlap between clusters is often desired, as interesting patterns might exist within a set of activities $\{a, b, c, d\}$, as well as within a set of activities $\{a, b, c, e\}$, and discovering on both subsets individually is faster than discovering once on $\{a, b, c, d, e\}$. The three projection set discovery methods introduced in this chapter exploit this and aim for overlapping projection sets.

The episode miner [LA14] is a related technique to LPM discovery, which discovers patterns that are less expressive (i.e. limited to partial orders), but is computationally less expensive. While the need for heuristic techniques to speed-up episode mining is limited compared to LPM discovery, in principle the three heuristics described to speed-up LPM discovery could also be used to speed-up the discovery of frequent episodes.

## 10.6 Conclusion

We have explored three different heuristics for the discovery of projection sets for speeding up Local Process Model (LPM) discovery. These heuristics enable the discovery of LPMs from event logs where it is computationally not feasible to discover LPMs from the full set of activities in the log. All three of them produce better than random projections on a variety of data sets. Projection discovery based on Markov clustering leads to the highest speed-up, while higher quality LPMs can be discovered using a projection discovery based on log statistics entropy. The Maximal Relative Information Gain based approach to projection discovery shows unstable performance with the highest gain in LPM quality over random projections on some event logs, while not being able to discover any projection smaller than the complete set of activities on some other event logs. In fact, we would like to explore event log properties that can serve as a predictor for the relative performance of these methods.

Furthermore, we investigated the use of subtrace mining in the generation of projections and constraints to aid the mining LPMs. Specifically, we extended the LPM algorithm to account for ordering constraints that we derived from patterns that were obtained through the SUBDUE algorithm. We evaluated our approach on three real-world event logs. The results show that mining LPMs with SUBDUE projections and constraints outperforms the current state-of-the-art techniques for LPM mining both in quality as well as in computation time. In particular, on two out of the three event logs, the speed-up in computation time compared to the current state-of-the-art technique is more than a factor 10.

# 11 Efficient Mining of LPMs with Time Gap and Event Gap Constraints

**Parts of this chapter have been published as:**

- Niek Tax, Natalia Sidorova, Reinder Haakma Wil, and M. P. van der Aalst: *Mining Local Process Models with Constraints Efficiently: Applications to the Analysis of Smart Home Data*. Proceedings of the International Conference on Intelligent Environments (IE), IEEE: pp. 56–63. (2018)

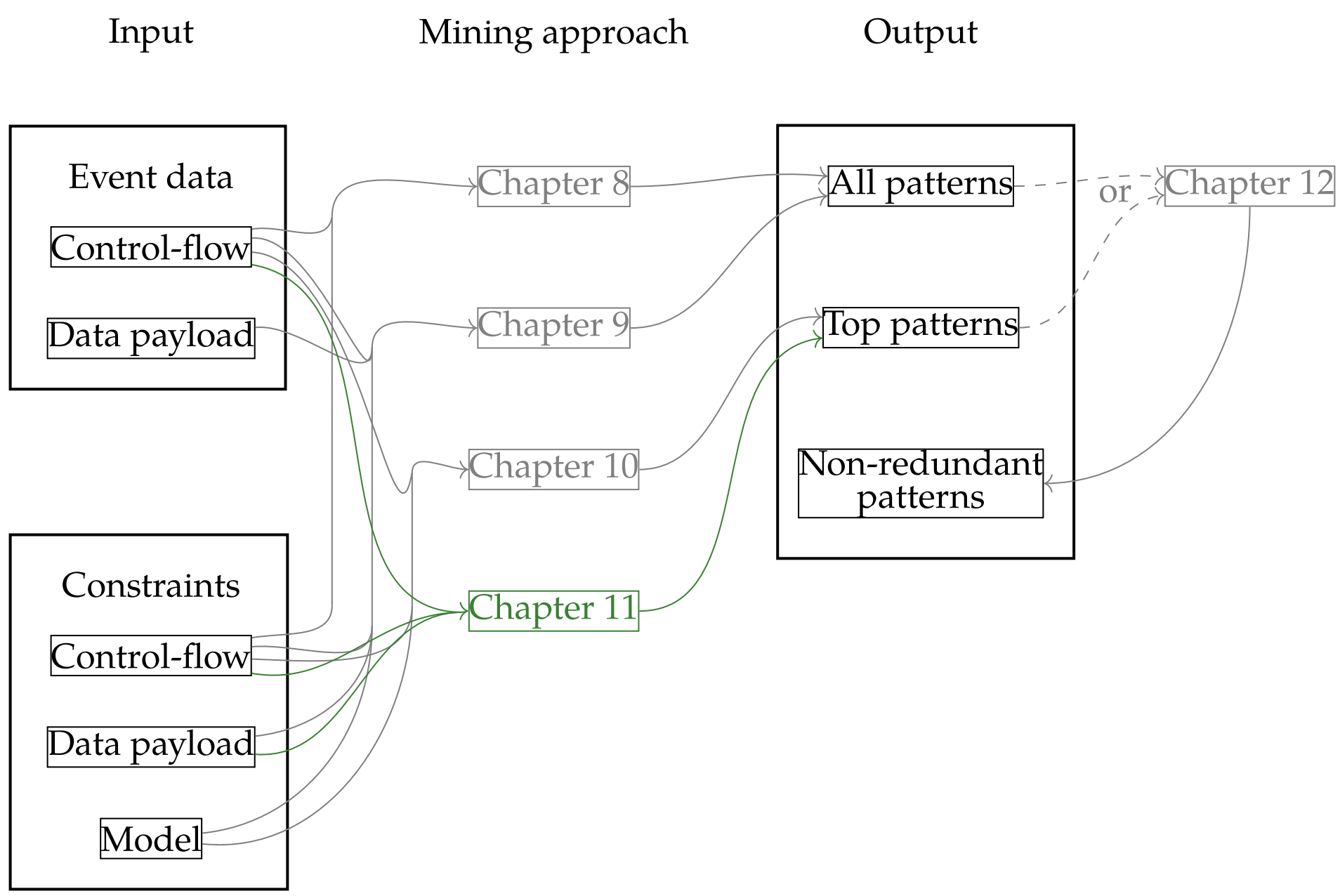

**Figure 11.1:** A taxonomy of local process model techniques.

In Chapter 9 we introduced local process model (LPM) mining using utility functions and constraints. In the case study in Section 9.2.3, we had already seen an application of constraint-based LPM mining in the area of smart home analysis, where LPM patterns can help to generate insight in the habits and routines of the inhabitants of a smart home. In this chapter, we will take a deeper look into applications of LPMs in the analysis of smart home data and into the types of constraints that are needed for such an analysis.

We now revisit the LPM pattern that we mined *with* and mined *without* using time-gap constraints in the smart home case study in Section 9.2.3. These patterns illustrate the need for time gap constraints when mining LPMs for the task of mining insights into human behavior from smart home data.

Figure 11.2a shows the top three LPMs mined from data of household A in an MIT smart home dataset [TIL04] without using any constraints. The LPMs are ranked based on *support* and *confidence* and mined using a *support threshold* of 10, i.e., LPMs in the resulting list of LPMs have occurred at least 10 times. All three LPMs follow the same structure: they specify the behavior that one event is followed by a repetition of a second type of event. For example, the second LPM communicates that 28 out of 34 *exhaust fan* events are followed by a looping behavior of opening the *third drawer in the kitchen*, while 98 out of 110 events of this drawer are preceded by an exhaust fan event.

The first LPM intuitively seems to make sense, as people generally wash their hands after visiting the toilet. However, the sensors described by the other two LPMs seem to be unrelated. We had shown a snippet of the corresponding event log from which these LPMs were mined in Table 9.1. The events of Table 9.1 combined form one instance of the second LPM of Figure 11.2a. However, the time gaps between these events are so large that there does not seem to be any relation between the *exhaust fan* (at 07:23) and the *kitchen drawer* (first occurrence at 11:07).

The problem of unrelated activities ending up together in an LPM can be mitigated either by setting a *time gap constraint* on the maximum time between two consecutive events in the log that fit the LPM, or by setting a *event gap constraint* that sets a maximum on number of events that do not fit the LPM behavior between two consecutive events that do fit the LPM. The effect of such constraints is that fragments in the log that fit the behavior that is described by an LPM are no longer considered to be an instance of the LPM when the constraint is violated, and are not included in the support of the pattern.

Figure 11.2b shows the resulting LPMs when we apply a time gap constraint of 2 minutes. Note that, in principle, these LPMs are also found when we mine LPM *without* time gap constraints. However, in that case, these LPMs are buried deep down in the ranked list of many thousands of LPMs. This becomes clear when comparing the support, confidence, determinism, and coverage values of the LPMs in Figure 11.2a with those in Figure 11.2b.

The first LPM when mining with time gap constraints shows that both *kitchen drawer 1* and *2* are almost always followed by *kitchen cabinet 1* within two minutes. When we look up the instances of this LPM in the event log, they all occur between 17:00 and 19:00, making it likely that this LPM is part of a *cooking* process. The second LPM shows that half of the *washing machine* events are followed within two minutes by the *laundry dryer*. This half of the washing machine events are likely to be the events where the washing machine is opened after the machine finished its washing program, while the other half of the washing machine events are the events where the washing machine was opened to put in the laundry. The third LPM shows that the *medicine cabinet* is generally followed by a sequence of multiple *sink faucet* events following it. These examples illustrate that the constraints help in finding sensible patterns of daily habits where the activities in the LPM are related to each other.

We will show that by using either *time gap constraints* or *event gap constraints* in the mining of LPMs, it is not needed to align the full event sequences to get an accurate count of the support of an LPM. Instead, it is possible to *extract* a smaller log consisting of shorter event sequences based on the constraint and get an accurate support count on this extracted log, which can be aligned faster as it contains shorter sequences. This leads to the situation that the usage of constraints in LPM mining does not only lead to *more insightful patterns* but also enables *more efficient mining*. Note that the algorithms that we introduce in this chapter are orthogonal to the projection set mining heuristics that we introduced in Chapter 10, while sharing the goal of efficient LPM mining. The main difference is that the algorithms of this chapter are specific to time-gap and event-gap constraints while the heuristics of Chapter 10 can be applied without formulating such constraints. Additionally, the heuristics of Chapter 10 operate by projecting on certain sets of activities in the log, while the algorithms that we introduce in this chapter operate by reducing the size of the traces in the log.

In this chapter, we present techniques to perform such extraction of smaller logs and show that these techniques can be used to mine of LPMs efficiently when using time or event gap constraints. In Section 11.1 we show how smaller event logs can be extracted for efficient mining of LPMs with event or time gap constraints. We evaluate our technique on a collection of real-life datasets from existing smart home experiments in Section 11.2. We describe related work in Section 11.3 and conclude this chapter in Section 11.4.

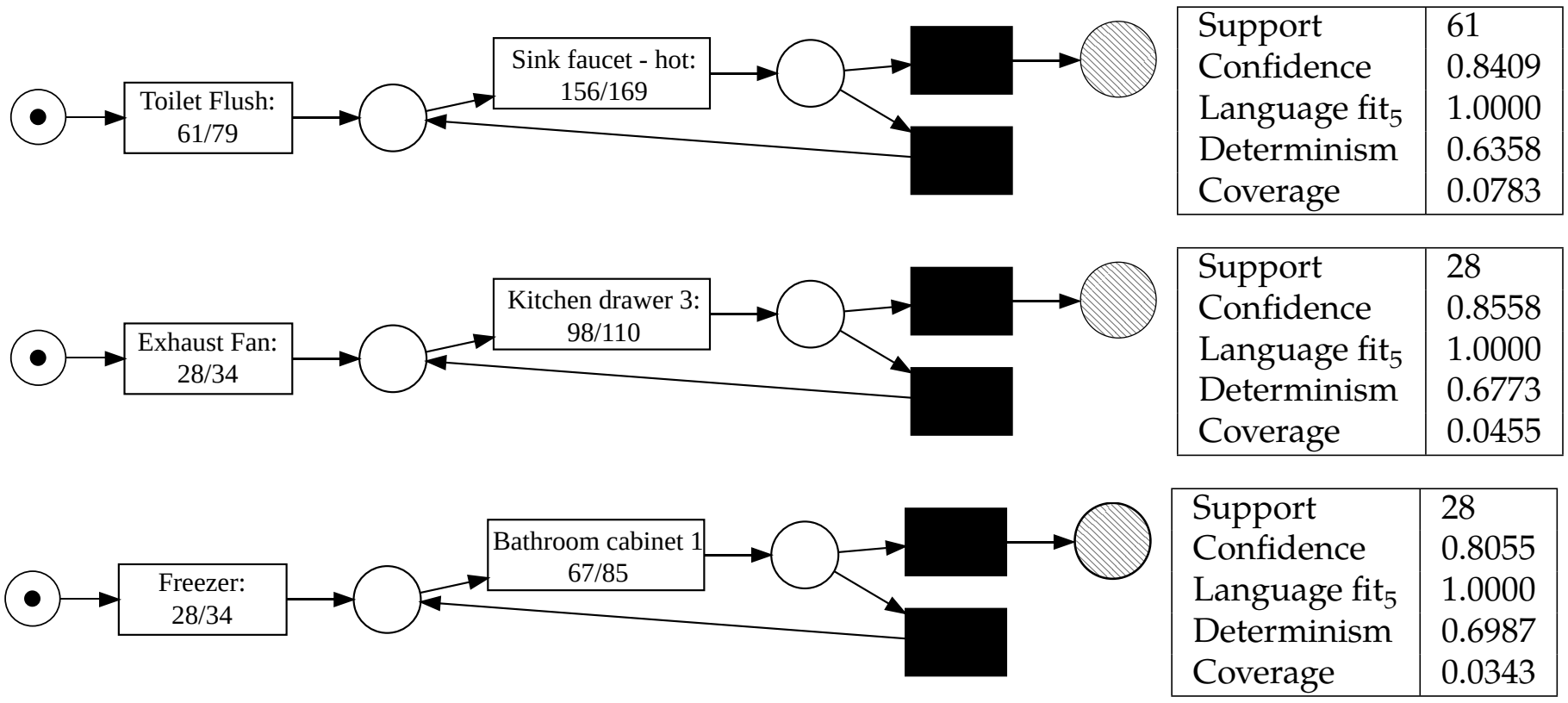

**(a)**

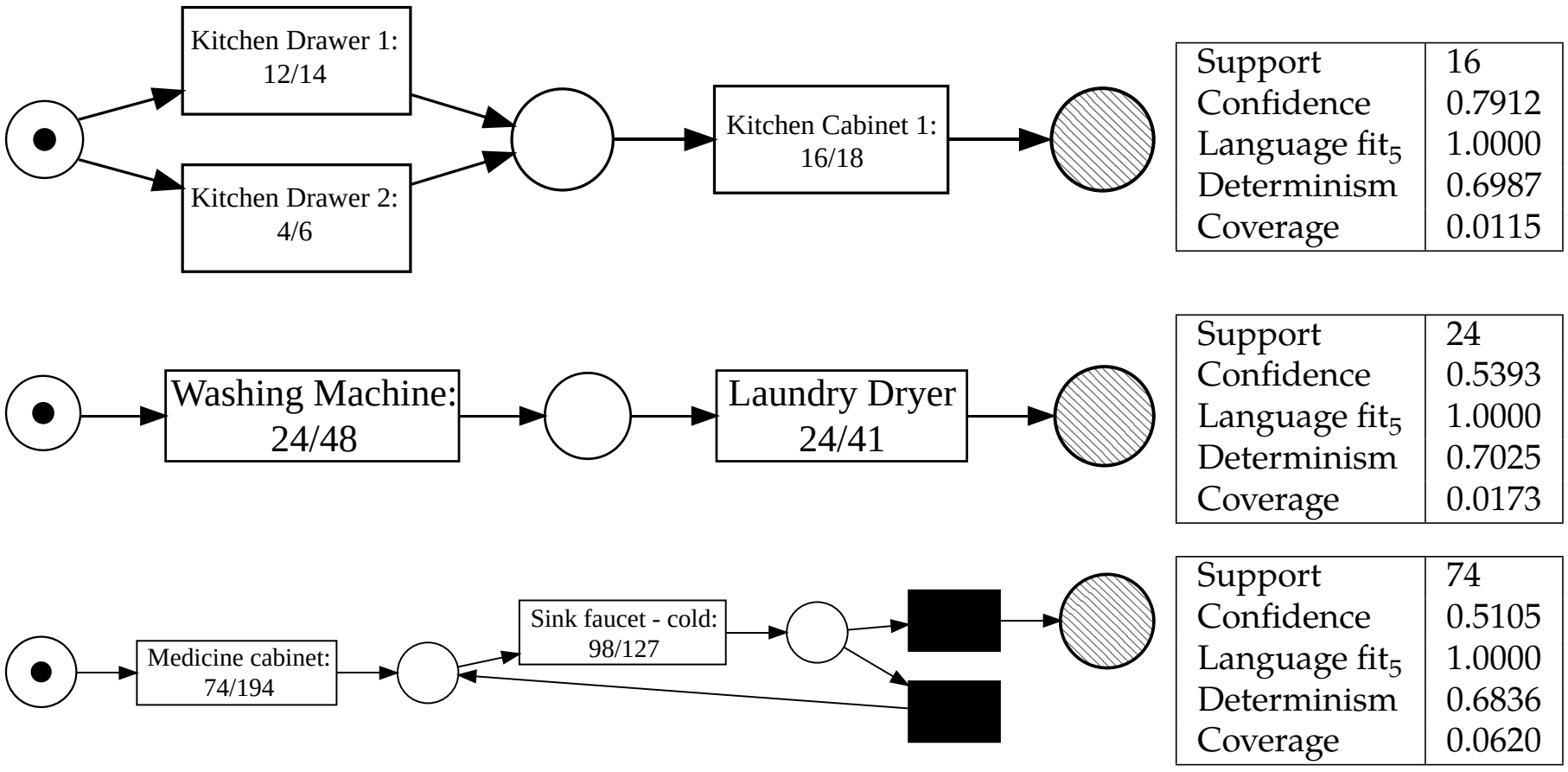

**(b)**

**Figure 11.2:** *(a)* The top three LPMs discovered without using time constraints, and *(b)* the top three LPMs discovered when we set a maximum time gap between two consecutive fitting events of 2 minutes and a maximum LPM instance time of 5 minutes.

## 11.1 Event Log Extraction for Efficient Support Counting

We consider two types of constraints: *event gap constraints* and *time gap constraints*. Both types of constraints put additional requirements on the instances of an LPM with respect to the definition of LPM instance in Definition 8.2 in order for the instance to be counted into the support.

**Event gap constraints** set an upper bound on the maximum number of allowed events that occur in between the events of an instance of a LPM in an event log. For some trace $\sigma^u$ from an unlabeled with log U and given labeling function l, instance $(l(\sigma^u), \lambda)$ is only counted into the constraint-based support of trace $l(\sigma)$ under event gap constraint *max_event_gap* if $\forall_{i \in \{2,\ldots,|\lambda|\}} \lambda(i) - \lambda(i-1) - 1 \leq$ *max_gap*. For example, for an unlabeled trace $\sigma^u$ and a labeling function l such that $l(\sigma^u) = \langle a, x, y, b, z, c \rangle$, the a, b and c events would be identified as an instance of LPM *LPM* $= \rightarrow(a, \wedge(b, c))$, however, the x, y, and z events are not part of the pattern instance, and form a *gaps* within the pattern instance. Therefore, instance $(\sigma^u, \lambda = \langle 1, 4, 6 \rangle)$ contributes to the constraint-based support when the *max_event_gap* parameter is set to 2 or higher, but not when *max_event_gap* $< 2$, since $\lambda(2) - \lambda(1) = 4 - 1 - 1 = 2 \not\leq$ *max_event_gap*. Note that the constraint is put on the *maximum number of consecutive gaps* and not on the *total number of gaps*, e.g., our example instance $(\sigma^u, \lambda = \langle 1, 4, 6 \rangle)$ contains three gaps in total (at indices 2,3, and 5), but at most only two consecutive gaps (indices 2 and 3). An event gap constraint of 0 requires all pattern instances to be consecutive sequences of events that are not interposed by any event that is not from the set of activities that is described by the LPM, i.e., *max_event_gap* $= 0$ does not allow for any event gaps in the instances.

**Time gap constraints** set an upper bound on the maximum time between two events that fit an instance of an LPM, using a parameter *max_time_gap*. For some trace $\sigma^u$ from an unlabeled with log U and given labeling function l, instance $(l(\sigma^u), \lambda)$ is only counted into the constraint-based support of trace $l(\sigma)$ under time gap constraint *max_time_gap* if $\forall_{i \in \{2,\ldots,|\lambda|\}} \pi_t(\sigma^u(\lambda(i))) - \pi_t(\sigma^u(\lambda(i-1))) \leq$ *max_time_gap*. For example, for an unlabeled trace $\sigma^u$ with $\pi_t(\sigma^u) = \langle 0, 1000, 1200, 2000, 2600, 2700, 2800 \rangle$ and a labeling function l such that $l(\sigma^u) = \langle a, x, b, c, a, b, c \rangle$ and for an LPM *LPM* $= \rightarrow(a, \wedge(b, c))$ the instance $(\sigma^u, \lambda = \langle 5, 6, 7 \rangle)$ is counted into the constraint-based support when setting *max_time_gap* to 120 time units since both the gap between 2600 and 2700 and the gap between 2700 and 2800 satisfy the time gap constraint. However, the instance $(\sigma^u, \lambda = \langle 1, 3, 4 \rangle)$ does violate the time gap constraint since $1200 - 0 = 1200$ and $1200 \not\leq 100$, and therefore this instance is not counted for the constraint-based support. Just like with event gaps, the time gap constraints are formulated in terms of the *maximum consecutive gap* and not in terms of the *total gap* (i.e., sum of gaps).

When a constraint is specified that the instances in the log of an LPM have to adhere to, this allows for the extraction of multiple *partial traces* from each trace in to log in such a way that these partial traces completely comply with the specified constraint. For example, given a *time gap constraint* of 500 time units and trace $\sigma$ such that $\pi_t(\sigma)=\langle 0, 1000, 1200, 2000, 2600, 2700, 2800\rangle$, the search for instances of an LPM be performed separately on partial traces $\langle\sigma(1)\rangle$, $\langle\sigma(2), \sigma(3)\rangle$, $\langle\sigma(4)\rangle$, and $\langle\sigma(5), \sigma(6), \sigma(7)\rangle$, instead of searching once on the full original trace, because any subsequence of the original trace that is not contained within any of these four partial traces violates the time gap constraint and can therefore not be an instance of the LPM. Because the computational complexity of the alignment algorithm that is used for support counting is exponential in the length of the trace [Aal12], counting support on the four partial traces is considerably faster counting support on the original traces. Therefore, extracting a modified log $U'$ that contains the partial traces in which the support counting procedure should be applied from a log $U$ can greatly speed up the time needed to count the support of an LPM.

While we have started by giving two specific type of gap constraints in the form of *event gap constraints* and *time gap constraints*, in the general sense, a *gap constraint* is a constraint that is defined in terms of some *distance function* and some *maximal gap*.

**Definition 11.1 (LPM Instance under Gap Constraint).** Given a distance function $D$ and an instance $I = (\sigma, \lambda)$ of process model $M \in \mathcal{M}$ in trace $\sigma \in L$ with landmark $\lambda$ and a maximum gap threshold $max_gap \in \mathbb{R}_{\geq 0}$, $I$ satisfies constraint $max_gap$ if $\forall_{i \in \{1,|\lambda|\}} D(\sigma(\lambda(i)), \sigma(\lambda(j))) \leq max_gap$. ◇

Note that Definition 11.1 does not specify a concrete distance function $D$. In the following subsections we will provide definitions of $D_{event}$ and $D_{time}$ for respectively event and time gap constraints. The task of mining LPMs *under the presence of a gap constraint* is to mine the minimal set of maximal non-overlapping instances (Definition 8.7) that satisfy the support and determinism thresholds as discussed in Chapter 8 and additionally satisfy the gap constraint.

We now continue by describing an event log extraction approach for LPM mining with *event gap constraints*, followed by describing event log extraction approaches for LPM mining with *time gap constraints*.

### 11.1.1 Event Log Extraction for Event Gap Constraints

When we calculate the *support* of an LPM in a trace $\sigma^u$ of an unlabeled event log $U$ using labeling function $l$, we know that each landmark of an instance of that LPM has to start at an index $i$ such that $l(\sigma^u)(i)$ matches one of the start activities of that LPM. The set of possible start activities of a process model is given by function $start : \mathcal{M} \rightarrow \mathcal{P}(\Sigma)$, i.e., $start(LPM) = \{\sigma(1) \mid \sigma \in \mathfrak{L}(LPM)\}$. For example,

```
Input: Unlabeled log U, labeling function l, process model LPM, parameter max_dist,
    distance function D
Output: novel event log U′
    Initialisation :
    events := {(σ, i) | (σ, i) ∈ A(U, l, start(LPM))}
    skipset := ∅, U′ := ∅
    Main Procedure:
    for (σ, i) ∈ events do
        if (σ, i) ∉ skipset then
            j := i, k := i, σ′ := ⟨⟩
            while k ≤ |σ| ∧ D(σ(k), σ(j)) ≤ max_dist) do
                if l(σ)(k) ∈ Activities(LPM) then
                    σ′ := σ′ · ⟨σ(k)⟩, j = k
                end if
                skipset := skipset ∪ {(σ, k)}
                k := k + 1
            end while
            U′ := U′ ∪ {σ′}
        end if
    end for
    return U′
```

**Algorithm 5:** Dynamic Log Extractor for Time Gap Constraints.

$start(\rightarrow(\times(a,d),\wedge(b,c))) = \{a,d\}$. Therefore, when we extract a modified event log $U'$ from the original U to count the constraint-based support of *LPM*, we can leverage the information regarding the possible starting activities of *LPM* by finding the matching events in the log and start looking forward at the in the log from there. Function $A(U,l,act)$ retrieves the set of events of a certain set of activities *act* from an event log, i.e., $A(U,l,act) = \{(\sigma,i) \mid \sigma \in U, i \in \{1,\dots,|\sigma|\}, l(\sigma)(i) \in act\}$. Note this function can be implemented such that the result can be obtained in $O(1)$ when we store a hash table from activities to all events of that activity.

Algorithm 5 describes an approach to generate a modified event log $U'$ from an event log U for an LPM *LPM*, such that the support count of *LPM* is identical in $U'$ as in U given distance function D and a maximum distance *max_dist*. Algorithm 5 can be used to mine LPMs with an *event gap constraint* by using the following *event distance function* $D_{event}$ as distance function D.

**Definition 11.2 (Event distance function).** Given traces $\sigma$, $\sigma'$ and indices $i, j \in \mathbb{N}$, the event distance function is defined as follows:

$$D_{event}(\sigma(i),\sigma'(j)) = \begin{cases} i-j & \text{if } \sigma = \sigma' \wedge j < i, \\ \infty & \text{otherwise.} \end{cases} \qquad \diamond$$

The algorithm iterates over all events that match one of the start activities of *LPM* and starts building a new trace for each such event. Each next event is added to the trace as long as it is in the set of activities of *LPM* and it adheres to the maximum event gap constraint. However, to avoid possible double-counting pattern instances, we keep a set of events *skipset*, which are already part of another trace. For example, given a trace $\sigma$ with $l(\sigma) = \langle a, b, a, b, c, x, x, a, b, c \rangle$ and LPM $LPM = \rightarrow(a, \wedge(b, c))$, $start(LPM) = \{a\}$, and *events* is filled with $\{(\sigma, 1), (\sigma, 3), (\sigma, 8)\}$. When we set $max_gap = 0$, then the trace built starting from $(\sigma, 1)$ results in $\sigma'_{(\sigma,1)} = \langle \sigma(1), \sigma(2), \sigma(3), \sigma(4), \sigma(5) \rangle$. It stops at event $\sigma(5)$, since event $\sigma(6)$ violates the gap constraint, as $l(\sigma)(6) = x$ and $x \notin Activities(LPM)$. Event $\sigma(3)$, however is already part of $\sigma'_{(\sigma,1)}$, therefore, $(\sigma, 3)$ is added to *skipset*, and no trace $\sigma'_{(\sigma,3)}$ is built. Then $\sigma'_{(\sigma,8)} = \langle \sigma(8), \sigma(9), \sigma(10) \rangle$ forms a second trace built from trace $\sigma$. Finally, the alignment-based support counting algorithm recognizes one instance of $LPM_3$ in $\sigma'_{(\sigma,1)} = \langle \sigma(1), \sigma(2), \overline{\sigma(3), \sigma(4), \sigma(5)} \rangle$ and one instance in $\sigma'_{(\sigma,8)} = \langle \overline{\sigma(8), \sigma(9), \sigma(10)} \rangle$, with the elements of the instance indicated with $\overline{\textit{overline}}$. Note that it is necessary to skip $\sigma'_{(\sigma,3)}$, since the alignment-based support counting approach would have recognized $\sigma'_{(\sigma,3)} = \langle \overline{\sigma(3), \sigma(4), \sigma(5)} \rangle$, leading to events $\sigma(3)$, $\sigma(4)$, $\sigma(5)$ being counted as instances of $LPM_3$ twice.

Note that the log $U'$ that Algorithm 5 extracts for a log $U$ is dependent on $start(LPM)$ (line 1) as well as on $Activities(LPM)$ (line 7). Since the LPM mining procedure needs to count the support of many LPMs, it is helpful to use *caching* to store the resulting log $U'$ for a given combination of $start(LPM)$ and $Activities(LPM)$, such that it does not need to be re-computed when at some later point the support of a differ LPM with the same activities and start activities has to be counted.

## 11.1.2 Event Log Extraction for Time Gap Constraints

Algorithm 5 can also be used to mine LPMs with time gap constraints by using the following *time distance function* $D_{time}$ as distance function D.

**Definition 11.3 (Time Distance Function).** Given traces $\sigma$, $\sigma'$ and indices $i, j \in \mathbb{N}$, the time distance function is defined as follows:

$$D_{time}(\sigma(i), \sigma'(j)) = \begin{cases} \pi_t(\sigma(i)) - \pi_t(\sigma(j)) & \text{if } \sigma = \sigma' \wedge j < i, \\ \infty & \text{otherwise.} \end{cases} \qquad \diamond$$

Where Algorithm 5 can be used to extract an event log $U'$ from a log $U$ that is *specific to a given set of activities*, Algorithm 6 describes an alternative approach to generate an event log $U'$ from $U$ in a *non-specific manner*. The benefit of this non-specific approach is that the event log does not have to be re-computed for each LPM, and as an effect, Algorithm 6 only has to generate log $U'$ once, while Algorithm 5 has to re-compute it for each LPM. However, by using information on the set of activities $\Sigma_M$ of the LPM *LPM*, Algorithm 5 is able to decrease the size of the resulting modified log $U'$ compared to $U$ to a higher degree than Algorithm 6

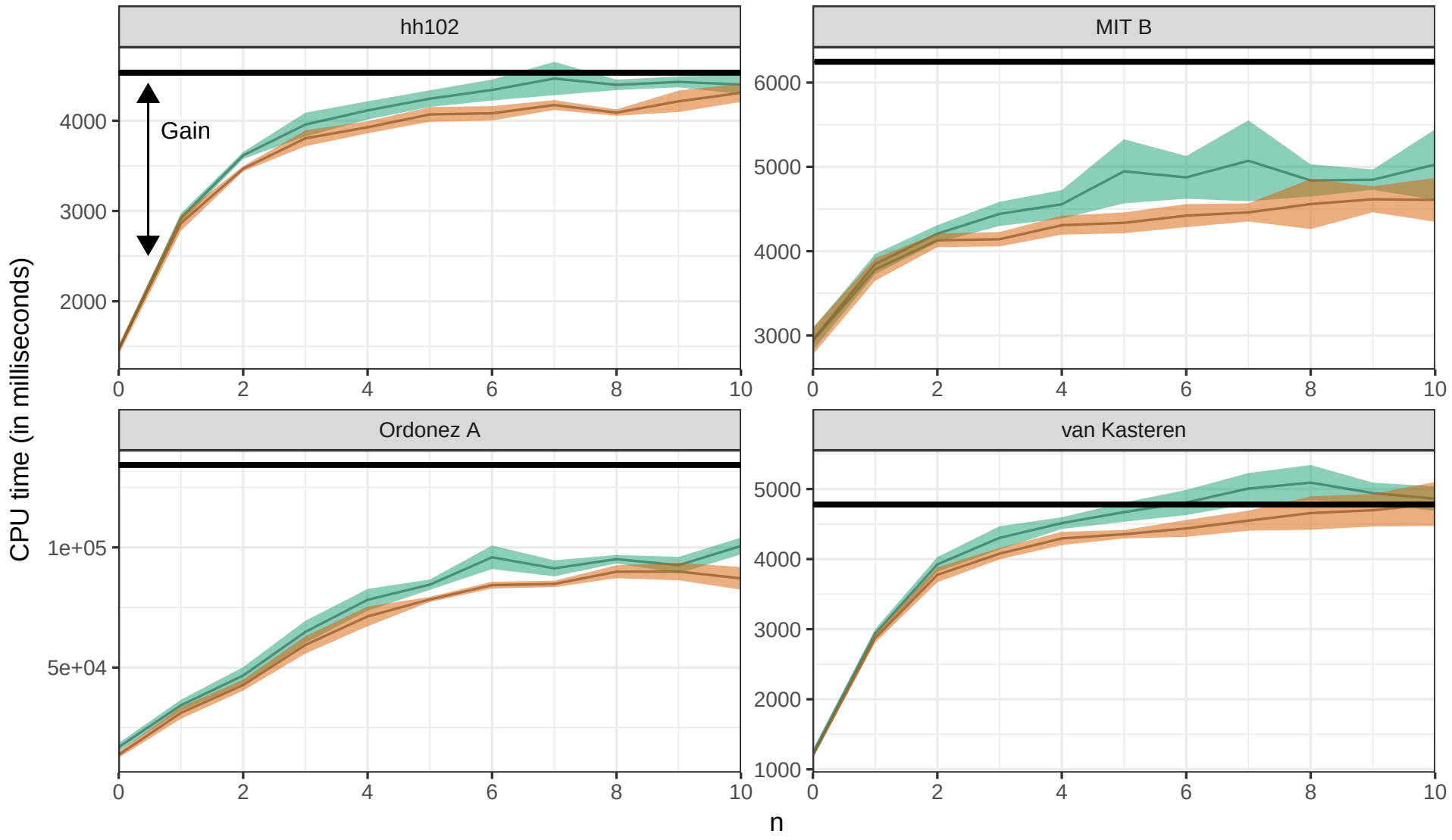

**(a)**

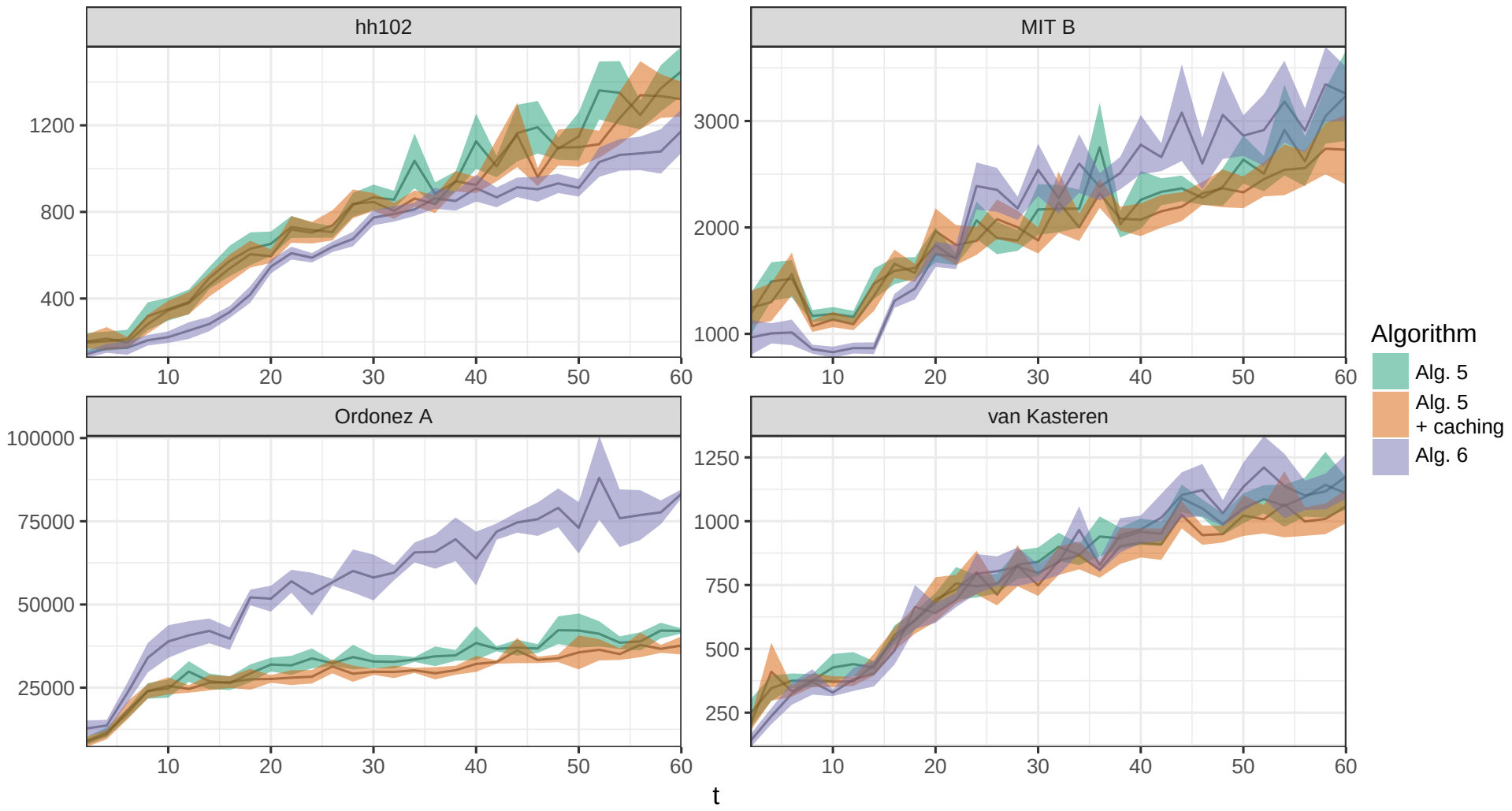

**(b)**

**Figure 11.3:** The CPU time in milliseconds needed for mining LPMs, per algorithm, when using *(a)* an event gap constraint of n events, and *(b)* a time gap constraint of t minutes. The black bar in subfigure *(a)* indicates the required CPU time when no constraints are used as a baseline. This baseline has been omitted from subfigure *(b)*, as for each data set this would yield the same value as the baseline in subfigure *(a)*.

```
Input: event log U, parameter max_gap
Output: novel event log U′
  Initialisation :
  U′ := ∅
  Main Procedure:
  for σ ∈ U do
      σ′ := ⟨⟩, i := 1
      while i ≤ |σ| do
          if i = 1 ∨ (π_t(σ(i)) − π_t(σ(i − 1))) ≤ max_gap then
              σ′ := σ′ · ⟨σ(i)⟩
          end if
          if i=|σ|∨(i≠1∧(π_t(σ(i))−π_t(σ(i−1)))>max_gap) then
              U′ := U′ ∪ {σ′}
              if (π_t(σ(i)) − π_t(σ(i − 1))) > max_gap then
                  σ′ := ⟨σ(i)⟩
                  if i = |σ| then
                      U′ := U′ ∪ {σ′}
                  end if
              end if
          end if
          i := i + 1
      end while
  end for
  return U′
```

**Algorithm 6:** Static Log Extractor for Time Gap Constraints.

for the same value of *max_gap*, leading to a faster support counting procedure. Therefore, the choice between Algorithms 5 and 6 is a trade-off between the time that is spent on preprocessing event log U into a smaller event log U′ on the one hand and the time of the alignment of the modified log on the other hand.

## 11.2 Experiments

We describe the experimental setup in Section 11.2.1, describe the effect of the proposed algorithms on computation time in Section 11.2.2, and show several Local Process Models (LPMs) mined from the event logs in Section 11.2.3.

### 11.2.1 Experimental Setup

We explore the speed-up in computation time of mining LPMs with time gap constraints and event gap constraints using the Algorithms 5 and 6 and compare them with the computation time needed to mine the LPMs when we do not pre-process the event log U into a smaller event log U′. Table 11.1 provides an overview of the real-life smart home datasets used in the experiments, the number of days of logging that they span, their number of events, and their number of activities. For each combination of an algorithm, dataset, and constraint, we measure the computation time twenty times, such that we can accurately estimate the mean and the confidence interval of the computation time needed for the given combination of algorithm, dataset, and constraint. We leave the support and determinism thresholds to the default values in the tool.

**Table 11.1:** An overview of the data used in the experiments.

| Dataset | Days | Events | Activities | Avg. Events/Day |
|---|---|---|---|---|
| CASAS household 102 [Coo+13] | 36 | 576 | 18 | 16.0 |
| MIT household B [TIL04] | 17 | 1962 | 68 | 115.4 |
| Ordónez et al. household A [OTS13] | 15 | 409 | 12 | 27.3 |
| van Kasteren et al. [Kas+08] | 23 | 220 | 7 | 9.6 |

### 11.2.2 Computation Time Results

Figure 11.3a shows the 95% confidence interval for the computation times of Algorithm 5 *with* and *without* caching the extracted log $U'$, for different values $n$ of parameter *max_gap*. Additionally, the computation time *without using any event gap constraint* is shown by the black horizontal bar. This horizontal bar at the same time represents the computation time that is needed for constraint-based LPM mining in the naive way, i.e., by first mining LPMs without constraints using the traditional LPM mining method, and then as a second step filtering out the pattern instances that violate the constraints and adjusting the support accordingly. Therefore, the difference between the lines for Algorithms 5 and 6 with the horizontal black line represents the gain in computational efficiency that is achieved with the techniques introduced in this chapter.

The results show that mining LPMs with event gap constraints can speed up the computation by up to 10x (in case of the Ordónez dataset) when setting *max_gap* to zero. Higher values of *max_gap*, as expected, lead to a lower speedup. Additionally, caching the extracted log $U'$ such that it does not have to be re-computed for succeeding LPMs that have the same alphabet of activities leads to significantly faster computation times on all datasets compares to not caching the extracted logs.

Figure 11.3b shows the 95% confidence interval for the computation times of Algorithms 5 and 6 on the four datasets for different time gap constraints, ranging from 2 to 60 minutes. The results show that Algorithm 5 outperforms Algorithm 6 for Ordónez A, as well as for high values for *max_gap* on MIT B and van Kasteren. The MIT B and Ordónez A datasets are the two datasets with the highest average number of events per trace. However, for CASAS hh102 and on for MIT B and van Kasteren with low values for *max_gap*, the time needed to extract a modified log $U'$ for each LPM instead outweighs the time that is saved on the support counting

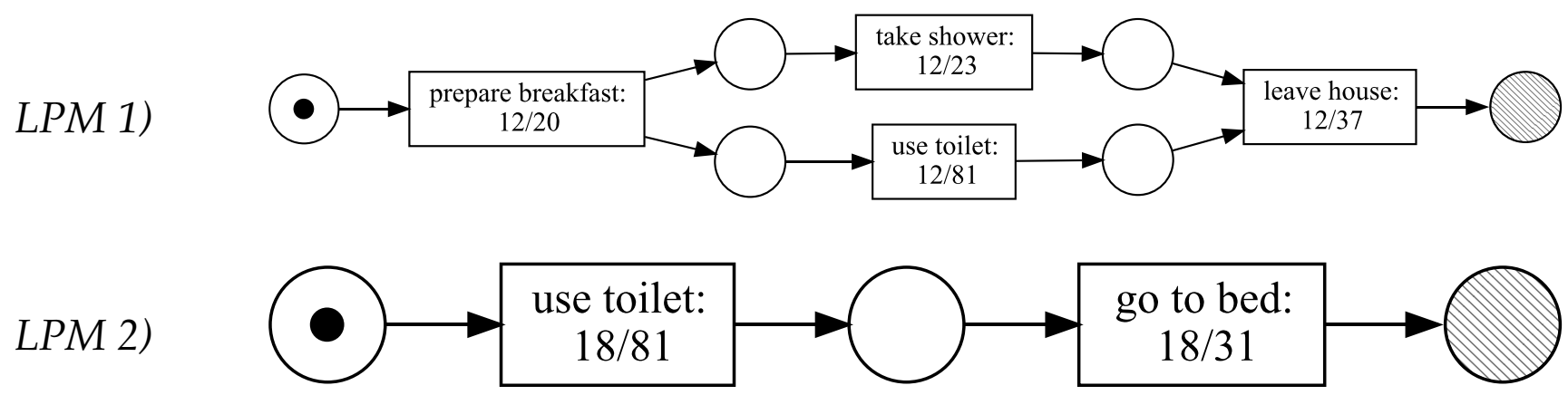

**Figure 11.4:** Two LPMs mined from the van Kasteren [Kas+08] dataset when using a maximum time gap constraint of 20 minutes.

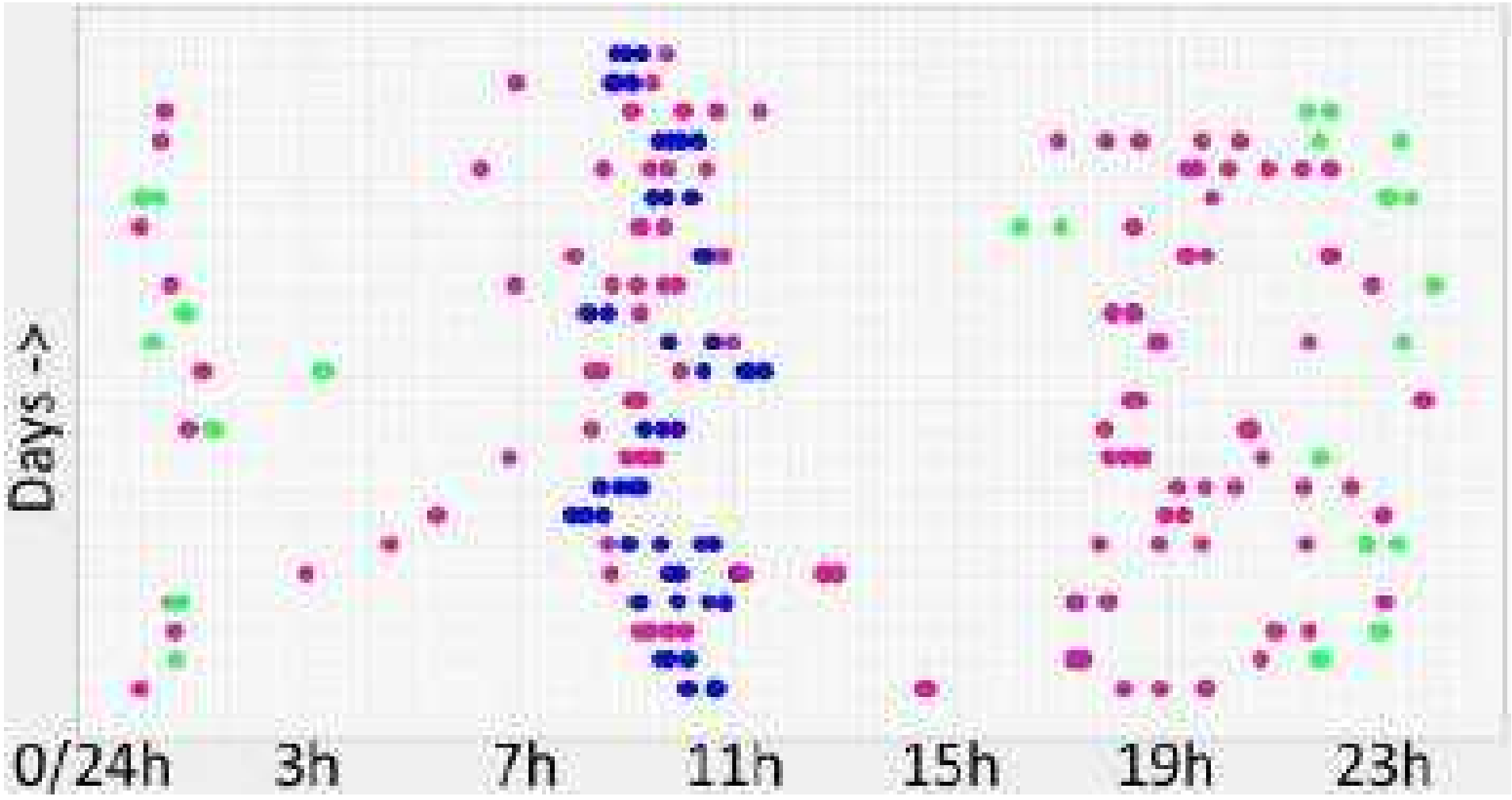

**Figure 11.5:** The van Kasteren data [Kas+08] in a plot with each dot representing one event. The blue and green dots respectively represent events that belong to LPMs 1 and 2 of Figure 11.4.

procedure, and Algorithm 6 which only extracts a modified log U′ once, as a result, outperforms Algorithm 5, even when caching is used. Furthermore, the results show that using caching with Algorithm 5 speeds ups up the computation by a small margin.

### 11.2.3 Local Process Model Results

Figure 11.4 shows the first two LPMs that are mined from the van Kasteren [Kas+08] dataset when we set a maximum time gap constraint of 20 minutes. The first LPM shows that 12 out of 20 *prepare breakfast* events are followed within 20 minutes by both *take shower* and *use toilet* in an arbitrary order, and finally the pattern finishes (again within 20 minutes) with a *leave house* event. The second LPM shows that 18 out of 81 *use toilet* events are followed within 20 minutes by *go to bed*.

The support counting procedure of the original LPM mining algorithm (Chapter 8) relates the patterns back to the log and finds the events that are instances of these patterns. Figure 11.5 plots the events of the van Kasteren log in a scatter plot with the days on the vertical axis and the time within the day on the horizontal axis. The events that are instances of LPM 1 are marked blue, the instances of LPM 2 are marked green, and all other events that are not an instance of either of the two LPMs are marked pink. The figure clearly shows that LPM 1 describes a morning routine, while LPM 2 describes an evening routine. Furthermore, it shows that the subject generally performs his waking up routine somewhere between 10h and 11h in the morning, while he performs his going-to-bed routine often around 23h, and sometimes after midnight. Furthermore, there is one clear anomaly in the going-to-bed routine, where the subject performed LPM 2 just after 16h in the afternoon. These results show that constraints enable the mining of LPMs that capture routines and habits from a smart home dataset in addition to enabling faster LPM mining.

## 11.3 Related Work

In this section, we describe related work in three areas. First, we start in Section 11.3.1 by describing related techniques in the area of mining patterns of daily life from ambient intelligence systems and from smart home environments. We then proceed with Section 11.3.2 in which we will give an overview of techniques for pattern mining with time gap and event gap constraints in the area of pattern mining. A more general discussion on constraint-based pattern mining techniques that either uses other types of constraints than time gap or event gap constraints or that does not specifically address the application domain of smart homes is outside the scope of this related work section. A more general discussion on constraint-based pattern mining techniques, without a restriction to the human behavior application domain or to event gap and time gap constraints, can be found in Section 9.3. Finally, we discuss work from the business process management domain regarding the analysis of time gaps in business process executions in Section 11.3.3.

### 11.3.1 Mining Insights from Smart Homes

Examples of applications for such analysis techniques in this domain include Nazerfard et al. [NRC11], who analyzes temporal relations between activities by clustering them based on their start time and duration, and then mining association rules between those clustered activities.

Jakkula and Cook [JCC07] mine patterns from a smart home data that describe the temporal ordering between two activities as patterns that are expressed in Allen's interval algebra. Lühr et al. [LWV07] focused on association rule mining to mine emergent patterns of human behavior.

### 11.3.2 Pattern Mining with Time Constraints

T-patterns [Mag00] are sequential patterns that specifically aim at generating insights into human behavior, and in addition to other sequential pattern mining techniques take into account the time differences between events. An overview of research focused on generating insights into human behavior using T-patterns, including but not limited research focused on smart home environments, is presented by Casarrubea et al. [Cas+15].

Rashidi et al. [Ras+11] proposed a novel sequence mining algorithm which they call the *Discontinuous Varied-Order Sequential Miner*, which in addition to sequential relations can express a subset of the parallel behavior that can be expressed with LPMs. Rashidi et al. demonstrate its applicability to mining insights in smart home environments. In later work, Rashidi et al. [RC13] improved their sequence mining method by making the support threshold dependent on the type of smart home sensor that registered the human activity. Jung et al. [JC15] proposed a sequential pattern mining technique for smart health services to detect normal and abnormal behavior.

### 11.3.3 Time gaps in the BPM domain

Suriadi et al. [Sur+15] proposed an analysis framework and tool to analyze the performance of a business process while putting special emphasis on gaps in time between consecutive events in the trace. The aim of the technique is to relate gaps in time to other data attributes of the events and cases and thereby find causes and explanations for why events take time.

Somewhat surprisingly, almost no work exists in the area of process discovery where time gaps between events or proximity between events in terms of time is taken into account. To our knowledge, the sole exception is the Fuzzy Miner [GA07]. The Fuzzy Miner creates process models without formal semantics, i.e., it outputs a graph with nodes representing business activities and directed edges between activities that are somehow related to each other, without specifying exactly which execution sequences are allowed by the process. Several measures have been proposed in the initial Fuzzy Miner paper [GA07] to determine which activities to connect with an edge, and the Ph.D. thesis of the first author [Gün09] provides an extended list of such measures. One of those measures, called *proximity correlation* by the authors, takes into account the time proximity between events. We envision that there is much to gain by using time gaps also in traditional process discovery

## 11.4 Conclusions

We have shown that existing Local Process Model (LPM) mining techniques are not able to mine patterns from smart home environment data such that the activities that are described by the LPM are truly related. Furthermore, we have shown that putting a *time gap constraint* on the maximum time between two events within an instance of the LPM, or putting an *event gap constraint* on the maximum number of non-fitting events between two events within an instance of an LPM, enables the mining of LPMs that describe routines or habits from smart home environment data. Moreover, we have shown that it is not needed to use the whole event log when counting the support of an LPM in the case that we are mining LPMs with time gap or event gap constraints, instead, it is possible to count the support on a smaller dataset that can be extracted from the full dataset, allowing for faster support counting while still guarantying the same support.

We have provided two algorithms to extract such an event log from the original event log. Algorithm 5 extracts an event log that is *specific* to the LPM, and therefore the event log extraction procedure needs to be applied again for each LPM for which we count the support. Algorithm 6 extracts an event log that is *generic* and can be used for each LPM, but as a consequence, it is able to shrink the event log to a lesser degree. We have shown with experiments on real-life smart home data sets that Algorithm 5 leads to higher speed up for large event logs, while repeatedly executing the event log extraction procedure does not outweigh the benefits of counting the support on a smaller data set when the original data set is already small, in which case Algorithm 6 leads to higher speedup.

# 12 Mining Non-Redundant Sets of LPMs

**Parts of this chapter have been published as:**

- Niek Tax and Marlon Dumas: *Mining Non-Redundant Local Process Models From Sequence Databases*. Under Review (2019)

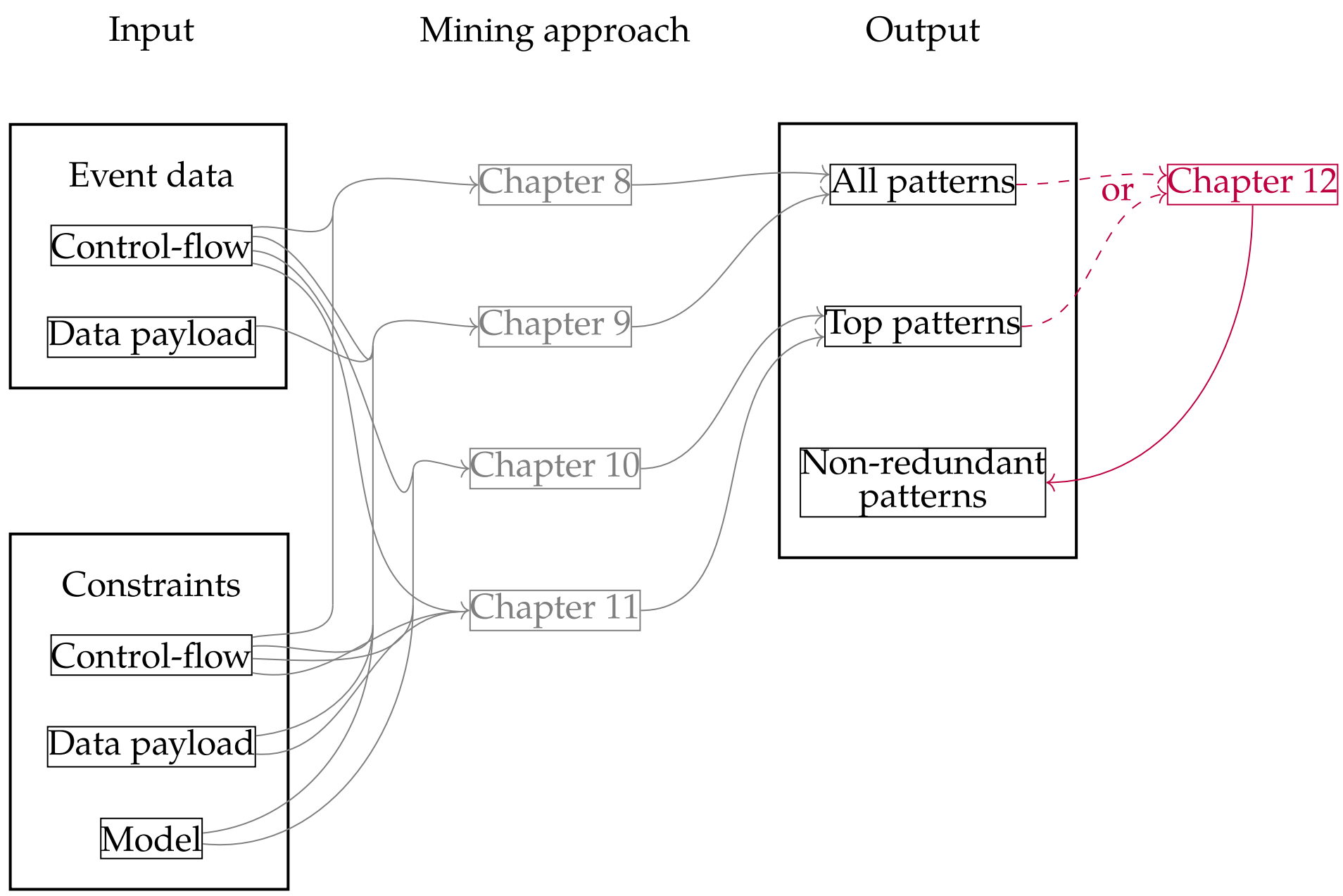

**Figure 12.1:** A taxonomy of local process model techniques.

In the general sense, the practical application of pattern mining techniques for data exploration is sometimes hindered by the fact that they often overload the analyst with a large number of patterns. To address this issue, several approaches have been proposed in the pattern mining literature to describe sequences in a smaller number of patterns. Such approaches include:

**Mining closed patterns [Din+09; YHA03b]** A pattern is a closed pattern if there does not exist an extension of it with the same support.

**Mining maximal patterns [Fou+14b]** A pattern is a maximal pattern if there does not exist an extension of it that meets the minimum support threshold.

**Mining statistically significant patterns [HZZ08; Low+13]** A pattern is a statistically significant pattern if does not just occur frequently, but if the frequency in which the activities in the pattern follow each other is statistically significantly more frequent than what would have been expected by chance.

**Mining compressed patterns [Lam+14; TV12]** The set of compressed patterns is a set of patterns if it reduces the entropy of the sequences according to the *minimal description length* principle.

However, the techniques mentioned above have been developed in the context of sequential patterns. Therefore, these approaches are limited by the fact that each pattern captures only one subsequence.

As introduced in Section 8.5, several pattern mining approaches aim to mine patterns that go beyond sequential patterns and where a pattern is able to capture multiple subsequences. For example, *episodes* [MTV97] extend sequential patterns with parallelism by allowing a pattern to incorporate *partial order relations*. As the issue of pattern overload does not just occur with sequential patterns, [Har+01] proposed a technique to mine closed episodes, while [Pei+06] showed that episode mining techniques cannot mine arbitrary partial orders, and proposes a technique to mine closed *partial order patterns*.

In Chapter 8, we presented local process models as a type of pattern mining that goes beyond existing types of patterns in its generalization capabilities. The approach shares common traits with the CloGSgrow algorithm [Din+09], in the sense that it mines gapped patterns and uses a notion of *repetitive support*, meaning that it counts multiple occurrences of a pattern within the same sequence. Given that the language of a sequential pattern is always a set of size one while the language of an LPM can be a set of arbitrary size, it is possible for a small set of LPMs to cover the behavior of a much larger set of sequential patterns. However, the original LPM mining technique still suffers from the pattern explosion problem because it is designed to extract one LPM at a time (in isolation). This algorithm leads to a set of redundant LPMs.

Table 12.1a shows an event log, and Table 12.1b shows the nine patterns produced by the CloGSgrow algorithm [Din+09] with a minimum support of three. In total, CloGSgrow requires 29 patterns to describe the behavior in this event log. Applying basic LPM mining with a minimum support of three and a minimum determinism of 0.5 leads to 717 patterns, two of which are shown in Figure 12.2. Figure 12.2a expresses that A is followed by B, C, and D, where the D can only occur after C, and the B can occur at any point after A. Figure 12.2b is equivalent to regular

**Table 12.1:** *(a)* An example event log, and *(b)* the patterns extracted with CloGSgrow [Din+09] using $min_sup = 3$.

| ID | Sequence |
|---|---|
| $\sigma_1$ | ⟨E, B, A, B, A, F, A, C, B, D⟩ |
| $\sigma_2$ | ⟨E, B, A, F, E, B, A, B, A, F⟩ |
| $\sigma_3$ | ⟨A, B, C, D, A, C, D, B, E, F⟩ |
| $\sigma_4$ | ⟨A, C, D, B, E, E, B, A, F⟩ |

| Sup | Pattern | Sup | Pattern | Sup | Pattern |
|---|---|---|---|---|---|
| 10 | ⟨A⟩ | 4 | ⟨A, C, D⟩ | 3 | ⟨B, B, A, F⟩ |
| 6 | ⟨A, A⟩ | 3 | ⟨A, E, F⟩ | 4 | ⟨B, B, A⟩ |
| 7 | ⟨A, B⟩ | 10 | ⟨B⟩ | 4 | ⟨B, B, F⟩ |
| 5 | ⟨A, B, A⟩ | 7 | ⟨B, A⟩ | 3 | ⟨B, E, F⟩ |
| 3 | ⟨A, B, A, B⟩ | 5 | ⟨B, A, B⟩ | 4 | ⟨D⟩ |
| 4 | ⟨A, B, A, F⟩ | 3 | ⟨B, A, B, F⟩ | 6 | ⟨E⟩ |
| 4 | ⟨A, B, B⟩ | 5 | ⟨B, A, F⟩ | 3 | ⟨E, B, A, B, A⟩ |
| 3 | ⟨A, B, B, F⟩ | 4 | ⟨B, A, A⟩ | 4 | ⟨E, B, A, F⟩ |
| 3 | ⟨A, C, B⟩ | 6 | ⟨B, B⟩ | 5 | ⟨E, F⟩ |

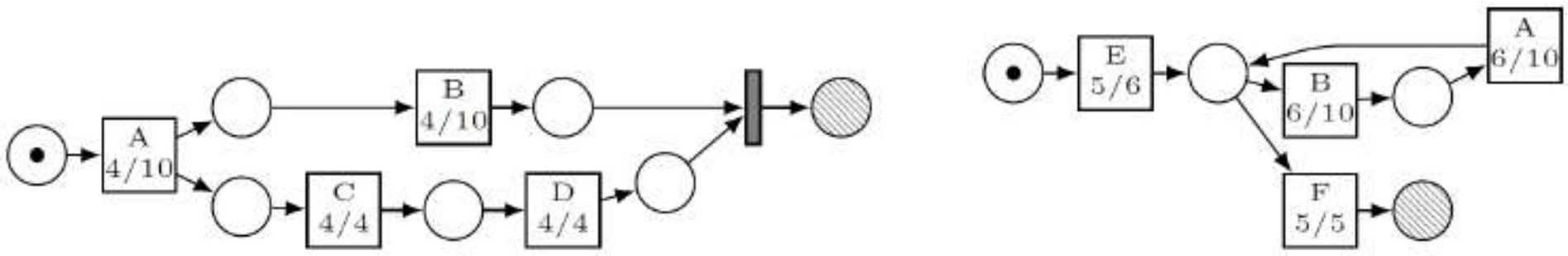

(a) support: 4 (b) support: 5

**Figure 12.2:** Two of the 717 Local Process Models (LPMs) mined from the event log of Table 12.1a using $min_sup = 3$ (visualized as Petri nets).

expression E(BA)*F. The numbers printed in the LPMs respectively indicate the number of events explained by the LPM patterns and the number of occurrences of each activity in the event log, e.g., 4 out of 10 occurrences of activity A in the event log are explained by LPM *(a)*. The four instances of LPM *(a)* are indicated in red in Table 12.1a, and the five instances of LPM *(b)* are indicated in blue. LPMs *(a)* and *(b)* together describe almost all behavior in the event log in a compact manner. While basic LPM mining with a minimum support of three results in 717 patterns, the desired output would be only the two LPMs of Figure 1. In this chapter, we propose heuristics to mine a set of non-redundant LPMs either from a set of redundant LPMs or from a set of gapped sequential patterns such as those produced by CloGSgrow.

Automated process discovery techniques mine a *single* process model that describes the behavior of an event log. The Inductive Miner [LFA13a] is a representative of this family of techniques. Figure 12.3 shows the process model produced by the Inductive Miner when applied to the event log of Table 12.1a. While process discovery techniques produce useful results over simple event logs, they create process models that are either very complex ('spaghetti'-like) or overgeneralizing (i.e., allowing for too much behavior, such as the process model of Figure 12.3) when applied to real-life datasets. In this chapter, we show that on real-life datasets, a set of LPMs can capture the event log more accurately and with lower complexity than a single process model discovered using existing automated process discovery techniques.

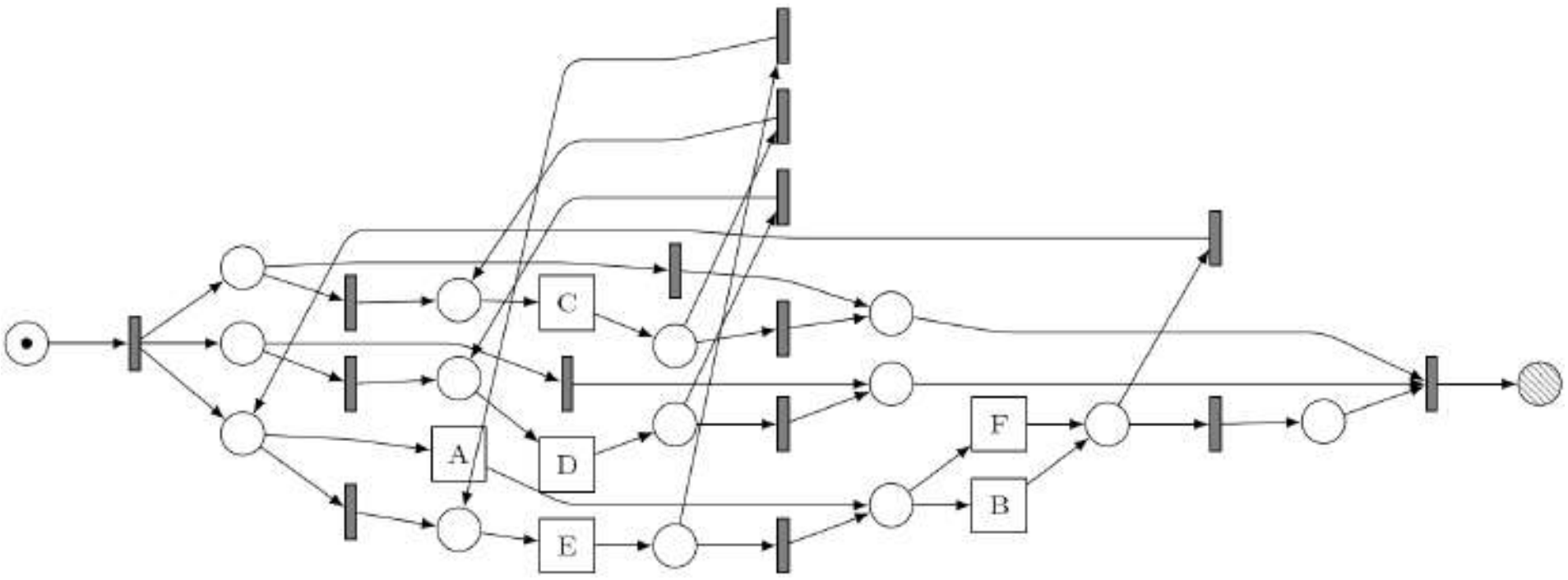

**Figure 12.3:** The resulting process model obtained by applying the Inductive Miner process discovery algorithm [LFA13b] to the event log of Table 12.1a.

This chapter is structured as follows. Section 12.1 outlines quality criteria for LPM lists. Section 12.2 presents the proposed heuristics to discover sets of non-redundant LPMs. Section 12.3 presents an empirical evaluation of the proposed heuristics using real-life event logs. Finally, Section 12.4 discusses related work, while Section 12.5 draws conclusions.

## 12.1 Quality Criteria for Local Process Model Sets

The LPM mining approach that we introduced in Chapter 8 returns a ranked collection of LPMs given an event log $L \in \mathcal{B}(\Sigma^*)$, i.e., a list $LPMS = \{LPM_1, LPM_2, \dots, LPM_n\}$ of LPMs where each $LPM_i \in \mathcal{M}$ satisfies the thresholds on the quality criteria with respect to some event log L. When n is very large, there might be too many patterns for the process analyst to look at. Although in practice the ordering of the patterns in the pattern list helps the process analyst to start the analysis with the patterns that are somehow more interesting according to the quality criteria, these quality criteria measure the quality of an LPM with respect to the log in isolation and do not take into account the other LPMs that were also discovered from that log.

We will now give an example to illustrate the need for quality criteria for LPMs that takes into account the *redundancy* of the whole collection of patterns. Consider the event log L shown in Table 12.1a and Local Process Model list *LPMS* that consist of the three LPMs of Figure 12.4, to which, compared to Figure 12.2, LPM *(c)* is added as one example of a redundant LPM from the list of 717 LPMs. There is overlap in the activities that are described by the LPMs, e.g., LPM *(a)*, *(b)*, and *(c)* all contain a transition that is labeled A. This means that there are multiple candidate patterns for which there is an instance in the log that contains this A-event.

Table 12.2 highlights the instances of the three LPMs in *LPMS* as found by function Γ. The events that are part of an LPM instance are shown in **bold** and a each instance is indicated by an $\overline{\text{overline}}$. As shown earlier, LPMs *(a)* and *(b)* combined explain all events except for the single E-event that is indicated in red. Notice that there is no overlap between LPMs *(a)* and *(b)* in the events that they explain: LPMs *(a)* and *(b)*, therefore, together they provide a near-perfect explanation of the event log.

The main problem with scoring LPMs in isolation is that an event of the event log can be part of a pattern instance of one LPM of the LPM list while at the same time being part of a pattern instance of another LPM. We will define what is means for an LPM to be non-redundant under the presence of another LPM and an event log.

**Definition 12.1 (Non-Redundant Instances of Multiple LPMs).** Given traces $\sigma, \sigma' \in \Sigma^*$, instance $(\sigma, \langle i_1, \dots, i_n \rangle)$ of a process model $M \in \mathcal{M}$ and instance $(\sigma', \langle i'_1, \dots, i'_m \rangle)$ of a process model $M' \in \mathcal{M}$ are *non-redundant* if $\sigma \neq \sigma'$ or if $i'_m < i_1 \vee i_n < i'_1$.

For example, instance $(\sigma_4, \langle 1, 2, 3, 4 \rangle)$ of LPM (a) and instance $(\sigma_4, \langle 3, 4 \rangle)$ of LPM (c) shown in Table 12.2 are not non-redundant, since events 3 and 4 are explained by both models. Therefore, both instances cannot both at the same time be part of a maximal non-redundant set of instances of an LPM list. In order to construct such a maximal non-redundant set of instances, the LPMs in the LPM list are in competition for the events in the log, i.e., each event can be explained only once. In Table 12.2, one could choose to assign $\sigma_4(3)$ and $\sigma_4(4)$ either to LPM *(c)* (i.e., instance $(\sigma_4, \langle 1, 2, 3, 4 \rangle)$) or to LPM(a) (i.e., instance $(\sigma_4, \langle 3, 4 \rangle)$). Choosing LPM *(c)* over *(a)* would lead to two events ($\sigma_4(1)$ and $\sigma_4(2)$, indicated in blue) remaining unexplained for, while instead choosing *(a)* leads to only $\sigma_4(5)$ unexplained. It is clear that there is redundancy in LPM list *LPMS*: LPM *(c)* does not contribute to the set of events that are explained as all three instances of LPM *(c)* are redundant either with respect to either LPM *(a)* or *(b)*. However, when the instances of LPMs are calculated in isolation as we did in Chapter 8, the redundancy in *LPMS* is not observed.

We now propose an approach to score an LPM list in such a way that each event *can only be part of a pattern instance of one LPM*, i.e., the LPMs in the LPM list *compete for the events of the event log* and the contribution of an LPM in the LPM list is only based on the events that it additional describes. This results in a set of instances of the LPMs $LPM_i$ in $LPMS = \langle LPM_1, LPM_2, \dots, LPM_n \rangle$ in a log L such that all the instances of any two LPMs $LPM_i, LPM_j \in LPMS$ are non-redundant. This novel evaluation approach for lists of LPMs is based on the construction of a global model from the LPMs in the list.

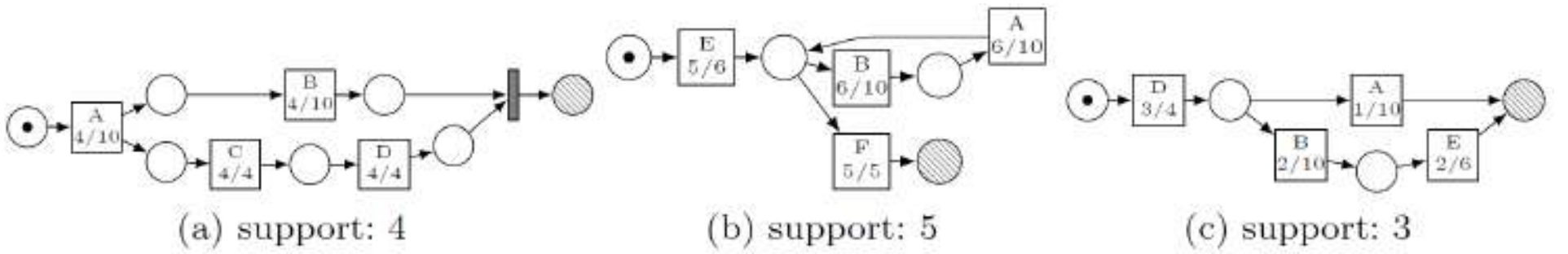

**Figure 12.4:** A Local Process Model set *LPMS* consisting of three example Local Process Models mined from event log L.

**Table 12.2:** An overview of the instances of the three Local Process Models in LPM list *LPMS* in event log L.

| Trace | LPM *(a)* | LPM *(b)* | LPM *(c)* |
|---|---|---|---|
| $\sigma_1$ | $\langle E, B, A, B, A, F, \overline{\mathbf{A}, \mathbf{C}, \mathbf{B}, \mathbf{D}} \rangle$ | $\langle \overline{\mathbf{E}, \mathbf{B}, \mathbf{A}, \mathbf{B}}, \overline{\mathbf{A}, \mathbf{F}}, A, C, B, D \rangle$ | $\langle E, B, A, B, A, F, A, C, B, D \rangle$ |
| $\sigma_2$ | $\langle E, B, A, F, E, B, A, B, A, F \rangle$ | $\langle \overline{\mathbf{E}, \mathbf{B}, \mathbf{A}, \mathbf{F}}, \overline{\mathbf{E}, \mathbf{B}, \mathbf{A}, \mathbf{B}, \mathbf{A}, \mathbf{F}} \rangle$ | $\langle E, B, A, F, E, B, A, B, A, F \rangle$ |
| $\sigma_3$ | $\langle \overline{\mathbf{A}, \mathbf{B}, \mathbf{C}, \mathbf{D}}, \overline{\mathbf{A}, \mathbf{C}, \mathbf{D}, \mathbf{B}}, E, F \rangle$ | $\langle A, B, C, D, A, C, D, B, \overline{\mathbf{E}, \mathbf{F}} \rangle$ | $\langle A, B, C, \overline{\mathbf{D}, \mathbf{A}}, C, \overline{\mathbf{D}, \mathbf{B}, \mathbf{E}}, F \rangle$ |
| $\sigma_4$ | $\langle \overline{\mathbf{A}, \mathbf{C}, \mathbf{D}, \mathbf{B}}, E, E, B, A, F \rangle$ | $\langle A, C, D, B, E, \overline{\mathbf{E}, \mathbf{B}, \mathbf{A}, \mathbf{F}} \rangle$ | $\langle A, C, \overline{\mathbf{D}, \mathbf{B}, \mathbf{E}}, E, B, A, F \rangle$ |

### 12.1.1 Merging Local Process Models into a Global Model

To summarize an event log in the form of LPMs, it is sufficient to have each event described by only one of the LPMs. To obtain an allocation of events to LPMs that provides an optimal number of explained events we transform the set of LPMs into a single process model by merging the places of the initial markings of each LPM in *LPMS* into a single place *mi*, and set as new initial marking $MI = \{mi\}$ of the merged model. Furthermore, we merge the places of the final markings of the LPMs in *LPMS* into a new place *mf*, which we set as new final marking $MF = \{mf\}$ the merged model. We will show how this global model can be used to detect instances of the LPMs in the event log. Formally, given an LPM list $LPMS = \langle LPM_1, LPM_2, \dots, LPM_n \rangle$ with each LPM $LPM_i$ being represented by an accepting Petri net $(N_i, M_{0_i}, M_{f_i})$, with $N_i = (P_i, T_i, F_i, \ell_i)$, we first transform each Petri net $N_i$ into $N_i' = (P_i', T_i, F_i', \ell_i)$ where:

- $P_i' = (P_i \cup \{mi\} \cup \{mf\}) \setminus (M_{0_i} \cup M_{f_i})$
- $F_i' = \{(n, n') | (n, n') \in F_i \wedge n \notin M_{0_i} \wedge n' \notin M_{f_i}\} \cup \{(mi, t) \mid (p, t) \in F_i \wedge p \in M_{0_i}\} \cup \{(t, mf) \mid (t, p) \in F_i \wedge p \in M_{f_i}\}$

A single sequence may contain occurrences of multiple LPMs in the set. Therefore, we add a silent transition $t_{bl}$ connecting the final place to the initial place. This allows the model to accept any concatenation of occurrences of LPMs. Furthermore, it can be the case that a sequence contains no instance of any of the LPMs. Therefore, we redefine the final marking of the merged model to its initial marking to allow it to accept the empty sequence $\langle\rangle$. Formally, given an LPM list $LPMS = \langle LPM_1, LPM_2, \dots, LPM_n \rangle$ with each LPM $LPM_i$ being represented by an ac-cepting Petri net $(N'_i, M_{0_i}, M_{f_i})$, with $N'_i = (P'_i, T_i, F'_i, \ell_i)$, the merged global Petri net representing *LPMS* is a Petri net $(P, T, F, \ell)$ such that:

- $P = \cup_{i=1}^{|LPMS|} P'_i$
- $T = (\cup_{i=1}^{|LPMS|} T_i) \cup \{t_{bl}\}$
- $F = (\cup_{i=1}^{|LPMS|} F'_i) \cup \{(mf, t_{bl}), (t_{bl}, mi)\}$
- $\ell = \begin{cases} \ell_i(t), & \text{if } \exists_{i \in \{1,\dots,n\}} t \in \ell_i \\ \tau, & \text{otherwise} \end{cases}$

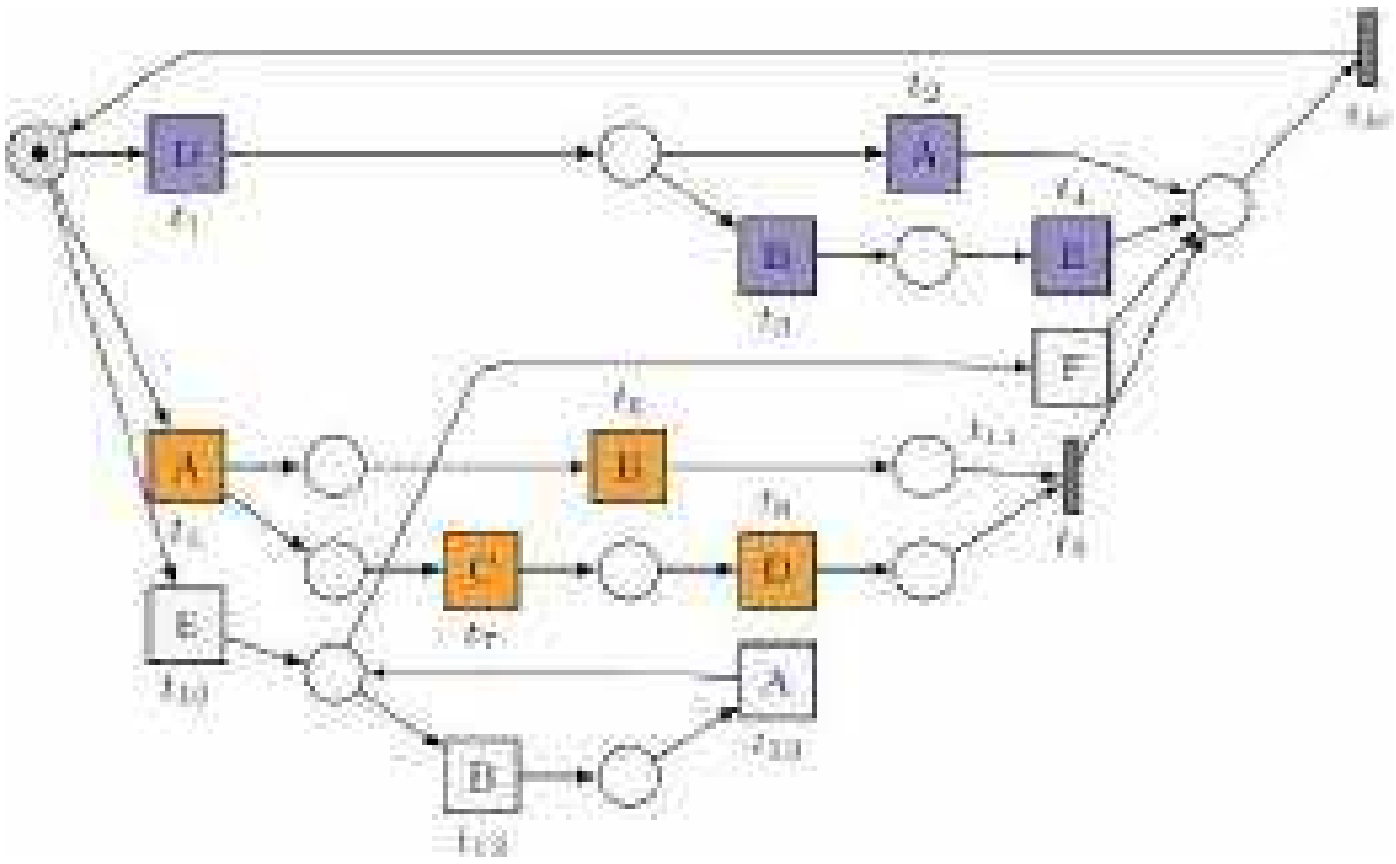

**Figure 12.5:** The global evaluation model for the evaluation of a Local Process Model set consisting of the Local Process Models of Figure 12.4.

Figure 12.5 shows the global process model obtained from *LPMS*. Orange transitions come from the LPM in Figure 12.5*(a)*, white ones from *(b)*, and violet ones from *(c)*.

12

## 12.1.2 Finding Instances of LPMs from an LPM list in an event log

We can calculate the degree to which the LPM list explains the L using alignments on the global model constructed from the LPM list *LPMS* and event log L. The events that are *explained* by *LPMS* are the ones that are mapped to a *synchronous move* in the alignments, while the unexplained events correspond to *moves in log*. However, we want to count *exact* and *complete* observations of the LPMs, while the *moves on model* option in the alignment search space allows for observations of LPM instances where activities are missing. To prevent that we count events that together only approximately (but not completely) form an execution of an LPM as being an instances of that LPM, we calculate alignments in such a way that we enforce each activity in the pattern instance to be represented by an event, i.e., we do not allow for *moves on model*. Therefore, we calculate alignments where we only allow *synchronous moves* and *moves in log*, however, we do additionally allow for *moves in model* in the specific case of silent transitions, as those transitions can only be fired through a *move in model* as they have no corresponding events is the event log. This is however no a limitation, as silent transitions are only in the model for routing purposes and do not describe any activity.

**Table 12.3:** An optimal alignment of sequence ⟨A, C, D, B, E, E, B, A, F⟩ on the global model of Figure 12.5, as obtained with the alignment approach of [AAD12].

| log | A | C | D | B | ≫ | ≫ | E | E | B | A | F |
|---|---|---|---|---|---|---|---|---|---|---|---|
| model | A<br>$t_5$ | C<br>$t_7$ | D<br>$t_8$ | B<br>$t_6$ | τ<br>$t_9$ | τ<br>$t_{bl}$ | ≫ | E<br>$t_{10}$ | B<br>$t_{12}$ | A<br>$t_{13}$ | F<br>$t_{11}$ |

Table 12.3 shows the alignment of the first sequence of the event log, ⟨A, C, D, B, E, E, B, A, F⟩, on the global Petri net that we constructed from the LPM list shown in Figure 12.5. The alignment starts with a synchronous move on activity a, which the model can mimic by firing $t_5$ (enabled in the initial marking). After that, the alignment likewise can perform synchronous moves on activities C, D, and B. Finally, to complete one instance of LPM *(a)*, silent transition $t_9$ is fired to join the two parallel branches. The event log cannot mimic $t_9$, leading to a model move, which is allowed since the transition is silent. Then, a model move is performed on silent transition $t_{14}$, leading to the final marking, where the process model could stop moving. However, the event log contains another instance of an LPM. The event log continues with two E-events, however, the process model has no way to mimic this with two consecutive firings of E, therefore it performs a synchronous move on one of the two E-events and a model move on the other one. The choice is arbitrary which E-event to consider for the synchronous move and which one for the model move, either choice leads to an optimal alignment. After that, synchronous moves on B, A, and F bring the model to the final marking, where it can end.

The alignment of L on the constructed global model allows us to lift the instance detection function Γ from a single LPM to an LPM list *LPMS*, detecting instances $(\sigma, \lambda)$ for which there exists an $LPM_i \in LPMS$ such that $\xi(\sigma, \lambda) \in \mathfrak{L}(LPM_i)$. Furthermore, we use $\Gamma^j_{LPMS}(L)$ to denote the set of instances $(\sigma, \lambda)$ that are assigned to LPM $LPMS(j)$. In the alignment of Table 12.3 we can clearly see how alignments segment the sequence into LPM instances: the model-row of the alignment starts with transitions $t_5$,$t_7$,$t_8$, and $t_6$, which originate from the LPM of Figure 12.4*(a)*, then it has an unmapped e in the sequence-row, and then continues with transitions $t_{10}$, $t_{12}$, $t_{13}$, and $t_{11}$, which originate from the LPM of Figure 12.4*(b)*. This indicates that alignments have segmented the sequence into first an instance $(\sigma, \langle 1, 2, 3, 4 \rangle)$ of the LPM of Figure 12.4*(a)* and then an instance $(\sigma, \langle 6, 7, 8, 9 \rangle)$ of the LPM of Figure 12.4*(b)*.

### 12.1.3 Measuring Coverage and Redundancy of LPM lists

We now continue by proposing quality measures based on the segmentation of an event log in terms of instances of LPMs in a LPM list. First, we define *coverage*, representing the ratio of events that can be explained by one of the LPMs in an LPM list *LPMS*:

$$coverage(\mathrm{L}, LPMS) = \frac{\sum_{\sigma' \in \Gamma_{LPMS}(\mathrm{L})} |\sigma'|}{\sum_{\sigma \in \mathrm{L}} |\sigma|} \tag{12.1}$$

For example, the coverage of our example event log L on the example LPM list *LPMS* of Figure 12.4 is $\frac{38}{39}$, due to the one unexplained e-event.

While coverage measures the share of events of L that is explained by *LPMS*, it does not measure the *degree of redundancy* of *LPMS*.

To measure the degree of redundancy we calculate escaping edges precision of the global model constructed from LPM list *LPMS* with respect to $\Gamma_{LPMS}(\mathrm{L})$ to capture the degree to which the model allows for too much behavior that was not used in the instances of the LPMs in *LPMS*. The obtained value depends on several aspects of *LPMS* which we can summarize in three observations:

**Observation 1** the precision measure punishes the presence of unnecessary LPMs in *LPMS*, i.e., LPMs that model behavior that was not seen in L. Unnecessary LPMs lead to lower precision, as they create additional escaping edges from the start state (i.e., the initial marking).

**Observation 2** the precision measure penalizes overlap in behavior between multiple LPMs in *LPMS*. To see that this is indeed the case, it is important to note that the implementation of Γ deterministically maps behavior in L states in the model. Therefore, when LPMs $LPM_1$ and $LPM_2$ both allow for a run $\langle \mathrm{a}, \mathrm{b}, \mathrm{c} \rangle$ (but both might additionally allow for other runs), all occurrences of $\langle \mathrm{a}, \mathrm{b}, \mathrm{c} \rangle$ in L are either mapped to $LPM_1$ or all occurrences are mapped to $LPM_2$, but not a mix of both. As a result, the overlap in behavior leads either to an escaping edge in $LPM_1$ or in $LPM_2$.

**Observation 3** the precision measure penalizes LPMs that, considered individually, capture too much unobserved behavior.

These three observations correspond to three types of redundancy that might be present in an LPM list. The first observation corresponds to redundancy that is caused by patterns that represent behavior that does not occur in the event log. The second observation corresponds to redundancy that is caused by multiple patterns that (partly) model the same behavior. The third observation corresponds to redundancy within a pattern, i.e., a pattern that models behavior that does not occur in the event log. Given these three observations, we can state that precision calculated on the global model constructed from a set of LPMs can be used to measure the degree of redundancy of that set. Note however that LPM lists without redundancy, i.e., all patterns occur in the event log, no behavior is modeled more than once, and the patterns do not allow for more behavior than what is seen in their instances, will yield a precision of 1. Therefore, precision is inversely related to redundancy, and hence it can be considered to be a measure of non-redundancy.

Since we are interested in LPM list that have both high *coverage* and low *redundancy*, we propose to additionally measure the quality of LPM lists using *F-score*: i.e., the harmonic mean of coverage and non-redundancy.

## 12.2 Local Process Model Collection Mining Approaches

We have implemented all algorithms for mining LPM lists that we are going to introduce in this section, as well as the evaluation approach for LPM lists that we had introduced in Section 12.1. All algorithms and techniques are openly available as part of the process mining tool ProM [Don+05] in the package *LocalProcessModelConformance*[31].

A straightforward approach to post-process the output of the basic LPM mining technique into a smaller set of patterns is to select only those LPMs that actually occur in the event log L according to the evaluation framework in Section 12.1 . To do so, we align the global model constructed from the LPMs in an LPM list *LPMS* to L and filter out any LPM that has no instances in the event log. Algorithm 7 shows the procedure for this LPM list post-processing. We will refer to this approach as the *alignment-based selection* of LPMs. Applying this filter to the LPM list shown in Figure 12.4 and the example event log would result in LPM *(c)* getting filtered out of the LPM list, as it does not have instances in L.

There may exist more than one optimal alignment. For example, sequence $\sigma_1 = \langle e,b,a,f,b,c,d \rangle$, can be aligned such that $\langle \overline{\mathbf{e},\mathbf{b},\mathbf{a},\mathbf{f}},b,c,d \rangle$ is one instance of LPM *(b)*, while alternatively it can be aligned such that $\langle e,b,\overline{\mathbf{a},f,\mathbf{b},\mathbf{c},\mathbf{d}} \rangle$ is one instance of LPM *(a)*, as both alignments provide an explanation for 4 out of the 7 events. When multiple optimal alignments exist, the alignment algorithm is deterministic in which optimal alignments it returns, i.e., identical sequences are always aligned to identical sequences of occurrences of LPMs. $\lambda_1 \cdot \gamma_1^{j_{k_1}} \cdot \lambda_2 \cdot \gamma_2^{j_2} \cdot \ldots \cdot \lambda_n \cdot \gamma_n^{j_n} \cdot \lambda_{n+1}$ such that all segments $\gamma_i^{j_i} \in \mathfrak{L}(LPMS(j_i))$ are assigned to LPM $j_i$, even when there exists an alternative LPM $LPM' \in LPMS$ with $LPM' \neq LPMS(j_i)$ such that $\gamma_i^{j_i} \in LPM'$. Therefore, *alignment-based selection* will select only one of such LPMs to represent $\gamma_i^{j_i}$, thereby reducing the number of instances of *LPM*, and potentially removing it if it has no instances left, thereby reducing the redundancy in the LPM list.

However, given two different sequences $\sigma_1$ and $\sigma_2$ ($\sigma_1 \neq \sigma_2$) with $hd^k(\sigma_1) = hd^k(\sigma_2)$ for some prefix length k, there is no guarantee that the events of $hd^k(\sigma_1)$

[31] https://svn.win.tue.nl/repos/prom/Packages/LocalProcessModelConformance/

**Input:** event log L, LPM list *LPMS*
**Output:** filtered LPM list *LPMS′*
*Initialisation* :
$i = 1$
$LPMS' = \langle\rangle$
*Main Procedure:*
**while** $i \leq |LPMS|$ **do**
  **if** $\{e \in \sigma | \sigma \in \Gamma^{i}_{LPMS}(L)\} \neq \emptyset$ **then**
    $LPMS' = LPMS' \cdot \langle LPMS(i) \rangle$
  **end if**
  $i = i + 1$
**end while**
**return** *LPMS′*

**Algorithm 7:** Alignment-based LPM selection.

and $hd^{k}(\sigma_2)$ are assigned to instances of the same LPMs. To see that this can cause redundancy to remain in the LPM list, consider $\sigma_1 = \langle e, b, a, f, b, c, d \rangle$ and $\sigma_2 = \langle e, b, a, f, b, c, d, a \rangle$, where $hd^{7}(\sigma_1) = hd^{7}(\sigma_2)$. As shown, for $\sigma_1$ there are two possible optimal alignments: as an instance of LPM *(a)* or as an instance of LPM *(b)*. However, $\sigma_2$, has only one optimal alignment, which is the following, where the $\langle \overline{e, b, a, f}, b, c, \overline{d, a} \rangle$, with $\langle e, b, a, f \rangle$ an instance of LPM *(b)* and $\langle d, a \rangle$ an instance of LPM *(c)*. However, if we would be mining from an event log $L = [\sigma_1, \sigma_2]$, the possible optimal alignment of $\sigma_1$ to LPM *(a)* would result in the alignment-based selection to create a redundant set of three LPMs consisting of LPMs *(a)*, *(b)*, and *(c)*, while the alignment of $\sigma_1$ to LPM *(b)* results in an LPM list consisting of only *(b)* and *(c)*.

To alleviate this cause of redundancy, a greedy approach to post-process an LPM list is proposed in Algorithm 8. The intuition behind this algorithm is as follows: first, we select the LPM from the set *LPMS* that explains the highest number of events in the event log L. Then, we filter out all the events from event log L that are already explained, resulting in a new event log $L'$. Iteratively, we search for the LPMs that explain the highest number of events that were still unexplained (i.e., are in $L'$), and update $L'$. We call this approach introduced by Algorithm 8 the *greedy selection* approach. The computational complexity of this algorithm is $\mathcal{O}(n^2)$ with n the number of starting patterns, as after each step in which one pattern is selected, all the other patterns that have not yet been selected need to be considered for the next selection step.

Algorithm 8 removes LPMs from the LPM list without taking into account how much behavior the LPMs itself allow for. The selection of only a small number of LPMs that all allow for many sequences over their activities may still result in a high degree of redundancy. In Algorithm 9 we propose a *direct approach* to greedily select the best combination of LPMs from the input LPM list *LPMS* that leads to the highest F-score according to the evaluation framework. We call this approach

```
Input: event log L, LPM list LPMS
Output: filtered LPM list LPMS'
  Initialisation :
  L' = L, LPMS' = ⟨⟩, candidate_LPMS = LPMS, continue_search = TRUE
  Main Procedure:
  while continue_search ∧ |candidate_LPMS| > 0 do
     i = 1, continue_search = FALSE, max_explained = 0, best_LPM = null
     while i ≤ |candidate_LPMS| do
        if |{e ∈ σ|σ ∈ Γ_candidate_LPMS(i)(L')}| > max_explained then
           max_explained = |{e ∈ σ|σ ∈ Γ_candidate_LPMS(i)(L')}|
           best_LPM = candidate_LPMS(i)
        end if
        i = i + 1
     end while
     if best_LPM ≠ null then
        continue_search = TRUE
        LPMS' = LPMS' · ⟨best_LPM⟩
        L' = L' ↾{e∈σ|σ∈Γ_best_LPM(L')}
        candidate_LPMS = candidate_LPMS↾{best_LPM}
     end if
  end while
  return LPMS'
```

**Algorithm 8:** Greedy alignment-based LPM selector.

the *greedy selection (F-score)* method.

Like Algorithm 8, the computational complexity of Algorithm 9 is also $\mathcal{O}(n^2)$ with n the number of starting patterns, as in both algorithms, all n patterns need to be considered for selection in each iteration and there are at most n possible iterations as each pattern can only be selected once. However, while both algorithms have the same worst case complexity, we expect Algorithm 9 to be slower on the average case, for two reasons. First, the process model on which the alignment needs to be calculated to evaluate the benefit of adding an LPM to the set of selected patterns grows in every iteration in Algorithm 9, where the process model is a global model that is constructed from a number of patterns that is growing with each iteration, while the size of the process model is stable in Algorithm 8, where it depends on the pattern under evaluation only. Secondly, the size of the event log is stable in Algorithm 9, while it shrinks with every step in Algorithm 8.

### 12.2.1 Re-mining of Selected Local Process Models

Algorithms 7-9 simply select a subset of LPMs *LPMS'* from an initial set of LPMs *LPMS*, however, the LPMs in the set themselves are left unchanged, i.e., $\forall_{LPM} \in \text{LPMS}' : LPM \in LPMS$. However, It can be the case that two LPMs $LPM_1, LPM_2 \in LPMS'$ are overlapping in the sequences that they allow for, i.e.,

```
Input: event log L, LPM list LPMS
Output: filtered LPM list LPMS'
  Initialisation :
  LPMS' = ⟨⟩, candidate_LPMS = LPMS, best_fscore = 0, continue_search = TRUE
  Main Procedure:
  while continue_search do
     continue_search = FALSE, i = 1, best_LPM = null
     while i ≤ |candidate_LPMS| do
        if Fscore(L, LPMS' · ⟨candidate_LPMS(i)⟩) > best_fscore then
           best_LPM = LPMS(i)
        end if
        i = i + 1
     end while
     if best_LPM ≠ null then
        continue_search = TRUE
        LPMS' = LPMS' · ⟨best_LPM⟩
        candidate_LPMS minus best_LPM.
     end if
  end while
  return LPMS'
```

**Algorithm 9:** Greedy F-score based LPM selector.

$\mathfrak{L}(LPM_1) \cap \mathfrak{L}(LPM_2) \neq \emptyset$. If such a case, even though $LPM_1$ and $LPM_2$ are both non-redundant patterns, it does indicate that part of the behavior allowed for by $LPM_1$ and $LPM_2$ is redundant. We refer to such type of redundancy as *within-LPM-redundancy*, as opposed to the *between-LPM-redundancy* that Algorithms 7-9 aim to address.

To mitigate *within-LPM-redundancy* from a selected set $LPMS'$, we propose to *re-mine* a process model from the set of occurrences of each LPM, by applying any existing *process discovery* algorithm to the set of pattern instances of an LPM. Algorithm 10 shows the re-mining procedure. *Re-mining* is orthogonal to the selection approaches of Algorithms 7-9, and can be used in combination with any LPM selection procedure. Although re-mining can be done with any process discovery algorithm, we use the Split Miner algorithm [Aug+17; Aug+18]. The Split Miner algorithm has linear time complexity in the number of events and polynomial time complexity in the number of activities of the event log [Aug+18]. Since we apply re-mining to the instances of a pattern in an event log, the number of activities is restricted to only those activities that occur in the pattern, which is a small number in the case of LPM patterns. Therefore, re-mining can in practice be applied efficiently.

**Input:** event log L, LPM list *LPMS*, Process discovery method $PD : \mathcal{B}(\Sigma^*) \rightarrow \mathcal{M}$
**Output:** re-mined LPM list $LPMS'$

```
    Initialisation :
    i = 1, LPMS' = ⟨⟩
    Main Procedure :
    while i ≤ |LPMS| do
        LPMS' = LPMS' · ⟨PD(Γ^i_LPMS(L))⟩
        i = i + 1
    end while
    return LPMS'
```

**Algorithm 10:** Re-mining of an LPM list.

### 12.2.2 Mining an LPM Collection Based on Sequential Pattern Mining

Ding et al. [Din+09] proposed the *CloGSgrow* algorithm to mine closed repetitive gapped sequential patterns from an event log. This technique shares several properties with LPM discovery, which also counts the support of a pattern in a way that is repetitive (i.e., a pattern can occur multiple times per sequence) and gapped (i.e., a pattern instance does not have to be a consecutive subsequence of events). Furthermore, both techniques share the property that instances of the patterns have to be *non-overlapping*, i.e., each event is part of at most one instance of the pattern.

As an alternative to LPM list mining approaches that work by post-processing LPM mining results, we explore an approach to combine multiple sequential patterns together to form more complex, non-sequential LPMs. Function $CloGSgrow(L, min_sup)$ returns a list of closed repetitive gapped sequential patterns ordered in decreasing order by their support. Each sequential pattern $SP$ in the list is represented by a tuple $\langle PT, PI \rangle$, where $PT \in \Sigma^*$ represents the sequence of the pattern and $PT$ represents the set of events of L that are part of a pattern instance of $SP$. $SP_{PT}$ refers to the sequence of sequential pattern $SP$ and $SP_{PI}$ refers to its instances.

Algorithm 11 describes such an LPM mining approach that relies on mining closed repetitive gapped sequential patterns with the CloGSgrow algorithm, and then merges the most similar patterns using a process discovery algorithm. First, an empty set of patterns *LPMS* is initialized. The algorithm first removes all patterns from the CloGSgrow patterns that do not at least explain one event that was not already explained by one of the patterns with more support. The algorithm then clusters together CloGSgrow patterns that are similar in the events of L that they describe, and then applies a process discovery technique to find a generalizing representation for each set of sequential patterns in the form of a process model by applying a process discovery technique. A distance measure *dist* is used for the clustering of sequential patterns, where a pattern is clustered together with another pattern if their distance is less than or equal to *max_dist*. In practice, we propose to use the Jaccard-distance for *dist*, measured between the sets of events of the event log that two sequential patterns describe.

```
Input: event log L, support threshold min_sup, distance threshold min_dist, distance
    measure dist : Σ × Σ → [0,1], Process discovery method PD : B(Σ*) → M
Output: LPM list LPMS
    Initialisation :
clusters = ⟨⟩, explained_events = ∅, LPMS = ⟨⟩
    Main Procedure :
seq_patterns = CloGSgrow(L, min_sup)
while i ≤ |seq_patterns| do
    if (seq_patterns(i)_PI \ explained_events) ≠ ∅ then
        j = 1, closest_clus_dist = ∞, closest_clus_ind = ∞
        explained_events = explained_events ∪ seq_patterns(i)_PI
        while j ≤ |clusters| do
            min_clus_dist = ∞, k = 1
            while k ≤ |clusters(j)| do
                min_clus_dist = min(min_clus_dist, dist(seq_patterns(i), [clusters(j)](k)))
            end while
            if min_clus_dist < closest_clus_dist then
                closest_clus_dist = min_clus_dist
                closest_clus_ind = j
            end if
            j = j + 1
        end while
        if closest_clus_dist < min_dist then
            clusters(closest_clus_ind) = clusters(closest_clus_ind) ∪ {seq_patterns(i)}
        else
            clusters = clusters · ⟨{seq_patterns(i)}⟩
        end if
    end if
    i = i + 1
end while
i = 0
while i ≤ |clusters| do
    LPMS = LPMS · ⟨PD({SP_PT | SP ∈ clusters(i)})⟩
    i = i + 1
end while
return LPMS
```

**Algorithm 11:** Mining an LPM list using CloGSgrow.

The computational complexity of Algorithm 11 is $\mathcal{O}(n)$ in the number of patterns that are found by the CloGSgrow algorithm in order to cluster them based on their similarity. Once the patterns have been clustered, process discovery is applied to each of them, which is polynomial in the number of activities. The computational complexity of this step therefore is dependent on the result of the clustering step, as the number of times that process discovery needs to be applied depends on the number of clusters and the computation time of each time that process discovery is applied depends on the number of activities that occurs in the CloGSgrow patterns in that cluster.

## 12.3 Evaluation

In this section, we evaluate and compare methods for mining LPM lists. Section 12.3.1 introduces the experimental setup and in Section 12.3.2 we discuss the quantitative results of these experiments. Finally, in Section 12.3.3 we present and discuss the patterns that we mined from one of the datasets of the evaluation.

### 12.3.1 Experimental Setup

In this evaluation, we compare algorithms 8, 9, 10, and 12 as introduced in Section 12.2 against four baseline techniques using a collection of real-life sequences databases. Furthermore, we explore the effect of the re-mining approach of algorithm 11. We now continue by detailing the baseline methods, the datasets used for the evaluation, and the evaluation methodology.

*Baseline Methods*

As first baseline approach, we simply select the top-k LPMs that are discovered by the LPM discovery algorithm as an LPM list. Comparison with this baseline gives insight into the effectiveness of the proposed LPM list mining techniques in reducing the redundancy from the originally mined set of LPMs. For event logs with more than 14 activities, we use an approximate heuristic LPM mining technique proposed in Chapter 10 for computational reasons. To evaluate the quality of a set of LPMs we need to calculate alignments, as discussed in Section 12.1. LPM mining can result in many thousands of patterns and the computation of alignments on a large global model that is constructed from so many patterns can become computationally infeasible. Since LPM patterns are ranked on a set of quality criteria such as *support* and *confidence*, we restrict the evaluation to the top 250 LPMs to make it computationally feasible to evaluate the pattern set.

Recall from Section 8.7.2 that Algorithm 4 provided us with a baseline heuristic approach to diversify the set of mined LPMs that works by looking only at the set of activities that are used by the LPMs. This approach starts by selecting the top LPM from the ranking of LPMs obtained by the original LPM mining procedure, and then iterates over the ranking of LPMs, thereby selecting each LPM where the minimal Jaccard-distance of the alphabet of activities in the LPM with the alphabet of activities in one of the already selected LPMs exceeds a minimum *diversity threshold*. We use the set of diversified LPMs obtained with this heuristic approach as a second baseline. This approach is solely based on the set of activities that are included in the LPMs, and in contrast to the approaches introduced in this chapter does not consider the control-flow properties of the LPM. Therefore, we expect this approach to be insufficient to reduce the redundancy of a set of LPMs. However, it is computationally efficient: it is $\mathcal{O}(n^2)$ in the number of starting patterns.

**Table 12.4:** An overview of the event logs used in the experiments.

| ID | Name | Source | Category | # sequences | # events | # activities | Perplexity |
|---|---|---|---|---|---|---|---|
| 1 | BPI'12 | van Dongen[32] | Business | 13087 | 164506 | 23 | 2.79 |
| 2 | SEPSIS | [Man+16b] | Business | 1050 | 15214 | 16 | 3.81 |
| 3 | Traffic Fine | de Leoni & Mannhardt[33] | Business | 150370 | 561470 | 11 | 1.54 |
| 4 | MIT B | [TIL04] | Human behavior | 17 | 1962 | 68 | 10.27 |
| 5 | Ordonez A | [OTS13] | Human behavior | 15 | 409 | 12 | 4.62 |
| 6 | Ordonez B | [OTS13] | Human behavior | 22 | 2334 | 12 | 4.18 |
| 7 | van Kasteren | [Kas+08] | Human behavior | 23 | 220 | 7 | 3.46 |
| 8 | Cook hh102 labour | [Coo+13] | Human behavior | 36 | 576 | 18 | 4.55 |
| 9 | Cook hh102 weekend | [Coo+13] | Human behavior | 18 | 210 | 18 | 5.14 |
| 10 | Cook hh104 labour | [Coo+13] | Human behavior | 43 | 2100 | 19 | 6.58 |
| 11 | Cook hh104 weekend | [Coo+13] | Human behavior | 18 | 864 | 19 | 5.68 |

As third baseline, we create an LPM list consisting of a single process model discovered with a traditional process discovery technique $PD : \mathcal{B}(\Sigma^*) \rightarrow \mathcal{M}$. For a given event log L, this creates an LPM list $LPMS = \langle PD(L) \rangle$. We use two variants: one where we apply the Inductive Miner [LFA13b] algorithm *PD*, and one we apply the Split Miner algorithm [Aug+17; Aug+18]. A comparison with this baseline gives insight in when the mining of *local patterns* is favorable instead of mining a single global model. Both process discovery algorithms are polynomial in the number of activities in the event log.

As a fourth and final baseline we compare our approaches with the sequential patterns that we obtain with CloGSgrow [Din+09] without using the merging approach of Algorithm 11. In order to compare sequential patterns with LPMs, we interpret each sequential pattern as if it is an LPM, i.e., we transform the sequential pattern into a strictly sequential Petri net where the transitions from left to right are labeled according to the sequential pattern.

### *Evaluation Datasets*

We perform experiments on a set of eleven real-life event logs, consisting of three event logs from the *business process management* domain and eight event logs originating from smart home environments. Mining process model descriptions of daily life from smart home event logs is a novel application of process mining that has recently gained popularity [LMM15; Szt+15; Tax+18b; TSA18]. Event data from human behavior has a high degree of variability, which has the effect that traditional process discovery methods that aim to discover a single global process model generate an overgeneralizing model, which motivates the mining of local models instead of a single global from such event logs.

[32] https://doi.org/10.4121/uuid:3926db30-f712-4394-aebc-75976070e91f

[33] https://doi.org/10.1007/s00607-015-0441-1

Table 12.4 provides an overview of the eleven event logs that we include in the experiments and lists their size in terms of the number of sequences, events, and activities. Furthermore, the table lists the *perplexity* of each event log as a measure of the degree of variability (i.e., the randomness) of the behavior in the sequences. The perplexity is calculated using a first order Markov model that is fitted on the event log, i.e., if the next activity of sequence element $t + 1$ can be accurately predicted from the activity of sequence element $t$, then the perplexity is low. The perplexity is the exponentiation of the entropy and offers an intuitive interpretation: if the perplexity is $k$, then the uncertainty is equal to the roll of a $k$-sided dice.

### *Evaluation Methodology*

We measure the coverage, the non-redundancy, and the F-score as introduced in Section 12.1 for each of the techniques and on each of the datasets. We use a different minimum support for each dataset, but we keep it consistent within the dataset and thus mine CloGSgrow patterns and LPMs both with the same minimum support. Furthermore, we are interested in the *complexity* of the resulting LPM lists. While in the sequential pattern mining field it is common to report the complexity of the obtained result in terms of the *number of patterns that are found*, this statistic is not sufficient for the case of LPM mining, since it does not take into account the complexity of the *individual LPMs* in the LPM list. When patterns can contain more complex constructs, like concurrency, loops, and choices, in addition to sequential ordering, patterns themselves can become complex in the sense that certain combinations constructs can require a high cognitive load to understand the pattern.

To measure the complexity of a set of patterns in a way that we take into account the complexity of the individual patterns themselves, we use two measures from the business process modeling field that have been developed to measure the complexity of a business process model. The first metric, the *extended Cardoso measure* [LA09], extends an earlier measure [Car05] for the complexity of control-flow graphs to Petri nets, and is based on the presence of certain splits and joins in the syntactical process definition. This measure captures the effect that the more a process model branches using either choices or into parallel paths, the harder it gets to understand the behavior modeled by the process model. The second measure, the *extended cyclomatic complexity* [LA09], extends the cyclomatic metric of [McC76] to Petri nets and is based on the size of the state-space of the process model. Both measures have been shown to have a relation with the understandability of a process model [Car06; GL07]. We measure the complexity on the global model that we construct from the LPM list. In addition, we also report the number of patterns.

**Table 12.5:** The results of the LPM collection mining methods aggregated over the eleven event logs (mean ± standard deviation).

| Method | Remining | Coverage | Non-Redundancy | F-score | # Patterns |
|---|---|---|---|---|---|
| *Baseline techniques* | | | | | |
| LPM mining | | 0.6971 ± 0.272 | 0.0625 ± 0.018 | 0.1095 ± 0.031 | 67353 ± 94804 |
| Heuristic selection | | 0.4538 ± 0.205 | 0.4021 ± 0.070 | 0.4023 ± 0.139 | *8.0909* ± 3.390 |
| Inductive Miner | | **1.0000** ± 0.000 | 0.0412 ± 0.033 | 0.0788 ± 0.062 | **1.000** ± 0.000 |
| Inductive Miner (20%) | | 0.9257 ± 0.108 | 0.0693 ± 0.032 | 0.1272 ± 0.056 | **1.000** ± 0.000 |
| Inductive Miner (50%) | | 0.8033 ± 0.191 | 0.1083 ± 0.071 | 0.1799 ± 0.108 | **1.000** ± 0.000 |
| Split Miner | | 0.3320 ± 0.304 | 0.1454 ± 0.163 | 0.1881 ± 0.201 | **1.000** ± 0.000 |
| CloGSgrow | | 0.9672 ± 0.086 | 0.0106 ± 0.013 | 0.0187 ± 0.006 | 4524 ± 3466 |
| *New approaches* | | | | | |
| Heuristic selection | ✓ | 0.4538 ± 0.205 | 0.4184 ± 0.064 | 0.4124 ± 0.144 | *8.0909* ± 3.390 |
| Alignment-based selection | | 0.6971 ± 0.272 | 0.1565 ± 0.043 | 0.2354 ± 0.066 | 43.4545 ± 22.967 |
| Alignment-based selection | ✓ | 0.6965 ± 0.272 | 0.1859 ± 0.040 | 0.2687 ± 0.058 | 26.6364 ± 14.438 |
| Greedy selection | | 0.5620 ± 0.230 | 0.3333 ± 0.069 | 0.3904 ± 0.108 | 14.4545 ± 6.251 |
| Greedy selection | ✓ | 0.5571 ± 0.226 | 0.3618 ± 0.083 | 0.4118 ± 0.121 | 14.4545 ± 6.251 |
| Greedy selection (F-score) | | 0.5063 ± 0.201 | **0.5766** ± 0.096 | **0.5191** ± 0.150 | 8.7273 ± 5.236 |
| Greedy selection (F-score) | ✓ | 0.4929 ± 0.189 | 0.5579 ± 0.105 | 0.5055 ± 0.146 | 8.7273 ± 5.236 |
| CloGSgrow merging | | 0.5557 ± 0.101 | 0.4768 ± 0.256 | 0.4596 ± 0.165 | 8.2727 ± 5.293 |

## 12.3.2 Results

Table 12.5 shows the mean coverage, non-redundancy, and F-score for each of the LPM list mining approaches averaged over all 11 event logs. It shows that LPM mining without reducing redundancy indeed results in very high numbers of patterns, results in very redundant LPM lists, and results in LPM lists that have a high coverage of 0.6971. Likewise, the set of CloGSgrow patterns without the merging post-processing steps results in even higher coverage, high redundancy, and high numbers of patterns. As expected, the number of patterns is higher for LPM mining than for CloGSgrow pattern mining, as the non-sequential nature of the patterns allow for more variations. The fact that lower coverage and lower redundancy was found for LPM patterns than for CloGSgrow patterns is likely due to the fact that we selected only the top 250 LPM patterns for computational reasons.

All of the proposed post-processing techniques for LPM lists and for sets of CloGSgrow patterns succeed in bringing down the redundancy and the number of patterns, but they come at the cost of coverage. The different proposed techniques differ in the trade off between coverage and redundancy that they provide. The greedy F-score based selection approach on average results in the lowest redundancy and F-score over the 11 event logs, explaining on average ±50% of the events in the log with a non-redundancy of 0.5766, however this approach reduces the coverage the most with respect to the original set of LPM patterns. On the other end of the spectrum, the alignment-based selection approach does not reduce coverage at all with respect to the original set of LPMs, but it is only able to reduce redundancy to some degree.

The original CloGSgrow patterns, which are not merged into complex non-sequential patterns using the proposed merging procedure, consistently have a high coverage and on average over all the datasets explain almost 97% of the events. Note that this coverage depends on the minimal support that is used in the mining, lower minimum support values correspond to higher coverage values, and, in the extreme, a minimum support of only 1 guarantees a coverage of 1.0. However, lower minimum support also results in higher number of patterns, leading to high redundancy. The CloGSgrow merging approach creates patterns out of the CloGSgrow patterns in a way that this redundancy is substantially reduced, from an average non-redundancy of 0.0106 to an average of 0.4768, however the coverage is significantly reduced to 0.5557. The reduction in coverage is caused by the application of the Split Miner [Aug+17; Aug+18], which doesn't guarantee the discovery of a fitting process model, thereby reducing coverage. The LPM lists that are mined with the CloGSgrow merging procedure on average cover slightly more events than the LPM lists that are discovered with the greedy F-score based approach, however, these LPM lists are on average more redundant. The high standard deviation of the redundancy of the LPM lists obtained with the CloGSgrow merging approach shows that the approach is unstable in the quality of the results, resulting in very redundant LPM lists of some event logs and very non-redundant LPM lists on others.

The Inductive Miner algorithm [LFA13b] provides a formal guarantee that all behavior of the L is contained in the model, therefore, a coverage of 1 is guaranteed. However, it leads to very imprecise process models. When the Inductive Miner with infrequency filtering (20% or 50%) is used, the coverage of the resulting LPM list decreases while non-redundancy slightly increases. The Split Miner [Aug+17; Aug+18] outperforms the Inductive Miner in terms of redundancy, however, the coverage of the resulting process models is unstable, even resulting in a coverage of 0 on some event logs. The coverage values of 0 are caused by process models being generated by the Split Miner where none of the sequences of the event log fits on the model without the need of skipping at least one activity somewhere in the model. The alignment-based selection approach post-processes the LPM mining results leading to lower redundancy, without any loss in coverage. In contrast, the greedy approach and the greedy F-score based approach are not able to post-process LPM mining results without loss in coverage, however, those two approaches are able to obtain a higher reduction of redundancy.

### *Detailed Results*

Table 12.6 shows more fine-grained results by showing the F-score obtained with each of the LPM list mining approaches on each of the 11 event logs individually. While the greedy F-score based selection approach on average results in the highest F-score, the approach based on merging CloGSgrow pattern outperform this approach on two of the eleven datasets (i.e., BPI'12 and MIT B). In contrast, the merged CloGSgrow patterns perform substantially less well on the Ordonez B, the van Kasteren, and the Cook hh102 weekend datasets. We conjecture that the relative performance of the greedy F-score based selection and the CloGSgrow merging approaches are related to the length of the frequent patterns in the data set.

**Table 12.6:** The F-score of the LPM list mining methods per event log.

| Method | R | event log (IDs as shown in Table 12.4) 1 | 2 | 3 | 4 | 5 | 6 | 7 | 8 | 9 | 10 | 11 |
|---|---|---|---|---|---|---|---|---|---|---|---|---|
| *Baseline techniques* | | | | | | | | | | | | |
| LPM mining | | 0.0866 | 0.1033 | 0.0615 | 0.0937 | 0.1103 | 0.0820 | 0.0605 | 0.1306 | 0.1163 | 0.1600 | 0.1521 |
| Heuristic selection | | 0.2491 | 0.4323 | 0.5056 | 0.1263 | 0.4709 | 0.4973 | 0.3238 | 0.4429 | 0.4459 | 0.5140 | 0.4854 |
| Inductive Miner | | 0.1713 | 0.1827 | 0.1939 | 0.0014 | 0.0059 | 0.0844 | 0.1267 | 0.0315 | 0.0037 | 0.1379 | 0.1326 |
| Inductive Miner (20%) | | 0.2184 | 0.2319 | 0.1804 | 0.0217 | 0.1003 | 0.1304 | 0.2169 | 0.1618 | 0.1204 | 0.1137 | 0.1463 |
| Inductive Miner (50%) | | 0.3142 | 0.2155 | 0.1669 | 0.0059 | 0.2488 | 0.3497 | 0.2303 | 0.2032 | 0.1209 | 0.0979 | 0.1329 |
| Split Miner | | 0.0000 | 0.0000 | 0.0000 | 0.0000 | 0.3230 | 0.3004 | 0.5347 | 0.0605 | 0.2855 | 0.0074 | 0.0371 |
| CloGSgrow | | 0.0187 | 0.0215 | 0.0236 | 0.0212 | 0.0209 | 0.0109 | 0.0205 | 0.0148 | 0.0131 | 0.0107 | 0.0301 |
| *Novel approaches* | | | | | | | | | | | | |
| Heuristic selection | ✓ | 0.2491 | 0.4323 | 0.5056 | 0.1281 | 0.4709 | 0.4973 | 0.4559 | 0.5217 | 0.4459 | 0.5140 | 0.4854 |
| Alignment-based selection | | 0.2070 | 0.1863 | 0.2292 | 0.1495 | 0.3227 | 0.1369 | 0.2535 | 0.3053 | 0.2829 | 0.2194 | 0.2909 |
| Alignment-based selection | ✓ | 0.2528 | 0.2323 | 0.3474 | 0.1576 | 0.3416 | 0.2234 | 0.3094 | 0.3229 | 0.3147 | 0.2319 | 0.2999 |
| Greedy selection | | 0.2451 | 0.4387 | 0.5222 | 0.1552 | 0.4275 | 0.3667 | 0.4820 | 0.4725 | 0.4414 | 0.5638 | 0.4275 |
| Greedy selection | ✓ | 0.2451 | 0.4425 | 0.5222 | 0.1555 | 0.4666 | 0.3993 | 0.5624 | 0.4863 | 0.4722 | 0.4599 | 0.4284 |
| Greedy selection (F-score) | | 0.3983 | **0.5750** | **0.6927** | 0.1777 | **0.6017** | **0.6507** | **0.6674** | **0.5670** | 0.5459 | **0.5758** | 0.5327 |
| Greedy selection (F-score) | ✓ | 0.3983 | **0.5750** | **0.6927** | 0.1763 | **0.6017** | 0.5228 | **0.6674** | **0.5670** | **0.5474** | **0.5758** | 0.5327 |
| CloGSgrow merging | | **0.4562** | 0.3785 | 0.4590 | **0.3331** | 0.5261 | 0.3212 | 0.3478 | 0.4866 | 0.1918 | 0.5433 | **0.5685** |

Where the mined LPMs are restricted to at most four activities for computational reasons, the CloGSgrow patterns do not have this restriction, and the patterns that meet the support threshold can be considerably larger. Therefore, long sequences of frequently repeated behavior can be captured in a single CloGSgrow pattern, where multiple LPMs are required, leading to a lower redundancy for the merged CloGSgrow patterns compared to the LPMs selected with the greedy F-score selection. On the other hand, when there are no long frequent patterns, the maximum size restriction of LPMs does not pose a problem. This conjecture is supported by Table 12.7, which shows the pattern length of the ten longest CloGSgrow patterns per event log, as for two of the three event logs on which the CloGSgrow merging procedure outperforms the greedy F-score based approach the mined CloGSgrow patterns were substantially lower than the ones that were mined for the other event logs.

The naive heuristic selection approach one average is outperformed in F-score only by the greedy F-score based approach and the CloGSgrow merging approach. This is remarkable, given that this approach does not take into account overlap in behavior between LPMs and merely looks at which activities are modeled. However, on specific datasets, such as the MIT B dataset, it returns a set of LPMs with considerably lower F-score than other approaches. Given that this naive approach has a very favorable runtime complexity compared to the other approaches, it can be useful as a starting point before moving to more involved but more computationally intensive techniques.

The re-mining procedure with the Split Miner algorithm decreases the redun-dancy and increases the F-score when applied to the results of the heuristic selection, alignment-based selection, and greedy selection. This comes at the price of a minor decrease in coverage. Surprisingly, the re-mining procedure has a negative effect on coverage, redundancy and F-score when used together with the greedy F-score based approach. Furthermore, the Inductive Miner performs much better for event logs 1,2,3, and 7, which are the datasets with the lowest perplexity. Furthermore, the Split Miner performs very well for event log 7. This shows that traditional process discovery approaches perform well when the event log is highly structured, while mining of *local patterns* instead of a *global model* performs better when there is less structure in an event log.

**Table 12.7:** The sequence length of the ten longest CloGSgrow patterns per event log.

| | event log (IDs as shown in Table 12.4) | | | | | | | | | | |
|---|---|---|---|---|---|---|---|---|---|---|---|
| Pattern | 1 | 2 | 3 | 4 | 5 | 6 | 7 | 8 | 9 | 10 | 11 |
| pattern 1 | 48 | 8 | 7 | 154 | 11 | 20 | 8 | 11 | 8 | 11 | 8 |
| pattern 2 | 47 | 8 | 6 | 153 | 11 | 19 | 8 | 11 | 8 | 11 | 8 |
| pattern 3 | 46 | 8 | 6 | 152 | 11 | 18 | 8 | 11 | 8 | 11 | 8 |
| pattern 4 | 45 | 8 | 6 | 151 | 11 | 17 | 8 | 11 | 8 | 11 | 8 |
| pattern 5 | 44 | 8 | 6 | 150 | 11 | 16 | 8 | 11 | 8 | 11 | 8 |
| pattern 6 | 43 | 8 | 6 | 149 | 11 | 15 | 8 | 11 | 8 | 11 | 8 |
| pattern 7 | 42 | 8 | 6 | 148 | 11 | 14 | 8 | 11 | 8 | 10 | 8 |
| pattern 8 | 41 | 8 | 5 | 147 | 11 | 13 | 8 | 11 | 8 | 10 | 8 |
| pattern 9 | 40 | 8 | 5 | 146 | 11 | 12 | 8 | 11 | 7 | 10 | 8 |
| pattern 10 | 39 | 8 | 5 | 145 | 11 | 11 | 7 | 11 | 7 | 10 | 8 |

Table 12.8 shows the patterns, the number of transitions and the complexity of the patterns in terms of extended Cardoso measure and extended cyclomatic complexity. For the LPM mining results, the number of LPMs reports the total number of patterns that are found, but the number of transitions, the extended Cardoso measure, and the extended cyclomatic complexity measure are reported on the top 250 LPMs, as discussed in Section 12.3.1.

The results show that all post-processing approaches for LPM mining results are effective in substantially bringing down the complexity of the LPM list, thereby creating much more understandable results for the user by reducing the pattern overload. It depends per event log which of the algorithms results in the most simple LPM list. It seems that while there are substantial differences between the approaches in computational complexity and in trade-off that the approaches offer between coverage and redundancy, in contrast, the differences between the approaches in pattern complexity are small and cannot be considered substantial.

Traditional process discovery techniques, compared to LPM lists, create only a single "pattern" and this single pattern has a low number of transition compared to the global model that we construct out of LPM lists that we obtain with the proposed

**Table 12.8:** The *number of LPMs* (L) / *number of transitions* (T) / *extended Cardoso measure* (CA) / *extended cyclomatic complexity* (CY) of the LPM list mining methods per event log.

| | | event log (IDs as shown in Table 12.4) | | | | | | | | | | | | | | | |
|---|---|---|---|---|---|---|---|---|---|---|---|---|---|---|---|---|---|
| Method | R | 1 | | | | 2 | | | | 3 | | | | 4 | | | |
| *Baseline techniques* | | L | T | CA | CY | L | T | CA | CY | L | T | CA | CY | L | T | CA | CY |
| LPM mining | | 112365 | 1021 | 530 | 313 | 5919 | 912 | 462 | 223 | 306872 | 729 | 290 | 208 | 6423 | 493 | 503 | 226 |
| Heuristic selection | | 5 | 17 | 16 | 8 | 13 | 44 | 44 | 19 | 6 | 18 | 15 | 9 | 10 | 37 | 38 | 177 |
| Inductive Miner | | 1 | 55 | 41 | 44 | 1 | 33 | 22 | 48 | 1 | 25 | 19 | 67 | 1 | 92 | 31 | 1491 |
| Inductive Miner (20%) | | 1 | 47 | 34 | 30 | 1 | 38 | 29 | 92 | 1 | 27 | 25 | 497 | 1 | 86 | 41 | 791 |
| Inductive Miner (50%) | | 1 | 44 | 41 | 27 | 1 | 39 | 31 | 67 | 1 | 24 | 24 | 438 | 1 | 75 | **29** | 312 |
| Split Miner | | 1 | 138 | 147 | - | 1 | 161 | 186 | - | 1 | 49 | 60 | - | 1 | 327 | 324 | 232 |
| CloGSgrow | | 6478 | 194340 | - | - | 4039 | 10893 | - | - | 2394 | 4325 | - | - | 12474 | 125728 | - | - |
| *Novel approaches* | | | | | | | | | | | | | | | | | |
| Heuristic selection | ✓ | 5 | 17 | 16 | 8 | 13 | 44 | 44 | 19 | 6 | 18 | 15 | 9 | 10 | **33** | 34 | **15** |
| Alignment-based selection | | 16 | 59 | 64 | 30 | 61 | 228 | 233 | 112 | 24 | 81 | 67 | 50 | 45 | 148 | 149 | 67 |
| Alignment-based selection | ✓ | 14 | 44 | 46 | 23 | 50 | 141 | 114 | 76 | 17 | 49 | 45 | 24 | 35 | 105 | 106 | 51 |
| Greedy selection | | 4 | **13** | **13** | 8 | 18 | 66 | 69 | **4** | 13 | 39 | 40 | 17 | 21 | 66 | 67 | 29 |
| Greedy selection | ✓ | 4 | **13** | **13** | 8 | 18 | 66 | 69 | **4** | 13 | 35 | 36 | 15 | 21 | 56 | 62 | 28 |
| Greedy selection (F-score) | | 6 | 28 | 28 | 13 | 2 | **9** | **10** | **4** | 3 | **12** | **10** | **6** | 20 | 76 | 77 | 34 |
| Greedy selection (F-score) | ✓ | 6 | 27 | 22 | 10 | 2 | **9** | **10** | **4** | 3 | **12** | **10** | **6** | 20 | 68 | 68 | 31 |
| CloGSgrow merging | | 3 | 14 | **13** | **6** | 9 | 45 | 38 | 23 | 8 | 34 | 29 | 13 | 7 | 64 | 61 | 46 |

| | | event log (IDs as shown in Table 12.4) | | | | | | | | | | | | | | | |
|---|---|---|---|---|---|---|---|---|---|---|---|---|---|---|---|---|---|
| Method | R | 5 | | | | 6 | | | | 7 | | | | 8 | | | |
| *Baseline techniques* | | L | T | CA | CY | L | T | CA | CY | L | T | CA | CY | L | T | CA | CY |
| LPM mining | | 151624 | 1013 | 509 | 252 | 136887 | 1145 | 504 | 276 | 12036 | 998 | 431 | 234 | 2865 | 850 | 383 | 193 |
| Heuristic selection | | 12 | 38 | 39 | 17 | 4 | **13** | **14** | 6 | 4 | 14 | 12 | 7 | 12 | 42 | 38 | 20 |
| Inductive Miner | | 1 | 47 | 50 | 214 | 1 | 25 | 17 | 72 | 1 | 23 | 21 | 225 | 1 | 37 | 27 | 22 |
| Inductive Miner (20%) | | 1 | 31 | 34 | 1823 | 1 | 21 | 17 | 14 | 1 | 20 | 17 | 96 | 1 | 29 | 21 | 18 |
| Inductive Miner (50%) | | 1 | 27 | 28 | 181 | 1 | 26 | 24 | 17 | 1 | 19 | 18 | 83 | 1 | 32 | **18** | 22 |
| Split Miner | | 1 | 59 | 60 | - | 1 | 84 | 84 | 34 | 1 | 27 | 27 | - | 1 | 32 | 22 | - |
| CloGSgrow | | 6065 | 10896 | - | - | 8963 | 70361 | - | - | 333 | 1630 | - | - | 628 | 5089 | - | - |
| *Novel approaches* | | | | | | | | | | | | | | | | | |
| Heuristic selection | ✓ | 12 | 38 | 39 | 17 | 4 | **13** | **14** | 6 | 4 | 14 | 12 | 7 | 12 | 35 | 32 | 17 |
| Alignment-based selection | | 43 | 146 | 144 | 72 | 91 | 366 | 323 | 179 | 20 | 87 | 83 | 44 | 33 | 117 | 114 | 53 |
| Alignment-based selection | ✓ | 16 | 89 | 87 | 41 | 55 | 161 | 161 | 76 | 14 | 48 | 48 | 21 | 23 | 79 | 78 | 37 |
| Greedy selection | | 12 | 38 | 39 | 19 | 13 | 44 | 44 | 19 | 7 | 22 | 19 | 14 | 18 | 59 | 61 | 26 |
| Greedy selection | ✓ | 12 | 30 | 30 | 16 | 13 | 41 | 41 | 18 | 7 | 21 | 19 | 12 | 18 | 54 | 54 | 24 |
| Greedy selection (F-score) | | 12 | 41 | 39 | 21 | 7 | 23 | 23 | 10 | 4 | **12** | **10** | **6** | 10 | 31 | 30 | 14 |
| Greedy selection (F-score) | ✓ | 12 | 41 | 39 | 21 | 7 | 22 | 23 | 12 | 4 | 13 | 12 | **6** | 10 | 31 | 30 | 14 |
| CloGSgrow merging | | 6 | **20** | **19** | **6** | 4 | 15 | **14** | 7 | 5 | 16 | 16 | **6** | 7 | **25** | 24 | **12** |

| | | event log (IDs as shown in Table 12.4) | | | | | | | | | | | |
|---|---|---|---|---|---|---|---|---|---|---|---|---|---|
| Method | R | 9 | | | | 10 | | | | 11 | | | |
| *Baseline techniques* | | L | T | CA | CY | L | T | CA | CY | L | T | CA | CY |
| LPM mining | | 2427 | 831 | 368 | 203 | 1208 | 899 | 427 | 215 | 2256 | 916 | 413 | 211 |
| Heuristic selection | | 10 | 37 | 32 | 18 | 6 | **19** | 19 | **9** | 7 | **22** | 22 | **10** |
| Inductive Miner | | 1 | 30 | **19** | 331 | 1 | 26 | **11** | 20 | 1 | 28 | 15 | 20 |
| Inductive Miner (20%) | | 1 | 29 | 24 | 107 | 1 | 38 | 25 | 23 | 1 | 24 | **11** | 18 |
| Inductive Miner (50%) | | 1 | **26** | 27 | 1172 | 1 | 32 | 18 | 22 | 1 | 25 | **11** | 20 |
| Split Miner | | 1 | 59 | 59 | - | 1 | 264 | 298 | - | 1 | 119 | 121 | - |
| CloGSgrow | | 2225 | 12609 | - | - | 3698 | 14931 | - | - | 2469 | 13589 | - | - |
| *Novel approaches* | | | | | | | | | | | | | |
| Heuristic selection | ✓ | 10 | 35 | 31 | 17 | 6 | **19** | 19 | **9** | 7 | **22** | 22 | **10** |
| Alignment-based selection | | 28 | 93 | 85 | 44 | 69 | 231 | 223 | 11 | 48 | 193 | 181 | 90 |
| Alignment-based selection | ✓ | 16 | 69 | 69 | 32 | 24 | 85 | 84 | 71 | 29 | 115 | 114 | 53 |
| Greedy selection | | 11 | 34 | 27 | 18 | 26 | 85 | 84 | 41 | 16 | 53 | 52 | 26 |
| Greedy selection | ✓ | 11 | 33 | 29 | 16 | 26 | 78 | 75 | 41 | 16 | 50 | 48 | 25 |
| Greedy selection (F-score) | | 11 | 34 | 31 | 16 | 13 | 42 | 42 | 20 | 8 | 27 | 27 | 12 |
| Greedy selection (F-score) | ✓ | 11 | 31 | 28 | **15** | 13 | 42 | 42 | 20 | 8 | 27 | 27 | 12 |
| CloGSgrow merging | | 22 | 97 | 97 | 527 | 12 | 39 | 37 | 31 | 9 | 29 | 30 | 26 |

LPM post-processing approaches. However, the models that are discovered with process discovery approaches often are more complex in terms of extended Cardoso measure and extended cyclomatic complexity, indicating that they might in fact be more difficult for the analyst to understand than the set of LPM patterns. The Split Miner generates a model with *improper completion* for many of the event logs, i.e., it allows for runs through the model that do not end with one token in the final marking. This is widely considered to be an undesirable property of a process model, and it disallows the calculation of the extended cyclomatic complexity (resulting in "-" values in the CY column of Table 12.8).

Note that we did not calculate complexity measures other than the number of LPMs and the number of transitions for CloGSgrow patterns. Because the patterns obtained with CloGSgrow are strictly sequential, it would not yield additional insights to calculate complexity measures for process models for them, and it suffices to compare the number of patterns and transitions. Compared to LPM mining, CloGSgrow generates a smaller number of patterns on almost all datasets. This is expected, as the non-sequential constructs that are allowed in LPMs allow for a wider variation of patterns. Two exceptions, however, are the *Cook hh104 labour* and the *Cook hh104 weekend datasets*, on which CloGSgrow generated more patterns than LPM mining even though the same minimal support was used for mining. The reason for this is that for computational reasons we used an approximate heuristic LPM mining approach for datasets with 15 or more activities, resulting in the situation that some LPM patterns that meet the minimum support are not found.

### *Computation Time*

In addition to the computational complexity of the approaches that we reported in Section 12.2, we now discuss experimental findings of the runtime of the approaches on the eleven datasets. Table 12.9 shows the running time (in seconds) of each approach on each of the 11 event logs, on an Intel i7 CPU @ 2.4GHz with 16 GB of memory.

The traditional process discovery approaches (the Inductive Miner and the Split Miner) are very fast and can generate a process model from the event log in less than a second. For techniques that post-process LPM mining results (i.e., heuristic selection, alignment-based selection, greedy selection, and greedy F-score based selection) the reported computation times include the time for mining the LPMs. LPM mining is slow for the traffic fine event log because this event log has many sequences and 11 activities, which is just below the threshold (14 activities) for switching to the heuristic approximate LPM mining approach. Alignment-based selection of LPMs is often slower than LPM mining itself. Heuristic selection is a fast procedure, adding at most a second to the computation time needed for mining the LPMs.

For the CloGSgrow merging approach the computation times include only the merging procedure and not the time required to mine the CloGSgrow patterns themselves. This is because we relied on our own implementation of the CloGS-grow algorithm and want to prevent implementation-specifics from influencing the results.

**Table 12.9:** The runtime (in seconds) of the LPM list mining methods per event log.

| | | event log (IDs as shown in Table 12.4) | | | | | | | | | | |
|---|---|---|---|---|---|---|---|---|---|---|---|---|
| Method | Remining | 1 | 2 | 3 | 4 | 5 | 6 | 7 | 8 | 9 | 10 | 11 |
| *Baseline techniques* | | | | | | | | | | | | |
| LPM mining | | *790* | *8* | 26824 | *142* | 123 | 700 | 8 | *18* | *11* | *589* | *189* |
| Heuristic selection | | 790 | 8 | 26825 | 142 | 124 | 701 | 8 | 19 | 12 | 589 | 190 |
| Inductive Miner | | 2 | 1 | 5 | 1 | 1 | 1 | 1 | 1 | 1 | 1 | 1 |
| Inductive Miner (20%) | | 2 | 1 | 4 | 1 | 1 | 1 | 1 | 1 | 1 | 1 | 1 |
| Inductive Miner (50%) | | 1 | 1 | 3 | 1 | 1 | 1 | 1 | 1 | 1 | 1 | 1 |
| Split Miner | | 1 | 1 | 1 | 1 | 1 | 1 | 1 | 1 | 1 | 1 | 1 |
| *Novel approaches* | | | | | | | | | | | | |
| Heuristic selection | ✓ | 791 | 9 | 26825 | 144 | 124 | 701 | 9 | 19 | 12 | 590 | 191 |
| Alignment-based selection | | 7202 | 323 | 53149 | 5414 | 1110 | 1071 | 19 | 476 | 2058 | 8994 | 6421 |
| Alignment-based selection | ✓ | 7203 | 325 | 53151 | 5421 | 1114 | 1073 | 20 | 478 | 2060 | 9001 | 6433 |
| Greedy selection | | 1542 | 841 | 28641 | 5570 | 201 | 852 | 532 | 130 | 86 | 774 | 316 |
| Greedy selection | ✓ | 1544 | 843 | 28643 | 5574 | 202 | 854 | 534 | 132 | 88 | 779 | 318 |
| Greedy selection (F-score) | | 8012 | 764 | 28757 | 6311 | 352 | 1103 | 308 | 331 | 199 | 1590 | 882 |
| Greedy selection (F-score) | ✓ | 8014 | 765 | 28760 | 6315 | 356 | 1105 | 310 | 333 | 201 | 1593 | 886 |
| CloGSgrow merging | | 1412 | 1632 | 17197 | 4983 | 1358 | 603 | 325 | 275 | 206 | 689 | 340 |

### 12.3.3 Case Study

In this section, we present and discuss the set of Local Process Models that we obtained with the greedy F-score based selection approach on the BPI'12 dataset, i.e., dataset 1 from Table 12.4. This event log originates from a financial loan application process at a large Dutch financial institution and every sequence represents the business activities that are performed for a single loan application.

The names of the activities in this event log all start with *A_*, *O_*, or *W_*. Activities starting with *A_* refer to the financial loan application themselves, i.e., they correspond to status changes of the financial loan application. Examples are *A_SUBMITTED*, which indicates that the application has been submitted, *A_PREACCEPTED*, which indicates that there is a provisional decision to accept the application but additional information from the applicant is required, and *A_ACTIVATED*, which indicates that an accepted financial loan has gone into effect and payments are being processed. Activities that start with *O_* correspond to offers that are communicated to the customer by bank employees. Examples include *O_SENT*, which indicates that a offer that was prepared by a bank employee has been sent to the customer, *O_CANCELLED*, which means that the bank employee cancelled an offer, and *O_DECLINED*, which indicates that the customer has refused an offer. Finally, activities that start with *W_* correspond to work items, i.e., tasks that are assigned to bank employees for the approval process. Examples include *W_ Afhandelen leads* (in English: "Processing leads"), which indicates that a bank employee follows up on an incomplete submission of a loan application by a potential customer, *W_Nabellen offertes* (in English: "Calling after offers") which indicates a call from a bank employee after transmitting an offer to a customer, and *E_Nabellen incomplete dossiers* (in English: "Calling after incomplete files"), which indicates a call from a bank employee to request additional information from the potential customer that is needed to asses the application.

Figure 12.6 shows the obtained LPM list, which consists of six patterns. The LPM list has a coverage of 0.447, indicating that almost half of the events in the event log are described by one of the six patterns. The precision of the LPM list is 0.359, resulting in an F-score of 0.3983.

The first LPM (in Figure 12.6a) shows that *O_SELECTED* and *A_FINALIZED* are executed concurrently and are ultimately followed by *O_CREATED* and *O_SENT*. This pattern occurs 5015 times in the event log, which corresponds to occurrences of *O_SELECTED* adhering to this behavior. For the other activities, *A_FINALIZED*, *O_CREATED*, and *O_SENT* a large majority of 5015 out of the 7030 events of each of these activities are described by this behavior.

Figure 12.6b shows that *O_ACCEPTED* is generally followed by both *A_REGISTERED* and *A_ACTIVATED*, but the order in which these two activities occur is not consistent. Finally, this behavior is followed by one or more occurrences of activity *W_Valideren aanvraag* (in English: "Validate request").

Figure 12.6c shows a pattern where *O_CANCELLED* is generally followed by multiple occurrences of *W_Nabellen offertes* (in English: "Calling after offers"). Looking at the numbers (2592 *O_CANCELLED* and 8127 *W_Nabellen offertes* events), we can conclude that on average the call center employees can the customer three times after an offer has been canceled.

Figure 12.6d shows that *W_Afhandelen leads* (in English: "Processing leads"), are in 4749 out of 5898 cases preceded either by an *A_DECLINED* (2234 times) or by *A_PREACCEPTED* (2515 times).

Figure 12.6e shows that *A_APPROVED* is sometimes (606 out if 2246 times) followed by *A_ACTIVATED* (316 times) or by *A_REGISTERED* (290 times).

Finally, Figure 12.6f shows that all occurrences of *A_SUBMITTED* are followed by *A_PARTLYSUBMITTED*, and that, likewise, all occurrences of *A_PARTLYSUBMITTED* are preceded by *A_SUBMITTED*.

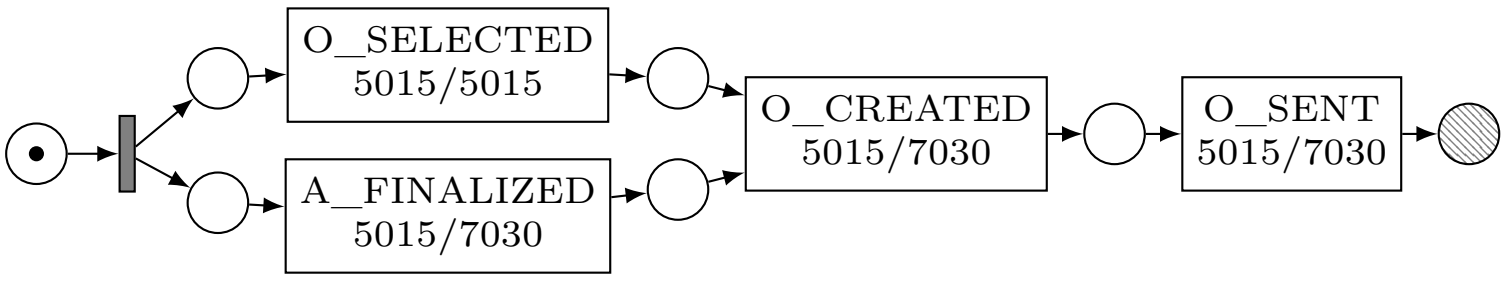

**(a)** support: 5015

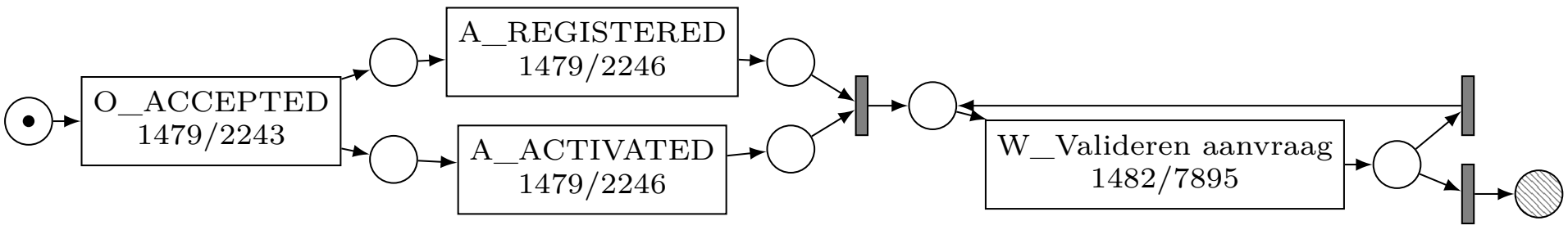

**(b)** support: 1479

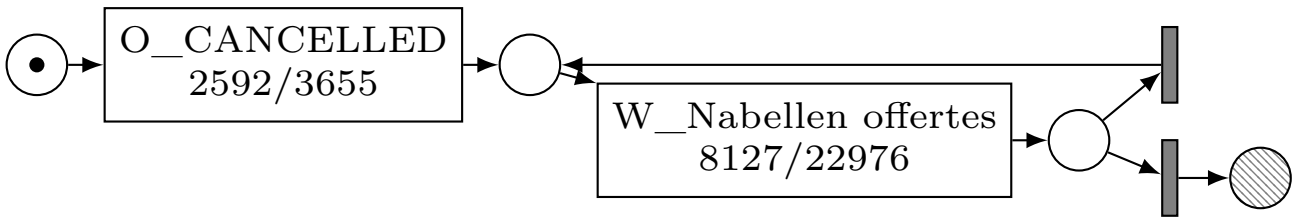

**(c)** support: 2592

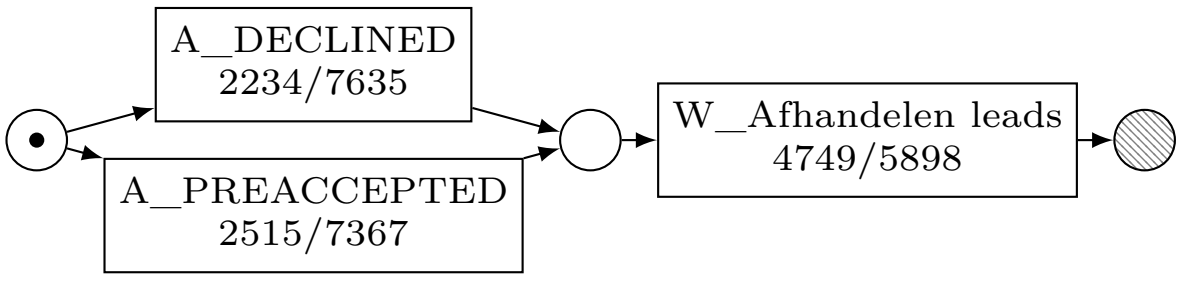

**(d)** support: 4749

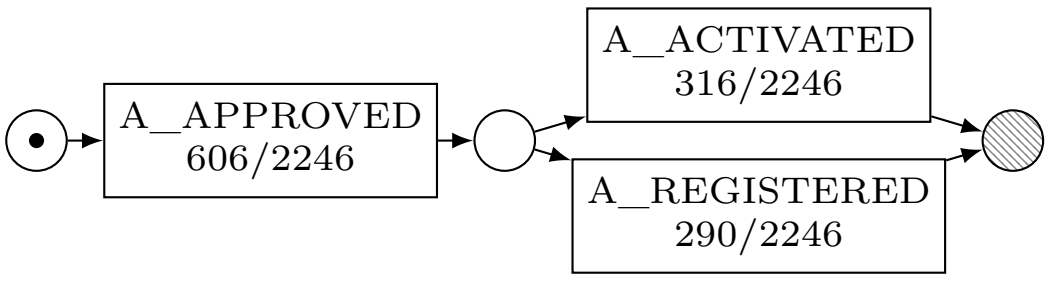

**(e)** support: 606

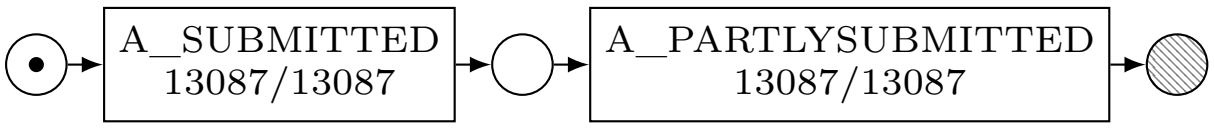

**(f)** support: 13087

**Figure 12.6:** The LPM list that is mined with the greedy F-score based selection approach from the BPI'12 dataset.

## 12.4 Related Work

Pesole et al. [Pes+92] developed the WORDUP algorithm back in 1992 to efficiently mine sequential patterns from DNA sequences that occur more frequently than what would be expected by chance. By mining such statistically significant patterns, a shorter list of patterns is obtained than what would have been obtained when mining patterns that simply meet a support threshold, as many frequent patterns that simply consist of frequent symbols might not be significant. The WORDUP algorithm as a first step extracts a first-order Markov model from the event log and uses this Markov model to assess the statistical significance of subsequences. Note that this approach is limited to strictly sequential patterns. DNA data typically consists of sequences of very small alphabet size. Later work on statistically significant sequential pattern mining by Low-Kam et al. [Low+13] improves the computational complexity for event logs with a large alphabet size.

Xin et al. [Xin+05] can be seen as early work in the area of compressed pattern mining, who applied ideas from the area of compression to the pattern mining task of frequent itemset mining. An important finding of this work is that mining compressed patterns is an NP-hard problem, regardless of the type of patterns that are being compressed. This motivated later work to look more into heuristic approaches. Yan et al. [Yan+05] proposed an approach based on generative modeling to reduce a set of itemsets into a set of k representative itemsets. This approach fits a Bernoulli distribution over the itemsets using maximum likelihood estimation and uses that probabilistic model to estimate the support of unseen itemsets that might be representative of multiple itemsets from the set of frequent itemsets even though it is not part of the set of frequent itemsets itself.

Many of the later techniques in compressed pattern mining are rooted in the minimum description length (MDL) principle. In the MDL approach, the pattern set is considered to be a compression dictionary that is used to replace each instance of a pattern by a novel symbol that represents the pattern. The description length is then defined as the number of bits of information that are needed to both store the compression dictionary and the compressed event log. Tatti and Vreeken extensively studied the problem of mining compressed patterns [Hei+09; MTV11; MVT12; TV08; TV12; VLS11]. Mampaey, Tatti, and Vreeken [MTV11; MVT12] proposed to use a class of probabilistic models called maximum-entropy models to select a compressed set of itemsets from the set of frequent itemsets. In [TV08] Tatti and Vreeken propose the PACK algorithm which uses decision trees combined to determine the shortest possible encoding and to select itemsets from a set of frequent itemsets. Around the same time, Heikinheimo et al. [Hei+09] proposed the LESS algorithm to greedily select low-entropy itemsets from the set of mined frequent itemsets in order to obtain a set of compressed itemsets. The KRIMP algorithm [VLS11] combines the main ideas of LESS and PACK into a single compressed itemset mining algorithm.

The KRIMP algorithm has later been adapted in order to be applied to select a set of compressed sequential patterns from a set of frequent sequential patterns, called the GoKRIMP [Lam+14]. An alternative algorithm to select a set of compressed sequential patterns from a set of frequent sequential patterns is SQS-Search [TV12]. The main difference between GoKRIMP and SQS-Search is that GoKRIMP allows the instances of the sequential patterns to be interleaved, while SQS-Search, like the approaches proposed in this chapter, does not. Note that by constructing the global model from the LPM patterns slightly differently, by putting all the individual LPMs in a parallel relation instead of in choice relation, we could obtain an LPM list evaluation approach that is similar to GoKRIMP rather than SQS-Search. The computational cost of aligning LPMs to a global model that is constructed in that way, however, prohibits the use of this approach in practice.

## 12.5 Conclusions

The foremost contribution of this chapter is a set of techniques to mine a non-redundant set of generalizing patterns captured as *Local Process Models* (LPMs) from an event log. We have shown that the generalizing capabilities of LPMs, coupled with the non-redundancy property, allow us to concisely summarize the event log via a small number of patterns compared to alternative methods. We have also introduced a framework for evaluating the quality of sets of LPMs, which takes into account both the share of the events in the event log that are *covered* by an LPM list as well as *the degree of redundancy* in the LPM list.

We have outlined multiple approaches to the posed problem, which differ in computational complexity and in their trade-off between *how many events are covered by the pattern set* and *the degree of redundancy of the pattern set*. We have shown that alignment-based selection can reduce the redundancy of a set of LPMs without loss in coverage. The greedy selection and the greedy F-score-based selection approach are able to achieve further reduction of redundancy at the cost of coverage.

The proposed method for mining LPM patterns has many applications. A direct application that we envisage is to discover repetitive routines from fine-grained event (e.g. clickstreams), which may be amenable to automation. Other applications includes exploratory analysis of smart home data in order to extract daily routines, as well as analysis of Web usage logs in order to identify typical usage patterns which may give rise to opportunities for Web site optimization. Investigating these applications is an avenue for future work.

# 13 The LocalProcessModelDiscovery Tool

**Parts of this chapter have been published as:**

- Niek Tax, Natalia Sidorova, Wil M. P. van der Aalst, and Reinder Haakma: *LocalProcessModelDiscovery: Bringing Petri Nets to the Pattern Mining World*. Proceedings of the 39th International Conference on Application and Theory of Petri Nets and Concurrency (Petri nets), Lecture Notes in Computer Science (LNCS), 10877, Springer: 374–384. (2018)

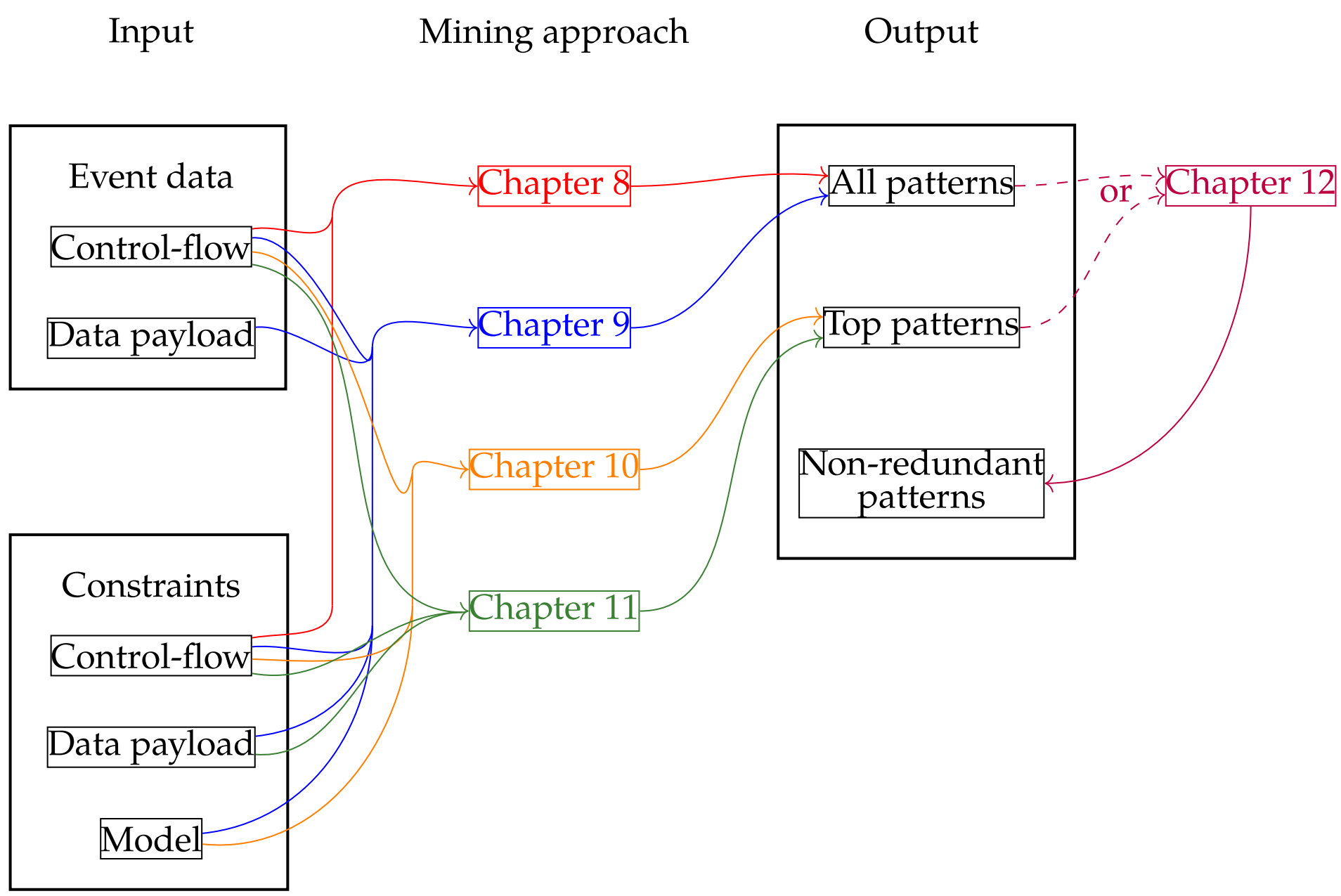

**Figure 13.1:** A taxonomy of local process model techniques.

13

After introducing Local Process Model (LPM) mining in the preceding chapters of Part II, we will use this final chapter to introduce the software tool that provides implementations of the techniques that have been introduced throughout Part II. This chapter is intended to read like a manual for the tool, by giving instruction on how to use it and how to set all the types of techniques and parameters that have been discussed in the previous chapters.

The tool is implemented as a package in the Java-based *process mining* framework ProM [Don+05] and is publicly available at `https://svn.win.tue.nl/repos/prom/Packages/LocalProcessModelDiscovery/` and in the ProM package manager. After installing the *LocalProcessModelDiscovery* package to ProM, the tool can be started by importing an event log in XES [IX16] format into ProM and then running the ProM plugin *Search for Local Process Models* using this event log as input. The algorithms for mining LPMs from event logs, i.e., frequent Petri net patterns form the core of the *LocalProcessModelDiscovery* tool.

### *Purpose of the Tool*

Mining of Local Process Models (LPMs) can be positioned in-between the research areas of *Petri net synthesis* and *process discovery* on the one hand and *frequent pattern mining* on the other hand. *Frequent pattern mining* [Han+07] techniques focus on extracting local patterns from data. *Sequential pattern mining* [Fou+17] techniques are a type of frequent pattern mining that focuses on the extraction of frequent patterns from sequence data. While process discovery [Aal16] and Petri net synthesis techniques aim to discover an *end-to-end process model*, sequential pattern mining techniques aim to extract a *set of patterns* where each pattern describes a subsequence that frequently occurs in the event log. Sequential pattern mining techniques can be used to generate insights from event data that only contain weak relations between the activities, i.e., that have a relatively high degree of randomness. From such event logs, process discovery and Petri net synthesis techniques generate either overgeneralizing models that allow for too much behavior, or generate a 'spaghetti'-model that is accurate in the allowed behavior but is not understandable and often overfitting.

Figure 13.2a gives an example of an overgeneralizing process model, showing the process model discovered with the Inductive Miner [LFA13b] from web-click data from the MSNBC.com news website[34] (note that most activities can be skipped and/or repeated allowing for any behavior). Figure 13.2b gives an example of a non-interpretable 'spaghetti'-like process model, discovered from the same dataset with the ILP Miner [Wer+09b]. The first LPM of Figure 13.2c, however, shows that activity 10 is generally followed by multiple instances of activity 2.

[34]http://kdd.ics.uci.edu/databases/msnbc/msnbc.data.html

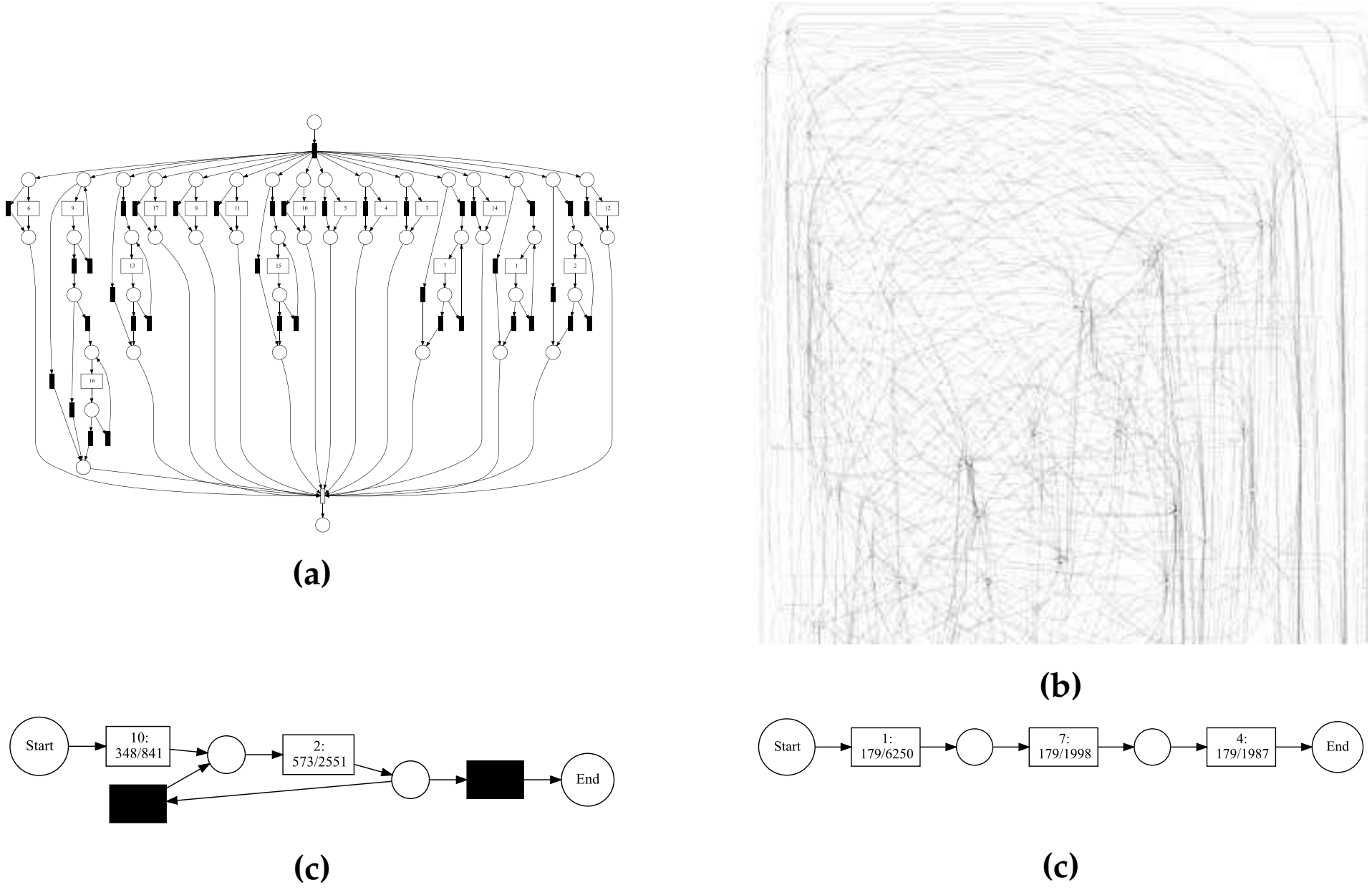

**Figure 13.2:** The Petri net model mined from the MSNBC dataset with *(a)* the Inductive Miner [LFA13b], *(b)* the ILP Miner [Wer+09b], and *(c)* two LPMs mined from the MSNBC dataset. Black transitions correspond to silent transitions.

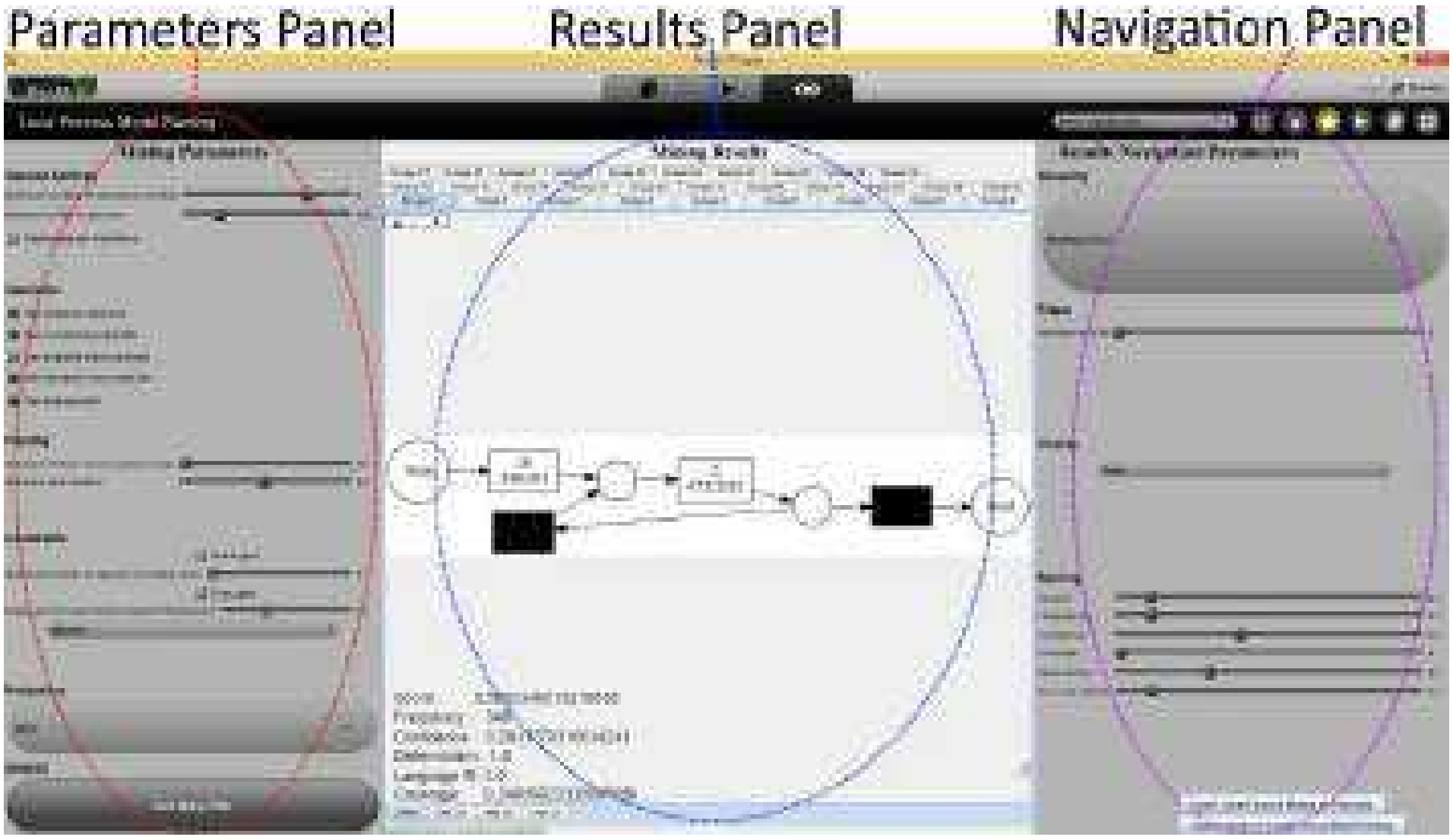

**Figure 13.3:** The home screen of the *LocalProcessModelDiscovery* tool.

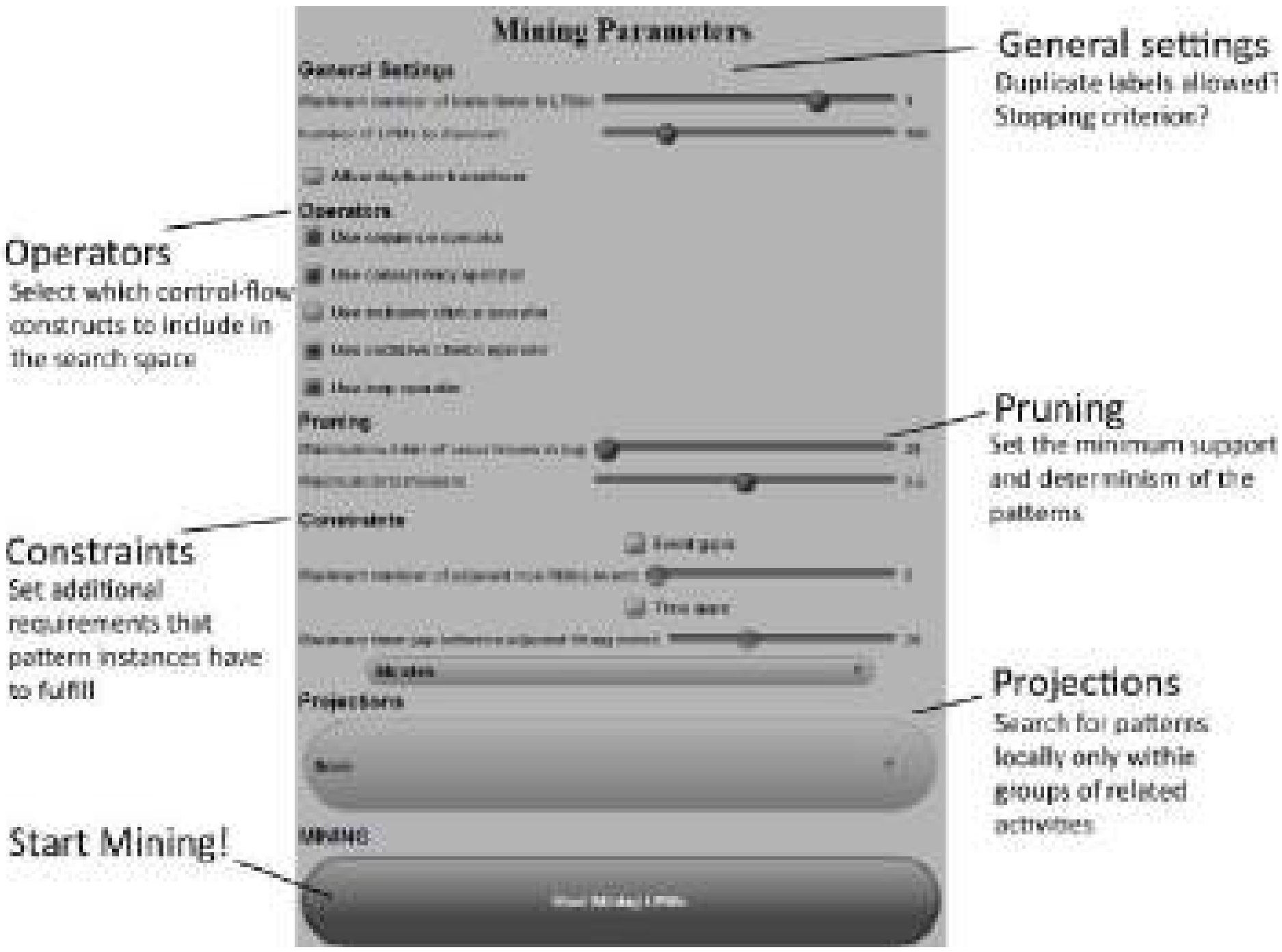

**Figure 13.4:** The parameters panel of the *LocalProcessModelDiscovery* tool.

# 13.1 The LocalProcessModelDiscovery tool

Figure 13.3 shows the main screen after launching the plugin *"Search for Local Process Models"* that is available in the ProM package *LocalProcessModelDiscovery*[35]. The main screen consists of three panels: on the left side a panel to configure the mining parameters, in the middle a panel that presents the mining results when mining has completed, and on the right side a panel to interact with and navigate through the mined local process models.

## 13.1.1 Configuring the Local Process Model Miner

Figure 13.4 shows the mining parameters panel.

Located at the top of the panel is a **maximum number of transitions in the LPMs** slider, which allows the user to set the maximum number of non-silent transitions for the local process models. The maximum number of non-silent transitions puts an upper bound on the number of expansion iterations in the local process model mining algorithm, therefore, setting this slider to higher values enables the tool to discover larger patterns at the cost of higher computation time. This slider implements the *max_iterations* as defined in Section 8.2.

The **number of LPMs to discover** slider lets the user specify a maximum number of patterns that he or she wants to obtain, allowing him or her to prevent mining an overload of patterns. Note that the local process model discovery algorithm returns a ranked list of patterns, therefore, the algorithm returns the patterns with the highest weighted score to the user. Note that the LPM mining algorithm in principle can find more LPMs that satisfy the support and determinism thresholds that were set by the user.

The **allow duplicate transitions** toggle allows the user to specify whether or not he or she wants to mine patterns that contain multiple transitions with the same label. Enabling this option comes at the price of higher computation cost.

The **operator** section of the parameter panel allows the user to include or exclude patterns with certain control-flow constructs in the search space of the mining procedure. Changing the selection of process tree operators has the effect that the definition of the expansion function $exp_2$ as we had defined in Section 8.2.4 by including or excluding expansions with certain process tree operators from the search space. Selecting fewer process tree operators restricts the types of LPMs that can be found during mining (to those that contain the selected operators), but also reduces the computation time.

In the **pruning** section of the parameter panel the user can specify a **minimum number of occurrences** of the pattern in the log (i.e., minimum support), and a minimum **determinism** value for the pattern. Patterns that do not meet these thresholds set by the user are not presented in the results and are not expanded into larger patterns, thereby pruning the search space of the local process model mining algorithm. We refer back to Section 8.3.1 for a more in-depth discussion on pruning.

[35]https://svn.win.tue.nl/trac/prom/browser/Packages/LocalProcessModelDiscovery/

The parameters section allows the user to specify **event gap constraints** and **time gap constraints** as we have introduced in Chapter 11. An event gap constraint puts a maximum on the number of non-fitting events in-between two fitting events of an instance of an LPM in the log. For example, a sequence $\langle a, b, x, x, c\rangle$ is considered to be an instance of a Petri net M with language $\mathfrak{L}(M) = \{\langle a, b, c\rangle\}$ when the event gap constraint is set to 2, but is not considered to be an instance when the event gap constraint is set to 1. For event logs containing timed events, time gap constraints can be used to specify an upper bound on the time difference between two consecutive events that fit the behavior of an LPM.

The computational complexity of mining LPMs is exponential in the number of activities in the event log. Several heuristic mining techniques have been proposed in Chapter 10 to make the mining of LPMs on event logs with large numbers of events feasible. These heuristic techniques work by detecting clusters of activities that frequently occur close to each other in the event log, and then mining the LPMs for each cluster individually, restricting the expansion steps of the LPM mining to activities that are part of the same cluster of activities. The **projections** section of the mining parameters panel allows the user to configure the tool to use these approaches. The dropdown selector for the projection allows can be set to *none* to mine LPMs without using projections sets, i.e., by exploring the full search space of $exp_2$ as introduced in Chapter 8. Alternatively, the dropdown selector can be set to *Markov-based*, *Entropy-based* or to *Maximal Relative Information Gain (MRIG)*, all of which we had introduced in Chapter 10.

When starting the *LocalProcessModelDiscovery* tool, it automatically sets the minimum support parameter and the projections configuration to a default value that is dependent on the event log, using information such as the number of events and the number of activities in the event log.

### 13.1.2 Interpreting Local Process Model Results

Figure 13.5a shows a grouping of local process models in the mining results panel. The figure shows a single local process model mined from a hospital event log that describes the behavior that a *lab test* and an *X-ray* are performed in arbitrary order, finally followed by an *echography*. Printed below the Petri net are the scores of the pattern in terms of the local process model quality criteria. The local process models are ranked by the aggregated score of the LPM (shown as **score**), and the tabs in the bottom of the panel can be used to navigate through the LPMs in the ranking. Typically, the mining procedure results in multiple local process models that specify behavior over the same alphabet of activities. By default, the resulting local process models are grouped by the alphabet of activities that they describe. The **group** tabs above the Petri net can be used to explore the LPMs that describe

different alphabets of activities.

### 13.1.3 Navigating Local Process Model Results

The navigation panel provides several functionalities to interactively navigate through the resulting local process models obtained through mining. A weighted average over the quality criteria of local process models is used to rank the resulting local process models, and the results are presented in the order of the ranking. The user can reconfigure the weights assigned to each of the quality criteria in the **ranking** section of the navigation panel, resulting in an updated ranking of local process models in the **results panel**.

The **overlay** functionality in the navigation panel allows the user to project data attributes of the events in the log onto the local process models. The overlay functionality consists of a drop-down selector where the user can select one of the global event attributes (i.e., an event attribute that is set for every event in the log). Figure 13.5b illustrates the overlay functionality and shows the mining results panel when selecting the *org:resource* event attribute for the local process model of Figure 13.5. The results show that the *X-ray* and *Echography* events that fit the behavior of this local process model are most frequently performed by employee *Alex*, while the *Lab Test* events that fit the behavior of this pattern are most frequently performed by employee *Jo*. Note that this does not say anything about the *X-ray* events that do not fit the behavior of this pattern, i.e., the X-ray events that are not performed concurrently to the *Lab Test* and before *Echography*.

The **filters** section of the results panel allows the user to filter out local process models from the results that do not comply with certain specifications that are provided by the user, such as a minimum number of **activities in the log**. In the **grouping** section, the user can select a strategy for grouping mined local process models into groups of local process models for the visualization in the results panel. By default, the **ranking-based** grouping strategy is used, which adds one local process model A to the same group as another local process model B if 1) the set of activities of A is a subset of the set of or equal to the activities of B and 2) A has a lower aggregated score than B.

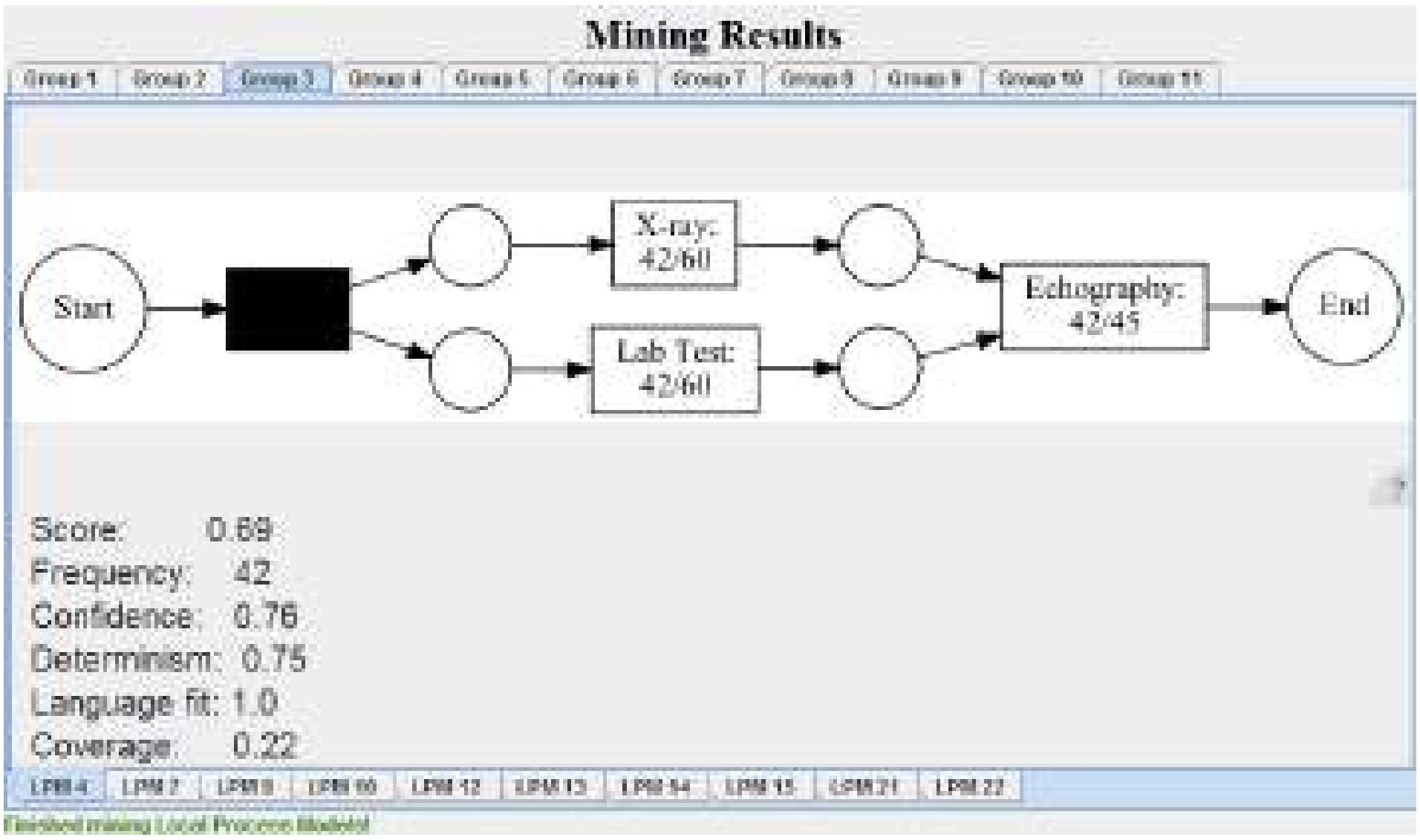

**(a)**

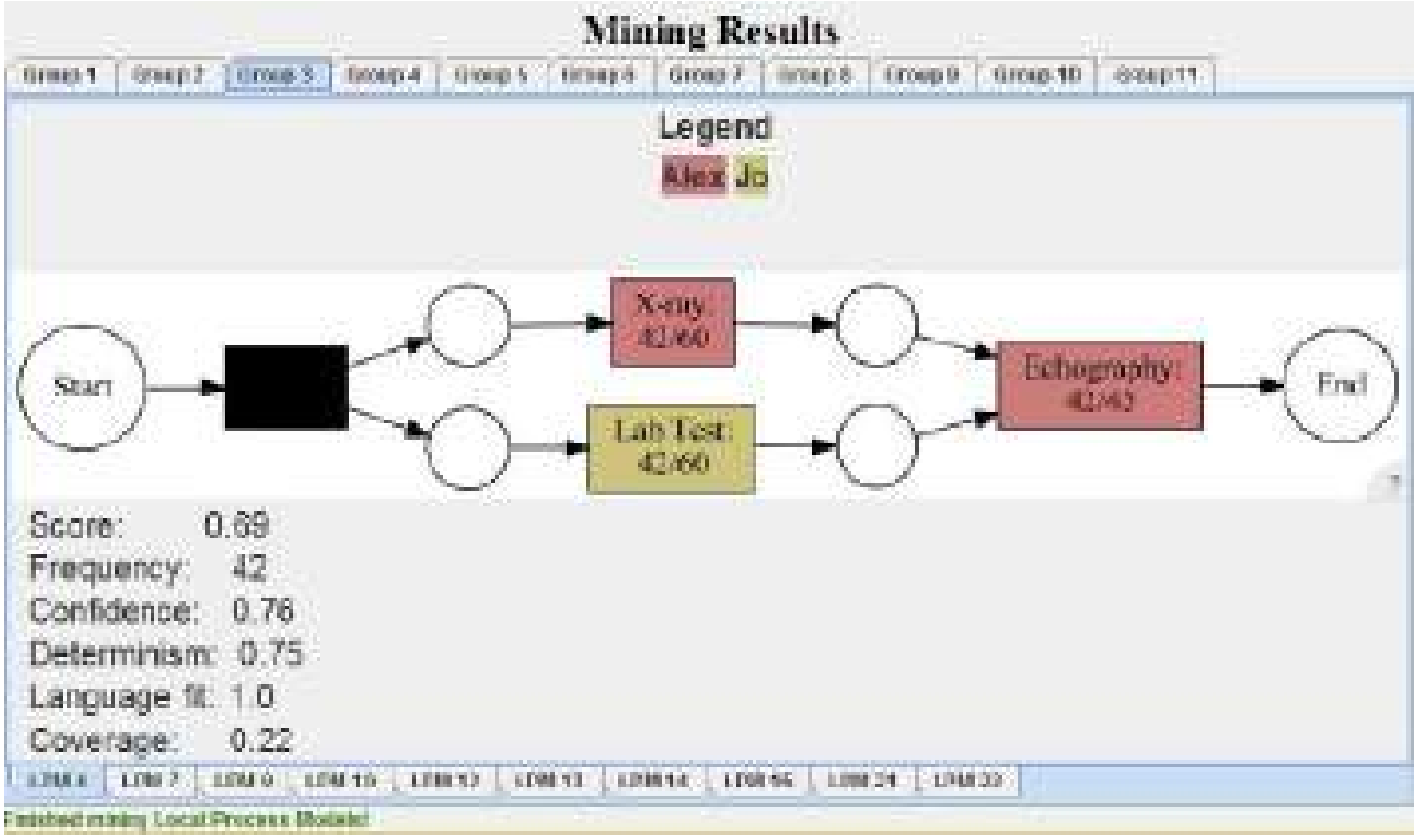

**(b)**

**Figure 13.5:** *(a)* An example LPM in the results panel of the *LocalProcessModelDiscovery* tool, and *(b)* an example of the *overlay* feature in the *navigation* panel, projecting the *resource* information on top of the pattern.

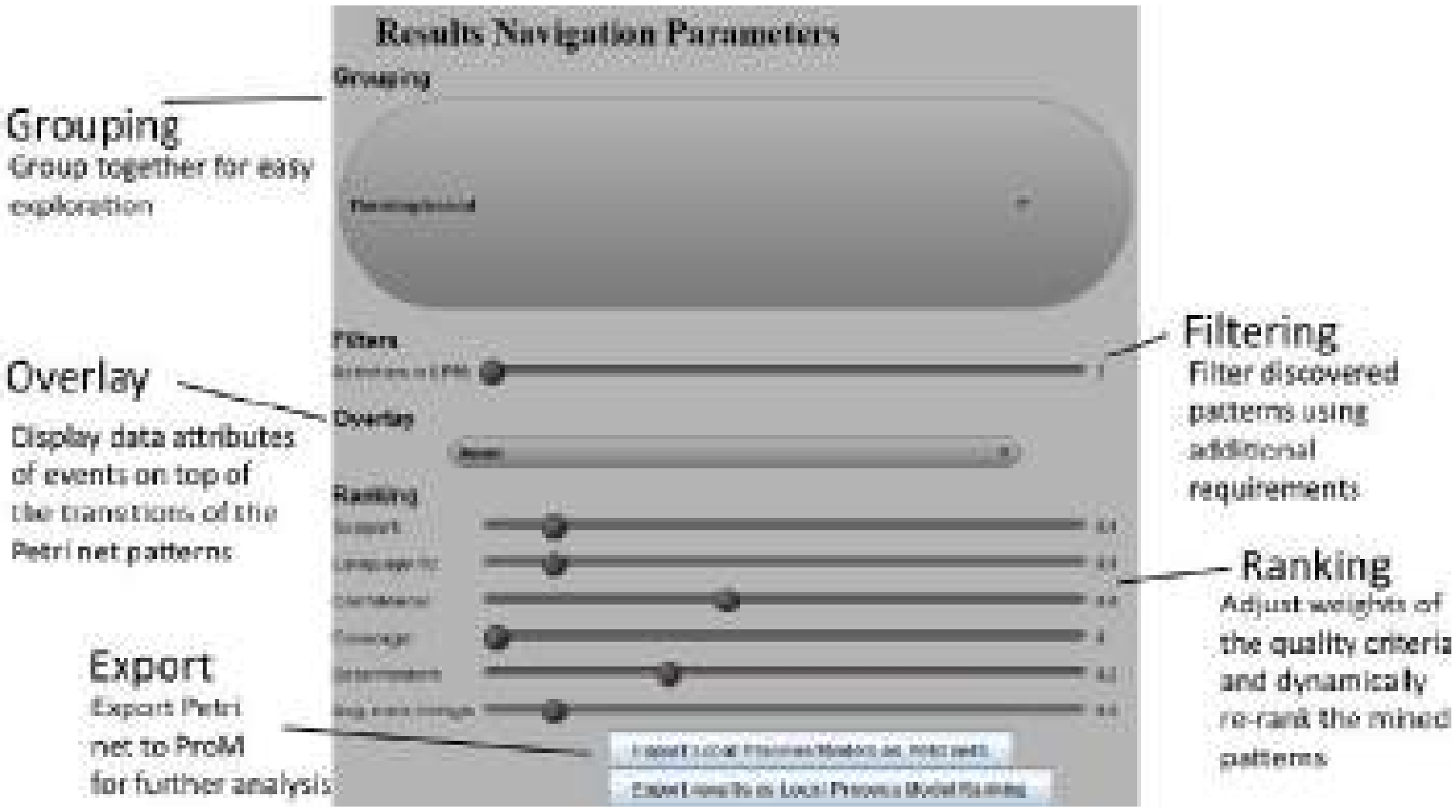

**Figure 13.6:** The navigation panel of the *LocalProcessModelDiscovery* tool, which allows the user to interact with the mining results and perform a more in-depth follow-up analysis.

## 13.2 Conclusion

This chapter presented the tool *LocalProcessModelDiscovery*, which provides implementations of the techniques that have been presented in Part II of this thesis. Local process models are positioned in-between process discovery and Petri net synthesis on the one hand, and frequent pattern mining on the other hand. The local process models that can be mined with this tool extend existing sequential pattern mining approaches: while sequential patterns are restricted to mining frequent sequential behavior, local process models allow the frequent patterns to describe a more general language over the activities by expressing the patterns as Petri nets. *LocalProcessModelDiscovery* supports the mining of local process models as well as functionality to navigate through the mining results and to relate discovered local process models back to the event log for a more in-depth analysis.

# Part III

# Closure

# 14 Conclusion

In this thesis, we have presented techniques and tools for the analysis of event data that originates from application domains where the behavior that was recorded was not constrained by some system. This contrasts the traditional application domain of process mining techniques: Business Process Management (BPM), where IT systems generally constrain what behavior is possible at which point in the process. When there is no system that constrains the allowed behavior in the process, the event log as a result tends to have higher variability. Throughout this thesis, we have used the analysis of human behavior in smart homes as prime example of an application domain with such high-variability event data.

At the core of this thesis is 1) a set of pre-processing techniques to enable the mining of insights from event data that is highly variable, and 2) a novel pattern mining approach called Local Process Models (LPMs) that combines ideas from the pattern mining field with behavior constructs from the process mining field.

The remainder of this concluding chapter of the thesis consists of four sections. First, in Chapter 14, we summarize the contributions of this dissertation with regard to the two parts in which we have structured this thesis. We continue with Section 14.2, in which we discuss limitations of the techniques that we have proposed in this thesis and use this to start a discussion on open issues regarding the three research goals that we have listed in the introduction of this thesis. In Section 14.3 we discuss several promising directions for future work. Finally, Section 14.4 concludes this thesis by reflecting on the broader context and outlook on process mining on event data that is highly variable.

## 14.1 Contributions

We structure the listing of contributions per part.

### 14.1.1 Part 1: Pre-processing of Event Data

**Dealing with unlabeled event logs and logs with labels of improper granularity**
We have developed a statistical approach to evaluate whether a given proposal for a refinement of a given event label into some more fine-grained event labels contributes positively to the degree of structure in the event log (described in Chapter 4). In Chapter 5 we propose an automated way to generate proposals to refine event labels in a way that is domain-specific to timestamp attributes. By combining this technique with the statistical test of Chapter 4 we are able to automatically refine event labels based on timestamp attributes. We have shown that this approach enables us to automatically refine event labels of a smart home event log into more specific event labels that partly consist of timestamp information and we have shown that the refined versions of an event label indeed are behaviorally different. To summarize: in Chapter 4 and Chapter 5 we respectively proposed techniques to assess whether an activity / event label in an event log is of the proper level of granularity and propose a method to refine an event label into more refined event labels by bringing in information from some event attribute into the event label. When there initially is no event label available at all, we propose to address this issue by choosing one of the event attributes as the starting event label and then iteratively applying the label refinement techniques of Chapter 4 and Chapter 5.

**Mining relations between time and control-flow** The label refinement technique of Chapter 5 brings elements from the timestamp data into the event label and thereby enables the discovery of relations between time and control-flow.

**Dealing with event logs that are unstructured due to chaotic activities** We have shown the devastating effect of what we call *chaotic activities* on the successful application of process discovery in Chapter 6. We have proposed a technique based on information entropy to filter such chaotic activities from event logs and we have shown that this has a large effect on the fitness and precision of the discovered process models.

**Dealing with event logs with events that are too fine-grained** We have shown that the degree of variability of an event log can depend on the level of granularity of events. Abstracting events to higher-level events can unveil structures in the event log that were not observable from the original low-level events. In Chapter 7 and Section 8.7 we have proposed techniques to automatically abstract event to higher-level events. The technique of Chapter 7 takes a supervised approach, i.e., it requires both low-level event data as well as some event data that is on the desired level of abstraction and from these examples it learns a mapping from low-level to high-level events that can be used to abstract some other unlabeled events for which no high-level interpretation is available. In contrast, in Section 8.7 we have proposed the use of LPMs for event abstraction, thereby yielding an unsupervised event abstraction approach.

### 14.1.2 Part 2: Local Process Models

**Mining frequent patterns in the form of process models** One of the main contributions of this thesis is a novel type of pattern mining, called *local process* *model* (LPM) mining, that focuses on mining small fragments of frequently occurring process behavior from event logs. These LPM patterns are expressed as process models and are thereby able to express rich ordering constructs like sequential ordering, (exclusive) choice, concurrency, and loops, which are commonly used in the business process modeling community. LPM mining is particularly well-suited to mine insights from highly variable event logs due to that fact that each pattern individually is only required to capture a single piece of frequent behavior. This contrasts existing process discovery approaches that mine a single process model from an event log, which can be either overgeneralizing or too complex in the case that the event log under study is highly variable. We introduced an initial base method for LPM mining in Chapter 8 and we have proposed a set of faster heuristic techniques for LPM mining in Chapter 10. Additionally, we have proposed an extension to the LPM mining framework in Chapter 9 that allows us to define the utility of LPM patterns in terms of data attributes and to mine patterns that have high utility.

**Mining LPM patterns that occurr within a pre-specified time interval** In Chapter 11 we propose a type of LPMs that can be used to discover relations between control-flow perspective and the time aspect. These constraint-based LPMs work by formulating a maximum time gap that is allowed between two events in order to still be considered part of the same pattern instance. Thereby, it gives insight into how frequently certain control-flow patterns occur in the event log given a maximum time constraint.

**Summarizing event data with a collection of LPM patterns** We have proposed a set of techniques in Chapter 12 to find a set of LPM patterns that collectively summarizes the behavior that was observed in an event log in a concise set of patterns.

## 14.2 Limitations and Open Issues

In this section, we discuss open issues and limitations of the approaches that we have described in this thesis. We start with limitations and open issues regarding the pre-processing techniques and then proceed with open issues and limitations regarding local process models.

### 14.2.1 Pre-processing

We identify the following limitations of the pre-processing techniques that we have proposed in this thesis.

*Label refinements*

**A Limited view on control-flow in the control-flow test for label refinements** One of the limitations of the control-flow test that we have proposed in Chapter 4 is the fact that it measures the impact of the label refinement on control-flow only through the directly follows/precedes and eventually follows/precedes log statistics. Even though the approach can in essence easily be extended to other log statistics, even doing so would leave the approach dependent on a pre-defined set of log statistics. Alternatively, one could think of approaches where one would specify a measure for structuredness of the behavior in the log and then measure the impact of a label refinement by assessing whether or not the structuredness of the log sufficiently increases as an effect of applying the label refinement. The event log entropy measure that we introduced in Chapter 6 could be a candidate for such a structuredness measure.

**Label refinements are limited to time-based refinements** The label refinement generation strategies that we have explored in Chapter 5 are limited to refinements based on the timestamp attribute of the events. Refining labels based on other types of event attributes thereby remains an unexplored direction. In principle, the mixture modeling approach that we applied for the circular continuous data attribute of time in Chapter 5 would likely also work for other, non-circular, continuous attributes by simply switching from a von Mises mixture model to a Gaussian mixture model, but this remains to be shown through experiments.

*Supervised event abstraction*

**Dependence on the availability of training data** The supervised approach that we have introduced in Chapter 7 requires training data to learn the mapping from low-level to high-level events. In practice, this in a smart home setting means that it is necessary to collect diary information with annotations of what the inhabitant of the smart home system did during the day. Collecting this information can be a costly and tedious task. A promising direction of research that in the future might mitigate this problem is the task of activity recognition from video data using cameras. Recognizing human activities from cameras is, in contrast to recognizing human activities from motion, pressure, and open/close sensors, not a household-specific problem, and instead one could train a model once and then deploy the model in multiple homes with cameras. However, this solution would be even more obtrusive and privacy-invasive than the sensors that are used in smart homes.

### 14.2.2 Local Process Models

We identify the following limitations of the local process model mining techniques that we have proposed in this thesis.

**Search space size of local process model expansions** In Chapter 8 we had formulated desired properties for the expansion function *exp* to expand an LPM into larger patterns. Informally speaking, these properties specified that we would ideally like to have an expansion function such that starting from the set of initial LPMs we can reach all possible process tree languages (Requirement 1) without doing duplicate work by generating equivalent languages more than once (Requirement 5). While we have proposed several improvements over some naive expansion function $exp_0$, the currently best expansion function ($exp_2$) satisfies Requirement 1 but still generates some equivalent languages and therefore does not satisfy Requirement 5. An open challenge is to develop more efficient expansion functions that are able to satisfy Requirement 1 while preventing language equivalent process trees, or at least reducing the generation of language-equivalent process trees with respect to $exp_2$.

**Computational complexity of identifying local process model instances** The current procedure to count instances of LPMs in an event log is based on alignments, which is known to be of exponential time complexity in the length of the trace as well as in the number of activities. An open challenge is to develop faster approaches to count/identify the instances of an LPM in an event log, either by developing techniques that are not alignment-based or by improving the alignment algorithm itself.

**Lack of clear conditions on when to use Local Process Models** We have seen many examples in several chapters of this thesis where traditional process discovery techniques fail to generate insights into the behavior that was registered in an event log, while some patterns could be successfully identified using LPMs. However, this thesis does not address the question of when exactly LPMs should be used instead of traditional process discovery techniques. An open challenge is to find clear criteria that a process analyst can use in order to decide when to use traditional process discovery techniques and when to use LPMs.

**Representational bias is limited to process trees** In this thesis, we have only explored LPMs that are expressed as process trees. Each type of process model notation comes with certain types of process behavior that can and some that cannot be described in that notation, which is commonly referred to as the *representational bias* of the process model notation. By choosing process trees as the underlying representation in the mining procedure of LPMs we rule out the possibility of some types of behavioral constructs that can just inherently not be expressed in the form of a process tree. Two examples of behavior that cannot be expressed by process trees are *long-term dependencies* and the *milestone pattern* (for an extensive overview of types of behavioral patterns we refer to [Aal+03]). An interesting direction for follow-up work would be to develop techniques for mining LPMs in a fundamentally different way, by using a different representation than process trees, thereby enabling LPMs that express other types of behavior that can currently not be discovered.

## 14.3 Future Work

In this section, we discuss ideas and challenges that emerged while writing this thesis.

### 14.3.1 A Unified Framework for Event Log Pre-processing

In this thesis we have discussed four types of pre-processing techniques: event abstraction, event refinement, label abstraction, and label refinement. The majority of existing techniques in the area of pre-processing for process mining fits one of those four categories in this framework. We have shown in Chapter 6 that all filtering techniques for process mining (event filtering, trace filtering, and activity filtering) fall into the category of event abstraction. An important drawback of current pre-processing techniques in process mining is the fact that it is currently left as a trial-and-error exercise to the process analyst: the process analyst is expected to iteratively try out a certain pre-processing method, then see if it gives satisfying results (e.g., does it enable the discovery of a "good" process model?), and if not, revert back and try a different pre-processing approach.

A more unified approach to event log pre-processing would be to start from formulating the success criteria by mathematically defining what a successfully pre-processed event log looks like. A successfully filtered event log conceptually fulfills two criteria:

**Structuredness** The resulting event log obtained through pre-processing is more structured than the input event log (which might have been weakly-structured), thereby enabling process mining techniques to unveil those structures in the event log. One can think of many possible measures to quantify the *structuredness* of an event log. Possible candidates are the entropy of the directly-follows and the directly-precedes relations of the activities (as defined in Chapter 6) or the perplexity of a Markov model that was fitted to the event log (as described in Chapter 12).

**Similarity** The resulting event log should be highly similar to the original event log.

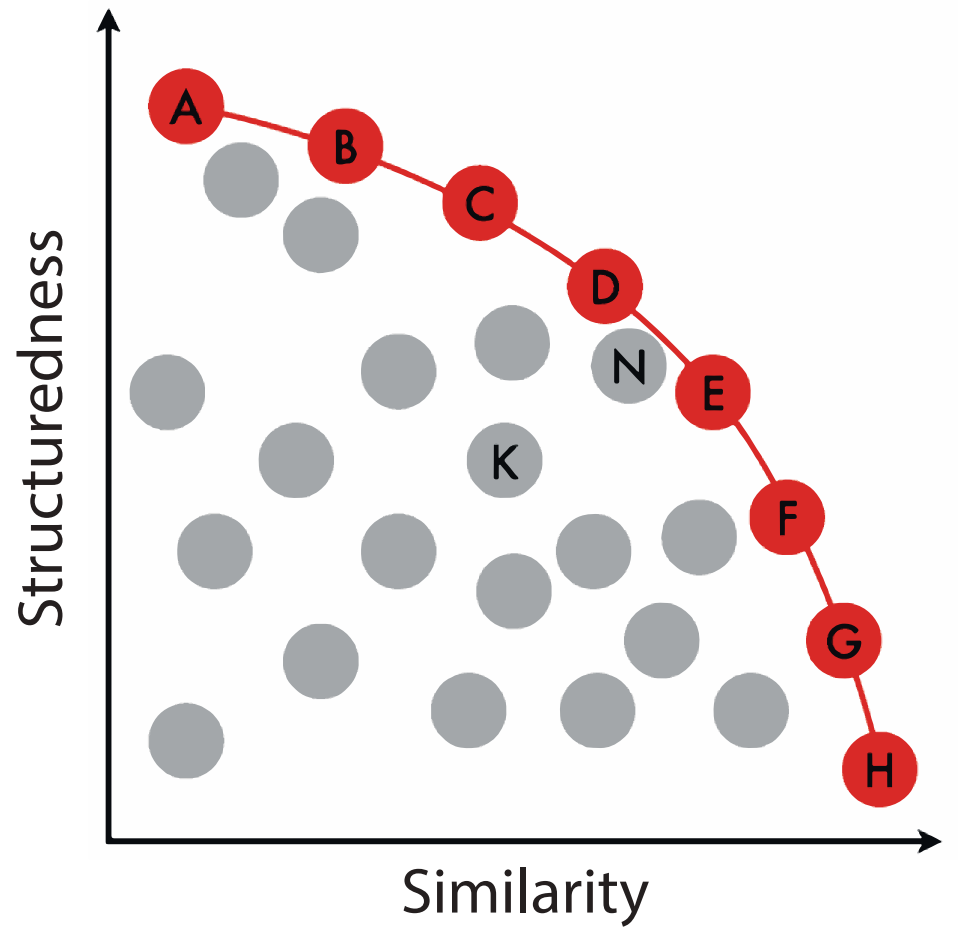

**Figure 14.1:** Pareto front of "optimal event logs" resulting from the combined application of pre-processing techniques.

Using these two criteria, event log pre-processing can be formulated as a multi-criteria optimization problem where the search space consists of some predefined set of possible event-log pre-processing steps (e.g., filtering activities Chapter 6, refining labels based on timestamps (Chapter 5), or filtering events [CRH17; FZA18; Lu+16b]) over which some multi-criteria optimization technique searches for the transformation of the original event log that is at the same time as structured as possible while being as similar as possible to the original event log. Popular techniques for multi-criteria optimization include evolutionary multi-objective optimization algorithms like Non-dominated Sorting Genetic Algorithm-II (NSGA-II) [Deb+02] (which in the process mining field is mostly known for being used in the ETMd [BDA12a] process discovery method) and Strength Pareto Evolutionary Algorithm 2 (SPEA-2) [Kim+04].

Multi-objective optimization algorithms return the set of Pareto optimal solutions, i.e., the set of solutions such that there has not been found any other solution that has both better structuredness and similarity. Figure 14.1 visualizes the Pareto front of event logs that such a multi-objective optimization method might return. The original event log (H) scores optimal similarity (since similarity is measured with respect to the original log), but behaviorally it might be highly variable, thereby making it difficult for mining approaches to find structure in the data. (A) to (G) represent Pareto optimal event logs that provide different trade-offs in their closeness to the original log (H) and their structuredness.

### 14.3.2 Structuredness Measures for Event Logs

Related to the previous area of future work, we would like to mention structuredness measures for event logs separately as an important area of future work. Currently, it is unknown how the degree of structure (or conversely, *variability*) should be quantified. We have seen that traditional process discovery techniques have trouble to generate insightful process models from event logs, and in such circumstances tend to generate either *spaghetti* models that consist of so many crossing edges that the model becomes non-iterpretable for human analysts, or *flower-like* that overgeneralize the behavior seen in the event log to such a degree that the model that allows for close to all possible behavior. However, structuredness is not a binary property, but it is a continuous spectrum, i.e., a given event log is not either structured or unstructured, but it is structured/unstructured to a certain degree. The central question is whether there exist measures of event log structuredness that can be used to accurately predict what type of analysis technique is most likely going to lead to useful insights into the behavior in the event log, e.g., whether process discovery should be applied (and if so, which algorithm?) or whether a pattern mining technique seems more promising.

### 14.3.3 Formal Foundations for Process Mining

In Chapter 3 we have started a formal analysis of fitness and precision measures in process mining where we started our analysis by formulating formal requirements for such measures in the form of axioms and used these axioms as a basis to validate whether existing measures measure what they are intending to measure. There seems to be room for a wider range of formal work in the process mining domain that follows the same paradigm of formulating requirements, formalizing these requirements into axioms, and proving or disproving whether existing techniques fulfill the formulated requirements. This approach could be applied to other types of quality criteria in process mining that were not yet included in Chapter 3 like simplicity and generalization. Additionally, this approach could be even applied more generally to process mining tasks, by for example formulating axioms regarding process discovery itself and verifying whether existing process discovery techniques do or do not satisfy certain desired properties regarding process discovery.

### 14.3.4 Process Discovery Methods that can Ignore Activities

The local process model techniques that we have proposed in this thesis generate process models that only model the behavior in the event log with respect to a subset of the activities in the event log. However, in the techniques that we have proposed in this thesis, these subsets of activities are quite small (often only four or five activities). We believe it to be a promising direction of research to develop process discovery techniques that automatically learn to ignore activities for which their occurrence seems to be independent of the other activities in the process. This would lead to process models that can be seen as local (in the sense that they do not explain all activities), but with a degree of locality that is less than the local process model mining techniques that are described in this thesis. Possibly, the chaotic activity detection/filtering approach that we have described in Chapter 6 can play a role in such process discovery techniques that are able to ignore activities.

### 14.3.5 Probabilistic Process Discovery

The process models that are generated by today's process discovery techniques only specify the language that is allowed by the model without specifying the likelihood of each trace to be generated by this process model. An interesting next step is the discovery of probabilistic process models that specify a probability distribution over sequences of business process activities. We conjecture that such probabilistic process models would be a suitable tool to express the behavior in highly variable event logs, as in such scenario's often many orderings of activity are possible in principle, but they might highly vary in their likelihood.

Adding probability to process models also strengthens the link between process mining with the fields of machine learning (more specifically: sequence models) and grammar inference. Both the sequence modeling and the grammar inference field have the aim to learn a probability distribution over sequences from a sample of sequences. This would also enable new methods to evaluate the quality of a process model for a given event log that takes into account the likelihood of a given trace and its frequency in the log. In [TZT18] and [TTZ18] we have made initial steps in this direction. In these works we have proposed a technique to add probability information into a process model in a *post-hoc* manner, i.e., a process model without probability information is first discovered using any existing process discovery technique and the probability information is inserted in a separate step. Furthermore, these works propose to evaluate the quality of a probabilistic process model with Brier score, which is a well-known evaluation measure from the field of probabilistic classification in Machine Learning.

## 14.4 A Broader Outlook

We conclude this thesis with a broader outlook on process mining, reflecting on our contributions in a broader context.

In recent years, we observe a trend where the process mining field is maturing at a rapid pace, both academically as well as commercially. Academically, we see that the process mining research community is growing, with more research groups joining the research field every year (last year, most noticeably the University of Melbourne and RWTH Aachen University). Further evidence can be seen in the organization of the first international conference on process mining[36] this year. Commercially, there has been a rapid growth in the number of companies that develop process mining tools, including Celonis[37], Lana Labs[38], Fluxicon[39], Minit[40], QRP[41] and many others.

There is a clear gap between what the academic process mining community and the commercial process mining community focus on. The commercial process mining tool vendors seem to mostly focus on generating nice looking visualizations of the process from the event data, while the types of process models are often rather simple and often lack formal semantics (e.g., many commercial tools use approaches that are similar in nature to the Fuzzy Miner [GA07]). In contrast, the academic process mining research community is much more focused on generating models with formal semantics and on supporting the identification of a richer set of workflow patterns.

When dealing with event logs that have high variability in the behavior, it is generally much easier to generate models that are "looking nice" and are not a large pile spaghetti (a model that consists of so many crossing edges that it becomes non-interpretable for human analysts) when resorting to types of models without formal semantics (like most commercial tools) in contrast to model types with formal semantics (like most techniques developed in academia). However, without formal semantics, the question "what does this model mean?" always remains and it is often unclear what behavior did occur and what behavior did not occur in the event log under study. For this reason, we believe that mining models with formal semantics is valuable.

Many of the techniques that are developed in this thesis contribute to the goal of mining models with formal semantics from event logs that are highly variable in their behavior and where, as a result, traditional process discovery techniques that generate models with formal semantics struggle to generate human-interpretable results. Mining models with formal semantics that are also human-interpretable from event logs with highly variable behavior will remain a challenge and an important area of research in the years to come. This thesis contributes a novel set of techniques that are proven useful for achieving this goal in certain conditions. In practice, it remains a tedious trial-and-error approach in which the process analyst / data scientist has to simply try out several pre-processing approaches and mining techniques and then manually assess which ones end up providing results that lead to actionable insights.

More specifically for the application domain of smart home environments and ambient intelligence, we believe that the main challenge in the coming years will be related to the question of how to actually use the patterns and the mining results in practice. Many mining techniques that have been developed for ambient intelligence, including many of the ones in this thesis, provide the user with patterns and with results, but do not explicitly tell the user what to do to achieve his or her goal. One possible use case that might motivate someone to mine his/her behavior could be aimed at behavioral change, where the person is aiming to change his/her habits into more healthy habits. Closing the gap between such a goal on the one hand and the mining results, on the other hand, is a vital direction of future research.

[36] https://icpmconference.org/
[37] https://www.celonis.com/
[38] https://lana-labs.com/
[39] https://fluxicon.com/disco/
[40] https://www.minit.io/
[41] https://www.qpr.com/solutions/process-mining